%% file: main.tex
\documentclass[11pt,dottedtoc,abstract=on,letterpaper,bibliography=totoc]{scrreprt}

\input{preamble/packages}
\input{preamble/commands}

\input{preamble/styles}

\begin{document}

\setcounter{page}{1}
\input{preface/preface.tex}

\cleardoublepage

\setcounter{page}{1}
\renewcommand{\thepage}{\arabic{page}}
\part{Preliminaries}
\renewcommand{\theHsection}{\thepart.\thechapter.\thesection}
\label{part:preliminaries}


\chapter{Introduction}
\pagestyle{dave-thesis}
\thispagestyle{dave-thesis-prefix}
\setlength\epigraphrule{0pt}
\label{chap:introduction}
\input{chapters/c1_introduction.tex}

\chapter{Background}
\thispagestyle{dave-thesis-prefix} 
\label{chap:background}
\input{chapters/c2_background.tex}

\cleardoublepage
\part{State Abstraction}

\label{part:state_abstraction}

\newpage
\chapter{Approximate State Abstraction}
\thispagestyle{dave-thesis-prefix}
\label{chap:approx_state_abstr}

\input{chapters/c3_sa_approx_state_abstraction.tex}

\newpage
\chapter{State Abstraction in Lifelong RL}
\thispagestyle{dave-thesis-prefix}
\label{chap:state_abstr_lifelong}
\input{chapters/c4_sa_lifelong_state_abstr.tex}

\newpage
\chapter{State Abstraction as Compression}
\thispagestyle{dave-thesis-prefix}
\label{chap:rlit}
\input{chapters/c5_sa_state_abstr_as_compression.tex}

\part{Action Abstraction}
\label{part:action_abstraction}

\newpage
\chapter{Finding Options that Minimize Planning Time}
\thispagestyle{dave-thesis-prefix}
\label{chap:options_for_planning}
\input{chapters/c6_aa_options_for_planning.tex}

\newpage
\chapter{The Expected-Length Model of Options}
\thispagestyle{dave-thesis-prefix}
\label{chap:elm_options}
\input{chapters/c7_aa_elm.tex}

\newpage
\chapter{Discovering Options for Exploration}
\thispagestyle{dave-thesis-prefix}
\label{chap:options_for_exploration}
\input{chapters/c8_aa_options_for_exploration.tex}

\part{State-Action Abstraction}
\label{part:state_action_abstraction}


\newpage
\chapter{Value Preserving State-Action Abstractions}
\thispagestyle{dave-thesis-prefix}
\label{chap:vpsa}
\input{chapters/c9_hierarchical_abstraction.tex}

\newpage
\chapter{Conclusion}
\thispagestyle{dave-thesis-prefix}
\label{chap:conclusion}

\input{chapters/c10_conclusion.tex}


\cleardoublepage
\bibliographystyle{plainnat}
\pagestyle{dave-thesis-prefix}
\bibliography{abstraction}

\end{document}

%% file: preamble/packages.tex
\usepackage[parts,pdfspacing,linedheaders]{classicthesis} 
\PassOptionsToPackage{usenames, dvipsnames}{xcolor}
\usepackage{enumerate, amsmath, amsthm, amssymb, mathrsfs, algorithm, pifont, subfig, csquotes, dashrule, tikz, bm, bbm, booktabs, verbatim, url, bbding, MnSymbol, tocloft} 
\usepackage{geometry}
\usepackage[framemethod=TikZ]{mdframed}
\usepackage[numbers]{natbib}
\usepackage{algpseudocode}
\usepackage{changepage}
\usepackage{chngcntr}
\usepackage{titlesec}
\usepackage{epigraph} 
\usepackage{pgfornament} 
\usepackage{environ}

\usepackage{setspace}

\usepackage{hyperref}

\usepackage{afterpage}

%% file: preamble/commands.tex
\newcommand{\rhoz}{\rho_{0}}
\newcommand{\eps}{\varepsilon}
\newcommand{\mc}[1]{\mathcal{#1}}
\newcommand{\indic}{\mathbbm{1}}
\newcommand{\bE}{\mathbb{E}}

\newcommand{\ra}{\rightarrow}
\newcommand{\la}{\leftarrow}

\newcommand{\KL}{D_{\text{KL}}}
\newcommand{\bs}[1]{\boldsymbol{#1}}

\newcommand{\phiop}{{\phi,\mc{O}_\phi}}

\newcommand{\PR}{\mathbb{P}}

\makeatletter
\@addtoreset{section}{part}
\makeatother
\newlength\mylen
\renewcommand\thepart{\arabic{part}}
\renewcommand\cftpartpresnum{Part~}
\newcommand{\bigmid}{\mathrel{\Big|}}

\definecolor{dblue}{RGB}{98, 140, 190}
\definecolor{dmblue}{RGB}{169, 193, 219}
\definecolor{dlblue}{RGB}{216, 235, 255}
\definecolor{dred}{RGB}{195, 112, 113}
\definecolor{partred}{RGB}{172, 60, 56}
\definecolor{dorange}{RGB}{230, 169, 132}
\definecolor{ddgreen}{RGB}{78, 132, 85}
\definecolor{dgreen}{RGB}{118, 167, 125}
\definecolor{dlgreen}{RGB}{154, 195, 157}
\definecolor{dtan}{RGB}{221, 215, 200}
\definecolor{dpink}{RGB}{207, 166, 208}
\definecolor{dyellow}{RGB}{255, 248, 199}
\definecolor{dgray}{RGB}{46, 49, 49}

\newcommand{\durl}[1]{\textcolor{dblue}{\underline{\url{#1}}}}

\newcommand{\spacerule}{\begin{center}\hdashrule{2cm}{1pt}{1pt}\end{center}}

\DeclareMathOperator*{\argmin}{arg\,min}
\DeclareMathOperator*{\argmax}{arg\,max}

\newcommand{\dbox}[1]{
\begin{mdframed}[roundcorner=4pt, backgroundcolor=gray!5]
\vspace{1mm}
{#1}
\end{mdframed}
}

\newmdenv[
  topline=false,
  bottomline=false,
  rightline = false,
  leftmargin=10pt,
  rightmargin=0pt,
  skipabove=16pt, 
  innertopmargin=0pt,
  innerbottommargin=0pt
]{innerproof}

\newenvironment{dproof}[1][Proof]{
\hypersetup{hidelinks}
\begin{proof}[\textbf{\textit{Proof of #1.}}]
\text{\vspace{2mm}} \begin{innerproof}}
{\end{innerproof}
\end{proof}
\vspace{8mm}}

\newcommand{\ddef}[2]
{
\vspace{2mm}
\begin{mdframed}[roundcorner=1pt, backgroundcolor=white]
\begin{definition}
\ #2
\end{definition}
\vspace{1pt}
\end{mdframed}
\vspace{-2mm}}

\newcommand\numberthis{\addtocounter{equation}{1}\tag{\theequation}}

\newtheorem{assumption}{Assumption}[chapter]

\newtheorem{corollary}{Corollary}[chapter]
\newtheorem{claim}{Claim}[chapter]
\newtheorem{example}{Example}[chapter]
\newtheorem{lemma}{Lemma}[chapter]

\newtheorem{remark}{Remark}[chapter]
\newtheorem{theorem}{Theorem}[chapter]
\newtheorem{definition}{Definition}[chapter]

\newcommand{\elone}[1]{\left\lVert#1\right\rVert_1}
\newcommand{\rademacher}[1]{\text{Rad}(#1)}
\newcommand{\phiall}{\Phi_{\text{all}}}
\newcommand{\Oall}{\mc{O}_{\text{all}}}
\newcommand{\mimoalg}{\textsc{A-MIMO}}
\newcommand{\momialg}{\textsc{A-MOMI}}

\newcommand{\oline}[1]{
\resizebox{0.9\textwidth}{6pt}{%
  \begin{tikzpicture}
        \pgfornament[symmetry=h]{#1}
    \end{tikzpicture}
}}

\newcommand*\circled[1]{\tikz[baseline=(char.base)]{\node[shape=circle,draw,inner sep=0.7pt] (char) {\footnotesize{#1}};}}

\newcommand{\subfhspace}{\hspace{6mm}}
\newcommand{\figvdim}{50mm}

%% file: preamble/styles.tex
\expandafter\def\expandafter\quote\expandafter{\quote\onehalfspacing}

\hypersetup{
    colorlinks=true,
    breaklinks=true,
    urlcolor=dblue,
    linkcolor=ddgreen,
    citecolor=dblue
}

\titleformat*{\part}{\centering\huge}
\titleformat*{\chapter}{\LARGE}
\titleformat*{\section}{\Large}

\titlespacing*{\section}{0mm}{4mm}{4mm}
\titlespacing*{\subsection}{0mm}{4mm}{4mm}
\titlespacing*{\paragraph}{0mm}{6mm}{6mm}

\numberwithin{footnote}{chapter}
\numberwithin{equation}{chapter}
\numberwithin{algorithm}{chapter}

\automark{chapter}

\clearpairofpagestyles 

\newpairofpagestyles{dave-thesis-prefix}{
\cfoot{-~\pagemark~-}
}
\newpairofpagestyles{dave-thesis}{
\cfoot{-~\pagemark~-}
\ohead{\textsc{\headmark}}
\ihead{\textsc{\partname\ \thepart}}
}

\AddToLayerPageStyleOptions{dave-thesis}
  {oninit=\KOMAoptions{headsepline}\setkomafont{headsepline}{\color{black}}}

%% file: preface/preface.tex
\pagenumbering{gobble}
\setcounter{page}{1}

\addtocontents{toc}{\protect\thispagestyle{dave-thesis-prefix}}
\addtocontents{lof}{\protect\thispagestyle{dave-thesis-prefix}}
\addtocontents{lot}{\protect\thispagestyle{dave-thesis-prefix}}

\input{preface/title_page.tex}
\newpage
\input{preface/copyright.tex}
\newpage
\renewcommand{\thepage}{\roman{page}}
\thispagestyle{dave-thesis-prefix}
\input{preface/signatures.tex}
\newpage
\input{preface/revisions.tex}
\newpage
\input{preface/abstract}
\newpage

\thispagestyle{dave-thesis-prefix}
\input{preface/cv.tex}
\thispagestyle{dave-thesis-prefix}
\input{preface/dedication}
\cleardoublepage

\input{preface/acknowledgements.tex}
\cleardoublepage
{
\thispagestyle{dave-thesis-prefix}
\setstretch{1.5}
\hypersetup{hidelinks}
\setcounter{tocdepth}{1} 
\thispagestyle{dave-thesis-prefix}
\tableofcontents
\thispagestyle{dave-thesis-prefix}
}

\input{preface/lists}

%% file: preface/title_page.tex
\begin{titlepage}
	
	\center 
	\vspace{20mm}
	\hfill\hfill
	\vspace{40mm}
	
	\textbf{A Theory of Abstraction in Reinforcement Learning} \\

	\vspace{25mm}

    David Abel \\
    
    \vspace{25mm}
    
    \begin{minipage}[c]{0.65\textwidth}
    \begin{center}
    A dissertation submitted in partial fulfillment of the requirements for the Degree of Doctor of Philosophy in the Department of Computer Science at Brown University
    \end{center}
    
    \end{minipage}

	\vspace{25mm}
	
	Providence, Rhode Island
	
	May 2020 

	\vfill\vfill
	
\end{titlepage}

%% file: preface/copyright.tex
\vspace{-50mm}
\vspace*{60mm}

\begin{center}
    {\Large \copyright\ David Abel} \\
    \vspace{16mm}
    This work is licensed under the \\
    \href{https://creativecommons.org/licenses/by/4.0/}{Creative Commons Attribution 4.0 International License}
\end{center}

\vspace{8mm}
\begin{center}
\includegraphics{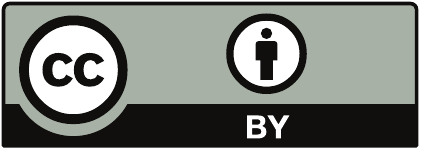}
\end{center}


%% file: preface/signatures.tex
\newcommand{\sig}[2]{
\begin{tabular}{ll}
      Date \rule{1in}{.05mm}\hspace*{.75in} & \rule{3.5in}{.05mm}\\
      \vspace{-5mm}&\vspace{-5mm} \\
      & \small{#1, #2}
\end{tabular}
}

\begin{center}
This dissertation by David Abel is accepted in its present form by \\
the Department of Computer Science as satisfying the
dissertation requirement \\
for the degree of Doctor of Philosophy. \\
\vspace{14mm}

\sig{Michael L. Littman}{Advisor}

\vspace{14mm}
Recommended to the Graduate Council
\vspace{14mm}

\sig{George Konidaris}{Reader}
\null\vskip.22in
\sig{Stefanie Tellex}{Reader}
\null\vskip.22in
\sig{Peter Stone}{Reader}
\null\vskip.22in
\sig{Will Dabney}{Reader}

\vspace{14mm}
Approved by the Graduate Council
\vspace{14mm}

\sig{Andrew G. Campbell}{Dean of the Graduate School}

\end{center}

%% file: preface/revisions.tex
\chapter*{Revisions}
\thispagestyle{dave-thesis-prefix}
    This version contains minor revisions from the original:
    \begin{itemize}
        \item April 27, 2020: Official version submitted to the university.
        \item October 13, 2020: Added revision page, moved abstract from first page, fixed typos in acknowledgements, fixed Silver et al. 2020, Bai \& Russell 2017, and Majeed \& Hutter 2019 references, fixed typo in Equation 2.2, fixed notation in Equation 9.83, fixed page number for Part 1.
        
        \item November, 2021: Minor formatting changes for printing.
        
	\item March, 2022: Reverted formatting changes from the printed version, fixed a floating comma in the references, adjusted the style of the header and footer, adjusted copyright, formatting changes to Chapter 3, caught autoref mistakes in Chapters 3, 4, 5, 7, and 9.
    \end{itemize}
    

%% file: preface/abstract.tex
\renewcommand{\abstractname}{\fontsize{10pt}{12pt}{Abstract of ``A Theory of Abstraction in Reinforcement Learning" \\
by David Abel, Ph.D, Brown University, May 2020.}}

\begin{abstract}
\thispagestyle{dave-thesis-prefix}
Reinforcement learning defines the problem facing agents that learn to make good decisions through action and observation alone.
%
To be effective problem solvers, such agents must efficiently explore vast worlds, assign credit from delayed feedback, and generalize to new experiences, all while making use of limited data, computational resources, and perceptual bandwidth.
%
Abstraction is essential to all of these endeavors. Through abstraction, agents can form concise models of their environment that support the many practices required of a rational, adaptive decision maker.
%
In this dissertation, I present a theory of abstraction in reinforcement learning.
%
I first offer three desiderata for functions that carry out the process of abstraction: they should 1) preserve representation of near-optimal behavior, 2) be learned and constructed efficiently, and 3) lower planning or learning time.
%
I then present a suite of new algorithms and analysis that clarify how agents can learn to abstract according to these desiderata.
%
Collectively, these results provide a partial path toward the discovery and use of abstraction that minimizes the complexity of effective reinforcement learning.
\end{abstract}

%% file: preface/cv.tex
\chapter*{Vita}

David Abel was born in Portland, Oregon. He received his Bachelors in computer science and philosophy from Carleton College, during which he spent a semester in Budapest, Hungary. He then attended Brown University, where he received a Masters in each of computer science and philosophy en route to his Ph.D. At Brown, he taught CS8: A First Byte of Computer Science and a Summer course he designed for high school students on AI and Society. David also spent time as a research intern at Microsoft Research in New York City, the University of Oxford, and DeepMind London. He received the Open Graduate Fellowship to pursue a Masters in philosophy and the Presidential Award for Excellence in Teaching.

\newenvironment{dpub}
{\begin{adjustwidth}{5mm}{0mm}}
{\dvspace\end{adjustwidth}}

During his Ph.D, David published the following papers: \emph{Value Preserving State-Action Abstractions} \cite{abel2020vpsa}, \emph{The Efficiency of Human Cognition Reflects Planned Use of Information Processing} \cite{ho2020_plan_to_plan}, \emph{The Value of Abstraction} \cite{ho_val_abstr2019}, \emph{The Expected-Length Model of Options} \cite{abel_winder2019elm}, \emph{Finding Options that Minimize Planning Time} \cite{jinnai2019opt}, \emph{Discovering Options for Exploration by Minimizing Cover Time} \cite{jinnai2019opt_explore}, \emph{State Abstraction as Compression in Apprenticeship Learning} \cite{abel2019rlit}, \emph{State Abstractions for Lifelong Reinforcement Learning} \cite{abel2018salrl}, \emph{Policy and Value Transfer in Lifelong Reinforcement Learning} \cite{abel2018transfer}, \emph{Bandit-Based Solar Panel Control} \cite{abel2018solar}, and \emph{Near Optimal Behavior via Approximate State Abstraction} \cite{abel2016near}.

%% file: preface/dedication.tex
{\newpage
\thispagestyle{dave-thesis-prefix}
\topskip0pt
\vspace*{\fill}
\begin{center}
\textit{To my fianc{\'e}e Elizabeth, my brother Mike, and my parents Mark \& Diane.}
\end{center}
\vspace*{\fill}
}

%% file: preface/acknowledgements.tex
\chapter*{Acknowledgements}
\thispagestyle{dave-thesis-prefix}
\pagestyle{dave-thesis-prefix}

One of the greatest pleasures of research is working with brilliant and thoughtful individuals from all over the world. I am grateful to my mentors for their encouragement and guidance over the years, my collaborators for sharing their time and ideas, and the support and friendship of many.

\section*{Mentors}

First, to my advisor, Michael Littman. From day one Michael was able to blend encouragement with structured advice. He always helped me feel like I was making progress and knew how to point me in the right direction. I regularly left our meetings with a clear objective and a wealth of excitement to learn about a new topic or work toward a new result. On top of this, Michael is both immeasurably wise, funny, and humble. I can recall many occasions where I pitched a new problem to him that I was stuck on---he quickly figured out a solution, and would carefully lead me to the conclusion in a way that nurtured my own sense of curiosity and discovery, and made it fun along the way. Some of my absolute favorite experiences in life are these moments Michael and I stared down a tricky problem on the whiteboard in his office. These conversations eventually grew into the publications and ideas that constitute this dissertation. Ultimately, Michael taught me what good science looks like. He shaped how I ask questions, how to do my best to be unbiased, and how to proceed in the presence of profound challenges and uncertainty. Richard Hamming's ``You and Your Research" suggests that great research is done with great courage: Through his guidance, Michael has inspired me to try to be courageous in research, to ask big questions, and to methodically seek out principled answers. He has also taught me to embrace my own instincts and curiosities, and to how to balance my own philosophy with existing literature. Michael: Thank you for your time, support, and trust, for giving me the freedom to explore, allowing me to make safe mistakes, and being such an outstanding mentor, advisor, and person. I am so profoundly lucky to have had the opportunity to work with you! I will never forget our time doing research together, and I look forward to continuing our collaborations and discussions in the future.

To George Konidaris, thank you for the many fascinating discussions on abstraction and AI, and for the countless pieces of life advice you have offered over the years. I was inspired by your research early on in my Ph.D, so it was a genuine delight to discover that you were coming to Brown, and even more so that I had the good fortune to collaborate with you for many years. I have such a deep admiration for your research, vision, and methods, and I am honored to have had the chance to work with you. Perhaps the moment in my Ph.D when I laughed the hardest was up against the ICML 2018 submission deadline for our work on transfer learning in RL. I used the term ``epistemic guise" in a draft of the paper, and received a note from you ``Dave what is an epistemic guise." I have been sure not used this phrase since (until now)! Thank you George for your time and mentorship!

Next, to my Masters advisor, Stefanie Tellex: Thank you for taking a chance on me as a researcher, and for inspiring me to pursue AI research. You first taught me how to read papers, how to give talks, how to formulate an appropriate research question, and many other crucial aspects of how to be an effective scientist. I can still remember pitching my first research problem; you were eager to engage with the ideas and gave me the tools I needed to systematize my study. I will always finish my talks with a contributions slide and include slide numbers, thanks to your many essential tips. Moreover, I will always remember the moment in the midst of a paper deadline when you paused, told my co-authors and me to bubble up a level and make sure we were doing okay as people. Thank you, Stefanie!

To Will Dabney, thank you for mentoring me at DeepMind, serving on my committee, and for a both fruitful and enjoyable collaboration---I learned a great deal from working with you and I look forward to our future research together!

To Peter Stone, thank you for your time in serving on my committee, your advice, and for your thoughtful questions that still have me thinking!

To Colin Day, or ``D-Day" (In Memoriam), I owe the initial spark of my love of math, elegant theory, and the belief that hard work can translate to deep understanding, even when facing challenges that seem insurmountable. You were an amazing teacher and mentor. Thanks, D-Day!

To Jake Stults (In Memoriam), my first computer science teacher. Thank you for cultivating my initial curiosity about computation, AI, and logic, for sharing music with me, and for your constant encouragement early on in life. Thanks, Jake!

To my other mentors, for their time, patience, advice, and helping me find my path: Jason Decker, Fernando Diaz, Owain Evans, David Liben-Nowell, James MacGlashan, Anna Moltchanova, Joshua Schechter, and Lawson L.S. Wong, thank you all!

\section*{Family, Friends, Collaborators, and Colleagues}

To my fianc{\'e}e, Elizabeth Thiry, I am forever grateful for your love and support throughout these five years. Thank you for listening to countless practice talks, inspiring me to take on new challenges, filling our days with epic experiences all the way from Sweden to Hungary to Boston, and always believing in me. The Ph.D has been such a blast and I look forward to the rest of our life together!

To my parents, Diane and Mark Abel, thank you for your love and support, all the way from 0th grade to 23rd! I am grateful for all you have done---I am so fortunate to have such wonderful parents. From early on in life you were both unwavering in your encouragement, helping me to ask questions and to have fun with learning and life more generally. Thank you for the many insightful conversations, trips, and experiences that have shaped nearly every aspect of my life, and for helping me find my way!

To my brother, Michael Abel, thank you for being such an amazing role model. I still remember the many deep discussions we had about science and philosophy as we grew up together. I vividly recall taking a drive together in the west hills of Portland one night and discussing different kinds of infinity for hours. You were always supportive of these discussions and helped me feel comfortable with wanting to think about philosophy and math (you made it fun, too!). I would not have had the confidence to pursue this degree without you as a role model. Most crucially, Mike is \textit{amazing} at balancing life. Despite the Ph.D being difficult, I always set aside time for friends, family, and personal time because of the example he set. Thank you, Mike!

To Nate Bowditch, thank you for all of the good times, for being an outstanding roommate, person, and friend, and for the many deep discussions and epic adventures we have had together!

To Pablo Leon-Luna, thank you for the many thoughtful conversations and good times, for inspiring me to pursue what I truly love, and for all your support and friendship over the years!

To Ellis Hershkowitz, thank you for helping me to discover the kind of research that I value, for the many productive collaborations, fascinating conversations, and good times!

To those that were especially close collaborators or friends throughout my Ph.D, I am so thankful for your time, support, and friendship: Cam Allen, Enrique Areyan, Dilip Arumugam, Kavosh Asadi, Akshay Balsubramani, Christina Donovan, Chris Grimm, Mark Ho, Yuu Jinnai, Ben LeVeque, Torben Noto, Nate Rahn, and Greg Yauney, thank you!

Lastly, I am grateful to all of the friends, extended family, colleagues, and collaborators that have helped me on this journey! Thanks to Belle \& Sid Abel (In Memoriam), Niko Adamstein, Lori Agresti, Suzanne Alden, S{\'e}b Arnold, Gabriel Barth-Maron, Kathy Billings, Lisa and Jason Bogardus, Ben Breen, Nick Brenner, Evan Cater, Lauren Clarke, Erica Clausen, Milan Cvitkovic, Genie DeGouveia, Marie desJardins, Sophia Faltin, Katie Franklin, Tomasz Kalbarczyk, Khimya Khetarpal, Brian Kimpson, Akshay Krishnamurthy, Rita \& Harry Krych (In Memoriam), Erwan Lecarpentier, Meg \& Alexander Leiken, Jason Liu, Jane Martin, Debbie Osterman, Matt Overlan, Alex Park, Katy Parsons, Bree \& Ohm Patel, Robby Plowman, Jesse Polhemous, Emily Reif, Mel Roderick, Mark Rowland, Hannah Roy, Brandon Saranik, Evan Schwed, Gunnar Sigurdson, Satinder Singh, Andy Smith, Katherine \& Julius Thiry, Raphael Townshend, Ally Wharton, Will Whitney, John Winder, Danfei Xu, and to all of the members of the iTeam, RLAB, \& Brown CS, thank you all!

%% file: preface/lists.tex
{
\setstretch{1.5}
\hypersetup{linkcolor=black}
\numberwithin{table}{chapter}
\listoftables
\thispagestyle{dave-thesis-prefix}
}

{
\numberwithin{figure}{chapter}
\setstretch{1.5}
\hypersetup{linkcolor=black}
\listoffigures
\thispagestyle{dave-thesis-prefix}
}

%% file: chapters/c1_introduction.tex
Suppose you take a walk in the woods. You find yourself surrounded by pine trees, chirping birds, a peaceful lake, and frogs eating delicious mealworms. A friend returns from a walk and relays a story of a goat miraculously walking up a steep mountain side. Your stomach grumbles and you deliberate over whether to eat an apple from your backpack or to start a campfire and cook a hot meal.

Consider just how many activities are at your finger tips: you could climb a tree, discuss philosophy with your friend, navigate to a nearby stream by listening for rushing water, or create a map of the territory. To engage in any of these practices in this complex and changing environment you must be capable of making hundreds of well-chosen decisions that move you toward a particular objective. Moreover, you must make these decisions while relying on imperfect memory and noisy sensory channels; some light hits your retina indicating the sun has risen, changes in sound pressure are processed through your ears to notify you of an impending thunderstorm, and nerve endings in your feet tell you your boots are wet. Somehow, you map this continual stream of observations to a choice of actions that moves you toward any of the above goals. How is this even remotely possible?

Central to an agent's ability to solve problems is the capacity to reason \textit{abstractly}---walking into a tree will cause pain while moving around it may not. Hence, representing particular patterns of visual stimuli as a tree that persists through time is immensely useful. Additionally, conveying to other agents that trees possess the ``do-not-walk-into" property is likely to be critical to the overall fitness of the community at large. Indeed, many central practices of agency rely on abstraction: speculating about and learning from hypothetical scenarios, overcoming new challenges because of their similarity to past experiences, and forming high-level plans spanning months or years that inform immediate action; all of these depend on a concise, adaptive, and abstract representational toolkit. For these reasons, the capacity to learn and make use of appropriate abstract representations is likely to be an essential cognitive skill of any intelligent agent, whether biological or artificial.

\begin{figure}[t!]
    \centering
    \includegraphics[height=\figvdim]{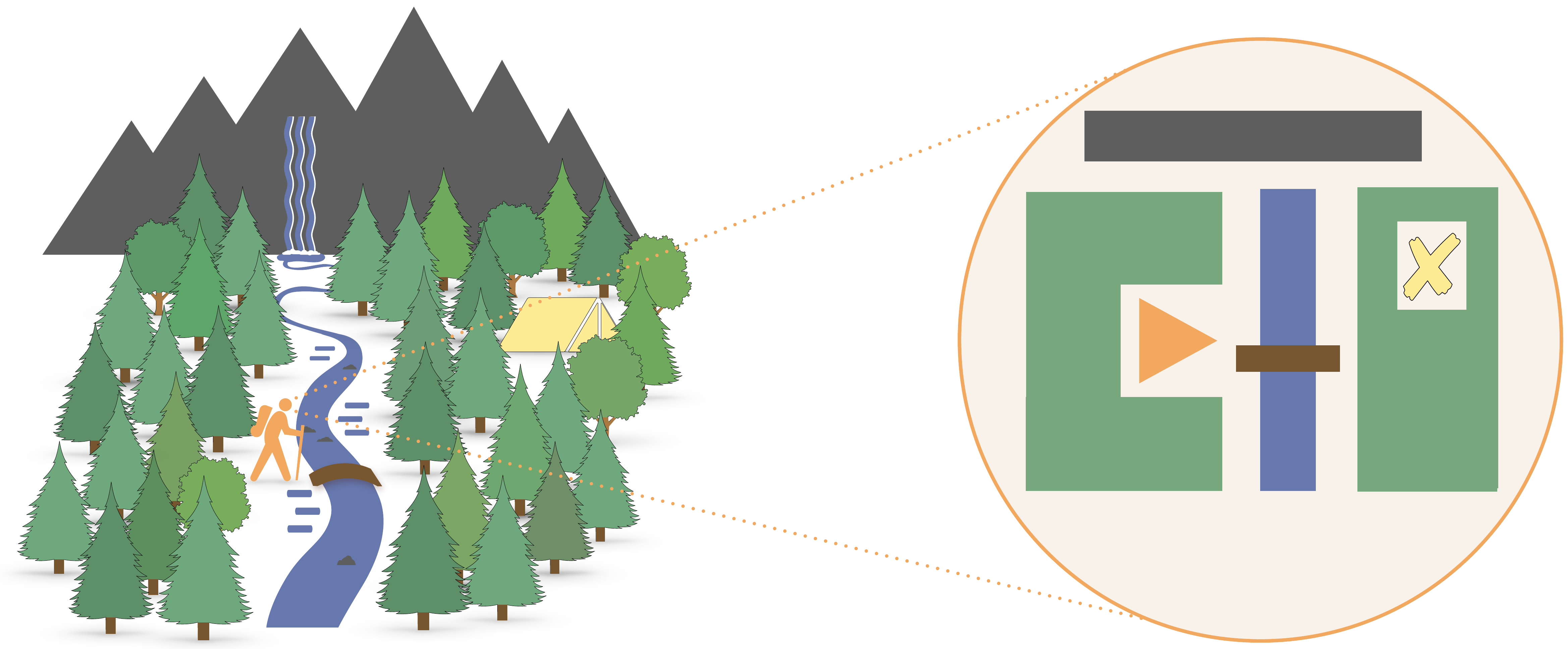}
    \caption{The process of abstraction.}
    \label{fig:abstract_forest}
\end{figure}

In the forest, we might imagine that a hiker trying to return to their tent may reason using the abstract representation pictured in \autoref{fig:abstract_forest}. With this smaller model that still retains relevant information, the hiker can carry out more valuable computation, explore less, draw more robust inferences, and predict further into the future. These benefits ultimately enable the hiker to safely navigate to their tent and to thrive in nature more generally. Where, though, does this model come from? And how can an agent discover such a model solely by interacting with their surroundings? These questions have long stood as a fundamental puzzle in the science of understanding intelligence.

This dissertation is about the study of abstraction and its role in effective agency. I ground this inquiry by concentrating on computational agents that must learn to solve problems from interaction alone, as captured by the reinforcement learning (RL) problem. Such an agent could consist of a finite state machine reacting to discrete symbols on a tape for the purpose of sorting a list, but also a robot or animal observing the world through sensors and a powerful action space that supports movement through and manipulation of the environment. In this remarkably general framework, we will find footing to make the study of abstraction concrete.

\section{The Reinforcement Learning Problem}
\label{sec:rl_intro}

RL defines the problem facing an agent that learns to make useful decisions through observation and action alone. The primary objects of interest in RL are computational agents, the worlds they inhabit, and the interactions thereof. An agent is understood as any entity capable of perception and action, where perception involves the receipt and processing of information from the environment, and action defines the process of committing to a choice from a set of alternative courses of behavior. I sharpen our use of the term ``world" in the next chapter, but broadly it is to be understood as a set of possible states of affairs, causal laws that move the world between these states, and an agent that makes decisions in the world based on a stream of observation.

Critical to RL is the assumption that one special observation of the world is a numerical \textit{reward signal} that corresponds to the immediate desirability of a given state of affairs. The objective of an RL agent is then simple: maximize future rewards. Richard S. Sutton and Michael L. Littman have articulated what is known as the \textit{reward hypothesis}, or \textit{reinforcement learning hypothesis}, that states the following:
\ddef{Reward Hypothesis}{The \textbf{reward hypothesis} states, ``all of what we mean by goals and purposes can be well thought of as maximization of the expected value of the cumulative sum of a received scalar signal (reward)" (\citeauthor{suttonwebRLhypothesis}, \citeyear{suttonwebRLhypothesis}).\vspace{2mm}}

Indeed, reward prediction and learning has long played a role in understanding human and animal cognition \cite{schultz1997neural,dayan2002reward,bayer2005midbrain}. For our present purpose, I assume the hypothesis to be valid, and proceed on the basis that the space of agents that effectively learn to maximize reward can be likened to the space of intelligent agents. I note, however, that a more thorough philosophical treatment of this hypothesis is of deep importance.

\begin{figure}[b!]
    \centering
    \includegraphics[height=\figvdim]{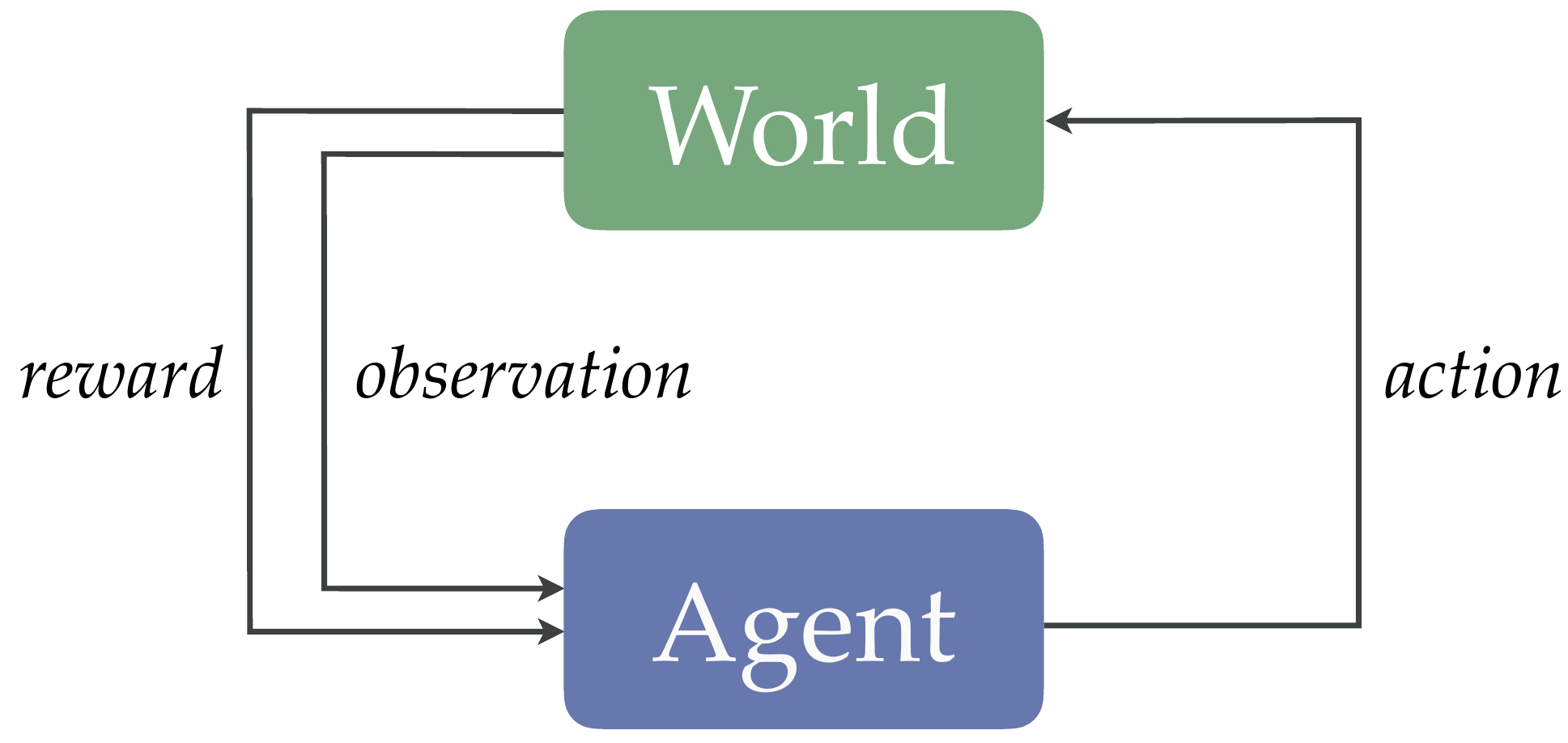}
    \caption{The RL problem.}
    \label{fig:rl}
\end{figure}

With these pieces in play, the RL problem is defined at a high level as follows.
\ddef{Reinforcement Learning Problem (High Level)}{The \textbf{reinforcement learning problem} is as follows. An RL agent interacts with its environment via the indefinite repetition of the following two discrete steps:
\begin{enumerate}
\item The agent receives an observation and a reward.
\item The agent learns from this interaction and executes an action.
\end{enumerate}
This process is pictured in \autoref{fig:rl}. The goal of the agent during this interaction is to make decisions so as to maximize its long term received reward.\vspace{2mm}}

Returning to our peaceful forest, we might imagine that the hiker occupies the world state pictured in \autoref{fig:abstract_forest}, and is learning about their surroundings to maximize reward. Depending on the reward generating process, the hiker will be incentivized to exhibit different kinds of behavior. For instance, the problem of navigating to a stream might be associated with a reward signal that increases as the hiker gets closer to the water. An effective agent, then, will learn to associate this positive signal with actions that move them toward the stream. Over time, the most effective agents can reach the stream without coming to harm along the way. To define the task of cooking food, we might attach positive reward signal to the experiences of eating tasty food---again, effective agents will be those that can prepare and eat food that is of sufficient levels of tastiness.

It is here that we find the remarkable expressivity of the RL problem: any goal-driven task can be defined in terms of a reward function that is positive when the goal is satisfied, and non-positive otherwise. Moreover, non-terminating behaviors such as controlling an elevator, balancing a pole, survival, or regulating energy on a power grid can \textit{also} be elicited with the right choice of reward function by similar mechanisms.

Richard S. Sutton describes both the appeal and challenge of RL as follows.
\begin{quote}
Part of the appeal of reinforcement learning is that it is in a sense the whole [artificial intelligence] problem in a microcosm. The task is that of an autonomous learning agent interacting with its world to achieve a goal. The framework permits the simplifications necessary in order to make progress, while at the same time including and highlighting cases that are clearly beyond our current capabilities, cases that we will not be able to solve effectively until many key problems of learning and representation have been solved. That is the challenge of reinforcement learning.
\\
    (\citeauthor{sutton1992introduction} \citeyear{sutton1992introduction}, p. 2)
\end{quote}

I am entirely sympathetic to Sutton's reasoning. Addressing the RL problem is of critical importance to establishing a holistic understanding of intelligence. Even roughly 30 years after the above quote, there are still many ``cases that are clearly beyond our current capabilities" at the heart of RL.
%
%
To be effective, RL agents must address a combination of three classical problems of machine learning:
\begin{enumerate}
    \item \textbf{Generalization:} Given experience from the past, how can an agent better act in the future?
    
    \textit{Example 1: The hiker spots an owl in the woods they have never before seen. How do they know it is an owl?}
    
    \textit{Example 2: You approach a door you have never before encountered and manage to open it within seconds.}
    
    \item \textbf{Exploration:} How can an agent systematically trade off between 1) exploiting what is known to be a reliably good choice with 2) making choices that may lead to new discoveries?
    
    \textit{Example: You visit your favorite restaurant, and deliberate whether to choose your go-to entr\'ee, or to try something new (that you might like even more!).}
    
    \item \textbf{Credit Assignment:} When feedback is delayed, how can an agent attribute credit to the most causally relevant decisions made previously?
    
    \textit{Example: You study for a test for weeks on end. Also, the night before the test, you eat a bowl of cereal. You ace the test. How can you determine that it was the studying that led to your success, and not the bowl of cereal?}
    
\end{enumerate}

Each of these problems individually is difficult, but in RL, agents must simultaneously address all three. I return to a more technical treatment of some of these problems throughout the dissertation after introducing the mathematical tools of RL in \autoref{chap:background}. Fortunately, however, abstraction can help address each of these challenges.

\section{Abstraction}
Indeed, understanding abstraction and its role in agency has long stood as one of the fundamental questions of artificial intelligence (AI), dating back to the famous workshop at Dartmouth that founded the field:

\begin{quote}
    The study is to proceed on the basis of the conjecture that every aspect of learning or any other feature of intelligence can in principle be so precisely described that a machine can be made to simulate it. An attempt will be made to find how to make machines...form abstractions and concepts. \\
    (\citeauthor{mccarthy2006proposal} \citeyear{mccarthy2006proposal}, p. 1)
\end{quote}

Since this workshop, the study of abstraction in AI and related fields has led to a profound appreciation for the role abstraction can play in both artificial and biological creatures. The focus of this dissertation is naturally on the former, though a large body of research in the cognitive, neuro, and psychological sciences examines the prevalence of abstraction in the representational and decision making practices of humans \cite{botvinick2009hierarchically,kemp2009structured,topin2015portable,solway2014optimal,botvinick2014model,werchan2015,quandt2017neural,eckstein2018evidence}.

The RL problem is perfectly suited to a scientific study of abstraction. Observation, on its own, is far too complicated for an agent to reason with while acting in a changing world. Thinking takes time. Processing, understanding, and reacting to every detail of a history of observations is computationally intractable. Additionally, in deliberating over possible futures, the space of sensible changes to the world that are worth considering is dramatically smaller than that of the possible future observation stream. Behavior, too, can often be defined at multiple levels of abstraction. For instance, an ant may act so as to follow its friend, or choose which precise muscles to twitch to propel its legs.

At a high level, the process of \textit{abstraction} can be divided into two broad categories: 1) state abstraction, which defines the practice of representing only the most relevant properties of the world, and 2) action abstraction, which defines the practice of forming a relevant set of long horizon behaviors available to an agent. In both cases, following \citet{giunchiglia1992theory}, I understand abstraction as ``the process of mapping a representation of a problem onto a new representation" (\citeyear{giunchiglia1992theory}, p. 1).

\begin{figure}[t!]
    \centering
    \subfloat[Reasoning in the environment.\label{fig:ground_mdp_forest}]{\includegraphics[height=\figvdim]{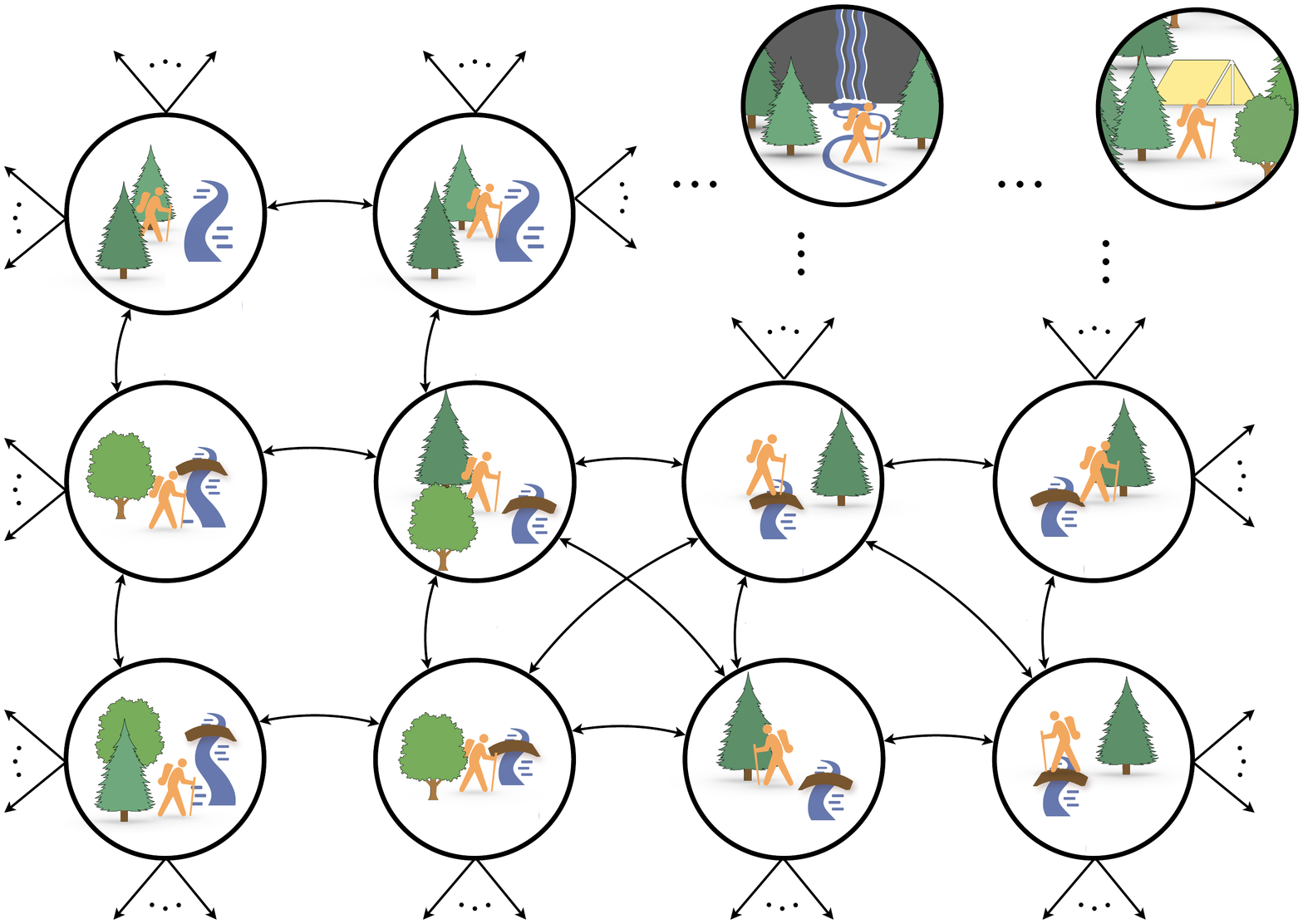}} \subfhspace
    \subfloat[Reasoning in the abstract.\label{fig:abstr_mdp_forest}]{\includegraphics[height=\figvdim]{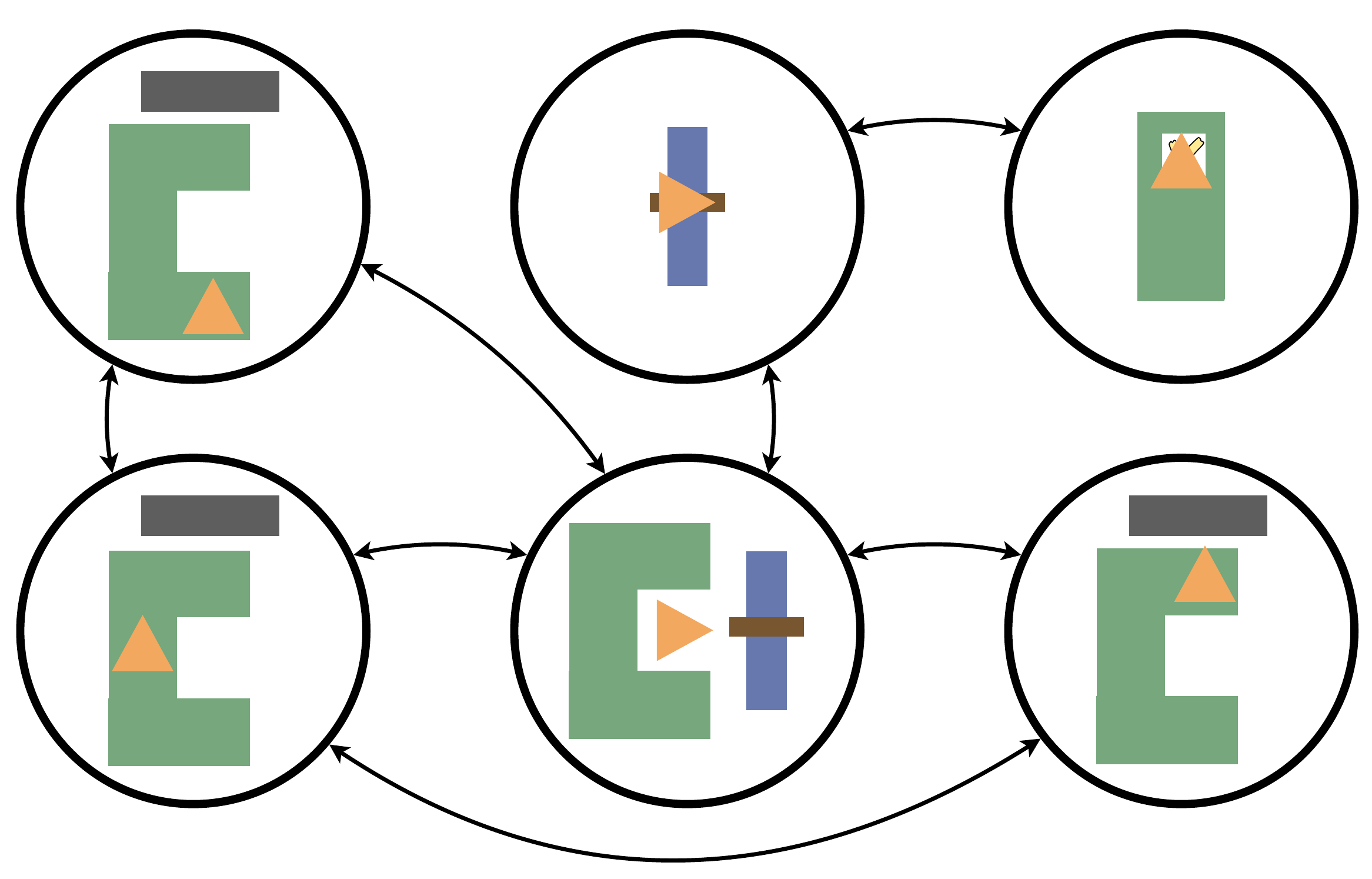}}
    \caption{A comparison of the problem posed to a hiker navigating to their tent in a forest in fine detail (left) and in the abstract (right).}
    \label{fig:abstr_forest_mdps}
\end{figure}

Let us return to the woods. Suppose our hiker is trying to navigate back to their camp, and can choose to represent this problem in great detail, or in the abstract, pictured in \autoref{fig:abstr_forest_mdps}. In each case, the hiker deliberates over possible future courses of action. In the first, however, the hiker's actions are modeled in terms of the smallest possible execution of behavior---a slight step in one direction, or a tilt of the head. In the second, decisions are considered that only change something \textit{substantive} about the environment---previously, the hiker was west of the bridge, and now, they are to the east of the bridge. Depending on the problem, different degrees and types of abstraction will be most effective.

A \textbf{state abstraction} determines which changes to the environment count as substantive. As the agent walks toward the bridge, for instance, the clouds shift overhead. The breeze picks up slightly, and a bird flies by. These small changes are likely to be irrelevant to the hiker's objective of crossing the bridge. Conversely, when the hiker has reached the river and can see the bridge, the state has changed in a relevant way. This strategy for reasoning in terms of abstract states of affairs alone is pictured in \autoref{fig:abstr_mdp_forest}. Note that only six states are required; the hiker might occupy the three distinct regions of the woods west of the river, the bridge, and their camp. These properties obscure many nuances such as weather, precise orientation, and even subtleties about the agent's physical or mental state. For some problems, these six states along would be entirely insufficient for permitting the representation of good behavior.

In \autoref{fig:ground_mdp_forest}, we instead see the diversity of state representations available if any slight change to the environment is perceived as a substantive change to world state. The agent could be next to the river, north of the bridge by five paces as opposed to four---or, the agent could be immediately under a tree as opposed to standing near it. For some problems, it is crucial to represent detailed aspects of the environment. For others, however, a state representation that mirrors the figure on the right is more effective.

An \textbf{action abstraction} determines the arrows moving between nodes in \autoref{fig:abstr_mdp_forest}. From the hiker's perspective, abstract actions are simply those behaviors that should be considered in choosing a course of action. Again supposing the hiker is trying to arrive safely at their camp, they may choose to navigate to a nearby hill to gain a view of the surroundings, or may navigate to a known landmark such as the bridge (from which they can quickly return to camp). In contrast, of course, the ``primitive" actions define the smallest possible choices available to the hiker---moving a toe, leg, or hand, for instance. Action abstraction has appeared under a variety of names such as skills, temporal abstraction, or macro-actions. I here use the general term of \textit{action abstraction} to capture each of these despite their differences.

Naturally, the two types of abstraction are intimately connected. As presented in \autoref{fig:abstr_mdp_forest}, there is an explicit sense in which they are related: by some accounts, an abstract action is just a behavior that takes an agent from one abstract state to another. This particular perspective connecting the two abstraction types has a rich history in AI, and will resurface several times throughout this dissertation. However, it is not the only sense in which the two forms of representation might be connected. \citet{konidaris2018skills} proves that algorithms that build an abstract state representation based on a given collection of abstract actions can preserve desirable properties. Indeed, dating back to the early work of \citet{dietterich2000hierarchical}, the two types of abstraction have been tied together. Near the end of the dissertation, I will return to a technical analysis of state-action abstractions (\autoref{part:state_action_abstraction}).

Finally, repeated application of state or action abstractions can induce \textbf{hierarchical abstraction}, through which entities can be represented at varying levels of granularity. In the forest, hierarchical abstraction may permit the hiker to reason using both the left \textit{and} the right representation, depending on the task at hand. As thinking time or data becomes more readily available, a more detailed representation may be used. If, however, a quick decision needs to be made, or if the world is fundamentally unpredictable in great detail at certain time horizons (for instance, it remains difficult to predict the weather a few days away), the hiker may opt for the representation on the right. Many of the fundamental open questions in the area center around hierarchical abstraction, with state and action abstraction serving as palatable chunks that can be analyzed independently. As with state-action abstraction, I will return to hierarchical abstraction near the end of the dissertation in \autoref{part:state_action_abstraction}.

\section{Thesis Statement}

With the main conceptual framework in play, I now highlight the central question addressed by this work:
\begin{center}
    \textit{How do reinforcement learning agents discover and make use of good abstractions?}
\end{center}

\noindent I answer this question by advancing the following thesis:
\vspace{1mm}
\input{misc/thesis_statement}

To defend this thesis, I introduce three desiderata that articulate which abstractions are useful in RL. At a high level, these desiderata state the following.
\begin{center}
    \emph{Good abstractions for RL are $\underbrace{\text{easy to discover}}_{\text{(D1)}}$ and $\underbrace{\text{enable efficient learning}}_{\text{(D2)}}$ of $\underbrace{\text{high value policies}}_{\text{(D3)}}$.}
\end{center}
I present more detail and justification for these desiderata in \autoref{sec:c2_desiderata}.

\section{Contributions}

The remaining defense of this thesis is organized as follows.

\paragraph{Part 1.} In \autoref{chap:background}, I provide necessary background on RL (\autoref{sec:c2_rl_background}) along with state abstraction (\autoref{sec:c2_state_abstraction}) and action abstraction (\autoref{sec:c2_action_abstr}). Then, I introduce and motivate the abstraction desiderata in more detail (\autoref{sec:c2_desiderata}).

\paragraph{Part 2.} The next part is dedicated to \textbf{state abstraction}. I present new algorithms and three intimately connected sets of analysis, each targeting the discovery of state abstractions that satisfy the introduced desiderata. In \autoref{chap:approx_state_abstr}, I develop a formal framework for reasoning about state abstractions that preserve near-optimal behavior. This framework is summarized by \autoref{thm:c3_approx_sa_main_result}, which highlights four such sufficient conditions for value-preserving state abstractions. Then, in \autoref{chap:state_abstr_lifelong}, I extend this analysis to the \textit{lifelong} RL setting, in which an agent must continually interact with and solve different tasks. The main insight of this chapter is the introduction of PAC state abstractions for the lifelong learning setting, along with results clarifying how to efficiently compute them. \autoref{thm:c4_pac_val_loss} illustrates the sense in which these abstractions are guaranteed to preserve good behavior, and \autoref{thm:c4_pac_sa_sample} shows how many previously solved tasks are sufficient to compute a PAC state abstraction. I highlight results from simulated experiments that illustrate the utility of the introduced types of state abstractions to accelerate learning and planning. Lastly, \autoref{chap:rlit} brings the tools of information theory to bear on state abstraction. I develop a tight connection between state abstraction and Rate-Distortion theory \cite{Shannon1948,berger1971rate} and the Information Bottleneck Method \cite{tishby2000information}, and exploit this connection to design new algorithms for efficiently constructing state abstractions that elegantly trade off between \textit{compression} and \textit{representation of good behavior}. I extend this algorithmic framework in a variety of ways, illustrating its power for discovering state abstractions that afford sample-efficient learning of good behavior.

\paragraph{Part 3.} I then turn to \textbf{action abstraction}. In \autoref{chap:options_for_planning}, I present analysis from \citet{jinnai2019opt} studying the problem of finding abstract actions that make planning as fast as possible---the main result states that this problem is NP-hard in general (under appropriate simplifying assumptions), and is even hard to approximate in polynomial time. Then, in \autoref{chap:elm_options}, I address the problem of constructing the \textit{predictive model} that accompanies high level behaviors in planning. Such a model enables an agent to estimate the outcome of executing the behavior in a given state (what will the world look like after I open this door?). In this chapter I introduce and analyze a new model for these high level behaviors, and prove that this simpler alternative is still useful under mild assumptions. I provide empirical evidence that indicates the new predictive model can serve as a suitable substitute for its more complicated counterpart. Lastly, in \autoref{chap:options_for_exploration}, I examine the potential for abstract actions to improve the exploration process. I describe an algorithm developed by \citet{jinnai2019opt_explore} that is based around the notion of constructing abstract actions that can easily reach all parts of the environment, and demonstrate that this algorithm can accelerate exploration on benchmark tasks.

\paragraph{Part 4.} Finally, I turn to the joint process of \textbf{state-action abstraction}. In \autoref{chap:vpsa}, I present a simple mechanism for combining state and action abstractions together. Using this scheme, I then prove which combinations of state and action abstraction can preserve representation of good behavioral policies in any finite MDP, summarized by \autoref{thm:c9_vpsa_main_result}. I next study the repeated application of these joint abstractions as a mechanism for constructing hierarchical abstractions. Under mild assumptions about the construction of the hierarchy and the underlying state-action abstractions, I prove that these hierarchies can \textit{also} preserve representation of globally near-optimal behavioral policies, as stated in \autoref{thm:c9_hierarch_loss}. I then conclude in \autoref{chap:conclusion} with reflections and directions forward. \\

\begin{figure}[h!]
    \centering
    \includegraphics[width=0.85\textwidth]{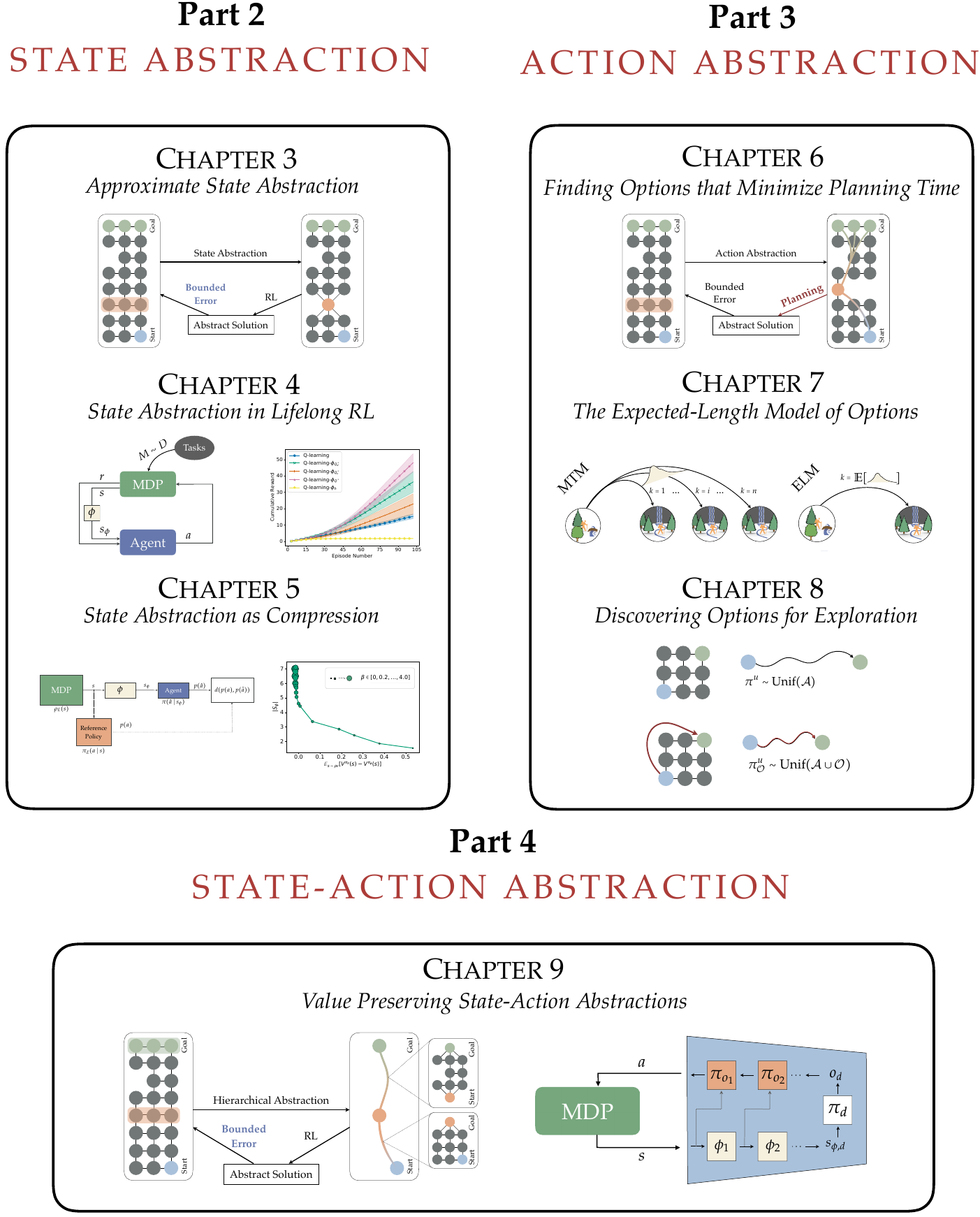}
    \caption{A visual overview of this dissertation.}
    \label{fig:c1_thesis_overview}
\end{figure}

Collectively, these results articulate a theory of abstraction in reinforcement learning. \autoref{fig:c1_thesis_overview} presents a visual overview of this dissertation.

\newpage

I now turn to providing necessary background and notation on RL and abstraction. For those familiar with RL, I recommend skipping to \autoref{sec:c2_state_abstraction}.

%% file: misc/thesis_statement.tex
\vspace{1mm}
\dbox{
\vspace{1mm}
\begin{center}
\large{\textsc{Thesis}}

\vspace{-1mm}
\oline{88} 

\vspace{4mm}

\begin{minipage}{0.9\linewidth}
\noindent \normalsize{By drawing on insights from computational complexity theory, decision-theoretic planning, and information theory, it is possible to design efficient algorithms for discovering abstractions that reduce the amount of experience or thinking time an RL agent requires to find a good solution.}
\end{minipage}
\vspace{4mm}
\end{center}
}

%% file: chapters/c2_background.tex
\begin{center}
\begin{minipage}{0.8\textwidth}
\textit{Parts of this chapter are based on ``Concepts in Bounded Rationality: Perspectives from Reinforcement Learning" \cite{abel2019philrl} and ``A Theory of State Abstraction for Reinforcement Learning" \cite{abeldc2019}.}
\end{minipage}
\end{center}
\vspace{2mm}

In this chapter, I bring clarity to the concepts of \textit{agent}, \textit{reward signal}, \textit{world}, and \textit{abstraction} by introducing the RL problem. In particular, I survey the key definitions and notation of the RL problem (\autoref{sec:c2_rl_background}) along with state (\autoref{sec:c2_state_abstraction}) and action abstraction (\autoref{sec:c2_action_abstr}).

\section{Reinforcement Learning}
\label{sec:c2_rl_background}

There are many possible choices for formalizing the agent-environment interaction. How is time to proceed---continuously, or in discrete rounds? What is the space of observations? Are all worlds of interest necessarily \textit{spatial} or filled with objects, at least in some capacity? With so many choices, it is not clear how to restrict attention to a suitable set of worlds. One natural response might be the space of \emph{computable} worlds, or perhaps those with polynomial-time laws that transition the world from one state to the next. Indeed, it is challenging to identify a set of worlds that is both suitably general while remaining restricted enough to be useful.

In computational RL, the space of relevant environments are those that may be modeled as a discrete-time Markov Decision Process (MDP)~\cite{puterman2014markov}. At a high level, the space of MDPs defines worlds in which the next reward and the probability of arriving at the next state of the world can be fully predicted by the \textit{current} world state (and perhaps, an agent's choice of action). Formally, an MDP is defined as follows.

\ddef{Markov Decision Process}{A discrete-time \textbf{Markov Decision Process} is a six tuple, $(\mc{S}, \mc{A}, R, T, \gamma, \rhoz)$, where:
\begin{itemize}
    \item $\mc{S}$: A set of states describing the possible configurations of the world.
    \item $\mc{A}$: A set of actions describing the possible choices available to an agent.
    \item ${R} : \mc{S} \times \mc{A} \times \mc{S} \ra \left[\textsc{RMin}, \textsc{RMax}\right]$: A reward function.
    \item ${T} : \mc{S} \times \mc{A} \ra \Delta(\mc{S})$: A transition function denoting the probability of arriving in the next state of the world after an action is executed in the current state.
    \item $\gamma \in [0,1)$: A discount factor, indicating an agent's preference between near-term and long-term rewards.
    \item $\rhoz \in \Delta(\mc{S})$: The probability of starting in each state.
\end{itemize}}

The ``Markov" in \textbf{M}DP indicates that the transition function, ${T}$, and reward function, ${R}$, both depend only on the current state of the world (and action), and \emph{not} the full state history. That is,
\begin{align}
    \label{eq:markov_t}
    T(s_{t+1} \mid s_t, a_t) &= p(s_{t+1} \mid s_t,a_t) = p(s_{t+1} \mid s_0, a_0,
    \ldots, s_t,a_t), \\
    {R}(s_t,a_t,s_{t+1}) &= {R}(s_t,a_t,s_{t+1} \mid s_0, a_0,
    \ldots, s_t,a_t).
    \label{eq:markov_r}
\end{align}
Here, and throughout this dissertation, I use $p(x)$ as shorthand for a probability mass function $\PR(X = x)$, where $X$ is a discrete random variable taking on values $x \in \mc{X}$.

\autoref{eq:markov_t} and \autoref{eq:markov_r} state that there exist functions that fully characterize the next state distribution and next reward from the current state and action alone. This assumption is remarkably useful for simplifying analysis while still retaining appropriate generality. Moreover, if any environment is \emph{not} Markov, it is typically feasible to roll the last $k \in \mathbb{N}$ steps of the world into a new memory-rich state representation, thereby yielding a Markov model. In this way, MDPs generalize Markov chains \cite{bremaud2013markov} and Markov reward processes \cite{reibman1989markov} by allowing an agent to influence the state distribution $T(s' \mid s,a)$ and reward $R(s,a,s')$ according to the agent's choice of action.

There are a few things to note about the reward function. First, there are three natural ways it may be expressed: $R(s)$, $R(s,a)$, and $R(s,a,s')$. Naturally, the third form is the most general, fully subsuming the first two. For this reason I introduce reward functions in the most general form, but will occasionally use $R(s)$ or $R(s,a)$ for brevity. Note that either of these are just shorthand or cases where all actions or all next states have the same reward for the given state. Second, while I have defined $R$ as a deterministic function, it can in general be a probability distribution with support $[\textsc{RMin}, \textsc{RMax}]$. Throughout the dissertation I will tend to treat $R$ as deterministic both in analysis and experiments unless otherwise noted. Lastly, I will sometimes assume the initial reward $r_0$ is sampled from some initial reward distribution $R(s_0)$ with the same support mentioned previously.

The central operation of RL is the repeated interaction between an agent and an MDP in discrete time steps. It is common to assume that the agent knows everything about the current \emph{state} of world: the agent has no uncertainty regarding which state it occupies, only what the reward and transition functions are. A more general formalism \emph{also} models hidden information, called the Partially Observable MDP (POMDP)~\cite{kaelbling1998planning}. In both POMDPs and MDPs, the agent interacts indefinitely with its environment with the goal of learning how to take actions that maximize long-term discounted reward. Throughout this dissertation, I make the standard assumption that the environment can be accurately modeled by an MDP, rather than a POMDP. Other work has considered a more general variant of the RL problem in non-Markovian settings~\cite{schmidhuber1991reinforcement,hutter2005universal,veness2011monte,leike2016nonparametric}. I focus only on agents learning in Markovian environments, though note that there is interesting and important work to be done in clarifying the role of abstraction in these general settings. I will often restrict attention only to finite MDPs, too, in which the state and action space are assumed to be finite.

From a methodological perspective, MDPs occupy an appropriate middle ground between simplicity and generality. I take it to be of fundamental importance to address prominent open questions in the context of simple formalisms for which those questions still remain open. By providing principled answers in these restricted settings, we can systematically build up our understanding and guide future research into richer settings rooted in first principles.

Under the assumption that RL agents will interact with an MDP, the RL problem can be stated more precisely as follows.
\ddef{Reinforcement Learning Problem (Formal)}{The \textbf{RL problem} is formalized as follows. An RL agent interacts with an MDP $M = (\mc{S}, \mc{A}, R, T, \gamma, \rhoz)$ by repeating the following four steps, letting $t=0$:
\begin{enumerate}
\item The agent receives a state $s_t \in \mc{S}$ and a reward $r_t \in \mathbb{R}$ from $M$.
\item The agent learns from this interaction and outputs an action, $a_t \in \mc{A}$.
\item The MDP outputs the next state, $s_{t+1} \sim T(s_{t+1} \mid s_t,a_t)$, and reward $r_{t+1} = R(s_t,a_t,s_{t+1})$.
\item Increment $t$.
\end{enumerate}}

The goal of an RL agent interacting with an MDP is to make decisions that maximize long term discounted reward:
\begin{equation}
\sum_{t=0}^{\infty} r_t \gamma^t.
\end{equation}

The standard objective of an RL agent is to solve for behavior that will prescribe what to do from \emph{any} state the agent might occupy. Note, though, that this is a stronger notion than what is strictly necessary. If the agent starts in state $s_0 \sim \rhoz$, then there may be some states of the environment that are difficult to reach. In this sense, it might be more effective to focus attention on those states that are likely to be visited during the agent's lifetime. This insight will emerge shortly when we discuss the quality of an agent's decision.

We ground this notion of behavior in all states in the MDP with a policy:
\ddef{Policy ($\pi$)}{A \textbf{policy}, $\pi : \mc{S} \ra \Delta(\mc{A})$, is a prescription for behavior for any state in the given MDP.
\vspace{2mm}}
Note that in the general case a policy can be stochastic. To make a decision, then, the agent samples $a \sim \pi(\cdot \mid s)$. Given that deterministic policies are often of interest, I will also use $\pi(s)$ to denote a deterministic policy.

To characterize the notion of expected long term expected discounted reward, we next introduce the value ($V$) and action-value ($Q$) functions.

\ddef{Value Function ($V$)}{\label{def:c2_value_function}The \textbf{value function} $V^\pi : \mc{S} \ra \mathbb{R}$, under a policy $\pi$ of a state $s \in \mc{S}$ is denoted
\begin{equation}
V^\pi(s) := \sum_{a \in \mc{A}} \pi(a\mid s) \sum_{s' \in \mc{S}} R(s,a,s') + \gamma T(s' \mid s,a) V^{\pi}(s').
\end{equation}}

\ddef{Action-Value Function ($Q$)}{The \textbf{action-value function}, $Q^{\pi} : \mc{S} \times \mc{A} \ra \mathbb{R}$, under a policy $\pi$ of a state $s \in \mc{S}$ and action $a \in \mc{A}$ is denoted
\begin{equation}
Q^\pi(s,a) := \sum_{s' \in \mc{S}} R(s,a,s') + \gamma T(s' \mid s,a) V^{\pi}(s').
\end{equation}}

I denote the value ($V$) and action-value ($Q$) functions under the optimal policy as $V^*$ and $Q^*$ respectively, which are determined by applying the $\max$ operator to the Bellman Equation \cite{bellman1956dynamic}:
\begin{equation}
V^*(s) := \max_a \left(\sum_{s'} R(s,a,s') + \gamma T(s' \mid s, a) V^*(s') \right).
\end{equation}
Since the support of $R$ is the real valued interval $[\textsc{RMin},\textsc{RMax}]$, I will denote
\begin{align}
    \textsc{QMax} &= \textsc{VMax} \leq \frac{\textsc{RMax}}{1-\gamma}, \\
    \textsc{QMin} &= \textsc{VMin} \geq \frac{\textsc{RMin}}{1-\gamma},
\end{align}
as upper and lower bounds on the value achievable in a discounted, infinite horizon RL problem. That is, for any state $s$ in any MDP,
\begin{equation}
    \textsc{VMin} \leq V^*(s) \leq \textsc{VMax}.
\end{equation}

In general, the value of a policy will serve as our primary method for evaluating an agent's behavior, and in many cases, for determining learning progress. Recently, \citet{bellemare2017distributional} propose an extension to the classical Bellman Equation that translates the expected future returns into a distribution over future returns. Later work has developed RL algorithms that learn relative to this distributional objective to great effect \cite{dabney2018distributional,dabney2018implicit,hessel2018rainbow}, and has given rise to new explanatory models of the role dopamine neurons play in reward error prediction \cite{dabney2020distributional}. While I do not attend to these directions, there is interesting work to be done in combining the ideas of abstraction and distributional RL.



\begin{figure}[t!]
\begin{center}
\includegraphics[scale=0.3]{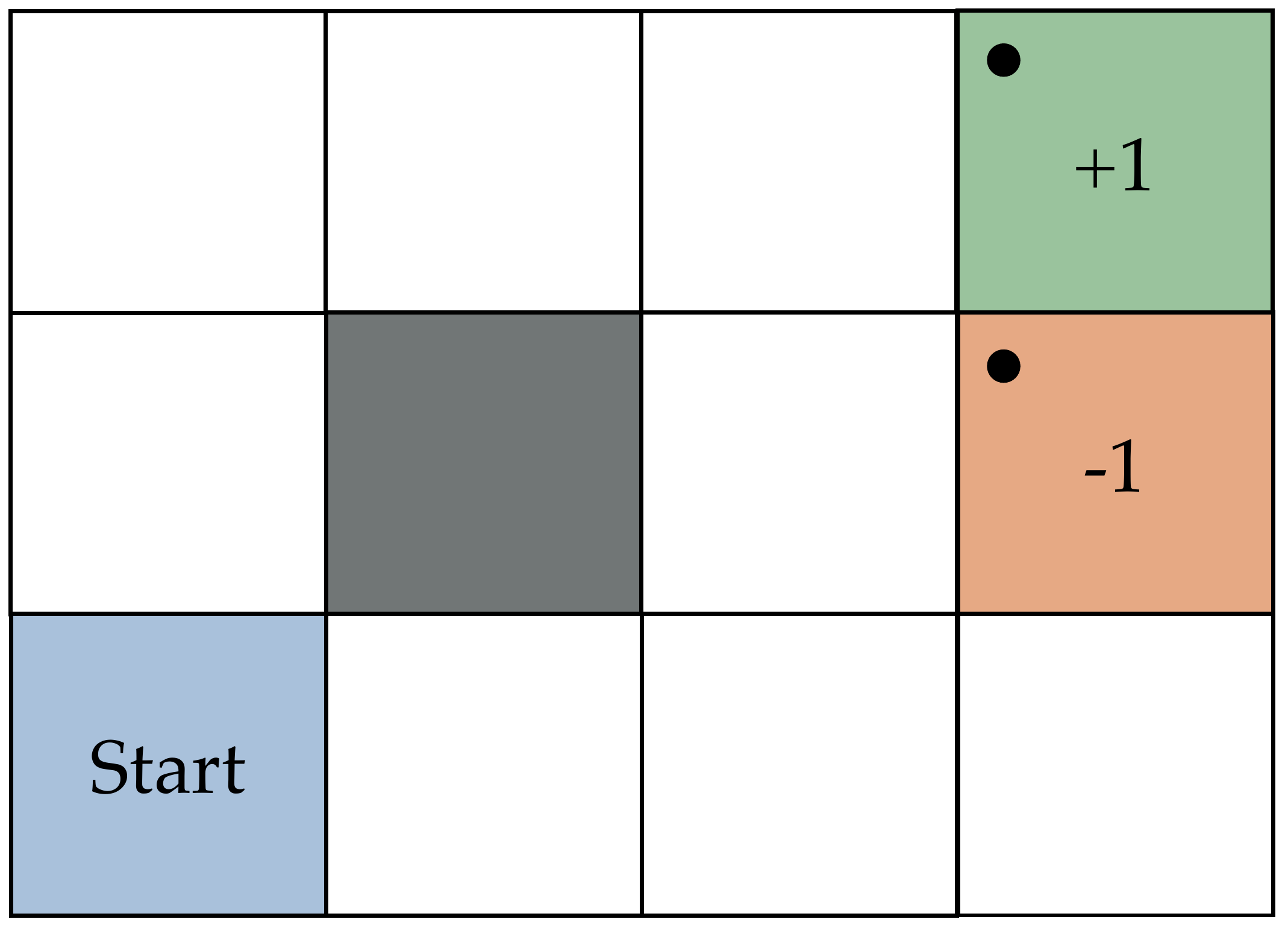}
\end{center}
\caption{The classical grid world by~\citet{russell1995modern}.} 
\label{fig:c2_rn_gridworld}
\end{figure}

\paragraph{Example.} Let us now consider an example: the Russell and Norvig grid world, used by the classic AI textbook \cite{russell1995modern}. The Russell and Norvig grid world is a discrete, $4 \times 3$ two-dimensional grid in which each state corresponds to the agent inhabiting one of the eleven empty grid cells (\autoref{fig:c2_rn_gridworld}). For the purpose of clarity, I adopt a factored representation for states. Specifically, each state will be defined as $(x, y)$, for $x \in \mathbb{N}_{[1:4]}$, $y \in \mathbb{N}_{[1:3]}$. Here, $(1,1)$ denotes the state in the bottom left corner with $x=1$ and $y=1$, with $x$ increasing as the agent moves to the right and $y$ increasing as the agent moves up. Naturally, an enumerated state space representation could be adopted, too, according to which the states are represented by a single number (and thus not imposing any notion of ``space" onto the problem). The grid world MDP is then defined as follows.

\begin{enumerate}
\item $\mc{S} = \{(1,1), (2,1), (3,1), (4,1), (1,2), (3,2), (4,2), (1,3), (2,3), (3,3), (4,3)\},$
\item $\mc{A} = \{\uparrow,\ \ra,\ \downarrow,\ \la\}$,
\item $R(s,a,s') = \begin{cases}
\hspace{6.57pt}1& s' = (4,3), \\
-1& s' = (4,2), \\
\hspace{6.57pt}0& \text{otherwise}, \\
\end{cases}$
\item $T(s' \mid s,a) = \indic\{s' = \texttt{grid\_\hspace{1pt}move}(s,a)\}$
where,
\[
\hspace{-16mm}\texttt{grid\_\hspace{1pt}move}(s ,a) = \begin{cases}
(s.x, \min(s.y + 1, 3)) &a =\ \uparrow \text{and}\ s.x \neq 2,\\
(\min(s.x + 1, 4), s,y)  &a =\ \ra \text{and}\ s \neq (1,2),\\
(s.x, \max(s.y - 1, 1)) &a =\ \downarrow \text{and}\ s.x \neq 2,\\
(\max(s.x - 1, 1), s,y) &a =\ \la \text{and}\ s \neq (3,2),\\
s& \text{otherwise},
\end{cases}
\]
\item $\rhoz(s) = \indic\{s = (1,1)\},$
\item $\gamma = 0.99.$
\end{enumerate}

That is, the reward function outputs zero for every transition unless the agent enters state (4,3) or (4,2), in which case it receives +1 and -1 respectively. The four actions move the agent in each cardinal direction with the exception of moving into the wall or edge of the environment, which yields no effect. Lastly, when the agent arrives in either (4,3) or (4,2), the episode ends and the agent moves back to $s_0$ to start the next episode.

From the perspective of the learning agent, it does not know that this grid world has a Cartesian coordinate system, and that the action associated with the symbol ``$\ra$" will typically increase its \texttt{x} coordinate. Instead, the agent must repeatedly experiment with the execution of different behaviors. With the initial state distribution $\rhoz$ only assigning mass to $(1,1)$, the agent is first presented with choosing an action in the lower left state. Without prior information, each state is equally informative, and so the agent might choose $\la$. Upon execution of this action, the MDP samples $s_1 \sim T(\cdot \mid s_0=(1,1), a_0=\ \la)$, and $r_1 = R(s_0=(1,1),a_0=\ \la,s_0'=(1,1))$. Now, after this single action execution, the agent occupies state $(1,1)$, since the $\la$ action moved the agent into the wall. After receiving the first bit of signal from the environment the agent has the opportunity to learn something. What effect did applying the ``$\la$" action in $s=(1,1)$ have? How much reward was received? As more data is gathered, agents will be better positioned to give high confidence answers to these questions for different $(s,a)$ pairs throughout the MDP.

\begin{figure}[b!]
    \centering
    \includegraphics[height=\figvdim]{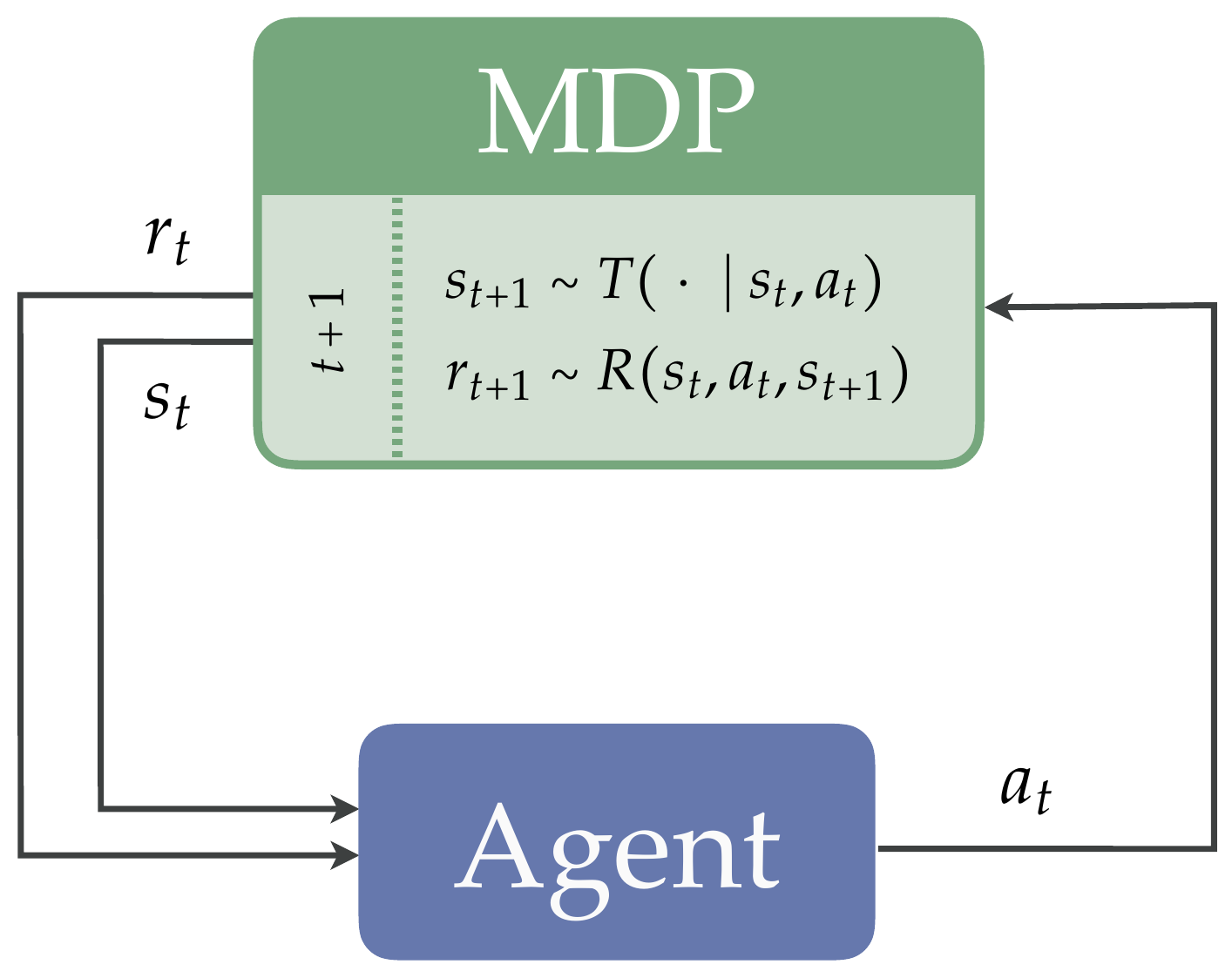}
    \caption{The RL problem when an agent interacts with an MDP.}
    \label{fig:c2_rl_mdp}
\end{figure}

This process continues indefinitely: the agent receives a state and reward pair from the environment ($s_t, r_t$), and chooses an action $a_t$. The MDP then transitions to the next state and generates the next reward. This process is pictured in \autoref{fig:c2_rl_mdp}.

It is often useful to allow the agent to periodically \textit{reset} by resampling $s_0 \sim \rhoz$. Such a system typically fixes a finite horizon $H \in \mathbb{N}$ and allows the agent to execute up to $H$ actions before resetting to a start state $s_0 \sim \rhoz$. This setting is referred to as \textit{episodic} RL, with each episode consisting of at most $H$ steps, as is the case in the grid world above. In some cases, arriving at special goal or trap states (such as (4,3) and (4,2) in the present example) can also have the effect of resetting the agent. All of this is in contrast to continual learning in which the agent repeatedly interacts with its environment and is never allowed to reset.

\subsection{Canonical RL Algorithms}

What, then, does it look like to \textit{solve} the RL problem? A typical solution comes in the form of a learning algorithm that captures a particular strategy for mapping a history of experiences, $(r_0, s_0, a_0, r_1, s_1, a_1, \ldots)$ to an action. In this way, the space of RL algorithms is roughly the space of all functions that map arbitrary length histories of experience to a choice of action. Good algorithms are those whose action selection becomes better with time, as measured by the sum of discounted rewards received.

\begin{figure}[t!]
    \centering
    \includegraphics[height=45mm]{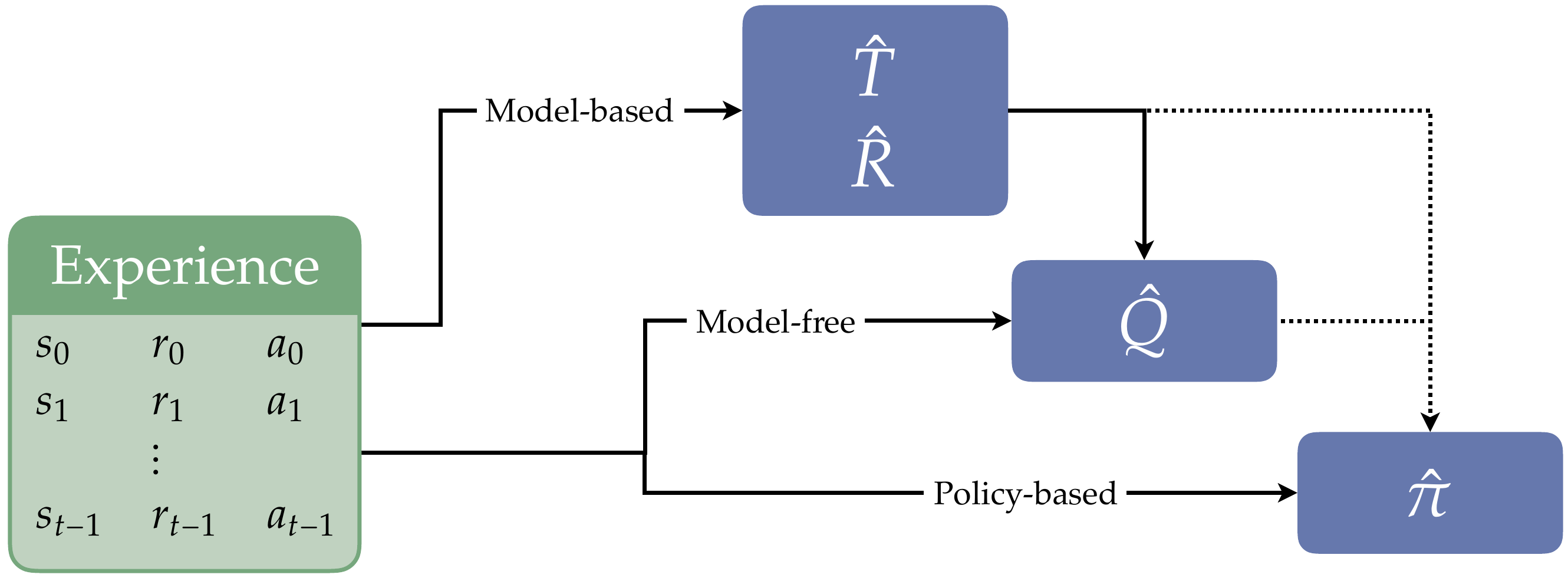}
    \caption{The different families of RL algorithms.}
    \label{fig:c2_rl_alg_types}
\end{figure}

RL algorithms can be divided into three broad categories: policy-based, model-free, and model-based. Each category is an answer to the question: ``which functions are being estimated during learning?". If an RL algorithm maintains estimates of the transition and reward functions, $\hat{T}$ and $\hat{R}$, then it is said to be model-based. If the reward and transition functions are not estimated, but the action-value function $Q$ is, then the algorithm is model-free. Lastly, if all that is estimated is the policy directly, then it is policy-based. These are relatively loose boundaries, however, as many algorithms often compute partial solutions to different functions or carry out implicit computational work that resembles construction of one of these functions \cite{vanseijen2015deeper}. A rough division between these three approaches is pictured in \autoref{fig:c2_rl_alg_types}.


In model-based RL, the transition and reward functions are typically estimated explicitly. Then, using these estimates $\hat{T}$ and $\hat{R}$, the agent often constructs a simulated MDP, $\hat{M}$, which can be used to do explicit search for good behavior, or to evaluate different policies. That is, given simulation access to an MDP $\hat{M}$ that is sufficiently similar to the environmental MDP $M$, the agent can perform computations on $\hat{M}$ to construct $\hat{\pi}^*$ or perhaps $\hat{Q}^*$, which can induce a policy by choosing the action with highest value.

In model-free and policy-based RL, the agent typically maintains an estimate of the action-value function $Q$ or a policy $\pi$ directly. Various mechanisms are adjusted in order to learn these functions faster by assigning credit more efficiently, generalizing more robustly, or exploring more elegantly.

There are good arguments to adopt each style of algorithm depending on the context. For instance, recent work by \citet{sun2019model} illustrates a gap in the efficiency between model-based and model-free approaches under a particular method of dividing the two families. On the other hand, it has proven difficult to estimate accurate models. For instance, even nearly-accurate one-step models are known to lead to an exponential increase in the error of $n$-step predictions as a function of the horizon~\cite{kearns2002near,brafman2002rmax}, though recent approaches show how to diminish this error through smoothness assumptions~\cite{asadi2018lipschitz}. Moreover, composing an accurate one-step model into an $n$-step model is known to give rise to predictions of states dissimilar to those seen during training of the model, leading to poor generalization~\cite{talvitie2017self}. Notably, model-free and policy-based methods have enjoyed a great deal of success when combined with deep neural networks, giving rise to so called ``deep RL" methods that learn effectively in a variety of challenging domains from Atari \cite{mnih2015human} to robotics \cite{levine2016end}. For more discussion on the relationship and interplay of these two families, see work by \citet{asadi2015strengths}. To build further intuition, I now introduce two classical RL algorithms, one model-free and one model-based.

\paragraph{Model-Based: R-Max.} One of the earliest successful model-based algorithms was that of \textit{Explicit Explore-Exploit} or $E^3$, developed by \citet{kearns2002near}. This pioneering work established early theoretical guarantees for the efficiency $E^3$, which has since paved the way for a long and fruitful sequence of new and improved algorithms. Around the same time, \citet{brafman2002rmax} introduced the algorithm R-Max, which uses a similar exploration strategy to $E^3$, and will be used in analysis and empirical study throughout this dissertation.

\input{algorithms/r_max}

R-Max makes decisions by exploring so as to seek out every opportunity for high-performing behavior. That is, R-Max initially supposes it inhabits the maximally rewarding MDP: all rewards are believed to be \textsc{RMax}, the maximal possible reward, and all transitions are assumed to be self-loops. Then, the algorithm acts according to the optimal policy in this optimistic MDP to explore and collect experience. Consequently, the algorithm will effectively try \textit{new} actions that it does not yet know about. R-Max uses all collected experience to inform empirical estimates of the reward and transition function for each $(s,a)$ pair it encounters. Once enough data is collected for a particular $(s,a)$ pair, the empirical estimate of the reward and transition replace the optimistic estimate, and the behavioral policy is recomputed based on this mixture of known and optimistic reward and transitions. Simplified pseudocode for R-Max is presented in \autoref{alg:r_max}, where $\texttt{act}(s_t,a_t)$ defines a single interaction with the MDP. For more details on R-Max, see \citet{brafman2002rmax} or Chapter 2.1 of \citet{Strehl2009}.

\paragraph{Model-Free: $Q$-learning.} The second and perhaps most canonical RL algorithm is called $Q$-learning, first introduced by \citet{watkins1992q}. $Q$-learning maintains an estimate of the $Q$ function for each state-action pair, and proceeds on the basis of performing one simple update to this $Q$ function estimate based on the last experience, $(s_{t-1}, a_{t-1}, r_{t-1}, s_t)$, and a learning rate $\alpha \in [0,1]$. That is, we first initialize a $Q$-function according to some protocol, such as choosing $Q$ values uniformly at random from the interval $\left[\textsc{QMin}, \textsc{QMax}\right]$. Or, more commonly, the initial $Q$ function is set to be either $0$, or is set optimistically where $\hat{Q}_0(s,a) = \textsc{QMax}$ for all $s,a$. Then, actions are chosen according to the \textit{greedy policy}, defined as follows.
\ddef{Greedy Policy}{The \textbf{greedy policy} with respect to some $Q$ function is:
\begin{equation}
    \pi_Q(a \mid s) = \indic\left\{a = \argmax_{a'}Q(s,a')\right\},
\end{equation}
where I assume that $\argmax$ returns a single entity, breaking ties consistently according to any fixed method.}

Note that if the given $Q$ is in fact $Q^*$, then $\pi_{Q^*}$ will be optimal. However, prior to learning $Q^*$, acting greedily will not sufficiently explore the environment as the agent may over commit to locally promising but globally poor solutions. Indeed, the main guarantee provided about $Q$-learning is that its asymptotic performance is optimal subject to the assumption that every state-action pair is experienced infinitely often in the limit (among other assumptions) \cite{watkins1992q}. Clearly, for many choices of initialization $\hat{Q}_0$, the greedy policy may force $Q$-learning to only experience a small subset of the state-action space, thus violating the conditions necessary to ensure convergence to optimal behavior.

The most common method of overcoming this difficulty is to pair $Q$-learning with an $\eps$-greedy policy, which chooses an action uniformly at random with some small $\eps \in [0,1]$ probability, defined as follows.

\ddef{$\eps$-Greedy Policy}{The \bs{$\eps$}-\textbf{greedy policy}, for $\eps \in [0,1]$, with respect to some $Q$ function is:
\begin{equation}
    \pi_{Q,\eps}(a \mid s) =
    \begin{cases}
        1-\eps& a = \argmax_{a'} Q(s,a'), \\
        \frac{\eps}{|\mc{A}| - 1}& \text{otherwise}.
    \end{cases}
\end{equation}}

\input{algorithms/q_learning}

Using an $\eps$-greedy policy (or another choice of stochastic policy, such as a softmax \cite{asadi2017alternative}) $Q$-learning makes decisions that are greedy with respect to its current estimate of the $Q$ function and slowly updates this estimate to be more accurate over time. Specifically, when an experience $(s_{t-1}, a_{t-1}, r_{t-1}, s_t)$ takes place, the following update is applied:
\begin{equation}
    \hat{Q}_{t}(s_{t-1}, a_{t-1}) = (1-\alpha)\hat{Q}_{t-1}(s_{t-1}, a_{t-1}) + \alpha (r_{t-1} + \max_{a' \in \mc{A}} \hat{Q}_{t-1}(s_t,a')).
\end{equation}
The full pseudocode of $Q$-learning is presented in \autoref{alg:q_learning}. For more detail on $Q$-learning, see the original work by \citet{watkins1992q}.


I will shortly contrast how these two algorithms behave in the grid world described earlier in the section. First, we must attend to the broader question: what does it look like to compare different RL algorithms? 

\subsection{Evaluation in RL}

As with other areas of machine learning, there are two broad approaches for evaluating and understanding an RL algorithm. First, theoretical guarantees may be established about an algorithm, often presented in the form of convergence to desirable fixed points, a bound on the sample complexity of exploration~\cite{kakade2003sample}, the algorithm's (Bayesian or frequentist) regret \cite{auer2007logarithmic}, or a KWIK bound on the algorithm~\cite{Li2011}. Second, empirical investigations of hypotheses relating to algorithms and their properties, or analysis on explanatory benchmark tasks. These may include visuals of learned policies, value functions, or representations, or, most commonly, learning curves illustrating the agents learning process.

In this section I provide a brief overview of these two approaches to evaluation. The focus of this dissertation is on algorithms (and the abstractions they use) that can reliably and quickly find near-optimal behavioral policies in any MDP. Naturally, there is much more to the RL problem that I can not cover here. In particular, we might also care about robustness, explainability, how an algorithm handles failures, safety, generalization, and many other properties of interest.


\paragraph{Sample Complexity of Exploration.} After conducting an experiment that involves a learning algorithm interacting with a chosen MDP, what is useful to have learned about the algorithm? There are many possible answers. The most pressing is typically related to the sample efficiency of the algorithm; how many experiences are needed until the algorithm will achieve a satisfactory level of performance? This notion is grounded in several different measures in RL, with the first being the sample complexity of exploration introduced by \citet{kakade2003sample}, based in part by the analysis done by \citet{kearns2002near} and \citet{brafman2002rmax}.

\ddef{Sample Complexity of Exploration~\cite{kakade2003sample}}{Let $\eps \in \mathbb{R}_{\geq 0}$ denote accuracy and $\delta \in \mathbb{R}_{\geq 0}$ denote an allowed failure probability. The expression,
\begin{equation}
\zeta\left(\frac{1}{\eps},\frac{1}{\delta},|\mc{S}|,|\mc{A}|,\frac{1}{1-\gamma},\textsc{RMax}\right)
\end{equation}
is a \textbf{sample complexity} bound for a learning algorithm $\mathscr{A}$ if the following holds. For any finite MDP $M$, $\textsc{RMax} > 0$: let $\mathscr{A}$ interact with $M$, starting in $s_0 \sim \rhoz$, resulting in the process $s_0, r_0, a_0, s_1, \ldots$, Then, with probability at least $1-\delta$, the number of time steps such that $V^{\mathscr{A}}(s_t) < V^*(s_t) - \eps$, is at most $\zeta(\frac{1}{\eps},\frac{1}{\delta},|\mc{S}|,|\mc{A}|,\frac{1}{1-\gamma},\textsc{RMax})$.}

The sample complexity captures how many mistakes we expect an algorithm to make in any finite MDP, if left to run indefinitely. That is, a sample complexity of $|\mc{S}|$ will tell us that the agent might make one mistake per state in the MDP. The sample complexity is intended to clarify an RL algorithm's effectiveness for exploring its environment while also learning to make good decisions in that environment. It determines, for a given $\eps$ and $\delta$, how many mistakes the agent is expected to make before acting in a near-optimal way. The Probably Approximately Correct in Markov Decision Processes (PAC-MDP)~\cite{Strehl2009} criterion expresses a desirable guarantee about an algorithm's sample complexity, inspired by the seminal work establishing the learnability of concepts in supervised learning introduced by~\citet{Valiant1984}. PAC-MDP algorithms are those that achieve a polynomial sample complexity \textit{and} computational complexity with respect to $\eps, \delta$ and the MDP parameters. For an early survey of PAC-MDP approaches, see work by~\citet{Strehl2009}.


\paragraph{Regret.} The regret compares an agent's total expected accumulated reward to that of the optimal policy from the time of the agent's first action execution. 

\ddef{Regret}{Let $\mu^*(M)$ denote the average reward of the optimal policy in MDP $M$:
\begin{equation}
\mu^*(M) = \mu^*(M,s) = \max_{\pi \in \Pi} (M, \pi, s).
\end{equation}
Further let $G(M,\mathscr{A},s_0,H)$ denote the total accumulated reward by the agent after $H$ steps in MDP $M$ with start state $s_0$:
\begin{equation}
G(M,\mathscr{A},s_0,H) = \sum_{t=0}^{H-1} r_t
\end{equation}
Then the following is the \textbf{regret} of an algorithm, $\mathscr{A}$, with finite horizon $H$, on MDP $M$.
\begin{equation}
\text{Regret}(M, \mathscr{A}, s_0, H) := H \mu^*(M) - G(M,\mathscr{A},s_0,H).
\end{equation}}

Regret differs from the sample complexity of exploration in several critical ways. First, the magnitude of each mistake the agent makes matters. While sample complexity counts the number of mistakes, the size of the mistake made will contribute to an agent's overall regret. Second, in measuring regret, the agent's long term behavior must approach the global optimum, and not a locally optimal policy for the region of the state space the agent is in. Regret is also commonly presented in two slightly different forms: \emph{Bayesian} regret, in which the agent is compared to optimal behavior relative to its prior, and \emph{Frequentist} regret, in which the agent is compared to the true optimal behavior. For more on these two measures and their relationship, see work by \citet{dann2017unifying}.

\paragraph{KWIK.} \citet{Li2011} introduced the Knows What It Knows (KWIK) criterion, which captures prominent elements of the PAC objective, but also incorporates an adversarial element. In KWIK, we suppose there exists an input set $\mc{X}$ and output set, $\mc{Y}$. A given hypothesis class, $\mc{H}$, contains a subset of possible functions from $\mc{X}$ to $\mc{Y}$. The agent's goal is to learn some target function, $h^* \in \mc{H}$ during the following repeated process:
\begin{itemize}
\item The agent and an adversary are given $\eps \in \mathbb{R}_{\geq 0}$, $\delta \in \mathbb{R}_{\geq 0}$, and $\mc{H}$.
\item The adversary selects the target function $h^* \in \mc{H}$.
\item Repeat:
\begin{itemize}
\item The adversary selects $x \in \mc{X}$ and gives it to the learner.
\item The learner predicts an output, $\hat{y} \in \mc{Y} \cup \perp$.
\item If $\hat{y} \neq \perp$, then it must be accurate: $|\hat{y} - h^*(x)| \leq \eps$, otherwise, the run is a failure.
\item If $\hat{y} = \perp$, then the learner observes $z \in \mc{Z}$ of the output, where $z = y$ in the deterministic case, and has noise determined by nature in the stochastic case.
\end{itemize}
\item The probability of a failed run must be bounded by $\delta$.
\item Over the course of a run, the total number of steps on which $\hat{y} =\perp$ must be bounded by $B(\eps, \delta)$.
\end{itemize}

KWIK has been used as a further evaluative criteria in the RL to bound the number of experiences needed for an agent to accurately learn the model, $T$ and $R$~\cite{diuk2008objects,walsh2009exploring,walsh2010integrating,walsh2011blending}. KWIK typically deals with learning the transition model or rewards directly, and so is naturally suited for model-based RL. It provides stronger guarantees than sample complexity, since both samples and the target function are chosen in an adversarial way.


\paragraph{Empirical Evaluation in RL.} Empirical evaluation in RL is naturally diverse. The most standard experiments assess aspects of the sample efficiency of a given RL algorithm. A typical experiment proceeds as follows. Choose an MDP, $M$, and a collection of learning algorithms, $\mathscr{A}_1, \ldots, \mathscr{A}_n$. Allow each of the $n$ algorithms to interact with the MDP for some number of steps, $H$. Then, compare the total cumulative reward received by each algorithm. In this way, we test the relationship between the amount of experience the algorithm has and its overall cumulative reward.


The results from such experiments are often represented as learning curves where the X-axis denotes experience and the Y-axis is some measure of performance on the task, either total reward accumulated or average reward accumulated per time step.

\begin{figure}[t!]
\begin{center}
\subfloat[Cumulative\label{fig:c2_grid_cumul_rew}]{\includegraphics[width=0.45\textwidth]{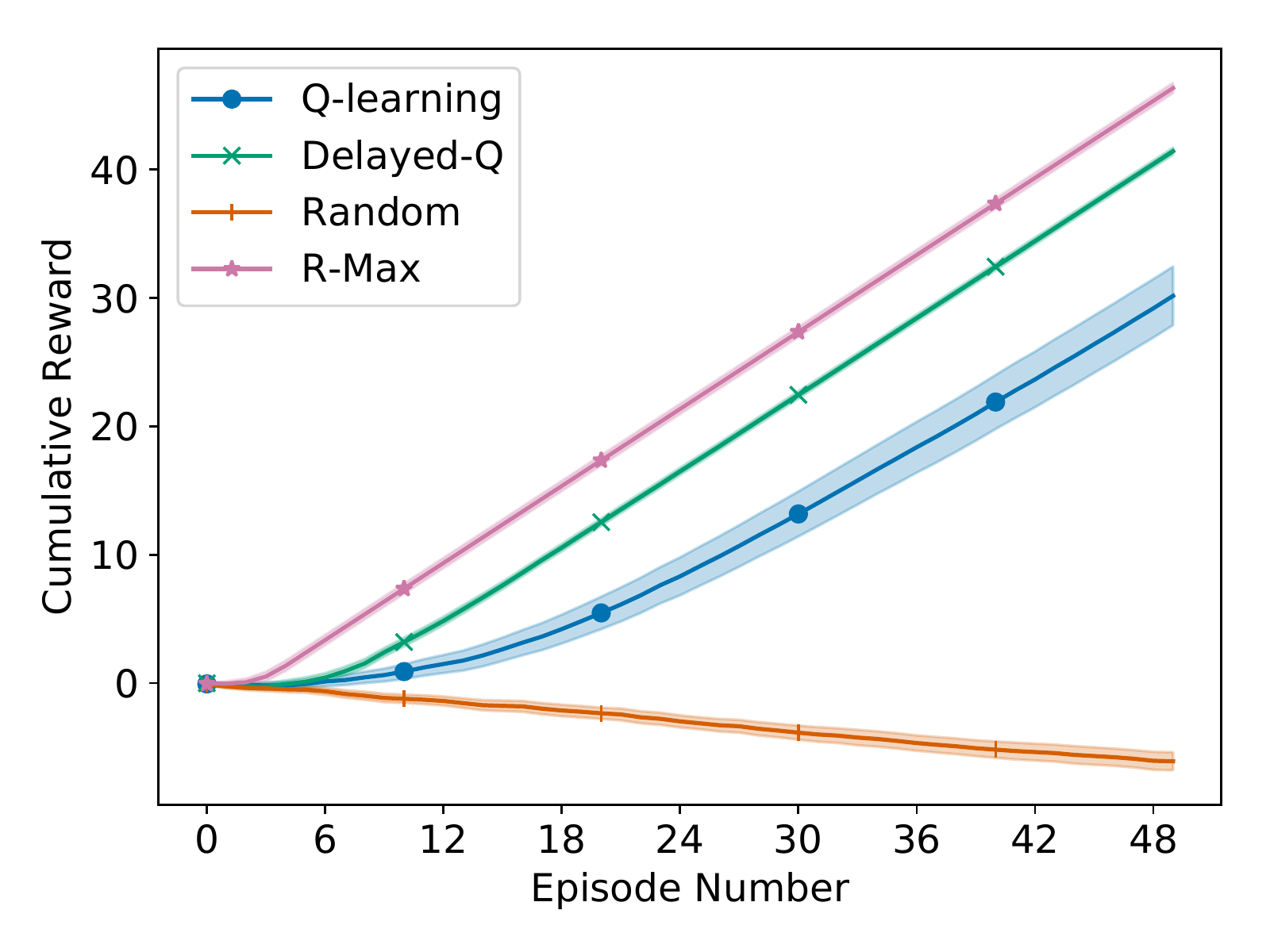}} \subfhspace
\subfloat[Average\label{fig:c2_grid_avg_rew}]{\includegraphics[width=0.45\textwidth]{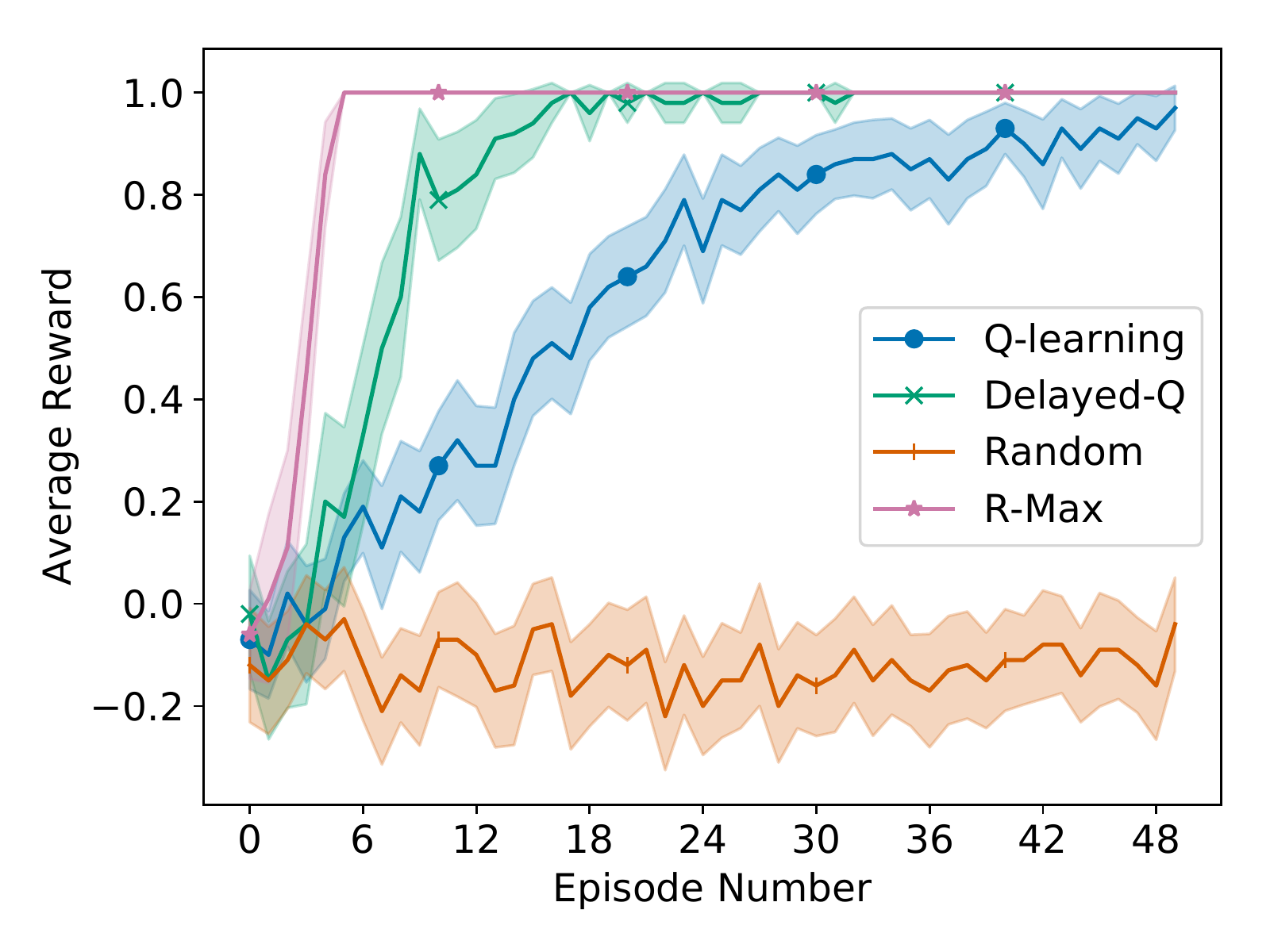}}
\caption{Example learning curves showing cumulative (left) and average (right) reward for a variety of RL algorithms.}
\label{fig:c2_rn_grid_learning_curves}
\end{center}
\end{figure}

Example learning curves for each of the two algorithms discussed previously (R-Max and $Q$-learning) acting in the Russell \& Norvig grid world from \autoref{fig:c2_rn_gridworld} are presented in \autoref{fig:c2_rn_grid_learning_curves}. I additionally include two other approaches for further contrast. First, a random actor (in orange) that does no learning, but simply always samples from the uniform distribution over actions. Second, a more sophisticated version of $Q$-learning called Delayed-$Q$ (green) \cite{strehl2006pac}, that enjoys similar theoretical guarantees to R-Max (both are PAC-MDP).

The two curves present the same results from a slightly different perspective. On the left, I show that mean cumulative reward over episodes, presented with 95\% confidence intervals from 100 runs of the experiment. On the right, I show the average reward per episode, again with 95\% confidence intervals (from the same 100 runs).

These plots give us insight into how the different learning algorithms perform on this grid world MDP. First note that on the first handful of episodes, all of the algorithms perform nearly identically, on average. This is because nearly any algorithm designed to learn on any MDP will require some number of samples before finding a reasonable behavioral policy. If an algorithm were to behave well from the outset, it is likely overfit to the task of interest and is susceptible to learn slowly in other MDPs, by No Free Lunch \cite{wolpert1996lack}. Second, note that R-Max and Delayed-$Q$ ultimately find the optimal policy in the sample budget allotted. From the definition of the problem, recall that the maximum reward receivable in any given episode is one. Therefore, note that the average plot indicates that both R-Max and Delayed-$Q$ are able to achieve one reward per episode after the first ten or so episodes---this indicates that they have both found near-optimal policies, and in a relatively sample efficient manner. In contrast, the random approach unsurprisingly never finds a reasonable policy. Since the $-1$ square in the grid is easier to reach on average than the $+1$ square, the random approach tends to lose reward over time. Lastly, we find that $Q$-learning ultimately does find a reasonable policy, but it requires more samples in order to produce high reward behavior.

These are roughly the questions we ask in running such an experiment: 1) What is the initial performance like? 2) How long does it take the agent to converge to its eventual best policy? 3) What is the value of the best policy the algorithm discovers? Learning curves like the ones presented can help answer each of these questions, thereby giving insight into the performance of the algorithm on the MDP of interest. Naturally, other questions are of interest as well, such as how well an algorithm can perform when given a new task, or its robustness to different hyper-parameter settings.


Beyond learning curves, there are many other facets of an RL algorithm that determine its effectiveness. Naturally this space is far too vast to summarize, but one important focus is on how well algorithms address each of the \textit{subproblems} of RL discussed briefly in the previous chapter. That is, in confronting the RL problem, algorithms must by necessity tackle several subproblems of deep interest to the broader machine learning and AI communities, including generalization, the explore-exploit dilemma, credit assignment, and planning. I next provide detail on one subproblem of special interest, planning. For more on the explore-exploit dilemma, see the recent book on bandits by \citet{lattimore2018bandit}. For more on credit assignment, see the recent work by \citet{harutyunyan2019hindsight}. Lastly, for more on generalization as understood in supervised learning, see the classic works by \citet{Valiant1984}, \citet{vapnik1971uniform}, or the book by \citet{shalev2014understanding}.

\paragraph{Planning.} A computational practice critical to RL is \emph{planning}. The key difference between RL and planning is that the full model of the environment is given as input in planning, so there is no uncertainty around $T$ or $R$. Planning was originally studied early on in AI by~\citet{bellman1956dynamic} and \citet{newell1957empirical} among others. The planning problem is commonly formalized in the context of a language (such as \textsc{Strips}~\cite{fikes1971strips}) that provides a high level scheme for expressing the consequences of decisions available to an agent in a given domain. Under the assumption that environments of interest may be modeled as MDPs, there is a decision-theoretic version of the planning problem tightly connected to RL.
\ddef{MDP Planning Problem}{The \textbf{MDP planning problem} is defined as follows: \texttt{given} an MDP $M$ as input, \texttt{return} a sequence of actions that achieves maximal expected discounted reward when executed in $M$,
\begin{equation}
\sum_{t=1}^{\infty} r_t \gamma^t.
\end{equation}
}
In the above version of the problem the required solution is a sequence of actions. Of course, other possible solutions may be of interest as well, such as a policy or value function, though naturally these tend to be harder to compute.

Depending on the constraints placed on the problem, the general (propositional) planning problem is known to be PSPACE-Complete~\cite{bylander1991complexity,bylander1994computational}, but solving for optimal behavior in an MDP is known to be P-Complete in the size of the environment~\cite{papadimitriou1987complexity,littman1995complexity}. Many problem representations are known to grow super-polynomialy with the number of variables that characterize the domain and robust action spaces are often large or even continuous, making many planning problems on the scale of the real world computationally intractable.

The standard algorithm in MDP planning is a dynamic programming algorithm called Value Iteration (VI), first introduced by \citet{bellman1957markovian}. As per the name, the idea of VI is to repeatedly propagate value to adjacent states, starting from some arbitrary initialized value function $V_0$ and terminating when the optimal value function is realized. Pseudocode for VI is presented in \autoref{alg:value_iteration}.

\input{algorithms/value_iteration}


This dissertation is ultimately about RL, rather than planning. Hence, VI will be treated a general purpose tool for planning in finite MDPs that can be called as a subprocedure by RL algorithms. In the context of model-based RL, such subprocedures are often called to produce a policy $\pi_{\hat{M}}^*$ that is optimal in the simulated MDP $\hat{M}$. Abstraction, as we will see, is particularly effective for accelerating planning, since planning with a well structured state-action space can be dramatically faster.

For further background on Markov Decision Process, see the text by~\citet{puterman2014markov}, and for more background on reinforcement learning, see texts by~\citet{kaelbling1996reinforcement,bertsekas1996neuro} and \citet{sutton1998reinforcement,sutton2018reinforcement}.

\section{State Abstraction}
\label{sec:c2_state_abstraction}

I next introduce the formalisms for state abstraction in RL, followed by a survey of prior research in the area. 

In an MDP, a state fully describes the current configuration of the environment down to the last detail. In a finite, discrete-time MDP, the default state representation is the set containing the states $s_1, s_2, \ldots, s_{|\mc{S}|}$. The actions, $\mc{A}$, change the state of the MDP according to the transition dynamics defined by $T$. However, this view on states is quite limiting, as the state representation lacks structure. For instance, we may instead suppose that two states can be deemed as similar or dissimilar to one another (standing near the bridge and on the bridge in the forest, for instance). It is precisely these similarities that underlie ontologies supporting \emph{objects}, \emph{properties}, \emph{relations}, and \emph{universals}. To facilitate this more general notion of \emph{state}, an MDP state is sometimes defined as a vector of variables or features, formalized as a Factored MDP~\cite{koller1999computing,guestrin2001max}, similar to the typical supervised and unsupervised learning settings. Other types of MDPs have been introduced that leverage some kind of implicit ontological structure, such as Relational MDPs~\cite{kersting2004bellman,gardiol2004envelope}, and Object-Oriented MDPs~\cite{diuk2008objects}, which explicitly carve the world into objects, their classes, and functions thereof.

Regardless of the mechanism for representing states, the goal of state abstraction is to reduce the size or complexity of the state space by grouping together similar states in a way that doesn't change anything important about the underlying problem being solved. Concretely, a state abstraction is defined as follows.
\ddef{State Abstraction}{A \textbf{state abstraction} is a function, $\phi : \mc{S} \ra \mc{S}_\phi$, that maps each true environmental state $s \in \mc{S}$ into an abstract state $s_\phi \in \mc{S}_\phi$.
\vspace{2mm}}

The abstract state corresponds to the agent's \emph{representation} of the current configuration of the environment; it is often not a perfect characterization in that the \emph{abstracted} state may throw away some information. In some cases, the underlying state space $\mc{S}$ may be continuous, and the abstracted space, discrete. For instance, we might let $\mc{S} = \mathbb{R}_{\geq 0}$, with $\mc{S}_\phi = \mathbb{N}$ induced by the abstraction function $\phi(s) = \lceil s\rceil$. The function $\phi$ may also just reduce a finite space $\mc{S}$ to a smaller one $\mc{S}_\phi$. In this sense, state representations that sharpen sensory observations into features or objects are carrying out a particular form of abstraction.

Determining what information to throw away is the central question behind the theory of state abstraction: how do effective agents come up with an appropriate abstract understanding of the environments they inhabit? I choose to systematize this question through the introduced state abstraction functions, sometimes called state \textit{aggregation} functions.

Why study such a broad question using such a specific and simple formalism? As with our choice of MDPs as the model of the environment, I take it to be important to attend to our question's simplest unanswered form. That way, any new results build a foundation upon which subsequent inquiry can take place. There are many other functions of interest that change the state representation, and it is important to understand each of them. These simple aggregation functions, $\phi$, are perhaps the simplest function that allow analysis and study, and for which new insights can bring clarity to the process of state abstraction in RL more generally. A natural direction for future work will investigate the more general classes of state abstraction that are expressive enough to include features, objects, and their kin.

\begin{figure}[t!]
    \centering
    \includegraphics[height=\figvdim]{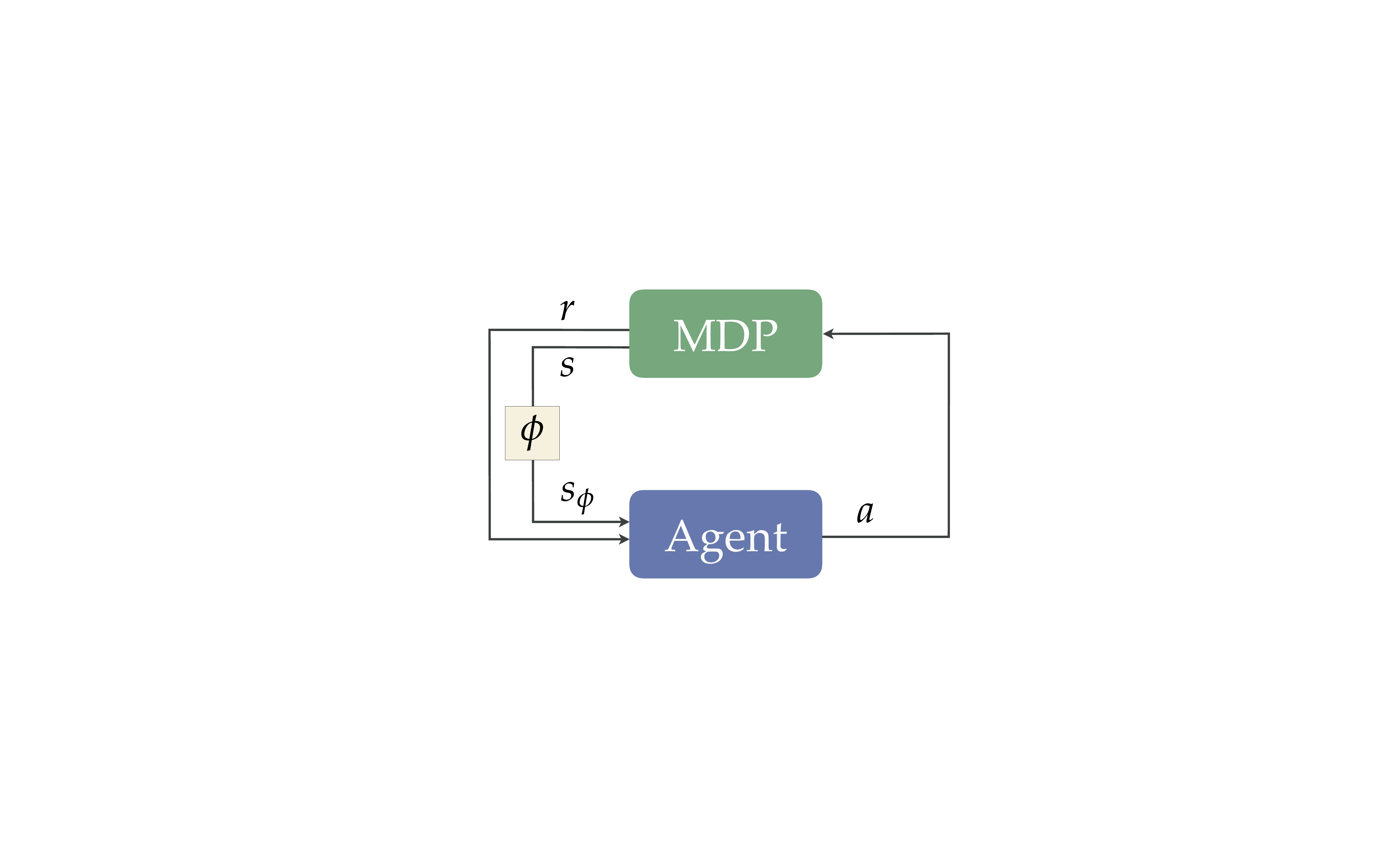}
    \caption{RL with a state abstraction.}
    \label{fig:c2_sa_rl}
\end{figure}

In general, an RL agent makes use of a state abstraction function as follows. Each time the MDP produces a state $s_t$, it is first passed through $\phi$ yielding the abstracted state $s_{\phi, t}$. Then, the agent takes $s_{\phi,t}$ as input, learns, and outputs an action $a_t$. This process is pictured in \autoref{fig:c2_sa_rl}. In this way, the agent never needs to know or confront the environmental state space. Moreover, this division between $\phi$ and the agent allows for the study of $\phi$ in a way that is agnostic to choice of RL algorithm.

It is also possible to define a new abstract MDP that is tightly connected to the original MDP. I refer to this new MDP as the \emph{abstract} MDP, $M_\phi$, drawing from the rich history of abstraction in MDPs \cite{knoblock1994automatically,dearden1997abstraction}. The abstract MDP is defined by three components: the original MDP $M$, a state abstraction $\phi$, and a \emph{weighting} function~\cite{ravindran2004approximate,van2006performance,li2006towards}, $w : \mc{S} \ra [0,1]$, such that:
\begin{equation}
    \forall_{s_{\phi} \in \mc{S}_\phi} : \sum_{s \in s_\phi} w(s) = 1,
\end{equation}
where I use $s \in s_\phi$ as shorthand for $s \in \{\bar{s} \in \mc{S} : \phi(\bar{s}) = s_\phi\}$.

Together, these three components induce an abstract reward and transition function as follows.
\ddef{Abstract Reward Function}{The \textbf{abstract reward function}, $R_\phi: \mc{S}_\phi \times \mathcal{A} \times \mc{S} \ra \left[\textsc{RMin}, \textsc{RMax}\right]$, is a weighted sum of the rewards of each of the ground states that map to the same abstract state:
\begin{equation}
R_\phi(s_\phi,a,s_\phi') = \sum_{s \in s_\phi} \sum_{s' \in s_\phi'} R(s,a,s') w(s).
\end{equation}}

\ddef{Abstract Transition Function}{The \textbf{abstract transition function}, $T_\phi: \mathcal{S}_\phi \times \mathcal{A} \ra \Delta(\mc{S}_\phi)$, is a weighted sum of the transitions of each of the ground states that map to the same abstract state:
\begin{equation}
T_\phi(s_\phi' \mid s_\phi,a) = \sum_{s \in s_\phi} \sum_{s' \in s_\phi'} T(s' \mid s,a) w(s).
\end{equation}}

With these two components in place, the abstract MDP is defined as follows.
\newpage
\ddef{Abstract MDP}{An \textbf{abstract MDP} is induced by the triple $(M, \phi, w)$, yielding: 
\begin{equation}
    M_\phi := (\mc{S}_\phi, \mc{A}, R_\phi, T_\phi, \gamma, \rhoz^\phi),
\end{equation}
where $\rhoz^\phi$ is the start state distribution $\rhoz$ projected into the abstract state space.
}

Of special interest is the best policy representable in this abstract state space. Here, ``best" is understood in terms of the \textit{environmental} value function, which defines the actual problem being solved. More formally,
\begin{equation}
    \pi_\phi^* = \argmax_{\pi_\phi \in \Pi_\phi} \bE_{s_0 \sim \rhoz}\left[V^{\pi_\phi}(\phi(s_0))\right].
\end{equation}
The policy $\pi_\phi$ is really a mapping from abstract states to actions, but can easily be turned into a policy over ground states and actions when paired with $\phi$. That is, for a given state $s$, the function composition $\pi_\phi(\phi(s))$ outputs an action.

This policy is particularly informative as it represents the best solution an RL algorithm reasoning with a state abstraction may try to discover. At the end of learning, we may ask about the \textit{value loss} incurred by the state abstraction, which corresponds to the gap in value between the true optimal policy and this optimal abstract policy. For instance, if we were interested in the policy that maximizes the expected ground value function under the start state distribution, it would be desirable to minimize the following quantity:
\begin{equation}
    \min_{\pi_\phi \in \Pi_\phi} \bE_{s_0 \sim \rhoz}\left[V^*(s_0) - V^{\pi_\phi}(s_0)\right].
\end{equation}
Much of the technical work of this dissertation focuses on ensuring that abstract policies that still achieve high value in the original problem still exist. While this property is not \textit{sufficient} to ensure an RL algorithm using $\phi$ can eventually learn good behavior, it is \textit{necessary}---if a good policy cannot even be represented, it certainly cannot be learned.

A few brief comments regarding notation and language are in order. Throughout the dissertation, I will use the term \emph{abstract state} or \emph{cluster} to refer to states in the abstract MDP, and \emph{ground} or \emph{environmental} state to refer to states in the original MDP $M$. I will occasionally abuse notation and allow $\pi_\phi(s)$ to act as shorthand for $\pi_\phi(s_\phi)$, and similarly $V^{\pi_\phi}(s)$ to abbreviate $V^{\pi_\phi}(\phi(s))$, and so on. Where it is needed for clarity, I include the use of $\phi$. Additionally, I let $\Pi_\phi$ denote the set containing all policies defined over the abstract state space $\mc{S}_\phi$ induced by a particular $\phi$. Finally, I use $\phiall$ to denote the space of all state abstraction functions.

\begin{figure}
    \centering
    \includegraphics[width=0.7\textwidth]{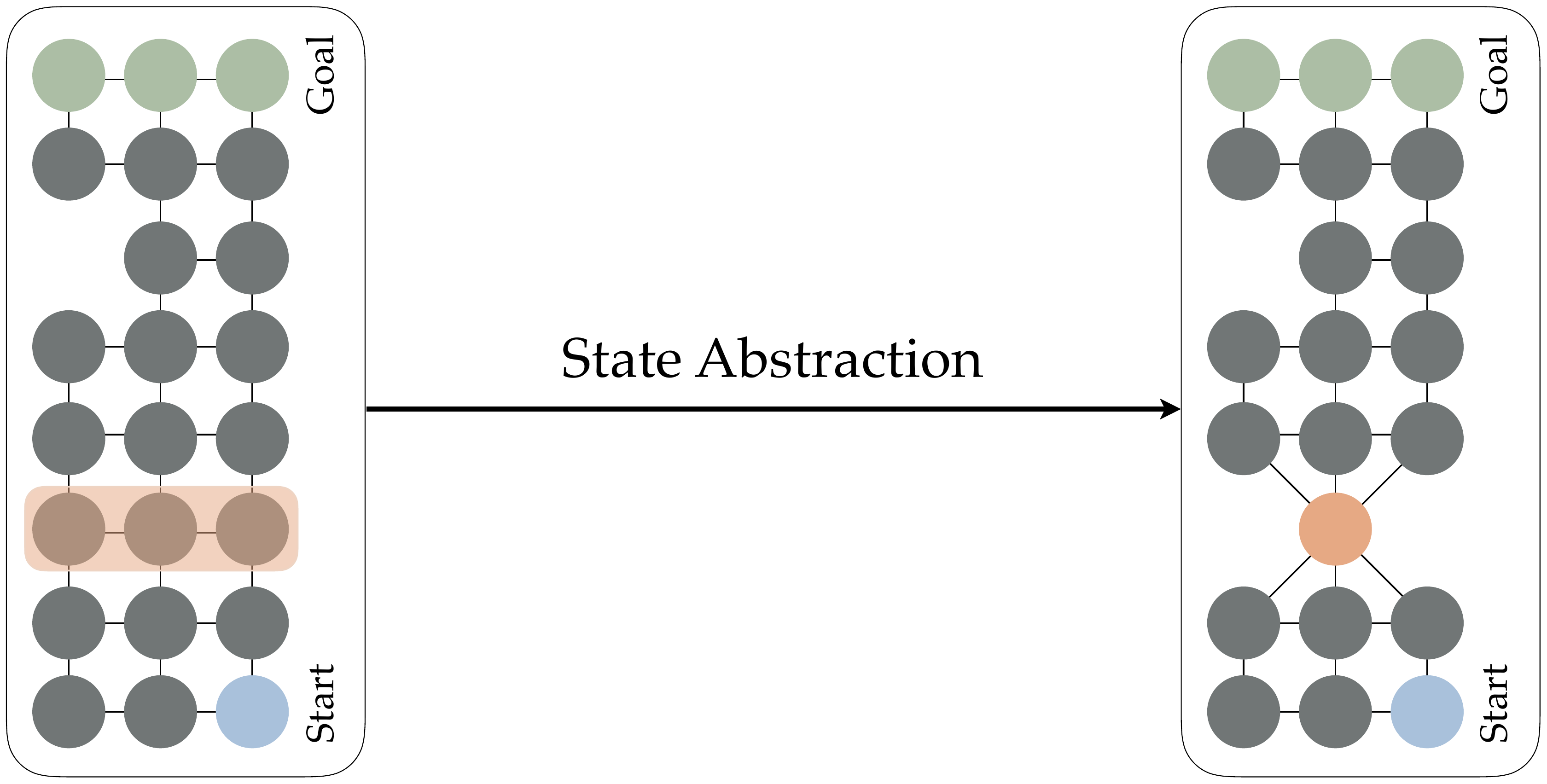}
    
    \caption{A simple grid world problem (left) and the abstracted problem induced by the state abstraction (right).}
    \label{fig:c2_state_abstr_ex}
\end{figure}

As a motivating example, let us suppose an agent is placed into a wide hallway with the goal of reaching the exit, which is placed at the far end of the hall. A traditional representation for this problem might yield a Cartesian grid: the agent has an \texttt{x} and a \texttt{y} coordinate, and the \texttt{up}, \texttt{down}, \texttt{left}, and \texttt{right} actions, and must navigate until its \texttt{y} coordinate is sufficiently large (and so has reached the exit of the hallway). This MDP and a corresponding abstract MDP, with a single row abstracted, is pictured in \autoref{fig:c2_state_abstr_ex}.

What is an effective state abstraction in this domain? Of course, there are many possible groupings. However, given the structure inherent in the problem, the agent's \texttt{x} coordinate is actually irrelevant for computing optimal behavior, or for computing the optimal value function $V^*$ and action-value function, $Q^*$, or (depending on the algorithm) for efficiently learning any of these quantities. Thus, consider the function $\phi$ that projects the ground state to an abstract state that only tracks the \texttt{y} coordinate. By this abstraction, all ground states with the same \texttt{y} coordinate belong to the same abstract state. In this way, the true state space of size $N \times M$ can be reduced to an abstract state space of size equivalent to the length of the hallway, $M$. Intuitively, imagine you are walking down a hallway without obstacles; if your mission was truly to make forward progress toward the exit of the hall, then there is little need to pay attention to horizontal movement.

A single application of this state abstraction is pictured in the right of \autoref{fig:c2_state_abstr_ex}. Again, this abstract state space throws away information. For instance, there is no longer a Markov transition model in the abstract space that will track perfectly with every trajectory in the ground MDP. When the agent moves into the clustered state (pictured on the right in orange), what happens when the agent executes the $\uparrow$ action? This is determined by choice of weighting function, $w$, that induces the abstract MDP. Hence, while there is no permissible Markov $T_\phi$ that can predict $T(s' \mid s, a=\uparrow)$, there surely exist Markov transition functions $T_\phi$ that can still support representation and discovery of good policies. Intuitively, such a state space will inevitably induce a \textit{wrong} but potentially \textit{useful} model, reminiscent of the classical adage of \citet{box1976science}.

To summarize, state abstraction is about translating the environmental state space into a new, more well behaved space. The central operation is \emph{aggregation}, which can apply to both continuous-to-discrete or large-to-small transformations. Different kinds of state abstractions are guaranteed to preserve certain properties, and are thus desirable for RL algorithms to discover and use. With our notation and concepts in place, I now turn to a survey of prior literature on state abstraction in RL.

\subsection{Prior Work on State Abstraction}
\label{sec:c2_state_abstr_prior_work}

The study of state abstraction in RL has a rich history, dating back to early work on approximating dynamic programs by~\citet{fox1973discretizing}, \citet{whitt1978approximations,whitt1979approximations2,axsater1983state}, and \citet{bertsekas1989}. Additionally, much of the literature has been heavily influenced by research on abstraction in planning \cite{sacerdoti1974planning,knoblock1994automatically,dean1997model}, and hierarchical RL~\cite{dayan1993feudal,kaelbling1993hierarchical,wiering1997hq,parr1998reinforcement,dietterich2000hierarchical,Hengst2002}. Indeed, the literature on state abstraction is vast. I here survey prior research that focuses on state abstraction in the context of RL in MDPs.

\paragraph{Bisimulation.}
The work of~\citet{fox1973discretizing} and \citet{whitt1978approximations,whitt1979approximations2} paved the way for understanding the value loss of state abstraction in MDPs. Fox and Whitt establish the first principles regarding state aggregation and its impact on value function representation and dynamic programming. Building on this work, \citet{dean1997model} developed an algorithm for finding states that resemble one another via the \textit{bisimulation} property \cite{larsen1991bisimulation}. Concretely, \citet{dean1997model} introduce the approximate bisimulation metric for partitioning an MDP's state space into clusters of states whose transition model and reward function are within $\eps \in \mathbb{R}_{\geq 0}$ of each other. In later work, \citet{givan1997bounded} use these ideas to develop an algorithm called Interval Value Iteration (IVI) that converges to the correct bounds on a family of abstract MDPs called Bounded MDPs, which summarize the space of possible MDPs the agent \emph{could} be in, given the agent's current knowledge of the MDP.

Since then, many approaches have adapted bisimulation in a variety of contexts related to state abstraction. \citet{ferns2004metrics,ferns2006methods} develop state similarity metrics for MDPs by bounding the value difference of ground states and abstract states for several bisimulation metrics that induce an abstract MDP. More recent work has since extended these bisimulation metrics to cooperate with action abstraction~\cite{castro2011automatic} and high-dimensional state spaces \cite{gelada2019deepmdp,castro2020scalable}. \citet{taylor2009bounding} further analyze the performance loss associated with using a bisimulation as part of an MDP homomorphism \cite{ravindran2003smdp}. In a similar vein, \citet{even2003approximate} study different distance metrics used in identifying state space partitions subject to $\eps$-similarity, providing value loss bounds for $\eps$-homogeneity subject to the $L_\infty$ norm. Even et al. also prove that the problem of finding the maximally compressing state abstraction is NP-hard, a result I return to in \autoref{chap:state_abstr_lifelong}. In more recent work, \citet{lehnert2018transfer,lehnert2019successor} combine ideas from bisimulation with the successor representation \cite{dayan1993improving} in the form of successor features \cite{barreto2017successor}. Lehnert and Littman present a novel combination of these two concepts in the form of the Linear Successor Feature Model (LSFM) and establish how LSFMs can underlie effective transfer, generalization, and model-based RL more generally.

\paragraph{Soft State Aggregation.} One of the earliest studies of state abstraction in RL was carried out by \citet{singh1995reinforcement}, who introduce \textit{soft state aggregation}. These soft forms of aggregation generalize the typical aggregation function to the class of stochastic functions $\tilde{\phi} : \mc{S} \ra \Delta(\mc{S}_\phi)$. Using $\tilde{\phi}$ to act, an RL agent samples $s_\phi \sim \tilde{\phi}(s)$, and learns functions based on abstract $s_\phi$, rather than $s$. Singh et al. present convergence guarantees for $Q$-learning with a fixed aggregation $\tilde{\phi}$, and a new heuristic method for adapting such an aggregation online during learning.

\paragraph{Model Selection.} A related and important body of work studies the problem of \textit{selecting} a state abstraction from a given class \cite{konidaris2009efficient,maillard2013optimal,odalric2013optimal,jiang2015abstraction,ortner2019regret}. \citet{ortner2014selecting} developed an algorithm for learning partitions in an online setting by taking advantage of the confidence bounds for $T$ and $R$ provided by UCRL~\cite{auer2007logarithmic}, a model-based RL algorithm that explores its environment efficiently. In earlier work, \citet{konidaris2009efficient} formulate the abstraction selection problem as a model selection problem, making use of the Bayesian Information Criterion \cite{schwarz1978estimating} to inform which abstraction to choose. This work focuses primarily on the selection of action abstractions, however, and will be discussed in more length in this later survey (\autoref{sec:action_abstr_survey}) Later, \citet{jiang2015abstraction} analyze the problem of choosing between two candidate abstractions for model-based RL. The core analysis again studies an algorithm that treats the choice of abstraction as a model selection problem, and analyzes the trade off between approximation error and estimation error produced by each abstraction. \citet{van2014efficient} study a similar problem in the context of Factored MDPs, in which the agent is explicitly given actions that move between candidate abstractions. \citet{diuk2009adaptive} study the closely related problem of \textit{feature selection} in MDPs, which has also received careful attention in prior work \cite{parr2008analysis,kroon2009automatic,snel2011fs}. Diuk et al. make use of the adaptive $k$-meteorologists problem to learn an appropriate set of features. Most recently, \citet{ortner2019regret} also study the problem of choosing an appropriate state abstraction from a given library during RL in MDPs without trap states. The key assumption is that at least one of the abstractions in the library induces an MDP---subject to this assumption, they develop an online algorithm that enjoys bounded regret relative to this best state abstraction in the library.

\paragraph{Learning What is Irrelevant.}
The early work of \citet{mccallum1995reinforcement} proposed the U-Tree algorithm that learns to represent state only in terms of its most relevant factors. The algorithm proceeds by carrying out statistical tests to determine which factors can be ignored, and which must be included in the tree. This notion of identifying irrelevant factors has been a key component of many abstraction methods, such as the work by \citet{jong2005state}. Here, Jong and Stone propose the property of policy irrelevance, which states roughly that states can be grouped together if they have the same optimal action. They then present statistical tests that can be deployed to determine which state variables may be safely ignored. \citet{menashe2018state} introduce an algorithm for abstracting continuous state, with the goal of inducing a small, tractable decision problem. They present Recursive Cluster-based Abstraction Synthesis Technique (RCAST), a technique for constructing a state abstraction that maps continuous states to discrete ones. Like the other tree-based methods discussed, RCAST uses $k$-d-trees to partition the state space. The key insight at the core of RCAST is to partition based on the ability to predict different \textit{factors} that characterize the state space.

\paragraph{Other Adaptive and Online Methods.} Several related approaches introduce algorithms for adaptively updating a state-space partitioning scheme, but do not base the clustering on notions of irrelevance. \citet{lee2004adaptive} use a form of adaptive vector quantization to repeatedly partition a continuous state space into a discrete one, thereby enabling classical RL algorithms like $Q$-learning to learn in a continuous space. To carry out this partitioning, Lee et al. make use of a form of vector quantization to rapidly cluster a continuous state space online in a computationally efficient manner using a Voronoi like tesselation of the state space. Then, the resulting state space can be used by $Q$-learning like algorithms (such as Temporal Difference or TD learning \cite{sutton1988learning}). \citet{krose1992adaptive} develop a similar approach in the context of controlling a robot to avoid collisions. Here, the state space is discretized using a Voronoi tesselation of the input state space, similar to \citet{lee2004adaptive}. \citet{nicol2012states} build on these approaches by extending the ideas of state-space quantization to the problem of state estimation in POMDPs, focused on applications in conservation biology. 
Lastly, \citet{cobo2011automatic,cobo2012automatic} study methods for finding state abstractions based on a \textit{demonstrator's} behavior---their method constructs abstract states that can be used to \textit{predict} what a demonstrator will do in each cluster. This idea will partially inform the study of \autoref{chap:rlit} as well. Similarly, \citet{akrour2018regularizing} proposes a method to simultaneously learn a clustering and learn behavior within each cluster. This coupled learning process is shown to be effective, as the existence of a sufficiently useful policy within each cluster is precisely the property needed to determine how to assign the clusters.

Most recently, \citet{du2019provably} and \citet{misra2019kinematic} study the process of learning a state abstraction online when learning in an observation-rich environment. Both approaches develop and analyze an algorithm for learning a state abstraction online assuming that the environment can be well modelled by a small, well-behaved state space, as modeled by a Block MDP introduced by \citet{du2019provably}. A Block MDP, roughly, is an MDP that is describable in terms of a small, well behaved state space, but is \textit{observed} through rich observations that are uniquely determined by their underlying latent state. Then, the algorithm of \citet{du2019provably} focuses on learning a mapping from this rich observation space to an estimate of this latent state space. The algorithm of \citet{misra2019kinematic} follows a similar approach, but searches for abstract state spaces in which all ground states share forward \textit{and} backward transitions are shared, called ``kinematic inseparability". Both algorithms enjoy guarantees on the sample complexity of RL while learning and exploiting these state abstractions.

\paragraph{A Unified Framework of State Abstraction in MDPs.}
\citet{li2006towards} presented a unifying framework for state abstraction in MDPs. They define five types of state abstractions that each ensure some property must hold between all the ground states in each abstract state. Formally, a state abstraction \emph{type} is defined with respect to a two-argument predicate on state pairs.

\ddef{State Abstraction Type}
{A \textbf{state abstraction type} is a collection of functions $\Phi_{p} \subseteq \phiall$ associated with a fixed predicate on state pairs, $p : \mc{S} \times \mc{S} \ra \{0,1\}$, such that when any $\phi_p \in \Phi_p$ clusters state pairs, the predicate must be true for that state pair:
\begin{equation}
\phi_p(s_1) = \phi_p(s_2) \implies p(s_1, s_2).
\end{equation}}

Several candidate types introduced in previous work are presented in \autoref{tab:c2_sa_table}, along with two notable properties I discuss in \autoref{chap:approx_state_abstr} and \autoref{chap:state_abstr_lifelong}. For simplicity, I abuse notation and let $\phi_p$ denote the type $\Phi_p$ for any predicate $p$. 

 \input{figures/tables/sa_table}


Li et al. analyze five state abstraction types, in many cases drawing from state abstractions introduced in prior work. These five classes are as follows.
\begin{enumerate}
    \item $\phi_{model}$: The ground reward function and the ground transition function into abstract states are the same,
    \begin{align}
          \phi_{model}(s_1) = \phi_{model}(s_2) \implies &\forall_{a \in \mc{A}} : R(s_1,a) = R(s_2,a),\\
          &\hspace{8mm} \text{and} \nonumber\\
          &\forall_{s_\phi \in \mc{S}, a \in \mc{A}} : \sum_{s' \in s_\phi} T(s' \mid s_1,a) = \sum_{s' \in s_\phi} T(s' \mid s_2,a).
    \end{align}
    \item $\phi_{Q^\pi}$: The $Q$ function under any policy is the same,
    \begin{equation}
          \phi_{Q^\pi}(s_1) = \phi_{Q^\pi}(s_2) \implies \forall_{a \in \mc{A}, \pi \in \Pi} : Q^\pi(s_1,a) = Q^\pi(s_2,a).
    \end{equation}
    \item $\phi_{Q^*}$: The $Q$ values for each action are the same,
    \begin{equation}
          \phi_{Q^*}(s_1) = \phi_{Q^*}(s_2) \implies \forall_{a \in \mc{A}} : Q^*(s_1,a) = Q^*(s_2,a).
    \end{equation}
    \item $\phi_{a^*}$: The optimal action is the same and the $Q$ value is the same for that action in each state.
      \begin{align}
          \phi_{a^*}(s_1) = \phi_{a^*}(s_2) \implies &\argmax_{x \in \mc{A}} Q^*(s_1,x) = \argmax_{y \in \mc{A}} Q^*(s_2,y),\\
          &\hspace{8mm} \text{and} \nonumber \\
          &|V^*(s_1) - V^*(s_2)| = 0.
    \end{align}
    \item $\phi_{\pi^*}$: The optimal policy chooses the same action in each state,
    \begin{equation}
    \phi_{\pi^*}(s_1) = \phi_{\pi^*}(s_2) \implies \pi^*(s_1) = \pi^*(s_2).
\end{equation}
\end{enumerate}

Li et al. further introduce the ordering operator, $\Phi_X \succeq \Phi_Y$, which states that any instance $\phi_X \in \Phi_X$ is also an instance of $\Phi_Y$. They prove the following ordering among the five introduced classes:
\begin{theorem}
(Theorem 2 from~\cite{li2006towards})
For any MDP, $\Phi_0 \succeq \Phi_{model} \succeq \Phi_{Q^\pi} \succeq \Phi_{Q^*} \succeq \Phi_{a^*} \succeq \Phi_{\pi^*}$.
\end{theorem}

That is, any element of $\Phi_{model}$ is also an element of $\Phi_{Q^\pi}$, and so on. This result is particularly useful for understanding planning performance in the abstracted models and for clarifying convergence conditions of different algorithms. They further prove that certain types of state abstraction preserve representation of optimal behavior.

\begin{theorem}
(Theorem 3 from~\cite{li2006towards})
Any state abstraction function of type $\Phi_{model}, \Phi_{Q^\pi}, \Phi_{Q^*}$ or $\Phi_{a^*}$ preserves the optimal policy. That is, if the abstraction $\phi$ is any of the above types, then the value loss is 0. Conversely, instances of $\Phi_{\pi^*}$ do not always preserve the optimal policy.
\end{theorem}

Subsequent study by~\citet{walsh2006transferring} investigates the power of state abstraction functions for supporting efficient transfer learning. That is, in the multitask RL setting, they seek to maximize the speedup ratio, which measures the reduction in time needed to find a good policy on the current MDP given that information may be transferred from a collection of source MDPs. Walsh et al. introduce the General Abstraction Transfer Algorithm (GATA), an algorithm for carrying out effective transfer of state abstractions across these MDPs by choosing a $\phi$ that maximizes this speedup ratio.

\paragraph{Representing State in Other Frameworks.}
Many previous works have proposed different kinds of state representation (beyond aggregation) for MDPs. \citet{dietterich2000hierarchical} introduced the MAXQ framework for decomposing value functions into a hierarchy represented as an acyclic graph over subtask policies. In subsequent work, \cite{dietterich2000state} further investigate the impact of state abstraction on learning under MAXQ. Dietterich highlights five conditions for safe state abstraction, under which a specific RL algorithm operating in the MAXQ framework is guaranteed to discovery a globally optimal policy. These five conditions primarily deal with picking up on the right notions of irrelevance, similar to those methods discussed above. \citet{andre2002state} also investigated a method for state abstraction in hierarchical reinforcement learning leveraging a programming language called ALISP that supports \emph{safe} state abstraction. Agents programmed using ALISP can ignore irrelevant parts of the state, achieving abstractions that maintain optimality.

\paragraph{State Representation and Value Function Approximation.}
A variety of other approaches to state representation learning have been proposed that parallel the objectives of state abstraction. For instance, \citet{whiteson2007adaptive} build an adaptive variant of the classical process of tile coding \cite{sutton1996generalization} for use in value function approximation. A related body of work studies the formation and discovery of appropriate basis functions for use in linear function approximation, typically applied to estimating $Q^*$. \citet{lagoudakis2003least}, \citet{mahadevan2006value,mahadevan2007proto}, and \citet{konidaris2011value} propose different kinds of basis functions (the polynomial, Laplacian, and Fourier, respectively) for use in value function approximation, each offering different desirable characteristics. Similarly, \citet{liang2016state} present empirical evidence that classical learning algorithms can achieve competitive performance to many deep RL algorithms. Their approach constructs features that are well suited to the structure of Atari games, including properties like relative object and color locations. The main result of the work shows that with a well crafted set of features, even a simple learning algorithm can achieve competitive scores in Atari games.

\paragraph{Value Preservation.}
Since the work of \citet{fox1973discretizing}, a long line of work has brought continued clarity the conditions under which state abstractions preserve value in MDPs. \citet{van2006performance} provide bounds on the suboptimality achieved by approximate VI in a well-behaved class of MDPs, in line with work on model minimization \cite{dean1997model,givan1997bounded,ravindran2002model,ravindran2003smdp,ravindran2004approximate}. Recent analysis extends these insights to non-Markovian environments~\cite{hutter2014extreme,hutter2016extreme,majeed2019performance}, invoking similar classes of state abstraction to those surveyed by \citet{li2006towards} but adapted to a more general class of environments. Again, the focus is on determining which classes of state abstraction are guaranteed to preserve representation of high value policies, a property that will be of central focus throughout the dissertation.

\citet{li2009unifying} also analyze a form of approximate state abstraction as applied to Delayed $Q$-learning (see Corollary 1 in Section 8.2.3). The main result here presents a modified sample complexity for Delayed $Q$-learning that only depends on the size of the abstract state space rather than the true state space. The key property to note, however, is that the analysis assumes Delayed $Q$-learning is interacting directly with an abstract MDP $M_\phi$, rather than interacting with $M$ and projecting states through $\phi$. This difference will return again in \autoref{chap:state_abstr_lifelong}, and particularly \autoref{corr:c4_m_phi_vs_learn_w_phi}. Additionally, \citet{lehnert2018value} explore the impact of horizon length on representation of value functions, with close ties to the diameter of an abstract state. They find that the value loss of any policy that optimizes with respect to an artificially-short horizon can achieve value similar to that of policies that take into account the full horizon, building on the results of \citet{jiang2015dependence}.

\paragraph{State Abstraction and Exploration.}
\citet{mandelefficient} focus on the exploration-exploitation dilemma in the context of state abstraction. In particular, they introduce a Bayesian method for clustering states to facilitate effective exploration while generalizing across the state space appropriately. The core contribution is the algorithm Thompson Clustering for Reinforcement Learning (TCRL), which addresses the large space of possible abstract state spaces by exploiting explicit structure present in the environment. As a result, TCRL narrows the search delicately to improve learning speed, enjoying Bayesian regret guarantees. A slight variant of TCRL achieves regret similarly to that of PSRL~\cite{osband2013more}. Separately, \citet{taiga2018approximate} study the exploration-exploitation dilemma from the perspective of approximate state abstraction. 

\citet{moore1994parti} introduced the Parti-Game algorithm, which uses a decision tree to dynamically partition a continuous state space based on the need for further exploration. That is, as data about the underlying environment is collected, state partitions are refined depending on a minimax score with respect to an adversary that prevents the learning algorithm from reaching the goal (and knows the current partitioning scheme). Parti-Game applies in tasks where 1) the transition function is deterministic, 2) the MDP is goal-based and the goal state is known, and 3) a local greedy controller is available. \citet{feng2004dynamic} also make use of a tree-based approach---this time, $k$-d-trees~\cite{friedman1977algorithm}---to dynamically partition a continuous MDP's state space into discrete regions. In contrast to Parti-Game, partitions are chosen based on value equivalence, thereby enabling a form of closure under the Bellman Equation.

\citet{chapman1991input} study tree-based partitioning as a means of generalizing knowledge in RL. Specifically, Chapman and Kaelbling propose the G algorithm, which constructs a data-dependent tree of $Q$-value partitions based on which $Q$ value can adequately summarize different regions of the state space. Over time, the tree will grow to sufficiently represent the needed distinctions in states. Further work uses decision trees of different forms to partition complex (and often continuous) state spaces into discrete models~\cite{uther1998tree}. \citet{asmuth2009bayesian} introduce the Best of Sampled Set (BOSS) algorithm, a Bayesian approach to exploration in RL that accommodates priors for clustering states. The algorithm itself resembles PSRL: maintain a posterior on models, sample from the posterior, and use the samples to inform decision making.

\paragraph{State Representation Learning.}
A separate but relevant body of literature investigates learning state \textit{representations} in the context of control and deep RL. For instance, \citet{jonschkowski2015learning} proposed learning state representations through a set of well chosen ontological priors, catered toward robotics tasks, including a simplicity prior and a causality prior (among others). These priors are then encoded into an optimization problem that seeks to jointly optimize over each of their prescribed properties. Similarly, \citet{karl2016deep} developed a variational Bayes method for learning a latent state-space representation of a Markov model, given high dimensional observations. Critically, this state space is of a simple Markov model, and does not involve decision making or rewards, which are critical aspects of learning state representations in MDPs~\cite{oh2017value}. For a full survey of recent state representation schemes for deep RL, see text by \citet{lesort2018state}, or the recent survey by \citet{bertsekas2018feature}.

\paragraph{State Abstraction and Planning} As a final note, state abstraction has also been applied extensively in the context of planning. Given the breadth of planning as a field, I again highlight several methods that are tightly connected to MDPs and RL. \citet{Hostetler2014} apply state abstraction to Monte Carlo Tree Search \cite{kocsis2006bandit,coulom2006efficient,james2017analysis} and expectimax search, giving value bounds of applying the optimal abstract action in the ground tree(s). Other work develops similar methods for incorporating state abstraction into Monte Carlo style planning algorithms~\cite{anand2015asap,anand2016oga,jiang2014improving}. \citet{dearden1997abstraction} also examine state abstraction for planning, focusing on abstractions that are quickly computed and offer bounded value. The primary analysis is on abstractions that remove negligible literals from the planning domain description, yielding value bounds and a mechanism for incrementally improving abstract solutions to planning problems.

As is hopefully apparent, the literature on state abstraction in RL is exceptionally both broad and deep. I have chosen to exclude those approaches that are better considered as joint state-action abstraction, as they will be discussed in \autoref{sec:c2_state_action_abstr}.

I now turn to a formal introduction of action abstraction.

\section{Action Abstraction}
\label{sec:c2_action_abstr}

Action abstractions describes methods that empower the \emph{action space} of an RL agent. With a well structured action space, decision making agents can probe more deeply in search, plan efficiently by focusing on progress toward a subgoal, or prune away irrelevant primitive action sequences based on knowledge of action-optimality correlations.
Of course, as with structuring state, there are many possible operations available to organize the action space. We might add new long horizon sequences of actions, prune away actions, or add actions that specifically try to satisfy some property or reach a subgoal with high probability. I here concentrate on what has become the most canonical formalism for action abstraction, the options framework introduced by~\citet{sutton1999between}. 


\begin{figure}[t!]
    \centering
    \subfloat[Four Rooms MDP ]{\includegraphics[width=0.4\textwidth]{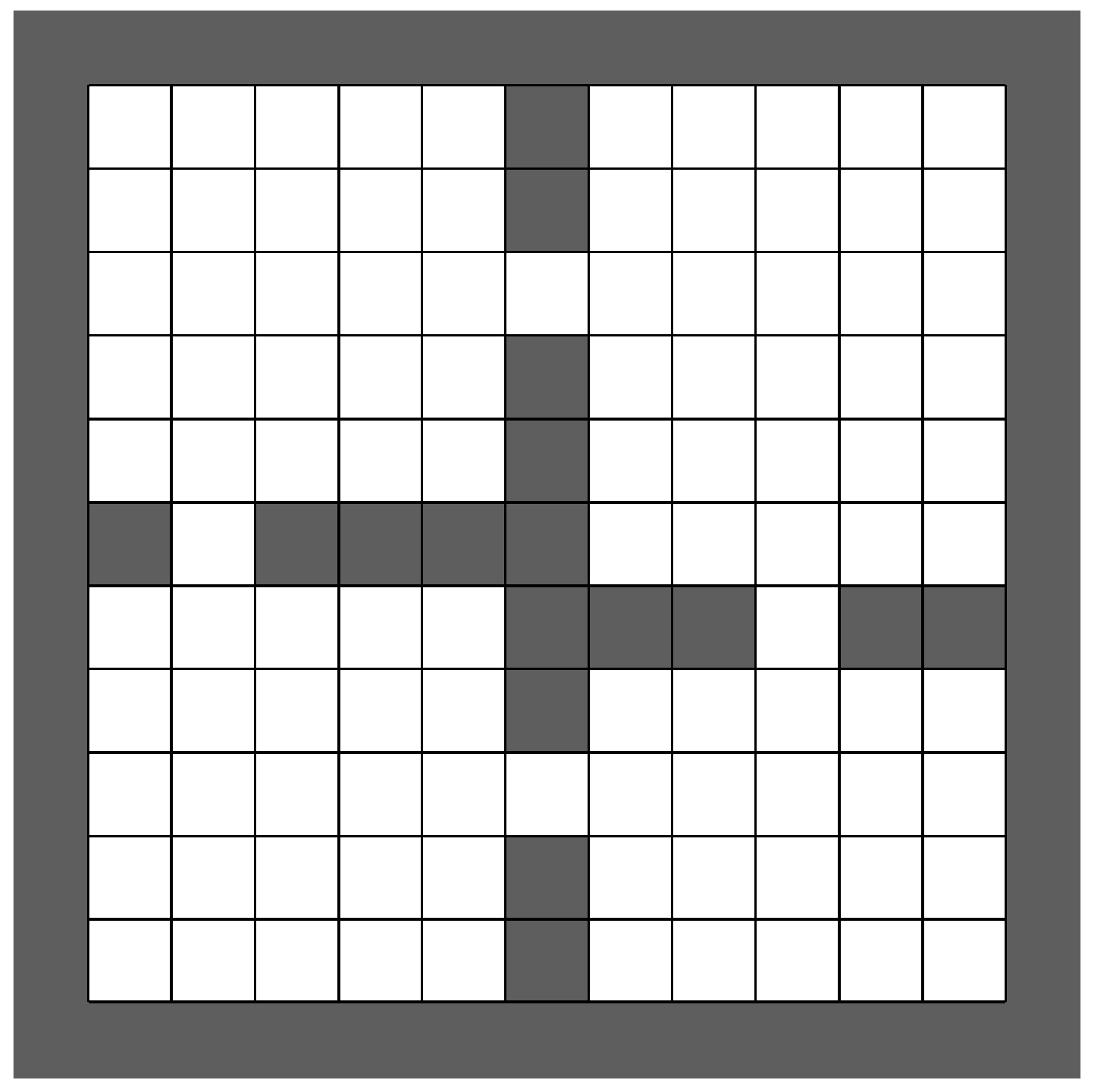}} \subfhspace
    \subfloat[Four Rooms MDP with Options]{\includegraphics[width=0.4\textwidth]{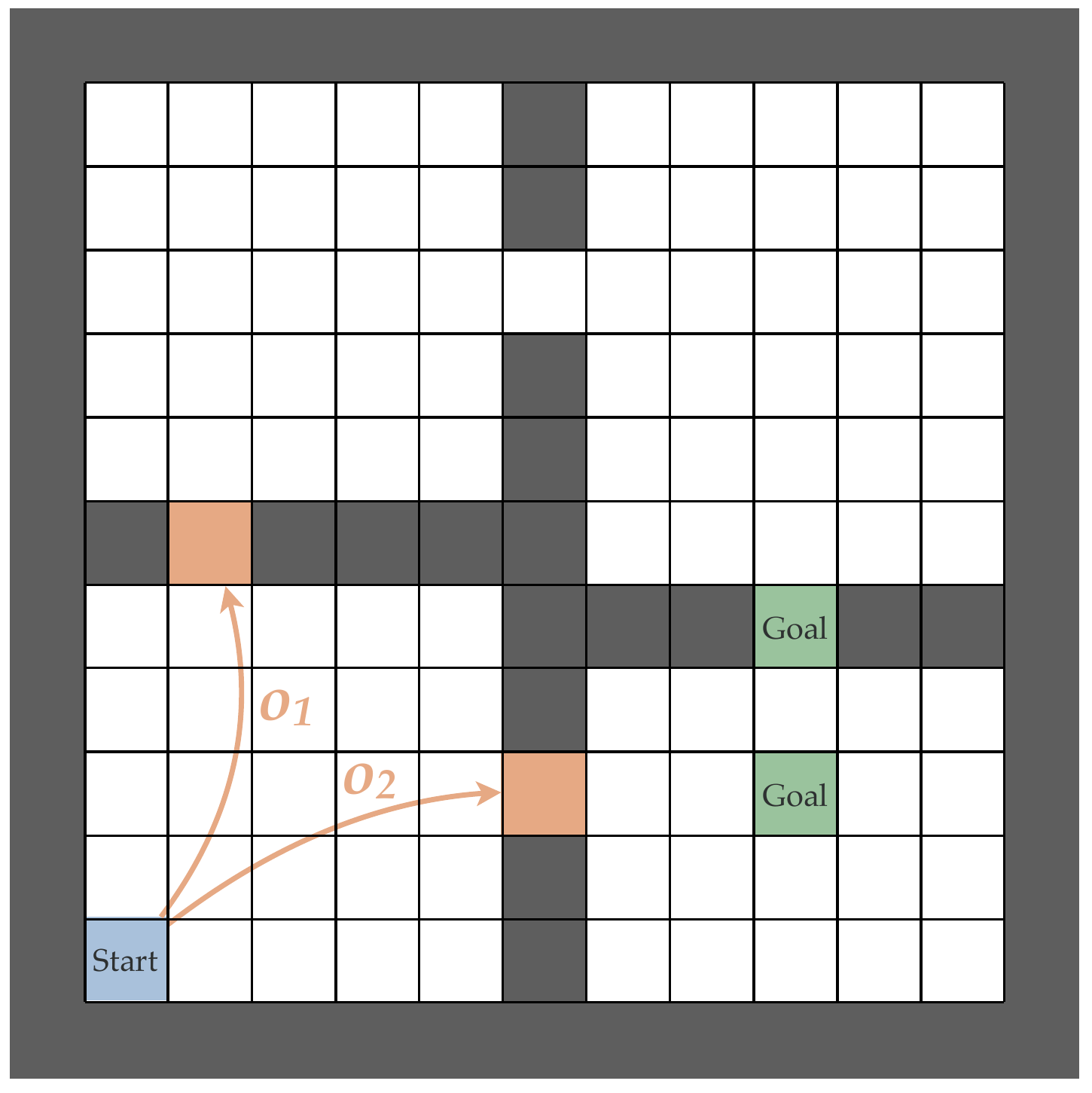}}
    \caption{The classical Four Rooms domain (left) extended by options (right).}
    \label{fig:c2_four_rooms}
\end{figure}

To introduce options, let us consider the example domain pictured in \autoref{fig:c2_four_rooms}. This problem is known as Four Rooms \cite{sutton1999between} and will be used as a canonical grid world problem for simple experiments and visuals throughout this dissertation. As with the original Russell \& Norvig grid world, the agent may move up, down, left, and right, with the goal of getting from a particular cell in the grid to another. In this case there are long walls that change the structure of the problem. The flow of movement throughout the environment is thought to be suggestive of certain types of action abstractions, such as those that take the agent to the doorways between rooms, or that transition between the rooms directly. These kinds of additional high level behaviors may be naturally expressed in terms of options. At a high level, an option is one prescription for an abstract behavior---in the language of our forest example from \autoref{chap:introduction}, options will constitute the behaviors at the level of ``move to the waterfall" or ``navigate back to camp", rather than ``rotate head".

More formally, an option is defined as follows.
\ddef{Option~\cite{sutton1999between}}{An \textbf{option} is a triple $o = (I_o, \beta_o, \pi_o)$, where:
\begin{itemize}
    \item $\mc{I}_o \subseteq \mc{S}$ is a subset of states denoting in which states the option is available to be executed,
    \item $\beta_o : \mc{S} \ra \Delta(\{0,1\})$, assigns a Bernoulli random variable to each state denoting the probability that the option terminates upon arriving in that state,
    \item $\pi_o : \mc{S} \ra \Delta(\mc{A})$ is a behavioral policy.
\end{itemize}}

Options denote abstract actions; the three components indicate where the option $o$ can be executed ($\mc{I}_o$), where it terminates ($\beta_o$), and what to do in between these two conditions ($\pi_o$). Options are known to aid in transfer~\cite{konidaris2007building,konidaris2009skill,brunskill2014pac,topin2015portable}, encourage better exploration~\cite{mcgovern1998acquire,csimcsek2004using,csimcsek2009skill,brunskill2014pac,bacon2017option,fruit2017exploration,machado2017laplacian,tiwari2019natural,jinnai2019opt_explore}, and make planning more efficient~\cite{mann2014scaling,mann2015approximate,jinnai2019opt}.

Action abstraction has historically been treated as a generic class of operations that change the action space of an agent. Throughout this thesis, I will formalize the process of action abstraction as a replacement of the primitive actions of an agent with some set of options, $\mc{O}$. I again take this class of operations to be of sufficient generality so as to characterize the important questions about abstraction, but not to be so general so as to limit analysis, understanding, and progress. Concretely, I define action abstraction as follows.

\ddef{Action Abstraction}{An \textbf{action abstraction} is a function $\omega : \mc{A} \mapsto \mc{O}$ that replaces the primitive actions $\mc{A}$ with a set of options $\mc{O}$.\vspace{2mm}}

An RL algorithm paired with an action abstraction chooses from among the available options, denoted $\Omega(s)$, at each time step. That is,
\begin{equation}
    \Omega(s) := \{o \in \mc{O} : s \in \mc{I}_o\}.
\end{equation}
Then, the agent runs the option until it terminates in some state $s'$ according to $\beta_o(s)$. Finally, the algorithm again chooses its next option from among the set $\Omega(s')$ and repeats this process indefinitely. 

With $\mc{O}$ \emph{replacing} the primitive action space, it is not necessarily the case that every policy over $\mc{S}$ and $\mc{A}$ may be represented. That is, the action abstraction may destroy an agent's ability to ever discover a near-optimal policy. Note, however, that the formalism is expressive enough to describe the case where the primitive actions are redefined in terms of options. For instance, the action $a_1$ can be translated into an option by constructing the option that initiates in every state, terminates with probability 1 in every state, and executes the policy $\pi_1 : \mc{S} \ra \{a_1\}$, effectively encodes $a_1$ as an option. However, by including options \emph{and} primitive actions, learning algorithms face a larger branching factor, and must search the full space of policies which can hurt learning performance~\cite{jong2008utility}. Hence, it is often prudent to restrict the action space \emph{only} to a set of options to avoid blowing up the search space.

\begin{figure}[t!]
    \centering
    \includegraphics[height=\figvdim]{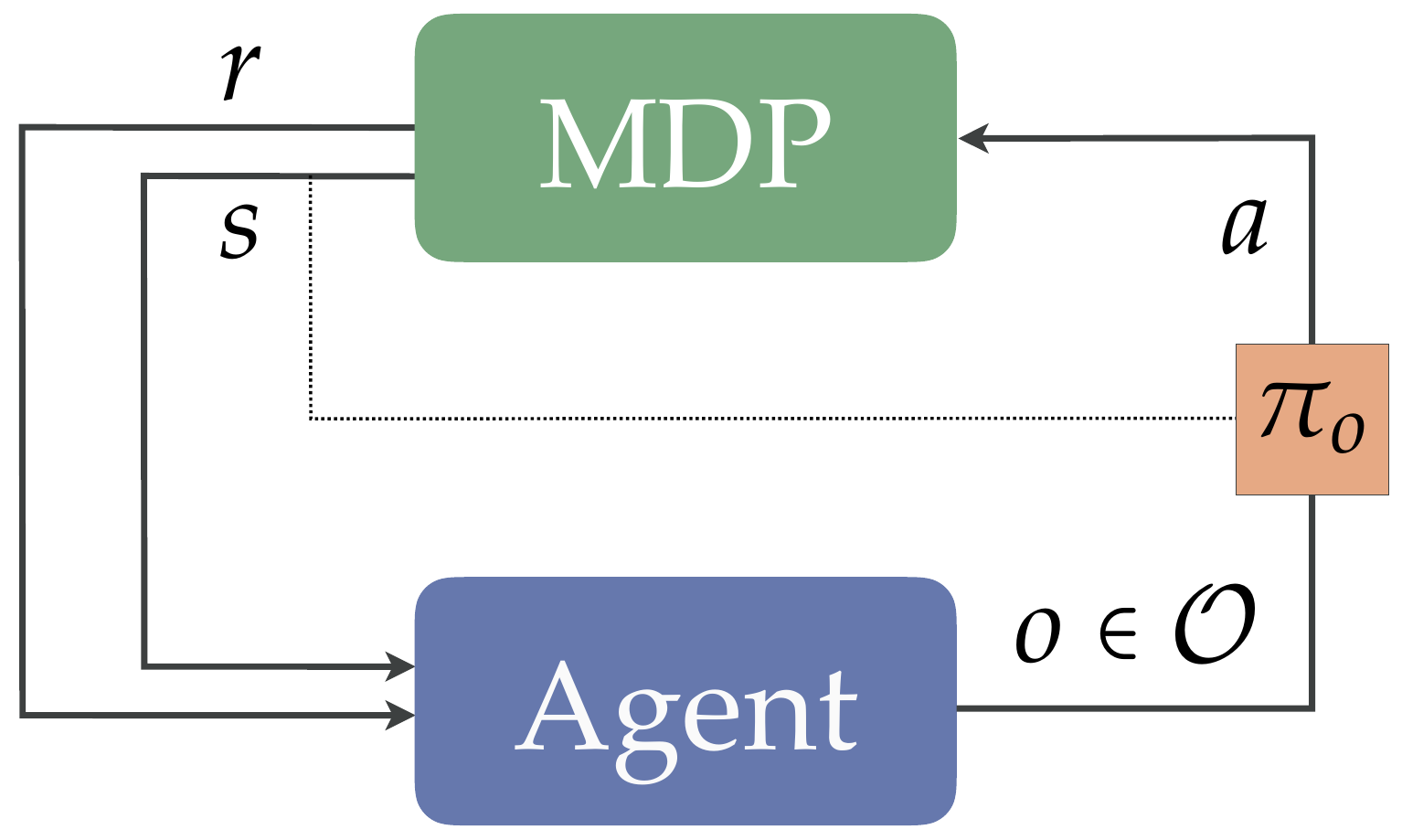}
    \caption{RL with action abstraction.}
    \label{fig:c2_aa_rl}
\end{figure}

Then, when the agent chooses from among available options, the agent commits to executing the policy associated with the option until the sampled terminating condition is true for a state the agent arrives in. For example, if the termination condition assigns zero probability to all states except $s_4$ (in which $\beta(s_4) = 1$), then the agent will execute the option's policy indefinitely until arriving in $s_4$. When the agent reaches $s_4$, the agent will stop executing the option policy, and will make its next choice of action or option. So, options facilitate action pruning of a certain form; when an option is selected, every state the agent arrives in up until termination, the actions not chosen by the option policy are effectively pruned. The resulting decision making problem slightly loses out on the Markov property, too, as any state encountered while executing the option will induce a different policy according to which option is currently being run. This process is pictured in \autoref{fig:c2_aa_rl}. As with state abstraction, I will occasionally use $\Oall$ to denote the set of all options to simplify notation.

Since the primary objects of interest in an action abstraction are the options introduced, $\mc{O}$, I will largely talk about action abstraction in terms of which options are added.

Options also give rise to new transition and reward functions based on where the options will terminate and the trajectory taken by the option's policy. The model was originally proposed for options by~\citet{precup1997multi,precup1998multi}, and is defined as follows.
\ddef{Multi-Time Model}{For a given $\gamma$ and option $o$, the \textbf{multi-time model} (MTM) defines the transition and reward function by:
\begin{align}
    T_\gamma(s' \mid s, o) &:= \sum_{k=0}^\infty \gamma^k \beta_o(s') \PR(S_k' = s' \mid S_0 = s, O = o) = \sum_{k=0}^\infty \gamma^k \beta_o(s') p(s', k \mid s, o),
    \label{eq:mtm} \\
    R_{\gamma}(s,o) &:= \underset{k, s_{1 \ldots k}}{\bE}\left[r_1 + \gamma r_2 \ldots + \gamma^{k-1} r_k \bigmid s, o\right].\label{eq:mtm_reward}
\end{align}}

This model allows for straightforward application of many standard RL and optimization algorithms to settings that incorporate options. In \autoref{chap:options_for_planning}, I will study how these models can be exploited to make planning faster, and in \autoref{chap:elm_options}, I motivate a simpler alternative to these models. As with state abstraction, it is natural to define an abstracted form of the environmental MDP in terms of the available options (and their models) at a given time step. Without further modification, the induced model is in fact a \textit{semi-}Markovian decision process, as the situation in which the agent enters state $s$ actually differs depending on whether the agent is executing some option $o$, or not \cite{sutton1999between}. In \autoref{chap:vpsa}, I will avoid the semi-Markovian nature of options by studying a restricted subclass of options that can only describe policies that are Markov in the original MDP.

In the Four Rooms MDP, let us consider two options that terminate in hallways, pictured in the right of \autoref{fig:c2_four_rooms}. These two options are defined as follows.
\begin{align*}
    o_1 &= \begin{aligned}
        \mc{I}(s)&\hspace{10mm} s = (1,1), \\
        \beta(s)&\hspace{10mm}
        \indic\{s = (3,6)\ \text{or}\ s = (6, 3)\}, \\
        \pi(s)&\hspace{10mm} \argmax_{a\in \mc{A}} \left(\indic\{s = (3,6)\ \text{or}\ s = (6, 3)\} + \gamma \sum_{s' \in \mc{S}} T(s' \mid s,a) V_{\indic\{s = (3,6)\ \text{or}\ s = (6, 3)\}}^*(s')\right),
    \end{aligned}
    \end{align*}
    and,
\begin{align*}
    o_2 &= \begin{aligned} 
        \mc{I}(s)&\hspace{10mm} s = (1,1), \\
        \beta(s)&\hspace{10mm}
        \indic\{s = (3,6)\ \text{or}\ s = (6, 9)\}, \\
        \pi(s)&\hspace{10mm} \argmax_{a\in \mc{A}} \left(\indic\{s = (3,6)\ \text{or}\ s = (6, 9)\} + \gamma \sum_{s' \in \mc{S}} T(s' \mid s, a) V_{\indic\{s = (3,6)\ \text{or}\ s = (6, 9)\}}^*(s')\right).
    \end{aligned}
\end{align*}

The first option $o_1$ will initiate in the bottom left corner, and will only terminate when the agent arrives in one of the two hallways leaving the lower left room, and executes the policy that moves the agent to one of the hallways as quickly as possible. So, the option might induce the following true trajectory:
\begin{equation}
    (1,1), \ra, (2,1), \uparrow, (2,2), \ra, (3,2), \ra, (4,2), \ra, (5,2), \uparrow, (5,3), \ra, (6,3).
\end{equation}

When the agent enters a new state while executing either option it draws a sample from the Bernoulli distribution $\beta_o(s)$. If the sample is 1, then the agent \emph{stops} following the option policy, and is again in a position to choose among its active options (and perhaps primitive actions). In the trajectory above, we define the option $o_1$ such that $\beta_o((6,3)) = 1$, and so the agent will stop executing the option at $(6,3)$. At this point, the agent resumes its regular action selection process.


With the basic notation and formalisms for action abstraction established, I next survey previous research in the area.

\subsection{Prior Work on Action Abstraction}
\label{sec:action_abstr_survey}

As with state abstraction, research on action abstraction in RL has a long and deep history. Early work concentrated on incorporating macro-operators \cite{korf1983learning,korf1985macro} that characterized relevant sequences of actions to accelerate both planning and RL~\cite{iba1989heuristic,singh1992scaling,mcgovern1997roles,mcgovern2001automatic,menache2002q,stolle2002learning}, often expressed through hierarchical structures \cite{dayan1993feudal,kaelbling1993hierarchical,hauskrecht1998hierarchical}. In the planning literature, it has long been understood that behavioral abstractions in the form of hierarchies can greatly accelerate planning speed, as in Hierarchical Task Networks \cite{currie1991plan,erol1994htn}. For these reasons, the primary focus of work on action abstraction in RL has been to unlock this same degree of highly efficient decision making.

In the late 1990s and early 2000s, three formalisms emerged for capturing similar notions of abstract behavior. \citet{parr1998hierarchical} proposed Hierarchies of Abstract Machines (HAMs) to specify abstract behaviors in terms of partially specified \textit{programs}. Around the same time, \citet{dietterich2000hierarchical} proposed the MAXQ framework for hierarchical RL discussed briefly in \autoref{sec:c2_state_abstr_prior_work}. Finally, \citet{sutton1999between} developed the options framework that much of this section has been focused on. Each of these three methods systematizes the study of behavioral abstractions in a slightly different way. Given the focus of this dissertation on options (for reasons discussed earlier in the chapter), I concentrate this survey on prior research on the options framework.

The initial options work grew out of the dissertation by \citet{precup2001temporal}. Since then, options have explored for their capacity to address many aspects of the RL problem, from aiding in transfer, to representation learning, to off-policy evaluation, to planning. This survey decomposes the use of options into these different desired effects, though in many cases options are intended to aid in more than one of these processes.

\paragraph{Option Discovery.} In contrast to planning, incorporating options into RL algorithms typically requires that options are learned online. It has thus long stood as an open question as to what precisely constitutes a good option for RL, and more generally how to learn good options through interaction alone. The option discovery literature studies variations of this problem. The emphasis has tended to be on forming objectives that give rise to algorithms that learn options during RL, though naturally the literature is diverse.

One of the most popular strategies for option discovery is based around the discovery of subgoals as a mechanism for inspiring options---once a set of subgoals (or often abstract states) is fixed, options may be defined that move the agent to the subgoal as quickly as possible. \citet{mcgovern2001automatic} propose one of the earliest option discovery methods based on estimating useful subgoals from a series of past trajectories. The main idea is to those states that always appear in successful trajectories, and not on any unsuccessful trajectories---if the number of trajectories is sufficiently large, then intuitively these states are likely to be useful subgoals. These subgoals are then used to inform options, which are shown to accelerate RL in benchmark tasks. \citet{digney1998learning} and \citet{stolle2002learning} introduce similar approaches that determine subgoals based on which states have high visitation count on past successful runs. By similar reasoning, \citet{menache2002q} propose the $Q$-cut algorithm for discovering useful subgoals in RL. Here, the subgoals are identified by estimating the MDP's transition graph and solving a min-cut problem online to identify bottleneck states. \citet{mannor2004dynamic} later developed an algorithm inspired by similar principles; the algorithm maintains an estimate of the MDP's transition graph and applies a form of state abstraction to determine clusters of states that have similar value functions. Using these clusters, options are then naturally defined as those policies that transition between the clusters, thus moving the RL agent between relevant subregions of the MDP. In a similar vein, \citet{castro2011automatic} construct an algorithm for discovering options that connects disparate states, with state-distance defined according to the bisimulation metrics discussed earlier in the chapter. In all of these works, options are empirically shown to accelerate RL on collections of benchmark tasks. Also along these lines, \citet{provost2006developing} learn discrete abstract state features in a continuous state environment. Then, options are constructed that explicitly target nearby feature changes, thereby again moving the agent through the state space. A separate but closely related pair of approaches by \citet{thrun1995finding} and \citet{pickett2002policyblocks} searches for high value policies in related tasks that share decisions in select states. These policies are then merged to form a general purpose option that is likely to accelerate learning across the different tasks.

Later, \citet{csimcsek2004using} build on the above approaches by explicitly studying the property of movement throughout the MDP, perhaps most closely related to the work of \citet{mannor2004dynamic}. Here, {\c{S}}im{\c{s}}ek and Barto use the property of \textit{relative novelty} to identify those states that are hard for the RL agent to reach from its current region of the state space. These hard to reach states (referred to as ``access states") closely resemble many of the bottleneck or subgoal discovery methods discussed above, but are unique in that they are defined so as to allow the agent to reach a new portion of the state space. The relative novelty of a state $s_t$ is said to be the ratio of novelty between the states before $s_{t-k}, \ldots, s_{t-1}$ and after $s_{t+1}, \ldots$. Hence, identifying novel subgoals is reduced to a classification problem in which subgoals are learned that have relative novelty above a particular well chosen threshold. This classification problem is solved in both a batch offline and online setting. Using the estimated subgoals, options are generated that allow RL agents to move more fluidly throughout the MDP. As with many of the approaches surveyed thus far, experiments are conducted contrasting learning with and without the learned options on grid world variations and the Taxi task introduced by \citet{dietterich2000hierarchical}. This was shortly followed by \citet{csimcsek2005identifying}, which blends the notion of access states with the min-cut method by \citet{menache2002q} but focuses on local rather than global cuts.

In follow up work, \citet{csimcsek2009skill} analyzes the graph theoretic property of \textit{betweenness centrality} \cite{freeman1977set}, and argues for its application in formalizing what it means to be a good option. The betwenness centrality assigns a real number to each vertex of a graph measuring roughly how easy it is to reach all other vertices from that vertex. Subgoal states, then, are defined as those states with high betweenness centrality in their local region of the state space. That is, if a state $s$ has considerably higher betweenness then all states reachable in some chosen $n \in \mathbb{N}$ steps from $s$, it is a good candidate for a subgoal. Options are then defined based around this subgoal as in prior work.

\citet{konidaris2009skill} present one of the first algorithms for option discovery in goal-based MDPs with continuous state based on the idea of \textit{chaining}. Concretely, the algorithm focuses on identifying options whose termination conditions always lie inside of at least one other option's initiation condition, thereby ensuring sequential application of options. The key idea is that the chain of options discovered ensure that the goal is eventually contained at the end of a chain, thus allowing an execution of the options to lead to the goal. More generally, the initiation and termination conditions of the options can be framed around arbitrary target events, such as those subgoals or bottlenecks discussed in prior work, rather than just the goal state. The chained options are shown to dramatically improve learning on the challenging continuous state Pinball domain, both when the ideal options are given up front and when the chained options are learned online. \citet{bagaria2020option} recently extend these ideas to coordinate with deep neural networks to great effect in environments with rich observations.

As discussed briefly earlier in the chapter, \citet{konidaris2009efficient} frame the problem of option discovery as one of model selection---given a collection of experiences collected during exploration, the objective is to determine the best option from a predetermined library of candidate options. In this work, Konidaris and Barto in fact study the joint process of selecting a state and action abstraction. The criteria used to determine the best abstraction is that of the Bayesian Information Criterion \cite{schwarz1978estimating}, which offers an elegant principle for determining the simplest, but most explanatory abstraction given the data. Moreover, it allows for the incorporation of prior knowledge or preference about the abstractions through the prior, as is standard in Bayesian methods. In experiments, the selected abstractions are shown to accelerate RL performance relative to learning without abstraction, providing strong evidence for the effectiveness of the approach. In follow up work, \citet{konidaris2010constructing} study the process of learning options given access to data generated by a demonstrator. Here, Konidaris et al. present CST, an algorithm for learning how to segment existing trajectories into options, again using the idea of target events from previous work. That is, given a list of target events, the agent will repeatedly try to formulate options for realizing these events when they occur. Given access to a trajectory generated by a demonstrator, the problem is to identify relevant change points throughout the trajectory that should be broken into target events, and hence, options. These ideas were later extended to the application of autonomous option discovery and use on a mobile-manipulator robot, which learned to pull levers to open doors and navigate through rooms \cite{konidaris2011autonomous}. For more on these approaches, see the dissertation by \citet{konidaris2011thesis}.

More recently, \citet{machado2017laplacian,machado2018eigenoption} develop a suite of algorithms for discovering options based around the graph Laplacian of the MDP's transition graph. The resulting options, called ``eigenoptions", are behaviors defined by Proto-value functions (PVFs), a spectral approach to representation learning in RL \cite{mahadevan2007proto}. The PVF captures, roughly, a reward-agnostic representation of the diffusion structure of transitions in an MDP. By similar reasoning to earlier work on subgoal discovery, the PVF may then be used to identify those regions of the state space that are distant in transition space. Machado et al. introduce the ``eigenpurpose", a mechanism for defining an intrinsic reward function that increases as the agent moves toward disparate regions of the state space, as defined by the PVF. Eigenoptions, then, are those options whose policies are optimal with respect to this intrinsic reward function. In subsequent work, \citet{machado2018eigenoption} extend these ideas to richer settings, allowing for the discovery of eigenoptions in stochastic MDPs and MDPs with high-dimensional state input. In a similar vein, \citet{eysenbach2019diversity} propose learning options such that the diversity of the trajectories produced by the set of options is maximized, thereby generating options that may explore infrequently visited states. Option discovery has also been studied in select other settings, including in inverse RL \cite{ng2000algorithms} by \citet{ranchod2015nonparametric}, and in active learning \cite{cohn1996active} by \citet{da2014active}.

\paragraph{Understanding the Impact of Options.} Alongside the option discovery problem, it has also long been of interest to characterize how options impact the RL problem. As much of the aforementioned work shows, when the right options are used, RL algorithms can be empowered in a dramatic way. \citet{jong2008utility} address this question by taking a close look at the impact options have on RL. Much of the focus is on inspecting the empirical effect different options have on RL algorithms. In particular, Jong et al. set out to clarify when and why options can help learning. One experiment conducted contrasts the learning performance of traditional $Q$-learning with two variants: 1) $Q$-learning with subgoal-based options, and 2) $Q$-learning with experience replay \cite{lin1992self}, a mechanism for improving an agent's capacity to assign credit across long time horizons. The results suggest that both variant (1) and (2) perform nearly identically, suggesting that options are serving the same role that experience replay can. In other experiments, results suggest that options can negatively affect learning performance, or have no change at all. In particular, when options are paired with the principle of optimism under uncertainty for exploration, learning time is increased.  The conclusion from this study is that it is not always straightforward that intuitively useful options will have the desired effect, and that sometimes they can even \textit{negatively} impact RL.

\paragraph{Options and Transfer Learning.} Similarly to state abstractions, options have long been studied as a mechanism for facilitating transfer across tasks. A collection of options can summarize many things about previous experience, including good default policies for exploration, areas of an environment to pursue or avoid, or which actions should appear in sequence. Early on, \citet{konidaris2007building} developed a method for building options that are transferable between similar tasks. A core practice of the method is to separate the problem representation into two types, a global problem-space representation, and a local \textit{agent-space} representation \cite{konidaris2006autonomous} that captures an agent-relative perspective on the environment's features. Each MDP the agent inhabits is characterized entirely by its problem-space representation, but the agent adopts the agent-space representation for use in transferring options across relevant situations. For instance, in facing several MDPs with keys and doors, a sensible agent-space option is one that collects the nearest agent-space key and takes it to the nearest door. In contrast, the problem-space representation defines these constituents by their absolute coordinates, and thus prohibits this same form of seamless transfer. Konidaris and Barto make use of given and learned options for accelerating learning in these key-door grid world MDPs to great effect.

Similarly, \citet{da2012learning} study the acquisition of \textit{parameterized} options. That is, each option policy is associated with some parameter that allows the option to adapt flexibly to specific aspects of the given domain. The proposed approach is cast as a series of regression problems, given data gathered on a collection of training MDPs. In particular, this data is used to estimate the geometry of policy space in terms of some number of low-dimensional manifolds. Then, a series of regression problems are solved that give rise to the parameters of each option. Experiments are conducted in a challenging dart throwing game, with results providing substantial evidence for the effectiveness of the learned parameterized options.

Separately, \citet{brunskill2014pac} study the option discovery problem in the lifelong setting. In particular, adopt the perspectives of Probably Approximately Correct (PAC) learning introduced in the seminal work by \citet{Valiant1984}. The first result is a highly general form of a PAC-MDP \cite{Strehl2009} guarantee adapted to the case of learning with options. Recall that a PAC-MDP algorithm is one that is said to have a polynomial bound on the number of mistakes made by the algorithm with high probability. Brunskill and Li first present (their Theorem 1) a PAC-SMDP guarantee, suited to the case where an agent inhabits an SMDP. Using this result, an extension of R-Max, SMDP-R-Max, is developed and analyzed that is again well suited to the case of learning with options. Finally, these insights are applied to the lifelong learning setting in which an agent will face a series of MDPs sampled from the same distribution, each sharing a state-action space \cite{wilson2007multi}.

\citet{thrun1995finding} performed one of the earliest studies of transferring abstract actions in RL. The work centeres around the SKILLS algorithm, which identifies correlated action sequences to group together into macro-operators called skills. In a similar vein, \citet{pickett2002policyblocks} propose Policy Blocks, an approach for generating useful options in lifelong RL. Given a set of optimal policies for some initial number of sampled MDPs, all possible policy combinations are enumerated and scored according to the size of their intersection with the solution policies. The $n$ best options found this way are then kept alongside the primitive actions during learning. 
\citet{topin2015portable} extend Policy Blocks to Object-Oriented MDPs~\cite{diuk2008objects}, thereby allowing for transfer to take place across tasks that share object structure. The main advantage is similar to the agent-space approach of \citet{konidaris2007building}: the components of each option can be defined in terms of objects and their relations that are guaranteed to exist across different tasks, thereby enabling high-level behaviors to be immediately applicable in new domains. \citet{macglashan2013multi} develops a suite of algorithms based on similar ideas, targeting the transfer of policies across tasks in continuous state settings, or when the tasks require different state representations.

Most recently, \citet{barreto2019option} propose the option keyboard, a framework for combining options together into compositions of novel behavior. The main perspective is to consider each option as a single key on a keyboard---then, when confronted with a new task that is a composition of previous tasks \cite{van2019composing}, a combination of keys can compose a ``chord" that can solve the new task. There are two key technical ideas underlying the option keyboard. First, a new perspective that views the process of combining distinct options as one of combining intrinsic reward functions that induce the options; Second, to make use of of \textit{generalized policy improvement} (GPI) first presented by \citet{barreto2017successor}, to inform the creation of new option policies. GPI is a mechanism for constructing a policy, $\pi$, from a collection of source policies, $\pi_1, \ldots, \pi_n$, such that $\pi$ is guaranteed to be no worse than any in the collection. Hence, the option keyboard first learns a collection of options designed to maximize independent intrinsic reward streams (also treated as generic cumulants). Then, these different options can be easily composed to form a new option that corresponds to any linear combination of the chosen intrinsic reward streams.

\paragraph{Options and Exploration.} As suggested by earlier work, options can, in ideal circumstances, dramatically improve the sample complexity of RL. Beyond the discovery and transfer work already surveyed, a recent line of work has explicitly concentrated on understanding how options impact exploration. For instance, \citet{fruit2017exploration} develop an algorithm for minimizing regret of learning options, building on SMDP-R-Max introduced \citet{brunskill2014pac}. Fruit and Lazaric propose SMDP-UCRL, an option-based RL algorithm that builds around UCRL \cite{auer2007logarithmic}. The main result of the work provides upper and lower bounds on the regret of this algorithm, along with additional analysis proving which cases the regret of option learning can be lower than that of traditional RL. However, SMDP-UCRL requires prior knowledge in the form of the distribution of reward and expected run time of each option. In follow up work, \citet{fruit2017exploration} build on SMDP-UCRL with Free-SMDP-UCRL, which no longer requires this prior knowledge, but still matches the original regret bound up to an additive constant.

\paragraph{Options and Neural Networks.} The focus of much the present survey has been on options in the context of finite MDPs, with occasional extensions into continuous state spaces. In many cases, to confront the complexity of rich state spaces, deep neural networks are exploited for their power in function approximation \cite{lecun2015deep}. To this end, \citet{bacon2017option} established an elegant adaptation of the policy gradient theorem \cite{sutton1999policy} to the problem of learning options. This result underlies the Option-Critic, a neural network architecture that supports end-to-end training of option policies and their termination conditions. Critically, learning options in the Option-Critic does not require any use of intrinsic rewards, which differentiates it from many of the option discovery approaches surveyed previously. Bacon et al. present strong empirical evidence that the Option-Critic can accelerate learning on challenging RL domains, even those with rich state spaces such as games from the Arcade Learning Environment (ALE) \cite{bellemare2013arcade}.

The Option-Critic work was later extended in several ways to cooperate with policy optimization methods \cite{zhang2019dac}, and to generate hierarchies of arbitrary depth \cite{riemer2018learning}. \citet{harb2018waiting} incorporate the notion of \textit{deliberation cost} into the training objective, drawing on ideas from bounded rationality \cite{simon1957models}. Here, as with the work of \citet{csimcsek2009skill}, the goal is to clarify what is meant by a good option. Harb et al. answer this question by arguing that options may be used to help resource-bounded agents make decisions efficiently under harsh computational constraints. This perspective leads to the introduction of the deliberation cost that acts as a regularizer to encourage options that execute for longer periods of time, following similar reasoning to \citet{mankowitz2014time}. Additionally, \citet{tiwari2019natural} offered pathways for incorporating the natural gradient \cite{amari1998natural} into the Option-Critic, yielding performance gains on benchmark tasks.

\paragraph{The Initiation and Termination of Options.} The Option-Critic is primarily concerned with learning the option policies and the termination condition, but assumes that the options are available everywhere (so $\mc{I}_o = \mc{S}$, for all $o$). \citet{khetarpal2020options} extend the Option-Critic to also incorporate the a generalization of the initiation condition called the interest function of each option. Similarly to the Option-Critic, a gradient-based update rule is developed that is suitable for learning options, their interest functions, and their termination conditions, resulting in the Interest-Option-Critic. 
On the termination side, \citet{harutyunyan2018learning} propose options that terminate in an ``off-policy" way, enabling unification of typical off-policy TD updates and option updates. This gives rise to a new option learning algorithm, $Q(\beta)$, that enables faster convergence by learning $\beta$ in an off-policy manner. Similarly, \citet{mankowitz2014time} study interrupting options, a means of improving a given set of options during planning. Their idea is to alter a given option's predefined termination condition based on information computed during planning. In this way, options can be iteratively improved via a Bellman like update (with interruption added).
They demonstrate that these new options also lead to a contraction-mapping that ensures convergence of the option value function to a fixed point. Their main contribution is to build regularization into this framework by encouraging their operator to choose longer options. Later,
\citet{mankowitz2016adaptive} propose Adaptive Skills Adaptive Partitions (ASAP), a framework for learning when to apply options and what they should do. ASAP is well suited for continuous state domains, and comes along with strong properties, including the correction of a misspecified model, and convergence to a locally optimal set of options and partitions.

\paragraph{Options and Planning.} Perhaps the greatest potential of options is their capacity to empower the planning capabilities of RL agents.
\citet{silver2012compositional} develop compositional option models, which enable recursive nesting of option models through a generalization of the Bellman operator. Using this new operator, Silver and Ciosek present an algorithm designed for goal-based MDPS that estimates the transition model and termination condition for both the goal and subgoals simultaneously. These options and their models are shown to greatly accelerate planning on the classical planning problems of Towers of Hanoi and and a navigation task. Separately, \citet{mann2014scaling,mann2015approximate} analyze the convergence rate of approximate dynamic programming with and without options. The main result of the work proves that options can improve the convergence rate of approximate value iteration. The degree of improvement depends on how long the options run for, whether the value function is initialized pessimistically, and the value of the policies associated with the options.

In a different vein, \citet{konidaris2014constructing,Konidaris2015} and \citet{james2018learning} use options to generate a well behaved abstract state representation, even if the underlying environment is continuous. Building on the chaining work discussed earlier, the main idea is again to draw on sequential composition from robotics \cite{burridge1999sequential}; that is, for each option, the termination condition of the option must lie entirely inside of the initiation condition of at least one other option. Using this idea, Konidaris et al. construct an abstract symbolic state representation based on the initiation-termination relationships in a given set of options. The result is an algorithm that can translate a complicated domain and a set of options into a domain that is representable in a classical planning language like \textsc{Strips}. Notably, the resulting discretized state space, along with the given options, is proven to have the property that any plan consisting of these options is \textit{feasible}---that the sequence of operators is in fact executable. \citet{silver2020learning} study a related problem, focusing on how to learn options that may then be exploited by a \textsc{Strips} planner given new, more complicated goals than those seen during the learning of the options.

Indeed, existing research on options is broad, exciting, and growing. Many open fundamental questions remain, a few of which I will study in \autoref{part:action_abstraction}.

\subsection{Other Forms of Abstraction}
\label{sec:c2_state_action_abstr}

State and action abstraction on their own have each been of long lasting interest in the RL literature. Many of the surveyed methods were in fact designed to carry out both types of abstraction simultaneously, or focus on one while carrying out the other implicitly. This is particularly true in the broader study of hierarchical abstraction, which allows for the representation of phenomena at different levels of granularity.

While my review of prior work concentrates on methods that focus on just state or action abstraction, there has been a rich history on both hierarchical abstraction and joint state-action abstraction.

I will differentiate between the process of state-action abstraction from hierarchical abstraction as follows. A single application of a state \textit{and} action abstraction will be defined as state-action abstraction, whereas a hierarchical abstraction is the repetition of $n > 1$ applications of state or action abstraction. Ultimately, this difference is purely for convenience, as a single state-action abstraction is effectively a shallow hierarchy.

\paragraph{State-Action Abstraction.} Together, state and action abstractions can distill complex problems into simple ones~\citep{jonsson2001automated,ciosek2015value}. As with the other types of abstraction, the literature on joint state-action abstraction is too broad to cover in its entirety. I instead highlight select works that are particularly relevant to the objectives of this dissertation.

Perhaps most relevant is are those approaches that inform state-action abstraction through \textit{MDP homomorphisms} \citet{ravindran2002model,ravindran2003relativized,ravindran2003smdp,ravindran2004approximate,ravindran2003thesis}.
MDP homomorphisms form a compressed representation of a given MDP by collapsing state-action pairs that can be treated as equivalent. First, \citet{ravindran2002model} adapt homomorphisms as used in finite state automata \cite{hartmanis1966algebraic} for application to MDPs, building on the model minimization techniques of \citet{dean1997model}. The main idea is to search through the state-action space for symmetries that allows for the formation of a functionally identical MDP. This tool is then exploited for the purpose of constructing homomorphisms and options that induce more compact representation of the original problem that is similar functionally.

In subsequent work, these ideas are expanded, building toward the more general family of SMDP-homomorphisms \citet{ravindran2003smdp} that allows for the discovery of symmetries between an MDP and SMDP. Later, \citet{ravindran2004approximate} generalize the previous frameworks to account for \textit{similarity} of state-action pairs, rather than equivalence, a move similar to the one I make in \autoref{chap:approx_state_abstr}. In this work, Ravindran and Barto introduce approximate MDP homomorphisms and prove the conditions under which they are guaranteed to preserve representation of good behavior---this result is one of the strongest of its kind, and heavily inspires the work of this dissertation. For more on MDP homomorphisms as a framework for abstraction, see the dissertation by \citet{ravindran2003thesis}. Lastly, in later years, \citet{majeed2019performance} extend the analysis of Ravindran and Barto to non-Markovian settings, proving the existence of several classes of value preserving homomorphisms---these classes closely resemble some of the families of state abstraction discussed earlier: for instance, one family studied groups histories of states together that induce similar value functions.

\citet{mugan2008towards,mugan2009autonomously,mugan2011autonomous} develop a holistic approach called Qualitative Learner of Action and Perception (QLAP) that autonomously discovers state and action abstractions, even in MDPs with an underlying rich state and action space. The main idea is to first learn a \textit{qualitative} state representation supposing that the agent can observe the value of different variables changing over time. By executing effectively random actions, a this qualitative representation learns measures such as \textit{magnitude} and \textit{change} variables of the observable quantities. Once enough data is collected, options are defined that explicitly modify the qualitative variables captured by the learned state representation. Experiments are conducted in the robotic simulator BREVE \cite{klein2003breve} in which a simulated humanoid robot is asked to manipulate objects on a tabletop. QLAP is able to successfully learn in a variety of tasks involving activities like pushing a block to a particular location picking up a block. 

Finally, \citet{bai2017efficient} develop a Monte Carlo planning algorithm that incorporates state and action abstractions to efficiently solve the Partially Observable MDP \citep{kaelbling1998planning} induced by these abstractions. That is, given an MDP with state space $\mc{S}$, the state-action abstraction induce a POMDP with observation space $\mc{S}_\phi$. The main results present guarantees on the performance of the algorithm: Theorem 1 shows that the value loss of their approach is bounded as a function of the state aggregation error \citep{Hostetler2014}, and Theorem 2 shows their algorithm converges to a recursively optimal policy for given state-action abstractions.

\paragraph{Hierarchical Abstraction.} Hierarchical abstraction captures methods that form representations---either of state, action, or both---at different levels of granularity. Like the other forms of abstraction, hierarchy has a rich history in RL, dating back to early work on feudal learning by \citet{dayan1993feudal}, hierarchical $Q$-learning by \citet{wiering1997hq}, HAMs by \citet{parr1998reinforcement}, and the MAXQ framework by \citet{dietterich2000hierarchical}. Since then, research has continued to establish the core principles of hierarchical abstractions, including the study of bayesian hierarchical RL \cite{cao2012bayesian}, model-based hierarchical RL \cite{jong2008hierarchical,li2017efficient,franccois2019combined}, model-free hierarchical RL \cite{kulkarni2016hierarchical,florensa2017stochastic,vezhnevets2017feudal,nachum2018data,nachum2018near,Levy2019LearningMH}, learning hierarchies in imitation learning \cite{le2018hierarchical}, from demonstration \cite{Mehta2011}, for transfer \cite{mehta2007automatic,mehta2008automatic,mehta2008transfer}, in multi-agent RL \cite{ghavamzadeh2006hierarchical}, and for planning \cite{konidaris2016constructing,gopalan2017planning,konidaris2018skills,winder2020planning}. For more on the early works of hierarchical RL, see the survey by \citet{barto2003recent}.

In summary: the literature on understanding abstraction and its role in RL is expansive, and far too broad to cover in this dissertation. I will return to direct comparisons where appropriate in subsequent chapters, again drawing the distinction between the four abstraction types: 1) state, 2) action, 3) state-action, and 4) hierarchical. A visual illustrating the intuitive difference between these four types of abstraction is presented in \autoref{fig:c2_diff_forms_of_abstr}.

\input{misc/abstraction_type_fig}

\section{Abstraction Desiderata}
\label{sec:c2_desiderata}

What is a \textit{good} abstraction? A natural route to answering this question is to measure an abstraction's utility in terms of the quality of the representations the abstraction induces, with a focus on how these representations change the RL problem. These considerations could be made with respect to a particular choice of RL algorithm (or perhaps family, such as model-free), or, in contrast, may be agnostic to the choice of RL algorithm. Indeed, the right abstraction for one type of algorithm may be entirely useless to another. Similarly, some abstractions may be effective in certain kinds of environments---those that abstract aggressively may be most appropriate in highly uncompressed worlds, for instance. Further, it might be the case that the properties underlying useful state abstraction differ from action.

Throughout this dissertation, I advocate for three simple properties that both state and action abstractions should have regardless of the choice of environment or RL algorithm. For this reason, these desiderata are not intended to be exhaustive. There are surely other properties we might hope abstractions possess depending on the broader context, domain, or resource requirements. Still, it is useful to highlight an initial set of properties that capture at least some of what is meant by ``good" abstraction---this is precisely the purpose of these desiderata. They are as follows.

\begin{enumerate}[(D1)]

\item \textsc{Efficient-Creation}: Computing or learning the abstraction should not be prohibitively difficult.

\emph{Measurement:} The most natural evaluation for D1 is to provide sample bounds or computational complexity results that illustrate what resources are required to accurately learn or construct the abstraction.

\item \textsc{Efficient Decision Making}: An abstraction should enable efficient decision making. That is, planning or learning with a good abstraction should be faster than planning or learning without it.

\emph{Measurement:} It is natural to measure such quantities in terms of the speed with which RL or its subproblems can be solved on MDPs of relevance. Concretely, an abstraction should lower the computational complexity of planning, or the sample complexity of RL. Throughout this dissertation I will sometimes use the \emph{size} of the induced abstract model as a proxy for this measure, as most worst case sample complexity, computational complexity, and regret bounds depend on the size of the MDP being solved.

\item \textsc{Near Optimality}: An abstraction should enable agents to discover policies that solve the original problem to a satisfactory degree.

\emph{Measurement:} I measure this property based on some variant of a bound on the \emph{value loss} of an abstraction discussed throughout the chapter. Such a bound captures the suboptimality of the best abstract policy in the environmental MDP. In the case of state abstraction, value loss is defined in a straightforward way.
\ddef{$\phi$ Value Loss}{A \textbf{value loss} bound of a state abstraction $\phi \in \phiall$ is any value $\tau \in \mathbb{R}$ such that
\begin{equation}
    \min_{\pi_\phi \in \Pi_\phi} \max_{s \in \mc{S}} V^*(s) - V^{\pi_\phi}(s) \leq \tau,
\end{equation}
with $V$ the ground MDP's value function and $\Pi_\phi$ the set of all policies over abstract states.}

Note that to extend this definition to an action abstraction $\omega$, we will require some extra machinery to define the ground value function under a policy over options, as it is not necessarily well defined due to the semi-Markovian nature of option execution. I remedy this fact in \autoref{chap:vpsa} with the introduction of joint state-action abstractions that are guaranteed to yield a policy class $\Pi_{\mc{O}_\phi}$ for which every entity has a well defined ground value function.

Other measures of optimality include \emph{recursive} optimality and \emph{hierarchical} optimality introduced by~\citet{dietterich2000hierarchical}. The value loss mentioned above is a bound on the global optimality, and so is stronger than either recursive or hierarchical optimality.

\end{enumerate}

Abstraction, broadly speaking, reduces the dimensionality of an entity. In the context of sequential decision making, abstraction reduces the representational complexity needed to support efficient learning in complex decision making problems. This process is captured by D2 (Efficient Decision Making) and D3 (Near-Optimality), with D1 further requiring that abstractions should be easy to create, given the computational and statistical budget available.

I take these three statements as guiding principles that help govern which abstractions to learn in RL. Collectively, they state the following:
\begin{center}
    \emph{Good abstractions for RL are $\underbrace{\text{easy to discover}}_{\text{(D1)}}$ and $\underbrace{\text{enable efficient learning}}_{\text{(D2)}}$ of $\underbrace{\text{high value policies}}_{\text{(D3)}}$.}
\end{center}

I now show that satisfying any \emph{one} or any \emph{two} desiderata is trivial. In each of the below remarks, I let $M$ denote the environmental MDP and $\tilde{M}$ denote the abstract MDP induced by a pair $\phi, \mc{O}$, with components $(\mc{S}_\phi, \mc{O}, R_{\phi,\omega}, T_{\phi,\omega}, \gamma, \rhoz^{\phi})$ and optimal policy $\tilde{\pi}^*$. The reward and transition functions resulting from $\phi$ and $\omega$ are defined as a straightforward combination of the MTM with the weighted average mechanism that underlies $R_\phi$ and $T_\phi$. For more detail on the construction of this abstract MDP, and particularly the components $R_{\phi,\omega}, T_{\phi,\omega}$, see \autoref{chap:vpsa}. Further, I let $I_{\phi, \omega}$ denote the identity abstraction, such that $I_{\phi, \omega}(M) = M$. More formally, $I_{\phi, \omega}$ is the pair $(\phi_I, \omega_I)$, where $\phi_I : s \mapsto s$, and $\omega_I : \mc{A} \mapsto \mc{O}_\mc{A}$, with $\mc{O}_\mc{A}$ the primitive actions redefined as options as per the scheme described earlier in the chapter.

\begin{remark}
All three desiderata are trivial to satisfy individually.
\end{remark}

\begin{dproof}[D1]
For D1 (efficient abstraction discovery), consider $I_{\phi, \omega}$. The abstraction is the identity function, and so requires no computation or learning. \qedhere
\end{dproof}
\vspace{-4mm}

\begin{dproof}[D2]
For D2 (supports efficient decision making), suppose we replace the ground state and actions space with a single state and single action: $|\mc{S}_\phi| = |\mc{O}| = 1$. Clearly, such a resulting MDP satisfies the first desiderata---it is trivial to plan or learn in the resulting MDP. \qedhere
\end{dproof}
\vspace{-4mm}

\begin{dproof}[D3]
For D3, consider $I_{\phi,\omega}$. The optimal policy for $M_{I_{\phi, \omega}}$ is exactly the optimal policy for $M$, thus preserving representation of high value policies. \qedhere
\end{dproof}
\vspace{-4mm}

I now show that any pair of desiderata are trivial to satisfy.

\begin{remark}
Any two desiderata are trivial to satisfy.
\end{remark}

\begin{dproof}[D1 \& D2]
For D1 and D3, we again consider the abstraction that induces an abstract MDP consisting of a single state and action. Planning and learning in this MDP are trivial, and the abstraction can be created without any computation or data. \qedhere
\end{dproof}
\vspace{-4mm}

\begin{dproof}[D2 \& D3]
For D2 and D3, suppose we solve for the optimal policy $\pi^*$ in $M$ and abstract according to the $\pi^*$-irrelevance abstraction that clusters states based on optimal action in each state \cite{jong2005state,li2006towards}. The resulting abstract MDP is as small as can be without losing the optimal policy, per the result of~\citet{li2006towards}, and so may be said to support quick learning (under the assumption that MDP size may be treated as a proxy for learning difficulty). Further, the abstract policy $\tilde{\pi}^*$ is guaranteed to be optimal when applied in the ground: $\max_{s \in \mc{S}} V^*(s) - V^{\tilde{\pi}^*}(s) = 0$. \qedhere
\end{dproof}
\vspace{-4mm}

\begin{dproof}[D1 \& D3]
For D1 and D3 we again invoke $I_{\phi, \omega}$. Clearly, the identity function is easy to compute and the optimal policy of $\tilde{M}$ will necessarily preserve optimality.\qedhere
\end{dproof}
\vspace{-4mm}

The case of interest is an abstraction that satisfies \emph{all three} desiderata. Really, though, none of the properties expressed by the desiderata are themselves boolean functions. They can each be satisfied to a different degree. Depending on the situation, it might be prudent to represent an near-optimal policy, or to ensure learning is as fast as possible. Thus, when we look for abstractions, our attention will be on those that achieve an appropriate trade off between the different properties.

In general, it is unclear whether there is a single optimal abstraction---it will largely depend on the broader objectives guiding the agent. Do we care about sample efficiency, safety, asymptotic performance, or reliability? Depending on these criteria, and on the resources available to the RL agent, different abstractions may be better suited to the given context. For this reason, much of the analysis in this dissertation is focused on understanding the interplay between these desiderata. 

I now turn to the primary technical contributions of this work, beginning with state abstraction.

\vspace{4mm}
\begin{center}
\begin{minipage}{0.65\textwidth}
\resizebox{0.9\textwidth}{6pt}{%
  \begin{tikzpicture}
        \pgfornament[symmetry=h]{85} 
    \end{tikzpicture}
}
\end{minipage}
\end{center}

%% file: algorithms/r_max.tex
\begin{algorithm}[b!]
\caption{R-Max}
\label{alg:r_max}
\textsc{Input:} $\eps$, $\delta$, $\gamma$, $s_0$ \\

\begin{algorithmic}[1]

\State{$m = f(\eps, \delta)$}
\State{initialize $\hat{T}, \hat{R}, n$}
\State{$\hat{Q} = \texttt{solve\_mdp}(\hat{R}, \hat{T}, \gamma)$}
\State{$t = 0$}
\While{True}
    \State{$a_t = \argmax_b \hat{Q}(s_t, b)$}
    \State{$r_t, s_{t+1} = \texttt{act}(s_t, a_t)$}
    \State{$n(s_t,a_t) = n(s_t,a_t) + 1$}
    \If{$n(s_t,a_t) = m$}
        \State{$\hat{Q}, \hat{R}, \hat{T} = \texttt{update\_model()}$}
    \EndIf
    \State{$t=t+1$}
\EndWhile
\end{algorithmic}
\end{algorithm}

%% file: algorithms/q_learning.tex
\begin{algorithm}[!b]
\caption{$Q$-Learning}
\label{alg:q_learning}
\textsc{Input:} $\alpha$, $\gamma$, $s_0$, $Q_0$, $\eps$ \\

\begin{algorithmic}[1]

\State{$t = 0$}
\While{True}
    \State{$a_t \sim \pi_{Q_t,\eps}(\cdot \mid s_t)$}
    \State{$r_t, s_{t+1} = \texttt{act}(s_t, a_t)$}
    \State{$Q_{t+1}(s_t, a_t) =  (1-\alpha)Q_{t}(s_{t}, a_{t}) + \alpha (r_{t} + \max_{a' \in \mc{A}} Q_{t}(s_{t+1},a'))$}
    \State{$t = t + 1$}
\EndWhile
\end{algorithmic}
\end{algorithm}

%% file: algorithms/value_iteration.tex
\begin{algorithm}[!t]
\caption{Value Iteration}
\label{alg:value_iteration}
\textsc{Input:} $M$, $\Delta$, $V_0$ \\

\begin{algorithmic}[1]

\While{True}
\For{$s \in \mc{S}$}
    \State{$V_{t+1}(s) = \max_{a \in \mc{A}} \sum_{s' \in \mc{S}} R(s,a,s') + \gamma T(s' \mid s,a) V_t(s')$}
\EndFor
\If{$\max_{s \in \mc{S}} |V_{t+1}(s) - V_{t}(s)| \leq \Delta$} \\

    \hspace{12mm}\Return{$V_{t+1}$}
\EndIf
\State{$t = t + 1$}
\EndWhile
\end{algorithmic}
\end{algorithm}

%% file: figures/tables/sa_table.tex
\begin{table}[!h]
    \centering
    \renewcommand*{\arraystretch}{1.3}
    \begin{tabular}{llll}
    \toprule
    Name& Predicate& Value Loss& Transitive \\
    \midrule
    $\phi_{Q^*}$& $\max_a |Q^*(s_1,a) - Q^*(s_2, a)| = 0$& 0& yes \\
    
    $\phi_{a^*}$& $a_1^* = a_2^*\ \text{and}\ V^*(s_1) = V^*(s_2)$& 0& yes \\
    
    $\phi_{\pi*}$& $\pi^*(s_1) = \pi^*(s_2)$& 0& yes \\
    
    \vspace{2mm}
    
    $\phi_{Q_\eps^*}$& $\max_a |Q^*(s_1,a) - Q^*(s_2, a)| \leq \eps$& $\frac{2\eps \textsc{RMax}}{(1-\gamma)^2}$& no\\ 
    
    \vspace{2mm}
    
    $\phi_{mult}$& $\max_a \left| \frac{Q^*(s_1,a)}{\sum_b Q^*(s_1,b)} - \frac{Q^*(s_2, a)}{\sum_{b} Q^*(s_2,b)}\right| \leq \eps$&  $2\eps\frac{|\mc{A}|\textsc{RMax} + k}{(1-\gamma)^2}$ & no\\ 
    
    \vspace{2mm}
    
    $\phi_{bolt}$& $\max_a \left| \frac{e^{Q^*(s_1,a)}}{\sum_be^{Q^*(s_1,b)}} - \frac{e^{Q^*(s_2, a)}}{\sum_{b}e^{Q^*(s_2,b)}}\right| \leq \eps$& $2\eps\frac{\left(|\mc{A}|\textsc{RMax} + \eps k + k\right)}{(1-\gamma)^2}$& no\\ 
    
    $\phi_{Q_d^*}$& $\forall_a: \lceil \frac{Q^*(s_1, a)}{d}\rceil = \lceil \frac{Q^*(s_2, a)}{d}\rceil$&$\frac{2d\textsc{RMax}}{(1-\gamma)^2}$& yes\\ 
    
    \bottomrule
    \end{tabular}
    \caption{A summary of several existing state abstraction types.}
    \label{tab:c2_sa_table}
\end{table}

%% file: misc/abstraction_type_fig.tex
\begin{figure}[h!]
    \centering
    \subfloat{\includegraphics[width=0.46\textwidth]{figures/diagrams/c1/state_abstr_ex.pdf}} \subfhspace \subfloat{\includegraphics[width=0.46\textwidth]{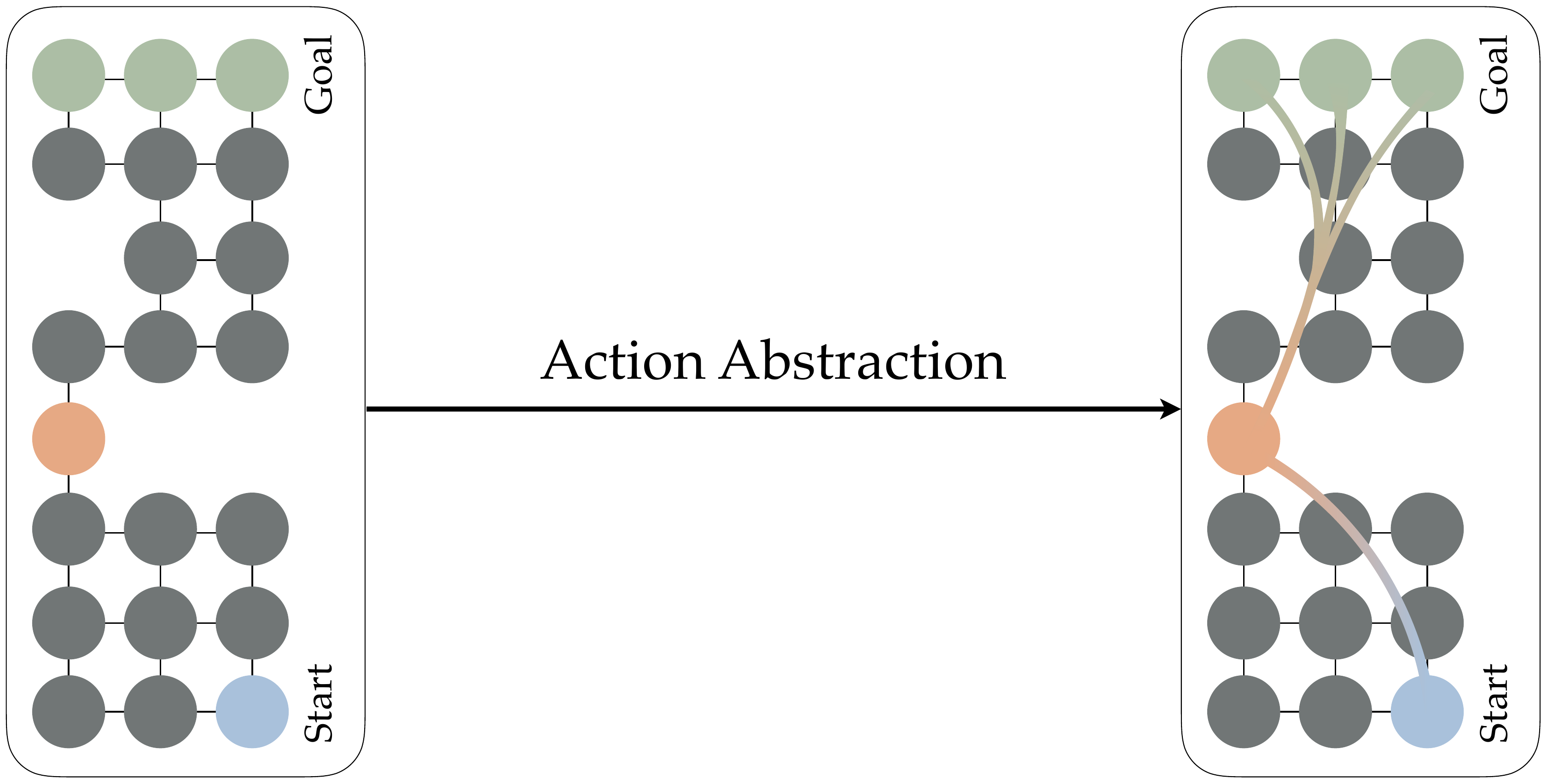}} \\
    \subfloat{\includegraphics[width=0.46\textwidth]{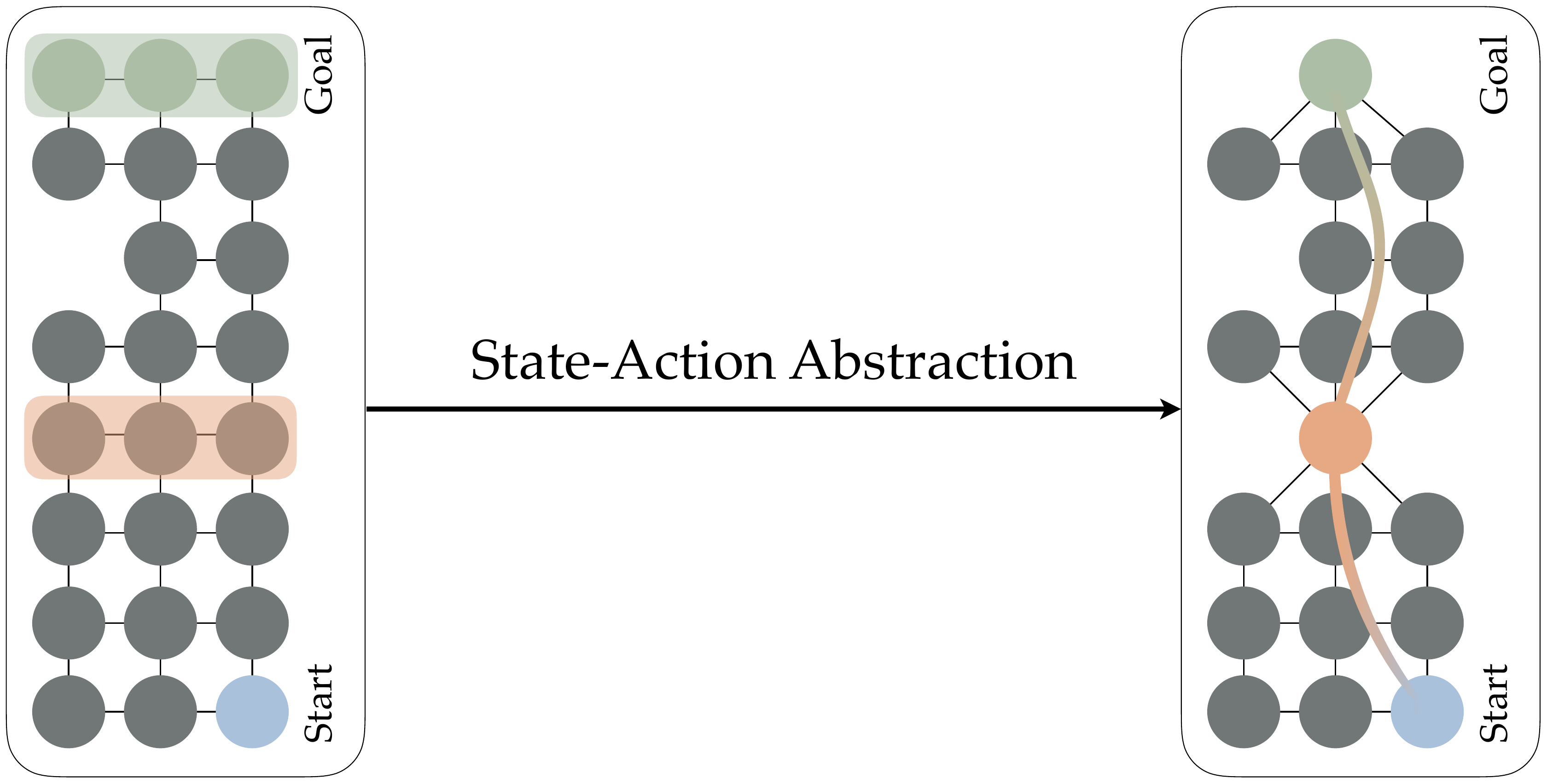}} \subfhspace \subfloat{\includegraphics[width=0.46\textwidth]{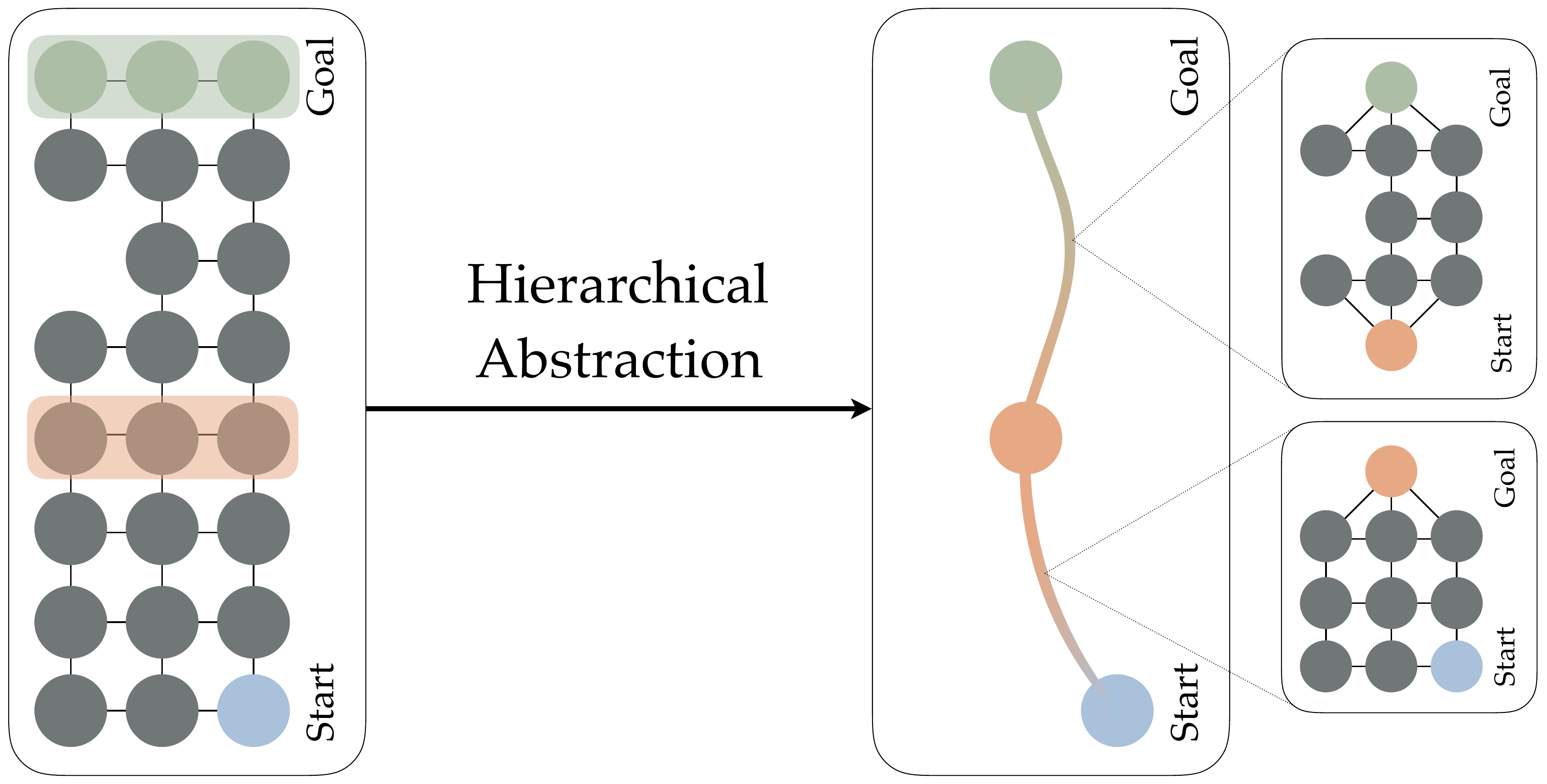}}
    \caption{The different forms of abstraction in MDPs.}
    \label{fig:c2_diff_forms_of_abstr}
\end{figure}

%% file: chapters/c3_sa_approx_state_abstraction.tex
\begin{center}
\begin{minipage}{0.8\textwidth}
\textit{This chapter is based on ``Near Optimal Behavior via Approximate State Abstraction" \cite{abel2016near}, jointly led by D. Ellis Hershkowitz, also in collaboration with Michael L. Littman.}
\end{minipage}
\end{center}
\vspace{2mm}

\begin{figure}[t!]
    \centering
    \includegraphics[width=0.7\textwidth]{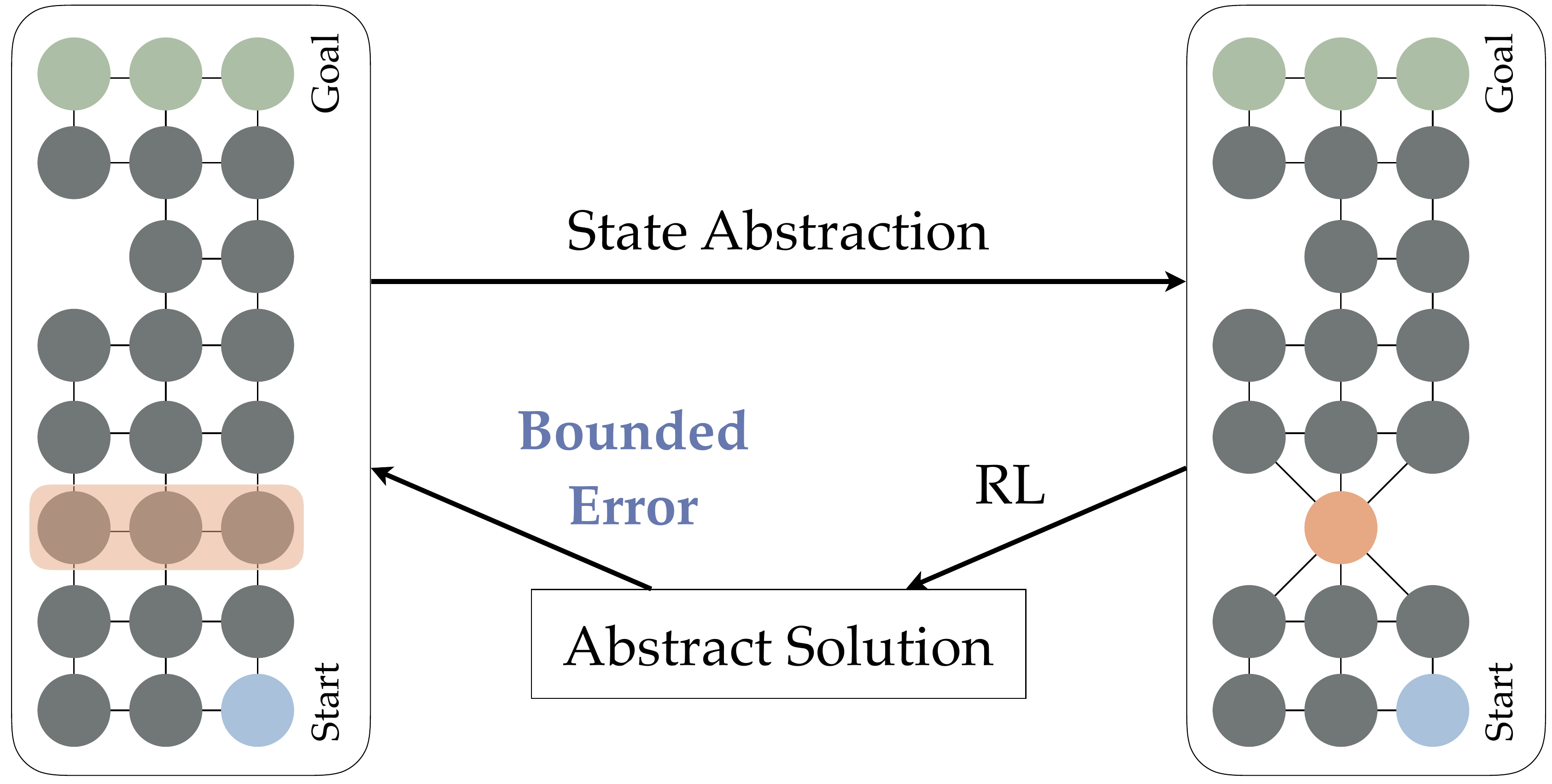}
    \caption{This chapter studies families of approximate state abstraction that induce abstract MDPs whose optimal policies have bounded value in the original MDP.}
    \label{fig:c3_overview}
\end{figure}

In this chapter I study which kinds of state abstraction are capable of preserving representation of good policies. Intuitively, abstraction of almost any kind throws away some amount of information. However, in RL, it is desirable (as per our third desiderata) that abstractions retain enough relevant information so as to allow RL agents to eventually learn to solve problems of interest. In light of this, I here introduce and analyze classes of \textit{approximate state abstraction} that are guaranteed to support representation of near-optimal behavior, as pictured in \autoref{fig:c3_overview}. Concretely, approximate state abstractions aggregate states based on degrees of similarity in terms of relevant functions like $Q$, $R$, and $T$. As we will see, these approximate state abstractions can jointly preserve representation of near-optimal behavioral policies while simultaneously reducing the size of the represented state space. In this way, this chapter is about state abstractions that satisfy desiderata D2 (efficient decision making) and D3 (near optimality). In the subsequent two chapters I show how to translate the main conceptual framework here introduced to accommodate all three desiderata.

I will first prove that approximate state abstraction can still preserve near-optimal behavior. The main result (\autoref{thm:c3_approx_sa_main_result}) shows that, by relaxing state aggregation criteria from equality to similarity, it is possible to achieve bounded value loss while offering three benefits. First, approximate abstractions make use of the kind of knowledge that we might expect a planning or learning algorithm to obtain without fully solving the MDP, and are thus more in line with the first desiderata. In contrast, exact abstractions often require solving for optimal behavior, thereby defeating the purpose of abstraction. Second, because of their relaxed criteria, approximate state abstractions can achieve greater degrees of compression than exact abstractions. This difference is particularly important in environments where no two states are identical. With a more compressed state space, many subprocedures critical to the overall effectiveness of an RL algorithm can be accelerated. For instance, planning in an abstract model $\hat{M}_\phi$ to compute an estimate of the optimal policy, $\hat{\pi}_\phi^*$, tends to be faster with a smaller state space. Third, because the state aggregation criteria are relaxed to near equality, approximate abstractions are able to tune the aggressiveness of abstraction by adjusting what they consider sufficiently similar states. I explicitly build an algorithmic framework around this idea in \autoref{chap:rlit}.

Furthermore, I empirically demonstrate the relationship between the degree of compression and error incurred on a variety of MDPs, illustrating a general trade off between compression and value-preservation that will serve as the focus of \autoref{chap:rlit}.

\section{Four Classes of Approximate State Abstraction}

I next describe four different types of approximate state abstractions that preserve near-optimal behavior by aggregating states on different criteria: $\phi_{Q^*_\eps}$, on similar optimal $Q$-values; $\phi_{model,\eps}$, on similarity of rewards and transitions; $\phi_{bolt,\eps}$, on similarity of a Boltzmann distribution over optimal $Q$-values; and $\phi_{mult,\eps}$, on similarity of a multinomial distribution over optimal $Q$-values. These four predicates are defined as follows.
\begin{align}
    p_{Q_\eps^*}(s_1,s_2) &\equiv \max_{a \in \mc{A}} |Q^*(s_1,a) - Q^*(s_2,a)| \leq \eps, \\
    p_{model,\eps}(s_1,s_2) &\equiv \max_{a \in \mc{A}} \left|R(s_1,a) - R(s_2,a)\right| \leq \eps_R\ \text{and} \nonumber \\
    &\hspace{4mm} \forall_{a \in \mc{A} ,s_\phi' \in \mc{S}_\phi, s' \in s_\phi'} : \left| T(s' \mid s_1,a) - T(s' \mid s_2,a) \right| \leq \eps_R,\\
    p_{mult}(s_1,s_2) &\equiv \max_{a \in \mc{A}} \left|  \frac{Q^*(s_1,a) - Q^*(s_2,a)}{\sum_{b} Q^*(s,b)} \right| \leq \eps, \\
    p_{bolt}(s_1,s_2) &\equiv \max_{a \in \mc{A}} \left|  \frac{e^{Q^*(s_1,a)} - e^{Q^*(s_2,a)}}{\sum_{b} e^{Q^*(s,b)}} \right| \leq \eps.
\end{align}

I now introduce the main theorem of this chapter that shows for each of the four classes of approximate abstraction, for \textit{any} finite MDP, the abstracted model preserves near-optimal behavior. More formally:
\begin{theorem}
\label{thm:c3_approx_sa_main_result}
There exist at least four types of approximate state aggregation functions, $\phi_{Q_\eps^*}$, $\phi_{model,\eps}$, $\phi_{mult}$ and $\phi_{bolt}$, for which the optimal policy in the resulting abstract MDP, applied to the environmental MDP, has suboptimality bounded by a function of ${\varepsilon}$:
\begin{equation}
\forall_{s \in \mc{S}}: V^*(s) - V^{\pi_\phi^*}(s) \leq 2 \eps \textsc{RMax} \eta_p,  
\label{eq:c3_icml2016_main_result}
\end{equation}
where $\eta_p$ depends on the predicate associated with state abstraction function types:
\begin{align}
\eta_{Q^*} &= \frac{1}{(1-\gamma)^2}, \\
\eta_{\text{model}} &= \frac{1 + \gamma \left( |\mc{S}_\phi| - 1 \right)}{(1-\gamma)^3}, \\
\eta_{\text{bolt}} &= \frac{\left(\frac{|\mc{A}|}{1-\gamma} + \eps k_{\text{bolt}} + k_{\text{bolt}}\right)}{(1-\gamma)^2}, \\
\eta_{\text{mult}} &= \frac{\left(\frac{|\mc{A}|}{1-\gamma} + k_{\text{mult}}\right)}{(1-\gamma)^2}.
\end{align}

For $\eta_{\text{bolt}}$ and $\eta_{mult}$, I also assume that the difference in the normalizing terms of each distribution is each bounded by some non-negative constant, $k_{\text{mult}} \in \mathbb{R}_{\geq 0}$, $k_{\text{bolt}} \in \mathbb{R}_{\geq 0}$ of $\varepsilon$:
\begin{align}
\left |\sum_i Q^*(s_1,a_i) - \sum_j Q^*(s_2,a_j) \right | &\leq k_{\text{mult}} \varepsilon, \\
\left |\sum_i e^{Q^*(s_1,a_i)} - \sum_j e^{Q^*(s_2,a_j)} \right | &\leq k_{\text{bolt}} \varepsilon.
\end{align}
\end{theorem}
Further, I note that $\eta_{model}$ of the original theorem has since been improved by a factor of $1/(1-\gamma)$ through Lemma 4 by \citet{taiga2018approximate} after the authors caught a mistake in our proof.

Naturally, the value bound of \autoref{eq:c3_icml2016_main_result} is vacuous for $2 \eps \eta_f \geq \frac{\textsc{RMax}}{1-\gamma} = \frac{1}{1-\gamma}$, since this is the maximum value achievable in any MDP (assuming $\textsc{RMin} = 0$). In light of this, observe that for $\eps = 0$, all of the above bounds are exactly 0. Any value of $\eps$ spanning between these two points achieves different degrees of abstraction, with different degrees of bounded loss. The degree of approximation (choice of $\eps$) changes the compression-value trade off made by the abstraction. In the closing section of the chapter (\autoref{sec:c3_experiments}), I present an empirical study of this relationship in a variety of benchmark MDPs. In each experiment the finding is consistent: as $\eps$ increases, the size of the abstract state space is reduced, and the value loss increases, though the rate at which this trade off is made differs across tasks. I return to a more technical treatment of this trade off in \autoref{chap:rlit}.

\section{Analysis}

I now introduce each approximate state aggregation family in more technical detail and prove the main result of the chapter. The proof strategy consists of proving a specific value loss bound for each of the four function types.

Let us consider an approximate version of Li et al.'s \cite{li2006towards} $\phi_{Q^*}$. In this abstraction, states are aggregated together when their optimal $Q$-values are within $\varepsilon$.
\ddef{$\phi_{Q^*_\eps}$}{
Let $\phi_{Q^*_{\eps}}$ define a type of state abstraction that, for fixed $\eps$, satisfies
\begin{equation}
\phi_{Q^*_{\varepsilon}}(s_1) = \phi_{Q^*_{\varepsilon}}(s_2) \implies \forall_{a \in \mc{A}} : \left|Q^*(s_1, a) - Q^*(s_2, a)\right| \leq \varepsilon.
\end{equation}
}

\begin{lemma}
\label{lem:c3_phi_Q}
When a $\phi_{Q^*_\eps}$ type abstraction is used to create the abstract MDP:
\begin{equation}
\forall_{s \in \mc{S}}: V^{\pi^*}(s) - V^{\pi_{\phi}}(s) \leq \frac{2\varepsilon \textsc{RMax}}{(1-\gamma)^2}.
\end{equation}
\end{lemma}

\input{proofs/c3/c3_approx_abstr_lemma_1.tex}

Now, consider an approximate version of \citet{li2006towards}'s $\phi_{model}$, where states are aggregated together when their rewards and transitions are within $\varepsilon$.
\ddef{$\phi_{model,\eps}$}{
We let $\phi_{model,\eps}$ define a type of state abstraction that, for fixed $\varepsilon_R \in [\textsc{RMin}, \textsc{RMax}]$ and $\eps_T \in [0,1]$ satisfies:
\begin{align}
\phi_{model,\eps}(s_1) = \phi_{model,\eps}(s_2) \implies &\forall_{a \in \mc{A}} : \left| R(s_1, a) - R(s_2, a)\right| \leq \eps_R \nonumber \\
&\hspace{8mm} \text{and} \nonumber \\
&\forall_{s_{\phi} \in \mc{S}_\phi, a \in \mc{A}} : \left|\sum_{{s}' \in s_{\phi}} \left[T(s' \mid s_1, a) - T(s' \mid s_2, a)\right] \right| \leq \eps_T.
\end{align}}

\begin{lemma}
\label{lem:c3_phi_model}
When $\mc{S}_\phi$ is created using a $\phi_{model,\varepsilon}$ type:
\begin{equation}
\forall_{s \in \mc{S}}: V^{\pi^*}(s) - V^{\pi_{\phi}}(s) \leq \textsc{RMax} \frac{2\eps_R + 2\gamma \eps_T \left( |\mc{S}_\phi| - 1 \right)}{(1-\gamma)^3}.
\end{equation}
\end{lemma}

\input{proofs/c3/c3_approx_abstr_lemma_2.tex}

As mentioned previously, the above bound has since been tightened by Lemma 4 of \citet{taiga2018approximate}. The new bound is
\begin{equation}
    \forall_{s \in \mc{S}}: V^{\pi^*}(s) - V^{\pi_{\phi}}(s) \leq \textsc{RMax} \frac{2\eps_R + 2\gamma \eps_T \left( |\mc{S}_\phi| - 1 \right)}{(1-\gamma)^2}.
\end{equation}

\subsubsection{Boltzmann over Optimal $Q$}
\label{sec:c3_bolt}

Next we introduce $\phi_{bolt,\eps}$, which aggregates states with similar Boltzmann distributions on $Q$-values. This type of state abstraction is appealing as a Boltzmann distribution over $Q$-values often shows up in exploration methods~\cite{sutton1998reinforcement}. We find this type particularly interesting for abstraction purposes as, unlike $\phi_{Q^*_\eps}$, it allows for aggregation when $Q$-value ratios are similar but their magnitudes are different.

\ddef{$\phi_{bolt,\eps}$}{
We let $\phi_{bolt,\eps}$ define a type of state abstraction that, for fixed $\varepsilon$, satisfies:
\begin{equation}
\phi_{bolt,\eps}(s_1) = \phi_{bolt,\eps}(s_2) \implies \forall_{a} \left|\frac{e^{Q^*(s_1,a)}}{\sum_b e^{Q^*(s_1,b)}} - \frac{e^{Q^*(s_2,a)}}{\sum_b e^{Q^*(s_2,b)}}\right| \leq \varepsilon.
\label{eq:c3_phi_bolt}
\end{equation}}

We also assume that the difference in normalizing terms is bounded by some non-negative constant, $k_{\text{bolt}}  \in \mathbb{R}_{\geq 0}$, of $\varepsilon$:
\begin{equation}
\left| \sum_{b \in \mc{A}} e^{Q^*(s_1,b)} - \sum_{b \in \mc{A}} e^{Q^*(s_2,b)} \right| \leq k_{\text{bolt}} \times\varepsilon.
\label{eq:c3_bolt_denom}
\end{equation}
\begin{lemma}
\label{lem:c3_phi_bolt}
When $\mc{S}_\phi$ is created using a function of the $\phi_{bolt,\eps}$ type, for some non-negative constant $k \in \mathbb{R}$:
\begin{equation}
\forall_{s \in \mc{S}} : V^{\pi^*}(s) - V^{\pi_{\phi}}(s) \leq \textsc{RMax} \frac{2\varepsilon\left(\frac{|\mc{A}|}{1-\gamma} + \varepsilon k_{\text{bolt}}  + k_{\text{bolt}} \right)}{(1-\gamma)^2}.
\end{equation}
\end{lemma}

\input{proofs/c3/c3_approx_abstr_lemma_3.tex}

\subsubsection{Multinomial over Optimal $Q$: $\phi_{mult,\eps}$}
\label{sec:c3_phi_mult}

Lastly, I consider a variant derived from a multinomial distribution over $Q^*$ for similar reasons to the Boltzmann distribution, with the multinomial offering the added appeal of simplicity.
\ddef{$\phi_{mult,\eps}$}{
Let $\phi_{mult,\eps}$ define a type of state abstraction that, for fixed $\varepsilon$, satisfies
\begin{equation}
\phi_{mult,\eps}(s_1) = \phi_{mult,\eps}(s_2) \implies \forall_{a} \left|\frac{Q^*(s_1,a)}{\sum_b Q^*(s_1,b)} - \frac{Q^*(s_1,a)}{\sum_b Q^*(s_1,b)}\right| \leq \varepsilon.
\end{equation}}

As with the Boltzmann class, I again assume that the difference in normalizing terms is bounded by some non-negative constant, $k_{\text{mult}} \in \mathbb{R}_{\geq 0}$, of $\varepsilon$:
\begin{equation}
\left |\sum_i Q^*(s_1,a_i) - \sum_j Q^*(s_2,a_j) \right | \leq k_{\text{mult}} \times \varepsilon.
\end{equation}
\begin{lemma}
\label{lem:c3_phi_mult}
When $\mc{S}_\phi$ is created using a function of the $\phi_{mult,\eps}$ type, for some non-negative constant $k_{\text{mult}} \in \mathbb{R}$:
\begin{equation}
\forall_{s \in \mc{S}_M} V^{\pi^*}(s) - V^{\pi_{\phi}}(s) \leq \textsc{RMax} \frac{2\eps \left(\frac{|\mc{A}|}{1-\gamma} + k_{\text{mult}}\right)}{(1-\gamma)^2}.
\end{equation}
\end{lemma}

\input{proofs/c3/c3_approx_abstr_lemma_4.tex}



\section{Experiments}
\label{sec:c3_experiments}

I next conduct experiments to highlight the impact state abstractions of the $\phi_{Q^*_\eps}$ type can have. I provide results for only $\phi_{Q^*_\eps}$ because, as per \autoref{lem:c3_phi_model}, \autoref{lem:c3_phi_bolt}, and \autoref{lem:c3_phi_mult}, the other three functions are reducible to particular $\phi_{Q^*_\eps}$ functions. The code for running these experiments is publicly available.\footnote{\url{https://github.com/david-abel/state_abstraction}}

I first explicitly construct an approximate state abstraction instance by approximating $Q^*$ through dynamic programming, then greedily aggregating ground states into abstract states that satisfy the $\phi_{Q^*_\eps}$ criteria. Since this approach represents an order-dependent approximation to the maximum amount of abstraction possible, I randomize the order in which states are considered across trials. Every ground state is equally weighted in its abstract state (that is, $w(s) = 1 / |\phi(s)|$).

\begin{figure}[t!]
\centering
\subfloat[Minefield]{
\includegraphics[width=0.42\columnwidth]{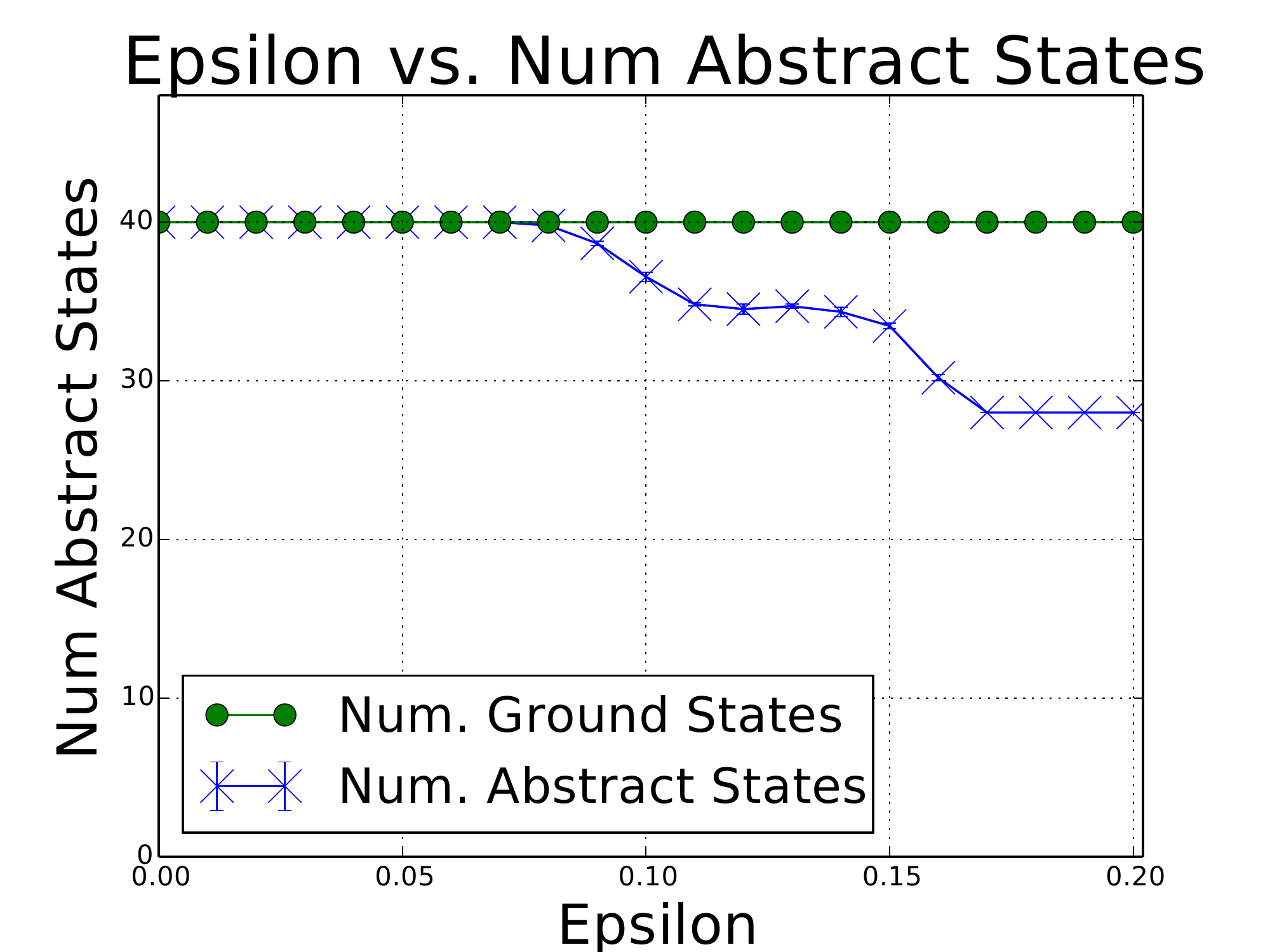}
}
\subfloat[Minefield]{
\includegraphics[width=0.42\columnwidth]{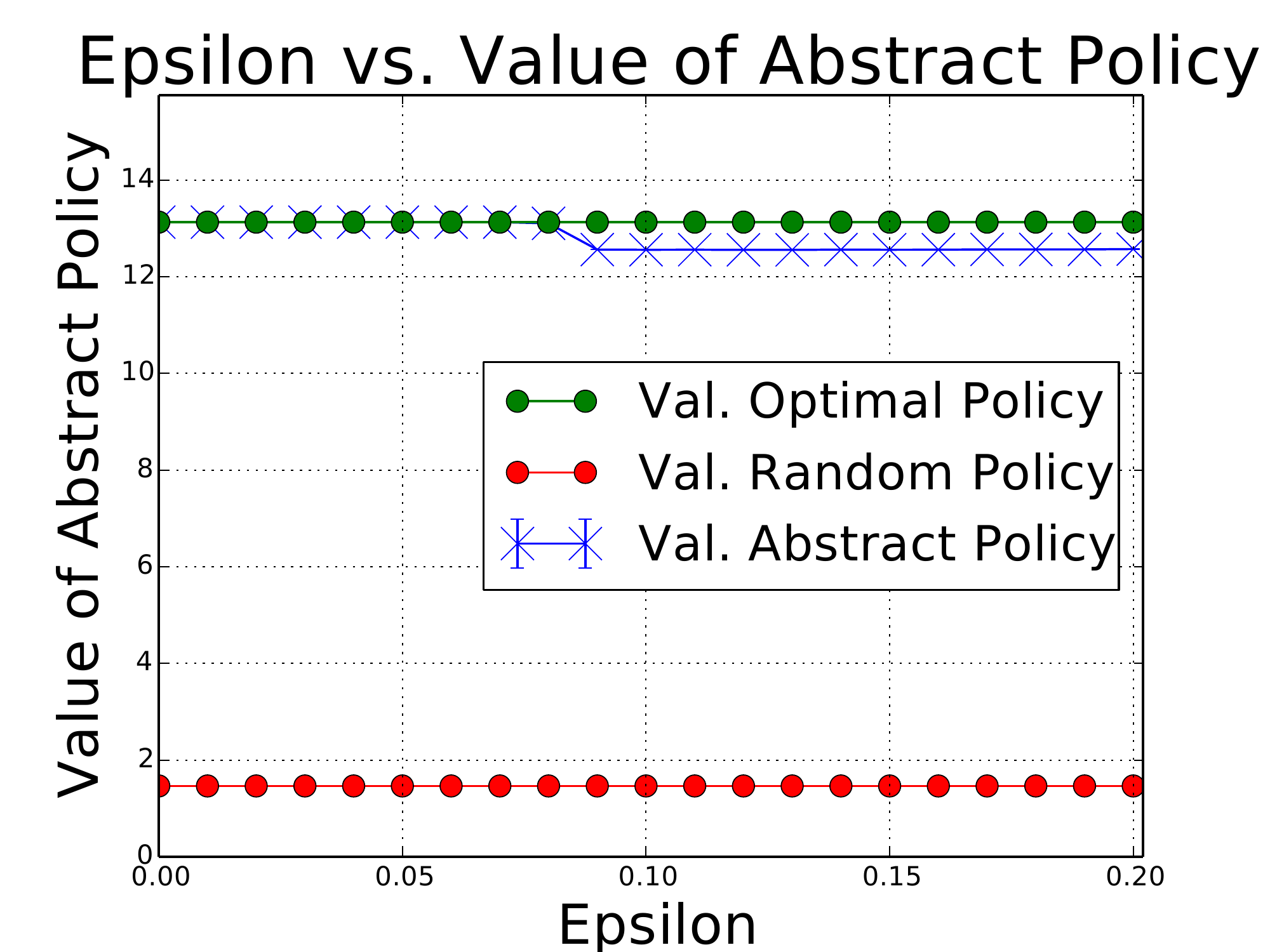}} \\
\subfloat[Taxi]{
\includegraphics[width=0.4\columnwidth]{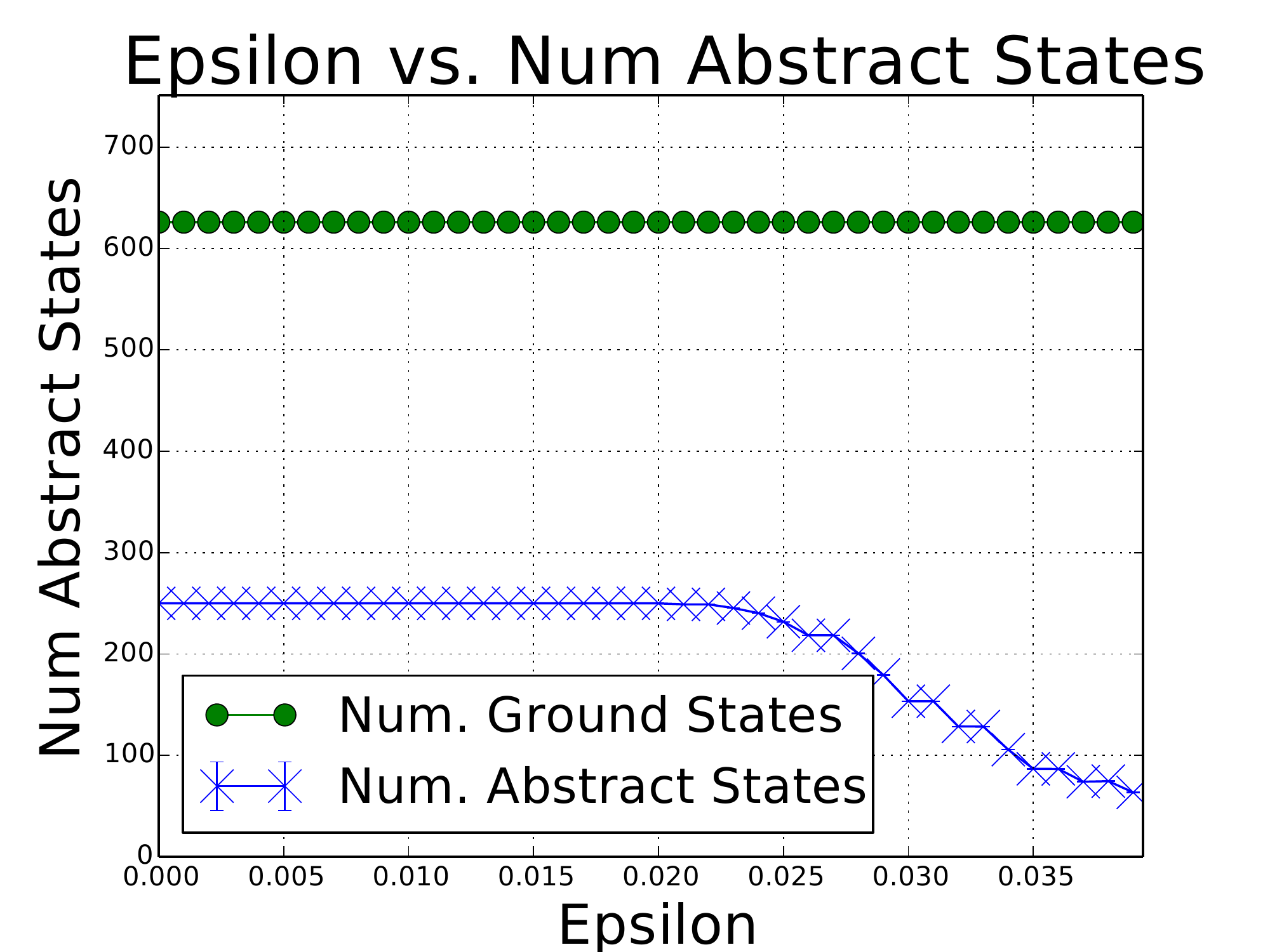}
}\subfhspace
\subfloat[Taxi]{
\includegraphics[width=0.4\columnwidth]{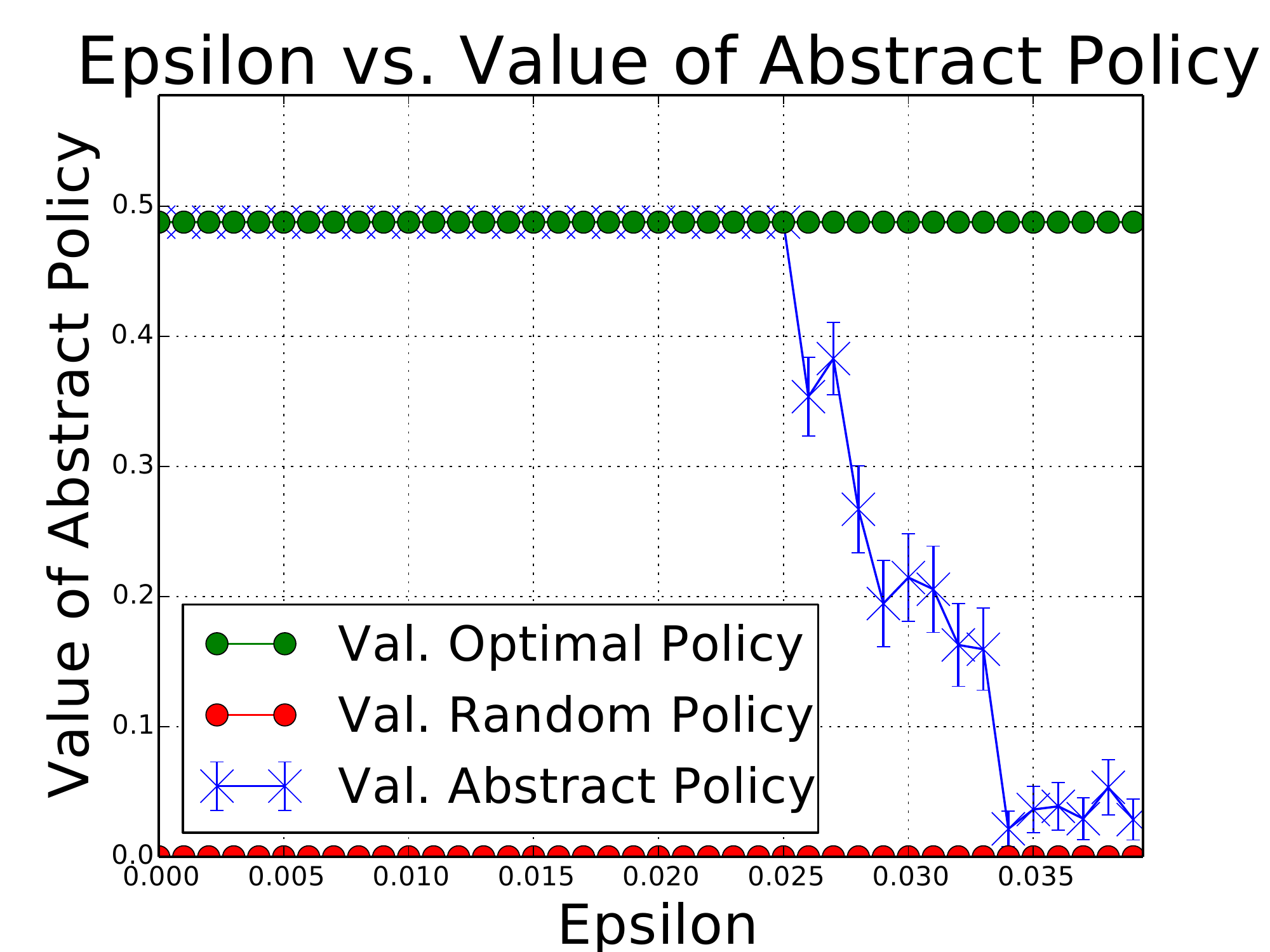}
} \\
\subfloat[Random]{
\includegraphics[width=0.4\columnwidth]{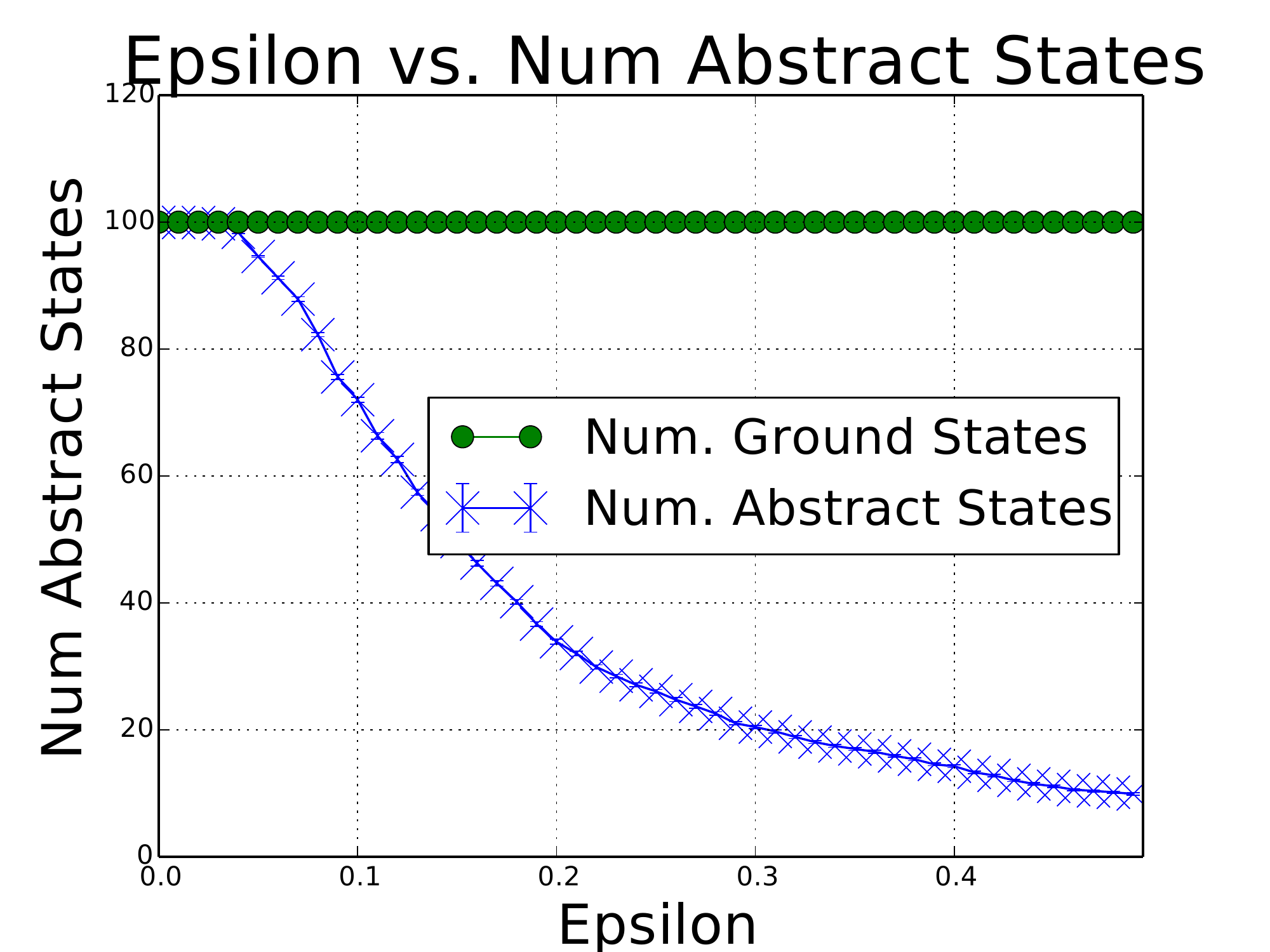}
}\subfhspace
\subfloat[Random]{
\includegraphics[width=0.4\columnwidth]{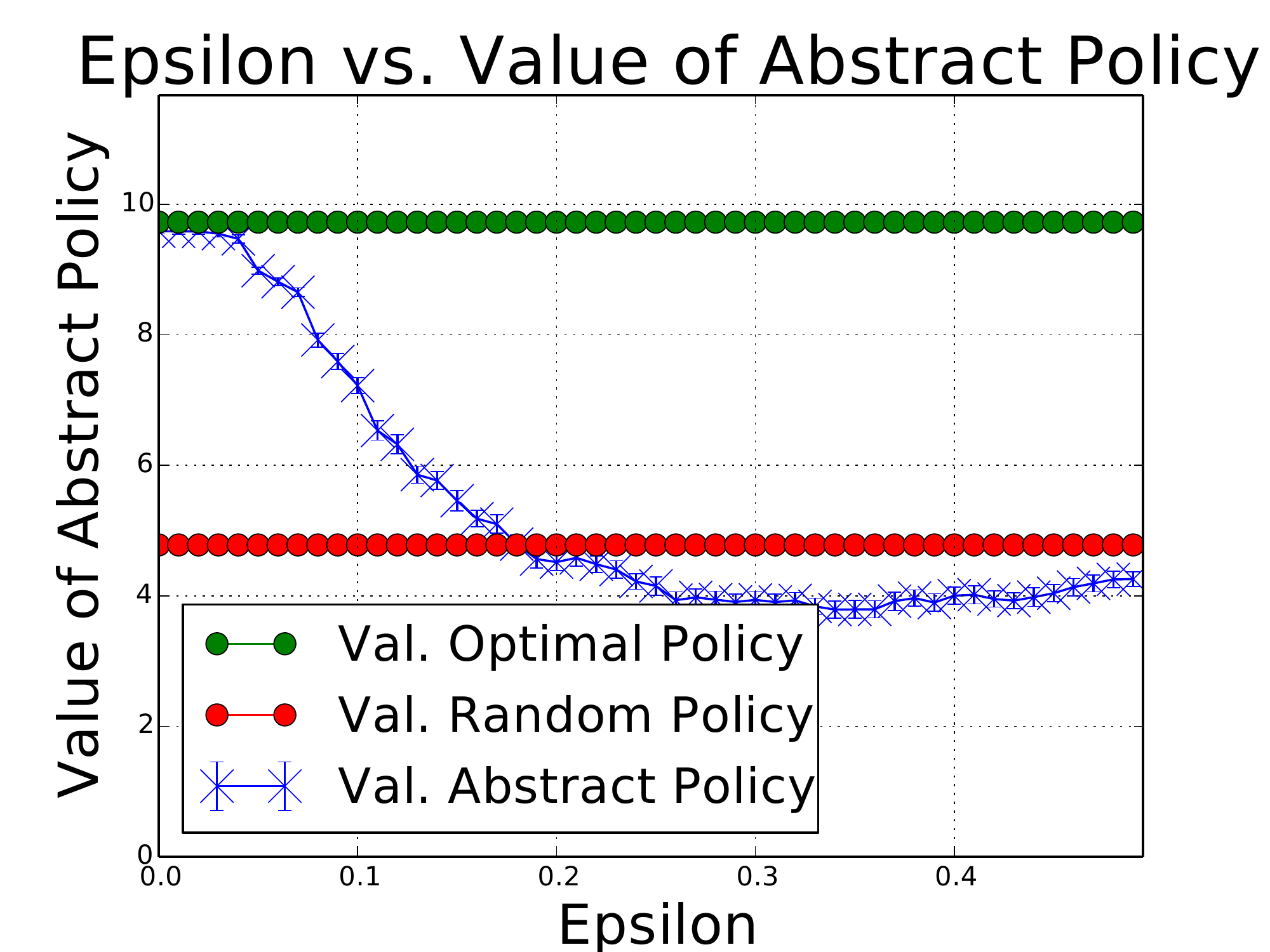}
}

\caption{$\varepsilon$ vs. Num States (left) and $\varepsilon$ vs. Abstract Policy Value (right).\label{fig:c3_empirical_results}}
\end{figure}

For each domain, I report the average number of abstract states and the value of the best abstract policy in the ground MDP, each with 95\% confidence intervals. First, I compare the number of states in the abstract MDP for different values of $\varepsilon$, shown in the left column of \autoref{fig:c3_empirical_results}. Second, I report the value under the abstract policy of the initial ground state, shown in the right column of \autoref{fig:c3_empirical_results}. In the Taxi and Random domains, 200 trials were run for each data point, whereas 20 trials were sufficient in Minefield.

These empirical results corroborate the main finding of this chapter---approximate state abstractions can decrease state space size while retaining bounded error. In Minefield, observe that as $\varepsilon$ increases from $0$, the number of abstract states is reduced, and optimal behavior is very nearly maintained. Similarly, in Taxi, when $\varepsilon$ is between $.02$ and $.025$, we observe a reduction in the number of states in the abstract MDP while value is fully maintained. For values of $\eps \geq .025$, increased reduction in state space size comes at a cost of value. Lastly, as $\varepsilon$ is increased in the Random domain, there is a smooth reduction in the number of abstract states with a corresponding cost in the value of the derived policy. When $\varepsilon = 0$, there is no reduction in state space size whatsoever (the ground MDP has 100 states), because no two states have identical optimal $Q$-values.

These experimental results also highlight a noteworthy characteristic of approximate state abstraction in goal-based MDPs. Taxi exhibits relative stability in state space size and behavior for $\varepsilon$ up to $.02$, at which point both fall off dramatically. I attribute the sudden fall off of these quantities to the goal-based nature of the domain; once information critical for achieving optimal behavior is lost in the state aggregation, solving the goal---and so acquiring any reward---is impossible. Conversely, in the Random domain, a great deal of near optimal policies are available to the agent. Thus, even as the information for optimal behavior is lost, there are many near optimal policies available to the agent that remain available. 

To summarize, in this chapter I motivated and introduced the family of approximate state abstraction. Each of the four types analyzed is guaranteed to preserve representation of good behavioral policies, with the degree of suboptimality induced a direct function of the amount of knowledge used to inform the state abstraction.

%% file: proofs/c3/c3_approx_abstr_lemma_1.tex
\begin{dproof}[Lemma~\ref{lem:c3_phi_Q}]
We first demonstrate that $Q$-values in the abstract MDP are close to $Q$-values in the ground MDP (\autoref{clm:c3_close_Qs}). We next use \autoref{clm:c3_close_Qs} to demonstrate that the optimal action in the abstract MDP is nearly optimal in the ground MDP  (\autoref{clm:optAbsActionNearOptGround}). Lastly, \autoref{clm:lmaFromClm} shows that \autoref{lem:c3_phi_Q} follows from \autoref{clm:optAbsActionNearOptGround}.

\begin{claim}
\label{clm:c3_close_Qs}
Optimal $Q$-values in the abstract MDP closely resemble optimal $Q$-values in the ground MDP:
\begin{equation}
\label{eq:Q*Claim1}
\forall_{s \in \mc{S}, a \in \mc{A}} : |Q^*(s, a) - Q_{\phi}^*(\phi_{Q^*_\eps}(s), a)| \leq \frac{\varepsilon}{1-\gamma}.
\end{equation}
\end{claim}

Consider a non-Markovian decision process of the same form as an MDP, $M_N =\langle \mc{S}_N, \mc{A}, R_N, T_N, \gamma, \rhoz \rangle$, parameterized by non-negative integer an $N \in \mathbb{Z}_{\geq 0}$, such that for the first $N$ time steps the reward function, the transition function, and state space are those of the abstract MDP, $M_\phi$, and after $N$ time steps the reward function, transition dynamics and state spaces are those of $M$. Thus,
\begin{align}
\mc{S}_N &= \begin{cases}
\mc{S}& \text{if } N =0 \\
\mc{S}_\phi& \text{o/w}
\end{cases} \\
R_N(s,a) &= \begin{cases}
R(s,a)& \text{if } N =0\\
R_{\phi}(s, a)& \text{o/w}
\end{cases}\\
T_N(s' \mid s,a) &= \begin{cases}
T_{\phi}(s' \mid s,a)& \text{if }N =0\\
\underset{{g \in G(s)}}{\sum} T_{\phi}(s' \mid g, a) w(g)& \text{if } N =1\\
T_\phi(s' \mid s,a)& \text{o/w}
\end{cases}
\end{align}
The $Q$-value of state $s$ in $\mc{S}_N$ for action $a$ is:
\begin{equation}
Q_N^*(s, a) = 
\begin{cases}
       Q^*(s, a) &  \text{if } N=0\\
       \underset{g \in G(s)}{\sum} Q^*(g,a) w(g)& \text{if } N =1\\
       R_{\phi}(s,a) + \gamma \underset{{s_{\phi}}' \in \mc{S}_\phi}{\sum} T_{\phi}(s_{\phi} \mid s,a) \max_{a'} Q_{N-1}^*({s_{\phi}}', a'). &\text{o/w}
\end{cases}
\end{equation}
We proceed by induction on $N$ to show that:
\begin{equation}
\label{eq:clm1Induct}
\forall_{N, s \in \mc{S}, a} |Q_N^*(s_N, a) - Q^*(s, a) | \leq \sum_{n=0}^{N-1} \varepsilon \gamma^{n},
\end{equation}
where $s_N =s$ if $N=0$ and $s_N =\phi_{Q^*_\eps}(s)$ otherwise. \\

\textbf{\textit{(Base Case: $\bs{N = 0}$)}} \\

\noindent When $N =0$, $Q_N^* =Q^*$, so this base case trivially follows. \\

\textbf{\textit{(Base Case: $\bs{N = 1}$)}} \\

\noindent By definition of $Q_N^*$, for any $s,a$,
\begin{align}
&Q_1^*(s,a) = \underset{g \in G(s)}{\sum} \left[ Q^*(g,a) w(g) \right].
\end{align}
Since all co-aggregated states have $Q$-values within $\varepsilon$ of one another and $w(g)$ induces a convex combination,
\begin{align}
&Q_1^*(s_N,a) \leq \varepsilon \gamma^n + \varepsilon + Q^*(s, a) \\
\therefore& \left| Q_{1}^*(s_N, a) - Q^*(s,a) \right| \leq \sum_{n=0}^{1}\varepsilon \gamma^n.
\end{align}
\textbf{\textit{(Inductive Case: $\bs{N > 1}$)}} \\

\noindent We assume as our inductive hypothesis that:
\begin{equation}
\forall_{s \in \mc{S}, a} |Q_{N-1}^*(s_N, a) - Q^*(s, a) | \leq \sum_{n=0}^{N-2} \varepsilon \gamma^n.
\end{equation}

\noindent Consider a fixed but arbitrary state, $s \in \mc{S}$, and fixed but arbitrary action $a$.
Since $N > 1$, $s_N$ is $\phi_{Q^*_\eps}(s)$.
By definition of $Q_{T}^*(s_N, a)$, $R_{\phi}$, $T_{\phi}$:
\[
Q_N^*(s_N, a) = \sum_{s \in G(s_N)}w(g)\ \times \left[ R(s,a) + \gamma \sum_{s' \in \mc{S}} T_{\phi}(s' \mid s,a) \max_{a'} Q_{N-1}^*(s', a')      \right].
\]
Applying our inductive hypothesis yields:
\[
Q_N^*(s_N, a) \leq \sum_{s \in G(s_N)}w(g) \times \biggl[ R(s,a)\ + \gamma \sum_{s' \in \mc{S}} T(s' \mid s,a) \max_{a'}(Q^*(s', a') + \sum_{n=0}^{N-2} \varepsilon \gamma^n) \biggr].
\]
Then,
\begin{equation}
Q_N^*(s_N, a) \leq \gamma\sum_{n=0}^{N-2} \varepsilon \gamma^n + \sum_{s \in G(s_N)}\left[ w(s)\ Q^*(s,a)\right].
\end{equation}
Since all aggregated states have $Q$-values within $\varepsilon$ of one another:
\begin{align}
&Q_N^*(s_N, a) \leq \gamma\sum_{n=0}^{N-2} \varepsilon \gamma^n + \varepsilon + Q^*(s, a), \\
\therefore\ &\left| Q_{N}^*(s_N, a) - Q^*(s,a) \right| \leq \gamma\sum_{n=0}^{N-1}\varepsilon \gamma^n.
\end{align}
Since $s$ is arbitrary we conclude Equation \ref{eq:clm1Induct}. As $N \rightarrow \infty$, $\sum_{n=0}^{N-1} \varepsilon \gamma^n \rightarrow \frac{\varepsilon}{1-\gamma}$ by the sum of infinite geometric series and $Q_N^* \rightarrow Q_{\phi}^*$. Thus, \autoref{eq:clm1Induct} yields \autoref{clm:c3_close_Qs}.

\begin{claim}
\label{clm:optAbsActionNearOptGround}

Consider a fixed but arbitrary state, $s \in \mc{S}$ and its corresponding abstract state $s_{\phi}=\phi_{Q^*\eps}(s)$.
Let $a^*$ stand for the optimal action in $s$, and $a^*_{\phi}$ stand for the optimal action in $s_{\phi}$:
\begin{align}
a^* = \argmax_a Q^*(s, a), \hspace{4mm}
a^*_{\phi} = \argmax_a Q_{\phi}^*(s_{\phi}, a).
\end{align}
The optimal action in the abstract MDP has a $Q$-value in the ground MDP that is nearly optimal:
\begin{equation}
\label{eq:Q*Claim2}
V^*(s) \leq Q^*(s, a^*_{\phi}) + \frac{2\eps}{1-\gamma}.
\end{equation}
\end{claim}
\noindent By \autoref{clm:c3_close_Qs},
\begin{align}
&V^*(s) = Q^*(s, a^*) \leq Q_{\phi}^*(s_{\phi}, a^*) + \frac{\varepsilon}{1-\gamma}.
\label{eq:Q*OptActionResult}
\end{align}
By the definition of $a^*_{\phi}$, we know that 
\begin{align}
Q_{\phi}^*(s_{\phi}, a^*) + \frac{\varepsilon}{1-\gamma} \leq Q_{\phi}^*(s_{\phi}, a^*_{\phi}) + \frac{\varepsilon}{1-\gamma}.
\end{align}
Lastly, again by \autoref{clm:c3_close_Qs}, we know
\begin{align}
Q_{\phi}^*(s_{\phi}, a^*_{\phi}) + \frac{\varepsilon}{1-\gamma} \leq Q^*(s, a^*_{\phi}) + \frac{2\varepsilon}{1-\gamma}.
\end{align}
Therefore, \autoref{eq:Q*Claim2} follows.

\begin{claim}
\autoref{lem:c3_phi_Q} follows from \autoref{clm:optAbsActionNearOptGround}.
\label{clm:lmaFromClm}
\end{claim}

\noindent Consider the policy for $M$ of following the optimal abstract policy $\pi^*_\phi$ for $t$ steps and then following the optimal ground policy $\pi^*$ in $M$:
\begin{equation}
\pi_{\phi,n}(s)=
\begin{cases}
\pi^*(s) &\text{if } n= 0\\
\pi_{\phi}(s) &\text{if } n > 0
\end{cases}
\end{equation}

\noindent For $n > 0$, the value of this policy for $s \in \mc{S}$ in the ground MDP is:
\[
V^{\pi_{\phi,n}}(s) = R(s, \pi_{\phi,n}(s)) +\ \gamma \sum_{{s}' \in \mc{S}}T_{\phi}(s, a, {s}')V^{\pi_{\phi,n-1}}({s}').
\]

\noindent For $n=0$, $V^{\pi_{\phi,n}}(s)$ is simply $V^*(s)$.

\noindent We now show by induction on $t$ that
\begin{equation}
\forall_{n \in \mathbb{Z}_{\geq 0}, s \in \mc{S}} : V^*(s) \leq  V^{\pi_{\phi,n}}(s) + \sum_{i=0}^{n}\gamma^i \frac{2\varepsilon}{1-\gamma}.
\end{equation}

\textbf{\textit{(Base Case: $\bs{n=0}$)}} \\

\noindent By definition, when $n=0$, $V^{\pi_{\phi,n}} = V$, so our bound trivially holds in this case. \\

\textbf{\textit{(Inductive Case: $\bs{n > 0}$)}} \\

\noindent Consider a fixed but arbitrary state $s \in \mc{S}$.
We assume for our inductive hypothesis that
\begin{equation}
V^*(s) \leq V^{\pi_{\phi,n-1}}(s) + \sum_{i=0}^{n-1}\gamma^i \frac{2\varepsilon}{1-\gamma}.
\end{equation}
By definition,
\[ 
V^{\pi_{\phi,n}}(s) = R(s, \pi_{\phi,n}(s)) + \gamma \sum_{s' \in \mc{S}}T(s' \mid s, a)V^{\pi_{\phi,n-1}}({s}').
\]
Applying our inductive hypothesis yields:
\[
V^{\pi_{\phi,n}}(s) \geq R(s, \pi_{\phi,n}(s)) + \gamma \sum_{{s}' \in \mc{S}}T(s' \mid s, \pi_{\phi,n}(s))\left(V^*({s}') - \sum_{i=0}^{n-1}\gamma^i \frac{2\varepsilon}{1-\gamma} \right).
\]
Therefore,
\begin{align}
V^{\pi_{\phi,n}}(s) &\geq -\gamma\sum_{i=0}^{n-1}\gamma^i \frac{2\varepsilon}{1-\gamma} + Q^*(s, \pi_{\phi,n} (s)).
\end{align}
Applying \autoref{clm:optAbsActionNearOptGround} yields:
\begin{align}
&V^{\pi_{\phi,n}}(s) \geq -\gamma\sum_{i=0}^{n-1}\gamma^i \frac{2\varepsilon}{1-\gamma} - \frac{2\varepsilon}{1-\gamma} + V^*(s) \\
\therefore\ &V^*(s) \leq V^{\pi_{\phi,n}}(s)  + \sum_{i=0}^{n}\gamma^i \frac{2\varepsilon}{1-\gamma}.
\end{align}
Since $s$ was arbitrary, we conclude that our bound holds for all states in $\mc{S}$ for the inductive case.
Thus, from our base case and induction, we conclude that
\begin{equation}
\forall_{n \in \mathbb{N}, s \in \mc{S}}: V^{\pi^*}(s) \leq  V^{\pi_{\phi,n}}(s) + \sum_{i=0}^{n}\gamma^i \frac{2\varepsilon}{1-\gamma}.
\end{equation}

\noindent Note that as $n \rightarrow \infty$, $\sum_{i=0}^{n}\gamma^i \frac{2\varepsilon}{1-\gamma} \rightarrow \frac{2\varepsilon}{(1-\gamma)^2}$ by the sum of infinite geometric series and $\pi_{\phi,n} \rightarrow \pi_{\phi}$.
Thus, we conclude Lemma~\ref{lem:c3_phi_Q}. \qedhere
\end{dproof}

%% file: proofs/c3/c3_approx_abstr_lemma_2.tex
\begin{dproof}[Lemma~\ref{lem:c3_phi_model}]

\noindent Let $B$ be the maximum $Q$-value difference between any pair of ground states in the same abstract state for $\phi_{model,\eps}$:
\begin{equation*}
B = \max_{s_1, s_2, a}  |Q^*(s_1, a) - Q^*(s_2, a)|,
\end{equation*}
where $s_1, s_2 \in s_{\phi}$. First, we expand:
\begin{equation}
B=\max_{s_1, s_2, a}      \biggl|R(s_1, a) - R(s_2, a)\ + \gamma \sum_{{s}' \in \mc{S}} \biggl[(T_{\phi}(s' \mid s_1,a) -T_{\phi}(s' \mid s_2, a))\max_{a'}Q^*({s}', a')\biggr]\biggr|
\end{equation}
Since difference of rewards is bounded by $\eps_R$:
\begin{equation}
B\leq \max_{s_1,s_2,a} \left|\eps_R + \gamma \sum_{s_{\phi} \in \mc{S}_\phi}\sum_{{s}' \in s_{\phi}} \biggl[(T(s' \mid s_1, a)\ - T(s' \mid s_2, a)) \max_{a'}Q^*({s}', a') \biggr]\right|.
\end{equation}
By similarity of transitions under $\phi_{model,\eps}$:
\begin{equation}
B \leq \eps_R + \gamma \textsc{QMax} \sum_{s_{\phi} \in \mc{S}_\phi} \eps_T \leq \eps_R + \gamma|\mc{S}|\eps_T \textsc{QMax}.
\end{equation}
Recall that \textsc{QMax} $\leq \frac{\textsc{RMax}}{1-\gamma}$. Hence:
\begin{equation}
B \leq \textsc{RMax}\frac{\eps_R + \gamma(|\mc{S}| - 1) \eps_T}{1-\gamma}.
\end{equation}
Since the $Q$-values of ground states grouped under $\phi_{model,\eps}$ are strictly less than $B$, we can understand $\phi_{model,\eps}$ as a type of $\phi_{Q^*,B}$. Applying \autoref{lem:c3_phi_Q} yields \autoref{lem:c3_phi_model}.
\qedhere
\end{dproof}

%% file: proofs/c3/c3_approx_abstr_lemma_3.tex
\begin{dproof}[Lemma \ref{lem:c3_phi_bolt}]

\noindent To prove the result, we make use of the approximation for $e^x$, with $\delta$ error:
\begin{equation}
 e^x = 1 + x + \delta  \approx 1 + x.
\label{eq:e_to_x_approx}
\end{equation}
We let $\delta_1$ denote the error in approximating $e^{Q^*(s_1,a)}$ and $\delta_2$ denote the error in approximating $e^{Q^*(s_2,a)}$. \\

\noindent By the approximation in Equation~\ref{eq:e_to_x_approx} and the assumption in Equation~\ref{eq:c3_bolt_denom}:
\begin{align}
\left|\frac{1 + Q^*(s_1,a) + \delta_1}{\sum_j e^{Q^*(s_1,a_j)}} - \frac{1 + Q^*(s_2,a) + \delta_2}{\sum_j e^{Q^*(s_1,a_j)} \underbrace{\pm k\varepsilon}_{\circled{a}}}\right| \leq \varepsilon \label{eq:bolt_with_approx}
\end{align}
Either term \circled{a} is positive or negative. First suppose the former. It follows by algebra that:
\begin{equation}
-\varepsilon \leq \frac{1 + Q^*(s_1,a) + \delta_1}{\sum_j e^{Q^*(s_1,a_j)}} - \frac{1 + Q^*(s_2,a) + \delta_2}{\sum_j e^{Q^*(s_1,a_j)} + \varepsilon k_{\text{bolt}} } \leq \varepsilon
\end{equation}
Moving terms:
\begin{multline}
-\varepsilon \left(k_{\text{bolt}}\varepsilon + \sum_j e^{Q^*(s_1,a_j)}\right) - \delta_1 + \delta_2 \leq \\
\varepsilon k_{\text{bolt}} \left(\frac{1+Q^*(s_1,a) + \delta_1}{\sum_j e^{Q^*(s_1,a_j)}}\right) + Q^*(s_1,a) - Q^*(s_2,a) \leq \\
\varepsilon \left(\varepsilon k_{\text{bolt}}  + \sum_j e^{Q^*(s_1,a_j)}\right) - \delta_1 + \delta_2
\label{eq:a_p_case}
\end{multline}
When \circled{a} is the negative case, it follows that:
\begin{equation}
-\varepsilon \leq \frac{1 + Q^*(s_1,a) + \delta_1}{\sum_j e^{Q^*(s_1,a_j)}} - \frac{1 + Q^*(s_2,a) + \delta_2}{\sum_j e^{Q^*(s_1,a_j)} - \varepsilon k_{\text{bolt}} } \leq \varepsilon
\end{equation}

\noindent By similar algebra that yielded Equation~\ref{eq:a_p_case}:
\begin{multline}
-\varepsilon \left(-\varepsilon k_{\text{bolt}}  + \sum_j e^{Q^*(s_1,a_j)}\right) - \delta_1 + \delta_2 \leq \\
-k_{\text{bolt}}\varepsilon\left(\frac{1+Q^*(s_1,a) + \delta_1}{\sum_j e^{Q^*(s_1,a_j)}}\right) + Q^*(s_1,a) - Q^*(s_2,a) \leq \\
\varepsilon \left(\varepsilon k_{\text{bolt}}  + \sum_j e^{Q^*(s_1,a_j)}\right) - \delta_1 + \delta_2
\label{eq:a_m_case}
\end{multline}

\noindent Combining Equation~\ref{eq:a_p_case} and Equation~\ref{eq:a_m_case} results in:
\begin{equation}
\left|Q^*(s_1,a) - Q^*(s_2,a)\right| \leq \varepsilon \left(\frac{|\mc{A}|}{1-\gamma} + \varepsilon k_{\text{bolt}}  + k_{\text{bolt}}  \right).
\label{eq:bolt_qs}
\end{equation}
Consequently, we can consider $\phi_{bolt,\eps}$ as a special case of the $\phi_{Q^*_\eps}$ type, with $Q^*$ similarity of
\begin{equation}
B = \varepsilon \left(\frac{|\mc{A}|}{1-\gamma} + \varepsilon k_{\text{bolt}}  + k_{\text{bolt}}  \right).
\end{equation}
Lemma~\ref{lem:c3_phi_bolt} then follows from Lemma~\ref{lem:c3_phi_Q}.
\qedhere
\end{dproof}

%% file: proofs/c3/c3_approx_abstr_lemma_4.tex
\begin{dproof}[Lemma~\ref{lem:c3_phi_mult}]

\noindent The proof follows an identical strategy to that of Lemma~\ref{lem:c3_phi_bolt}, but without the approximation $e^x \approx 1+x$. \qedhere
\end{dproof}

%% file: chapters/c4_sa_lifelong_state_abstr.tex
\begin{center}
\begin{minipage}{0.8\textwidth}
\textit{This chapter is based on `State Abstractions for Lifelong Reinforcement Learning" \cite{abel2018salrl}, joint with Dilip Arumugam, Lucas Lehnert, and Michael L. Littman.}
\end{minipage}
\end{center}
\vspace{2mm}

\begin{figure}
    \centering
    \includegraphics[height=\figvdim]{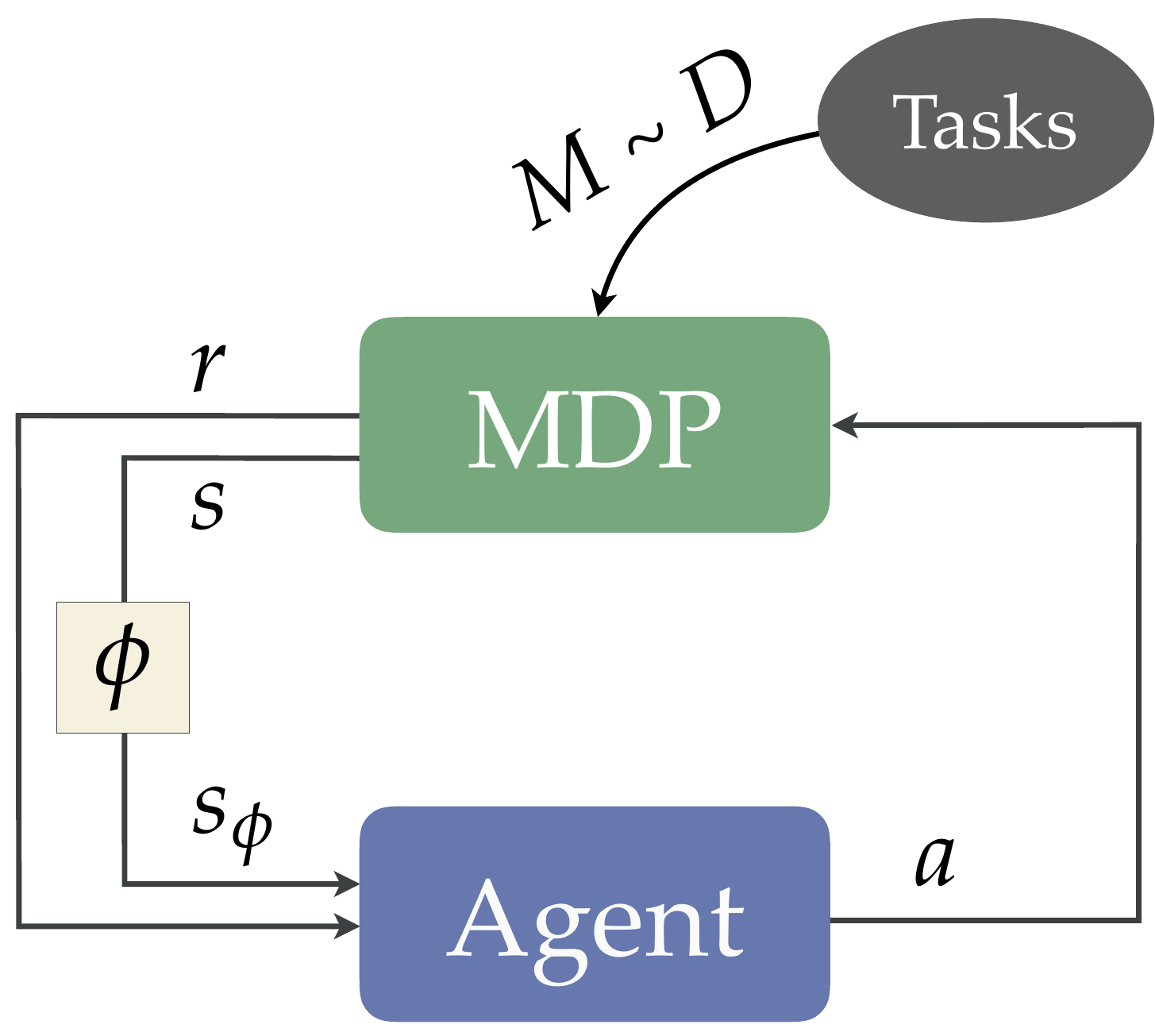}
    \caption{Lifelong Reinforcement Learning with State Abstraction.}
    \label{fig:c4_rl_lifelong_sa}
\end{figure}

In the previous chapter, I motivated the family of \textit{approximate} state abstraction, which can preserve representation of good behavior without requiring a perfect solution to the MDP of interest. In this chapter, I extend these results to the case where an agent is presented with a continuous stream of tasks to solve. That is, I study state abstraction in the context of \textit{lifelong} RL.

Indeed, a long standing goal of AI is to understand how autonomous agents can accumulate and make use of knowledge across a variety of related tasks, or perhaps from a continual stream of experience. This setting (and its kin) has appeared under a variety of names including multitask learning, lifelong learning, transfer learning, or continual learning. The premise of each of these settings is to agents that must interact with and solve many different tasks over the course of a lifetime, as studied by~\citet{thrun1996learning,wilson2007multi,isele2016using,walsh2006transferring} and~\citet{wilson2007multi}. There are important nuances that separate each of these specific settings, but the spirit is largely the same. I will henceforth use the term \textit{lifelong} learning to capture the many variations of this setting.

Lifelong \textit{reinforcement learning} presents a particularly difficult set of challenges as it forces agents not only to generalize within an MDP, but also across MDPs. Drawing from prior literature, I offer the following definition of the lifelong RL setting.
\ddef{Lifelong  reinforcement learning}
{In \textbf{lifelong reinforcement learning}, the agent receives $\mc{S}, \mc{A}, \rhoz$, horizon $H$, discount factor $\gamma$, and query access to a fixed but unknown probability distribution over reward-transition function pairs, $D$. The agent samples $(R_i, T_i) \sim D$, and interacts with the MDP $(\mc{S}, \mc{A}, R_i, T_i, \gamma, \rhoz)$ for $H$ time steps, starting in state $s_0 \sim \rhoz$. After $H$ time steps, the agent resamples from $D$ and repeats.
}

The tools of abstraction are particularly well-suited to assist in lifelong RL, as state abstraction can capture relevant task structure across MDPs that can aid in information transfer and accelerate learning. Equipped with the right state abstraction, then, an agent might be able to learn in an extremely sample efficient manner, even on unseen tasks. It is precisely this insight that I investigate in this chapter. For more motivation and background on lifelong RL, see work by~\citet{brunskill2015online}. 

\section{Transitive PAC State Abstractions}

Concretely, I propose two new classes of state abstraction that together are easy to compute in the lifelong setting. That is, the joint family of 1) transitive, and 2) PAC state abstractions are efficient to compute, can be estimated from a finite number of sampled and solved tasks, and preserve expected near-optimal behavior in lifelong RL. These are the first state abstractions to satisfy this collection of properties. I close the chapter with a negative result, however: PAC-MDP algorithms~\cite{Strehl2009} such as R-Max~\cite{brafman2002rmax} are not guaranteed to interact effectively with an abstracted MDP, suggesting that additional work is needed to leverage this idea to yield sample efficient learning. Finally, I conduct several simple experiments with standard algorithms to empirically corroborate the effects of state abstraction on lifelong RL.

I will make use of two PAC-MDP algorithms from prior literature: R-Max~\cite{brafman2002rmax} and Delayed $Q$-learning~\cite{strehl2006pac}. As discussed in \autoref{chap:background}, a PAC-MDP algorithm comes with a guarantee that it will only make a polynomial number of mistakes with high probability, thereby ensuring that it explores the MDP efficiently. Delayed $Q$-learning is a model free algorithm that also makes heavy use of optimism to inform its exploration strategy.

In general, computing the approximate state abstraction that induces the smallest possible abstract state space for that predicate is known to be NP-Hard~\cite{even2003approximate}. Indeed, this result limits the potential utility of state abstractions, as reducing the size of the abstract state space is the main goal of state abstraction; a smaller state space is desirable as the reduced MDP is (typically) easier to solve.

I first introduce \textit{transitive} state abstractions, a restricted class of approximate state abstractions that can be computed efficiently. Intuitively, transitive state abstractions induce an equivalence class on ground states; transitivity guarantees that the predicate $p$ associated with the type satisfies the implication $\left[p(s_1, s_2) \text{and}\ p(s_2, s_3)\right] \implies p(s_1,s_3)$. Many existing state-abstraction types are transitive. However, the \textit{approximate} abstraction types introduced in the previous chapter are not transitive. Thus, I next introduce a transitive modification of each of the approximate state-abstraction types.
\ddef{$\phi_{Q^*_d}$}{For a given $d \in [0,\textsc{VMax}]$, $\bs{\phi_{Q^*_d}}$ denotes a state-abstraction type with predicate:
\begin{equation}
    p_{M}^d(s_1, s_2) := \forall_{a \in \mc{A}}: \left\lceil \frac{Q_M^*(s_1, a)}{d}\right\rceil = \left\lceil \frac{Q_M^*(s_2, a)}{d}\right\rceil.
\end{equation}}

Intuitively, the abstraction discretizes the interval from $[0, \textsc{VMax}]$ by buckets of size $d$. Then, a pair of states satisfy the predicate if the $Q^*$-values for all actions fall in the same discrete buckets. Note that this predicate is transitive by the transitivity of being-in-the-same-bucket. As I will show in the next section, the above type affords representation near-optimal behavior as a function of $d$.

The second new family of state abstractions introduced are those that are suitable for application to a distribution of MDPs. The motivation for these state abstractions is to identify a mechanism for extending a state abstraction that preserves representation of good policies in one MDP to the case of many MDPs. In particular, I will later show that PAC abstractions ensure, with high probability over the task distribution $D$, that a high value policy is representable. The family is defined as follows, inspired by PAC learning~\cite{Valiant1984}.
\ddef{PAC State Abstraction}{$\phi_p^\delta$ is \textbf{a PAC state abstraction} belonging to type $\phi_p$ such that, for a given $\delta \in (0, 1]$, and a given distribution over MDPs $D$, the abstraction groups together nearly all state pairs for which the predicate $p$ holds with high probability over the distribution. More formally, for an arbitrary state pair $(s_1,s_2)$, let $\rho^p_x$ denote the predicate that is true if and only if $p$ is true over the distribution with probability $1-x$:
\begin{equation}
\rho_x^p(s_1, s_2) := \underset{M \sim D}{\PR}\{p_M(s_1,s_2) = 1\} \geq 1-x.
\end{equation}
Then $\phi_p^\delta$ is a PAC state abstraction if there exists an $\eps \in (-\delta, \delta)$ such that, for all $s_1, s_2$:
\begin{equation}
\PR \left\{ \rho^p_{\delta + \eps}(s_1, s_2) \equiv \phi_p^\delta(s_1) = \phi_p^\delta(s_2)\right\} \geq 1-\delta.
\end{equation}}

\section{Analysis}

I now present our main theoretical results on each of the two new abstraction types. These results summarize how to bring efficiently computable, value-preserving state abstractions into lifelong RL. I first analyze transitive abstraction, then PAC abstractions.

\subsection{Transitive State Abstractions}
\label{sec:c4_theory_q1}

I first show that transitive state abstractions can be computed efficiently.

\begin{theorem}
\label{thm:c4_sa_trans_comp}
Consider any \textit{transitive} predicate on state pairs, $p$, that takes computational complexity $c_p$ to evaluate for a given state pair. The state abstraction type $\phi_p$ that induces the smallest abstract state space can be computed in $\mc{O}(|\mc{S}|^2 \cdot c_p)$.
\end{theorem}

\input{proofs/c4/c4_transitive_predicate_comp.tex}

The intuition here is that we can avoid many computations by relying on transitivity. Any one query made of a state pair predicate can yield information about all state pairs in the equivalence class. Critically, the complexity of $c_p$ dictates the overall complexity of computing $\phi_p$.

Recall that most known approximate state-abstraction types are \textit{not} transitive (see \autoref{tab:c2_sa_table}, for instance). Hence, I next show that there exists an approximate state-abstraction type---with a transitive predicate---with bounded value loss:

\begin{theorem}
\label{thm:c4_sa_qdstar_vl}
The $\phi_{Q^*_d}$ abstraction type is a subclass of $\phi_{Q_\eps^*}$ introduced in the previous chapter, with $d = \eps$, and therefore, for a single MDP, the optimal abstract policy $\pi_\phi^*$ resulting from $\phi_{Q^*_d}$ ensures
\begin{equation}
    V^*(s_0) - V^{\pi_\phi^*}(s_0) \leq \frac{2d \textsc{RMax}}{(1-\gamma)^2}.
\end{equation}
\end{theorem}

\input{proofs/c4/c4_qd_star_val_loss}

Thus, the $\phi_{Q^*_d}$ class represents a reasonable candidate for state abstractions as it can be computed efficiently and posses a value loss that scales according to a free parameter, $d$. When $d=0$, the value loss is zero, and the abstraction collapses to the typical $\phi_{Q^*}$ irrelevance abstraction from~\citet{li2006towards}. Note that predicates defining other existing abstraction types, such as $\phi_{a^*}$~\cite{li2006towards}, also have natural translations to transitive predicates using the same discretization technique. While most of the main theoretical results are agnostic to choice of predicate, I concentrate on $Q$ based abstractions due to their simplicity and utility. Notably, none of these state abstractions require exact knowledge of $Q^*$: I always approximate it based on knowledge of prior tasks. Our results shed light on when it is possible to employ approximate knowledge of this kind for use in decision making.

Recall, however, that the primary goal of state abstraction is to reduce the size of the agent's representation over problems of interest. A natural question arises: if one were to solve the full NP-Hard problem of computing the maximally compressing state abstraction of a particular class, how much more compression can be achieved over the transitive approximation? Intuitively: Is the transitive abstraction going to compress the state space? The following result addresses this question.

\begin{theorem}
\label{thm:c4_sa_size_loss}
For a given $d$, the function belonging to the transitive abstraction type $\phi_{Q^*_d}$ that induces the smallest possible abstract state space size is at most $2^{|\mc{A}|}$ times larger than that of the maximally compressing instance of type $\phi_{Q_\eps^*}$, for $d=\eps$. Thus, letting $\mc{S}_d$ denote the abstract state space associated with the maximally compressing $\phi_{Q^*_d}$, and letting $\mc{S}_\eps$ denote the abstract state space associated with the maximally compressing $\phi_{Q_\eps^*}$,:
\begin{equation}
    |\mc{S}_\eps| \cdot 2^{|\mc{A}|} \geq |\mc{S}_d|.
\end{equation}
\end{theorem}

\input{proofs/c4/c4_sa_state_space_size.tex}

The above result shows that the non-transitive, maximally compressing state space size can in fact be quite smaller than the transitive approximation (by a factor of $2^{|\mc{A}|}$).

\subsection{PAC Abstractions}
\label{sec:theory_q2}

Next, I analyze PAC abstractions for the purpose of extending state abstractions to the lifelong setting. I first show that, for any abstraction type $\phi_p$, its PAC variant achieves bounded value loss (with high probability) as a function of the single task loss of $\phi_p$:
\begin{theorem}
\label{thm:c4_pac_val_loss}
Consider any state-abstraction type $\phi_p$ with value loss $\tau_p$. That is, in the traditional single task setting, letting $\pi^*_p$ denote $\pi^*_{\phi_p}$:
\begin{equation}
    \forall_{s \in \mc{S}} : V^*(s) - V^{\pi^*_p}(s) \leq \tau_p.
\end{equation}
Then, the PAC abstraction $\phi_p^\delta$, in the lifelong setting, induces a policy $\pi_{p,\delta}^*$ with expected value loss:
\begin{equation}
    \forall_{s \in \mc{S}} : \underset{M\sim D}{\bE}\left[V_M^*(s) - V_M^{\pi_{p,\delta}^*}(s)\right] \leq \eps (1-3\delta)\tau_p + 3\delta \textsc{VMax}.
\end{equation}
\end{theorem}

\input{proofs/c4/c4_lifelong_sa_val_loss.tex}

The value loss may be quite high, as up to $3\delta \textsc{VMax}$ value can be lost in the worst case. Accordingly, it is important to be cautious in selection of $\delta$. This bound is not tight, however, so in practice the value loss is likely to be lower.

Next, I show how to compute PAC abstractions from a finite number of sampled tasks.

\begin{theorem}
\label{thm:c4_pac_sa_sample}
Let $\mathscr{A}_p$ be an algorithm that given an MDP $M = ( \mc{S}, \mc{A}, R, T, \gamma )$ as input can determine if $p(s_1,s_2)$ is true for any pair of states, for any state abstraction type. Then, for a given $\delta \in (0,1]$ and $\eps \in (-\delta, \delta)$, it is possible to compute a PAC abstraction $\hat{\phi}_p^\delta$ after $m \geq \frac{\ln \left( \frac{2}{\delta} \right) }{\eps^2}$ sampled MDPs from $D$.
\end{theorem}

\input{proofs/c4/c4_pac_sample_bound.tex}

Note that this result assumes oracle access to the true predicate, $p(s_1,s_2)$, during the computation of $\hat{\phi}_p^\delta$. The analogous case in which $p$ can only be estimated via an agent's interaction with its environment is a natural next step for future work.

Given the ability to compute PAC abstractions from a finite number of samples, I now study the interplay between state abstractions and PAC-MDP algorithms for efficient RL.

\begin{theorem}
\label{thm:c4_rmax_broken}
Consider an MDP $M$ and an instance of R-Max~\cite{brafman2002rmax} that breaks ties using round-robin selection over actions. Now, consider R-Max paired with a state-abstraction function $\phi$ with value loss bounded by $\eps_\phi \in \mathbb{R}_{\geq 0}$. If R-Max interacts with $M$ by projecting any received state $s$ through $\phi$, then R-Max is no longer guaranteed to be PAC-MDP in $M$ (even relative to mistakes defined by $\eps_\phi$). In fact, the number of mistakes made by R-Max can be arbitrarily large.
\end{theorem}

\input{proofs/c4/c4_rmax_broken.tex}

The above result is a surprising \textit{negative} result---it suggests that there is more to the abstraction story than simply projecting states into the abstract. Specifically, it is indicative of future work that clarifies how to form abstractions that preserve the right kinds of guarantees.

\begin{figure}[t!]
    \centering
   {\includegraphics[height=\figvdim]{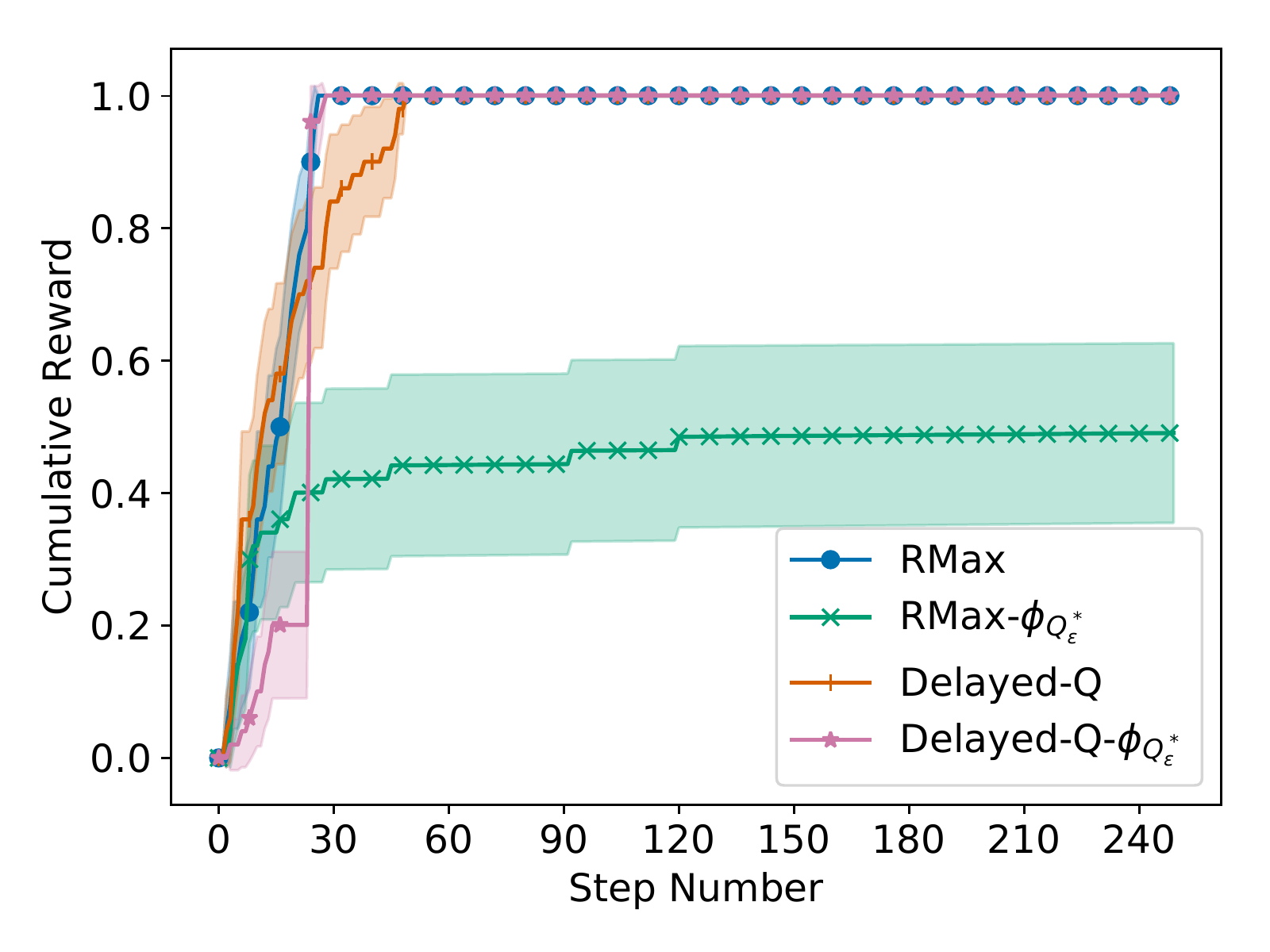}}
    \caption{Results averaged over 50 runs on the pathological three chain MDP introduced in the proof of \autoref{thm:c4_rmax_broken}.}
    \label{fig:c4_3_chain}
\end{figure}

To communicate this piece more directly, I conduct a simple experiment in the $3$-chain problem introduced in the proof of \autoref{thm:c4_rmax_broken}. Here I run R-Max and Delayed $Q$-learning with and without $\phi_{Q_\eps^*}$, with abstraction parameter $\eps=0.01$. Each agent is given 250 steps to interact with the MDP. The results are shown in \autoref{fig:c4_3_chain}. R-Max, paired with abstraction $\phi_{Q_\eps^*}$, fails to learn a anywhere close to a near-optimal policy. In fact, it is possible to control a parameter in the MDP such that R-Max performs arbitrarily bad. It remains an open question as to whether $\phi$ preserves the PAC-MDP property for Delayed $Q$~\cite{strehl2006pac}.

To highlight this point further, I next show that projecting an MDP to the abstract state space via $\phi$ and learning with $M_\phi$ is non-identical to learning with $M$ and projecting states through $\phi$:

\begin{corollary}
\label{corr:c4_m_phi_vs_learn_w_phi}
Consider any RL algorithm $\mathscr{A}$ whose policy updates during learning and a fixed choice of state abstraction, $\phi$. Let $\mathscr{A}_\phi$ denote the algorithm yielded by projecting all incoming states to $\phi(s)$ before presenting them to $\mathscr{A}$, and let $M_\phi = (\mc{S}_\phi, \mc{A}, T_\phi, R_\phi, \gamma, \rhoz^\phi)$, denote the abstract MDP induced by $\phi$ on $M$, with $w(s)$ the uniform weighting function.

There exists an MDP $M$ such that the process yielded by $\mathscr{A}_\phi$ interacting with $M$ is not identical to $\mathscr{A}$ interacting with $M_\phi$, even if $\pi_{\mathscr{A}_\phi}^t(s) = \pi_{\mathscr{A}}^t(s)$ for all $t$ and $s$. That is, there exists choice of $\phi$ and $M$ such that the expected trajectory of these two processes is non-identical. Formally, there exists a time step $t \in \mathbb{N}$, MDP $M$, and $\phi \in \phiall$ such that
\begin{equation}
    \underset{\mathscr{A}, s_0 \sim \rhoz^\phi}{\bE}\left[s_t \mid s_0\right] \neq \underset{\mathscr{A}_\phi, s_0 \sim \rhoz^\phi}{\bE}\left[s_t \mid s_0 \right],
\end{equation}
where $s_t$ is the state the agent arrives in after $t$ time steps.
\end{corollary}

\input{proofs/c4/c4_m_phi.tex}

These results illustrate a peculiarity to using state abstractions in RL: abstracting during interaction is distinct from offline abstraction. This result is reminiscent of Theorem 4 and Theorem 5 by~\citet{li2006towards} that describe the impact $\phi$ can have on convergence guarantees of well known RL algorithms. An important direction for future work is to provide a cohesive framework that preserves both PAC and convergence guarantees, whether the abstractions are used offline or online.

To summarize the analysis in this chapter: any state abstraction that belongs to both the transitive class \textit{and} the PAC class is: (1) efficient to compute, (2) can be estimated from a polynomial number of sampled and solved problems, (3) and preserves near-optimal behavior in the lifelong RL setting. The identification of such a class of desirable state abstractions for lifelong RL is the main contribution of this chapter, gesturing toward state abstractions that can trade off between all three desiderata. Alongside these positive results, I have highlighted shortcoming of state abstractions in the final two results, raising open questions about how to generalize state abstractions to work well with existing PAC-MDP algorithms.

I now move on to an empirical study evaluating the utility of these abstractions.

\section{Experiments}

I conduct two sets of simple experiments with the goal of illuminating how state abstractions of various forms impact learning and decision making. The code for running these experiments is publicly available.\footnote{\url{https://github.com/david-abel/rl_abstraction}}
\begin{itemize}
    
    \item \textit{Learning with and without $\phi_p^\delta$}: I investigate the impact of different types of abstractions on $Q$-Learning~\cite{watkins1992q} and Delayed $Q$-learning~\cite{strehl2006pac} in different lifelong RL task distributions.
    
    \item \textit{Planning with and without $\phi_p^\delta$}: Second, I explore the impact of planning via VI (\autoref{alg:value_iteration}) with and without a state abstraction, intended to be suggestive of the potential to accelerate model-based algorithms with good state abstractions.
\end{itemize}

In each case, I compute various types of $\phi_p^\delta$ according to the sample bound from \autoref{thm:c4_pac_sa_sample}, with $\delta = 0.1$, and the PAC parameter $\eps=0.1$ (the worst case $\eps$). I experiment with ($\phi_{Q^*}^\delta$), approximate ($\phi_{Q_\eps^*}^\delta$), and transitive ($\phi_{Q^*_d}^\delta$) state abstractions from the $Q$ similarity classes across each of the above algorithms. I experiment with probably approximate $Q$ based abstractions because their value loss bound is known, tight, and a small function of the approximation parameter, and (2) They have known transitive variants and are thus simple to compute, as shown in \autoref{thm:c4_sa_trans_comp}. Further, if a $Q^*$ based abstraction presents no opportunity to abstract (the reward or transition function change too dramatically across tasks), then \autoref{thm:c4_pac_sa_sample} indicates that $\phi$ will abstain from abstracting.


\subsubsection{Lifelong RL}

Each learning experiment proceeds as follows. For each agent, at time step zero, sample a reward function from the distribution. Then, let the agent interact with the resulting MDP for 100 episodes. When the last episode finishes, reset the agent to the start state $s_0$, and repeat. All learning curves are averaged over samples from the distribution. Thus, improvements to learning from each $\phi$ are improvements \textit{averaged over the task distribution}. In all learning plots we report 95\% confidence intervals over the sampling process (both samples from the distribution and runs of each agent in the resulting MDP).

\textbf{Color Room}: I first conduct experiments testing $Q$-learning and Delayed $Q$ learning on an $11\times11$ Four Rooms variant, adapted from~\citet{sutton1999between}. In this task distribution, goal states can appear in exactly one of the furthest corner of each of the three non-starting rooms (that is, there are three possible goal locations) uniformly at random. Transitions into a goal state yield $+1$ reward with all other transitions providing $+0$. Goal states are set to terminal. To explore the impact of abstraction, I augment the problem representation by introducing an irrelevant state feature: \textit{color}. Specifically, each cell in the grid can have a color \textit{red, blue, green}, or \textit{yellow}. All cells are initially \textit{red}. The agent is given another action, \texttt{paint}, that paints the entire set of cells to one of the four colors uniformly at random. No other action can change the color of a cell. The color has no impact on either reward or transitions, and so is fundamentally irrelevant in decision making. We are thus testing the hypothesis as to how effectively the sample based PAC abstractions can pick up on the irrelevant characteristics and still support efficient but high performance learning. Given the inherent structure of the Four Rooms domain we also experiment with an intuitively useful hand-coded state abstraction, $\phi_h$, that assigns an abstract state to each room for a total of four abstract states. The agents all start in the bottom left cell.

\begin{figure}[t!]
\centering
    \subfloat[$Q$-Learning, Color Room]{\includegraphics[scale=0.45]{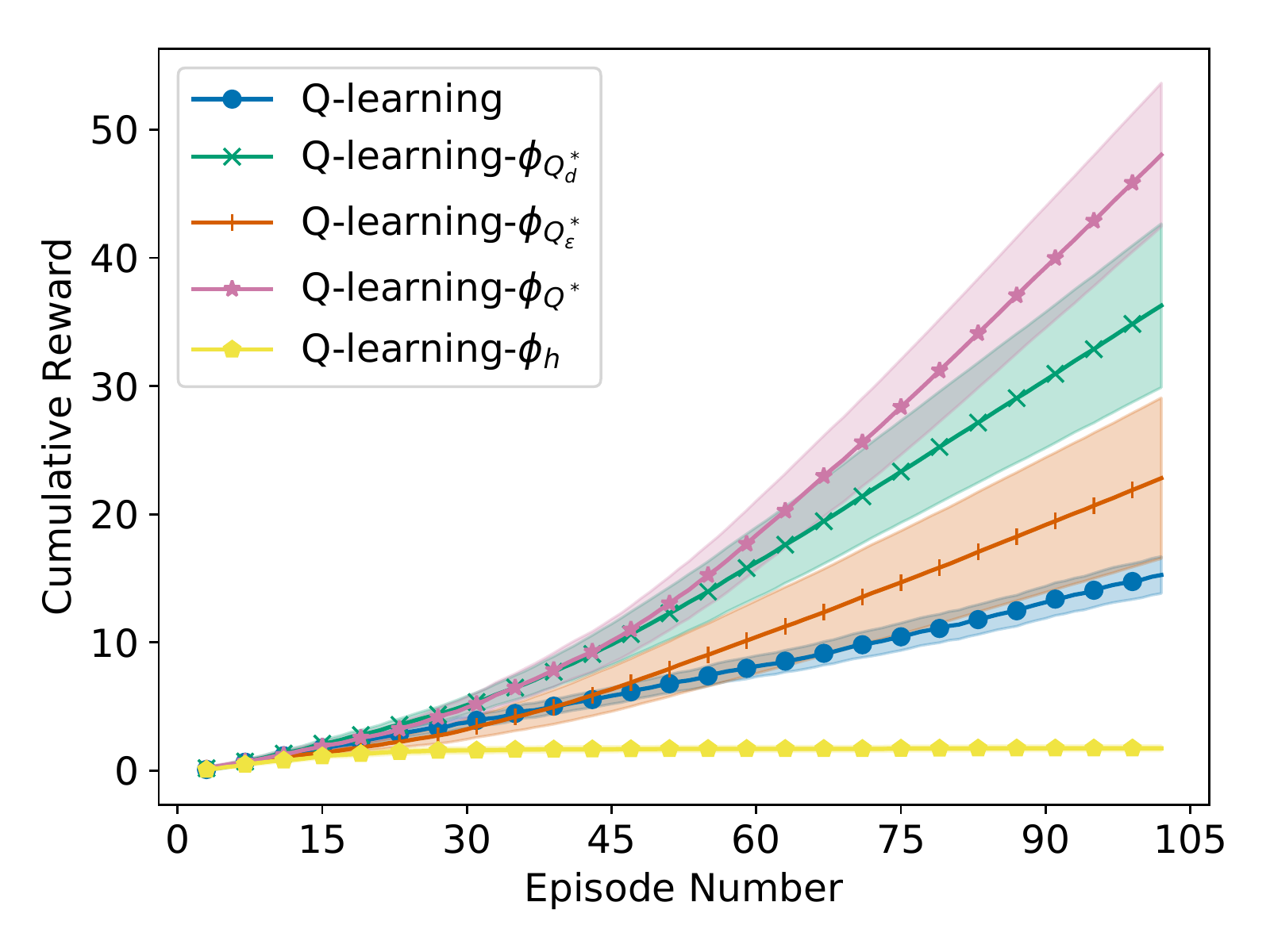}} \subfhspace
    \subfloat[Delayed $Q$, Color Room]{\includegraphics[scale=0.45]{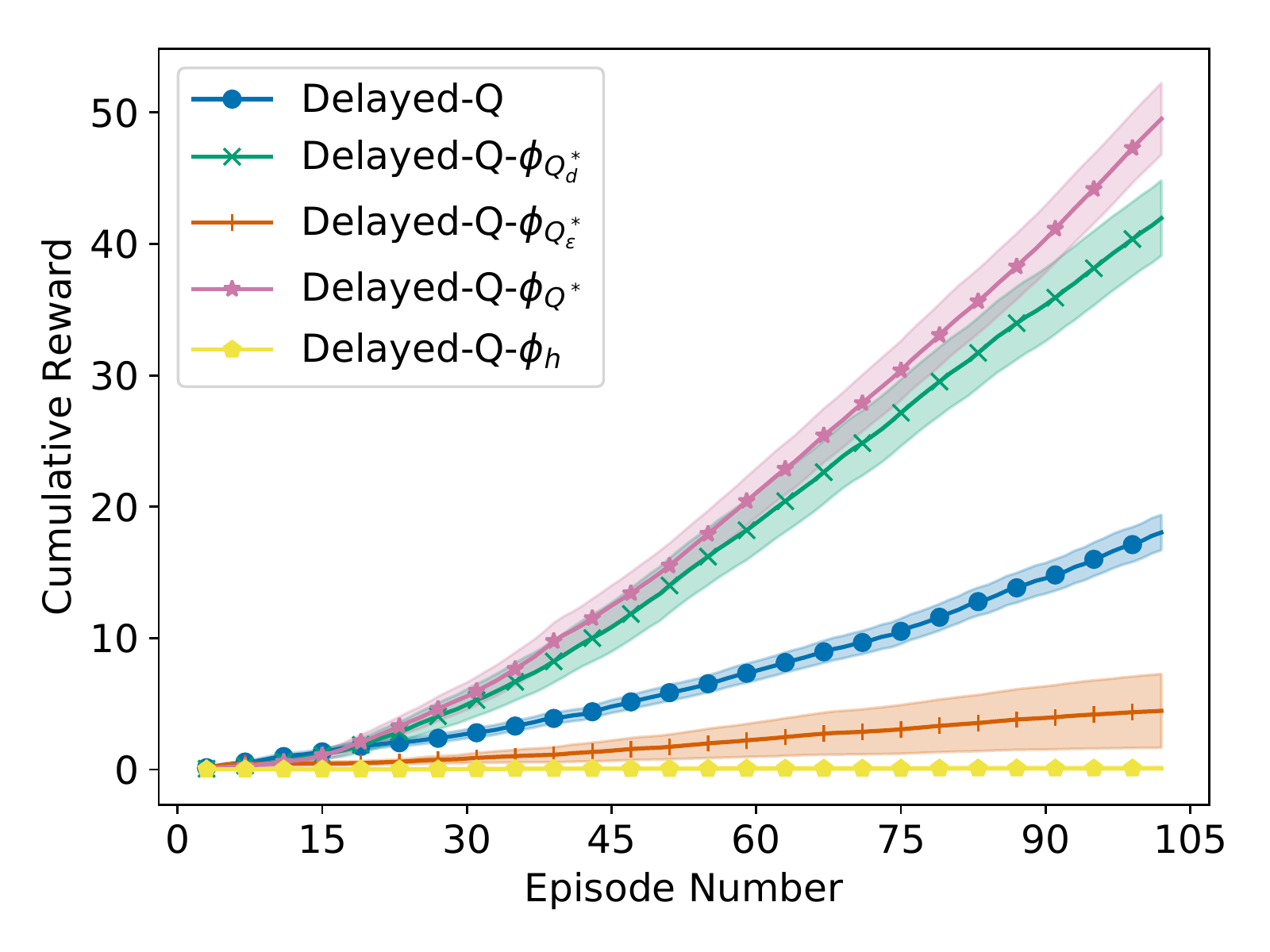}} \\
    
    \subfloat[$Q$-Learning, Upworld]{\includegraphics[scale=0.45]{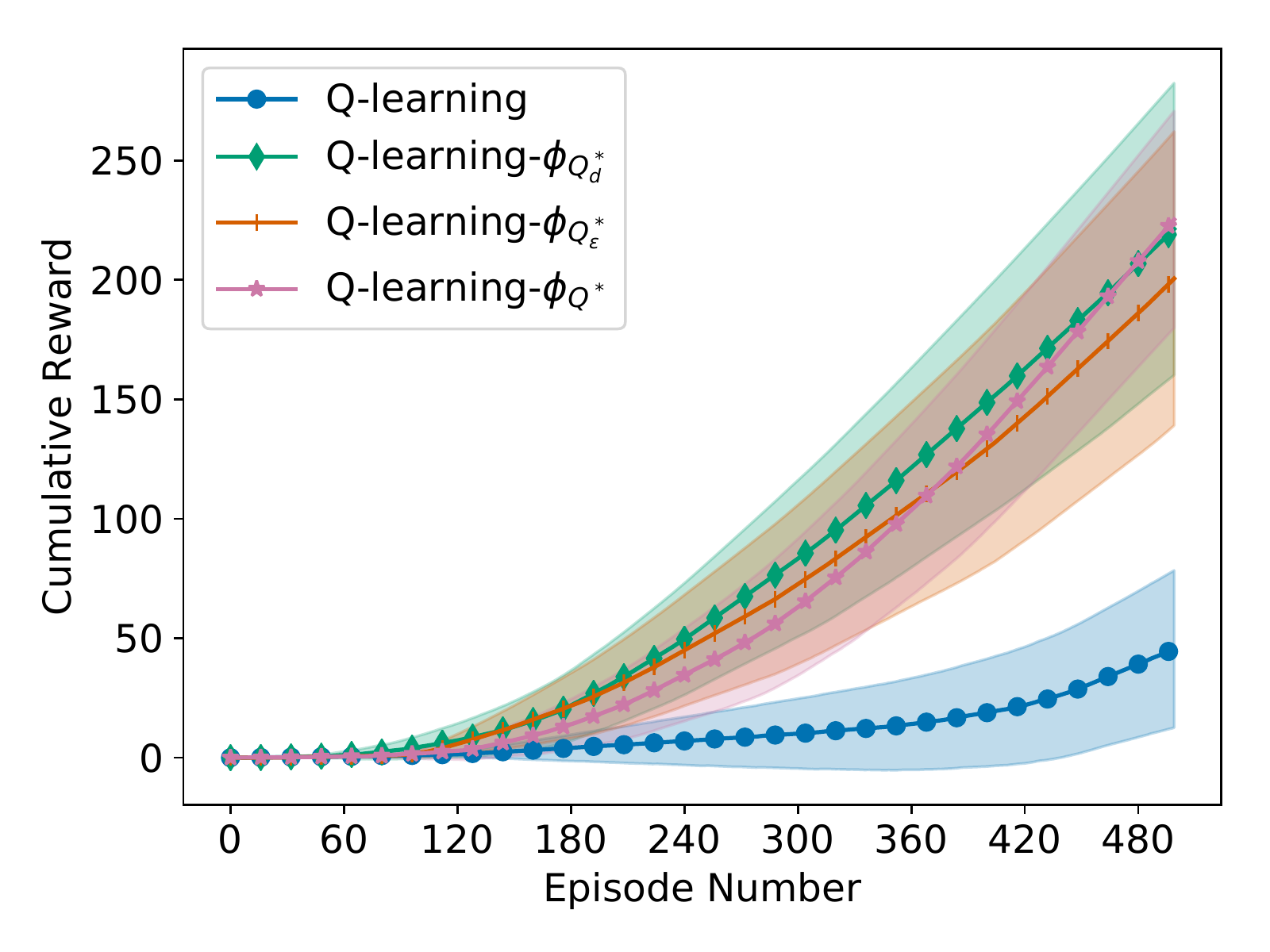}} \subfhspace
    \subfloat[Delayed $Q$, Upworld]{\includegraphics[scale=0.45]{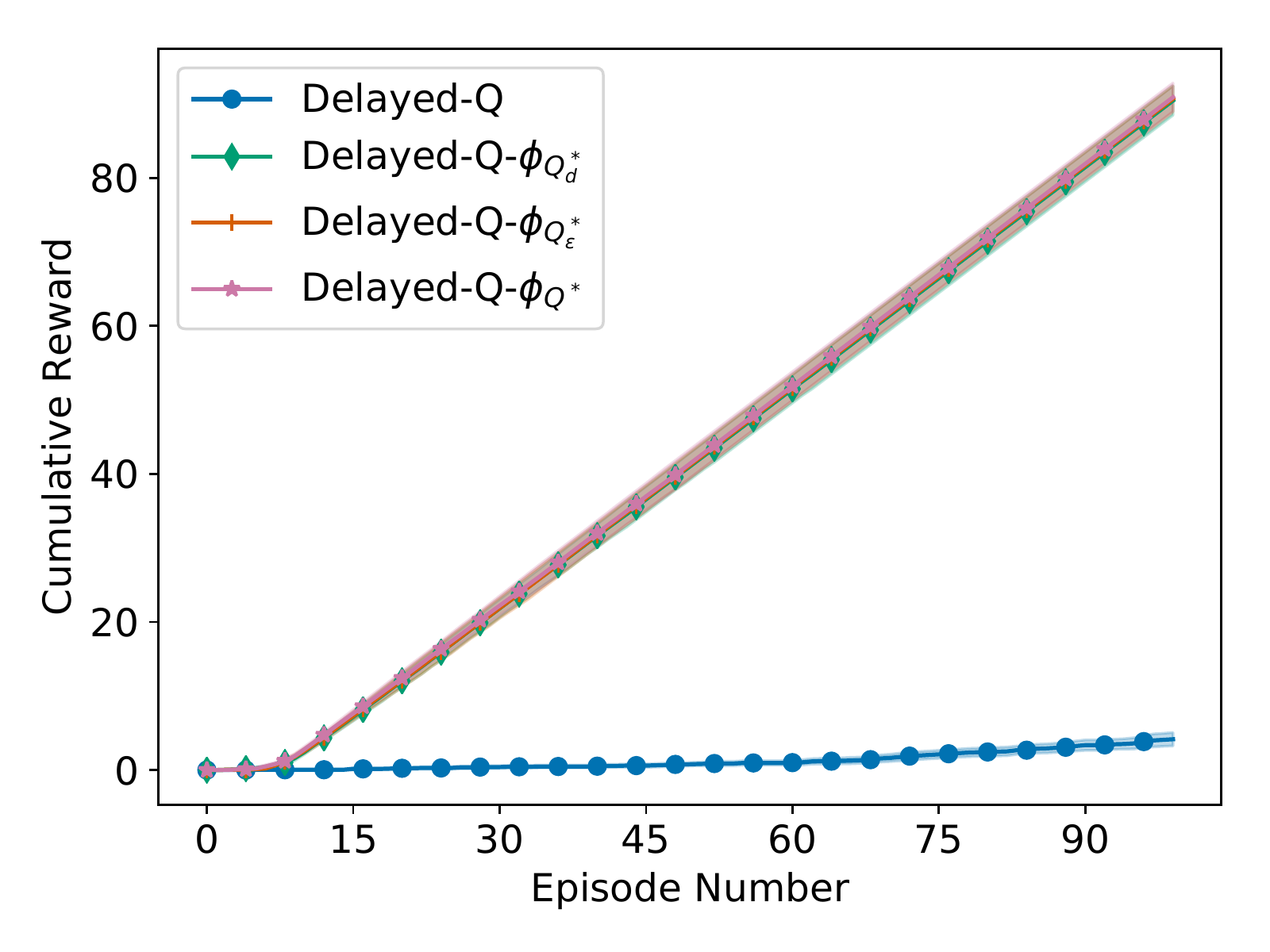}} \\
  
\caption{Cumulative reward averaged over 100 task samples from the Colored Four Rooms task distribution (top) and the Upworld task distribution (bottom).}
\label{fig:c4_color_results}
\end{figure}

The top row of \autoref{fig:c4_color_results} shows results for algorithms run on the Colored Four Rooms task distribution. For $Q$-learning the data suggest that all three PAC abstractions achieve improvement in mean cumulative reward, averaged across 100 task samples. Notably, the slope of the learning curves are similar as well, suggesting that the policies discovered after 100 episodes are comparable in value. Notably, the variants with the abstraction tend to find better policies more quickly. In the case of Delayed-$Q$, the new transitive PAC abstraction (green) finds even further improvement over the baseline algorithm, both in terms of learning speed and the value of the policy used near the end of learning.

The bottom row of \autoref{fig:c4_color_results} present results for the same learning set up with the Upworld MDP distribution. This domain is an extremely simple 30 $\times$ 11 grid world, where one of the 30 possible goals in the top row is active at any given time. The agent always starts in the bottom left corner. As expected, the data suggest that the improvement from the abstraction in this domain are dramatic, as there is great opportunity to abstract. The performance of both baseline algorithms is dominated by any of the approaches that use state abstraction.

\begin{figure}[!t]
    \centering
    \includegraphics[width=0.5\textwidth]{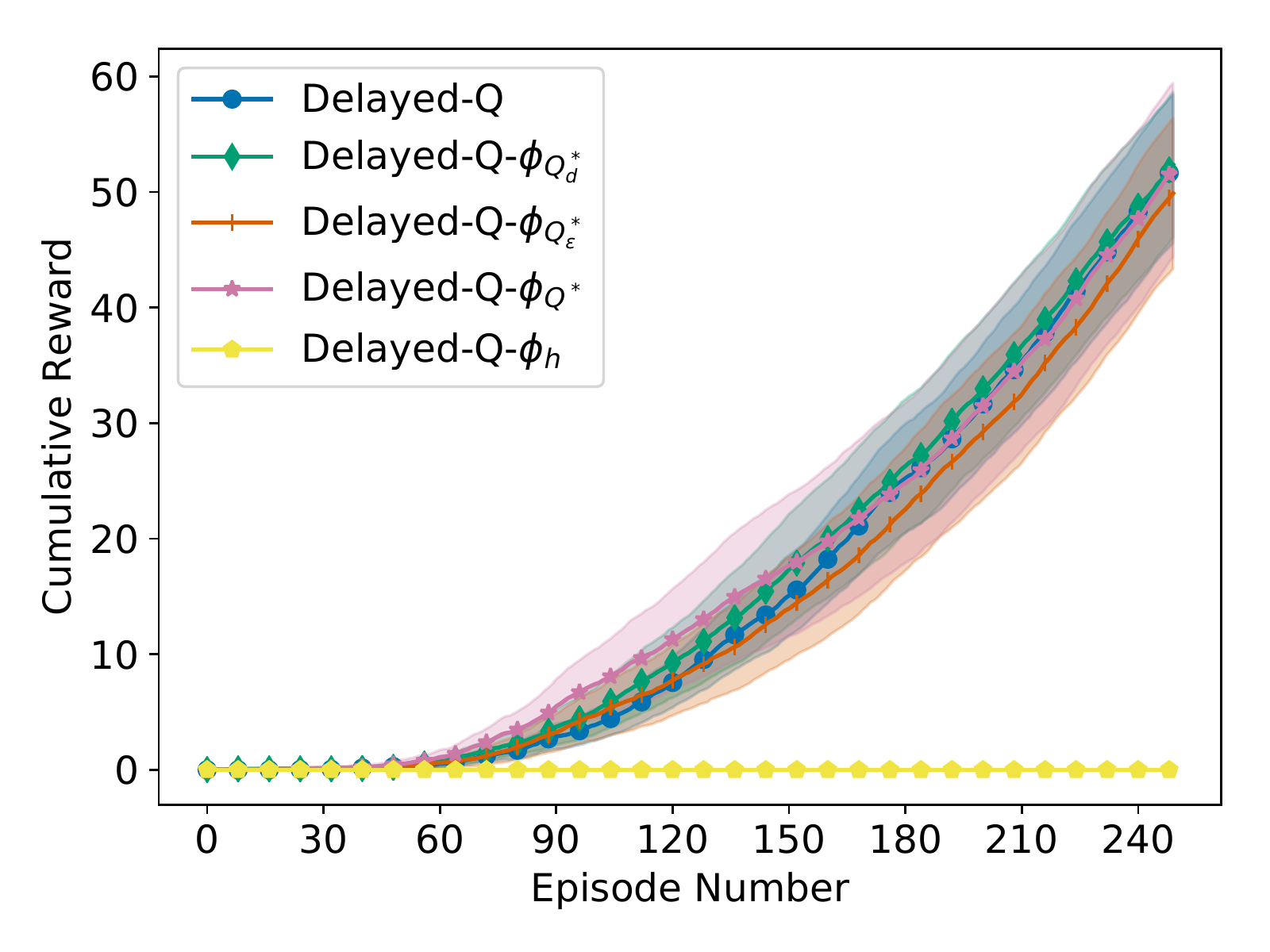}
    \caption{Delayed $Q$-learning on a 15 $\times$ 15 Four Rooms task distribution.}
    \label{fig:c4_big_four}
\end{figure}

\begin{figure}[h]
\centering
\subfloat[Planning in Upworld]{\includegraphics[scale=0.4]{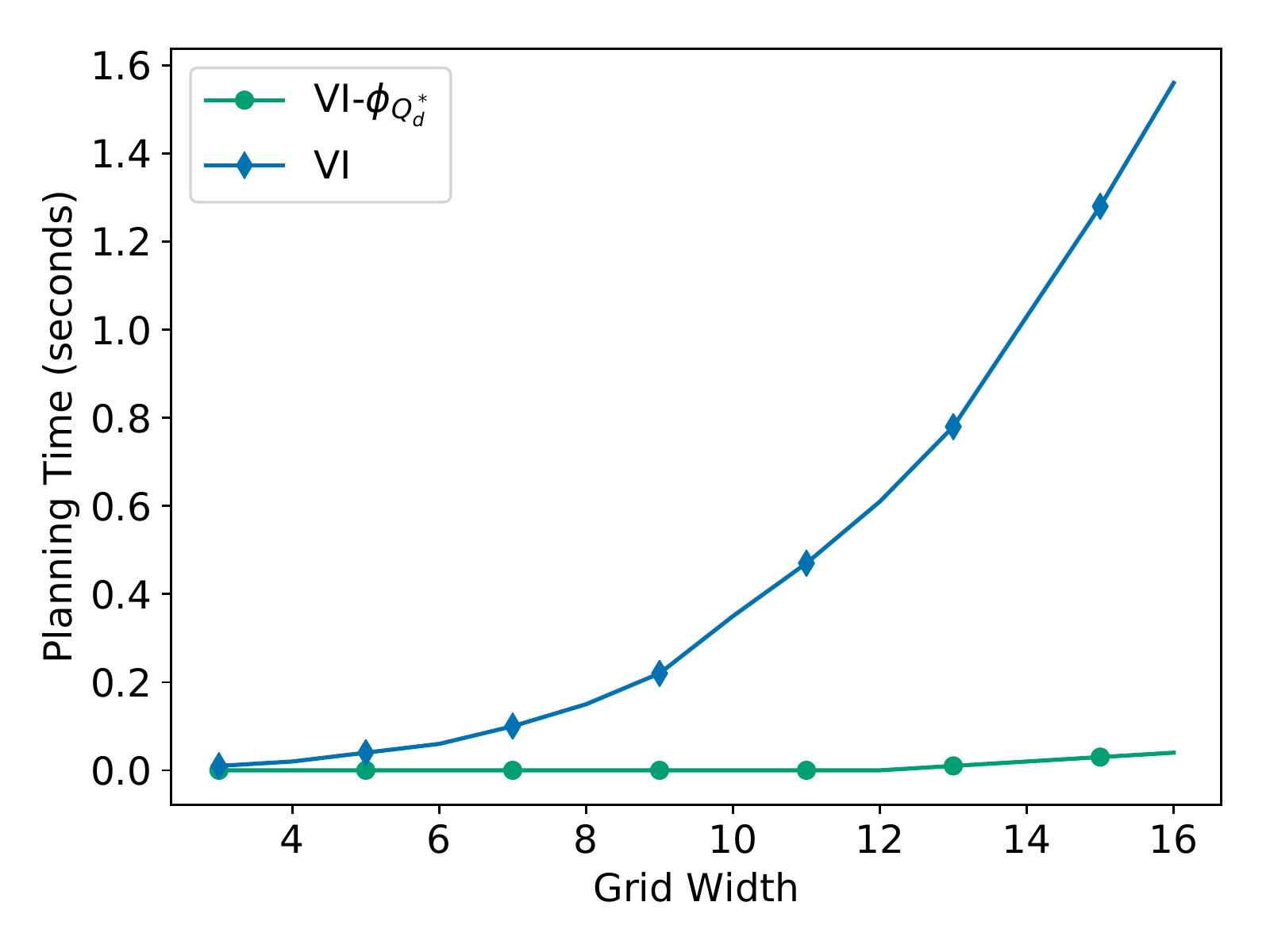}} \subfhspace
\subfloat[Planning in Four Rooms]{\includegraphics[scale=0.4]{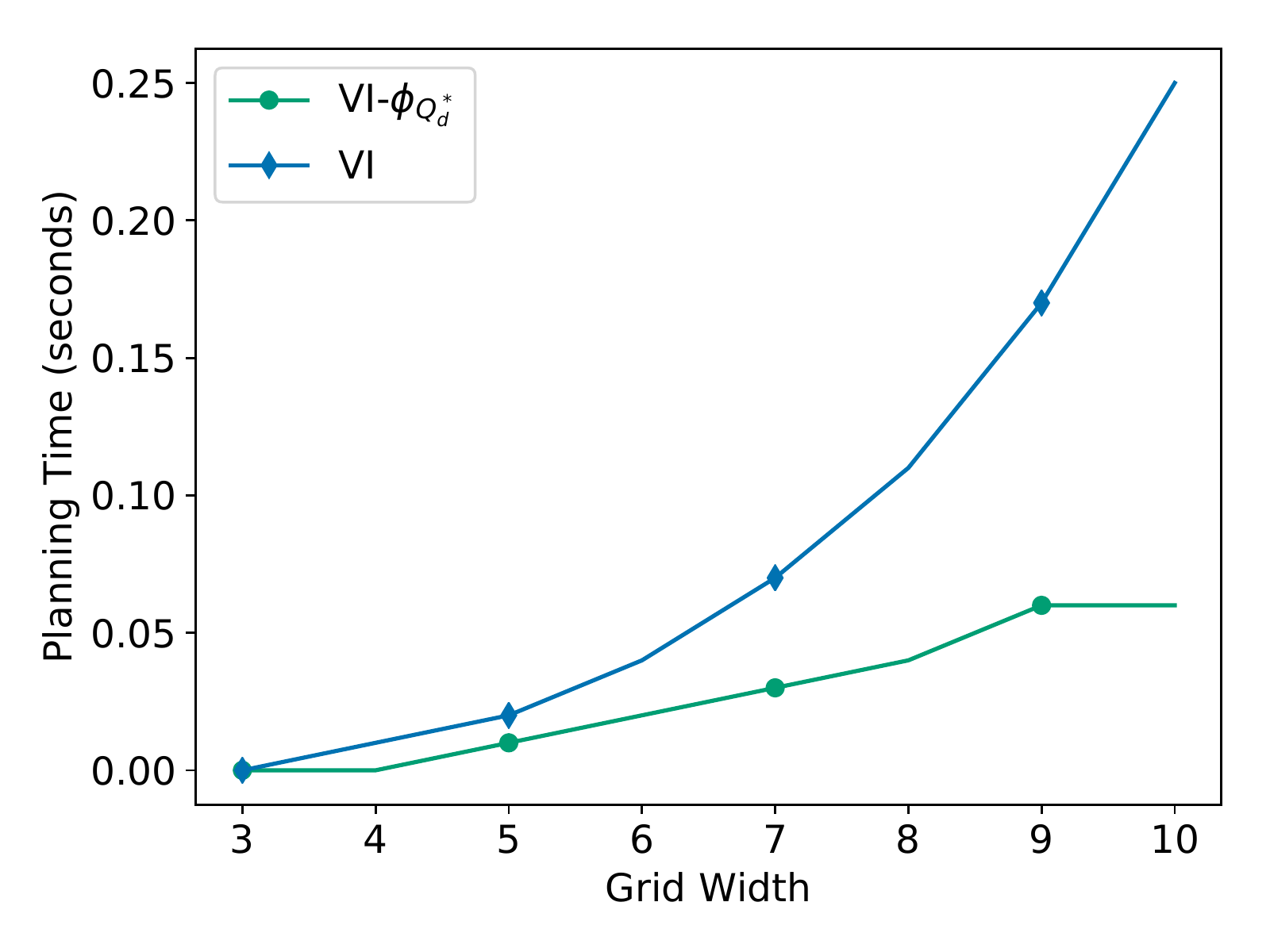}} \\
\subfloat[Planning in Color Rooms]{\includegraphics[scale=0.4]{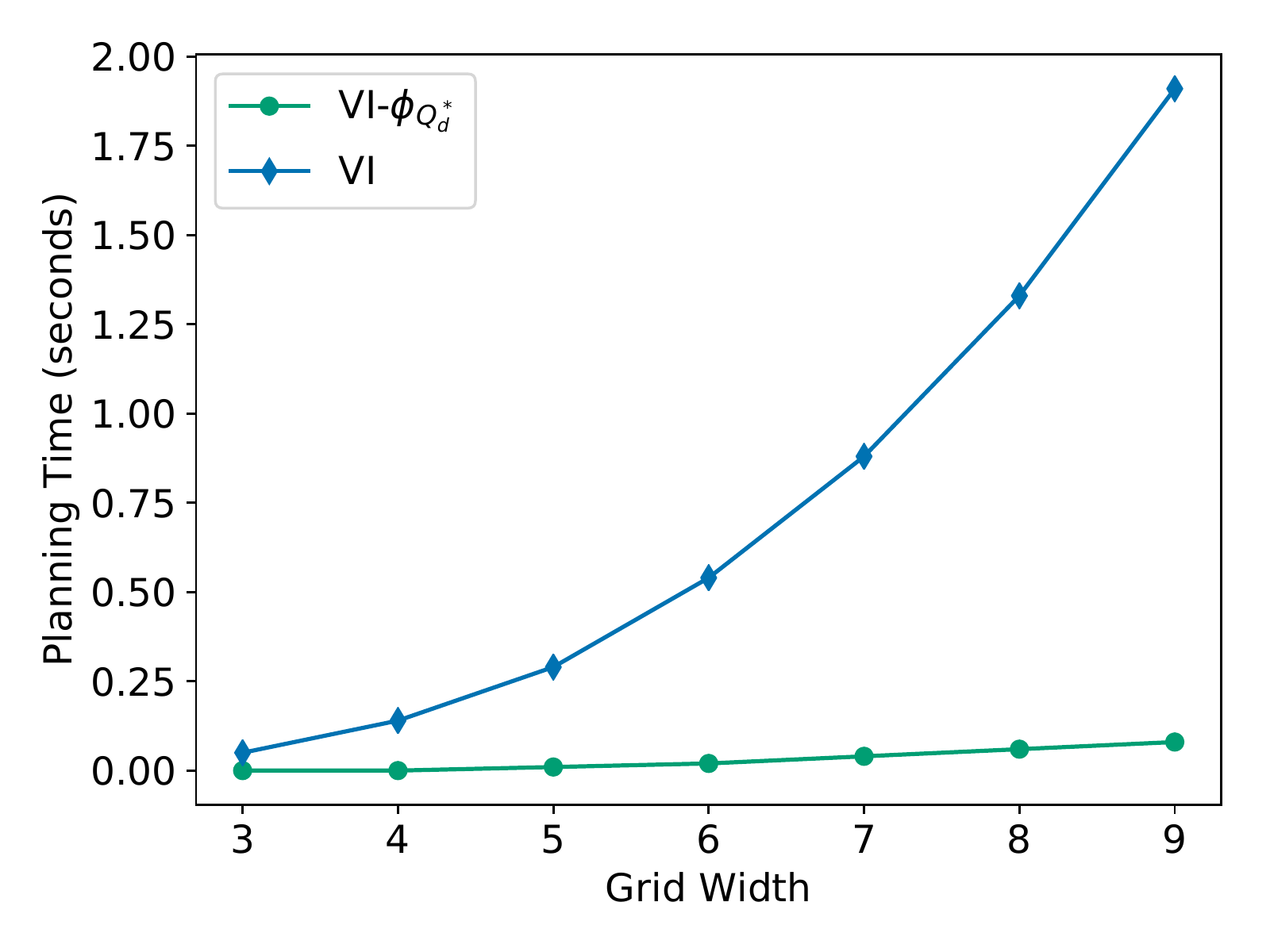}} \subfhspace
\subfloat[Planning in a Random MDP]{\includegraphics[scale=0.4]{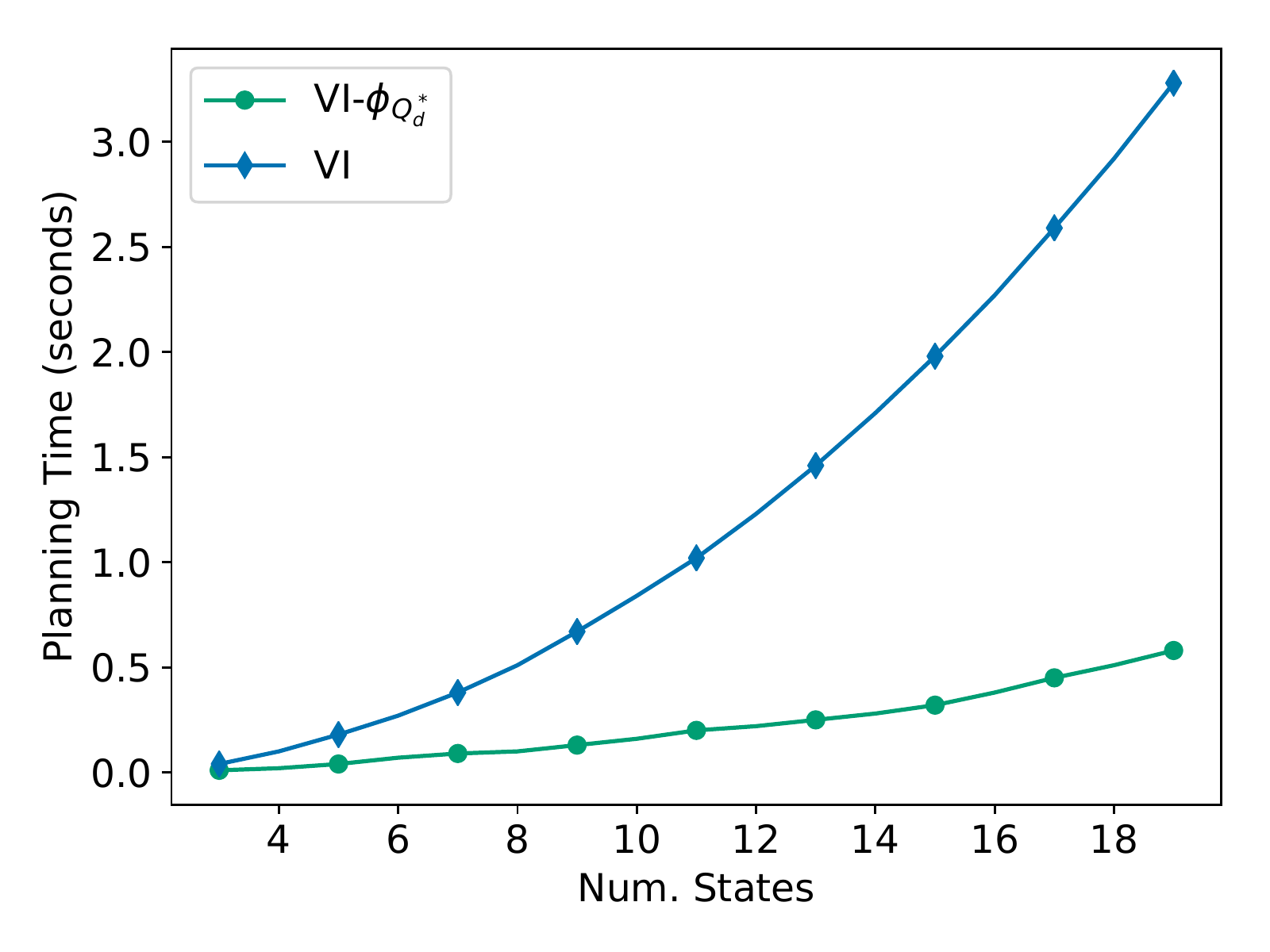}} \\

\caption{Planning time for Value Iteration with and without a state abstraction as the environmental state space grows.}
\label{fig:c4_plan_results}
\end{figure}

\textbf{Four Rooms}: I next conduct an experiment in a larger $15\times 15$ Four Rooms variant in which color and \texttt{paint} are removed to explore the degree to which the irrelevant variables explain the learning improvement found in the previous experiment. I evaluate Delayed $Q$-learning again with the same process for constructing state abstractions, again with 250 episodes, with 100 steps per episode. Here, the results suggest that the abstractions do effectively nothing to change learning---no irrelevant variables are included in the problem, and so only a few states are clustered. The original MDP has around 200 states, with the abstract state spaces averaging around 150 states. Consequently, learning is largely unchanged. However, not that in the case of $\phi_h$, when each state in the same room is clustered together, learning is devastated. This again highlights the importance of delicately choosing a state abstraction. Even when the other state abstractions did not accelerate learning, they at least did not negatively impact it, either.

\subsubsection{Planning}

To give further evidence of the potential benefits offered by state abstraction I next contrast the time taken to plan with and without the state abstraction. Indeed, the benefits of state abstraction to planning have been well studied~\cite{knoblock1994automatically,Hostetler2014,jiang2014improving,anand2015asap,anand2016oga}. I next study the impact of giving VI (\autoref{alg:value_iteration}) a state abstraction in four simple problems. The first is the $10\times 30$ Upworld grid problem from \autoref{chap:approx_state_abstr}, the second is the standard Four Rooms domain, the third is the Color Rooms domain from the previous experiment, and the final is the Random MDP from \autoref{chap:approx_state_abstr}. In each MDP, I vary the size of the underlying state space and contrast the time taken for VI to converge with and without a state abstraction.

Results are presented in \autoref{fig:c4_plan_results}. The findings are expected: in both Upworld and the Color Rooms MDPs, there are opportunities to abstract aggressively, thereby significantly lowering the computation needed to run VI to convergence. In the other two domains, there is some opportunity to abstract, but not as much, and consequently the benefits to planning time are not as dramatic. In all cases, the ground value of the computed policy is identical. Thus, the data suggest that planning can be accelerated when there is opportunity to abstract. Consequently, model-based RL algorithms employing the appropriate state abstractions may be able to plan more efficiently.

In this chapter, I focused on bringing state abstraction theory out of the traditional single task setting and into lifelong RL. I introduce two new complementary families of state abstractions, (1) transitive state abstractions, and (2) PAC abstractions. Together, they characterize state abstractions that can be feasibly obtained (satisfying D1) while still preserving near-optimal behavior (satisfying D3). Additionally, I drew attention to several shortcomings of learning with abstractions, building on those studied by~\citet{li2006towards} and~\citet{gordon1996chattering}, suggesting pathways for realizing the full potential of abstraction in RL. Moreover, the experimental evidence suggests that both planning and learning can be made more efficient when these abstractions are used---in this sense, these state abstractions target the satisfaction of all three desiderata. 

%% file: proofs/c4/c4_transitive_predicate_comp.tex
\begin{dproof}[Theorem \ref{thm:c4_sa_trans_comp}]
Let $c_p$ denote the computational complexity associated with computing the predicate $p$ for a given state pair. Consider the algorithm consisting of the following four rules for constructing abstract clusters (which define the abstract states) using queries to each of the $|\mc{S}|^2$ state pairs. Let $(s_i, s_j)$ denote the current state pair:

\begin{enumerate}
\item If $p(s_i, s_j)$ is true, and neither state is in an abstract cluster yet, make a new cluster consisting of these two states.
\item If $p(s_i, s_j)$ is true and only one of the states is already in a cluster, add the other state to the existing cluster. 
\item If $p(s_i, s_j)$ is true and both $s_i$ and $s_j$ are in different cluster, merge the clusters.
\item If $p(s_i, s_j)$ is false, add each state not yet in a cluster to its own cluster.
\end{enumerate}

Running this algorithm makes one query per state pair, of which there are $|\mc{S}|^2$. Thus, the complexity is $O\left(|\mc{S}|^2 \cdot c_p\right)$.

From steps 1-3, after iterating through the possible state pairs, there cannot exist a state pair $(s_x, s_y)$ such that $p(s_x, s_y)$ is true but $s_x$ and $s_y$ are in different clusters. Further, by transitivity, when we apply the cluster merge in step 3, we are guaranteed that every state pair in the resulting cluster necessarily satisfies the predicate. Thus, we compute the smallest clustering definable by $p$. \qedhere

\end{dproof}

%% file: proofs/c4/c4_qd_star_val_loss.tex
\begin{dproof}[Theorem \ref{thm:c4_sa_qdstar_vl}]
For any two state-action pairs that satisfy the predicate $p_M^d$, we know by definition of the predicate that for each action $a$, there exists a $Q_{lower}$ such that:
\begin{align*}
	Q_{lower} \leq Q^*(s_1,a) \leq Q_{lower} + d, \\
	Q_{lower} \leq Q^*(s_2,a) \leq Q_{lower} + d.
\end{align*}
Therefore, for each action $a$:
\begin{equation}
	|Q^*(s_1,a) - Q^*(s_2,a)| \leq d.
\end{equation}
Therefore, $\phi_{Q_d^*}$ is a subclass of $\phi_{Q_\eps^*}$. \qedhere
\end{dproof}

%% file: proofs/c4/c4_sa_state_space_size.tex
\begin{dproof}[Theorem \ref{thm:c4_sa_size_loss}]
Let $M$ be an arbitrary MDP. Consider a set of states $\tilde{S} \subset \mc{S}$ clustered together under $\phi_{Q^*_\eps}$ and, in particular, consider the $Q$-values of all states in $\tilde{S}$ for a particular action, $a \in \mc{A}$. Note that, by construction of $\phi_{Q^*_\eps}$, for any 
\begin{align*}
\forall_{s,s' \in \tilde{S}} : |Q^*(s, a) - Q^*(s',a)| \leq \eps,
\end{align*}

Recall that, intuitively, $\phi_{Q^*_d}$ is a discretization of the interval $[0, \textsc{VMax}]$ where $d$ controls the placement of boundaries, forming buckets of $Q$-values. The $Q$-values for all states in $\tilde{S}$ and for action $a$ reside in a single sub-interval of length $\eps$.

Letting $d = \eps$, the placement of boundaries that form $\phi_{Q^*_d}$ could break the $\eps$-interval of $Q$-values for the non-transitive cluster $\tilde{S}$ no more than once, resulting in the creation of at most two new state clusters in $\phi_{Q^*_d}$. Repeating the process for each action, these separations within the original cluster compound, resulting in at most $2^{|\mc{A}|}$ such subdivisions and, accordingly, $2^{|\mc{A}|}$ clusters in $\phi_{Q^*_d}$ for each cluster in $\phi_{Q^*_\eps}$. \qedhere

\end{dproof}

%% file: proofs/c4/c4_lifelong_sa_val_loss.tex
\begin{dproof}[Theorem \ref{thm:c4_pac_val_loss}]
By definition of PAC abstractions, with probability $1-\delta$, the abstraction function $\phi_p^\delta$ aggregates if and only if $\rho^p_{\delta + \eps}$, for some small $\eps \in (-\delta, \delta)$.

Then, with probability $1-\delta$, there is at least a $1-\delta-\eps$ chance that the predicate holds for a particular state, by definition of $\rho^p_{\delta}$. Thus, by definition of $\rho^p_{\delta}$, with probability $(1-\delta)(1-\delta-\eps)$, the state abstraction correctly aggregates, and consequently the inherited value loss $\tau_p$ bound holds. If the abstraction incorrectly aggregates, the value loss can be up to $\textsc{VMax}$.

Letting $\eps = \delta$, we see that the PAC loss is at worst upper bounded by a convex mixture of $\tau_p$ with probability $(1-3\delta)$, and with probability $3\delta$, is \textsc{VMax}. Thus, the expected value loss of $\phi_p^\delta$ is:
\begin{equation}
    \forall_{s \in \mc{S}} : \underset{M\sim D}{\bE}\left[V_M^*(s) - V_M^{\pi_{p,\delta}^*}(s)\right] \leq \eps (1-3\delta)\tau_p + 3\delta \textsc{VMax}. \qedhere
\end{equation}
\end{dproof}

%% file: proofs/c4/c4_pac_sample_bound.tex
\begin{dproof}[Theorem \ref{thm:c4_pac_sa_sample}]
We are given as input a $\delta \in (0,1]$, a distribution over MDPs $D$, and the algorithm $\mathscr{A}_p$ which, given an MDP $M$ and a state pair outputs $p_M(s,s').$

Consider an arbitrary pair of states $s$ and $s'$. For $m$ sampled MDPs, the algorithm $\mathscr{A}_p$ can produce a sequence of $m$ predicate evaluations:
\begin{equation}
    p_1(s,s'), \cdots, p_m(s,s').
\label{eq:p-seq}
\end{equation}
Let $\hat{p}$ be the empirical mean over the predicate sequence:
\begin{equation}
    \hat{p} = \frac{1}{m} \sum_{i=1}^m p_i(s,s').
\end{equation}
The clustering algorithm is quite simple: for our input $\delta \in (0,1]$, cluster all state pairs $(s,s')$ such that $\hat{p}(s,s') \geq 1-\delta$ after $m$ samples.

We now prove that, for a particular setting of $m$, the resulting cluster assignments constitute a state abstraction that clusters a pair of states only if the predicate is true with high probability.

First, let $\overline{p}$ denote the probability that $p$ is true over the distribution:
\begin{equation}
    \overline{p}(s,s') = \PR_{M \sim D}\{p(s,s') = 1\}.
\end{equation}
Using Hoeffding's Inequality, we upper bound the probability that $\hat{p}$ deviates from $\overline{p}$ by more than some small $\eps \in (0,\delta)$:
\begin{equation}
    \PR \left\{ \left|\hat{p}(s,s') - \bE \left[ \hat{p}(s,s') \right] \right| \geq \eps \right\} = \PR \left\{ |\hat{p}(s,s') - \overline{p}(s,s')| \geq \eps \right\} \leq 2e^{-2m \eps^2}.
\end{equation}

\noindent Thus, for $\delta = 2e^{-2m \eps^2}$:
\begin{equation}
    \PR \left\{ |\hat{p}(s,s') - \overline{p}(s,s')| < \eps \right\} > 1 - \delta.
    \label{eq:sa_hoeff}
\end{equation}
Rewriting:
\begin{align}
    &\PR \left\{ | \hat{p}(s,s') - \overline{p}(s,s') | < \eps \right\} > 1 - \delta \\
    &\hspace{8mm}\iff \nonumber\\
    &\PR \left\{  -\eps < \hat{p}(s,s') - \overline{p}(s,s') < \eps \right\} > 1 - \delta,
\end{align}
By algebra, note that, when $m \ge \frac{\ln \frac{2}{\delta}}{\eps^2}$, the condition of Equation~\ref{eq:sa_hoeff} holds.

Let $\rho^p_\delta$ denote the predicate that is true if and only if $p$ is true over the distribution with high probability for a given $\delta$:
\begin{equation}
    \rho^p_\delta(s_1, s_2) = \begin{cases}
        1& \overline{p} \geq 1-\delta \\
        0& \text{otherwise.}
    \end{cases}
\end{equation}
Now, we form our state abstraction under the following rule:
\begin{equation}
    \hat{\phi}_p^\delta(s_1) = \hat{\phi}_p^\delta(s_2) \equiv \hat{p}(s,s') > 1 - \delta.
\label{eq:clustering_rule}
\end{equation}
If, after $m$ samples, $\hat{p}$ were identical to $\overline{p}$, then we would have:
\begin{equation}
    \forall_{s,s'} : \PR_{M \sim D} \{\rho^p_\delta(s,s') \equiv \hat{\phi}_p^\delta(s_1) = \hat{\phi}_p^\delta(s_2)\} \geq 1-\delta.
\end{equation}
Hence, $\hat{p}$ deviates from $\overline{p}$ by at most $\epsilon$ with probability $1-\delta$. Thus, for some $\eps \in (-\delta, \delta)$, $\hat{p} + \eps = \overline{p}$. Therefore, the clustering rule defined by Equation~\ref{eq:clustering_rule} ensures there exists an $\eps$ such that, with high probability, we cluster according to:
\begin{equation}
    \forall_{s_1, s_2} : \rho^p_{\delta + \eps}(s_1, s_2) \equiv \phi_p^\delta(s_1) = \phi_p^\delta(s_2).
\end{equation}

We conclude that, for $m \ge \frac{\ln \frac{2}{\delta}}{\eps^2}$ sampled and solved MDPs, we compute a lifelong PAC state abstraction $\hat{\phi}_p^\delta$.\qedhere
\end{dproof}

%% file: proofs/c4/c4_rmax_broken.tex
\begin{dproof}[Theorem \ref{thm:c4_rmax_broken}]

Consider the simple three state chain:
\begin{center}
\begin{tikzpicture}
    \node [draw, circle] (s0) at (2, 0) {$s_0$};
    \node [draw, circle] (s1) at (4, 0) {$s_1$};
    \node [draw, circle] (g) at (6, 0) {$g$};

    \draw[out=80,in=30,looseness=7] (s0) edge node[right]  {$+\kappa$} (s0);
    \draw[out=80,in=30,looseness=7] (s1) edge node[right]  {$+\kappa$} (s1);
    \draw[out=80,in=30,looseness=7] (g) edge node[right]  {$+\textsc{RMax}$} (g);
    \draw[-] (s0) edge node  {} (s1);;
    \draw[-] (g) edge node  {} (s1);
\end{tikzpicture}
\end{center}
    
The agent has three actions, \texttt{left}, \texttt{right}, and \texttt{loop}, associated with their natural effects (\texttt{left} in $s_0$ is a self loop with reward $0$, while \texttt{right} moves the agent to $s_1$, and so on).

In states $s_0$ and $s_1$, let the reward for \texttt{loop} be some small constant $\kappa$, and let the \texttt{loop} action in $s_3$ yield $\textsc{RMax}$ reward.

Let $\gamma=0.95$, $s_0$ define the initial state, and $\kappa=0.001$. Suppose the agent reasons using an instance of $\phi_{Q_\eps^*}$ with $\eps=0.1$. Then, observe that all three states may be clustered, since:
\begin{align}
    \forall_{s \in \{s_0, s_1, s_2\}} : \max_{a_1, a_2} Q^*(s,a_1) - Q^*(s, a_2) \leq \eps.
\end{align}

That is, note that the $Q^*$ values of each state-action pair are roughly as follows:
\begin{center}
    \begin{tabular}{lll}
    $Q^*(s_0,\texttt{right}) \approx 0.90$& $Q^*(s_1,\texttt{right}) \approx 0.95$ &$Q^*(s_2,\texttt{right}) \approx 0.95$ \\
    $Q^*(s_0,\texttt{left}) \approx 0.86$& $Q^*(s_1,\texttt{left}) \approx 0.86$ & $Q^*(s_2,\texttt{left}) \approx 0.90$ \\
    $Q^*(s_0,\texttt{loop}) \approx 0.87$ & $Q^*(s_1,\texttt{loop}) \approx 0.91$ & $Q^*(s_2,\texttt{loop}) \approx 1.0$ \\
    \end{tabular}
\end{center}
\vspace{2mm}

Therefore, for $\eps=0.1$, a valid clustering assigns $\phi(s_0) = \phi(s_1)$.

To break ties, we suppose R-Max chooses actions according to a \textit{round-robin} policy, starting with action \texttt{left}. Thus, in the abstract, R-Max first chooses left, then right, then self loop, then left, right, self loop, and so on, until each state-action pair is known.

In the above problem, this sequence of actions will \textit{never} lead the agent out of state $s_0$ or $s_1$. Let $m$ denote the parameter given to R-Max that determines how many samples per state-action pair are sufficient for the pair to be considered known. Therefore, after $m$ executions of these three actions across states $s_0$ and $s_1$, R-Max with $\phi$ will compute a transition model that assigns zero probability to arriving in $g$ from the aggregated state $\phi(s_0)$. Further, the action \texttt{loop} will have the largest reward associated with it---$\kappa$, a reward chosen to be arbitrarily small---which is thus arbitrarily worse than the goal reward. So, R-Max with $\phi$ will make an unbounded number of mistakes even when $\phi$ is a state abstraction that ensures bounded value loss. \qedhere
\end{dproof}

%% file: proofs/c4/c4_m_phi.tex
\begin{dproof}[Corollary \ref{corr:c4_m_phi_vs_learn_w_phi}]
Note that when $M_\phi$ is computed directly, the functions $R_\phi$ and $T_\phi$ assume a fixed weighting function $w(s)$.

Again let us consider the three state chain MDP from the previous proof. During typical interaction between $M$ and $\mathscr{A}_\phi$, no such fixed weighting function exists \textit{for any algorithm $\mathscr{A}$ that updates its policy}. That is, the distribution of states the agent finds itself in will change as its policy changes, and therefore, $w(s)$ must change, too, thereby updating $T_\phi$. Conversely, in the true MDP, $T$ remains fixed.

Thus, the process of $\mathscr{A}_\phi$ interacting with $M$ induces a sequence of interactions with abstract MDPs whose transition and rewards change along with the policy the agent follows. Thus, for some non-identity $\phi$, for any algorithm $\mathscr{A}$ whose policy changes over time, the resulting expected interaction may be different.\qedhere
\end{dproof}

%% file: chapters/c5_sa_state_abstr_as_compression.tex
\begin{center}
\begin{minipage}{0.8\textwidth}
\textit{This chapter is based on ``State Abstraction as Compression in Apprenticeship Learning" \cite{abel2019rlit} joint with Dilip Arumugam, Kavosh Asadi, Lawson L.S. Wong and Michael L. Littman, and ``Learning State Abstractions for Transfer in Continuous Control" \cite{asadi2020sa_control} led by Kavosh Asadi, also with Michael L. Littman.}
\end{minipage}
\end{center}
\vspace{2mm}

In the previous two chapters, I analyzed classes of state abstraction functions that can reduce the size of the underlying state space while simultaneously preserving representation of good policies. This dual-objective closely parallels the mission of \textit{information theory}, which presents a rigorous formalism for understanding communication in the presence of noise. The key results of information theory are centered around the act of \textit{compression}---how an entity can be reduced in size while preserve its essence. There is striking similarity between this process and that of abstraction. A natural line of reasoning, then, seeks to establish more explicit contact between the tools of information theory and the process of abstraction. Indeed, cognitive neuroscience has suggested that perception and generalization are tied to efficient compression~\cite{fattneave1954,sims2016rate,sims2018efficient}, termed the ``efficient coding hypothesis" by~\citet{barlow1961possible}.

The goal of this chapter is to adopt the viewpoint that state abstraction for sequential decision making can be understood as a process of compression. From this new perspective, I will introduce a new algorithm for constructing state abstractions that imports many of the desirable characteristics enjoyed by some of the key algorithms of information theory. Precisely, I draw a parallel between \textit{state abstraction} as used in reinforcement learning and \textit{compression} as understood in information theory. This parallel is heavily inspired by the seminal work of~\citet{Shannon1948}, \citet{blahut1972computation}, \citet{arimoto1972algorithm} and~\citet{tishby2000information}, and draws on insights from related work on understanding the relationship between abstraction and compression~\cite{botvinick2015reinforcement,solway2014optimal}.

While the perspective I here introduce is intended to be general, I will restrict the initial study by concentrating on the learning problem when a \textit{demonstrator} is available, as in Apprenticeship Learning~\cite{atkeson1997robot,abbeel2004apprentice,argall2009survey}, which simplifies aspects of the model. I will later build toward the regular RL setting after this initial framework is established.

\begin{figure}[t!]
    \centering
    \includegraphics[height=\figvdim]{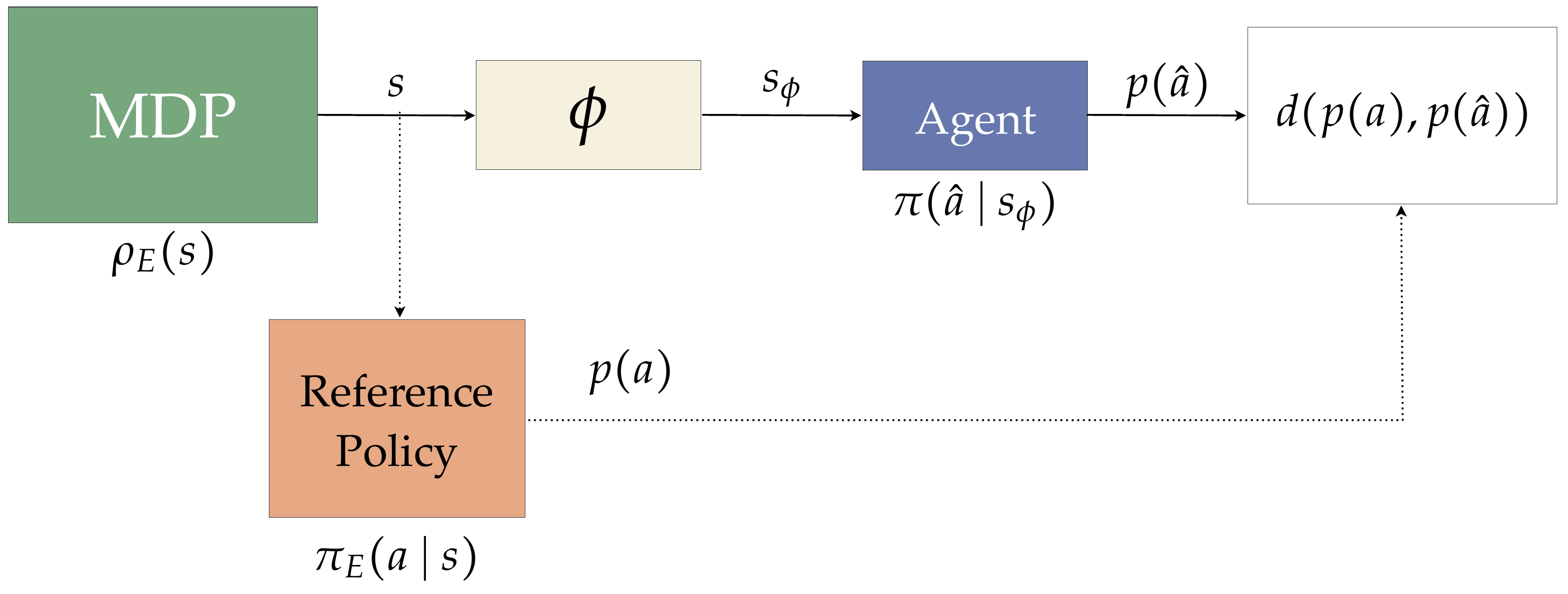}
    \caption{The proposed framework for trading off compression with value via state abstraction.}
    \label{fig:c5_rl_abstr_rate_dist}
\end{figure}

Concretely, I introduce a new objective function that explicitly balances state-compression and performance. The main result of this chapter proves this objective is upper bounded by a variant of the Information Bottleneck objective adapted to sequential decision making. I introduce Deterministic Information Bottleneck for State abstraction (\textsc{Dibs}), an algorithm that outputs a \textit{lossy} state abstraction optimizing the trade off between compressing the state space and preserving the capacity for performance in that compressed state space. I present empirical results that showcase the relationship between compression and performance captured by the algorithm in a traditional grid world, along with an extension to high-dimensional observations via experiments with the Atari game Breakout. Then, I introduce several extensions to this framework that relax critical assumptions, allowing for more general application of the proposed methods.

First, I present a brief survey of information theory.

\section{Information Theory}

Information theory offers foundational results about the limits of compression~\cite{Shannon1948}. The core of the theory clarifies how to communicate in the presence of noise, culminating in seminal results about the nature of communication and compression that helped establish the science and engineering practices of computation. In Shannon's words: ``The fundamental problem of communication is that of reproducing at one point either exactly or approximately a message selected at another point" (\citeyear{Shannon1948}, p. 1). One focus of information theory is on constructing coder-decoder pairs that can faithfully communicate messages with zero or low error, even in the presence of noise.

The seminal results all center around the definition of \textit{entropy}, sometimes called the Shannon entropy:
\ddef{Entropy}{The \textbf{entropy} of a discrete random variable $X$, with alphabet $\mc{X}$, is given by:
\begin{equation}
H(X) := -\sum_{x \in \mc{X}} p(x) \log p(x).
\end{equation}}

The entropy measures, roughly, the surprise inherent in a random variable. Throughout this chapter, I use $\log$ as shorthand for $\log_2$.

Entropy may be extended to account for joint and conditional probability distributions as follows.

\ddef{Joint Entropy}{The \textbf{joint entropy} of two discrete random variables $X$ and $Y$, with alphabets $\mc{X}$ and $\mc{Y}$, is given by:
\begin{equation}
H(X, Y) := -\sum_{x \in \mc{X}}\sum_{y \in \mc{Y}} p(x,y) \log p(x,y).
\end{equation}
}

\ddef{Conditional Entropy}{The \textbf{conditional entropy} of $X$ given $Y$ (again two discrete random variables with alphabets $\mc{X}$ and $\mc{Y}$) is given by:
\begin{equation}
H(X \mid Y) := -\sum_{x \in \mc{X}} \sum_{y \in \mc{Y}} p(x,y) \log p(x \mid y).
\end{equation}}

An additional quantity of relevance is the Mutual Information between two discrete random variables:

\ddef{Mutual Information}{The \textbf{mutual information} of two discrete random variables $X$ and $Y$ is given by:
\begin{equation}
    I(X;Y) := \sum_{x \in \mc{X}} \sum_{y \in \mc{Y}} p(x,y) \log \frac{p(x,y)}{p(x)p(y)}.
\end{equation}}

\begin{figure}
    \centering
    \includegraphics[width=0.4\textwidth]{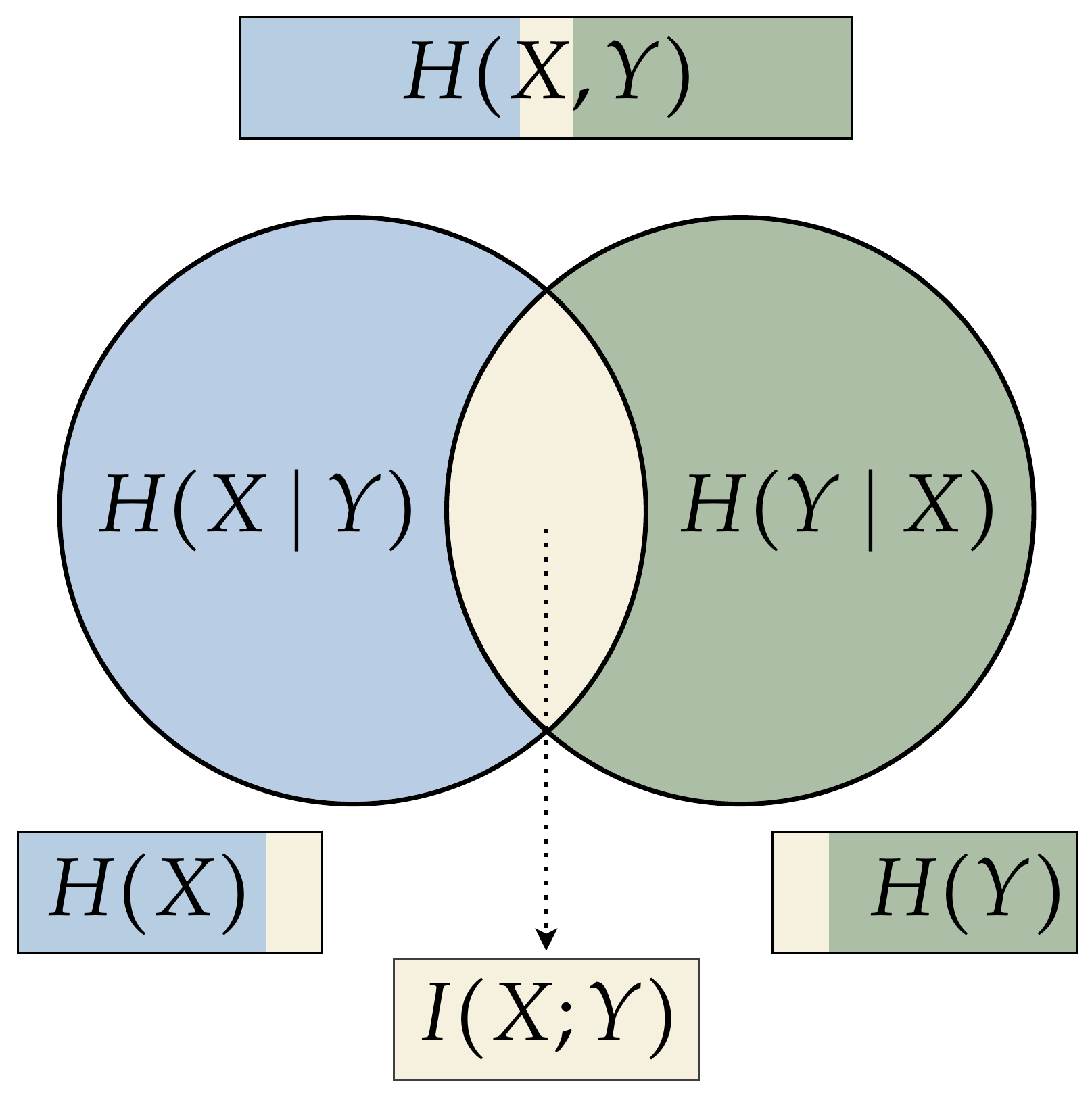}
    \caption{The basic quantities of information theory and their relations.}
    \label{fig:c5_entropy_venn_diagram}
\end{figure}

Together with the joint and conditional entropy, the theory offers an elegant, interlocking set of relations between these basic quantities. Their relations are pictured in \autoref{fig:c5_entropy_venn_diagram}.

A final quantity that is of special interest is the \textit{relative entropy}, also called the Kullback-Leibler divergence (KL divergence). The KL divergence expresses the error associated with choosing the probability distribution $p(\tilde{x})$ to approximate the probability distribution $p(x)$.

\ddef{Kullback-Leibler divergence}{The \textbf{KL divergence} between two probability distributions $p(x)$ and $q(x)$ is given by:
\begin{equation}
    \KL(p\mid\mid q) = \sum_{x \in \mc{X}} p(x) \log \frac{p(x)}{q(x)}.
\end{equation}}

Observe that another interpretation for the mutual information of two random variables $X$ and $Y$ is that it expresses the KL divergence between the joint $p(x,y)$ and the independent: $p(x) p(y)$ distributions:
\begin{equation}
    I(X;Y) = \KL(p(x,y) \mid \mid p(x) p(y)) = \sum_{x \in \mc{X}} \sum_{y \in \mc{Y}} p(x,y) \log \frac{p(x,y)}{p(x)p(y)}.
\end{equation}

Further note two other properties of the KL divergence: 1) The $\KL$ between two probability distributions that do not have overlapping support is $\infty$, and 2) $\KL$ is not a metric, because there exist choices of $p$ and $q$ such that $\KL(p \mid \mid q) \neq \KL(q \mid \mid p)$. Still, the KL divergence is an exceptionally useful measure.

For more background on information theory, see the book by \citet{cover2012elements}.

\subsection{Rate-Distortion Theory}

Of particular relevance to state abstraction is Rate-Distortion (RD) theory, which is a subfield of information theory that studies the trade off between a code's ability to compress (rate) and represent the original signal (distortion)~\cite{Shannon1948,berger1971rate}. 

\begin{figure}[t]
    \centering
    \includegraphics[width=0.7\textwidth]{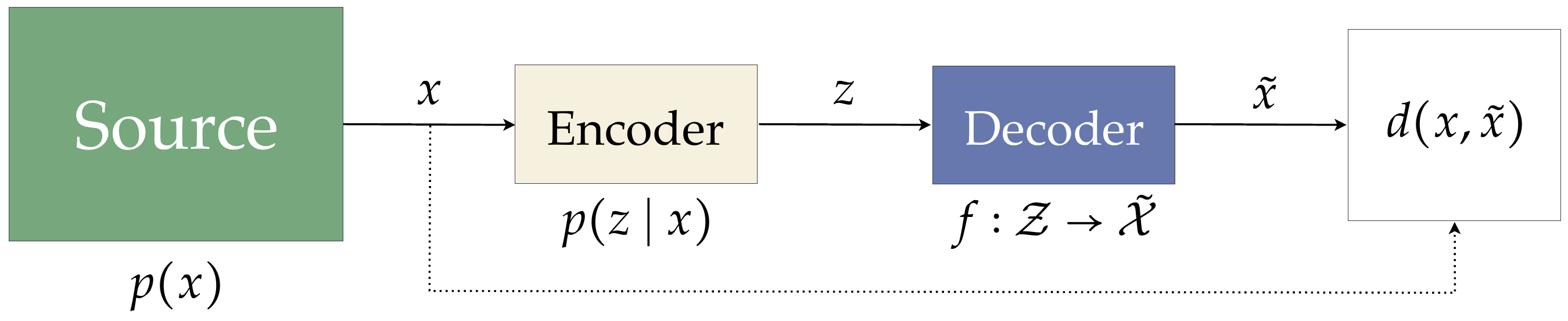}
    \caption{The usual Rate-Distortion setting.} 
    \label{fig:c5_rate_dist_setup}
\end{figure}

The typical RD setting is pictured in \autoref{fig:c5_rate_dist_setup}: an information source generates $x \in \mc{X}$, which is coded via $p(z \mid x)$ to $z \in \mc{Z}$, and decoded via a deterministic function $f : \mc{Z} \ra \tilde{\mc{X}}$. \emph{Distortion} is defined with respect to a chosen distortion metric, $d : \mc{X} \times \tilde{\mc{X}} \ra \mathbb{R}_{\geq 0}$, where typically $\mc{X} = \tilde{\mc{X}}$. The information \emph{rate}, $R$, denotes the number of bits in each code word. So, with a coding alphabet $\tilde{\mc{Z}} = \{0,1\}^n$, the rate is $n$. Shannon and Kolmogorov (see~\citet{berger1971rate} for more background) offer a lower bound on the trade off between Rate and Distortion: given a level of distortion, $D$, the following function defines the smallest rate that achieves expected distortion of at most $D$:
\begin{equation}
    R(D) = \min_{p(\tilde{x} \mid x) : \bE[d(x, \tilde{x})] \leq D} I(X;\tilde{X}).
    \label{eq:c5_rate_dist_lower_bound}
\end{equation}
Intuitively, \autoref{eq:c5_rate_dist_lower_bound} tells us that as bits are added to the code, algorithms can more faithfully reconstruct the original source messages.

For a given information source, it is natural to consider how to compute a coder-decoder pair that achieves one of the minimal points defined by the Rate-Distortion function. Finding this point presents the following optimization problem:
\begin{equation}
    \min_{p(\tilde{x} \mid x)} \underbrace{I(X;\tilde{X})}_{Rate} + \beta \underbrace{\bE_{p(x,\tilde{x})} \left[d(x,\tilde{x})\right]}_{Distortion},
    \label{eq:c5_ba_opt}
\end{equation}
with a Lagrange multiplier $\beta \in \mathbb{R}_{\geq 0}$ expressing the relative preference between minimizing rate and distortion. As $\beta$ gets closer to $0$, rate becomes more important, while as $\beta$ approaches $\infty$, minimizing distortion is prioritized. Note that it is desirable to identify coder-decoder pairs that live exactly on this curve, as any point living above indicates that either more compression can take place (lower rate) or more accurate reconstruction can take place (lower distortion).

Blahut-Arimoto (BA) is a simple iterative algorithm that converges to the global optimum of this optimization problem~\cite{arimoto1972algorithm,blahut1972computation}. BA alternates between the following two steps, for a given $\beta \in \mathbb{R}_{\geq 0}$:
\begin{align}
    p_{t+1}(\tilde{x}) &= \sum_{x \in \mc{X}} p(x) p_t(\tilde{x} \mid x), \\
    p_{t+1}(\tilde{x} \mid x) &= \frac{p_{t+1}(\tilde{x}) \exp(-\beta d(x, \tilde{x}))}{\sum_{x' \in \tilde{\mc{X}}} p_{t+1}(x') \exp(-\beta d(x, x'))}.
\end{align}

BA is known to converge to the global optimum with convergence rate:
\begin{equation}
O\left(|\mc{X}| |\tilde{X}| \sqrt{\log(|\tilde{X}|)} / \eps\right),
\end{equation}
for $\eps$ error tolerance~\cite{arimoto1972algorithm}. The computational complexity of finding the exact solution for a discrete, memoryless channel is unknown. For a continuous memoryless channel, the problem is an infinite-dimensional convex optimization which is known to be NP-hard~\cite{Sutter2015EfficientAO}.

\subsection{The Information Bottleneck Method}

Note, however, that for us to make use of the Blahut-Arimoto algorithm for computing an optimal coder-decoder pair (for a given $\beta$), a distortion metric $d$ is required. However, as discussed by \citet{tishby2000information}, this requirement places \textit{all} of the burden of relevant information onto choice of metric; RD theory defines ``relevant" information by choice of a distortion function---codes are said to capture relevant information if they achieve low distortion. How, though, should such a metric be chosen if the signal being transmitted were to represent an image? It is not obvious whether precise pixel values are the important thing to preserve, rather than the overall contents of the image at the level of objects, scenes, and relations. A simple pixel inversion or image rotation will surely yield extremely high increase for many natural choices of metric, but in many contexts such transformations don't actually destroy relevant information.

\begin{figure}[t!]
    \centering
    \includegraphics[width=0.7\textwidth]{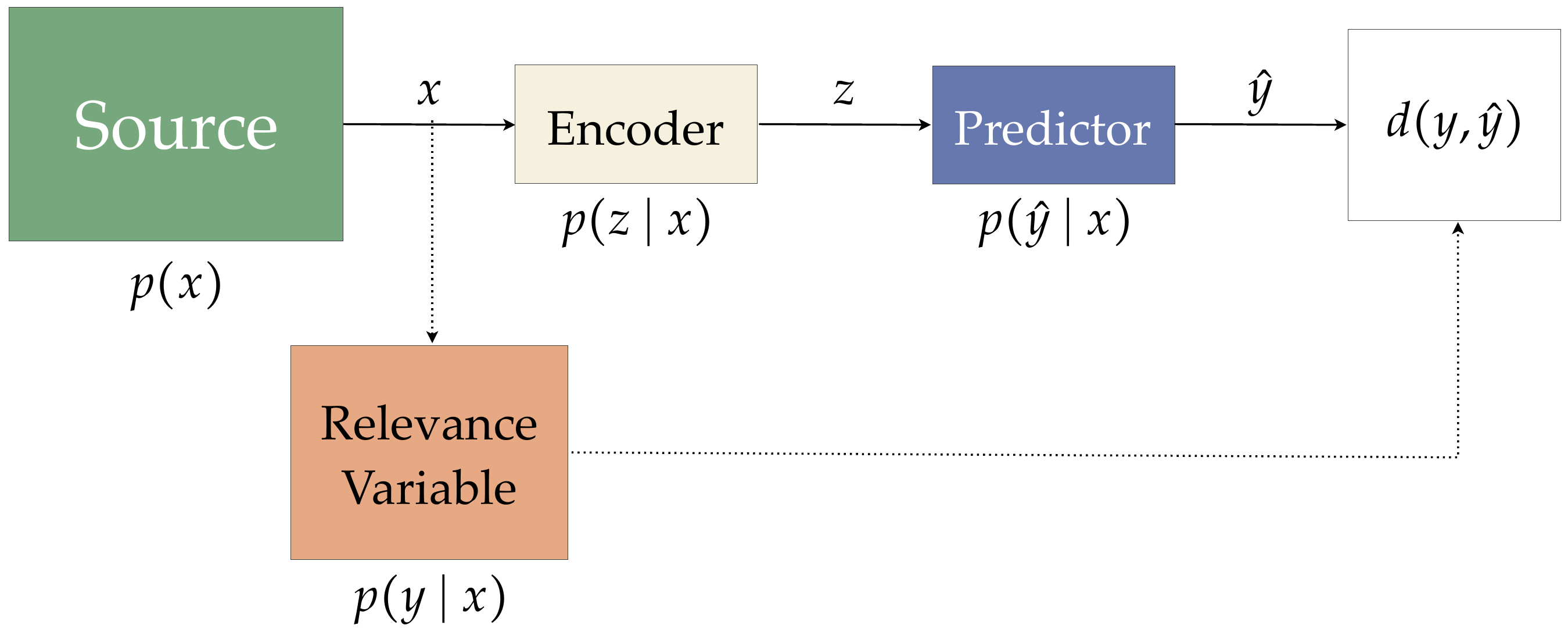}
    \caption{The Information Bottleneck.}
    \label{fig:c5_info_bottleneck_setup}
\end{figure}

In this sense, the choice of metric determines what counts as relevant information. In general, it is desirable to allow for a less restrictive choice of relevant information.

The Information Bottleneck (IB) Method is one possible remedy to this problem. The IB defines relevant information according to how well a random variable $Y$ can be predicted from each $\tilde{x} \in \tilde{\mc{X}}$, as pictured in \autoref{fig:c5_info_bottleneck_setup}. For the IB to make sense, we must suppose that $I(X;Y) > 0$, and that the coder--decoder scheme has access to the joint probability mass function (pmf) $p(x,y)$. IB then recasts the RD lower bound in \autoref{eq:c5_rate_dist_lower_bound} in terms of prediction of $Y$ given $\tilde{X}$. The optimal assignment to the distribution $p(\tilde{x} \mid x)$ is then given by minimizing:
\begin{equation}
    \mc{L}\left[p(\tilde{x} \mid x)\right] = I(\tilde{X};X) - \beta I(\tilde{X};Y),
\label{eq:c5_ib_opt_function}
\end{equation}
where $\beta \in \mathbb{R}_{\geq 0}$ is again a Lagrange multiplier attached to the meaningful information. Like BA, choice of $\beta$ determines the relative preference between compression (rate) and predicting $Y$ (distortion); when $\beta = 0$, the coder can ignore $Y$ entirely, and so is free to compress arbitrarily. Conversely, as $\beta \ra \infty$, the coder must prioritize prediction of $Y$, requiring more bits in the coding alphabet.

\citet{tishby2000information} offer a convergent algorithm for solving the above optimization problem.
\begin{theorem}(Appears as Theorem 5 by~\citet{tishby2000information})
\label{thm:c5_tishby_ib}
\autoref{eq:c5_ib_opt_function} yields the following optimization problem:
\begin{equation}
    \min_{p(\tilde{x} \mid x)} \mc{L}_{\text{IB}}\left[p(\tilde{x}\mid x);p(\tilde{x});p(y\mid\tilde{x})\right] = \min_{p(\tilde{x} \mid x)} \left( I(X; \tilde{X}) + \beta \underset{p(x,\tilde{x})}{\bE}\left[\KL(p(y \mid x) \mid\mid p(y \mid \tilde{x})\right]\right).
    \label{eq:c5_tishby_obj} 
\end{equation}

The algorithm consists of the following three steps, which, when repeated, converge to a local minima of the above optimization problem, with $Z(\beta, x)$ a normalizing term:
\begin{equation}
    \begin{cases}
    &p_t(\tilde{x} \mid x) \la \frac{p_t(\tilde{x})}{Z(\beta, x)} \exp(-\beta \KL(p(y \mid x) \mid \mid p_t(y \mid \tilde{x}))), \\
    &p_{t+1}(\tilde{x}) \la \sum_x p(x) p_t(\tilde{x} \mid x), \\ 
    &p_{t+1}(y \mid \tilde{x}) \la \sum_y p(y \mid x) p_t(x \mid \tilde{x}).
    \end{cases} 
    \label{eq:c5_ib_iterative}
\end{equation}
\end{theorem}
It is important to note that the algorithm only presents a \textit{locally optimal} solution to the above optimization problem. To the best of our knowledge, there is no known efficient algorithm for computing the global optimum. \citet{mumey2003optimal} show that a closely related problem to finding the global optimum in the above is in fact NP-hard, suggesting that local convergence or approximation is likely our best option. Additionally, if the support of $p(y \mid x)$ and $p(y \mid \tilde{x})$ does not \textit{exactly} overlap, then $\KL$ is trivially infinity, leading to vacuous updates. It is thus important that application of their algorithm be applied in a context with overlapping supports.

\paragraph{The Deterministic IB (DIB).} \citet{strouse2017deterministic} extend IB by focusing on deterministic coding functions where $p(\tilde{x} \mid x) \equiv f : \mc{X} \ra \tilde{\mc{X}}$. Given the equality $I(X;Y) = H(X) - H(X\mid Y)$, note that when the coder is a deterministic function $f$, we can replace the mutual information term in the objective by the entropy of the latent space:
\begin{equation}
    \min_{f(x)} \mc{L}_{\text{DIB}}\left[f(x); p(\tilde{x}); p(y \mid x) \right] =
    \min_{f(x)} \left(H(\tilde{X}) + \beta \underset{p(x)}{\bE}\left[\KL(p(y \mid x) \mid\mid p(y \mid \tilde{x}))\right] \right).
\end{equation}
Given that state abstractions are often deterministic, I will primarily be focused on this extension.

\section{Analysis: State Abstraction as Compression}

I now adapt the Information Bottleneck to construct state abstractions for sequential decision making problems. The proposed framework is pictured in \autoref{fig:c5_rl_abstr_rate_dist}, with the information generating source defined as $\rho_E$, the stationary distribution in the given MDP induced by the expert policy, $\pi_E$. That is, the source distribution is defined as the $\gamma$-discounted stationary distribution $\rho_E(s)$, for each $s \in \mc{S}$, for a given start state distribution $\rho_0$, as:
\begin{equation}
    \rho_E(s) := \sum_{t=0}^\infty \gamma^t \PR\{s_t = s \mid \rho_0, \pi_E\}
    \label{eq:c5_station_distr}
\end{equation}
My goal is to answer the following question: \textit{How many abstract states are needed for an agent to faithfully make similar decisions to an expert demonstrator?} To answer this question, I cast the Rate-Distortion trade off as one between (1) the size of the abstract state space $|\mc{S}_\phi|$, and (2) the value of the best policy representable using $\mc{S}_\phi$ compared to $\pi_E$.

One might wonder why such a question cannot be answered by assigning one abstract state to each action, as is captured by the ${\pi^*}$-irrelevance abstractions studied by \citet{jong2005state} and \citet{li2006towards}. First, if the demonstrator policy is stochastic, no such abstraction exists. Second, we are ultimately interested in state abstractions that facilitate effective \textit{learning}; if the abstraction were given to an arbitrary RL algorithm, we would like learning to be made easier. Highly aggressive abstraction types like ${\pi^*}$ destroy guarantees and make aspects of learning harder~\cite{li2006towards,abel2017abstr}. Lastly, $\pi^*$-irrelevance only captures lossless abstraction; through RD, we can build a toward theory of lossy compression for RL.

More formally, I introduce and study the following objective:
\ddef{Compression-Value Abstraction or CVA Objective}{The objective function, $\mc{J}$, for a given Lagrange multiplier $\beta \in \mathbb{R}_{\geq 0}$, is defined as:
\begin{equation}
    \mc{J}\left[\phi\right] := |\mc{S}_\phi| + \beta \underset{\rho_E(s)}{\bE}\left[ V^{\pi_E}(s) - V^{\pi_\phi^*}(\phi(s))\right].
\end{equation}
}
The goal is to define an algorithm that efficiently minimizes the above objective.

Eventually, I will introduce an algorithm that minimizes an upper bound on the CVA objective. To related this upper bound, we require the following definition, denoting the size of the non-negligibly used portion of an alphabet under a pmf:
\ddef{pmf-Used Alphabet Size}{The pmf-used alphabet size of $\mc{X}$ is the number of elements whose probability under $p(x)$ is greater than some negligibility threshold $\delta_{min} \in (0,1)$:
\begin{equation}
    |\mc{X}|_{p(x)}^{\delta_{min}} := \\ \min\left\{|\{x \in \mc{X} : p(x) > \delta_{min}\}|, |\mc{X}|\right\}.
\end{equation}}
This notion of alphabet size generalizes the usual method of measuring the size of a state space. When we think about the CVA objective, the state space size will be thought of in relation to this notion of state space size, under a given state distribution.

\subsubsection{DIB Upper Bounds the CVA Objective}

Recall that the given MDP $M$ paired with the fixed control policy $\pi_E$ defines an information-generating source. At each time step, a state is sampled from $\rho_E$ and given to a learning agent through a state-abstraction function, $\phi : \mc{S} \ra \mc{S}_\phi$, which projects each state to each abstract state $s_\phi$. I make the additional simplifying assumption that there exists a fixed policy $\pi_E$ that controls the MDP. The agent's goal is to perform as well as the demonstrator using as small of a state space as possible, as reflected by $\mc{J}$. This reference policy $\pi_E$ may be the optimal policy $\pi^*$, but it could also be something else, such as the agent's policy on a previous episode.

I now construct the IB and DIB analogue objectives. First, let $I(S;S_\phi)$ denote the rate, where $S$ is a random variable indicating the probability of arriving in each state under $\rho_E$, and $S_\phi$ is a random variable indicating the probability of arriving in each abstract state under $\rho_E$ and projecting each ground state through $\phi$. Second, following the IB, let $\KL(\pi_E(\cdot \mid s) \mid \mid \pi_\phi(\cdot \mid s_\phi))$ denote the distortion for a given state $s$. The total distortion, then, is the KL in expectation under $\rho_E$:
\begin{equation}
    \bE_{s \sim \rho_E}\left[\KL(\pi_E(\cdot \mid s) \mid \mid \pi_\phi(\cdot \mid \phi(s_\phi)))\right].
\end{equation}

Further, suppose there exists a fixed, deterministic mapping from $\mc{S}_\phi$ to $\tilde{\mc{S}}$, with $\tilde{\mc{S}} = \mc{S}$. Thus, the distribution $p(\tilde{x} \mid x)$ is simply $p(s_\phi \mid s)$, which I henceforth abbreviate as $\phi$. Consequently, the following alignments emerge between the present objects of study (abstractions, policies) and those studied by the IB:
\begin{equation}
    p(\tilde{x}\mid x) \leadsto \phi, \hspace{4mm} p(\tilde{x}) \leadsto \rho_\phi, \hspace{4mm} p(y \mid \tilde{x}) \leadsto \pi_\phi,
\end{equation}
where $\rho_\phi$ is the stationary distribution over abstract states induced by $\pi_\phi$ and $\phi$. Thus, per \autoref{thm:c5_tishby_ib}, I next construct an objective function $\hat{\mc{J}}$ based on the IB:
\begin{equation}
\label{eq:c5_sa_ibm_lagrange}
    \hat{\mc{J}}_{\text{IB}}\left[\phi ; \rho_\phi ; \pi_\phi \right] := I(S;S_\phi)\ + \underset{{s \sim \rho_E, s_\phi \sim \phi(s)}}{\mathbb{E}}[\beta \KL(\pi_E(\cdot \mid s) \mid\mid \pi_\phi(\cdot \mid s_\phi))]. 
\end{equation}

If we choose to use the DIB instead, then we only consider deterministic state abstraction functions $\phi : \mc{S} \ra \mc{S}_\phi$, and so $H(S_\phi \mid S) = 0$. Therefore, the DIB analogue objective is expressed as:
\begin{equation}
    \hat{\mc{J}}_{\text{DIB}}\left[\phi ; \rho_\phi ; \pi_\phi \right] := H(S_\phi)\ + \underset{{s \sim \rho_E}}{\mathbb{E}}[\beta \KL(\pi_E(\cdot \mid s) \mid\mid \pi_\phi(\cdot \mid \phi(s)))]. 
    \label{eq:c5_dib_proxy_objective}
\end{equation}

With these objectives in place, I now build toward the main theorem of this chapter, which relates $\hat{\mc{J}}_{\text{DIB}}$ to $\mc{J}$. To prove the theorem, I first introduce two key lemmas. The first relates the entropy of a pmf to the maximum size of the alphabet used by that pmf:
\begin{lemma}
\label{lem:c5_entropy_sa_size}
    Consider a discrete random variable $X$, with alphabet $\mc{X}$ and some pmf $p(x)$. For a given threshold $\delta_{min} \in (0,1)$, the pmf-used alphabet size of the alphabet is bounded, relative to some pmf $p(x)$:
    \begin{equation}
         |\mc{X}|_{p(x)}^{\delta_{min}} \leq \frac{H(X)}{\delta_{min}\log \left(\frac{1}{\delta_{min}}\right)}.
    \end{equation}
\end{lemma}

\input{proofs/c5/c5_entropy_bounds_alphabet_size.tex}

This bound is relatively loose; we know trivially that $H(X) \leq \log_2 |\mc{X}|$. Thus, in the worst case, the bound can be up to $\tilde{O}\left(1 / \delta_{min}\right)$ times larger than the true alphabet. Still, this result allows us to relate the entropy of a random variable with its used alphabet size. Further, by definition, the entropy of the abstract stationary distribution, $H(\rho_\phi)$, gives us a lower bound on the number of bits needed to represent the used parts of $\mc{S}_\phi$. In this way, the entropy as a measure of compression is exploiting the fact that the most probable state can be written as \texttt{0}, the second most probable state as \texttt{10}, and so on. Thus, a lower entropy is already indicative of reducing $|\mc{S}_\phi|$. Further, in experiments, we will find this upper bound is loose relative to the size of the abstract state space the algorithm produces.

Next, I introduce a second lemma that relates the expected KL divergence between two policies to the difference in value achieved by the policies, in expectation under some state distribution:
\begin{lemma}
\label{lem:c5_kl_val_bound}
 Consider two stochastic policies, $\pi_1$ and $\pi_2$ on state space $\mc{S}$, and a fixed probability distribution over $\mc{S}$, $p(s)$. If, for some $k \in \mathbb{R}_{\geq 0}$:
    \begin{equation}
     \underset{p(s)}{\bE}\left[\KL(\pi_1(\cdot \mid s) \mid \mid \pi_2(\cdot \mid s))\right] \leq k,
    \end{equation}
    then:
    \begin{equation}
         \underset{p(s)}{\bE}\left[ V^{\pi_1}(s) - V^{\pi_2}(s)\right] \leq \sqrt{2k}\textsc{VMax},
    \end{equation}
    where $\textsc{VMax}$ is an upper bound on the value-function. 
\end{lemma}

\input{proofs/c5/c5_exp_kl_val_bound.tex}

The above bound relates the distortion measure present in IB to that of the CVA objective. Note that this bound is vacuous for values of $k \geq \frac{1}{2}$.

With these lemmas in place, I now present the theorem.
\begin{theorem}
    \label{thm:c5_main_objective_bound}
    A variation of the DIB objective $\hat{\mc{J}}_\text{DIB}$ is an upper bound for the CVA objective, $\mc{J}$, where state space size is treated as $|\mc{S}_\phi|_{\rho_\phi(s)}^{\delta_{min}}$. Formally, for all $\phi \in \Phi$:
    \begin{equation}
    |\mc{S}_\phi|_{\rho_\phi(s)}^{\delta_{min}} + \beta \underset{\rho_E(s)}{\bE}\left[V^{\pi_E}(s) - V^{\pi_\phi^*}(s)\right] \leq \frac{H(\rho_\phi)}{\delta \log{\frac{1}{\delta}}} + 2\textsc{VMax}\beta \underset{\rho_E(s)}{\bE}\left[ \KL(\pi_E(s) \mid\mid \pi_\phi^*(s)\right].
    \end{equation}
\end{theorem}

\input{proofs/c5/c5_proxy_rlit_objective.tex}


This theorem tells us that the optimization problem presented by $\hat{\mc{J}}_{\text{DIB}}$ can be well approximated by the usual IB method. I thus introduce Determinstic Information Bottleneck for State abstractions (\textsc{Dibs}, presented in \autoref{alg:dibs}), a simple iterative algorithm that adapts the DIB to Apprenticeship Learning with state abstractions. \textsc{Dibs}\ outputs a state-abstraction--policy pair in finite time that computes a local minimum of $\hat{\mc{J}}_\text{DIB}$, which we know from \autoref{thm:c5_main_objective_bound} is an upper bound on $\mc{J}$. The pseudocode presented is for the \textit{deterministic} variant of the IB, as often state abstraction functions are treated as deterministic aggregation functions~\cite{li2006towards}. The stochastic variant, which I call \textsc{Sibs}, will also be of interest, as soft state aggregation has been explored as well~\cite{singh1995reinforcement}.

\input{algorithms/dibs_alg.tex}

\section{Experiments}
\label{sec:c5_experiments}
I now describe several experiments that explore the power of \textsc{Dibs}\ for constructing abstractions that trade off between compression and value. First, I study the traditional Four Rooms domain discussed in \autoref{chap:background}. Second, I present a simple extension to \textsc{Sibs} that scales to high-dimensional state spaces and evaluate this extension in the Atari game Breakout using the Arcade Learning Environment (ALE) \cite{bellemare2013arcade}. The code for running these experiments is freely available for reproduction and extension.\footnote{\url{https://github.com/david-abel/rl_info_theory}}

\subsection{Four Rooms}

I first investigate the power of \textsc{Dibs} to appropriately trade off between value loss and compression.

The first experiment focuses on the Four Rooms grid world domain discussed in \autoref{fig:c2_four_rooms}. Recall that the agent interacts with an $11 \times 11$ grid with walls dividing the world into four connected rooms. The agent has four actions, \texttt{up}, \texttt{left}, \texttt{down}, and \texttt{right}. Each action moves the agent in the specified direction with probability $0.9$ (unless it hits a wall), and orthogonally with probability $0.05$. The agent starts in the bottom left corner, and receives $+1$ reward for transitioning into the top right state, which is terminal. All other transitions receive 0 reward. I set $\gamma$ to 0.99. For simplicity, I set the expert policy $\pi_E$ to be the optimal policy, with an additional $\eps = 0.05$ probability of taking an action at random to ensure that an arbitrary stochastic policy over the action space has overlapping support with the expert policy.

In Four Rooms, I run \textsc{Dibs}\ and \textsc{Sibs} to convergence and compare the value of $\pi_{\phi,\textsc{Dibs}}$ and $\pi_{\phi,\textsc{Sibs}}$ to the value of the demonstrator policy for $\beta$ between 0 and 4, incrementing by $0.2$. I determine convergence as per line 9 of \autoref{alg:dibs}: if all updating functions change by no more than $\Delta$, the algorithm has converged. I set $\Delta$ to 0.001, an arbitrarily chosen small constant.

\begin{figure}[b!]
    \centering
    \subfloat[Trade off made by the $\phi$'s output by \textsc{Dibs} for different values of $\beta$. \label{fig:c5_rate_dist_four_rooms}]{  \includegraphics[width=0.45\textwidth]{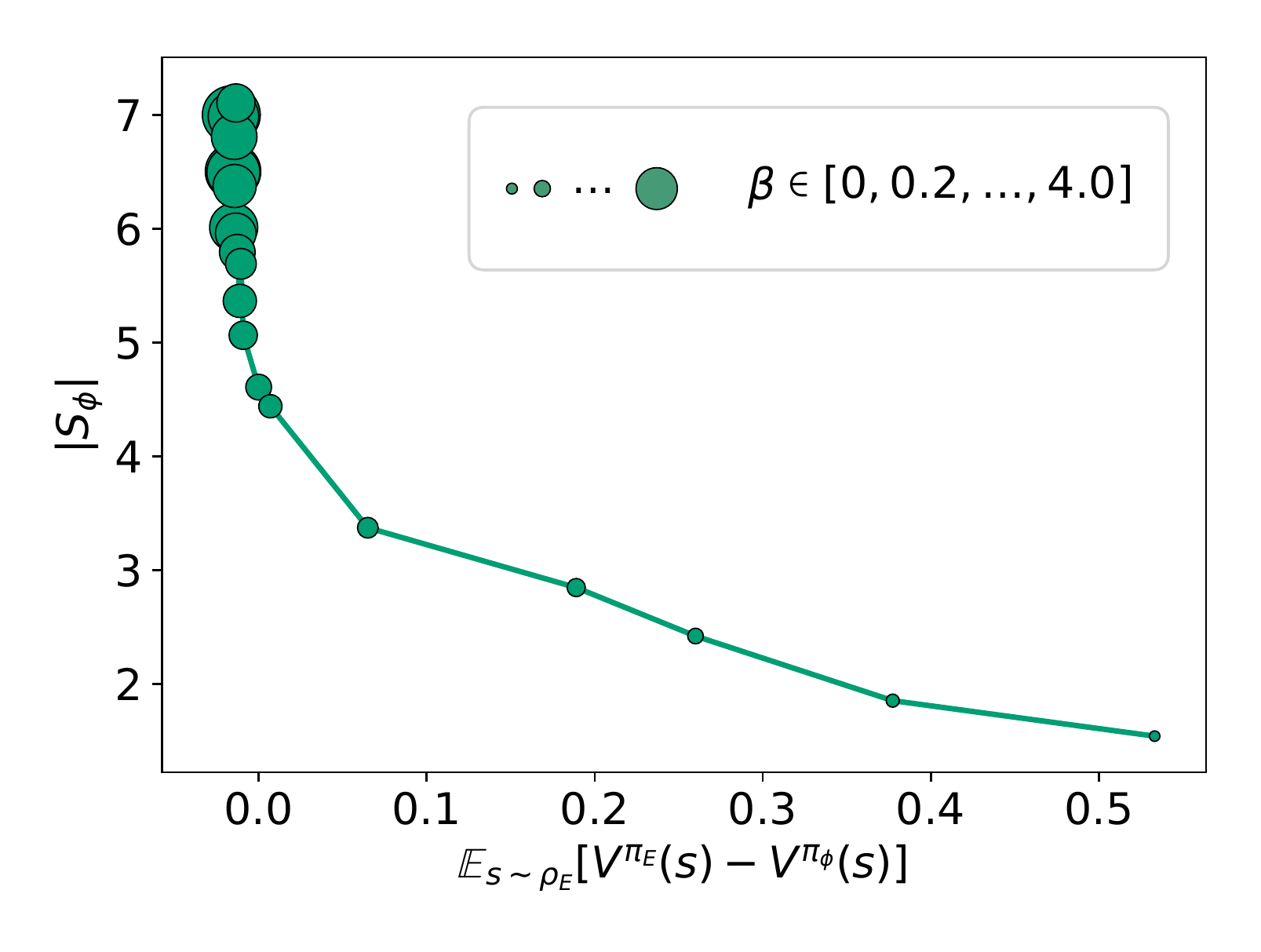}} \subfhspace
    \subfloat[Four Rooms: $V^\pi(s_0)$ vs. $\beta$\label{fig:c5_four_rooms_beta_val}]{\includegraphics[width=0.45\textwidth]{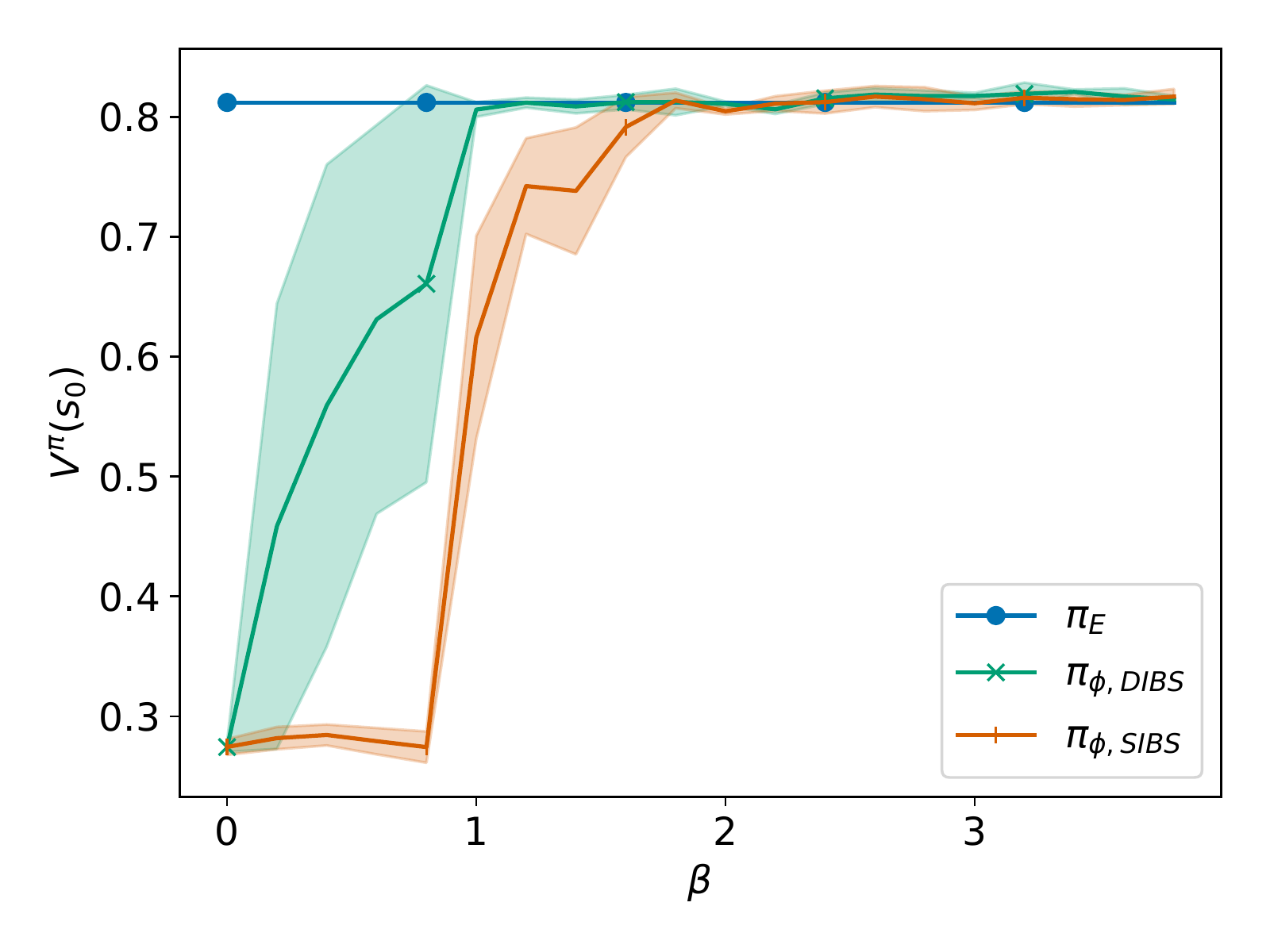}} 

    \caption{(a) The average rate-distortion trade off made by \textsc{Dibs} as $\beta$ varies, and (b) The average value of the $\phi,\pi_\phi$ pairs found by \textsc{Dibs}\ for different values of $\beta$.}
    \label{fig:c5_four_room_results}
\end{figure}

\autoref{fig:c5_rate_dist_four_rooms} illustrates the rate-distortion trade off made by 500 different runs of \textsc{Dibs}\ with different settings of $\beta$ ranging from $0.0$ to $4.0$: each point indicates the size of the average abstract state space (y-axis) and value of the abstract policy relative to the demonstrator (x-axis), achieved by the computed state abstraction for the given $\beta$. The large values of $\beta$ correspond to larger circles, with the maximal value of $\beta = 4.0$ appearing in the top left of the curve, and the smallest of $\beta=0.0$ appearing at the bottom right. Since the demonstrator is in fact suboptimal due to the $\eps$-randomness, it is possible for the abstract policy to do slightly better than the demonstrator in terms of reward, as is the case for all points with x-value less than 0.0. However, there is no incentive in the objective for this to be the case, as reward is not yet incorporated into learning in any way. Observe that when $\beta = 0$, \textsc{Dibs}\ prioritizes compression, yielding a one-state MDP on average (the far right point). As $\beta$ increases (which moves along the line to the left), the algorithm gradually tips the trade off from prioritizing compression to prioritizing performance. As $\beta$ increases, we see the abstract policy achieve the same value as the demonstrator. Also of note is that a two-state abstract space is capable of representing a policy (which could be stochastic) that is nearly as effective as the expert policy. With a one state MDP, however, the best policies found by \textsc{Dibs}\ still yield around $0.45$ expected value loss relative to $\pi_E$.

\autoref{fig:c5_four_rooms_beta_val} offers a slightly different perspective on the same results. Here, I present the average value of the abstract policy achieved as a function of $\beta$, with $\beta$ again varying from $0$ to $4.0$. The results are averaged over the same 500 runs, with the lines indicating the average and shaded regions denoting 95\% confidence intervals. Notably, when $\beta$ is 0, the algorithm tends to find a policy that achieves significantly worse than the expert. With \textsc{Dibs}, as $\beta$ increases, we see rapid improvement in the quality of the discovered policy, up until $\beta = 1$, at which point the abstract policy achieves almost identical performance to the expert. In contrast, the stochastic variant \textsc{Sibs} sees effectively no improvement in the quality of the policy until $\beta = 1$. I conjecture that this is due to the more difficult optimization problem presented, as the space of probabilistic state abstraction may be much harder to search through than the space of deterministic ones. Still, as $\beta$ increases past one, both $\pi_{\phi,\textsc{Dibs}}$ and $\pi_{\phi,\textsc{Sibs}}$ nearly match the value of the control policy.

\begin{figure}[t!]
    \centering

    \subfloat[$\phi$ with $\beta = 0$, $|\mc{S}_\phi| = 1$ \label{fig:c5_beta_0}]{\includegraphics[width=0.23\columnwidth]{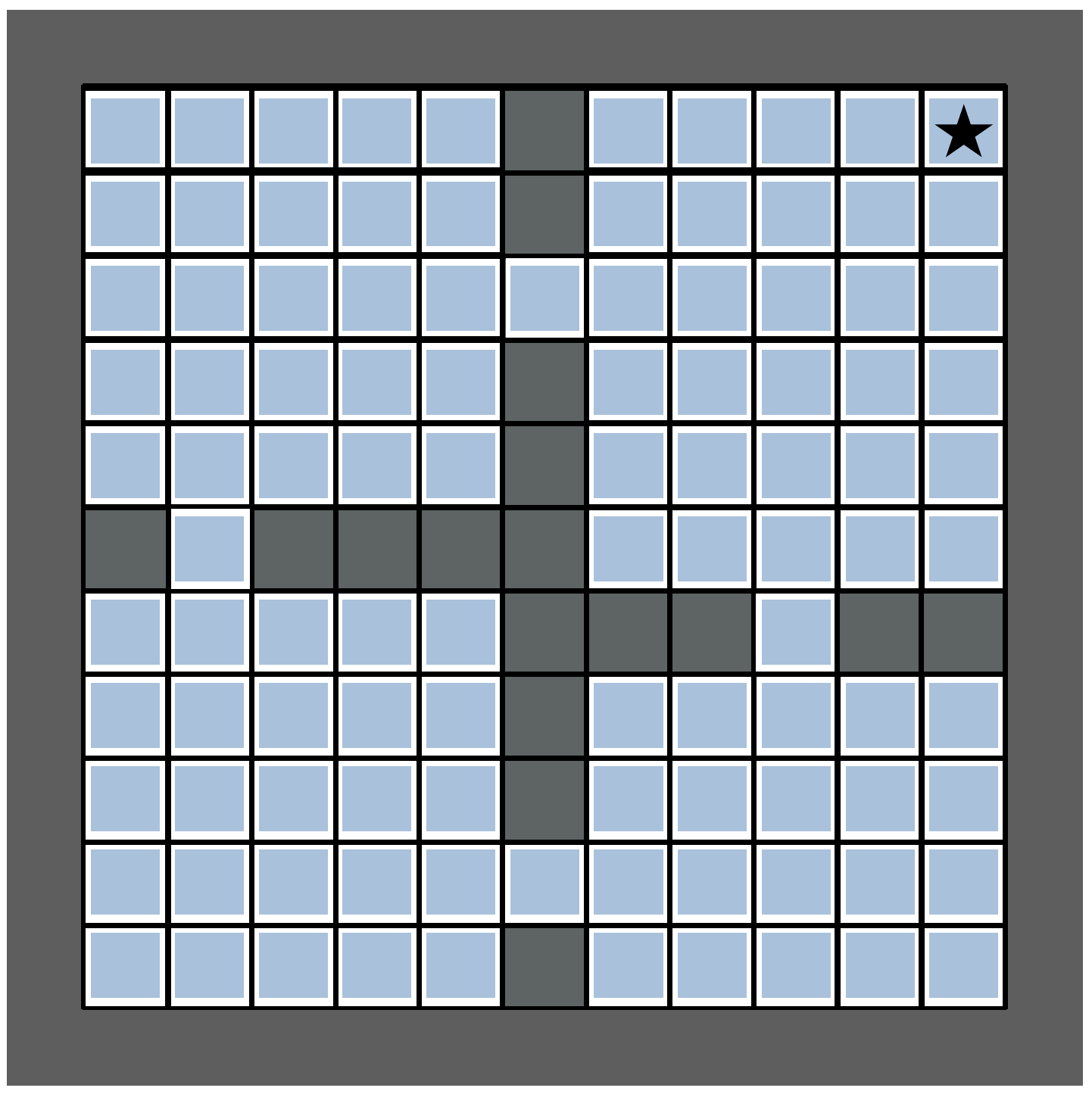}} \hspace{3mm}
    \subfloat[$\phi$ with $\beta = 1$, $|\mc{S}_\phi| = 4$ \label{fig:c5_beta_1}]{\includegraphics[width=0.23\columnwidth]{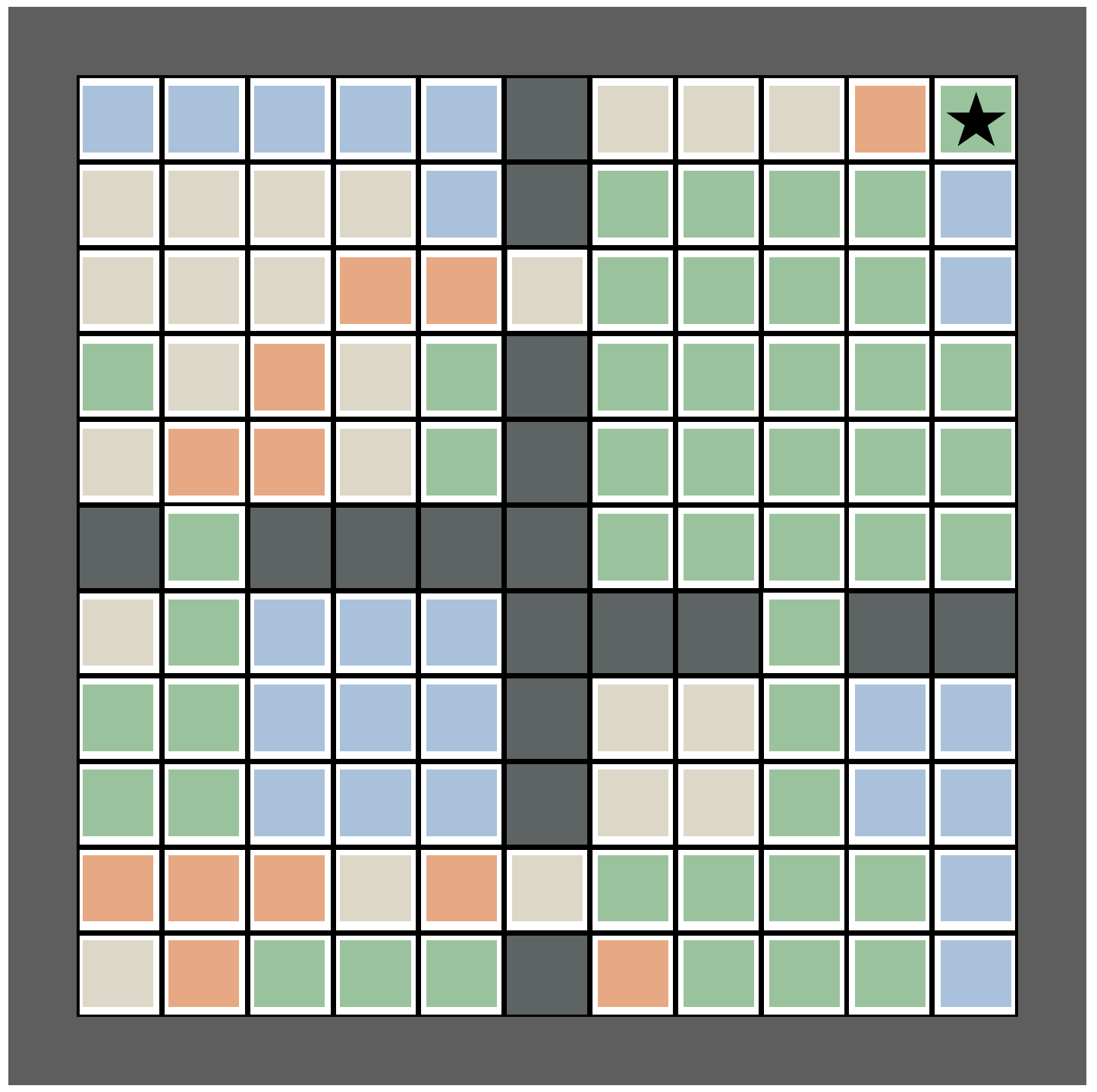}} \hspace{3mm}
    \subfloat[$\phi$ with $\beta = 2$,  $|\mc{S}_\phi| = 5$ \label{fig:c5_beta_2}]{\includegraphics[width=0.23\columnwidth]{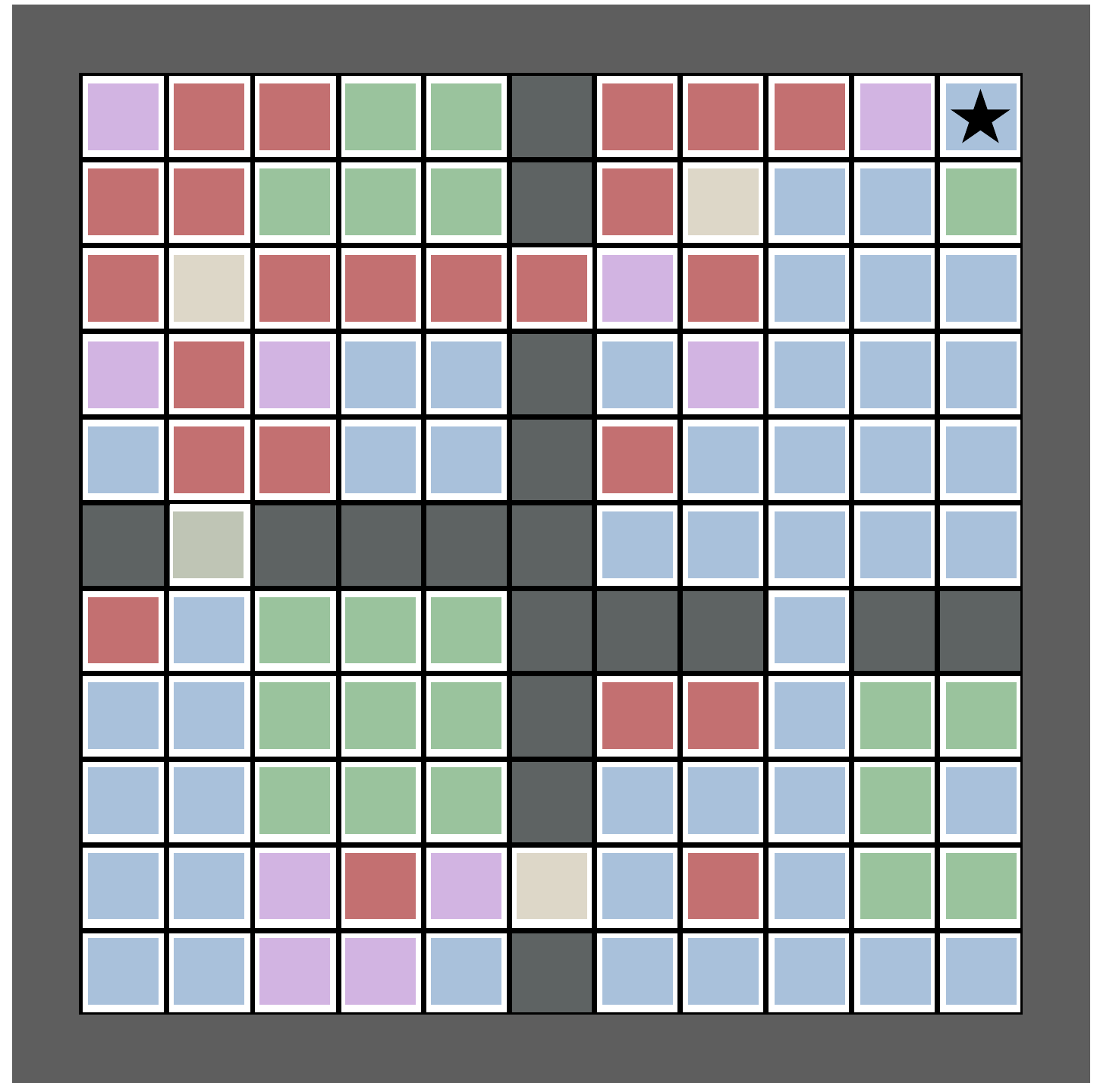}} \hspace{3mm}
    \subfloat[$\phi$ with $\beta = 20$, $|\mc{S}_\phi| = 9$ \label{fig:c5_beta_20}]{\includegraphics[width=0.23\columnwidth]{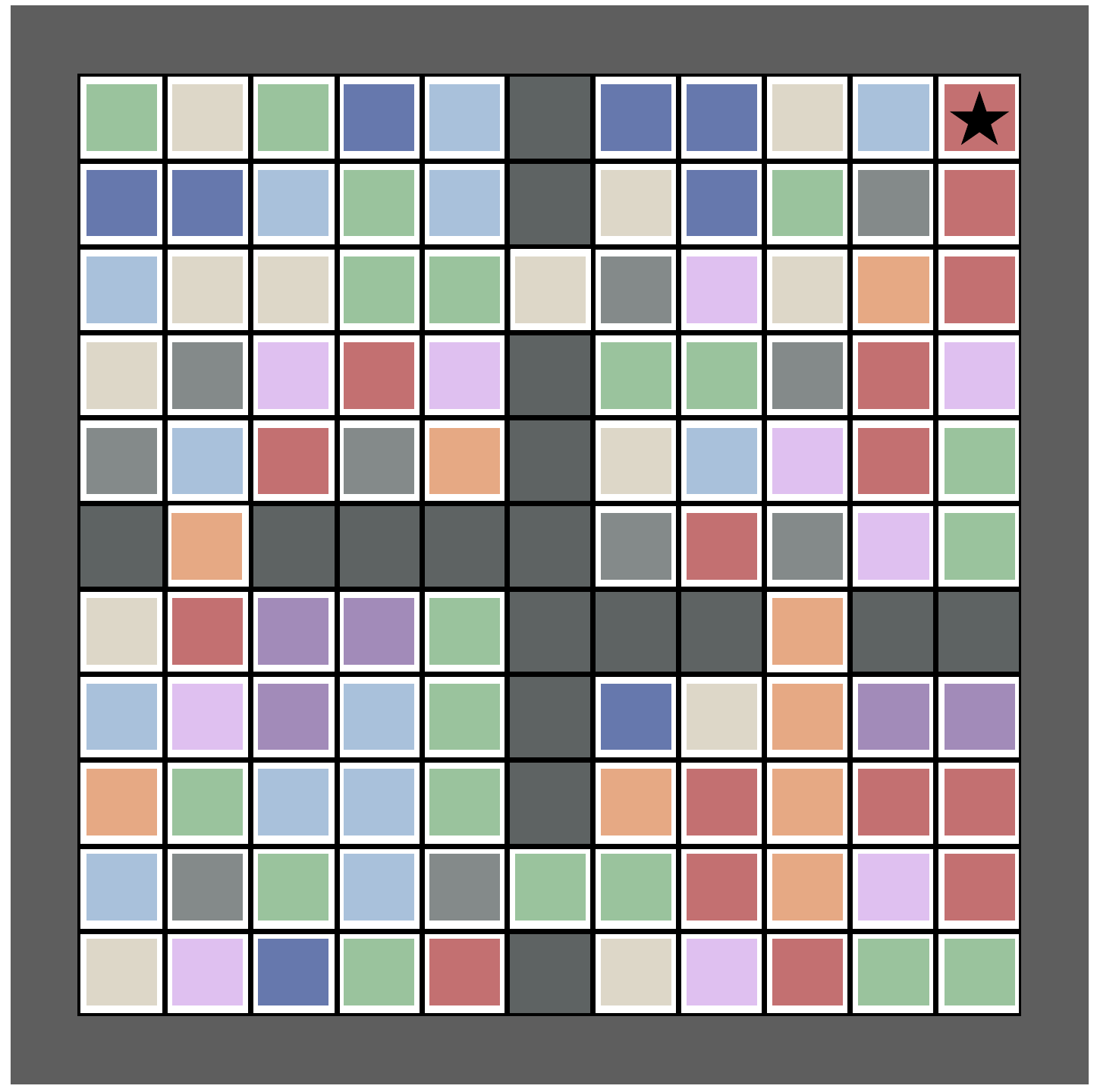}} 
    \caption{The state abstractions found by \textsc{Dibs}\ in the Four Rooms domain when (a) $\beta=0$, (b) $\beta = 1$, (c) $\beta = 2$, and (d) $\beta = 20$ .}
    \label{fig:c5_four_room_visuals}
\end{figure}

\autoref{fig:c5_beta_0}, \autoref{fig:c5_beta_1}, \autoref{fig:c5_beta_2}, and \autoref{fig:c5_beta_20} show the state abstractions found by \textsc{Dibs} for $\beta=0$, $\beta = 1$, $\beta = 2$, and $\beta = 20$ respectively. Notably, each of these state abstractions were sufficient for effectively solving the problem escept for the $\beta=0$ case. For the abstraction in \autoref{fig:c5_beta_1}, there are four abstract states, which is sufficient for nearly representing a $\pi^*$-irrelevance abstraction of the demonstrator policy (``move up" in green, ``move right" in tan, and so on). As $\beta$ is increased, observe that the abstraction yields far more states, though the quality of the optimal policy is already as high as it can go (there is simply less pressure to compress). Note that these experiments only examine the effect of the state abstractions found by \textsc{Dibs} on representation of the optimal policy. However, intuitively, the abstract state spaces pictured in \autoref{fig:c5_beta_1}, \autoref{fig:c5_beta_2}, and \autoref{fig:c5_beta_20} will give rise to different degrees of \textit{learning} difficulty, even though they can each represent a near-optimal policy.

I now investigate the impact of these state abstractions on learning. In particular, I study how the resulting state abstractions change learning for simple RL algorithms in the Four Rooms grid world. The experiment proceeds as follows. First, before any RL algorithm interacts with an MDP, I construct state abstractions using \textsc{Dibs} using different values of $\beta$, with all other settings as described in the previous experiment. Then, I give the resulting state abstractions to $Q$-learning on Four Rooms and contrast the different algorithm-$\phi$ pairs' behavior.

\begin{figure}
    \centering
    \subfloat[$Q$-learning, 100 episodes]{\includegraphics[width=0.45\textwidth]{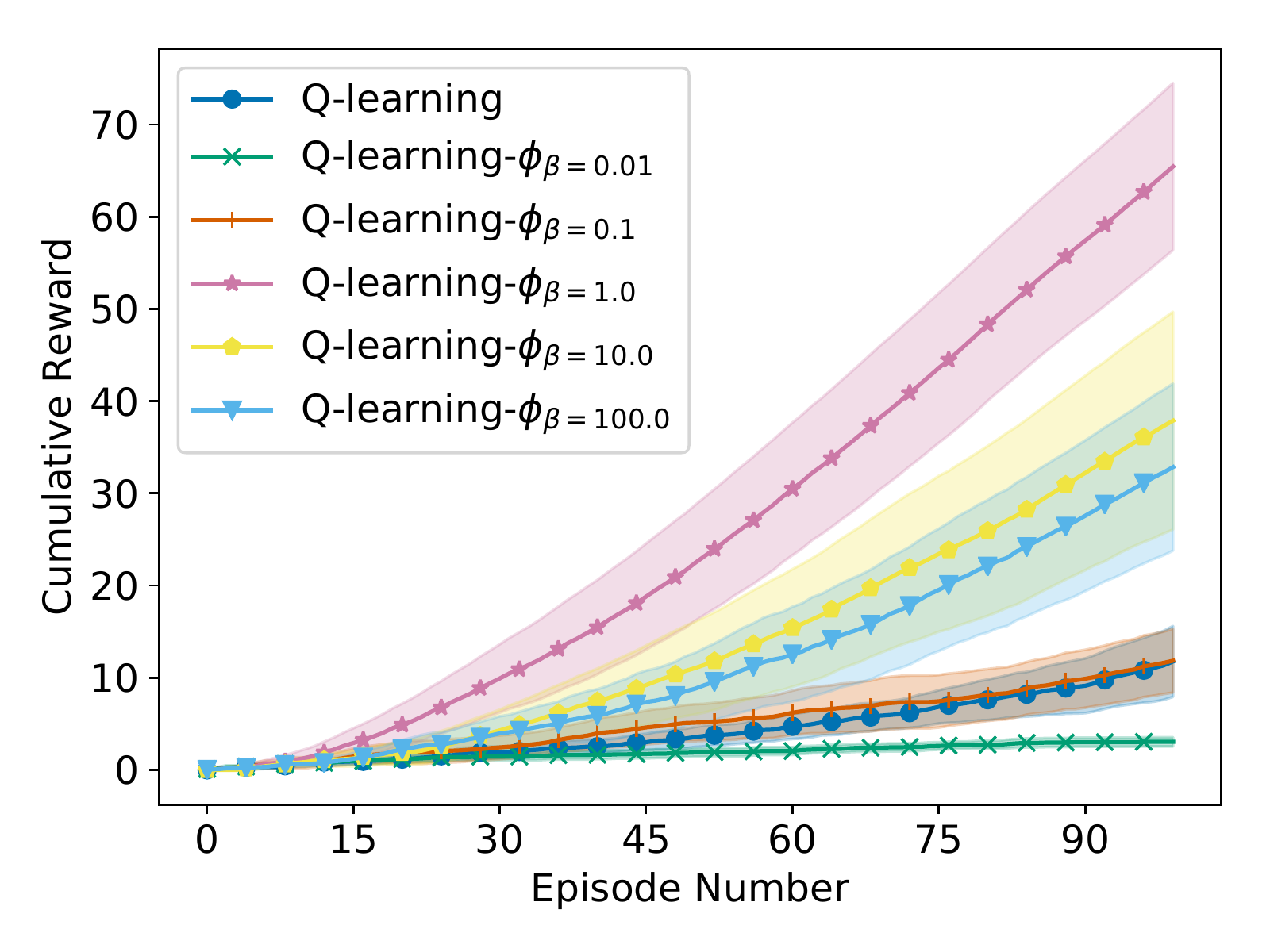}} \subfhspace
    \subfloat[$Q$-learning, 1,000 episodes]{\includegraphics[width=0.45\textwidth]{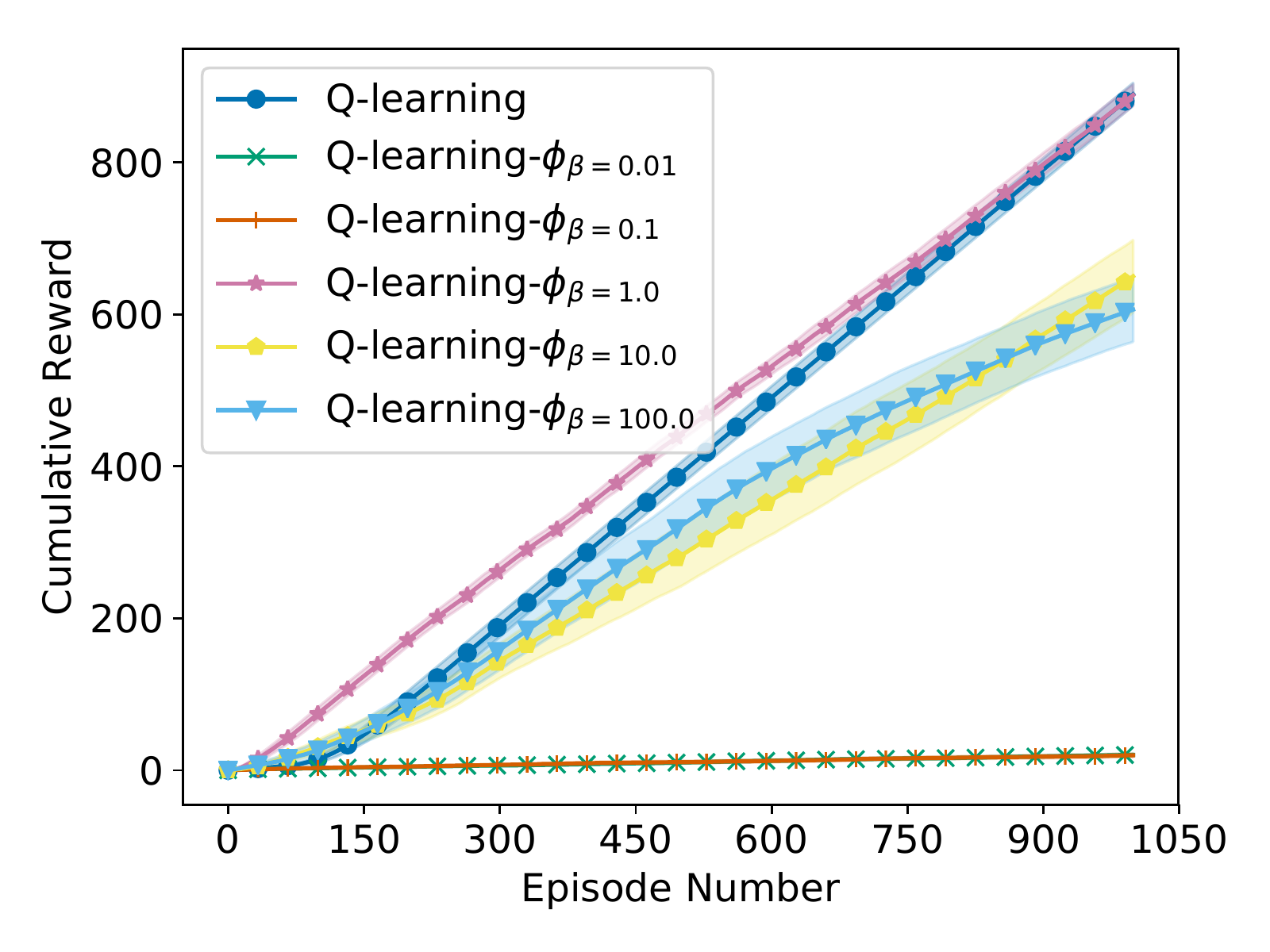}}
    
    
    \caption{A comparison of how $\phi$ with different choice of $\beta$ impacts simple RL in the Four Rooms domain.}
    \label{fig:c5_dibs_learning_grid}
\end{figure}

Results are presented in \autoref{fig:c5_dibs_learning_grid}. Each plot is the result of 25 runs of the experiment, indicating the mean cumulative reward of different learning algorithm--$\phi$ combinations. The $\beta$ parameter was chosen to range from $0.01$ up to $100.0$, with each order of magnitude in between. Observe that, with a smaller data set (left), $Q$-Learning is able to more quickly find a better policy using state abstractions resulting from middle choices of $\beta$. In particular, the best performance within 100 episodes is obtained by the approach using $\beta=1.0$. As $\beta$ decreases, performance decays rapidly, and as $\beta$ increases, performance worsens as well. In contrast, with a \textit{larger} data set---that is, when the number of episodes is set to 1,000---$Q$-learning is able to ultimately overtake all approaches using a state abstraction, but less efficiently than when $\beta$ is set to the ideal value of $1$ for the domain. This is precisely because $Q$-learning is guaranteed to eventually find the optimal policy, whereas the variations using $\phi$ will learn a best-in-class policy, but relative to a restricted class. This is precisely a bias-complexity trade off; choice of $\beta$ is determining how large the abstract state space is, and thus the space of possible $Q$ functions learnable by the algorithm.

\subsection{Breakout}

I next translate the proposed algorithmic framework into domains with high-dimensional observations. To do so, I turn to variational autoencoders (VAEs) \cite{kingma2014auto}. In a VAE, we are concerned with learning a compact latent data representation, $z$, that captures a high-dimensional observation space, $x$, through the use of two parameterized functions $q_\psi(z \mid x)$ and $p_\theta(x \mid z)$. The pair represent a probabilistic encoder and decoder, typically captured by two separate neural networks, where the former maps data to a latent representation and the latter maps from $z$ to the original observation. Traditionally, the two models are trained jointly to optimize the evidence lower bound objective (ELBO), which maximizes $\mathbb{E}_{q_\psi(z \mid x)}[\log p_\theta(x \mid z)]$ to facilitate reconstruction of the original data and minimizes $\KL(q_\psi(z \mid x) \mid \mid p(z))$ to keep $q_\psi(z \mid x)$ close to some prior $p(z)$ over latent codes. Both $q_\psi(z \mid x)$ and $p(z)$ are commonly treated as Gaussian to use the Gaussian reparameterization trick~\cite{kingma2014auto}. Since the ELBO is optimized in expectation over the data distribution, $p(x)$, we can leverage a known result regarding the KL divergence
\cite{kim2018disentangling}:
\begin{equation}
    \label{eq:c5_kl_mi_ub}
    \mathbb{E}_{p(x)}[\KL(q_\psi(\cdot \mid x) \mid \mid p(z))] = I(X;Z) + \KL(q (z) \mid \mid p(z)) \geq I(X;Z),
\end{equation}
where $q(z) = \mathbb{E}_{p(x)}[q_\psi(z \mid x)]$. I treat $x$ as the ground state representation $\mc{S}$ and $z$ as the abstract state $\mc{S}_\phi$. I derive a new objective function that serves as a variational upper bound to the stochastic IB (\textsc{Sibs}) objective derived in \autoref{eq:c5_sa_ibm_lagrange}:
\begin{align}
    \label{eq:c5_vae_objective}
    &\min_\phi \underset{{s \sim \rho_E}}{\mathbb{E}}[\KL(q_\psi(\cdot \mid s) \mid\mid p(s_\phi))\ + \beta \underset{{s_\phi \sim \phi(s)}}{\mathbb{E}}[\KL(\pi_E(\cdot \mid s) \mid\mid \pi_\phi(\cdot \mid s_\phi))]], \\
    &\geq \min_\phi I(S;S_\phi)\ + \beta \underset{{s \sim \rho_E, s_\phi \sim \phi(s)}}{\mathbb{E}}[\KL(\pi_E(\cdot \mid s) \mid\mid \pi_\phi(\cdot \mid s_\phi))],
\end{align}
where the upper bound follows directly from \autoref{eq:c5_kl_mi_ub}.

To make use of this upper bound, we first
create a demonstrator policy $\pi_E$ for the Atari game Breakout~\cite{bellemare2013arcade}
using A2C~\cite{Mnih2016AsynchronousMF}. A Gaussian VAE agent is then trained with a separate architecture that has the same first four layers as the A2C agent before mapping out to a mean and covariance (in $\mathbb{R}^{25}$). Instead of reconstructing states, this decoder serves as an abstract policy network, mapping to a final distribution over the primitive actions, $\pi_\phi(a \mid s)$, which is really $\pi_\phi(a \mid z)$. The model is trained via \autoref{eq:c5_vae_objective} for $2000$ episodes using the Adam optimizer \cite{Kingma2014AdamAM} with a learning rate of 0.0001. During training, the expert's policy ($\pi_E$) controls the MDP.

\begin{figure}
    \centering
    \subfloat[Breakout: Mean Reward vs. $\beta$\label{fig:c5_brkout_beta_val}]{\includegraphics[width=0.45\textwidth]{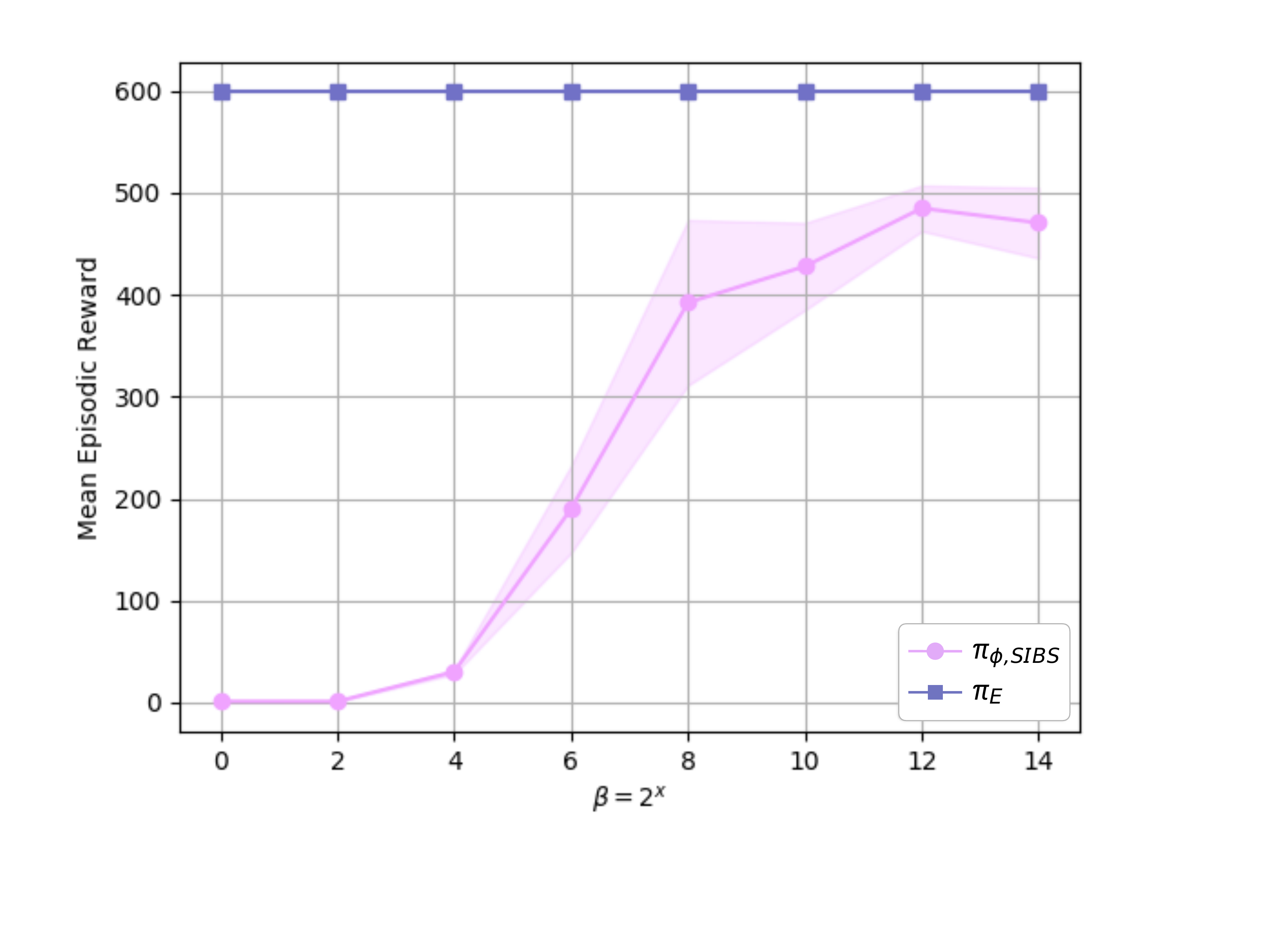}} \hspace{6mm}
    \subfloat[Original (top) and $\phi$ with $\beta = 2$ (middle) and $\beta = 2048$ (bottom) \label{fig:c5_brkout_beta}]{\includegraphics[width=0.45\linewidth]{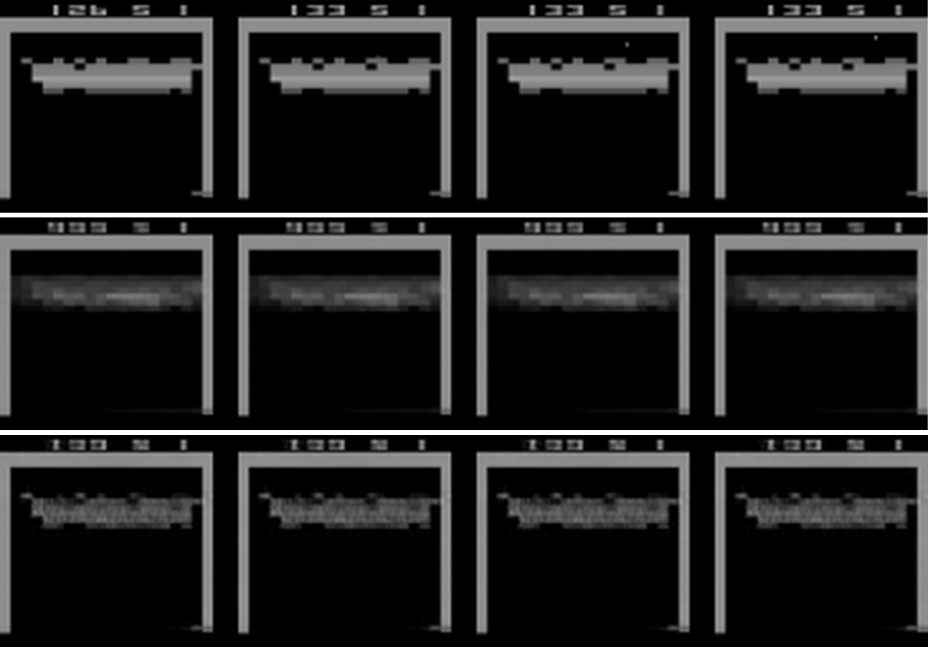}}
    \caption{(a) The mean reward over 100 evaluation episodes of $\phi, \pi_\phi$ combinations found by the VAE-approximation to \textsc{Sibs} for different values of $\beta$, and (b) attempted state reconstructions using fixed abstractions found when $\beta = 2$ and $\beta = 2048$.} 
    \label{fig:c5_brkout}
\end{figure}

\autoref{fig:c5_brkout} presents results showcasing the effect of $\beta$ on compression and performance in. The data suggest a relationship similar to that of \autoref{fig:c5_four_room_results}) between choice of $\beta$, success in approximating the demonstrator policy, and the nature of the resulting state abstraction. In the visualizations of the abstraction, observe that the quality of state reconstruction (each state is a row of four consecutive game screens) is compromised under a low setting of $\beta$ (prioritizing compression), whereas a higher value of $\beta$ preserves more information (paddle position and shape of bricks), leading to higher-quality reconstruction. Seeing that agent performance converges below the expert policy $\pi_E$, it is possible that increasing the size of the latent bottleneck may close the gap. Of critical importance to this setup is the use of a reconstruction network to visualize the latent state information, shown in \autoref{fig:c5_brkout_beta}. Due to the method of reconstruction, it may be difficult to determined what information is truly represented as opposed to what information is easily captured in the reconstruction.


So far, I have articulated a new formalism for treating state abstraction as compression in MDPs. To make the setting concrete, I adopt the apprenticeship learning perspective and assumed query access to an expert control policy $\pi_E$. Then, using this new perspective, I introduced a new objective and proved that it is well-approximated by an IB-like objective, giving rise to a convergent algorithm for computing state abstractions that trade off between compression and value.

\section{Extensions}
\label{sec:c5_rlit_extensions}

Many questions remain. First, it is natural to be interested in the case when $\pi_E$ does not control the MDP, but rather the RL agent's actions determine which states are occupied. In this case, unfortunately, no fixed state distribution is likely to be available. Second, the current formulation concentrates on constructing abstract state spaces that can represent high value policies, but do nothing to ensure that such policies are easy for RL algorithms to discover. An important direction going forward is to identify the role that a well structured state space plays in ensuring low sample complexity RL. Third, much of the chapter has focused on discrete state spaces. It is important to understand whether similar ideas can apply to continuous state spaces, too. Lastly, it is natural to consider the lifelong or multitask setting, in which an RL agent must learn to solve a variety of related tasks. I next discuss study some of these extensions.

\subsection{Agents Controls the MDP}
Relaxing the assumption that $\pi_E$ controls the MDP is essential for extending these ideas to traditional RL. I here propose a path toward removing the control policy $\pi_E$ by focusing on an intermediate goal: define an algorithm with the same properties as \textsc{Dibs}, but with the learning agent's \textit{non-stationary} policy controlling the underlying MDP instead of $\pi_E$. Ultimately, this will give rise to an algorithm that, after $T < \infty$ iterations, can produce an abstraction--policy pair such that, for some state distribution $p(s)$:
\begin{equation}
    \bE_{p(s)} \left[V^*(s) - V^{\pi_\phi^{T}}(s)\right] \leq f(\beta, T).
\end{equation}
The most challenging aspect of this setup is that the source distribution is no longer fixed, since the agent's policy will change over time as the agent learns and updates both $\phi$ and $\pi_\phi$. To this end, I present the following lemma that suggests that two policies that deviate by a bounded amount are guaranteed to share similar stationary distributions.

\begin{lemma}
    \label{lem:c5_l1_pol_l1_rho}
    Given two policies $\pi_1$ and $\pi_2$, if $\sup_{s \in \mc{S}} \sum_{a \in \mc{A}}|\pi_1(a \mid s)-\pi_2(a \mid s)|\leq \Delta$, for $\Delta \in \mathbb{R}_{\geq 0}$, then:
    \begin{equation}
        \sum_{s \in \mc{S}}|\rho_{\pi_1,s_0}(s)-\rho_{\pi_2,s_0}(s)|\leq \frac{\Delta \gamma}{1-\gamma},
    \end{equation}
    where $\rho_{\pi,s_0}$ denotes the stationary distribution over states under $\pi$, starting in state $s_0$.
\end{lemma}

\input{proofs/c5/c5_steady_state_distr_val_loss.tex}

\begin{corollary}
As a simple corollary of \autoref{lem:c5_l1_pol_l1_rho}, observe that,
\begin{eqnarray}
    |V^{\pi_1}(s)-V^{\pi_2}(s)| &=&|\sum_{t\in \mathbb{N}}\sum_{s \in \mc{S}}\rho^{t}_{\pi_1,s_0}(s)R(s)-\sum_{t}\sum_{s\in \mc{S}}\rho^{t}_{\pi_2,s_0}(s)R(s)|\\
    &\leq&\textsc{RMax}\sum_{t\in \mathbb{N}}(1-\gamma)\sum_{s \in \mc{S}}|\rho^{t}_{\pi_1,s_0}(s)-\rho^{t}_{\pi_2,s_0}(s)|\\
    &\leq&\textsc{RMax}(1-\gamma)\sum_{t\in \mathbb{N}}\Delta \gamma^t t\\
    &=&\textsc{RMax}\frac{\gamma\Delta}{1-\gamma}=\Delta\gamma \textsc{VMax}
\end{eqnarray}
\end{corollary}

This lemma is useful as it suggests that policies that are similar to one another will have similar stationary state distributions, too. In particular, in the case where $\pi_E$ no longer controls the MDP, but rather $\pi_\phi$ does, the state distribution of relevance will be $\rho_{\phi,t}$. If, however, over time we can ensure that after some $t > N$ updates, $\pi_{\phi,t} \approx \pi_E$, then we can construct a convergent algorithm for the agent-in-control setting. I leave this analysis as an open question.

I offer an initial variant of this algorithm that I call agent-controlled \textsc{Dibs}\ (AC-\textsc{Dibs}). I conduct an experiment similar to that of the previous section in the Four Rooms domain. Here, I run the entire process of AC-\textsc{Dibs}\ for $N$ rounds, each time to convergence, but letting the agent's \textit{initial} policy for that round define the stationary state distribution.

\begin{figure}[t!]
    \centering
    \includegraphics[width=0.45\textwidth]{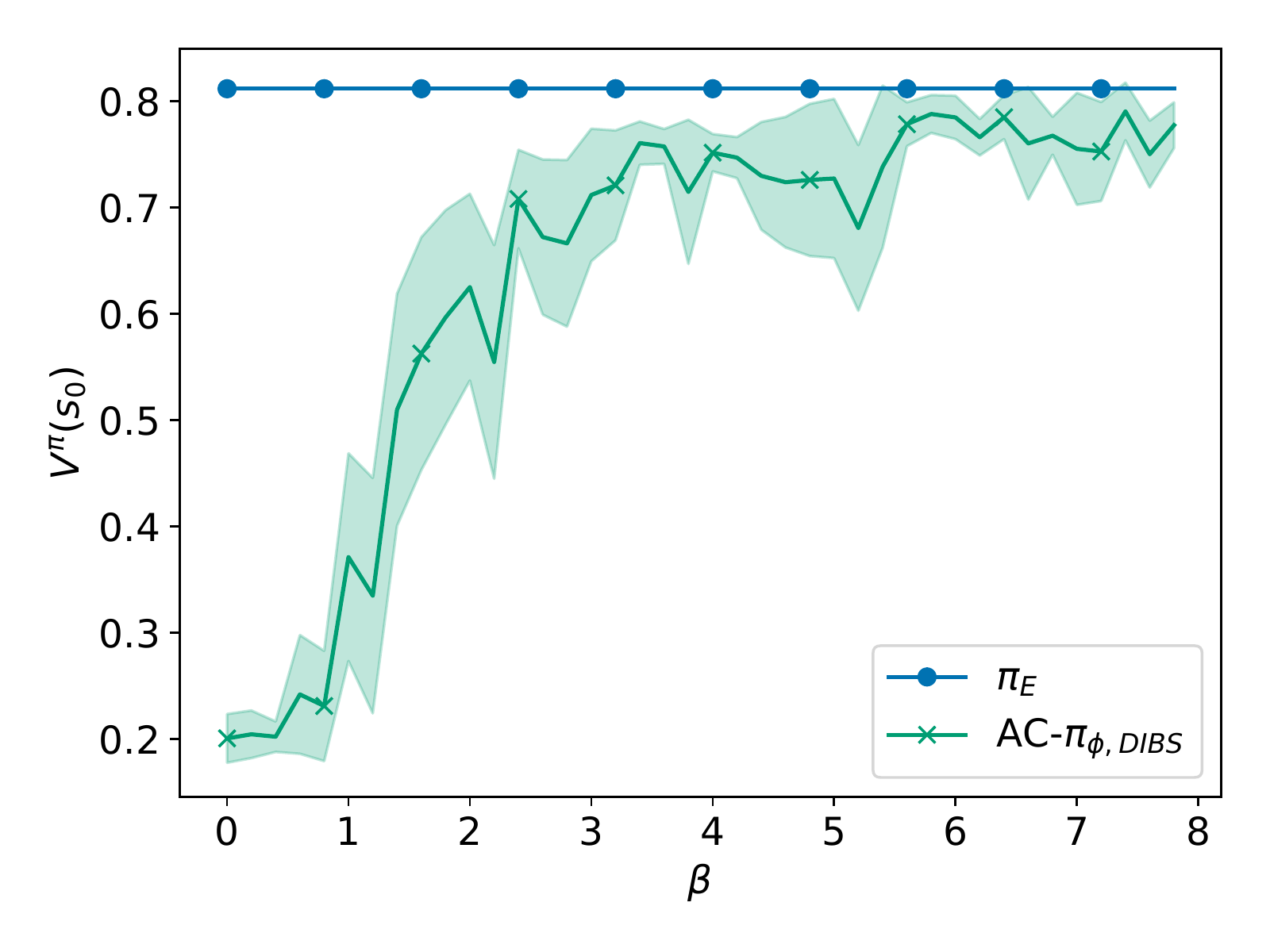}
    \caption{The value of the abstract policy found by AC-\textsc{Dibs} for values of $\beta$ between 0 and $4$.}
    \label{fig:c5_ac_dibs}
\end{figure}

Results are presented in \autoref{fig:c5_ac_dibs}. Surprisingly, AC-\textsc{Dibs}\ \textit{always} converges, yielding policies that are similar in value to $\pi_E$ for $\beta > 2$. This finding supports the previous speculation, along with \autoref{lem:c5_l1_pol_l1_rho}, that there is a feasible route to defining a convergent form of \textsc{Dibs} when the agent's policy controls the underlying MDP.

\subsection{Multiple MDPs}
Second, I use our framework to compute a single abstraction sufficient for representing the demonstrator policy across distinct but potentially related tasks. Concretely, I suppose we are given a set of MDPs $\mc{M}$, each sharing a state and action space, but are allowed to vary in $T$, $R$, and $\gamma$.

I conduct an experiment in which $|\mc{M}|=4$, where the four task are defined by a goal being in each of the four corners of the grid world. I run \textsc{Dibs}\ for each MDP in $\mc{M}$, for a fixed $\beta$, and form a global abstraction $\phi_\mc{M}$ by taking the intersection across each computed state abstraction. That is, for any state pair $(s_1, s_2)$, for $\phi_i$ computed by \textsc{Dibs}\ on each MDP, I define $\phi_\mc{M}$:
\begin{equation}
    \phi_\mc{M}(s_1) = \phi_\mc{M}(s_2) \equiv \bigwedge_{i=1}^{|\mc{M}|} \{\phi_i(s_1) = \phi_i(s_2)\}.
    \label{eq:c5_mt_dibs}
\end{equation}

\begin{figure}[t!]
    \centering
     \subfloat[$\beta = 0$ \label{fig:mt_dibs_beta_0}]{\includegraphics[width=0.3\columnwidth]{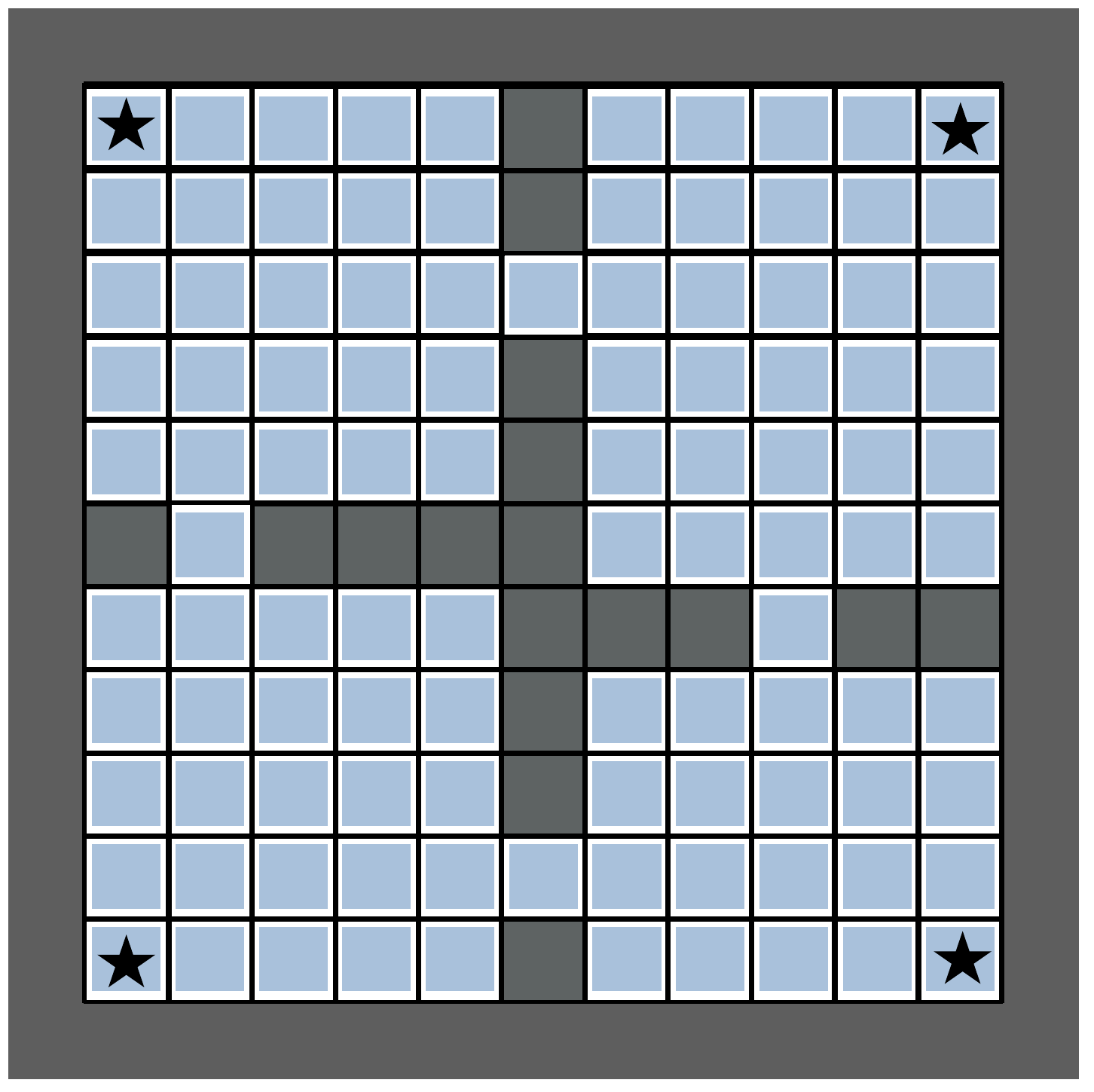}}\subfhspace
    \subfloat[$\beta = 10$ \label{fig:mt_dibs_beta_1}]{\includegraphics[width=0.3\columnwidth]{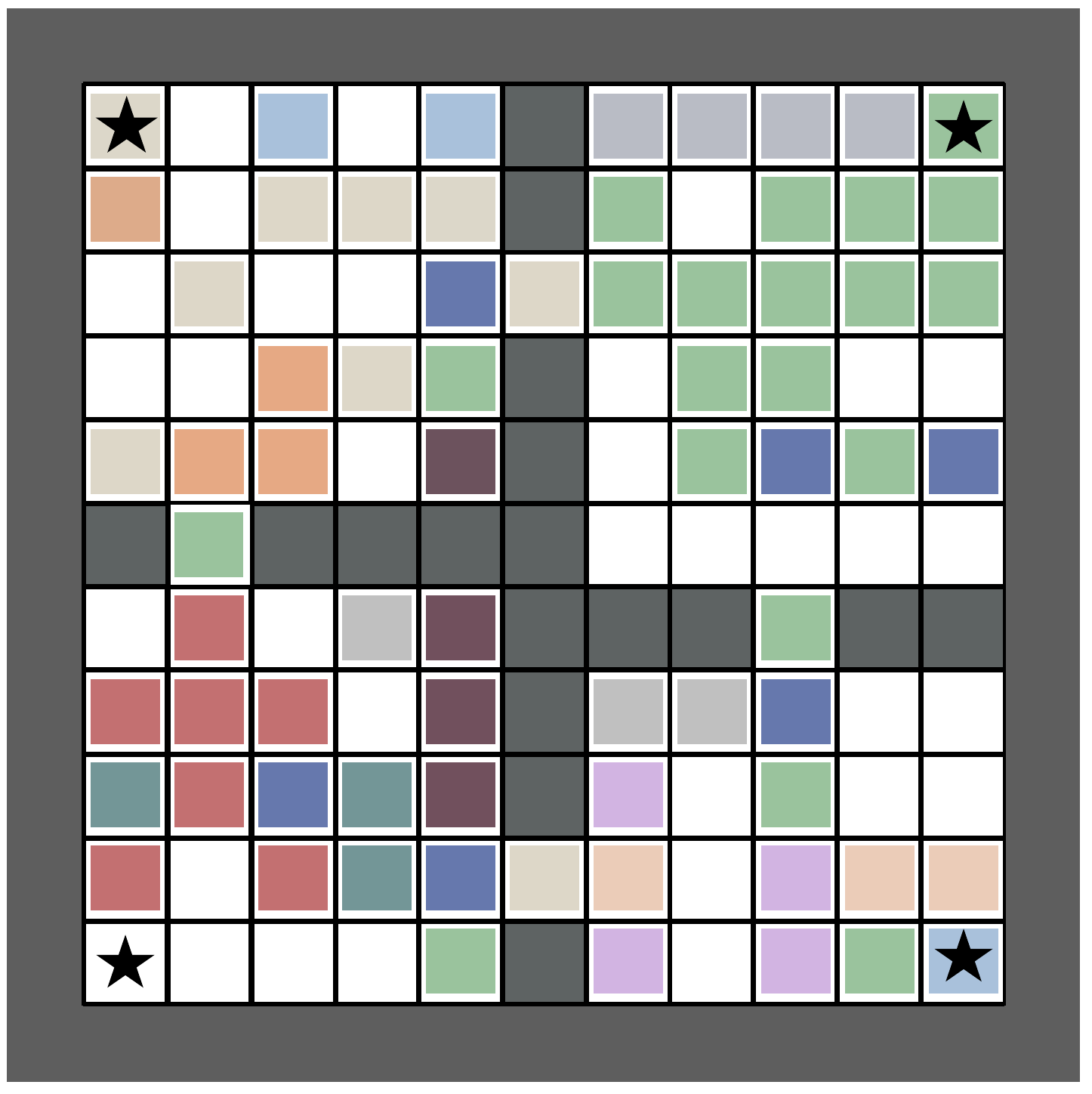}}\subfhspace
    \subfloat[$\beta = 20$ \label{fig:mt_dibs_beta_20}]{\includegraphics[width=0.3\columnwidth]{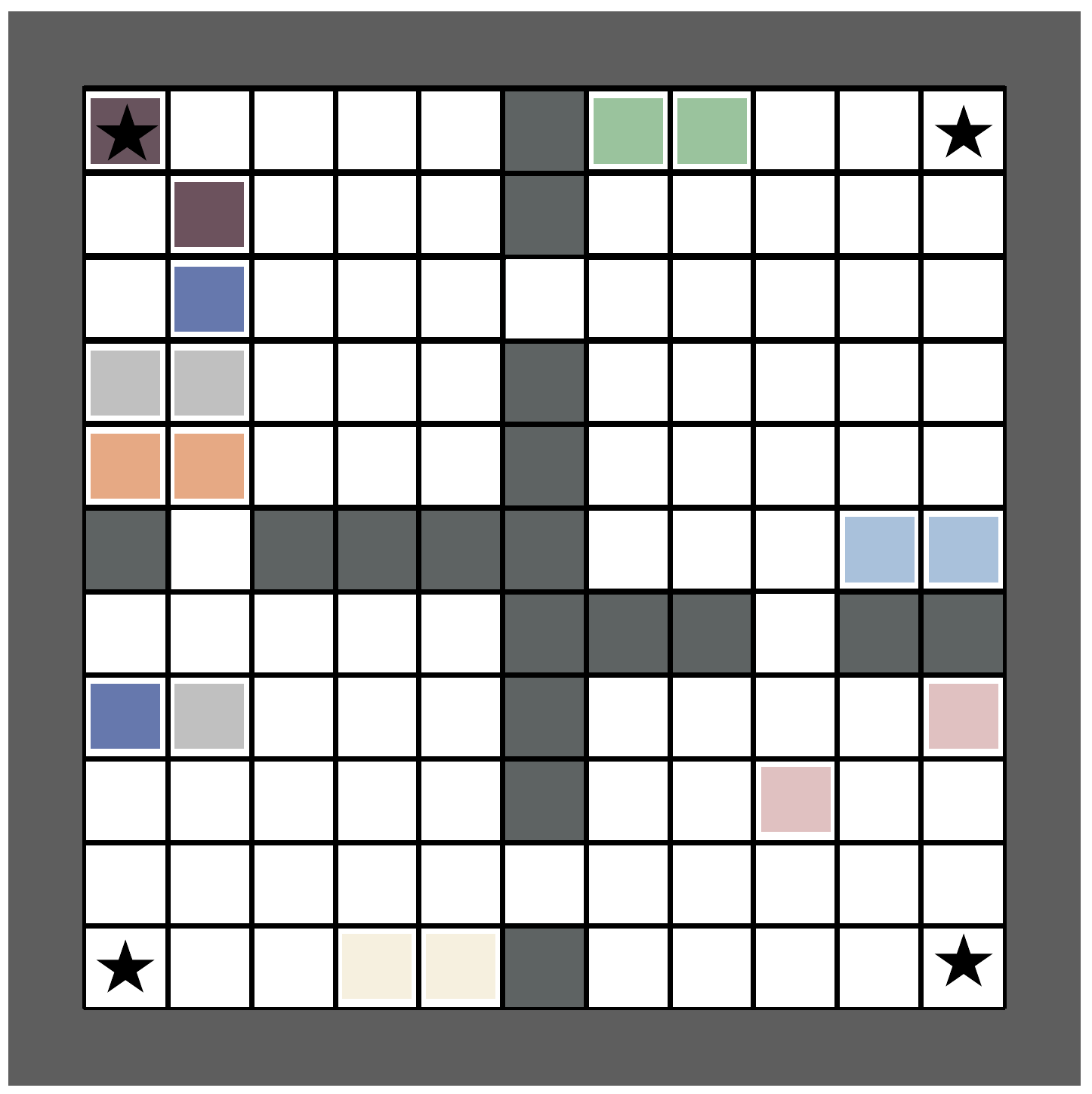}}
    \caption{State abstractions computed by \textsc{Dibs}\ for a collection of MDPs using \autoref{eq:c5_mt_dibs} for different values of $\beta$.}
    \label{fig:c5_lifelong_sa}
\end{figure}

\autoref{fig:c5_lifelong_sa} shows $\phi_\mc{M}$ for different values of $\beta$. All cells with the same color are grouped into the same state, except for white: all white states are each treated as their true ground state. Note that the abstraction becomes far more detailed as $\beta$ increases. When $\beta$ is close to 0, the algorithm prioritizes \textit{compression}, as is reflected by \autoref{fig:mt_dibs_beta_0}, which only has a single state. Conversely, as $\beta$ increases, the algorithm adds more distinctions between states, only grouping those that are close to one another or the near the same wall. When $\beta = 10$, the abstraction groups large regions of contiguous states together, such as the central group of states in the top right room, and many states in the center of the bottom left room. When $\beta = 20$, we find even less compression, but still see a few small contiguous regions that have some structural similarities. Critically, none of the abstractions are perfect. It is unknown what might constitute an ideal abstraction in this case, nor how well our proposed algorithm might approximate such an ideal.

\subsection{Learning $\phi$ in Continuous State Environments}

The investigation in this part of the dissertation so far has focused on discrete MDPs. It is important, however, to consider whether similar ideas are applicable when the underlying environmental state and action space are continuous, too. In this final section, I study the case when the underlying state space is continuous, summarizing work first presented by \citet{asadi2020sa_control}.

Concretely, I here introduce a new objective function that may be solved with standard stochastic gradient descent. Thus, we will develop a procedure simple for learning a state abstraction that maps a continuous state space into a discrete one given access to some number of trajectories of behavior on several training tasks. I then provide a generalization error bound on the quality of the learned state abstraction, assuming a fixed distribution over states $\rho$ is used in both training and testing (even if the MDP changes). I then conduct an empirical study to validate the usefulness of the learned state abstraction in two kinds of experiments: 1) using $\phi$ to accelerate RL on the exact MDP in which $\phi$ was learned, and 2) using $\phi$ to accelerate RL on MDPs similar (but not identical to) the MDP in which $\phi$ was learned. That is, in the most general case, the agent will be allowed to collect data on some set of training MDPs, and then use this data to inform a choice of state abstraction for use in future related tasks.

One important difference from previous settings is that I will here focus on \textit{stochastic} state abstraction functions, which I denote $\tilde{\phi}: \mathcal{S}\ra \Delta(\mc{S}_\phi)$ that defines a function from ground states $s\in \mc{S}$ to a probability distribution over abstract states $\mc{S}_\phi$~\cite{singh1995reinforcement}. Any policy over abstract states, denoted $\pi_{\tilde{\phi}}$ will first sample $s_\phi \sim \tilde{\phi}(\cdot \mid s)$, then act given this $s_\phi$.

\paragraph{The New Objective.}
The objective is similar in spirit to the CVA and \textsc{Dibs} introduced earlier in the chapter (\autoref{alg:dibs}) adapted to continuous state spaces. Here, we again look to map ground states in which the optimal policy is similar into a common abstract state. The difficulty is that only finitely many pairs ($s_i, \pi^*(s_i)$) can be sampled, thus making it challenging to determine how to cluster any state $s'$ not seen during training. This is especially challenging when the state space is continuous, as conservatively we expect the agent to never inhabit the same state twice.

The objective is based around the idea of forming a state abstraction and policy pair that are most likely to have generated the set of trajectories seen during training. Concretely, we introduce the following objective that measures the probability of trajectories $\tau^{i}$ generated by an estimate of the optimal policy $\hat{\pi}^{*}$. In a single MDP, the goal is to maximize this probability if an agent were to use the function $\tilde{\phi}$ and policy, $\pi_{\tilde{\phi}}$, over abstract states. This optimization problem is formulated as follows.
\begin{equation}
    \argmax_{\tilde{\phi},\pi_{\tilde{\phi}}} \prod_{i=1}^{M} p(\tau^{i}, \tilde{\phi}, \pi_\phi) = \argmax_{\tilde{\phi},\pi_{\tilde{\phi}}} \sum_{i=1}^{M} \log p(\tau^{i}, \tilde{\phi},\pi_{\tilde{\phi}}),
\end{equation}
where:
\begin{align}
    \log p(\tau^{i}, \tilde{\phi}, \pi_{\tilde{\phi}}) &= \log \prod_{j=1}^{T(\tau^i)} \pi(a^{\tau^i}_{j} \mid s_{j}) p(s^{\tau^i}_{j+1} \mid s^{\tau^i}_{j},a^{\tau^{i}}_{j}),\\
    &=\sum_{j=1}^{T(\tau^i)} \log \sum_{s_\phi \in \mc{S}_\phi} \tilde{\phi}(s_\phi \mid s_{j}^{\tau^i})\pi_{\tilde{\phi}}(a_{j}^{{\tau}^i}\mid s_\phi) + \underbrace{\sum_{j=1}^{T(\tau^i)}\log p( s^{\tau^i}_{j+1}\mid s^{\tau^i}_{j},a^{\tau^i}_{j})}_{\circled{a}}.
\end{align}
Note that term $\circled{a}$ is not a function of the optimization variables, and may be dropped, yielding the following.
\begin{equation}
\argmax_{\tilde{\phi},\pi_{\tilde{\phi}}} \sum_{i=1}^{M}\log p(\tau^{i},\tilde{\phi},\pi_{\tilde{\phi}}) = \argmax_{\tilde{\phi},\pi_{\tilde{\phi}}} \sum_{i=1}^{M}\sum_{j=1}^{T({\tau}^i)}\log \sum_{s_\phi \in \mc{S}_\phi} \tilde{\phi}(s_\phi \mid s_{j}^{{\tau}^i})\pi_{\tilde{\phi}}(a_{j}^{{\tau}^i}\mid s_\phi).
\end{equation}

In general, we imagine that the agent is tasked with solving a set of $K$ different MDPs, only some of which are seen during training. This extension changes the objective as follows.
\begin{equation}
\argmax_{\tilde{\phi},\pi_{\tilde{\phi}}} \sum_{k=1}^K \sum_{i=1}^{M}\sum_{j=1}^{T({\tau}^i)}\log \sum_{s_\phi \in \mc{S}_\phi} \tilde{\phi}(s_\phi \mid s_{j}^{{\tau}^i})\pi_{\tilde{\phi}}(a_{j}^{{\tau}^i}\mid s_\phi,k).
\end{equation}
If the solution to the optimization problem is accurate enough, then states that are clustered into a single abstract state generally have a similar policy in the MDPs used for training.

Although it is possible to jointly solve this optimization problem, for simplicity let us assume $\pi_{\tilde{\phi}}$ is fixed and provided to the learner. We then parameterize the abstraction function $\phi$ by a vector $\theta$ representing the weights of a neural network, $\tilde{\phi}(\cdot \mid s;\theta)$. Further, we use softmax activation to ensure that $\tilde{\phi}$ outputs a probability distribution.

A good setting of $\theta$ can be found by performing stochastic gradient ascent on the objective above, as is standard when optimizing neural networks \cite{lecun2015deep}:
\begin{equation}
    \theta\!\leftarrow\!\theta + \alpha \nabla_{\theta} \sum_{k=1}^K\sum_{i=1}^{M}\sum_{j=1}^{T({\tau}^i)}\log \sum_{s_\phi \in \mc{S}_\phi}\tilde{\phi}(s_\phi \mid s_{j}^{{\tau}^i};\theta)\pi_{\tilde{\phi}}(a_{j}^{{\tau}^i}\mid s_\phi ,k).
\end{equation}
As in the experiments earlier in the chapter, we here use the Adam optimizer \cite{Kingma2014AdamAM}.

\paragraph{Analysis.} Ideally, as per the abstraction desiderata, we would like the abstractions learned by our procedure to support efficient discovery of good policies. When the state space is continuous, this is particularly challenging, as the agent is constantly encountering ground states never seen during training. In light of these difficulties, we next prove that a $\tilde{\phi}$ learned according to our procedure on some finite data set of $n$ experiences can still ensure a form of bounded expected value loss, assuming the state distribution is unchanged.

Concretely, let us suppose that the probability distribution used to generate states during both training and evaluation is some fixed $\rho$. Then, it is natural to consider how well the learned state abstraction will support learning under the friendly assumption that future states will \textit{also} be sampled according to $\rho$. In particular, I next present a generalization error bound on the dissimilarity between $\pi^*$ and $\pi_{\tilde{\phi}}$ for the learned $\tilde{\phi}$, in expectation under the sampling distribution $\rho$. This error bound can be decomposed into three components: 1) the training error (in training $\tilde{\phi}$), 2) the Rademacher Complexity \cite{bartlett2002rademacher} of $\Phi$, and 3) the size of the data set used to train $\tilde{\phi}$, $n$. More formally, the result is as follows.

\begin{theorem}
For any error probability $\delta \in (0,1)$, $n$ the size of the training data set, $\Delta$ the training error, and for some fixed distribution on states $\rho$ using during training, the following holds with probability $1-\delta$,
\begin{equation}
\begin{aligned}
    & \bE_{s \sim \rho}\left[\elone{\big(\pi^{*}(\cdot\mid s) - \pi_{\tilde{\phi}}(\cdot\mid s)\big)}\right] \leq \frac{\Delta}{2}+
    2\sqrt{2}\rademacher{\Phi}+\sqrt{\frac{2\ln{\frac{1}{\delta}}}{n}}.
\end{aligned}
\end{equation}
\end{theorem}

The proof was first introduced by \citet{asadi2020sa_control}---see Theorem 1. The main power of this bound comes from the application of Rademacher Complexity \cite{bartlett2002rademacher}, which measures the richness of the function family $\Phi$. This result indicates the effect of the capacity of the model family (smooth neural networks, in our case), and of the sample size ($n$) on the overall gap in quality between the abstract policy and the optimal policy. This result is powerful as it tells us that states sampled from the distribution $\rho$, even those not seen during training, are likely to be well accommodated by the learned state abstraction.

\paragraph{Experiments.}

I now summarize empirical results from two sets of experiments that investigate the utility of the proposed approach.
\begin{enumerate}

    \item Single Task: We collect an initial data set $\mc{D}_{\text{train}}$ of size $n$ to be used to train the state abstraction $\tilde{\phi}$ based on MDP $M$. Then, we evaluate the performance of tabular $Q$-learning on $M$ given this state abstraction. These experiments provide an initial sanity check as to whether the state abstraction can facilitate learning at all.
    
    \item Multi-Task: We next consider a collection of MDPs $\{M_1, \ldots, M_k\}$. We collected an initial data set $\mc{D}_{\text{train}}$ of $(s,a)$ tuples from a (strict) subset of MDPs in the collection. We used $\mc{D}_{\text{train}}$ to construct a state abstraction $\phi$, which we then gave to $Q$-learning to learn on one or many of the remaining MDPs. Critically, we evaluate $Q$-learning on MDPs \textit{not} seen during the training of $\phi$.
\end{enumerate}

For each of the two experiment types, we evaluate in three different domains, Puddle World, Lunar Lander, and Cart Pole. Open source implementations of Lunar Lander and Cart Pole are available by \citet{openai_gym2016}. We contrast the learning efficiency of tabular $Q$-learning paired with the learned state abstraction (green) with $Q$-learning using a linear function approximator (blue) \cite{melo2008analysis,konidaris2011value}. The features used by the linear approximator are those provided by the standard implementation of each domain. In Puddle World, the state is a pair of real numbers denoting spatial coordinates; in Cart Pole, the state is constituted by four real numbers indicating the location, velocity, angle, and angular velocity of the pole; and in Lunar Lander, the state is eight real numbers indicating things like the position and angle of the lander. Full parameter settings and other experimental details are available in our code.\footnote{\url{https://github.com/david-abel/continuous_state_sa}}

\begin{figure}
    \centering
  \hspace{7mm} \subfloat[Puddle World~\label{fig:c5_puddle}]{\includegraphics[width=0.24\textwidth]{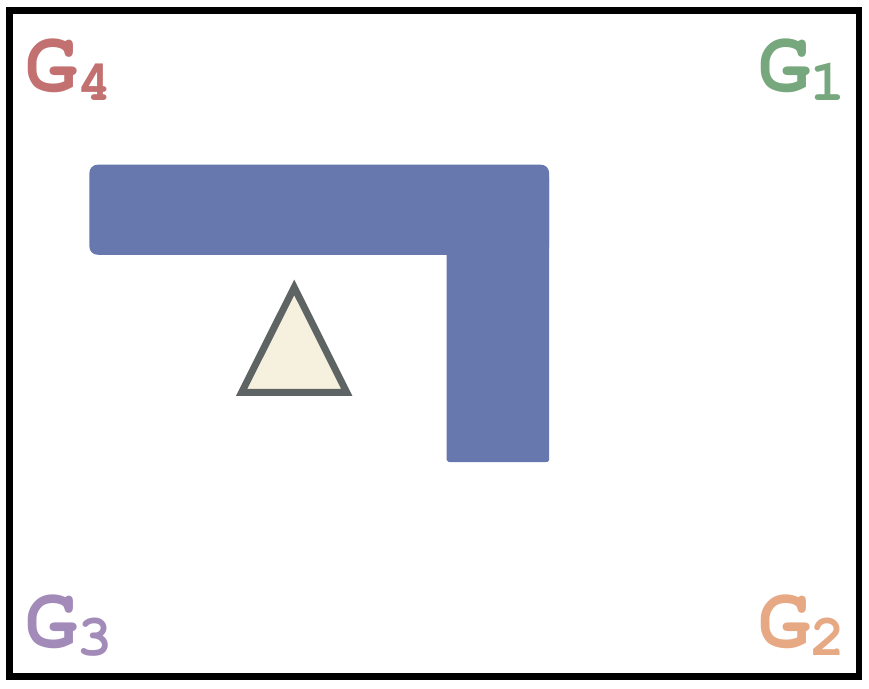}} \hspace{12mm}
    \subfloat[Lunar Lander~\label{fig:c5_lunar}]{\includegraphics[width=0.24\textwidth]{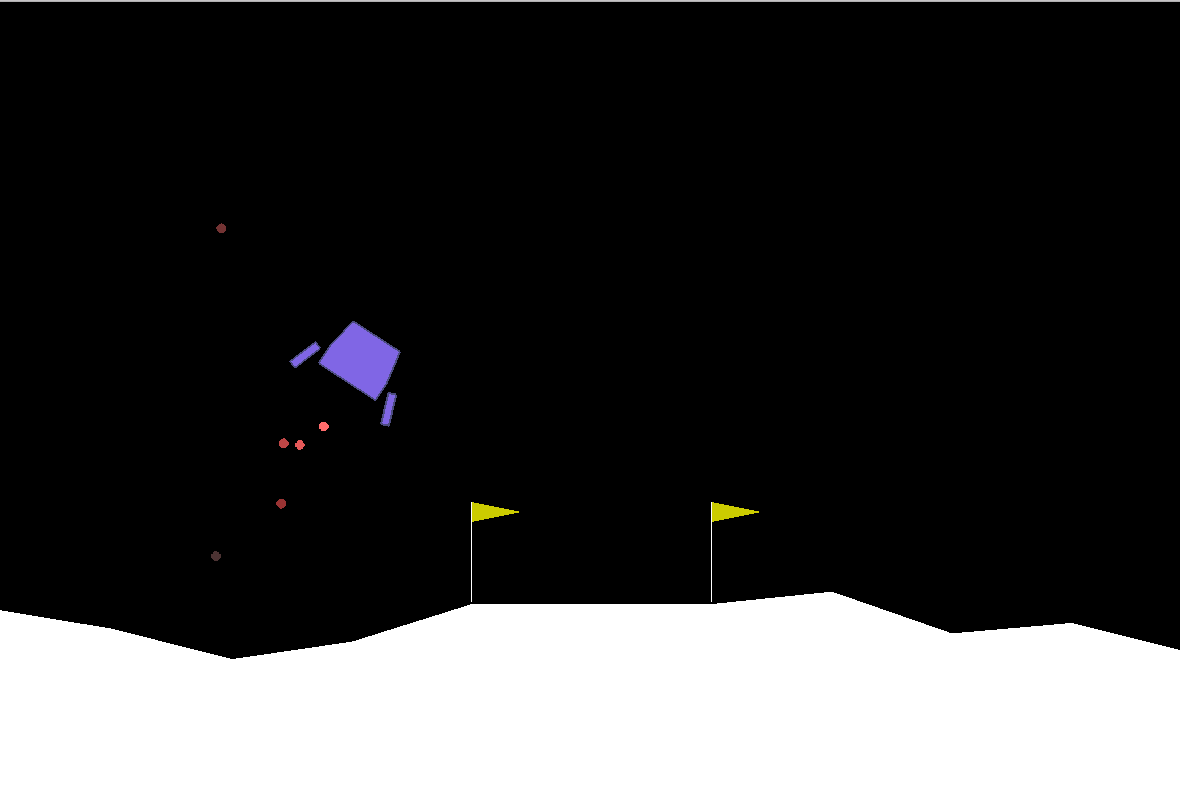}}\hspace{8mm}
    \subfloat[Cart Pole]{\includegraphics[width=0.24\textwidth]{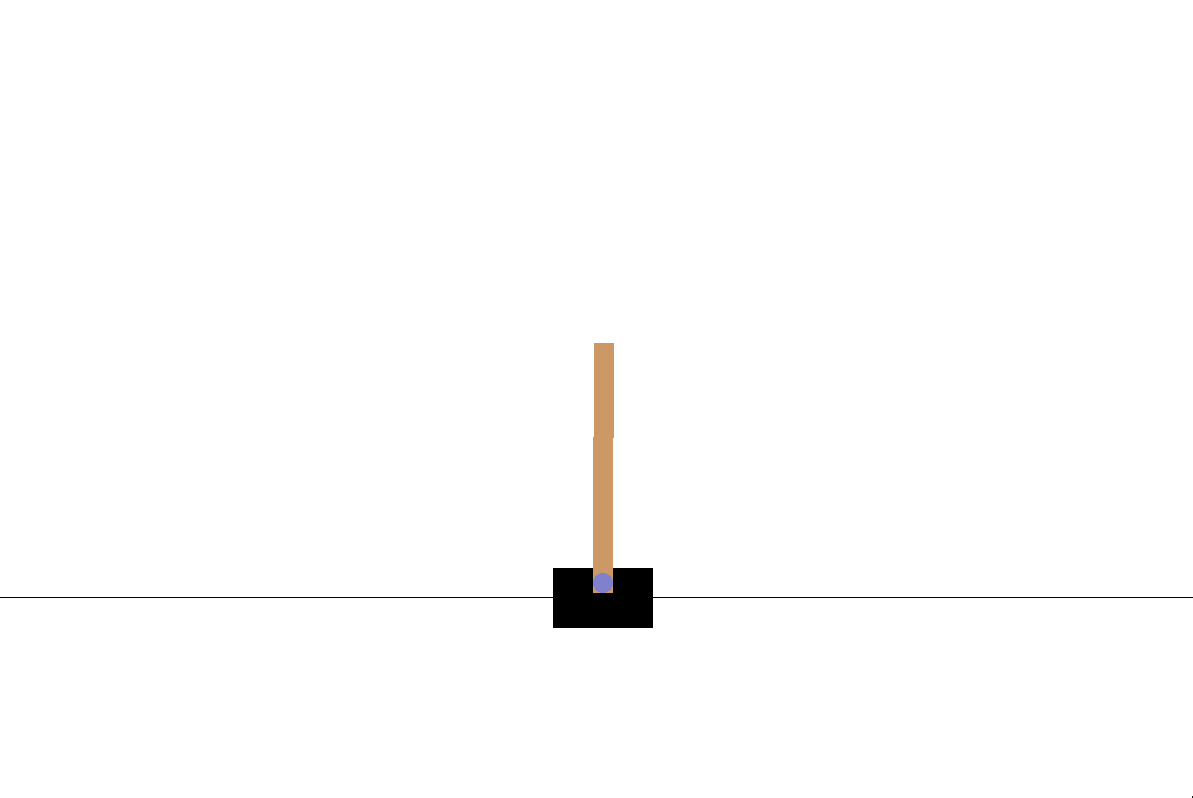}} \\
    
    \subfloat[Single Task Puddle World \label{fig:c5_puddle_results}]{\includegraphics[width=0.3\textwidth]{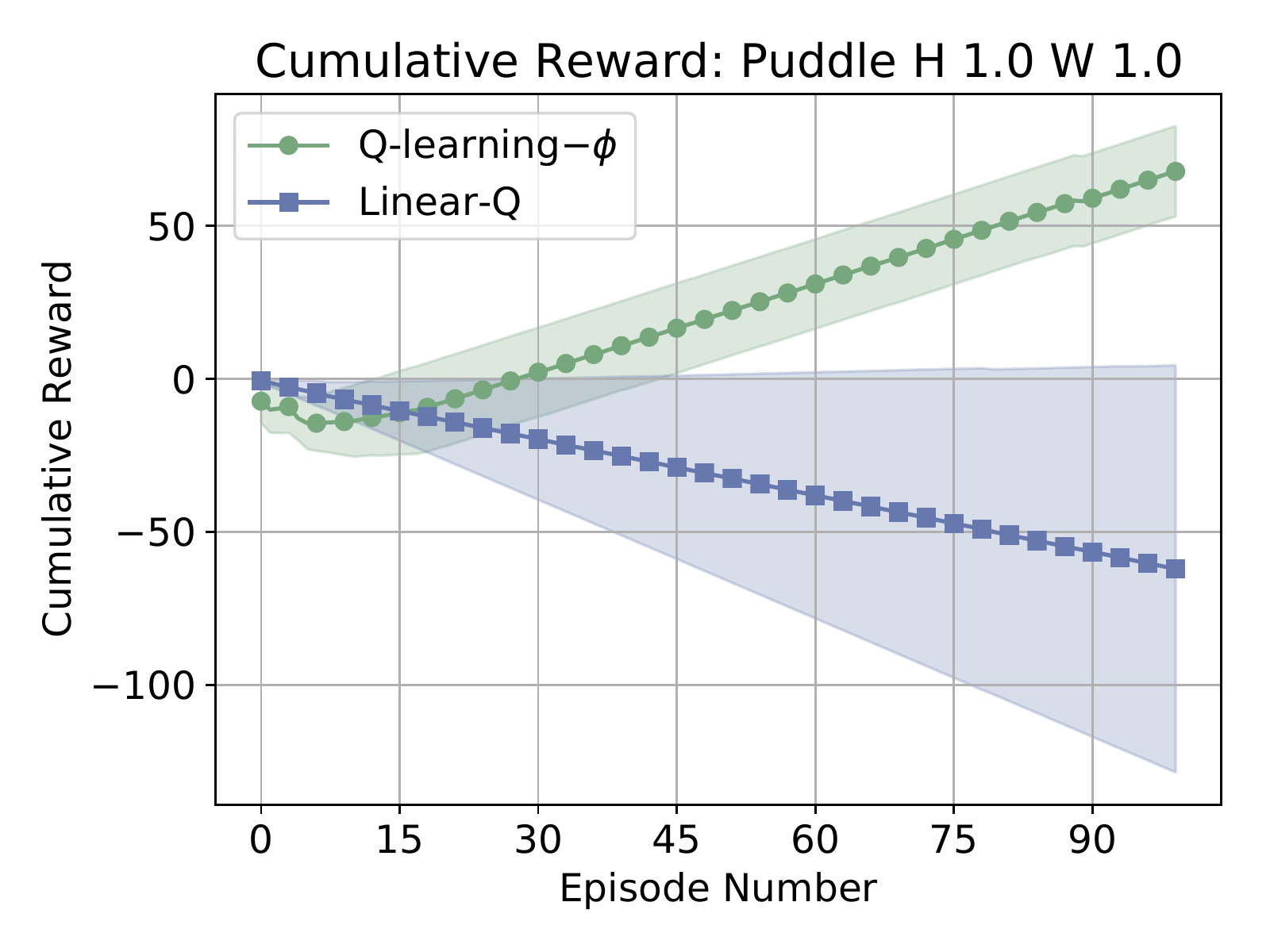}}
    \subfloat[Single Task Lunar Lander \label{fig:c5_lunar_results}]{\includegraphics[width=0.3\textwidth]{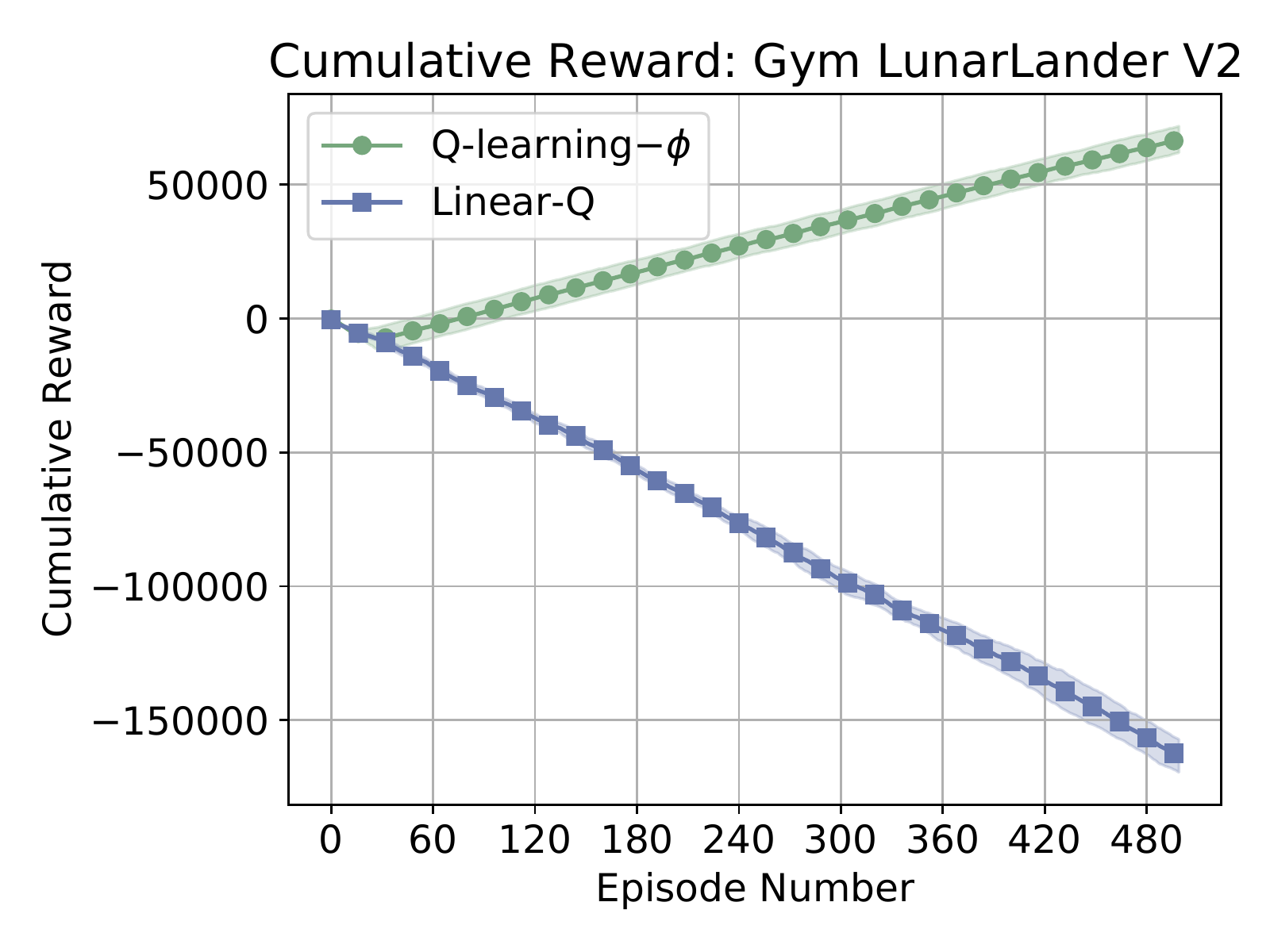}}
    \subfloat[Single Task Cart Pole \label{fig:cartpole_results}]{\includegraphics[width=0.3\textwidth]{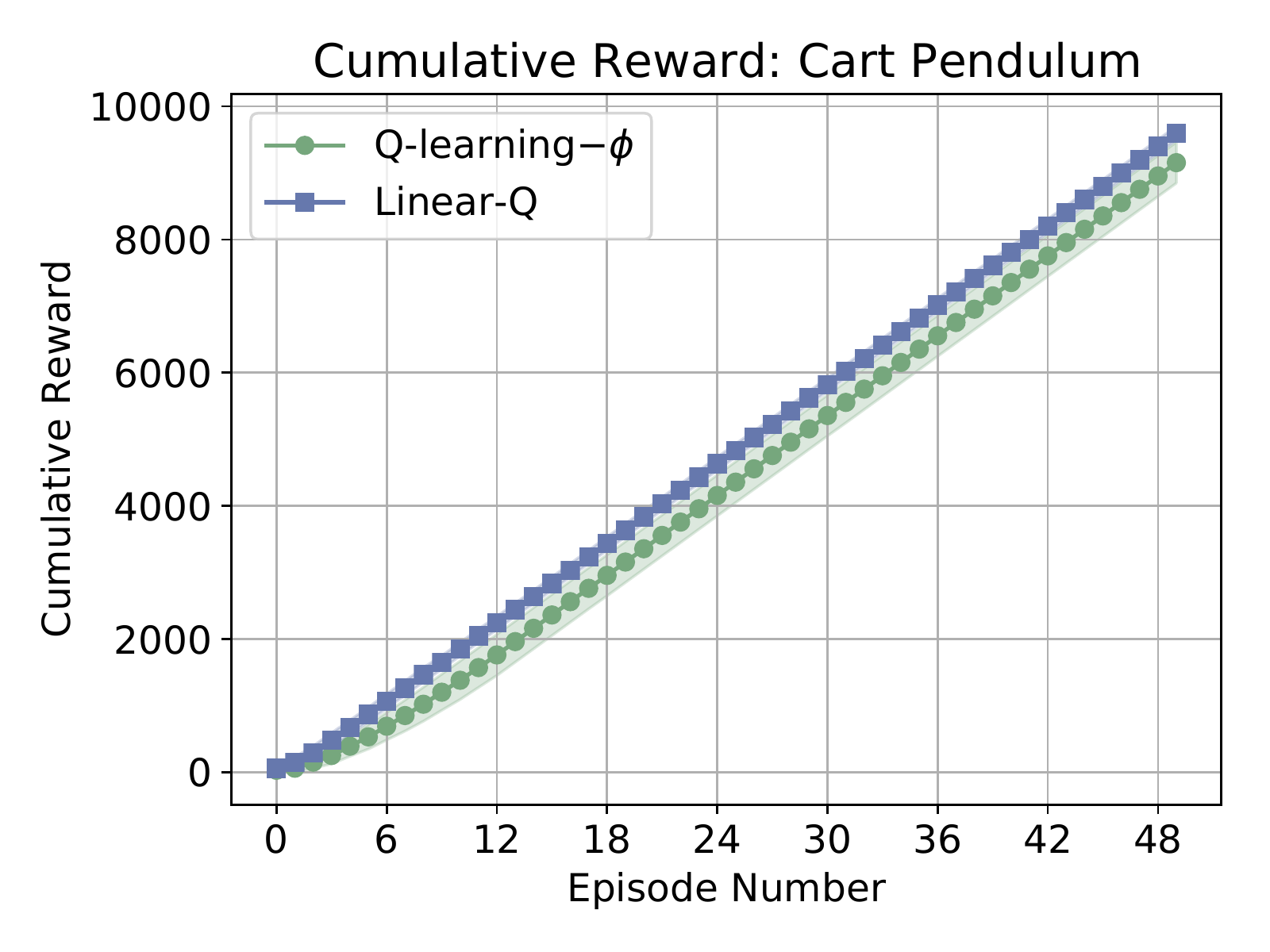}} \\
    
    \subfloat[Puddle World Transfer \label{fig:c5_puddle_transfer}]{\includegraphics[width=0.3\textwidth]{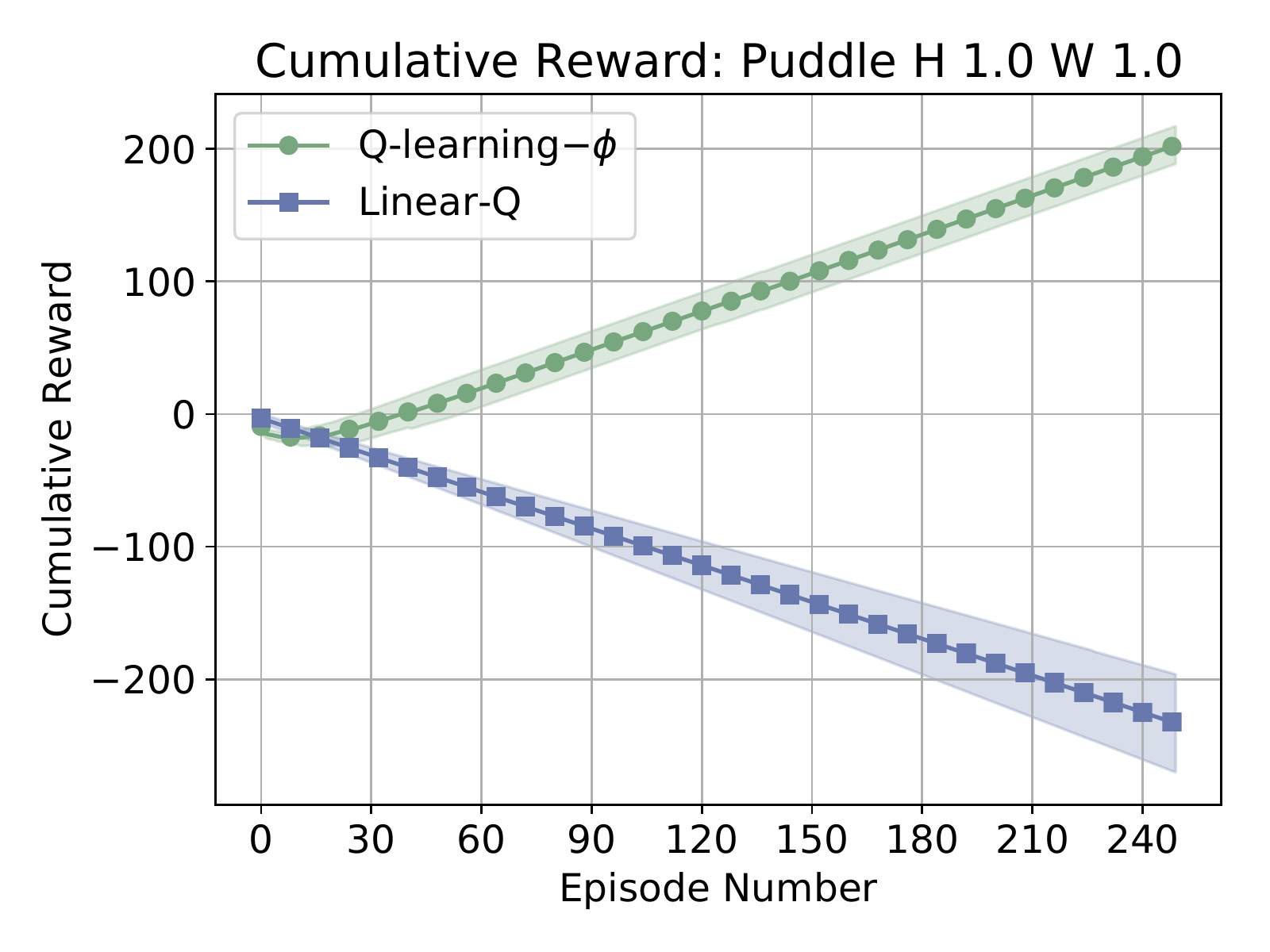}}
    \subfloat[Lunar Lander Transfer \label{fig:c5_lunar_transfer}]{\includegraphics[width=0.3\textwidth]{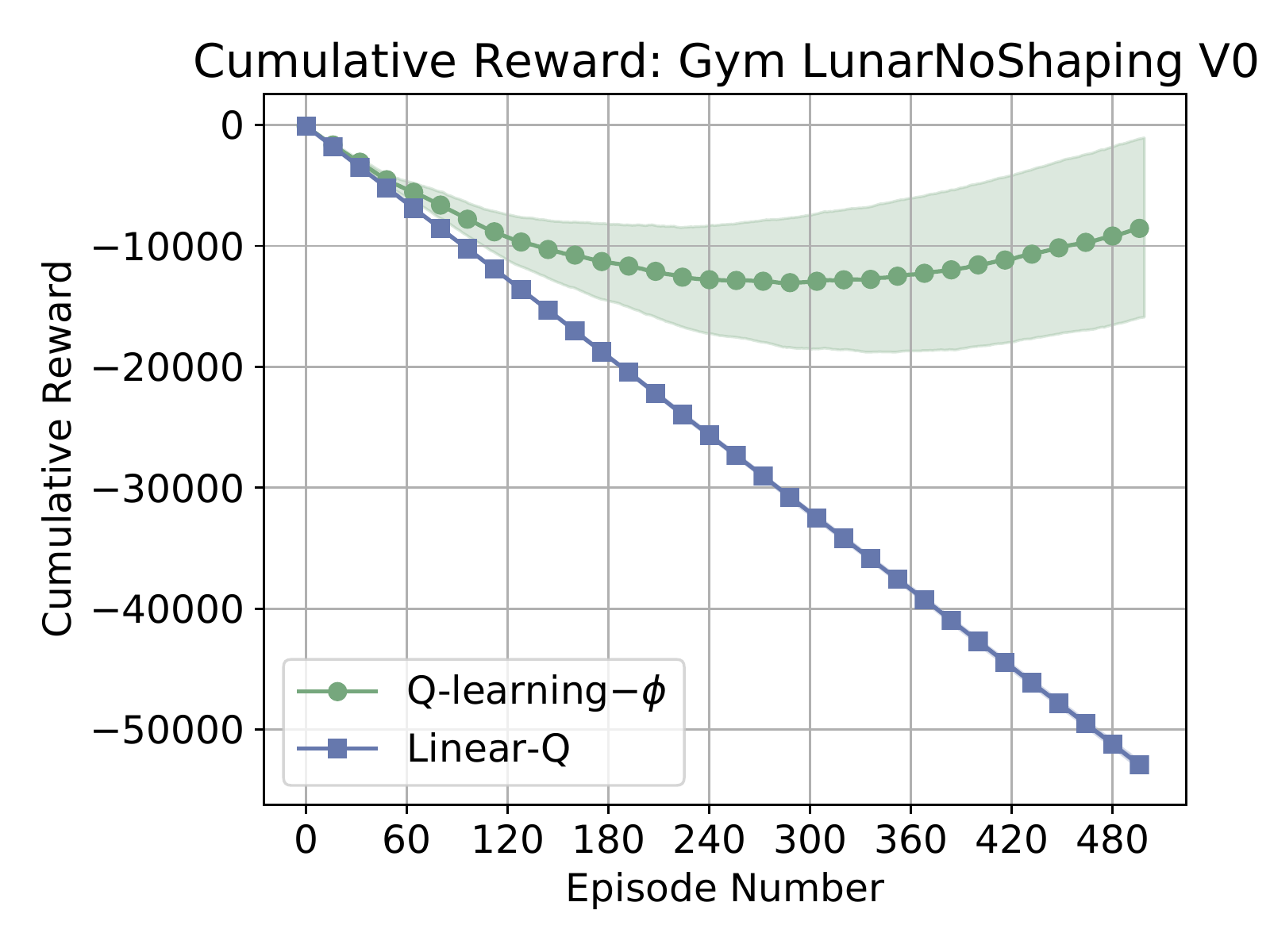}}
    \subfloat[Cart Pole Transfer \label{fig:c5_cartpole_transfer}]{\includegraphics[width=0.3\textwidth]{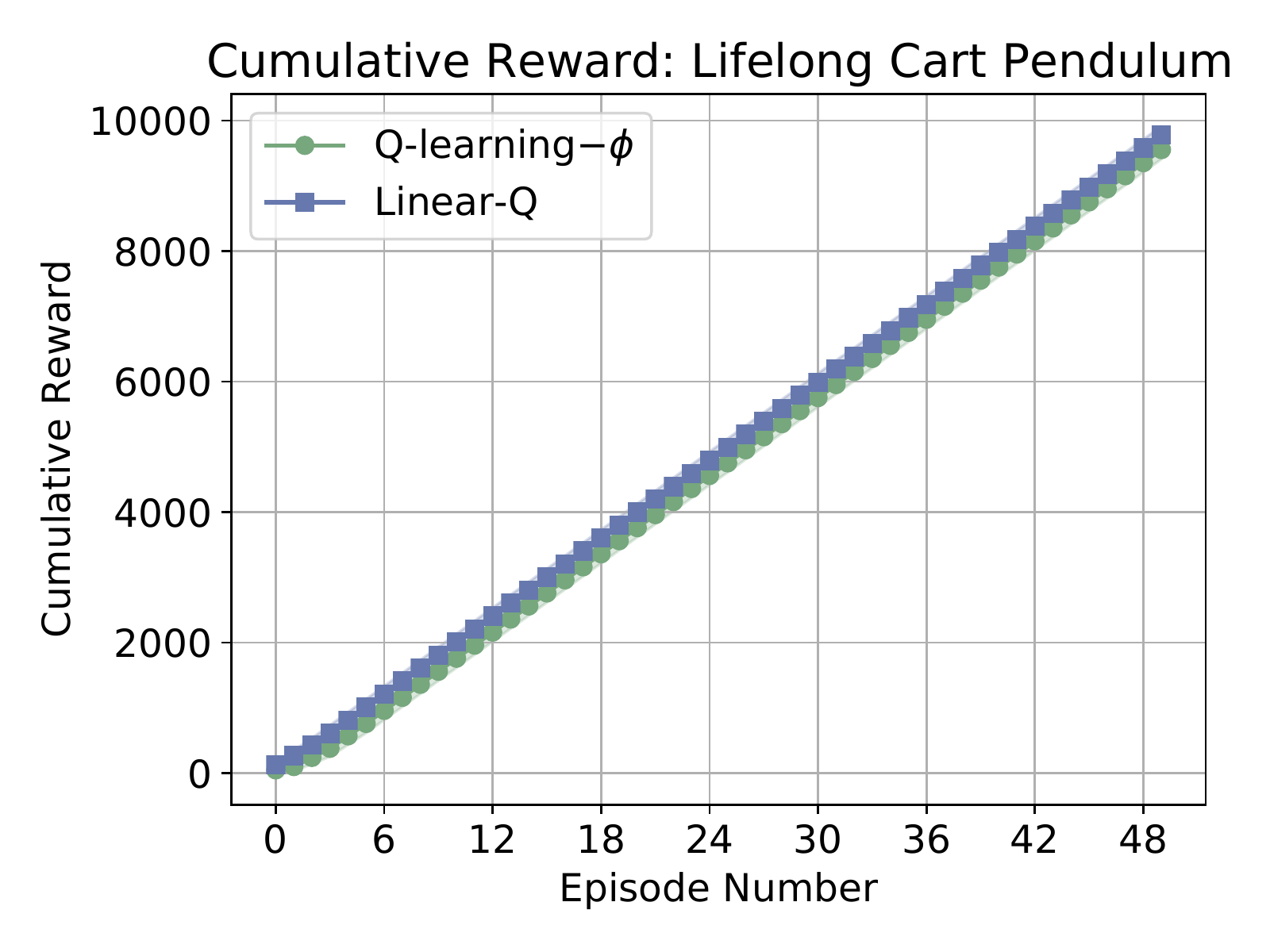}}
    \caption{Learning curves for the single task experiments (top) and the transfer experiments (bottom).}
    \label{fig:c5_cont_sa_results}
\end{figure}

\autoref{fig:c5_cont_sa_results} presents results for both experiments, with the single task results in the middle row and the transfer results in the bottom row. Each figure presents the mean cumulative reward per episode with 95\% confidence intervals, averaged over 25 instances. In Puddle World, we see that the learned state abstraction is capable of consistently supporting extremely sample efficient learning in both the single task and transfer case---by around episode 30 in both cases, tabular $Q$-learning reliably converges to a policy that effectively takes the agent directly to the goal while avoiding the puddle. The same is true of Lunar Lander, only more samples are required in the transfer case. In contrast, for the given sample budgets, the approach using the linear function approximator is unable to improve its policy at all. Lastly, in Cart Pole, we see both approaches are able to find near-optimal policies in relatively few samples.

\begin{figure}[b!]
    \centering
    \subfloat[Tabular $Q$]{\includegraphics[width=0.4\textwidth]{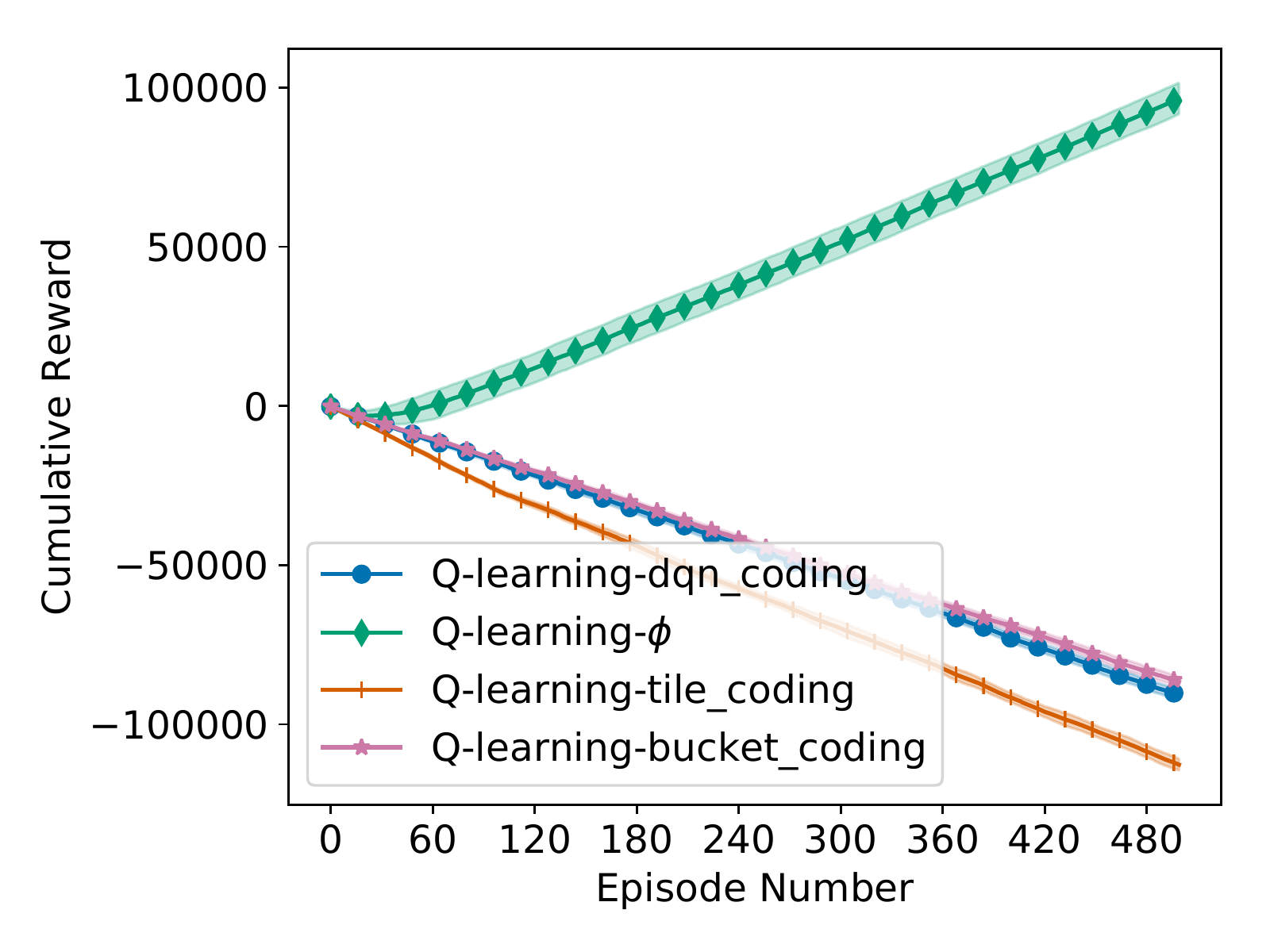}} \subfhspace
    \subfloat[Linear $Q$]{\includegraphics[width=0.4\textwidth]{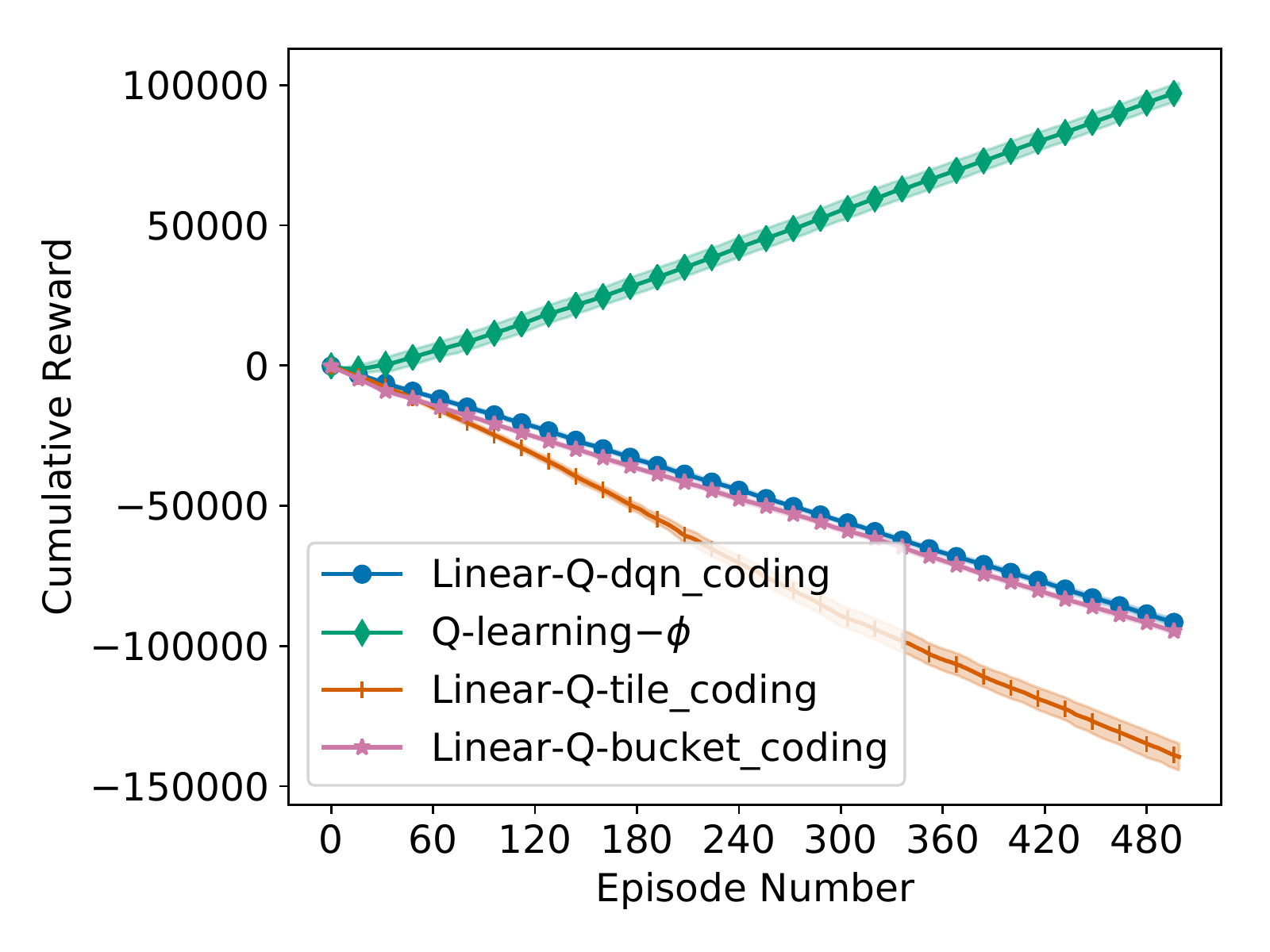}} 
    \caption{A comparison of different state discretization methods in Lunar Lander.}
    \label{fig:c5_lunar_reps}
\end{figure}

In a final experiment, we contrast the learning performance of tabular $Q$-learning using several different state-discretization methods on the single task variant of the Lunar Lander domain. First, we use our same approach, shown in green. Second, we contrast tile coding (orange) \cite{sutton1996generalization}, a naive form of discretization we call bucket coding (pink), and an approach that uses state features learned by a Deep $Q$-Network \cite{mnih2015human} trained on the same task (blue). Results are presented in \autoref{fig:c5_lunar_reps}. The data are quite clear: on average, the state abstraction learned by our approach is sufficient for enabling extremely sample efficient learning on Lunar Lander. In contrast, none of the other state representations support the learning of good behavioral policies. For more on these ideas, experiments, and the algorithm, see original work by \citet{asadi2020sa_control}.


To summarize, this chapter draws on the tools of information theory to cast the process of state abstraction as a form of \textit{compression}. I introduced an algorithmic framework that can efficiently produce state abstractions that trade off between compression and value preservation. Through a variety of visuals, empirical study, and analysis, I have demonstrated the power of this approach to discover good state abstractions for RL.

This brings Part 2 to a close. I next shift focus to action abstraction.

\vspace{4mm}
\begin{center}
\begin{minipage}{0.65\textwidth}
\resizebox{0.9\textwidth}{6pt}{%
  \begin{tikzpicture}
        \pgfornament[symmetry=h]{85} 
    \end{tikzpicture}
}
\end{minipage}
\end{center}

%% file: proofs/c5/c5_entropy_bounds_alphabet_size.tex
\begin{dproof}[Lemma \ref{lem:c5_entropy_sa_size}]
First, recall the definition of the entropy $H(X)$ of a discrete random variable $X$, taking on values $x \in \mc{X}$,
\begin{equation}
    H(X) := - \sum_{x \in \mc{X}} p(x) \log_2 p(x).
\end{equation}
Then, for a given maximum entropy $H(X) \leq N$ and $\delta_{min} \in \mathbb{R}_{> 0}$, we seek an upper bound on $|\mc{X}|_{p(x)}^{\delta_{min}}$.  \\

\noindent That is, we would like to upper bound the following quantity:
\begin{equation}
    \max_{p(x) : H(X) \leq N} |\mc{X}|_{p(x)}^{\delta_{min}}.
\end{equation}
Note that this quantity is maximized by a uniform distribution that applies $\delta_{min}$ mass to each element in the support, across the largest alphabet where $H(X) = N$:
\begin{align}
    H(X) &= -\sum_{x \in \mc{X}'} p(x) \log_2 p(x) \\
    &= -\sum_{x \in \mc{X}'} \delta_{min} \log_2 \delta_{min} \\
    &= -|\mc{X}'| \delta_{min} \log_2 \delta_{min} \\
    &= |\mc{X}'| \delta_{min} \log_2 \frac{1}{\delta_{min}}.
\end{align}
Therefore, for a given pmf $p(x)$ with entropy $H(X)$, and a minimum threshold of probability, the minimum size of the alphabet $\mc{X}$ is upper bounded:
\begin{equation}
    |\mc{X}| \leq \frac{H(X)}{\delta_{min} \log_2 \frac{1}{\delta_{min}}}. \qedhere
\end{equation}
\end{dproof}

%% file: proofs/c5/c5_exp_kl_val_bound.tex
\begin{dproof}[Lemma \ref{lem:c5_kl_val_bound}]
Recall the total variation distance (TVD) between our two policies for a given state $s$ is defined as:
\begin{equation}
    TV(\pi_E(\cdot \mid s), \pi_\phi(\cdot \mid s)) = \sup\limits_{a \in \mc{A}} |\pi_E(a \mid s) - \pi_\phi(a \mid s)|.
\end{equation}
Furthermore, recall that TVD relates to the $L_1$ norm and the KL divergence:
    \begin{align}
        TV(\pi_E(\cdot \mid s), \pi_\phi(\cdot \mid s)) &= \frac{1}{2} \sum\limits_{a \in \mc{A}} |\pi_E(\cdot \mid s) - \pi_\phi(\cdot \mid s)| \\
        &\leq \sqrt{\frac{1}{2} \KL(\pi_E(\cdot \mid s) \mid \mid \pi_\phi(\cdot \mid s))},
        \label{eq:pinsker}
    \end{align}
where the inequality in Equation~\ref{eq:pinsker} is formally known as Pinsker's inequality. With this inequality in place, we expand the expectation in the value bound:
\begin{align*}
&\bE_{p(s)}[V^{\pi_E}(s) - V^{\pi_\phi}(s)] \numberthis \label{eq:exp_kl_val_last_step} \\
&\leq \sum_{s \in \mc{S}} p(s) \left(\sum\limits_{a \in \mc{A}} |\pi_E(a \mid s) - \pi_\phi(a \mid s)| (R(s,a) + \gamma\sum\limits_{s'} T(s' \mid s,a)|V^{\pi_E}(s') - V^{\pi_\phi}(s')|\right) \\
&= \sum_{s \in \mc{S}} \sum\limits_{a \in \mc{A}} p(s)  |\pi_E(a \mid s) - \pi_\phi(a \mid s)| \left(R(s,a) + \gamma\sum\limits_{s'} T(s' \mid s,a)|V^{\pi_E}(s') - V^{\pi_\phi}(s')|\right)
\end{align*}
Then, applying the upper bound on the possible value $\textsc{VMax} = \textsc{RMax} / (1-\gamma)$ to Equation~\ref{eq:exp_kl_val_last_step}:
\begin{equation}
\bE_{p(s)}[V^{\pi_E}(s) - V^{\pi_\phi}(s)]\leq \textsc{VMax}\ \bE_{p(s)} \left[|\pi_E(a \mid s) - \pi_\phi(a \mid s)|\right].
\end{equation}
Then, by Pinsker's inequality, we conclude:
\begin{align}
\bE_{p(s)}[V^{\pi_E}(s) - V^{\pi_\phi}(s)] &\leq 2\textsc{VMax}\ \bE_{p(s)} \left[\sqrt{\frac{1}{2} \KL(\pi_E(a \mid s) \mid \mid \pi_\phi(a \mid s))}\right] \\
&\leq \sqrt{2k}\textsc{VMax}. \qedhere
\end{align}
\end{dproof}

%% file: proofs/c5/c5_proxy_rlit_objective.tex
\begin{dproof}[Theorem \ref{thm:c5_main_objective_bound}]
The proof follows from \autoref{lem:c5_entropy_sa_size} and \autoref{lem:c5_kl_val_bound}. Consider the $\phi$ that minimizes $\hat{\mc{J}}_\text{DIB}$, yielding the value of at most $N + \beta k$, where:
\begin{align}
    N &:= H(S_\phi) \\
    k &:= \underset{{s \sim \rho_E}}{\bE}\left[D_{KL}(\pi_E(a \mid s) \mid \mid \pi_\phi(a \mid \phi(s))\right].
\end{align}
Then, by \autoref{lem:c5_entropy_sa_size}, we know:
\begin{equation}
    |\mc{S}_\phi|_{\rho_\phi(s)}^{\delta_{min}} \leq \frac{N}{\delta_{min} \log_2 \frac{1}{\delta_{min}}}.
\end{equation}
By \autoref{lem:c5_kl_val_bound}, we know:
\begin{equation}
    \bE_{\rho(s)}\left[V^{\pi_E}(s) - V^{\pi_\phi}(s)\right] \leq \sqrt{2k} \textsc{VMax}.
\end{equation}
Therefore, since both quantities are non-negative, we conclude:
\begin{align}
    \mc{J}[\phi] &= |\mc{S}_\phi|_{\rho_\phi(s)}^{\delta_{min}} + \beta \bE_{\rho(s)}\left[V^{\pi_E}(s) - V^{\pi_\phi}(s)\right] \\
    &\leq \frac{N}{\delta_{min} \log_2 \frac{1}{\delta_{min}}} + \beta \sqrt{2k}\textsc{VMax}.
\end{align}
Thus, we can upper bound the quantities in $\mc{J}$ as a function of the quantities in $\hat{\mc{J}}_\text{DIB}$. \qedhere
\end{dproof}

%% file: algorithms/dibs_alg.tex
\begin{algorithm}[!h]
\caption{\textsc{DIBS}}
\label{alg:dibs}
\textsc{Input:} $\pi_E, \rho_E,  M, \beta, \Delta, \text{\textit{iters}}$ \\
\textsc{Output}: $\phi, \pi_\phi$ \\

\begin{algorithmic}[1]


\State{$\forall_{s} : \phi_{0}(s) = \texttt{random.choice}([1,|\mc{S}|])$} \Comment{Initialize}
\State{$\forall_{s} : \pi_{\phi_{0}}(a \mid s_\phi) \sim \mathrm{Unif}(\mc{A})$}
\State{$\forall_s : \rho_{\phi,0}(s_\phi) \sim \mathrm{Unif}([1,|\mc{S}|])$}


\For{$t = 0$ to \textit{iters}} \Comment{Iterative updates}
    \State{$j_{t+1}(\phi_t(s)) = \log \rho_{\phi,t}(\phi_t(s)) - \beta \KL(\pi_E(\cdot \mid s) \mid \mid \pi_{\phi,t}(\cdot \mid \phi_t(s)))$}
    
    \State{$\phi_{t+1}(s) = \argmax_{s_\phi} j_{t+1}(s_\phi)$}
    \State{$\rho_{\phi,t+1}(s_\phi) = \sum_{s :\phi_t(s) = s_\phi} \rho_E(s)$}
    \State{$\pi_{\phi,t+1}(a \mid s_\phi) = \frac{ \sum_{s :\phi_t(s) = s_\phi} \pi_E(a \mid s) \rho_E(s)}{\sum_{s :\phi_t(s) = s_\phi} \rho_E(s)}$}

    \If{$\max_{f \in \{\pi_\phi, \phi, \rho_\phi\}} \mathrm{L_1}(f_t, f_{t+1}) \leq \Delta$} \Comment{Check convergence}
        \State{\texttt{break}} 
    \EndIf
\EndFor \\
\Return{$\phi_{t+1}, \pi_{\phi,t+1}$}
\end{algorithmic}
\end{algorithm}

%% file: proofs/c5/c5_steady_state_distr_val_loss.tex
\begin{dproof}[Lemma \ref{lem:c5_l1_pol_l1_rho}]
We bound the difference between the two state distributions after $t$ steps as follows. First, expanding:
\begin{align}
	\sum_{s' \in \mc{S}}|\rho^{t}_{\pi_1,s_0}(s')-\rho^{t}_{\pi_2,s_0}(s')|=&\sum_{s'\in \mc{S}}|\PR(S_{t}=s'\mid s_0,\pi_1)-\PR(S_{t}=s' \mid s_0,\pi_2)| \\
	&-\sum_{s\in \mc{S}}\PR(S_{t-1}=s \mid s_0,\pi_2)\sum_{a \in \mc{A}}\pi_2(a \mid s)T(s' \mid s,a)|. \nonumber
\end{align}
Then, by algebra:
\begin{eqnarray}
    &&\sum_{s'\in \mc{S}}|\rho^{t}_{\pi_1,s_0}(s')-\rho^{t}_{\pi_2,s_0}(s')|\\
	&\leq &\sum_{s'\in \mc{S}}|\sum_{s\in \mc{S}}\PR(S_{t-1}=s \mid
	s_0,\pi_1)\sum_{a \in \mc{A}}\Big(\pi_1(a \mid s)-\pi_2(a \mid s)\Big)T(s' \mid s,a)| \\
	&+&\sum_{s'\in \mc{S}}|\sum_{s \in \mc{S}}\Big(\PR(S_{t-1}=s \mid s_0,\pi_1)-\PR(S_{t-1}=s \mid s_0,\pi_2)\Big)\sum_{a\in \mc{A}}\pi_2(a \mid s)T(s' \mid s,a)|. \nonumber
\end{eqnarray}

Continuing,
\begin{eqnarray}
    &&\sum_{s' \in \mc{S}}|\rho^{t}_{\pi_1,s_0}(s')-\rho^{t}_{\pi_2,s_0}(s')|\\
	&\leq &\sum_{s\in \mc{S}}\PR(S_{t-1}=s \mid s_0,\pi_1)\sum_{a\in \mc{A}}\Big|\pi_1(a \mid s)-\pi_2(a \mid s)\Big|\sum_{s'\in \mc{S}} T(s' \mid s,a)\\
	&+&\sum_{s \in \mc{S}}\Big|\PR(S_{t-1}=s \mid s_0,\pi_1)-\PR(S_{t-1}=s \mid s_0,\pi_2)\Big|\sum_{a \in \mc{A}}\pi_2(a \mid s)\sum_{s' \in \mc{S}} T(s' \mid s,a)\nonumber\\
	&\leq& \Delta +\sum_{s\in \mc{S}}\Big|\PR(S_{t-1}=s \mid s_0,\pi_1)-\PR(S_{t-1}=s \mid s_0,\pi_2)\Big|\\
	&=&\Delta + \sum_{s'\in \mc{S}}|\rho^{t-1}_{\pi_1,s_0}(s')-\rho^{t-1}_{\pi_2,s_0}(s')|
\end{eqnarray}
Applying the above bound, and using induction, we have:
\begin{equation}
    \sum_{s'\in \mc{S}}|\rho^{t}_{\pi_1,s_0}(s')-\rho^{t}_{\pi_2,s_0}(s')|\leq t\Delta
\end{equation}
Therefore,
\begin{eqnarray}
	&&\sum_{s\in \mc{S}}|\rho_{\pi_1,s_0}(s)-\rho_{\pi_2,s_0}(s)|\\
	&=&\sum_{s\in \mc{S}}|(1-\gamma)\sum_{t \in \mathbb{N}}\gamma^t \rho^{t}_{\pi_1,s_0}(s)-(1-\gamma)\sum_{t\in \mathbb{N}}\gamma^t \rho^{t}_{\pi_2,s_0}(s)|\\
	&\leq &(1-\gamma) \sum_{t\in \mathbb{N}}\gamma^t \sum_{s\in \mc{S}}|\rho^{t}_{\pi_1,s_0}(s)-\rho^{t}_{\pi_2,s_0}(s)|\\
	&\leq &(1-\gamma)\sum_{t\in \mathbb{N}}\gamma^t t\Delta = (1-\gamma)\frac{\gamma\Delta}{(1-\gamma)^2}=\frac{\gamma \Delta}{1-\gamma}. \qedhere
\end{eqnarray}
\end{dproof}

%% file: chapters/c6_aa_options_for_planning.tex
\begin{center}
\begin{minipage}{0.8\textwidth}
\textit{This chapter is based on ``Finding Options that Minimize Planning Time" \cite{jinnai2019opt} led by Yuu Jinnai, joint with D. Ellis Hershkowitz, Michael L. Littman, and George Konidaris.}
\end{minipage}
\end{center}
\vspace{2mm}

Action abstraction defines the process of forming high level behaviors such as ``go to the bridge", in place of ``rotate right leg so many degrees". Such a mechanism is deeply connected to many other important practices of agency, including the discovery and manipulation of useful subgoals, efficient long-horizon planning, and credit assignment. The primary formalism I adopt for capturing action abstraction is the \textit{options} framework \cite{sutton1999between}. Several other names describe roughly the same process, including skills, temporal abstraction, and macro-actions. I treat options as sufficiently general to capture all of these, but of course there are subtleties to each particular type. For further background on action abstraction and options, see \autoref{sec:c2_action_abstr}.

\begin{figure}
    \centering
    \includegraphics[width=0.7\textwidth]{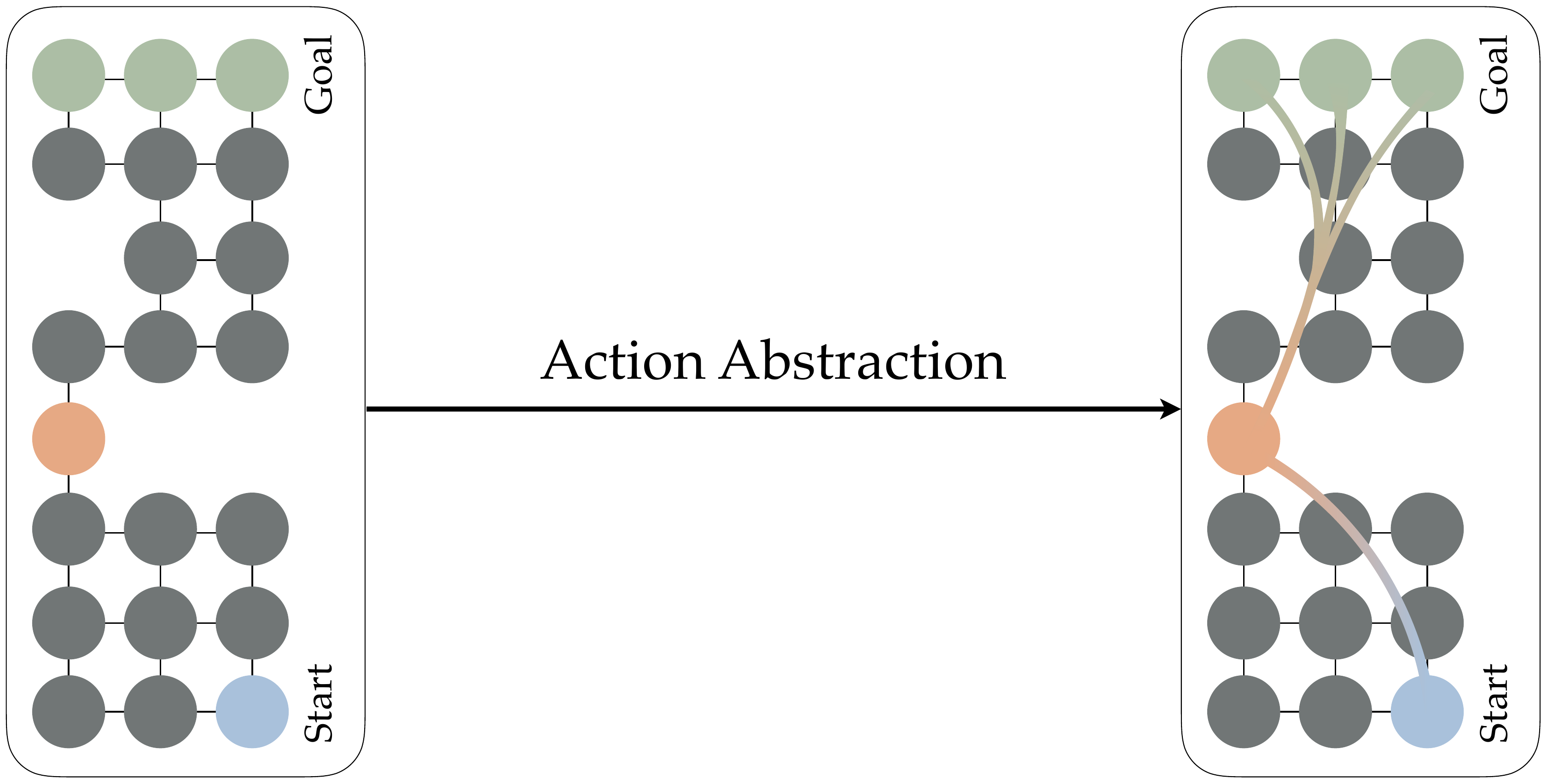}
    \caption{Action abstraction.}
    \label{fig:c6_action_abstr}
\end{figure}

As with state abstraction, my objective in this part of the dissertation is to bring formal clarity to the discovery of \textit{good} action abstractions (those that satisfy the abstraction desiderata---see \autoref{sec:c2_desiderata}).

In this chapter, I first study one notion of ``good action abstraction". I present analysis of the problem of finding options that make the process of \textit{planning} as efficient as possible. To make this problem concrete I will make several simplifying assumptions that allow for appropriate analysis. In particular, I will ground the speed of planning in terms of how many iterations of VI (\autoref{alg:value_iteration}) are required to return an accurate value function for the whole state space. I prove that this problem is NP-hard and hard to approximate well. Fortunately, these hardness results also come along with two approximation algorithms that each nearly match the lower bound of approximation hardness under friendly assumptions. One of the algorithms is based on an approximation of set cover by \citet{chvatal1979greedy}, and the other is based on the procedure by \citet{archer2001two}. In simple experiments, the options found by these approximation algorithms are nearly competitive (in terms of acceleration of VI) with those options found by solving the NP-hard problem.

Why might we care about this problem? Well, again, there are many things abstract actions can do to enhance a learning agent's capabilities. Intuitively, carrying out long-horizon simulations of the right kind can be immensely useful, so long as the simulations are sufficiently well informed. Imagine boiling a pot of water. By setting a particular degree of heat beneath the flame, it is easy to predict that in some number of minutes that the water will boil with extremely high probability. It is not necessary to know precisely how long it will take, or exactly how much water will evaporate before you turn the heat down. To be useful, it is simply sufficient to know that the heat will lead to water boiling in a reasonable time frame. It is this activity that carries a great deal of promise, and for which the options formalism is well suited to study. Hence, a general negative result illustrating that it is difficult to find the right options suggests that there is more nuance to the problem of discovering good options than simply optimizing relative to a given task---I liken this result to a version of the No Free Lunch theorem \cite{wolpert1996lack}, according to which no learning algorithm can dominate all others on all problems. In a sense, the hardness results presented here suggest that no option discovery algorithm can efficiently find the right options on all problems, but perhaps on a well chosen subset, such a task is easier. To summarize, the results of this chapter are useful for understanding the limitations of option discovery, evaluating option discovery methods, and for guiding future option discovery algorithms. I return to this discussion later in the chapter.

\section{Formalizing The Problem}

I now formalize what it means to find the set of options that is optimal for planning. I will then use this formalism to establish hardness results for computing options that help with planning, both in the worst and approximate cases. The main positive result is the existence of an approximation algorithm with a principled theoretical foundation.

To ground this study, I will restrict attention to a particular notion of planning in several ways. First, I only study the acceleration of VI, rather than the full scope of planning algorithms. Indeed, I take VI to be both sufficiently general and canonical to capture the rough structure of many planning algorithms. Second, I ignore any increase to the branching factor. This is a big component, as adding options will help reduce the number of iterations of VI, but will necessarily increase the number of actions evaluated at each state. Hence, the proposed model only captures a part, but not the whole picture, of planning acceleration. Finally, I concentrate only on finite MDPs. While these restrictions limit the scope of the result, it is important to establish this the computational difficulty of this problem in a restricted setting before considering the more general cases.

Precisely, I will show that the problem of finding options that minimize the number of iterations required by VI,
\begin{enumerate}
    \item is $2^{\log^{1 - \epsilon} n}$-hard to approximate for any $\epsilon > 0$ unless $\text{NP} \subseteq \text{DTIME}(n^{\text{poly} \log n})$,\footnote{This is a standard complexity assumption: see, for example, \citet{dinitz2012label}} where $n$ is the input size;
    \item is $\Omega(\log n)$-hard to approximate even for deterministic MDPs unless PTIME = NP;
    \item has a $O(n)$-approximation algorithm;
    \item has a $O(\log n)$-approximation algorithm for deterministic MDPs.
\end{enumerate}

In \autoref{sec:algorithms}, I present \momialg{}, a polynomial-time approximation algorithm that has $O(n)$ suboptimality in general and $O(\log n)$ suboptimality for deterministic MDPs.
Note that the expression $2^{\log^{1-\epsilon}n}$ is only slightly smaller than $n$: if $\epsilon=0$ then $\Omega(2^{\log n}) = \Omega(n)$. Thus, \momialg{} is close to the best possible approximation factor. In addition, I will consider the complementary problem of finding a set of $k$ options that minimize the number of VI iterations until convergence. I show that this problem is also NP-hard, even for a deterministic MDP. After establishing these complexity results, I highlight a brief empirical study comparing the performance of two heuristic approaches for option discovery: betweenness options \cite{csimcsek2009skill} and eigenoptions \cite{machado2017laplacian}, with those options discovered by the new approximation algorithm.

\section{Options and Value Iteration}

I here study the \textit{value-planning problem}, defined as follows.
\ddef{Value-Planning Problem}{The \textbf{value-planning problem} is defined as follows: \texttt{given} an MDP $M = (\mc{S}, \mc{A}, R, T, \gamma, \rhoz)$ and an $\epsilon \in \mathbb{R}_{\geq 0}$, \texttt{return} a value function, $V_t$ such that $|V_t(s) - V^*(s)| < \epsilon$ for all $s \in \mc{S}$.\vspace{2mm}}

As discussed in \autoref{sec:c2_action_abstr}, options have a well defined transition and reward model for each state named the multi-time model~\cite{precup1998multi}:
\begin{align}
    T_\gamma(s' \mid s,o) &= \sum_{k=0}^\infty \gamma^k \beta_o(s') p(s', k \mid s, o), \\
    R_\gamma(s,o) &= \underset{k, s_{1 \ldots k}}{\bE}\left[r_1 + \gamma r_2 \ldots + \gamma^{k-1} r_k \bigmid s, o\right].
\end{align}
To run VI with options, it is natural to substitute the MTM for the standard reward and transition function and apply the same exact operations. The algorithm then computes a sequence of functions $V_0, V_1, \ldots, V_t$ using the Bellman Equation on the MTM:
\begin{equation}
    V_{i+1} (s) = \max_{o \in A \cup \Omega(s)} \left(R_\gamma(s,o) + \sum_{s' \in S} T_\gamma(s' \mid s,o) V_{i}(s') \right).
\end{equation}
Throughout this chapter, I will assume that the model of each option is given to the agent and ignore the computational cost for computing the model for the options. This is yet another assumption that will help simplify the analysis, but will also restrict the scope of the result---indeed, understanding the total difficulty of option discovery (model computation and all), is of deep importance. In \autoref{chap:elm_options}, I develop an alternative model to the MTM that is simpler to estimate while retaining desirable properties.

Here, I study the problem of choosing a subset of options $\mc{O}'$ from a given set $\mc{O}$ to add to $\mc{A}$ that minimizes the number of iterations required for VI to converge.\footnote{To ensure that $|V^*(s) - V_{i}(s)| < \epsilon$ for each $s \in \mc{S}$, run VI until $|V_{i+1}(s) - V_{i}(s)| < \epsilon (1 - \gamma) / 2 \gamma$ for each $s \in \mc{S}$ \cite{williams1993tight}.}

\ddef{$L(\mc{O})$}{
The \textbf{number of iterations} $\bs{L(\mc{O})}$ of VI using the joint action set $\mc{A} \cup \mc{O}$, with $\mc{O}$ a non-empty set of options, is the smallest $b$ at which $|V_b(s) - V^*(s)| < \epsilon$ for all $s \in \mc{S}$.\vspace{2mm}
}

\begin{figure}[t!]
    \centering
    \includegraphics[width=0.36\columnwidth]{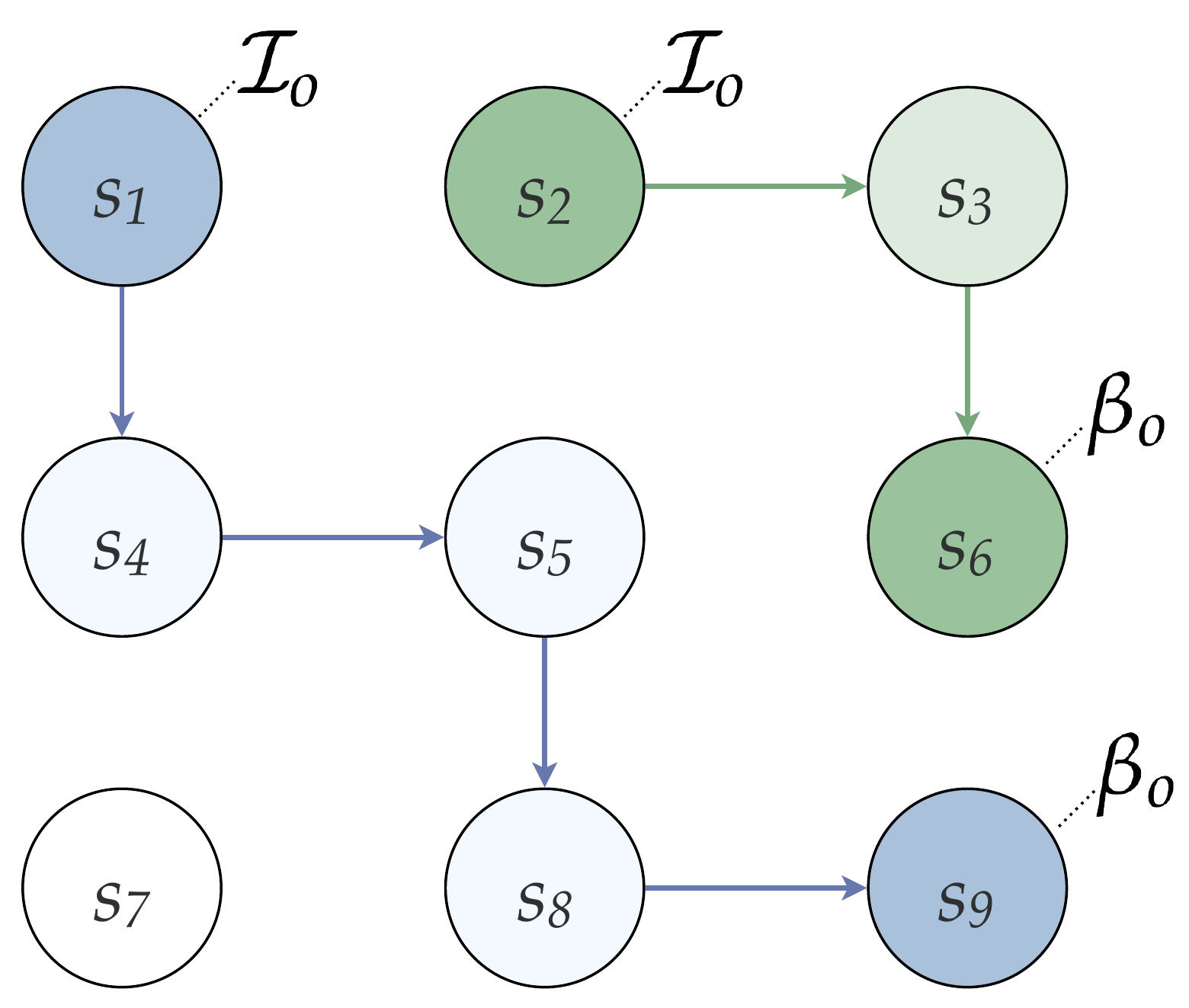}
    \caption{A single option can encode multiple unrelated behaviors.}
    \label{fig:c6_point_option_example}
\end{figure}

\paragraph{Point options.}
Due to the generality of the options framework, a single option can in fact encode several completely unrelated sets of different behaviors. For example, consider the nine-state MDP pictured in \autoref{fig:c6_point_option_example}. In this MDP, I include the initiation, policy, and termination of a single option. The option initiations in $s_1$ and $s_2$, and terminates in $s_9$ and $s_6$. However, the policy executed from $s_1$ and $s_2$ ultimately produce entirely independent trajectories. Consequently, this single option defines two separate behaviors.
%
For this reason, it can be difficult to reason about the impact of adding a single option in the traditional sense---it might be the case that one option in fact defines arbitrarily many separate behaviors. In fact, as the MDP grows larger, a combinatorial number of behaviors can emerge from a single option. It can thus be difficult to address the question: which \textit{single} option helps planning the most? Thus, for the purposes of the analysis, we choose to focus attention on a special subclass of options that only allow for a single continuous stream of behavior:
\ddef{Point option}{
\label{def:c6_point_option}
    A \textbf{point option} is any option $o$ whose initiation set $\mc{I}_o$ and termination condition $\beta_o$ correspond to one state each:
    \vspace{-3mm}
\begin{align}
        |\{s \in \mc{S} : \mc{I}_o(s) = 1\}| &= 1, \\
        |\{s \in \mc{S} : \beta_o(s) > 0\}| = |\{s \in \mc{S} : \beta_o(s) = 1\}| &= 1.
\end{align}
\vspace{-3mm}}

I let $\mc{O}_p$ denote the set containing all point options. Note that trivially $\mc{O}_p \subset \Oall$. Additionally observe two key properties of point options: 1) an arbitrary collection of options in a finite MDP can be represented as a collection of point options, and 2) a point option is simply a regular option paired with a particular kind of state abstraction that groups together all states in the initiation and termination sets (assuming deterministic termination). For these reasons, point options are surprisingly general.

\section{Complexity Results}
\label{sec:complexity}

The main results of this chapter focus on two computational problems, first introduced by \citet{jinnai2019opt};
\begin{enumerate}
    \item \textsc{MinOptionMaxIter} (MOMI): Given a set of options $\mc{O}$, which subset $\mc{O}' \subseteq \mc{O}$ allows
    VI to converge in at most $\ell$ iterations?
    \item \textsc{MinIterMaxOption} (MIMO): Given a set of options $\mc{O}$, which subset of $k$ or fewer options $\mc{O}_k' \subseteq \mc{O} $ will minimizes the number of iterations of VI to convergence?
\end{enumerate}

More formally, the two problems are defined as follows.

\ddef{MOMI}{The \textbf{MinOptionMaxIter} problem is defined as follows: \texttt{given} an MDP $M$, a non-negative real-value $\epsilon$, and an integer $\ell$, \texttt{return} $\mc{O}$ that minimizes $|\mc{O}|$ subject to $\mc{O} \subseteq \mc{O}_p$ and $L(\mc{O}) \leq \ell$.
\vspace{2mm}}

\textsc{MinIterMaxOption} (MIMO).

\ddef{MIMO}{The \textbf{MinIterMaxOption} problem is as follows: \texttt{given} an MDP $M$, a non-negative real-value $\epsilon$, and an integer $k$, \texttt{return} $\mc{O}$ that minimizes $L(\mc{O})$, subject to $\mc{O} \subseteq \mc{O}_p$ and $|\mc{O}| \leq k$.
\vspace{2mm}}

I now present the main result of the chapter, which states that both MOMI and MIMO are NP-hard.
\begin{theorem}
    MOMI and MIMO are NP-hard.
    \label{thm:c6_mimo_momi_np-hard}
\end{theorem}

\noindent This result was first proven by \citet{jinnai2019opt}---see Theorem 1. For proofs of all results introduced in this chapter, see original work by \citet{jinnai2019opt}.

\subsubsection{Generalizations of MOMI and MIMO}
It is natural to consider whether the hardness results of \autoref{thm:c6_mimo_momi_np-hard} extend to more general settings. Let us now turn to extensions of these problems that offer significant coverage of settings relevant to finding the optimal options for accelerating VI.

First, consider the case where the given options are \textit{not} point options, but rather may be an arbitrary subset of $\Oall$. When the given set of options is in fact $\Oall$, MOMI is solved trivially, since the optimal option for accelerating VI is one that initiates everywhere and executes $\pi^*$. However, in practice, the set $\Oall$ is likely to be inaccessible. Instead, it is often preferable to focus on classes of options that can be constructed with a restricted computational or sample budget. To capture this variant of the problem, let us now introduce a generalization of MOMI:
\ddef{MOMI$_{gen}$}{The \textbf{MOMI$_{\textbf{gen}}$} problem is defined as follows: \texttt{given} an MDP $M$, a non-negative real-value $\epsilon$, $\mc{O'} \subseteq \mc{O}_{all}$, and an integer $\ell$, \texttt{return} $\mc{O}$ minimizing $|\mc{O}|$ subject to $L(\mc{O}) \leq \ell$ and $\mc{O} \subseteq \mc{O'}$.
\vspace{2mm}}

In this way, $\mc{O}'$ can denote options that satisfy other criteria, such as those that can be constructed or estimated given some resource budget, or obtain of other desirable properties. As is expected, this problem is again NP-hard, since both MOMI$_{gen}$ and MIMO$_{gen}$ are supersets of MOMI and MIMO respectively.

\begin{theorem}
    MOMI$_{gen}$ and MIMO$_{gen}$ are NP-hard.
\end{theorem}


Another relevant relaxation of MIMO and MOMI is to move to the multitask or lifelong setting discussed in \autoref{chap:state_abstr_lifelong}---how does this problem change when our goal is to identify options that accelerate planning on multiple MDPs? More concretely, given a distribution over MDPs $D$, we would like to compute the smallest set of options $\mc{O}_D'$ that minimize the \textit{expected} number of iterations to solve $M \sim D$. I refer to this problem as MOMI$_{multi}$, defined as follows.

\ddef{MOMI$_{multi}$}{The \textbf{MOMI$_{\textbf{multi}}$} defines the following computational problem: \texttt{given} a probability distribution over MDPs $D$, $\mc{O}' \subseteq \mc{O}_{all}$, a non-negative real-value $\epsilon$, and an integer $\ell$, \texttt{return} $\mc{O}$ that minimizes $|\mc{O}|$ such that $\bE_{M \sim D}[L_M(\mc{O})] \leq \ell$ and $\mc{O} \subseteq \mc{O'}$.
\vspace{2mm}}

As expected, the same extension can be applied to MIMO, too.


\begin{theorem}
    MOMI$_{multi}$ and MIMO$_{multi}$ are NP-hard.
\end{theorem}

The proof follows from the fact that MOMI$_{multi}$ is a superset of MOMI$_{gen}$ and MIMO$_{multi}$ is a superset of MIMO$_{gen}$.

In light of the computational difficulty of both problems, the appropriate approach is to find a suitable approximation algorithms.
However, even approximately solving MOMI is hard. More precisely:
\begin{theorem} {\ }
    \label{thm:c6_MOMIHardness}
    \begin{enumerate}
    \itemsep0em 
        \item MOMI is $\Omega(\log n)$ hard to approximate even for deterministic MDPs unless P = NP.
        \item MOMI$_{gen}$ is $2^{\log^{1-\epsilon}n}$-hard to approximate for any $\epsilon>0$ even for deterministic MDP unless $\text{NP} \subseteq \text{DTIME}(n^{poly \log n})$.
        \item MOMI is $2^{\log^{1-\epsilon}n}$-hard to approximate for any $\epsilon>0$ unless $\text{NP} \subseteq \text{DTIME}(n^{poly \log n})$.
    \end{enumerate}
\end{theorem}


Note that an $O(n)$-approximation is achievable by the trivial algorithm that returns a set of all candidate options. Thus, \autoref{thm:c6_MOMIHardness} roughly states that there is no polynomial time approximation algorithms other than the trivial algorithm for MOMI. 

In the next section we show that an $O(\log n)$-approximation is achievable if the MDP is deterministic, and the agent is given the set containing all point options. Thus, together, these two results give a formal separation between the hardness of abstraction in MDPs with and without stochasticity.

In summary, the problem of computing optimal behavioral abstractions for accelerating VI is computationally intractable.

\section{Approximation Algorithms}
\label{sec:algorithms}

I now provide polynomial-time approximation algorithms, \mimoalg{} and \momialg{}, to solve MOMI and MIMO respectively. Both algorithms have bounded suboptimality that is slightly worse than a constant factor for deterministic MDPs.

The analysis requires several assumptions. First, there is exactly one absorbing state $s_g \in \mc{S}$ with $T(s_g \mid a, s_g) = 1$ and $R(s_g, a) = 0$. Second, that every optimal policy eventually reaches $s_g$ with probability 1. Third, there is no cycle with a positive reward involved in the optimal policy's trajectory. That is, $V^\pi_{+}(s) := {\bE}[ \sum_{t=0}^{\infty} \max\{0, R(s, a)\}] < \infty$ for all policies $\pi$. Note that we can convert a problem with multiple goals to a problem with a single goal by adding a new absorbing state $s_g$ to the MDP and adding a transition from each of the original goals to $s_g$.

Unfortunately, these algorithms are computationally more involved than solving the MDP itself through standard methods, and are thus unlikely to be practical. Instead, they are useful for analyzing and evaluating options discovered by heuristic algorithms. If the option set found by an option discovery method outperforms the option set found by one the following approximation algorithms (in planning performance), then it is strong evidence that the option set found by the heuristic is close to the optimal option set (for that MDP). The approximation algorithms are guaranteed to have bounded suboptimality if the MDP is deterministic, so any heuristic method that provably exceeds our algorithm's performance will also guarantee bounded suboptimality. Further, these algorithms may be a useful foundation to help guide future option discovery methods.


\paragraph{Approximation Algorithm: \momialg.} I now describe a polynomial-time approximation algorithm, \momialg{}, that uses set cover to solve MOMI.
The overview of the procedure is as follows.
\begin{enumerate}
    \item Compute an asymmetric distance function $d_\epsilon(s, s'): \mc{S} \times \mc{S} \rightarrow \mathbb{N}$ representing the number of iterations for a state $s$ to reach its $\epsilon$-optimal value if we add a point option from a state $s'$ to a goal state $s_g$.
    \item For every state $s_i$, compute a set of states $X_{s_i}$ within $\ell - 1$ distance of reaching $s_i$. The set $X_{s_i}$ represents the states that converge within $\ell$ steps if a point option is added from $s_i$ to $s_g$.
    \item Let $\mc{X}$ be a set of $X_{s_i}$ for every $s_i \in \mc{S} \setminus X^+_{g}$, where $X^+_{g}$ is a set of states that converges within $\ell$ without any options.  
    
    \item Solve the set-cover optimization problem to find a set of subsets that covers the entire state space using the approximation algorithm by \citet{chvatal1979greedy}. This process corresponds to finding a minimum set of subsets $\{X_{s_i}\}$ that makes every state in $\mc{S}$ converge within $\ell$ steps.
    
    \item Generate a set of point options with initiation states set to one of the center states in the solution of the set-cover, and termination states set to the goal. 
\end{enumerate}

The distance function $d_{\epsilon}: \mc{S} \times \mc{S} \rightarrow \mathbb{N}$, is defined as follows.

\ddef{Distance $d_\epsilon(s_i, s_j)$}{The asymmetric distance $\bs{d_\epsilon(s_i, s_j)}$ is one minus the number of iterations for $s_i$ to reach $\epsilon$-optimal if a point option is added from $s_j$ to $s_g$.
\vspace{2mm}}

More formally, let $d'_\epsilon(s_i)$ denote the number of iterations needed for the value of state $s_i$ to satisfy $|V(s_i) - V^*(s_i)| < \epsilon$, and let $d'_\epsilon(s_i, s_j)$ be an upper bound of the number of iterations needed for the value of $s_i$ to satisfy $|V(s_i) - V^*(s_i)| < \epsilon$, if the value of $s_j$ is initialized such that $|V(s_j) - V^*(s_j)| < \epsilon$. Let $d_\epsilon(s_i, s_j) := \min(d'_\epsilon(s_i) - 1, d'_\epsilon(s_i, s_j))$.
For simplicity, I use $d$ as shorthand for $d_\epsilon$. 

Note that we need to solve the MDP once to compute $d$. The quantity $d(s, s')$ can be computed once the MDP is solved without any options and have stored all value functions $V_i$ for $i=1,\ldots, b$ until convergence as a function of $V_1$: $V_i(s) = f(V_1(s_0), V_1(s_1),\ldots)$. If a point option is added from $s'$ to $s_g$, then $V_1(s') = V^*(s')$.
Thus, $d(s, s')$ is the smallest $i$ such that $V_{i}(s)$ is $\epsilon$-optimal if we replace $V_1(s')$ with $V^*(s')$ when computing $V_i(s)$ as a function of $V_1$. With these pieces in play, we can now state the properties of A-MOMI.

\begin{theorem}
    \momialg{} has the following properties:
    \begin{enumerate}
        \itemsep0em 
        \item \momialg{} runs in polynomial time.
        \item It guarantees that the MDP is solved within $\ell$ iterations using the option set acquired by \momialg{} $\mc{O}$. 
        \item If the MDP is deterministic, the option set is at most $O(\log n)$ times larger than the smallest option set possible to solve the MDP within $\ell$ iterations.
    \end{enumerate}
\end{theorem}

Note that the approximation bound for a deterministic MDP will inherit any improvements to the approximation algorithm for set cover. Set cover is known to be NP-hard to approximate up to a factor of $(1 - o(1)) \log n$ \cite{dinur2014analytical}, thus there may be an improvement on the approximation ratio for the set cover problem, which will also improve the approximation ratio of \momialg.



\paragraph{Approximation Algorithm: \mimoalg.} The outline of the approximation algorithm for MIMO (\mimoalg{}) is as follows.
\begin{enumerate}
    \item Compute $d_\epsilon(s, s'): \mc{S} \times \mc{S} \rightarrow \mathbb{N}$ for each pair of states.
    
    \item Using this distance function, solve an asymmetric $k$-center problem, which finds a set of center states that minimizes the maximum number of iterations for every state to converge.
    
    \item Generate point options with initiation states set to the center states in the solution of the asymmetric $k$-center problem and termination conditions to the goal.
\end{enumerate}


As in \momialg{}, we first compute the distance function, which is the most computationally demanding part of the algorithm. Then, we use $d$ to solve the asymmetric $k$-center problem~\cite{panigrahy1998ano} on $(\mc{U}, d, k)$ to get a set of centers, which we use as initiation states for point options.
The asymmetric $k$-center problem is a generalization of the metric $k$-center problem where the function $d$ obeys the triangle inequality, but is not necessarily symmetric:

\ddef{AsymKCenter}{Th \textbf{asymmetric $\bs{k}$-center problem} is defined as follows: \texttt{given} a set of elements $\mc{U}$, a function $d: \mc{U} \times \mc{U} \rightarrow \mathbb{N}$, and an integer $k$, \texttt{return} $\mc{C}$ that minimizes $P(\mc{C}) = \max_{s \in U} \min_{c \in \mc{C}} d(s, c)$ subject to $|\mc{C}| \leq k$.
\vspace{2mm}}

We solve this problem using a polynomial-time approximation algorithm proposed by \citet{archer2001two}. The algorithm has a suboptimality bound of $O(\log^*k)$\footnote{$\log^*$ is the number of times the logarithm function must be iteratively applied before the result is less than or equal to 1.} where $k < |\mc{U}|$. It is known that the problem cannot be solved within a factor of $\log^* |\mc{U}| - \theta(1)$ unless P=NP~\cite{chuzhoy2005asymmetric}. As the procedure by \citet{archer2001two} often finds a set of options smaller than $k$, we generate the rest of the options by greedily adding $\log k$ options at once. Finally, we generate a set of point options with initiation-states set to one of the centers and the termination state set to the goal state of the MDP. That is, for every $c \in \mc{C}$, we generate a point option starting from $c$ to the goal state $s_g$.

\begin{theorem}
    \mimoalg{} has the following properties:
    \begin{enumerate}
        \itemsep0em 
        \item \mimoalg{} runs in polynomial time.
        \item If the MDP is deterministic, it has a bounded suboptimality of $O(\log^* k)$.
        \item The number of iterations to solve the MDP using the option set acquired is upper bounded by $P(\mc{C})$.
    \end{enumerate}
\end{theorem}

With the primary analysis established, I now turn to an empirical study of these algorithms and the options they construct.

\section{Experiments}
\label{sec:experiments}

I next turn to an empirical study that examines the performance of VI using options generated by the approximation algorithms on simple grid worlds. The first domain is the same Four Rooms grid world studied in previous chapters, and the second is a 9$\times$9 grid world with no walls. In both domains, the agent's goal is to reach the top right corner.

\begin{figure}[b!]
    \centering
    \newcommand{\optfigsize}{0.31}
    \newcommand{\opthspace}{\hspace{3mm}}
    \subfloat[Optimal, $k = 2$]{\includegraphics[width=\optfigsize\textwidth]{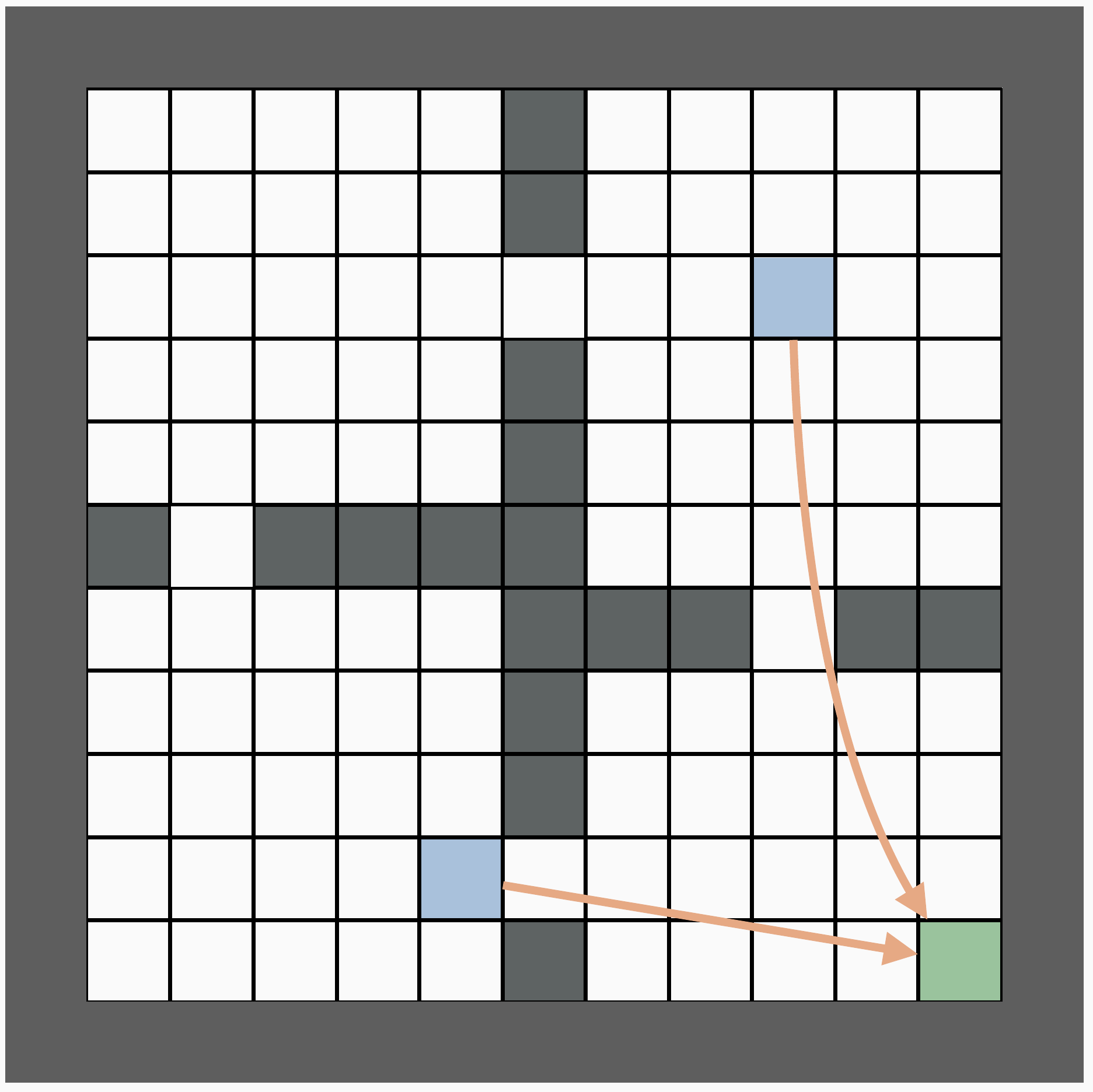} \label{mimo2op}} \opthspace
    \subfloat[Optimal, $k = 4$]{\includegraphics[width=\optfigsize\textwidth]{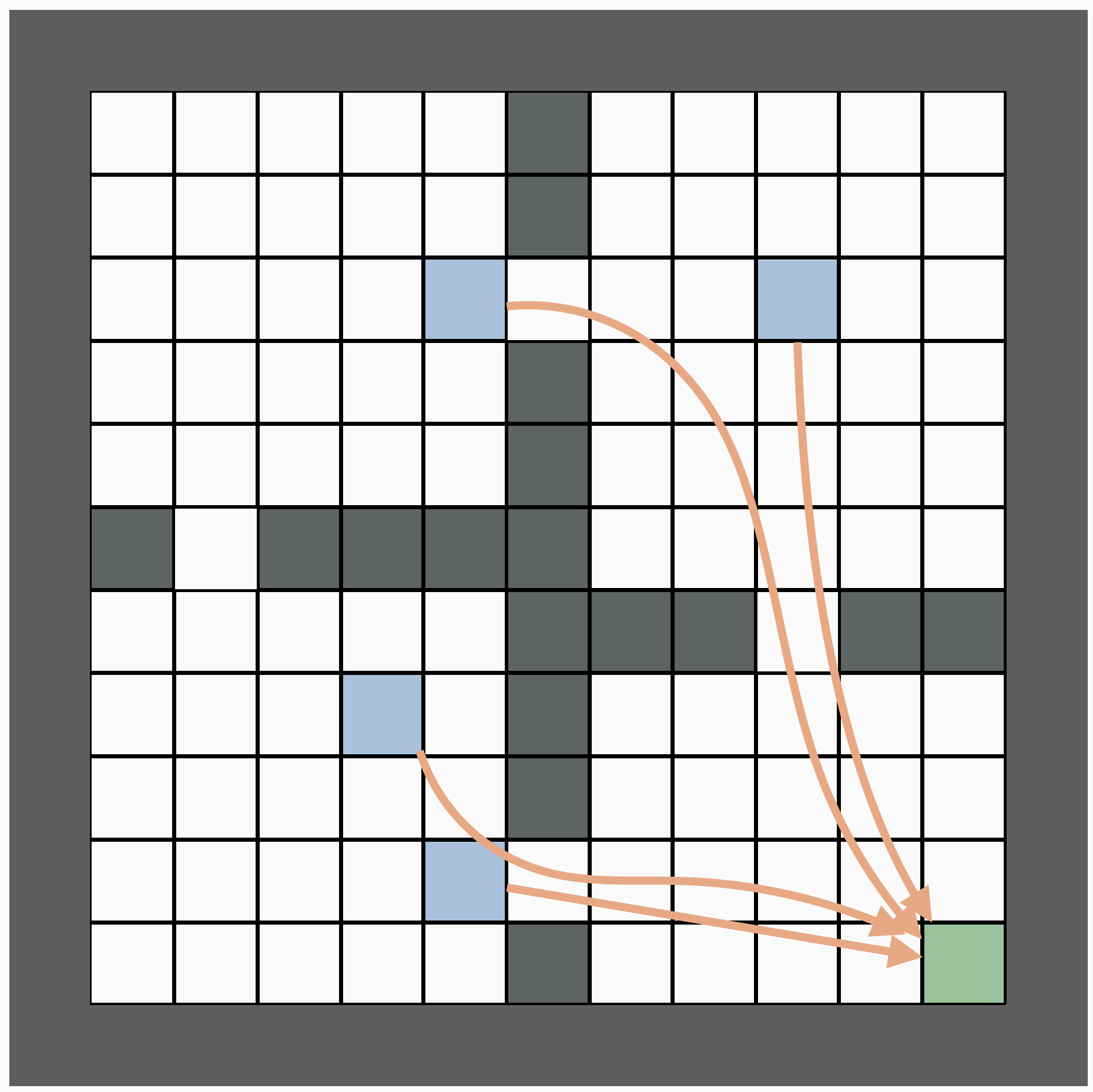} \label{mimo4op}} \opthspace
    \subfloat[Betweenness, $k=4$]{\includegraphics[width=\optfigsize\textwidth]{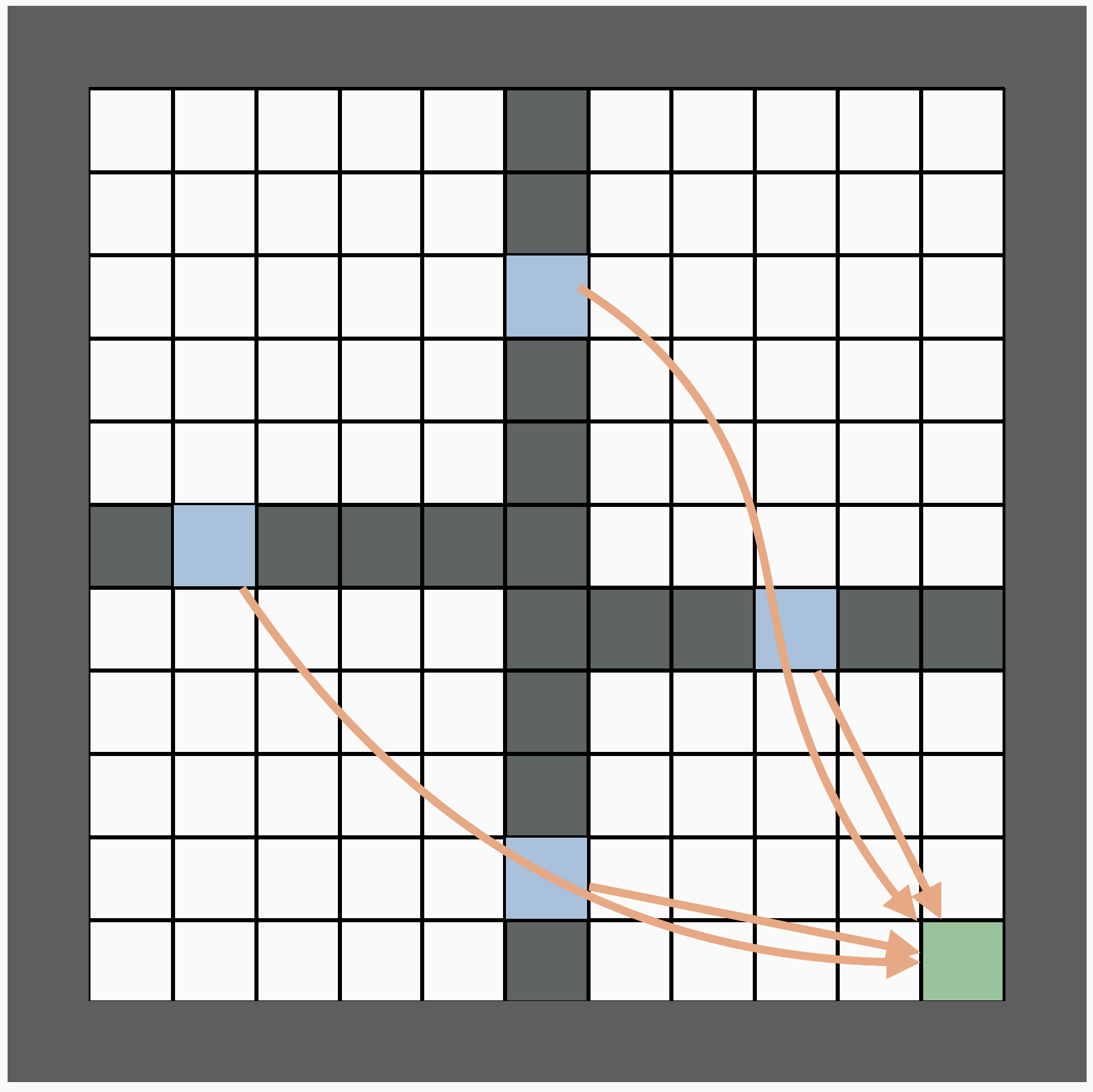} \label{mimo4bet}} \\
    \subfloat[Approximation, $k = 2$]{\includegraphics[width=\optfigsize\textwidth]{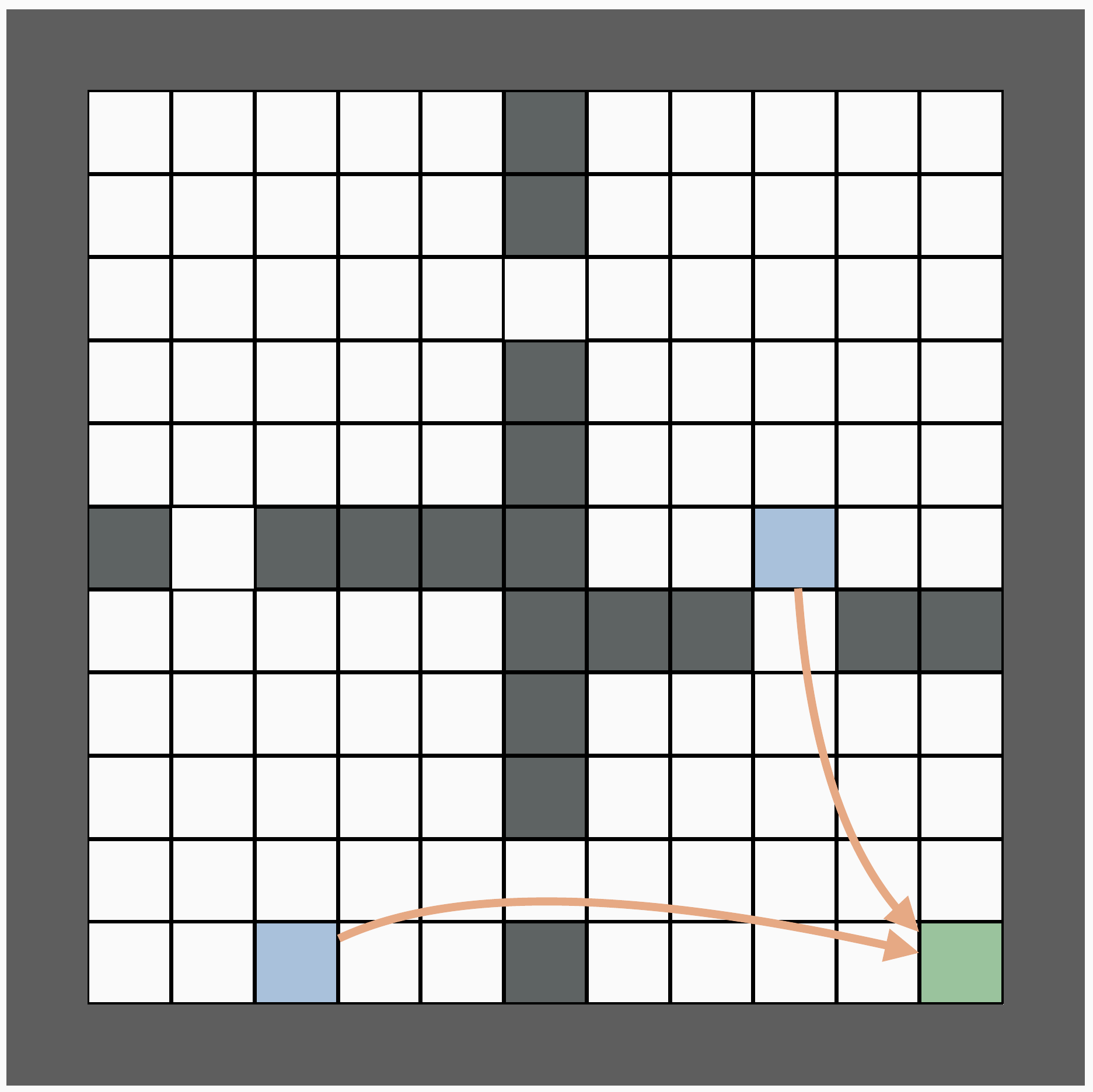} \label{mimo2ap}} \opthspace
    \subfloat[Approximation, $k = 4$]{\includegraphics[width=\optfigsize\textwidth]{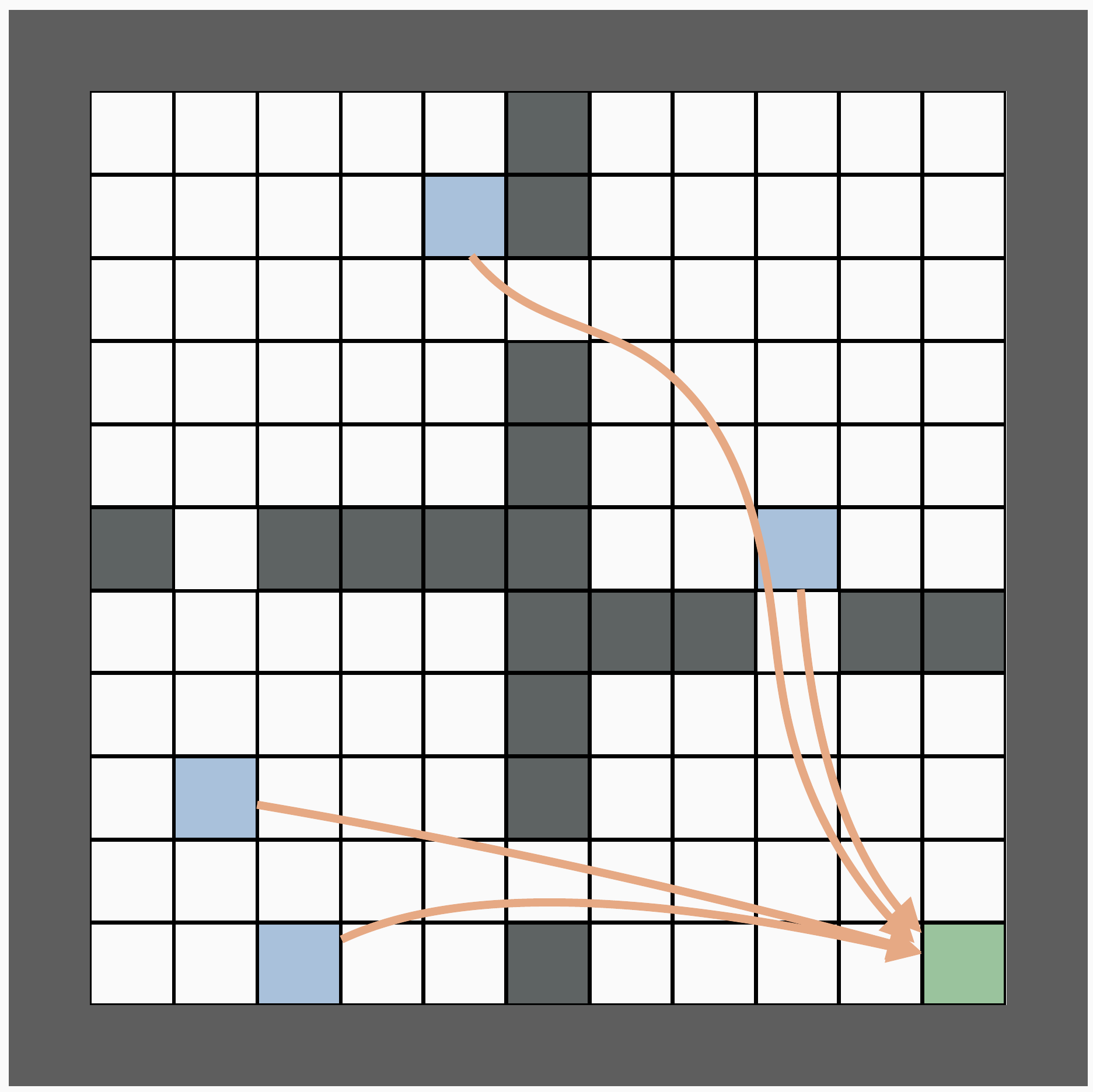} \label{mimo4ap}} \opthspace
    \subfloat[Eigenoptions, $k=4$]{\includegraphics[width=\optfigsize\textwidth]{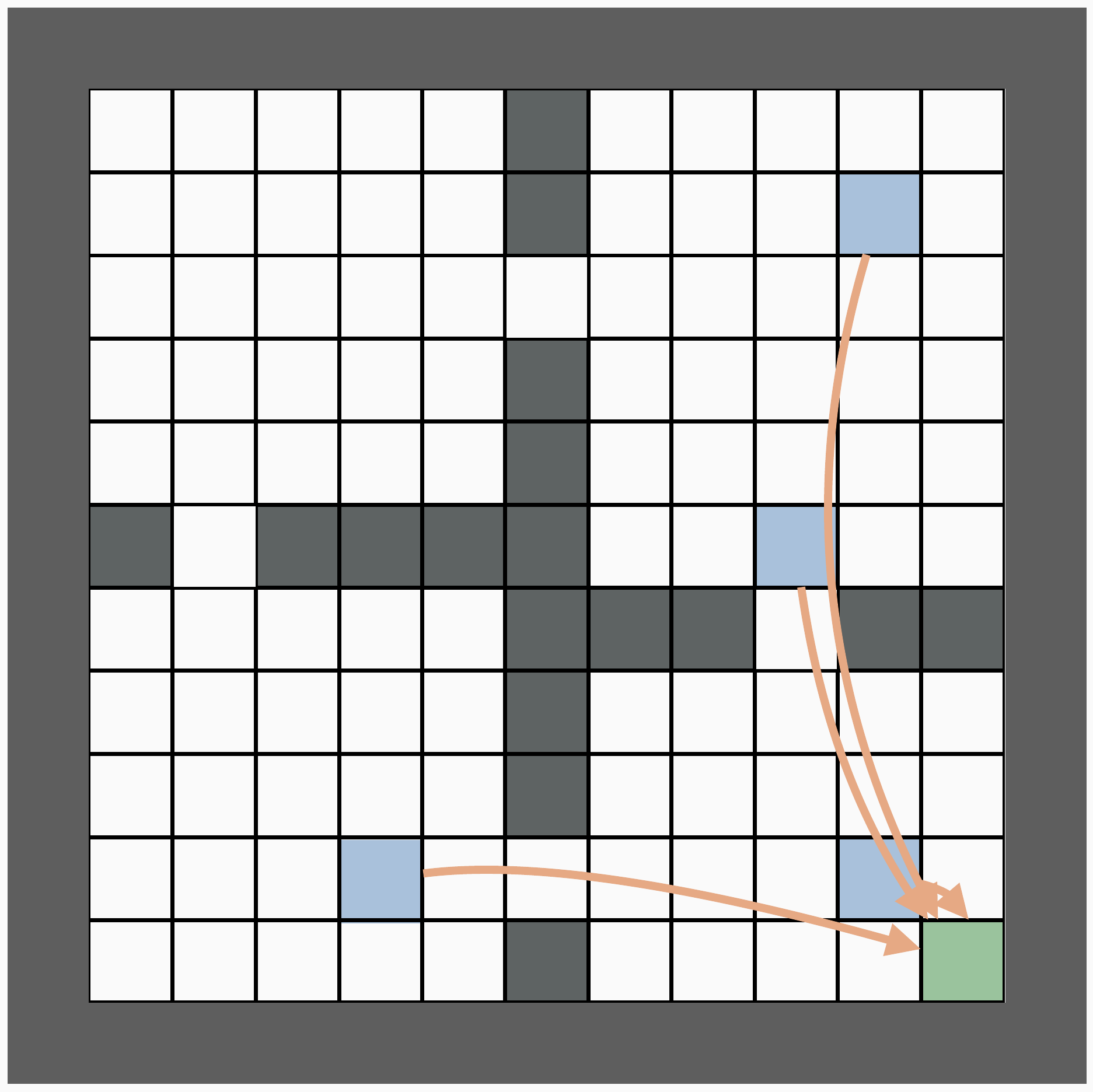} \label{mimo4eg}} 

    \caption{Qualitative comparison of the optimal point options with options generated by the approximation algorithm \mimoalg{}.}
    \label{fig:c6_fourroom-viz}
\end{figure}

\noindent\textbf{Visualizations.} First, we visualize a variety of option types, including the optimal point options, those found by our approximation algorithms, and several option types proposed in the literature.
To generate these visuals, we compute the optimal set of point options by enumerating every possible set of point options and picking the best. We also contrast these options with eigenoptions \cite{machado2017eigenoption} and options based on  betwenness by \citet{csimcsek2009skill}. \autoref{fig:c6_fourroom-viz} shows the optimal and bounded suboptimal set of options computed by \mimoalg{} for $k=2$ and $k=4$.

\autoref{mimo4bet} shows the four bottleneck states with highest shortest-path betweenness centrality in the state-transition graph. Observe that the optimal options are quite close to the bottleneck states, suggesting that bottleneck states are also useful for planning as a heuristic to find useful subgoals. \autoref{mimo4eg} shows the set of subgoals discovered by graph Laplacian analysis following the method of \citet{machado2017laplacian}. Both eigenoptions and betweenness options are designed for use in RL, rather than planning, but there is still meaningful qualitative difference in the options discovered. Indeed, one potential reason to acquire good options to plan is for their eventual use in model-based RL.

While such visuals can only highlight qualitative differences in the different methods, it is still apparent that the approximation algorithm and the optimal algorithm find reasonably similar sets of options. That is, contrasting \autoref{mimo2op} with \autoref{mimo2ap}, we see that the options generated are in fact relatively similar---both initiate in the bottom left and top left rooms respectively, and are around 10 steps away from the goal. The same is true when $k=4$: note that in \autoref{mimo4op} and \autoref{mimo2op} there is roughly one option in each room in both cases.

\begin{figure}[t!]
    \centering
    \newcommand{\figsizeq}{0.45}
    \subfloat[Four Rooms (MIMO)]{\includegraphics[width=\figsizeq\textwidth]{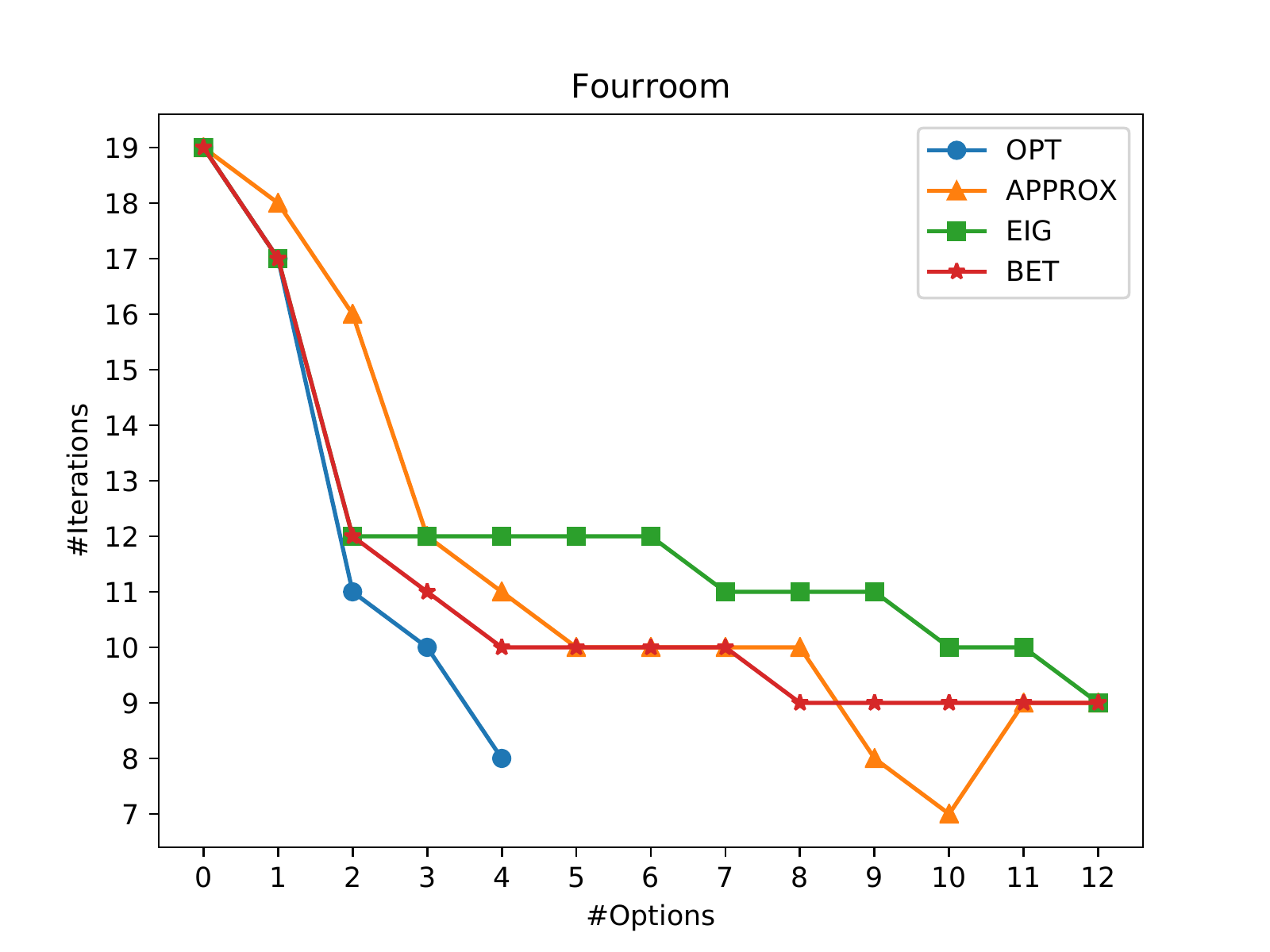} \label{fig:mimofour}} \subfhspace
    \subfloat[$9\times 9$ grid (MIMO)]{\includegraphics[width=\figsizeq\textwidth]{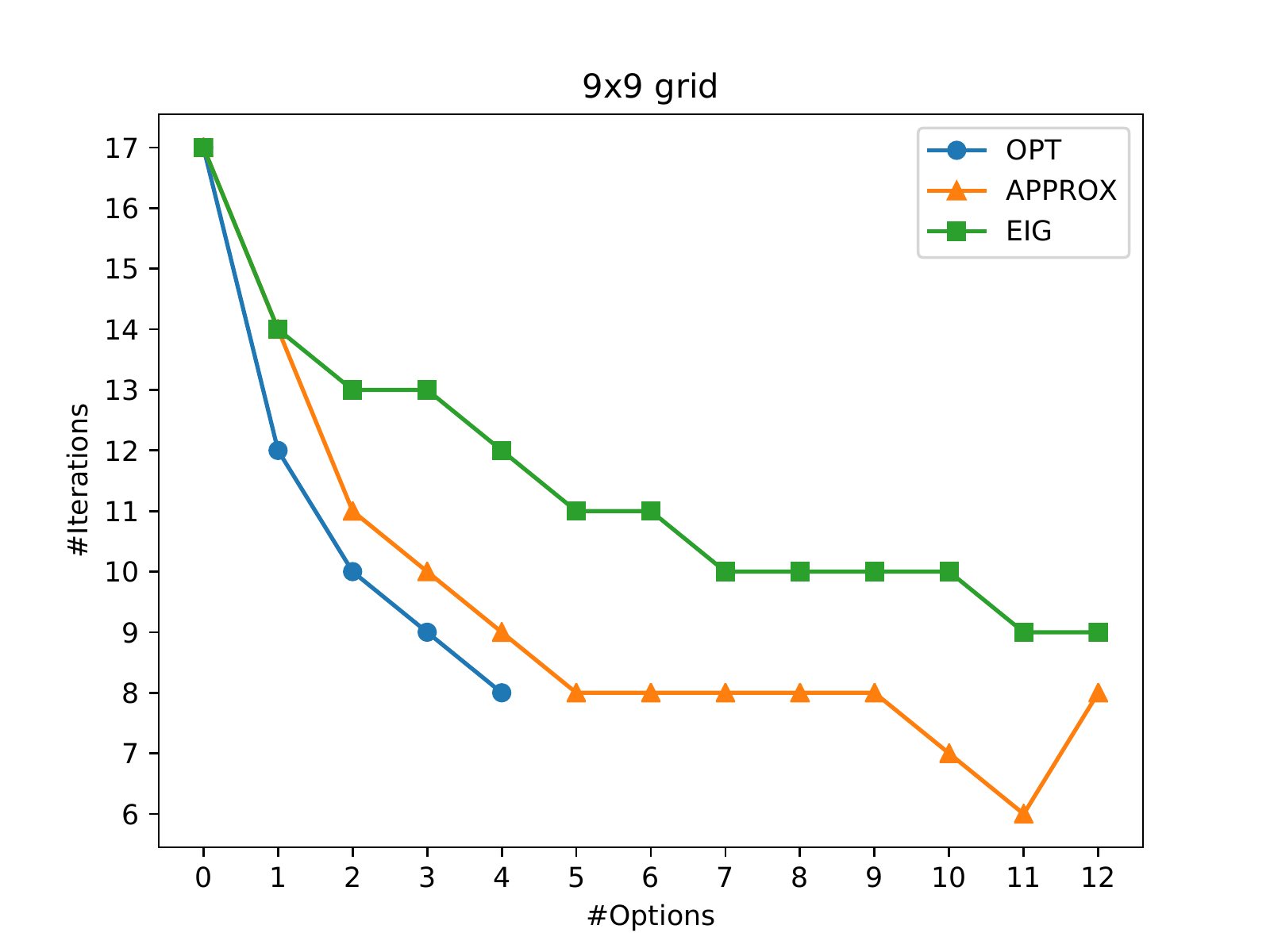} \label{fig:mimogrid}} \\
    \subfloat[Four Rooms (MOMI)]{\includegraphics[width=\figsizeq\textwidth]{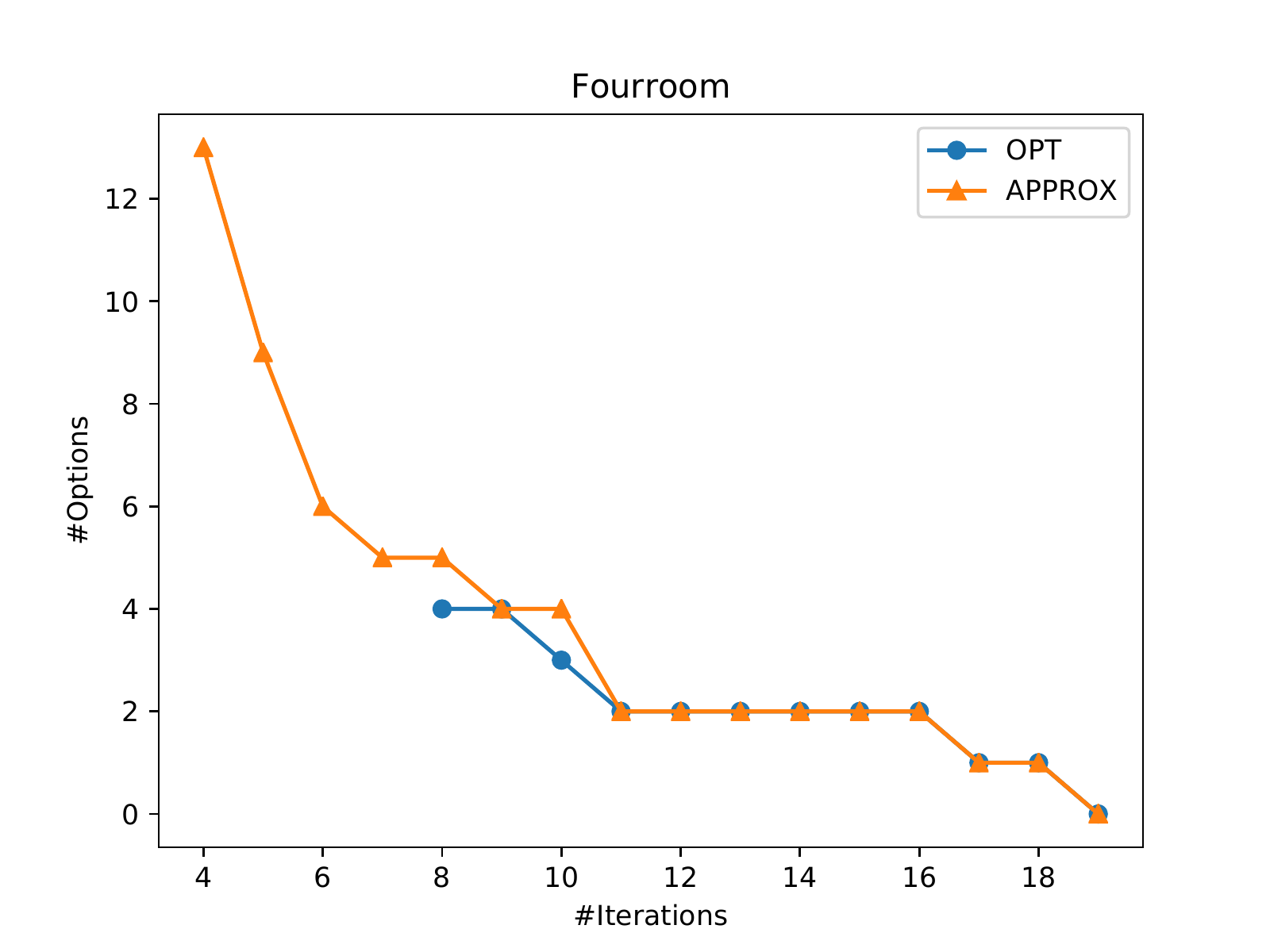} \label{fig:momifour}} \subfhspace
    \subfloat[$9\times 9$ grid (MOMI)]{\includegraphics[width=\figsizeq\textwidth]{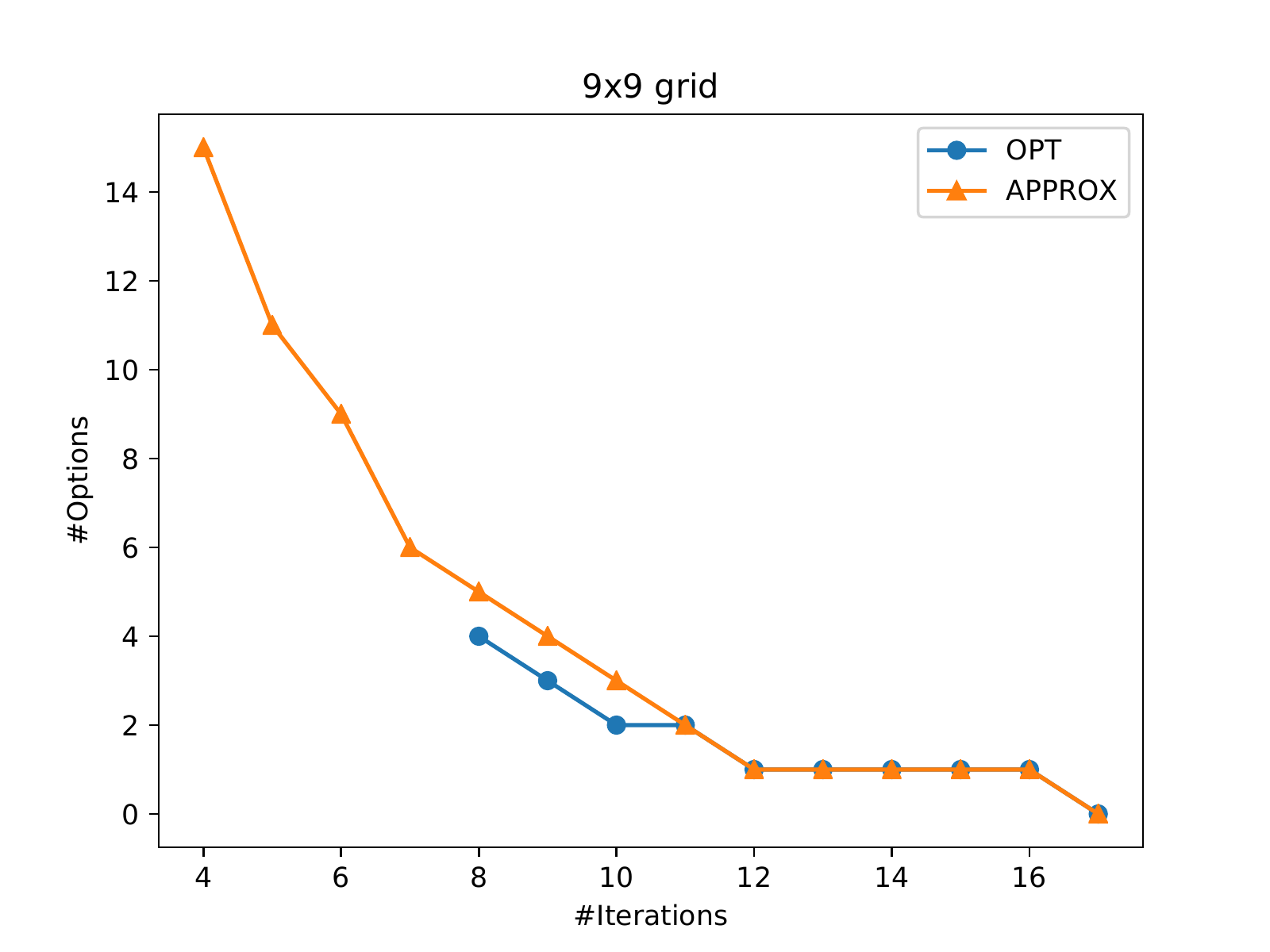} \label{fig:momigrid}}
    \caption{Quantitative evaluation comparing the planning speed up resulting from the options computed by solving MIMO and MOMI in various ways.}
    \label{fig:c6_numiter-op}
\end{figure}

\noindent\textbf{Quantitative Evaluation.} Next, we turn to a quantitative evaluation to directly contrast the impact the discovered options have on the speed of VI. Specifically, we run VI using the set of options generated by \mimoalg{} and \momialg{} and compare their performance to the optimal set of options found by solving the full NP-hard problem.
\autoref{fig:mimofour} and \autoref{fig:mimogrid} present the number of iterations on the Four Rooms and $9\times 9$ grid using a set of options of size $k$.
The experimental results suggest that the approximation algorithm tends to find a set of options slightly worse in performance than the optimal ones.
For betweenness options and eigenoptions, we evaluated every subset of options among the four and present results for the best subset found.
Because betweenness options are placed close to the optimal options, the performance is close to optimal especially when the number of options are small.

In addition, we used \momialg{} to find a minimum option set to solve the MDP within the given number of iterations.
\autoref{fig:momifour} and \autoref{fig:momigrid} show the number of options generated by \momialg{} compared to the minimum number of options. As the data indicate, the optimal approach and \momialg{} find options of similar quality, suggesting that in these simple problems, the approximation algorithm is as effective as solvine thee full NP-hard problem.

In this chapter, I address a fundamental question concerning the use of action abstractions that help accelerate planning in MDPs.
This led to two problem formulations for finding options: 1) minimize the size of option set given a maximum number of iterations (MOMI) and 2) minimize the number of iterations given a maximum size of option set (MIMO).
The main results prove that these two problems are both computationally intractable under several assumptions---we here suppose the option models are given, the branching factor is ignored, and that VI is sufficiently general to capture planning algorithms. These assumptions do limit the scope of these results. Fortunately, each problem also permits a polynomial-time algorithm for MDPs with bounded reward and goal states, with bounded optimality for deterministic MDPs. Experimental data support the usefulness of the approximation algorithms, though it is important to be mindful of their computational costs.

%% file: chapters/c7_aa_elm.tex
\begin{center}
\begin{minipage}{0.8\textwidth}
\textit{This chapter is based on ``The Expected-Length Model of Options" \cite{abel_winder2019elm} jointly led by John Winder, also with Marie desJardins and Michael L. Littman.}
\end{minipage}
\end{center}
\vspace{2mm}

\begin{figure}[b!]
    \centering
    \subfloat[Multi-Time Model\label{fig:c7_mtm_ex}]{\includegraphics[width=0.55\textwidth]{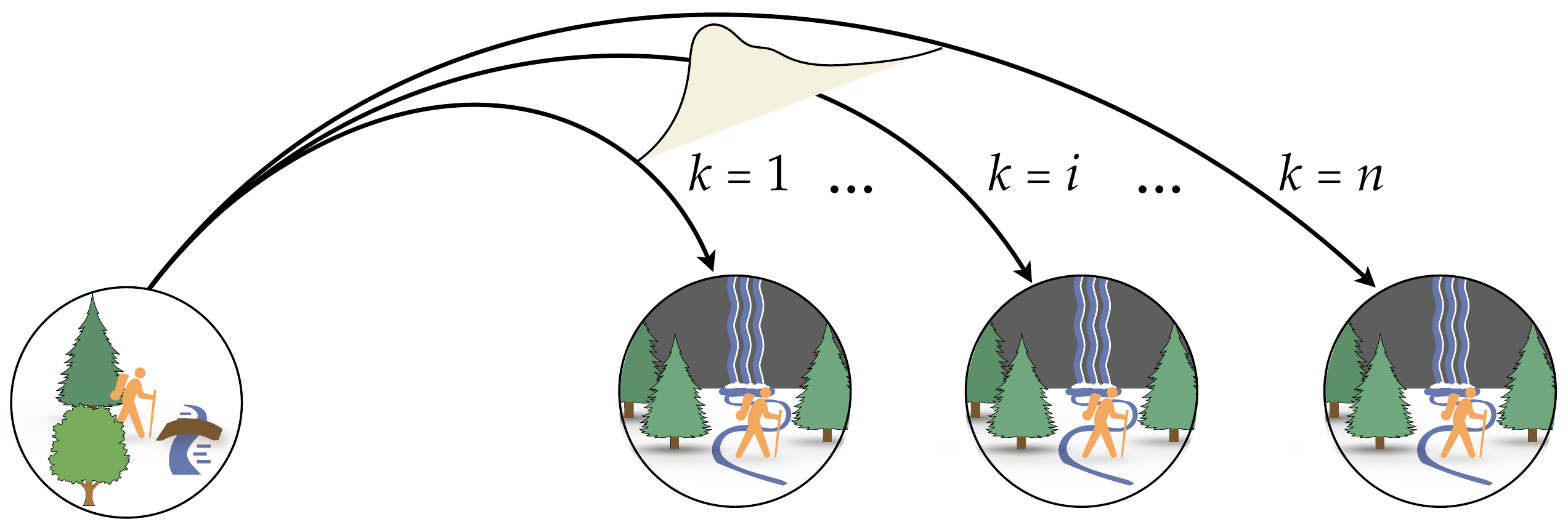}} \subfhspace
    \subfloat[Expected-Length Model\label{fig:c7_elm_ex}]{\includegraphics[width=0.35\textwidth]{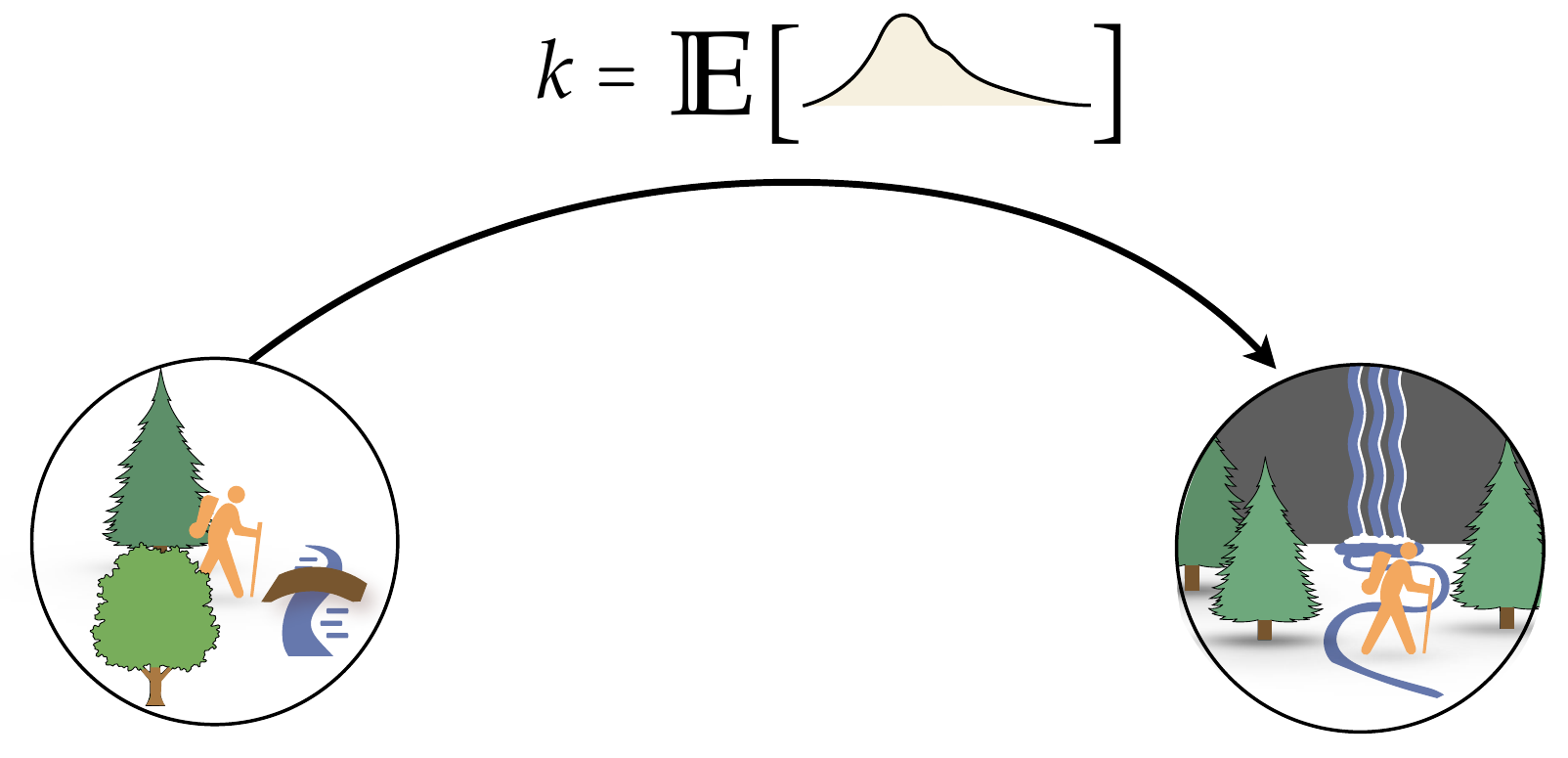}}
    \caption{An intuitive illustration of the MTM (left) and ELM (right).}
    \label{fig:c7_elm_mtm_forest}
\end{figure}

Making accurate long horizon predictions about the effects of an action can improve an agent's ability to make meaningful decisions. For instance, the hiker in the forest from \autoref{chap:introduction} is sure to rely on the ability to predict which high level behaviors will lead them to the bridge, the tent, or the waterfall. With such predictive power, agents can take into account the long-term outcomes of an action, and use this information to make informed plans that account for contingencies, uncertainty, and ultimately help determine which actions will maximize value.

However, learning models that are suitable for use in making long horizon predictions is challenging. Even $\eps$-accurate one-step models are known to lead to an exponential increase in the error of $n$-step predictions as a function of the horizon~\cite{kearns2002near,brafman2002rmax}, though recent approaches show it is possible to diminish this error through smoothness assumptions about the environment~\cite{asadi2018lipschitz}.
Moreover, composing an accurate one-step model into an $n$-step model is known to give rise to predictions of states dissimilar to those seen during training of the model, leading to poor generalization~\cite{talvitie2017self}. By encoding only relevant long horizon sequences of behavior, options offer one promising approach for supporting the discovery of accurate long term models.
How to obtain an option model tractably, however, remains an open question. To this end, this chapter studies the problem of efficiently computing option models from experience.

I first discuss the sense in which the traditional multi-time model (MTM) of options~\cite{precup1997multi,precup1998multi}, is highly parameterized, and thus difficult to compute or learn under reasonable constraints. Intuitively, the density modeled by the MTM tracks the outcome of a given option over all possible time steps (\autoref{fig:c7_mtm_ex}), which can be impractical to compute even in small domains.
In light of this difficulty, I motivate the construction of an alternate model that I call the expected-length model (ELM). The main idea behind the ELM, and indeed, this chapter, is that it is not necessary to model the full joint distribution of possible outcomes of an option, as in the MTM. Instead, it is sufficient to estimate 1) how long, on average, an option takes to run, and 2) a categorical distribution over states where the option terminates. The ELM is formed by combining these two pieces of information (\autoref{fig:c7_elm_ex}) .

I then prove that in goal-based MDPs, the differences in value functions induced by the MTM and the ELM is bounded as a function of the horizon and other relevant properties of the option and environment. I then conduct an empirical study contrasting the performance of the ELM with the MTM in a variety of MDPs. The findings from these experiments suggest that in the right kinds of problems, the ELM is a suitable replacement for the MTM.

\section{The Expected-Length Model}
I now introduce the new option model and motivate its properties. The ELM is defined as follows.

\ddef{Expected-length model of options}{The \textbf{expected-length model} (ELM) for a given option $o$ in state $s$ supposes that the distribution of time steps taken by the option can be well approximated by its expected value, ${\mu_k}$:
\begin{align}
    T_{\mu_k}(s' \mid s, o) &:= \gamma^{\mu_k} p(s' \mid s, o), \label{eq:elm} \\
    R_{\mu_k}(s,o,s') &:= \gamma^{\mu_k}\bE\left[r_1 + r_2 \ldots + r_{\mu_k} \mid s, o\right], \label{eq:elm_reward}
\end{align}
where $p(s' \mid s, o)$ denotes the probability of terminating in $s'$, given that the option was executed in $s$.
}

Modeling only the expected number of time steps throws away information---it ignores, essentially, the particulars of how executing the option can play out.
Consider an agent in the usual Four Rooms domain with an option for moving from the top-left room to the top-right one. Suppose the primitive actions are stochastic, with a small probability of moving the agent in the wrong direction.
Due to the non-zero probability of slipping, the option may sometimes take five, ten, or even more steps to reach the top-right room.
Instead of modeling the full distribution of the number of time steps taken, ELM averages over these quantities (represented by $\mu_k$), and models the transition as taking place over this expected number of time steps. 
I provide additional intuition for ELM in \autoref{sec:intuition} by working through a concrete example.

The main result of this chapter demonstrates that this process of distillation is acceptable and desirable, leading to simpler models and often improving the rate at which models are learned.
Specifically, I prove that, under mild assumptions, ELM induces similar value functions to MTM, where the bound depends on primarily on the amount of stochasticity in the MDP (and the option's trajectory). From experimental evidence, I conclude that ELM option models can perform competitively to MTM.

\section{A Simple Example}
\label{sec:intuition}

Let us first develop intuition behind ELM through an example, concentrating on the transition model.

\begin{figure}[b!]
    \centering
    \subfloat[\label{fig:c7_elm_example}]{\includegraphics[width=0.31\columnwidth]{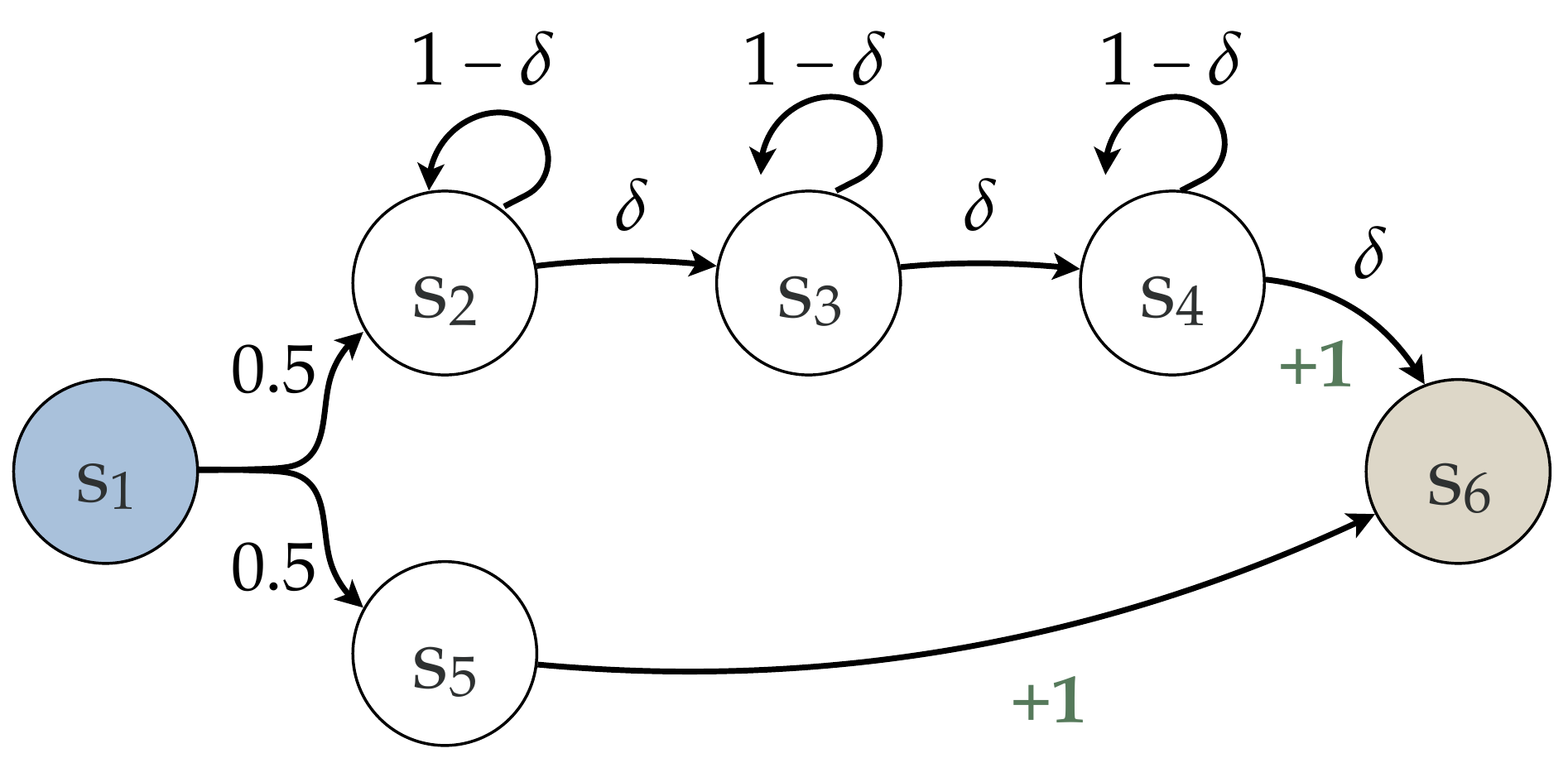}} \subfhspace
    \subfloat[Model Diff. $\delta = 0.4$ \label{fig:c7_split_pr_k_elm_mtm}]{\includegraphics[width=0.3\columnwidth]{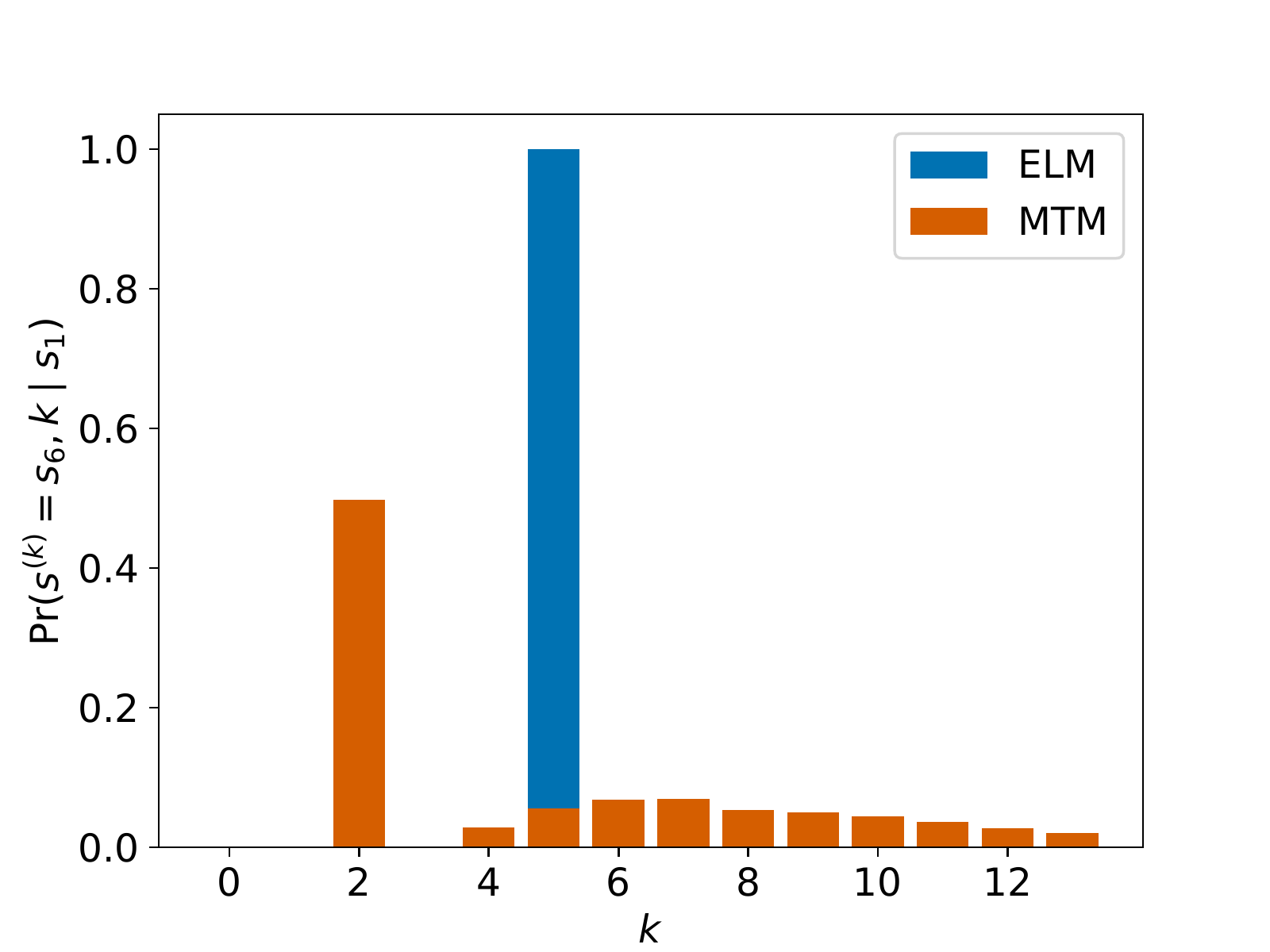}} \subfhspace
    \subfloat[Value Diff. vs. $\delta$ \label{fig:c7_split:value_diff}]{\includegraphics[width=0.3\columnwidth]{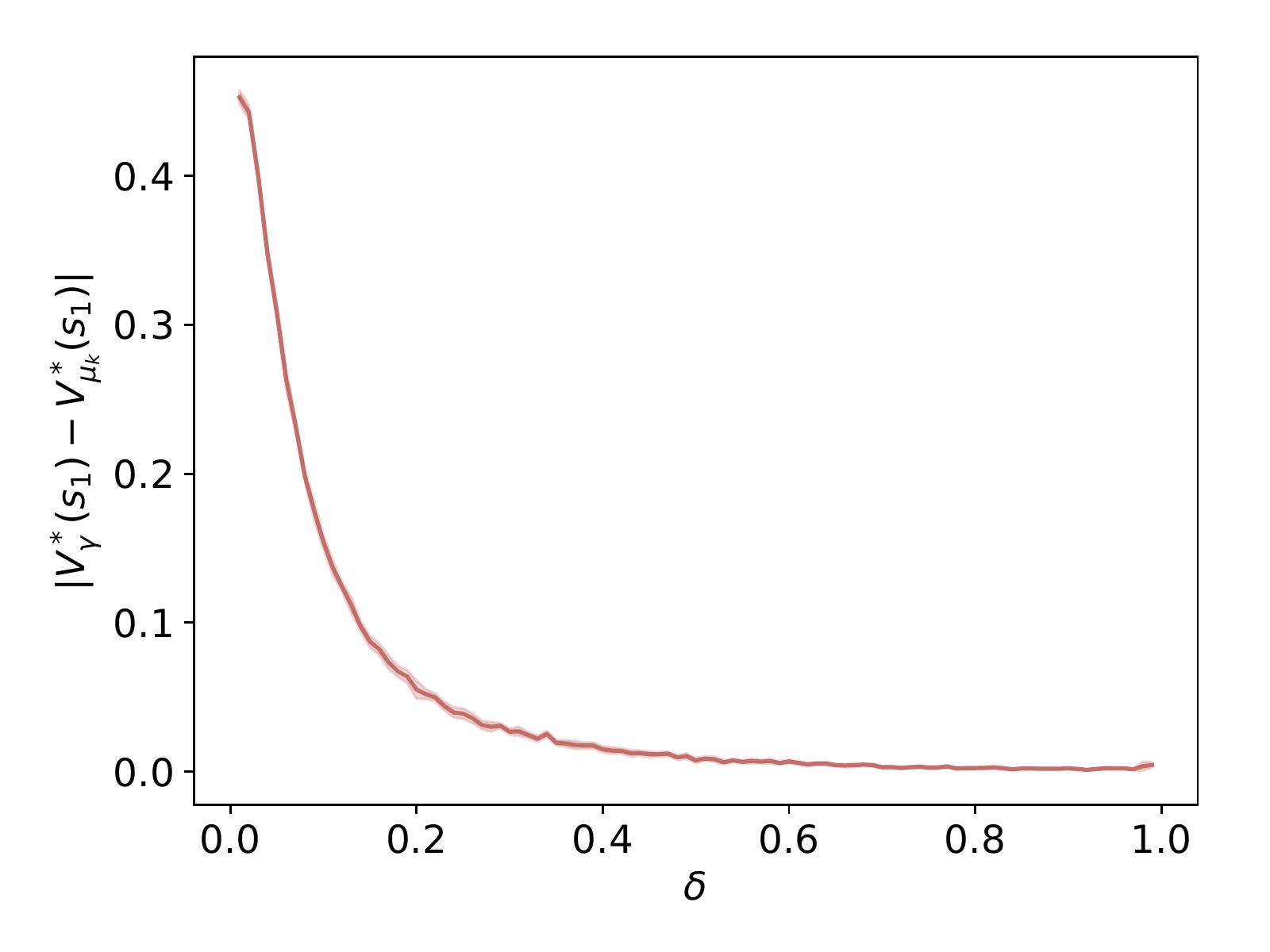}}

    \caption{An illustration of the difference between ELM and MTM.}
    \label{fig:c7_elm_mtm_diff_example}
\end{figure}

\begin{example}
Consider the six-state MDP in \autoref{fig:c7_elm_example}, chosen to accentuate the differences in ELM and MTM. Suppose an option initiates in $s_1$ (shown in blue), and terminates in $s_6$ (shown in tan). For simplicity, suppose $\beta_o(s) = \indic\{s = s_6\}$. The option policy is depicted by the arrows---when the option executes its policy in $s_1$, it lands in $s_2$ with probability $\frac{1}{2}$ and $s_5$ with probability $\frac{1}{2}$. In $s_2$, when the option executes its policy, the agent stays in $s_2$ with probability $1-\delta$, and transitions to $s_3$ with probability $\delta$ (and so on for $s_3$ and $s_4$). Conversely, in $s_5$, the option transitions to $s_6$ with probability 1.
\end{example}

Consider the process of estimating the option model at $s_6$: $T_\gamma(s_6 \mid s_1, o)$ under the MTM. To construct a proper estimate, the MTM must estimate the probability of termination in each state over all possible time steps to determine $\PR(s^{(1)} = s_6 \mid s_1, o), \PR(s^{(2)} = s_6 \mid s_1, o), \ldots$. This computation involves estimation over arbitrarily many time steps; in some cases, like this one, we might find a closed form based on convergence of the geometric series, but agents cannot always intuit this fact from limited data. In contrast, the ELM models this distribution according to $\mu_k$, the average number of time steps.

Given the true MDP transition function $T$, I run $n$ rollouts of the option to termination.
Supposing each rollout reports $(s, o, r, s', k)$, with $r$ the cumulative reward received and $k$ the number of time steps taken, it is natural to estimate $\mu_k$ with the maximum likelihood estimator (MLE) $\hat{\mu}_k = \frac{1}{n}\sum_{i=1}^n k_i$.
Additionally, we can estimate $p(s' \mid s, o)$, the probability that $o$ terminates in $s'$, by modeling it as a categorical distribution with $\ell = |\mc{S}|$ parameters. Then, it is feasible to estimate each $\ell_i$ with an MLE.

To summarize:
\begin{itemize}
    \item The ELM estimates $\mu_k$ and $p(s' \mid s,o)$, for each $s'$ of relevance, by using an MLE based on data collected from rollouts of the option.
    \item The MTM must estimate the probability of terminating in each state, at each time step. It is unclear how to capture this infinite set of probabilities of value economically.
\end{itemize}

\autoref{fig:c7_split_pr_k_elm_mtm} shows their differences in the quantity $\PR(s^{(k)} = s_6 \mid s_1, o)$, for each time step $k$.
The MTM (in orange) distributes the transition probability across many lengths $k$.
Approximately half of the time, $s_6$ is reached in two steps via the bottom route through $s_5$; the rest of the probability mass is spread across higher values, reflecting longer paths (via $s_2$).
The ELM (in blue) instead assumes the option takes $\mu_k = 5$ steps. For both models, each non-zero bar represents a parameter that needs to be estimated, giving a sense of the difficulty in estimating each distribution.

Next consider the mean value difference under each model averaged over 20 runs, presented with 95\% confidence intervals in \autoref{fig:c7_split:value_diff}. Observe that this value difference which decreases to nearly 0.15 as $\delta$ tends to $1$ (with $\textsc{VMax} = 1.0$). This trend is predicted by the analysis I conduct in \autoref{sec:analysis}, which suggests that the higher the variance over expected number of time steps, the more the ELM deviates from the MTM.

\subsubsection{Difficulty of Finding Option Models} \label{sec:mtm_is_hard}

The goal of the ELM is to simplify the MTM to be able to estimate and compute the model of a given option more efficiently.

\paragraph{Estimation.} Learning an option's MTM involves estimating a complicated probability distribution.
Specifically, the general case requires parameters for the (potentially unbounded) number of time steps taken to reach a given $s'$ conditioned on initiating $o$ in $s$, for each $s' \in \mc{S}$.
For such cases, a common assumption to make in analyzing complexity is to model the process out to some finite horizon. One reasonable approximation might involve limiting the sum inside the MTM to the first $\lambda = (1-\gamma)^{-1}$ steps as an artificial horizon, thereby yielding $\lambda|\mc{S}|^2$ parameters to estimate.
In contrast, ELM requires learning the parameters of a categorical distribution indicating the probability of terminating in each state.
With one multinomial for each state, any learning algorithm must estimate $2|\mc{S}|^2$ total parameters.
Depending on the stochasticity inherent in the environment, option policy, and option-termination condition, estimating this smaller number of parameters is likely to be considerably easier $(\lambda \gg 2)$.

\paragraph{Computation.} The MTM requires performing the equivalent computation of a Bellman backup until the option is guaranteed to have terminated \textit{just to compute the option's reward function} (\autoref{eq:mtm_reward}).
Due to the decreasing relevance of future time steps due to $\gamma$, one might again only compute out to $\lambda$ time steps to determine $R_{\gamma}$ and $T_{\gamma}$.
Thus, computing $R_{\gamma}$ is roughly as hard as computing the value function of the option's policy (at least out to $\lambda$ time steps), requiring computational hardness similar to that of an algorithm like Value Iteration (\autoref{alg:value_iteration}), which is known to be $O(|\mc{S}|^2 |\mc{A}|)$ per iteration, with a rough convergence rate of $\tilde{O}(\lambda|T|)$ for $|T|$ as a measure of the complexity of the true transition function~\cite{tseng1990solving,littman1995complexity}.
Conversely, ELM is well suited to construction via Monte Carlo methods.
Consider a single simulated experience $e = (s, o, r, s', t)$, of the initial state, the option, termination state, cumulative reward, and time taken.
This experience contains each data point needed to compute the components of option $o$'s ELM (\autoref{eq:elm} and \autoref{eq:elm_reward}), all sampled directly from the appropriate distributions. With the ELM, option models can be learned from these simulations, $\mc{E}$, with each $e \in \mc{E}$ needing only labels of where the option began, where it ended, how much reward it received, and how long it took.
It is therefore sufficient to run a number of rollouts proportional to the desired accuracy when using ELM.
Relying on such methods for computing the MTM again requires estimating a potentially large number of parameters, which is often untenable.

In considering both estimation and computation, note that these are not conclusive analyses of the computational and statistical difficulty of obtaining each model, but take the insights discussed to serve as sufficient motivation for further exploration of the ELM. For instance, there is some similarity in determining the MTM and TD($\lambda$) when $\lambda = 1$~\cite{sutton1988learning}, so such estimation can be feasible (see, for instance, Chapter 4 of~\citet{parr1998hierarchical}).

I now turn to the primary analysis of the chapter, which illustrates the deviation between the MTM and the ELM for each of the transition, reward, and value functions.

\section{Analysis} \label{sec:analysis}

The main result of this chapter bounds the value difference between the ELM and MTM in goal-based MDPs. This theorem holds under the following two assumptions.

\begin{assumption}
All MDPs considered are goal-based. That is, in each MDP, there is a unique goal state $s_g$ such that $\forall_{s \in \mc{S}, a\in\mc{A}} : R(s,a,s_g) = 1$, where $T(s_g \mid s_g, a) = 1$. All other rewards are zero.
\label{asmptn:mdp_is_gb}
\end{assumption}
This assumption is used to bound the difference of the ELM and MTM reward functions, but is not required to bound their transition functions.

The next assumption is useful in the analysis, but conceptually there is no reason it cannot be removed in future.

\begin{assumption}
For every option, the termination probability is non-zero in every state, bounded below by a fixed constant $\beta_{\min} \in (0,1]$.
\label{asmptn:beta_min}
\end{assumption}

Indeed, while these assumptions slightly limit the scope of the analysis of the ELM, I take the setting to still be sufficiently interesting to offer insights about learning and using option models. Naturally, the relaxation of each assumption is a sensible direction for future work.


At a high level I now show that the following claims hold under these two assumptions:
\begin{enumerate}
    \item \autoref{lem:c7_tgamma_tkappa}: $||T_\gamma - T_{\mu_k}||_\infty$ is bounded.

    \item \autoref{lem:c7_rkappa_rgamma}: $||R_\gamma - R_{\mu_k}||_\infty$ is bounded in goal-based MDPs.

    \item \autoref{thm:c7_vkappa_vgamma}: $||V_\gamma^* - V_{\mu_k}^*||_\infty$ is bounded in goal-based MDPs.
\end{enumerate}

I begin with the two lemmas that show the transition and rewards of the ELM are reasonable approximations of the MTM in goal-based MDPs.
 
\begin{lemma}
\label{lem:c7_tgamma_tkappa}
Under \autoref{asmptn:beta_min}, the expected-length transition model is sufficiently close to the transition model of the multi-time model. More formally, for any option $o \in \mc{O}$, for some real $\tau > 1$, for $\delta = \frac{\sigma_{k,o}^2}{\tau^2}$, and for any state pair $(s,s') \in \mc{S} \times \mc{S}$, with probability $1-\delta$: 
\begin{equation}
    |T_\gamma(s' \mid s,o) - T_{\mu_k}(s' \mid s,o)| \leq \gamma^{{\mu_{k,o}} - \tau} (2\tau + 1) e^{-\beta_{\min}}.
\end{equation}
\end{lemma}

\input{proofs/c7/c7_tkappa_tgamma.tex}

\begin{lemma}
\label{lem:c7_rkappa_rgamma}
Under \autoref{asmptn:mdp_is_gb} and \autoref{asmptn:beta_min}, ELM's reward model is similar to MTM's reward model. More formally, for a given option $o$, for $\delta = \frac{\sigma_{k,o}^2}{\tau^2}$, for some $\tau > 1$, for any state $s$:
\begin{equation}
    |R_\gamma(s,o) - R_{\mu_k}(s,o)| = |T_\gamma(s_g\mid s,o) - T_{\mu_k}(s_g\mid s,o)|.
\end{equation}
Thus, by \autoref{lem:c7_tgamma_tkappa}, with probability $1-\delta$:
\begin{equation}
    |R_{\gamma}(s,o) - R_{\mu_k}(s,o)| \leq \gamma^{{\mu_{k,o}} - \tau} (2\tau + 1) e^{\beta_{\min}}.
\end{equation}
\end{lemma}
%
\input{proofs/c7/c7_rkappa_rgamma.tex}
%
%
%
Notably, \autoref{lem:c7_tgamma_tkappa} does not depend on \autoref{asmptn:mdp_is_gb}---it applies to any finite MDP. Hence, it is likely that the ELM reward function is similar to the MTM in a more general class of MDPs than goal-based problems, but I leave such a direction open for future work. With these lemmas in place, I now present the main result of the chapter.

\begin{theorem}
\label{thm:c7_vkappa_vgamma}
In goal-based MDPs, the value of any policy over options under ELM is bounded relative to the value of the policy under the multi-time model, with high probability.

More formally, under \autoref{asmptn:mdp_is_gb} and \autoref{asmptn:beta_min}, for any policy over options $\pi_o$, some real valued $\tau > 1$, $\eps = \gamma^{{\mu_{k,o}} - \tau} (2\tau + 1) e^{-\beta_{min}}$, $\delta = \frac{\sigma^2}{\tau^2}$, for any state $s \in \mc{S}$, with probability $1-\delta$:
\begin{equation*}
    |V_\gamma^{\pi_o}(s) - V_{\mu_k}^{\pi_o}(s)| \leq \frac{\eps(1-\gamma^{\mu_k}) + \gamma^{\mu_k} \frac{\eps}{2}\textsc{RMax}}{(1-\gamma^{\mu_k})(1-\gamma^{\mu_k} + \frac{\eps}{2} \gamma^{\mu_k})}.
\end{equation*}
\end{theorem}
\input{proofs/c7/c7_vkappa_vgamma.tex}

Thus, in goal-based MDPs, the value of the two models is bounded. The dominant terms in the bound are $\tau$ and $\gamma^{\mu_k - \tau}$, which roughly capture the variance over the number of time steps taken by the option and the length of the option's execution. When the option's execution is nearly deterministic, $\tau$ is close to $1$, and the bound collapses to $3\gamma^{\mu_k}$. Therefore, the bound is tightest when 1) the option or MDP is not very stochastic, and 2) the option executes for a long period of time. Further, the bound is quite loose; the proof of \autoref{lem:c7_tgamma_tkappa} uses Chebyshev's inequality, which does not sharply characterize concentration of measure, and the proof relies on at least one other loose approximation. Hence, in practice, it is likely that the two models will be closer; the experiments I next provide further support for the closeness of the two models in a variety of traditional MDPs.

Finally, for clarity, note that the typical convergence guarantees of the Bellman Operator are preserved under the ELM. The property follows naturally from the main result of~\citet{littman1996generalized}, since the ELM is still a well-formed transition model, and $\gamma^{\mu_k} \in [0,1)$ for any $\gamma \in [0,1)$:
\begin{remark}
The Bellman Operator using the ELM (in place of the MTM) converges to the fixed point $V^*_{\mu_k}$.
\end{remark}


To summarize, in goal-based MDPs, the ELM gives rise to value functions that do not deviate too dramatically from the MTM. The degree of deviation in these value functions scales with the stochasticity inherent to the underlying MDP and option, and inversely with the expected-length of the option. I now discuss experiments investigating the utility of the ELM in RL.

\section{Experiments} \label{sec:c7_experiments}

I next describe the findings from simulated experiments that offer further support for the hypothesis that the ELM is suitable replacement for the MTM in specific MDPs. 

\paragraph{Methodology.} Each experiment is framed as a hierarchical model-based RL problem. That is, the agent reasons with a collection of primitive actions \textit{and} options. All option models are initially unknown, and thus the learning agent estimates each options' reward and transition model from past experience. Throughout, the baseline RL algorithm used is R-Max \cite{brafman2002rmax}. As discussed in \autoref{sec:c2_rl_background}, R-Max treats unknown $(s,a)$ pairs (or in this case, $(s,o)$ pairs), as providing maximum reward until they become ``known'' by being executed some $m$ number of times. It is here that the MTM and the ELM differ in application: a transition under the MTM requires adding and updating as many parameters as needed across all $k$ possible time steps, while a transition under the ELM needs only update its running average, $\mu_k$. This is in part due to the number of parameters involved in R-Max, and it is worth noting that a tighter concentration inequality than Hoeffding's (used in R-Max to determine $m$) may yield similar performance. The policy for each option is computed by running VI (\autoref{alg:value_iteration}) over the approximate option models.

Each experiment consists of 30 independent trials and is conduced as follows. Every trial, I first sample a new reward function $R_i$ from a prespecified distribution over tasks, $D$.
Each reward function induces a goal-based problem that provides a large positive reward at goal states, a negative reward at any transition into a failure state, and zero otherwise. The presence of negative rewards deviates slightly from \autoref{asmptn:mdp_is_gb}, but not in a way that prevents straightforward application of the ELM. Indeed, expanding \autoref{lem:c7_rkappa_rgamma} to more general classes of reward functions is an important open question.

Each trial consists of 300 episodes, terminating at either a goal state, a failure state, or upon reaching a pre-determined maximum number of steps.
The hierarchies used are based on options or MAXQ task hierarchies from existing literature \cite{dietterich2000hierarchical}.
I set $m=5$ for the confidence parameter in R-Max.
Across all MDPs, $\gamma=0.99$, and all transitions are stochastic with probability $4/5$ of an action succeeding, otherwise transitioning with probability $1/5$ to a different adjacent state (as if another action had been selected).

Experiments are conducted in the following domains: the standard Four Rooms domain seen in previous chapters; Bridge Room, a grid world with a large central room containing pits (failure states) spanned by a bridge with two longer safe corridors on either side; the Taxi domain studied in the experiments from \autoref{chap:approx_state_abstr}, for which tasks are defined by hierarchical options composed of other options; and, the discrete Playroom domain~\cite{singh2005intrinsically,konidaris2018skills}, also using a hierarchy of options.

The Bridge Room domain is a grid world with a large central room contains a bridge of traversable cells that are flanked by failure states. The agent starts on one side of the bridge, with the goal state across the bridge.
Two corridors on either side of the central room offer safe but longer pathways from the start state to the bridge. The agent is only given options for moving to the doorways between rooms.
The bridge is short but crossing it is dangerous due to the stochasticity inherent in the environment.
The optimal policy, then, is to move through either corridor around the bridge room.

The discrete Playroom domain~\cite{singh2005intrinsically,konidaris2018skills} defines a complex sequential decision problem.
The agent has three effectors: 1) an eye, 2) a hand, and 3) a marker. Each marker is moved separately. The environment contains music and lights (both start off) and several objects that can be interacted with if both the hand and eye are over them. There is a switch that turns the lights on or off, a green button that turns music on, a red button that turns music off, a ball that can be thrown towards the marker, a bell that rings when hit by the ball, and a monkey that cries only when the lights are off, the music is on, and the bell rings; the goal is to make the monkey cry. Following~\citet{konidaris2018skills}, the agent plans over the interact primitive action and options for moving each effector to each object.

\begin{figure}[b!]
    \subfloat[\label{fig:c7_experiment_four_rooms}Four Rooms.]{\includegraphics[width=0.45\textwidth]{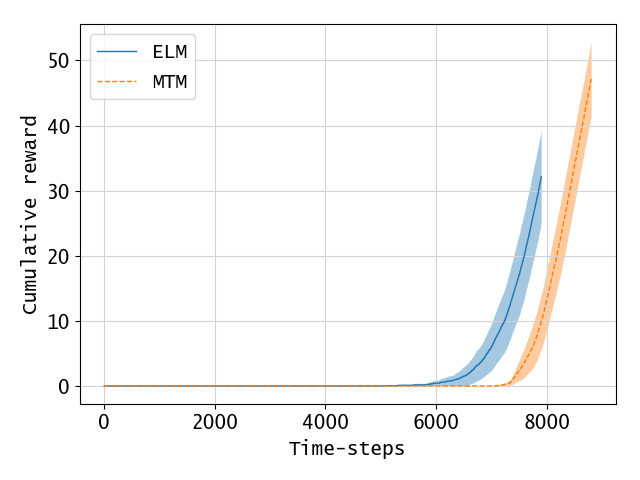}} \subfhspace
    \subfloat[\label{fig:c7_experiment_bridge_rooms_both}Bridge Room.]{\includegraphics[width=0.45\textwidth]{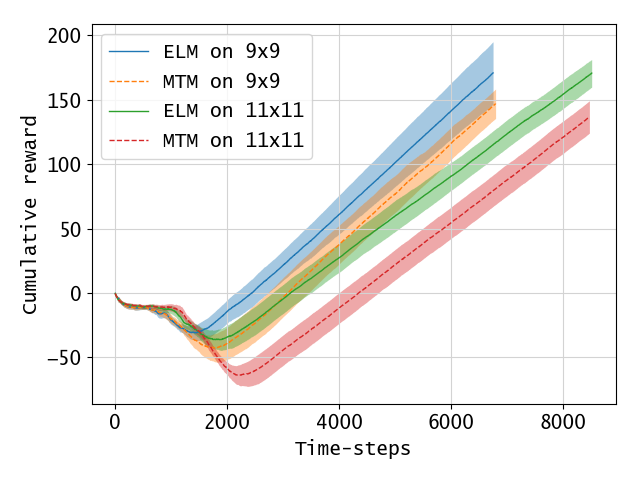}}
    
    \caption{\label{fig:c7_experimentgroup1}Learning with options in grid worlds.}
\end{figure}

\begin{figure}[t!]
    \subfloat[\label{fig:c7_experiment_taxi_1and2}Taxi, one and two passenger variants.]{\includegraphics[width=0.45\textwidth]{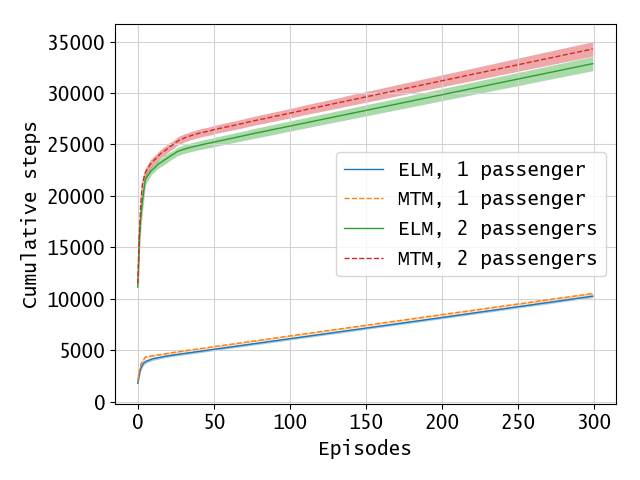}} \subfhspace
    \subfloat[\label{fig:c7_experiment_taxi_3}Taxi, three passengers.]{\includegraphics[width=0.45\textwidth]{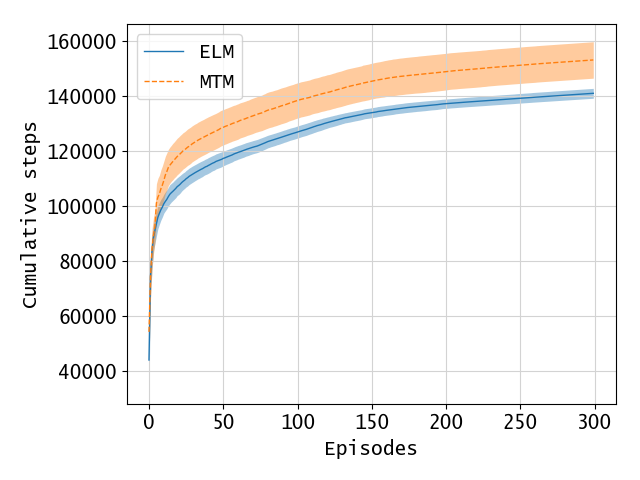}}
\caption{\label{fig:c7_experimentgroup2}Learning with options in Taxi.}
\end{figure}

\begin{figure}[t!]
    \centering
    \includegraphics[width=0.44\textwidth]{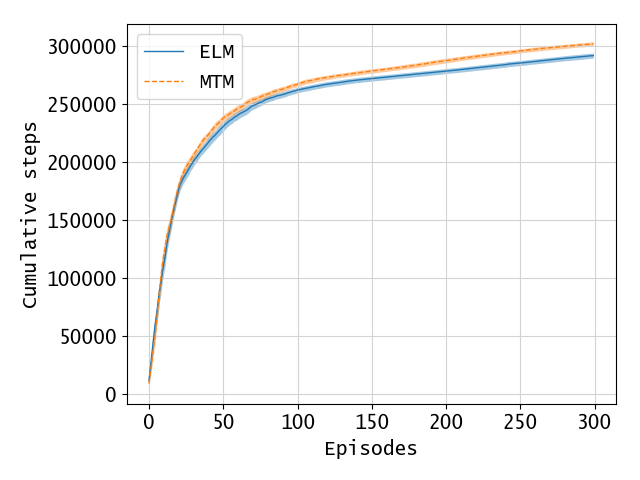}
    \caption{\label{fig:c7_experiment_playroom}Learning with options in Playroom.}
\end{figure}

\paragraph{Results.} I contrast discounted cumulative reward (performance) and time steps (sample complexity) achieved by the ELM compared to the MTM in each of the above domains.
\autoref{fig:c7_experimentgroup1}, \autoref{fig:c7_experimentgroup2}, and \autoref{fig:c7_experiment_playroom} present performance curves with 95\% confidence intervals.
Overall, the data suggest that the ELM and the MTM attain the same asymptotic performance across each domain, reflecting the fact that they both eventually converge to policies with similar values for each task.
Further, the results suggest that the ELM often requires fewer absolute samples to achieve the same quality behavior. This fact is reflected by the learning curve of the ELM terminating earlier than the MTM when plotted over time steps in \autoref{fig:c7_experiment_four_rooms}, given that both approaches are run for a consistent number of episodes. This result suggests that policies formed using ELM reach the goal earlier in learning, since the agent more quickly finds a good policy. That is, since the goal state is terminal, R-Max using the ELM tends to find the goals more quickly, and thus experiences shorter episodes than the variant using the MTM.

\begin{figure}[t!]
    \centering
    \includegraphics[width=0.45\columnwidth]{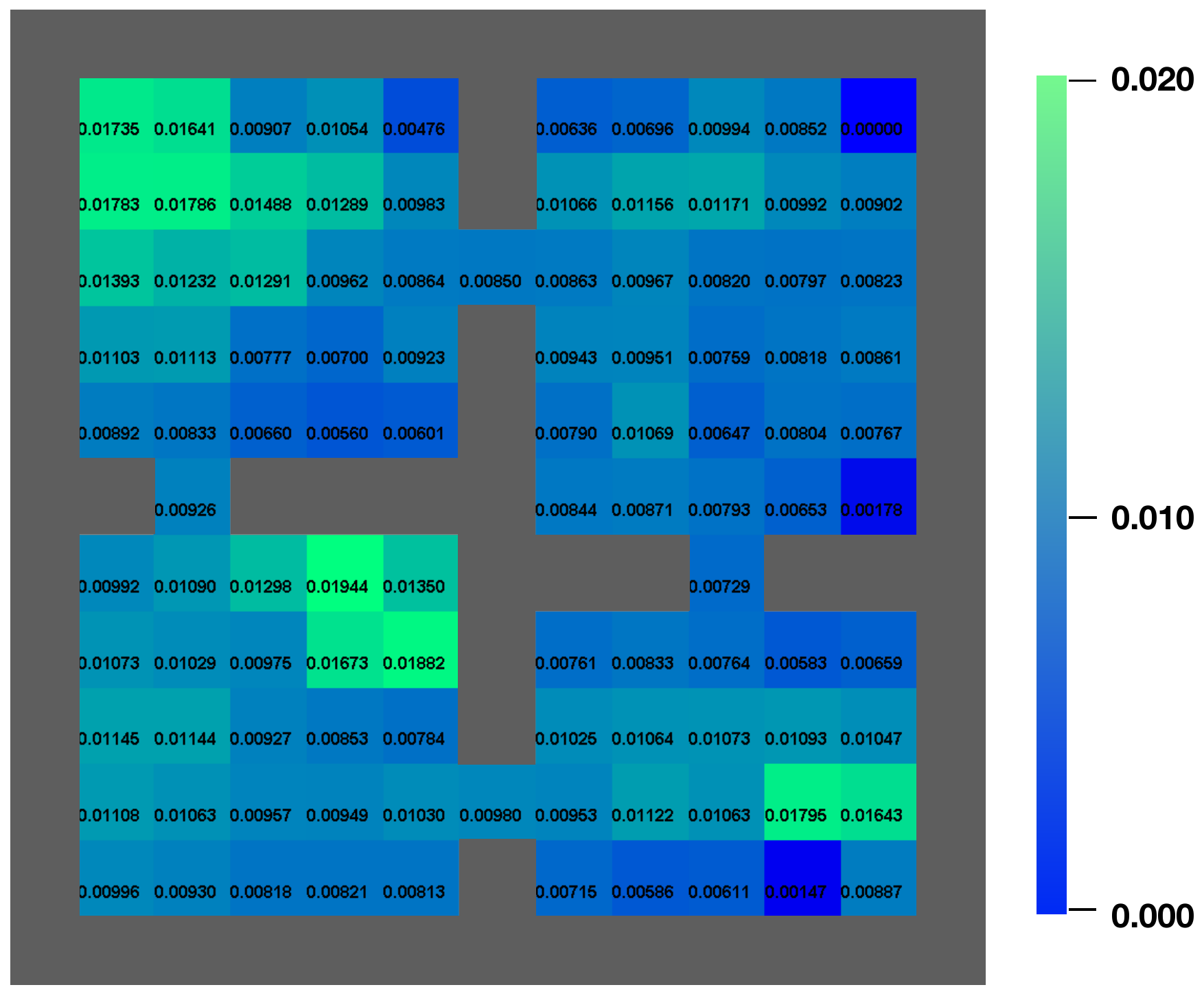}
    \caption{The difference in value between ELM and MTM in the Four Rooms task.}
    \label{fig:c7_fourrooms_vfs}
\end{figure}

\autoref{fig:c7_fourrooms_vfs} displays the difference of the value functions learned under these models in the Four Rooms MDP. In this problem, $\max_s V^*(s) = 1 / (1-0.99) = 1 / .01 = 100$. The largest gap between $V_\gamma^*$ and $V_{\mu_k}^*$ is just around $0.02$, confirming their similarity in this MDP.
Most importantly, despite the difference in the value functions, the policies generated from both are identical; both the MTM and ELM are able to support the discovery of a high-value policy. This is further evidence that the models learned under the ELM are nearly-correct, but per the learning curves in \autoref{fig:c7_experiment_four_rooms}, can often be acquired sooner.

\autoref{fig:c7_experiment_bridge_rooms_both} presents results on two variants of the Bridge Room domain.
The inflection points in the learning curves reflect the average point in learning when option models are considered known by R-Max. Observe that the ELM consistently tends to converge earlier in learning, reflecting its ability to quickly estimate the ELM, and thus more quickly make use of the available options to plan. 
In the 9$\times$9 variant of Bridge Room, however, the results are not statistically significant. For this smaller domain, the bridge is short enough that the approach using the ELM may get lucky and cross the bridge safely several times.
If this were to occur, the agent will learn to expect higher reward from the bridge option, negatively impacting the ELM's overall performance until it eventually learns the impact of occasionally falling into a pit. In other words, the high-level behavior of choosing to cross the bridge, rather than take the long route around it, tends to yield high variance outcomes over reward. Hence, the confidence interval of the ELM on 9$\times$9 in \autoref{fig:c7_experiment_bridge_rooms_both} widens as the ELM is less consistent across trials; this domain was used precisely to exhibit this potential downside of the ELM. As suggested by the analysis, domains with high stochasticity are likely to prove problematic for the ELM.

For the Taxi domain, I consider the cumulative number of samples as task complexity increases from one to three passengers. In each case, observe that both approaches are able to learn models in relatively few episodes.
In the case of one and two passengers (\autoref{fig:c7_experiment_taxi_1and2}), the approaches achieve similar performance, and the benefit of the ELM over the MTM is statistically significant but minimal.
For the largest Taxi task involving three passengers (\autoref{fig:c7_experiment_taxi_3}), the results are similar but have lower variance.

\autoref{fig:c7_experiment_playroom} presents results, again measuring the cumulative steps taken in the discrete Playroom domain.
Here, the patterns observed in the other examples recur, though the two approaches diverge later than in the Taxi experiments. This behavior is due to the immense state-action space that must be learned for the effector-moving options. That is, even as the option models are being learned, ELM's practical effect is apparent---favoring expected length leads to the generation of overall shorter plans.

Let us take stock. In this chapter, I proposed a simpler option model, the ELM, to replace the standard MTM. The analysis and experiments presented in this chapter showcase the ELM's potential for retaining a reasonable approximation of the MTM while removing the overhead in its construction. The main theorem (\autoref{thm:c7_vkappa_vgamma}) bounds the value difference of MTM and ELM in goal-based problems, and the experimental findings corroborate the claim that the ELM can be a suitable replacement for the MTM for RL, especially when the environment is not very stochastic.

%% file: proofs/c7/c7_tkappa_tgamma.tex
\begin{dproof}[Lemma~\ref{lem:c7_tgamma_tkappa}]
Let $T_\gamma(s' \mid s,o)$ denote the multi-time model, and let $T_{\mu_k}(s' \mid s,o)$ denote the expected length model.

For a fixed but arbitrary state-option-state triple $(s,o,s')$:
\begin{align}
    &|T_\gamma(s' \mid s,o) - T_{\mu_k}(s' \mid s,o)|\\
    &= |\sum_{t=1}^{\infty} \gamma^t \PR(S_t = s' \mid s, o) \beta(s')  - \gamma^{\mu_k} \sum_{t=1}^{\infty} \PR(S_t = s'\mid s, o) \beta(s') | \\
    &= |\sum_{t=1}^{\infty} \gamma^t \PR(s_t = s' \mid s, o) \beta(s') - \gamma^{\mu_k} \PR(S_t = s'\mid s, o) \beta(s') | \\
    &= |\sum_{t=1}^{\infty} (\gamma^t - \gamma^{\mu_k}) \PR(S_t = s' \mid s, o) \beta(s') | \\
    &= |\sum_{t=1}^{\infty} (\gamma^t - \gamma^{\mu_k}) \PR(S_t = s' \mid s, o) \beta(s')|
\end{align}

Note that $\PR(s_t = s' \mid s, o)\beta(s') $ is bounded above:
\begin{equation}
    \PR(S_t = s' \mid s, o)\beta(s') \leq (1 - \beta_{min})^t,
\end{equation}
since, in order to be in state $s_t$ at time $t$ we have to \textit{not} terminate in each of $s_1, \ldots s_t$. Further, we know that:
\begin{equation}
    (1-x)^t \leq e^{-xt}
\end{equation}
for any $x \in [0,1]$. Therefore:
\begin{equation}
\PR(S_t, = s' \mid s, o) \beta(s') \leq e^{-\beta_{min}t}
\end{equation}

So, rewriting:
\begin{align}
    |T_\gamma(s' \mid s,o) - T_{\mu_k}(s' \mid s,o)| &= |\sum_{t=1}^{\infty} (\gamma^t - \gamma^{\mu_k}) \PR(s_t = s', \beta(s') \mid s, o)| \\
    & \leq |\sum_{t=1}^{\infty} (\gamma^t - \gamma^{\mu_k}) e^{-\beta_{min}t}|.
\end{align}

Thus:
\begin{align}
    |T_\gamma(s' \mid s,o) - T_{\mu_k}(s' \mid s,o)| &\leq |\sum_{t=1}^{\infty} (\gamma^t - \gamma^{\mu_k}) e^{-\beta_{min}t}|
\end{align}
Let $K$ denote the random variable indicating the number of time steps taken by the option. Now, note that by Chebyshev's inequality, we know that for any $\tau > 1$:
\begin{equation}
    \PR\{|K - {\mu_k}| \geq \tau\} \leq \frac{\sigma^2}{\tau^2}.
\end{equation}
Thus, letting $\delta = \frac{\sigma^2}{\tau^2}$, we find that:
\begin{equation}
    \PR\{|K - {\mu_k}| \leq \tau\} \geq 1 - \frac{\sigma^2}{\tau^2} = 1-\delta.
\end{equation}

Thus, with probability $1-\delta$: 
\begin{align}
    |T_\gamma(s' \mid s,o) - T_{\mu_k}(s' \mid s,o)| &\leq \left|\sum_{t = {\mu_k} - \tau}^{{\mu_k} + \tau} (\gamma^t - \gamma^{\mu_k}) e^{-\beta_{min}t}\right| \\
    &\leq \sum_{t = {\mu_k} - \tau}^{{\mu_k} + \tau} \left|(\gamma^{t} - \gamma^{\mu_k})\right| e^{-\beta_{min}t} \\
    &\leq \sum_{t = {\mu_k} - \tau}^{{\mu_k} + \tau} |\gamma^{{\mu_k}-\tau}| e^{-\beta_{min}t} \\
    &= \gamma^{{\mu_k} - \tau} \sum_{t = {\mu_k} - \tau}^{{\mu_k} + \tau} e^{-\beta_{min}t} \\
    &\leq \gamma^{{\mu_k} - \tau} (2\tau + 1) e^{-\beta_{min}}
\end{align}

Therefore, for $\delta = \frac{\sigma^2}{\tau^2}$:
\begin{equation}
    \PR\{|T_\gamma(s' \mid s,o) - T_{\mu_k}(s' \mid s,o)| \leq \gamma^{{\mu_k} - \tau} (2\tau + 1) e^{-\beta_{min}}\} \geq 1-\delta.\qedhere
\end{equation}
\end{dproof}

%% file: proofs/c7/c7_rkappa_rgamma.tex
\begin{dproof}[Lemma~\ref{lem:c7_rkappa_rgamma}]
In the goal-based MDP considered all rewards are either 0 or 1 (\autoref{asmptn:mdp_is_gb}). Thus, if a given option \textit{cannot} reach the goal state, the two reward models are identical, since all accumulated rewards by the option will be zero:
\begin{equation}
    |R_\gamma(s,o) - R_{\mu_k}(s,o)| = 0.
\end{equation}
Conversely, if the option \textit{can} reach the goal state, then the expected reward of the option is just the probability, under the relevant transition model ($T_\gamma$ or $T_{\mu_k}$) of reaching the goal. Therefore, more generally:
\begin{align}
    R_\gamma(s,o) &= T_\gamma(s_g \mid s,o), \\
    R_{\mu_k}(s,o) &= T_{\mu_k}(s_g, s,o).
\end{align}
Consequently, by definition:
\begin{equation}
    |R_\gamma(s,o) - R_{\mu_k}(s,o)| = |T_\gamma(s_g \mid s,o) - T_{\mu_k}(s_g \mid s,o)| 
\end{equation}
Thus, we conclude by applying \autoref{lem:c7_tgamma_tkappa}, for $\delta  = \frac{\sigma^2}{\tau^2}$, for any $s$ and $o$:
\begin{equation}
    \PR\left\{|R_\gamma(s,o) - R_{\mu_k}(s,o)| \leq \gamma^{{\mu_k} - \tau} (2\tau + 1) e^{\beta_{min}}\right\} \geq 1 - \delta. \qedhere
\end{equation}
\end{dproof}

%% file: proofs/c7/c7_vkappa_vgamma.tex
\begin{dproof}[Theorem~\ref{thm:c7_vkappa_vgamma}]
Let
\begin{equation}
    \eps = \gamma^{{\mu_k} - \tau} (2\tau + 1) e^{-\beta_{min}},
\end{equation}
and again let $\delta = \frac{\sigma^2}{\tau^2}$. By \autoref{lem:c7_tgamma_tkappa} and \autoref{lem:c7_rkappa_rgamma}, we know that the reward and transition models are bounded, each with probability $1-\delta$:
\begin{align}
    |R_\gamma(s,o) - R_{\mu_k}(s,o)| &\leq \eps, \\
    |T_\gamma(s' \mid s,o) - T_{\mu_k}(s' \mid s,o)| &\leq \eps.
\end{align}
Then, let
\begin{equation}
    V_{\gamma,\eps}^{\pi_\gamma}(s) = R_\gamma(s,o) + \gamma^{\mu_k} \sum_{s' \in \mc{S}} \left(\PR(s' \mid s, o) + \eps\right) V_{\gamma,\eps}^{\mu_k}(s').
\end{equation}
Note that, by the transition model bound above:
\begin{equation}
    ||V_{\gamma}^{\pi_\gamma} - V_{\mu_k}^{\pi_\gamma}||_\infty \leq ||V_{\gamma,\eps}^{\pi_\gamma} - V_{\mu_k}^{\pi_\gamma}||_\infty
    \label{eq:c7_sup_vgam}
\end{equation}

Then, by Lemma 4 by~\citet{strehl2005theoretical}, we upper bound the right hand side of Equation~\ref{eq:c7_sup_vgam} with probability $1-\delta$, for any option $o$, any policy $\pi$, for any state $s$:
\begin{equation}
    |Q_{\gamma,\eps}^\pi(s,o) - Q_{\mu_k}^\pi(s,o)| \leq \frac{(1-\gamma^{\mu_k})\eps + \gamma^{\mu_k} \frac{\eps}{2} \textsc{RMax}}{(1-\gamma^{\mu_k})(1-\gamma^{\mu_k} +\frac{\eps}{2} \gamma^{\mu_k})}.
    \label{eq:c7_sim_lem}
\end{equation}
By combining \autoref{eq:c7_sup_vgam} and \autoref{eq:c7_sim_lem}, we conclude the proof. \qedhere
\end{dproof}

%% file: chapters/c8_aa_options_for_exploration.tex
\begin{center}
\begin{minipage}{0.8\textwidth}
\textit{This chapter is based on ``Discovering Options for Exploration by Minimizing Cover Time" \cite{jinnai2019opt_explore} led by Yuu Jinnai, also in collaboration with Jee Won Park and George Konidaris.}
\end{minipage}
\end{center}
\vspace{2mm}

\newcommand{\algname}{Covering options}

One of the central challenges of RL is the \textit{explore-exploit} dilemma, discussed briefly in \autoref{chap:background}. At a high level, the dilemma highlights the fact that RL agents acting in an unknown world must simultaneously discover \textit{new} things about their surroundings while using what is already known to make good decisions. Such a trade off is especially challenging in environments where the vast majority of rewards are zero with only occasional signal that differentiates good and bad decisions.

Let us return once again to the hiker camping in the forest from \autoref{chap:introduction}. The last few nights, they unfortunately developed a bit of back pain from sleeping inside the tent in a thin sleeping bag. However, the hiker also brought a hammock and a (now broken, sadly) inflatable air mattress. The sun is beginning to set and our hiker is deliberating over possible sleeping configurations. They are presented with a variety of choices: they could move their tent to a new location with softer or flatter ground, rotate their sleeping bag, find a suitable spot to hang the hammock, or try to fix the air mattress. Critically, the hiker does not know in advance which of these activities will help them reliably get a good night of rest and let their back recover. Testing each activity requires time and energy as well, as it is not trivial to hang a hammock or fix an air mattress. Additionally, there is uncertainty over the desirability of each outcome, and the only way to get true signal about the impact of each decision is to actually experiment with a particular sleeping configuration for a period of time. How is the hiker to proceed?

This is precisely an instance of the explore-exploit dilemma: the hiker could exploit what they know now and continue sleeping in their current spot where they have consistently been safe, warm, and comfortable except for the recent back pain. Or, they can explore new activities, facing the uncertainty inherent in each of the new sleeping modes. Depending on the hiker's willingness to experiment, or the potential severity of the outcomes, different strategies might make sense.

Implicit in the hiker's situation is that they are already aware of the other relevant sleeping strategies even if they have not actually tried them before. This is a key source of the utility of action abstraction in facing down the explore-exploit dilemma. If the agent did not have the capacity to consider the ``set-up-and-sleep-in-hammock" option, but rather faced down the exponential policy space formed by every permutation of primitive actions, then the problem is effectively hopeless (should I sleep in my sleeping bag, or execute some choice of actions $a_i, a_j, a_k, a_\ell, \ldots, \forall_{i, j, k, \ell, \ldots}$?).

In this way, options have the potential to dramatically alter the exploration problem. Thus, options that are able to accelerate exploration are directly in line with the second desiderata (efficient decision making). Long-horizon actions can enable agents to focus on a particular objective conditioned on a consistent intent, giving rise to directed exploration strategies that prune away irrelevant action sequences. 

\begin{figure}
    \centering
    \includegraphics[width=0.6\textwidth]{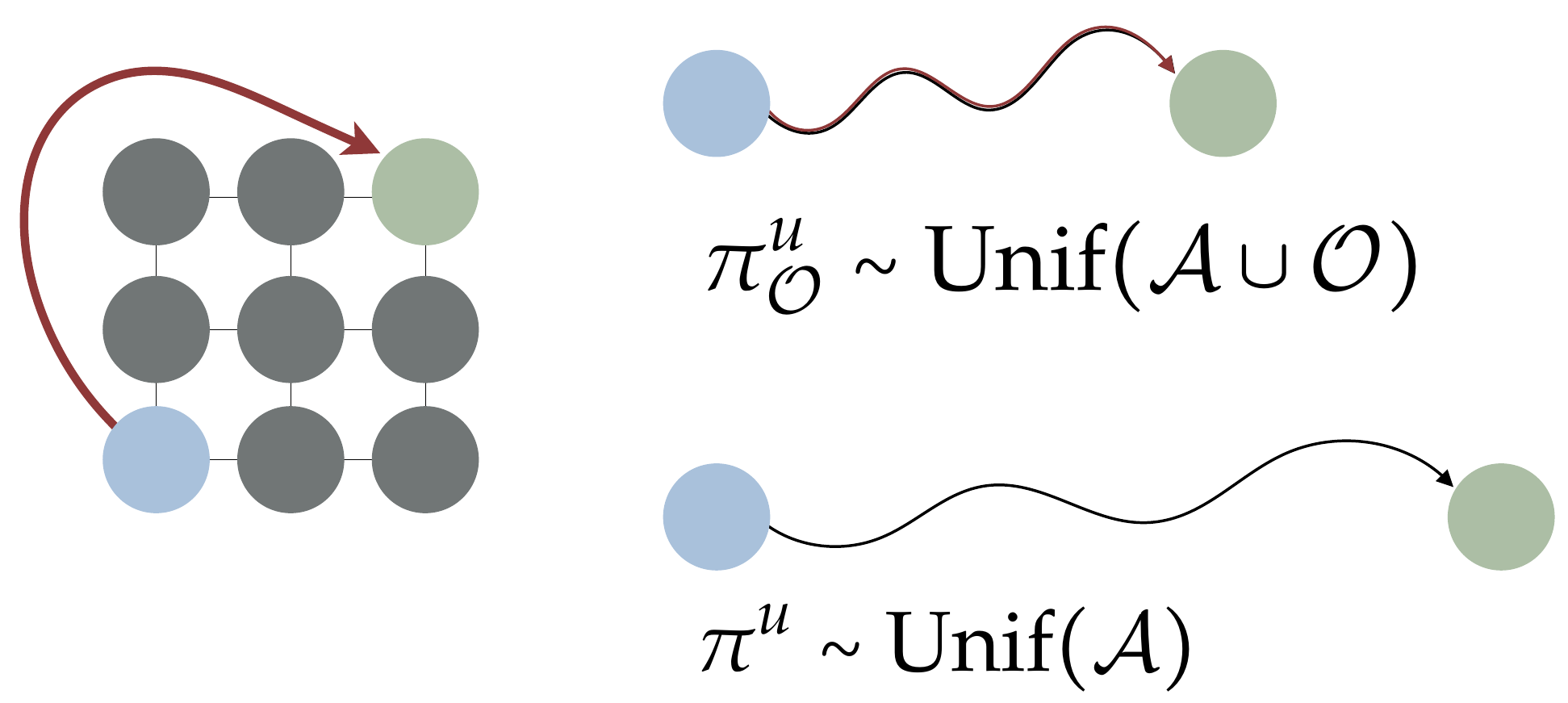}
    \caption{An example illustrating the main idea behind covering options: the expected length of the random walk between relevant states can be reduced by well chosen options.}
    \label{fig:c8_cover_opt_ex}
\end{figure}

In light of these intuitions, this chapter studies the problem of discovering options that can aid in exploration. I describe a new polynomial time algorithm first introduced by \citet{jinnai2019opt_explore} that improves exploration in finite, sparse reward MDPs by constructing \textit{covering options} that minimize the expected number of steps to reach a (previously unknown) rewarding state without task-dependent reward information. An intuitive illustration of this idea is pictured in \autoref{fig:c8_cover_opt_ex}. I suggest that this algorithm is in line with both the first and second abstraction desiderata, since the options do not require unreasonable resource budgets, and are aimed at improving the sample efficiency of RL. Concretely, the proposed algorithm finds a set of options that reduce the expected \textit{cover time} \cite{broder1989bounds} of a random walk over the combined action space, $\mc{A} \cup \mc{O}$.

Computationally, this problem is equivalent to finding a set of edges to add to the MDP's transition graph that minimize the expected cover time, which is known to be a hard combinatorial optimization problem \cite{braess1968paradoxon,braess2005paradox}. In light of this difficulty, the algorithm instead seeks to minimize an upper bound of the expected cover time given as a function of the algebraic connectivity of the graph Laplacian \cite{fiedler1973algebraic,broder1989bounds,chung1996spectral} using the heuristic method by \citet{ghosh2006growing}. I study the practical utility of the proposed option discovery algorithm in six finite goal-based MDPs, finding that covering options can indeed accelerate exploration.

The proposed option discovery algorithm makes the assumption that the behavior of an RL agent at the beginning of its learning process may be modeled as a random walk induced by a fixed stationary policy. Without any other knowledge about a given problem, this is sensible: a reasonable default exploration policy is to execute actions uniformly at random (at least until more data is collected). Additionally, this random walk will help establish a simple form of \textit{worst-case} analysis. Surely other more nuanced exploration methods will be faster than a fixed policy. Under this assumption, we build toward an upper bound on the expected cover time of a random walk in the MDP.

As in \autoref{chap:options_for_planning}, this chapter studies \textit{point options} that initiate and terminate in exactly one (possibly different) state each. For more on point options, see \autoref{chap:options_for_planning}, and specifically \autoref{def:c6_point_option}. Adding a point option to an MDP corresponds to inserting a single edge into the graph induced by the MDP's transition dynamics. Throughout the chapter, I will refer to the state $s$ with $\beta_o(s) = 1$ as the subgoal state.

\section{Cover Time}
\label{sec:cover-time}

In this section, I motivate the use of cover time as a mechanism for studying exploration in RL. The cover time is the time required for a random walk to visit all the vertices in a graph \cite{broder1989bounds}. To be precise in grounding this concept, let us first introduce several additional definitions.

First, assume we are given a discrete Markov chain $\{X_t\}$ with state space $V$ denoting the vertices of graph $G$. The hitting time $H_{ij}$, where $i,j \in V$, is defined as
\begin{equation}
    H_{ij} := \inf\big\{t \in \mathbb{N}: X_t = j\ \text{and}\ X_0 = i\big\}.
\end{equation}
In other words, $H_{ij}$ is the greatest lower bound on the number of time step $t$ required to reach state $j$ after starting at state $i$. The cover time starting from state $i$ is then defined as
\begin{equation}
   C_i := \max_{j \in V } H_{ij},
\end{equation}
and the expectation of cover time, $\bE[C(G)]$, is the expected cover time of trajectories induced by the random walk, maximized over the start state \cite{broder1989bounds}. Thus, the expected cover time bounds how likely a random walk will lead to a rewarding state, formalized in the following result.

\begin{theorem}
\label{thm:c8_value_cover_time}
    Given an MDP $M$ that encodes a goal-based MDP with goal $g$, where a non-positive reward $r_c < 0$ is given for entering non-goal states and $\gamma = 1$. Let $W$ be a random walk transition matrix, $W(s, s') = \sum_{a \in A} \pi(s) T(s' \mid s, a)$ then:
    \begin{equation}
        \forall_g: V_g^\pi(s) \geq r_c \bE[C(G)].
    \end{equation}
    where $\bE[C(G)]$ is the expected cover time of a transition matrix $W$.
\end{theorem}

The proof was first introduced by \citet{jinnai2019opt_explore}---see Theorem 1.

Intuitively, this result suggests that a smaller expected cover time may translate to more efficient exploration. More formally, let $P$ be a random walk induced by a fixed policy $\pi$ in an MDP with start state distribution $\rhoz$. \citet{broder1989bounds} prove that the expected cover time $\bE[C(G)]$ of $P$ is bounded by a function of the second largest eigenvalue of the random walk matrix $\lambda_{k-1}(P)$ as follows.
\begin{equation}
\label{eq:bound}
    \bE[C(G)] \leq \frac{n^2 \ln n}{1 - \lambda_{k-1}(P)} (1 + o(1)),
\end{equation}
where $n = |V|$ and $k$ is the number of eigenvalues. The normalized graph Laplacian of an unweighted undirected graph is defined as:
\begin{equation}
    \mathcal{L} := I - T^{-1/2} A T^{-1/2},
\end{equation}
where $I$ is an identity matrix \cite{chung1996spectral}.
The random walk matrix can be written in terms of the Laplacian:
\begin{equation}
    P = T^{-1} A = T^{-1/2} (I - \mathcal{L}) T^{1/2}.
\end{equation}

Note that since $P$ and $I - \mathcal{L}$ are \textit{similar} matrices, they have the same eigenvalues and eigenvectors. Therefore, $\lambda_{k-1}(P) = 1 - \lambda_2(\mathcal{L})$, where $\lambda_2(\mathcal{L})$ is the second smallest eigenvalue of $\mathcal{L}$. By \autoref{eq:bound},
\begin{equation}
\label{eq:boundl}
    \bE[C(G)] \leq \frac{n^2 \ln n}{\lambda_{2}} (1 + o(1)).
\end{equation}
Hence, the larger the $\lambda_{2}(\mathcal{L})$ is, the smaller the upper bound on the expected cover time. %

The second smallest eigenvalue of $\mathcal{L}$ is known as the \textit{algebraic connectivity} of the graph, with its corresponding eigenvector referred to as the \textit{Fiedler vector} \cite{fiedler1973algebraic}.
There are several operations that can be applied to the graph to increase the algebraic connectivity. For instance, if we add well chosen nodes to the graph, the connectivity changes, but not necessarily in a way that improves the cover time. Alternatively, we might reroute edges in the graph. However, in general it is undesirable to reroute edges as this changes the space of representable policies and may destroy an agent's ability to represent a high value policy. Finally, we might add entirely new edges to the graph---in RL, this is effect of adding options to the primitive action space. Therefore, adding edges is a reliable way to reduce the cover time without potentially sacrificing optimality. Naturally, if more information about the problem is available (such as which primitive actions may be pruned on a per-state basis), other operations may be considered as well to lower cover time.

\subsection{Cover Time Experiments}

\begin{figure}[t!]
    \centering
    \subfloat[(Connectivity) $\sim$ (cover time)]{\label{fig:c8_connectivity-covertime} \includegraphics[width=0.45\textwidth]{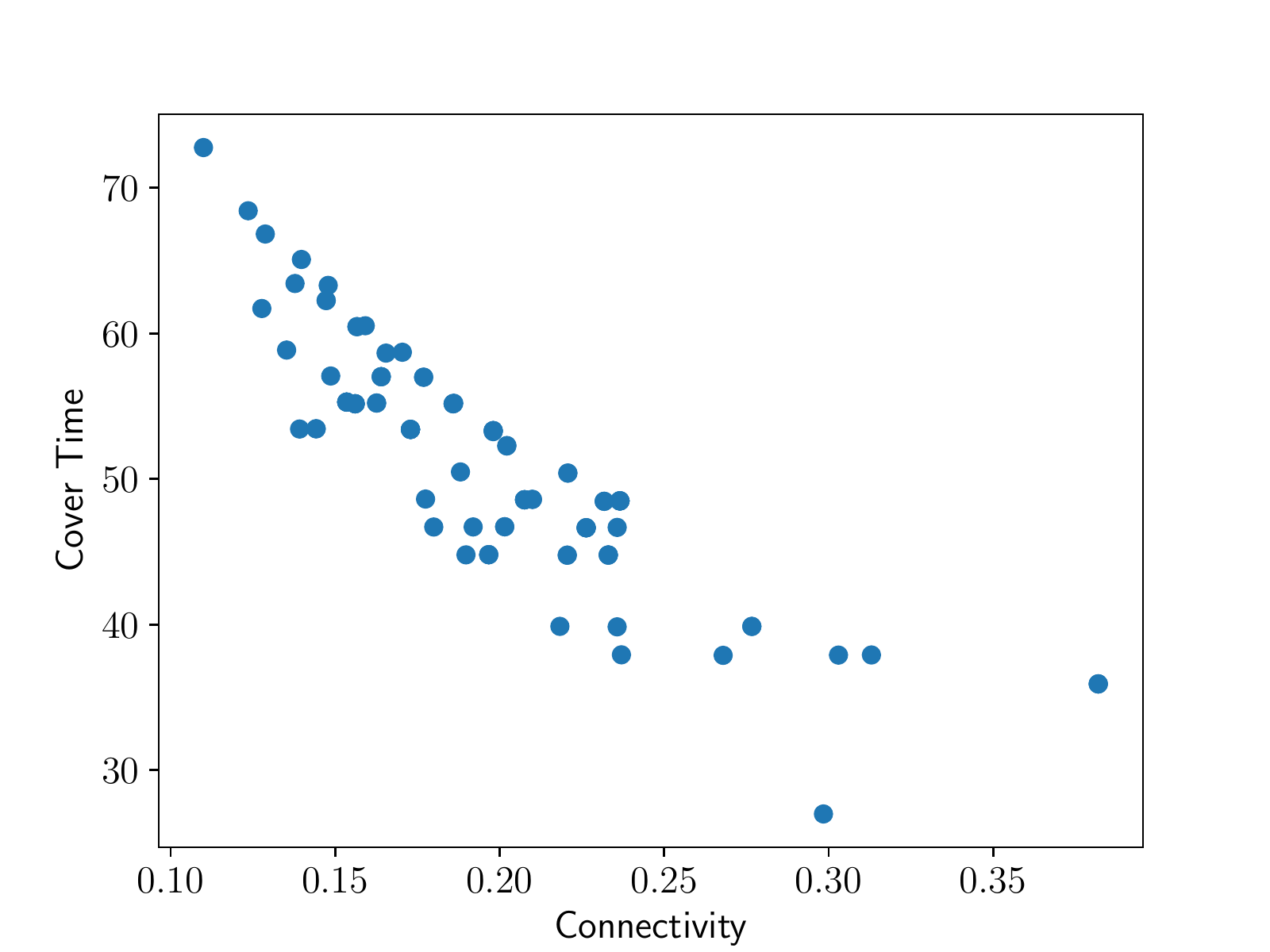}} \subfhspace
    \subfloat[(cover time) $\sim$ (cost of the policy)]{\label{fig:c8_covertime-hits} \includegraphics[width=0.45\textwidth]{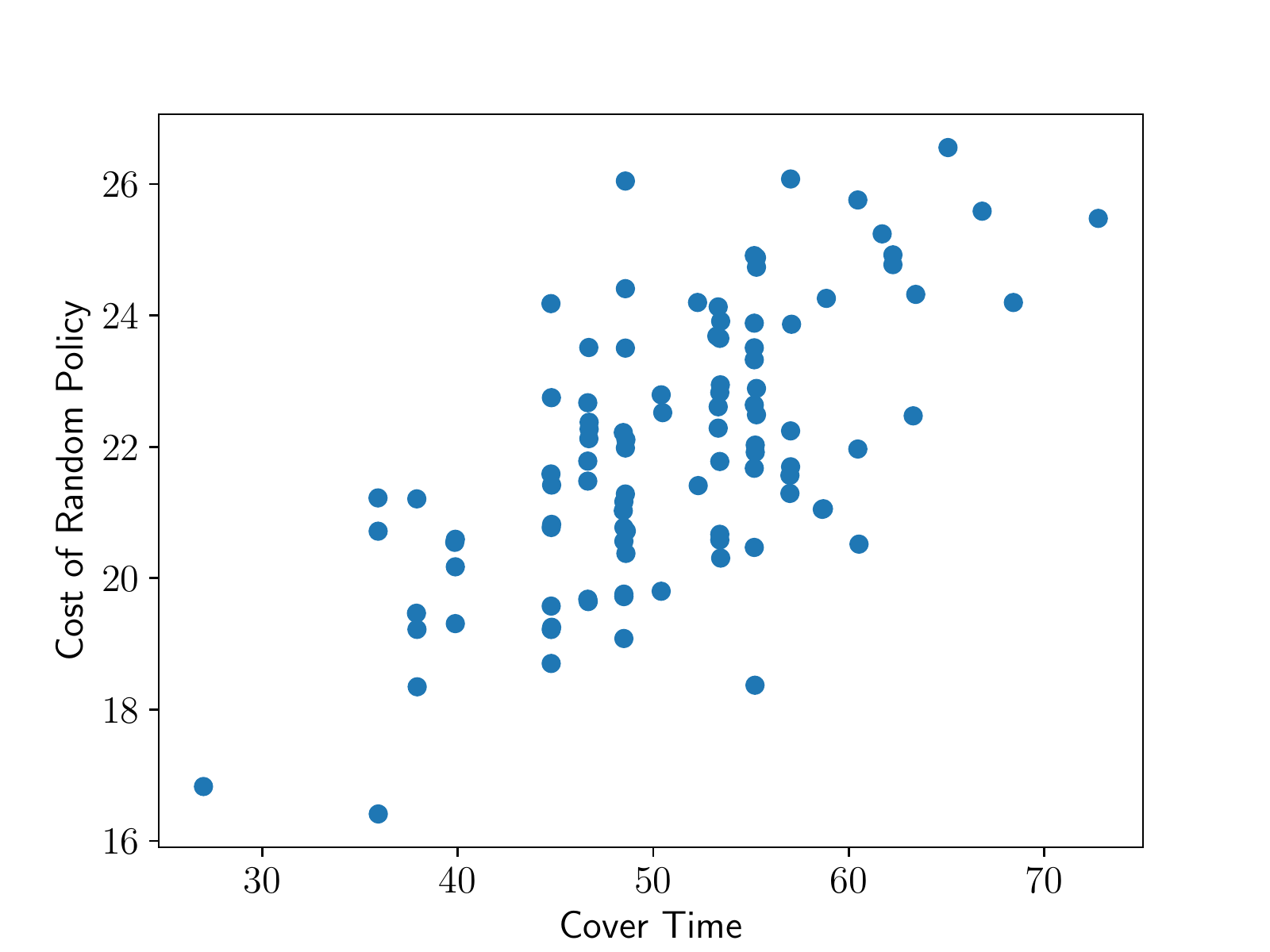}}
    \caption{The relationship between (a) algebraic connectivity ($\lambda_2$) and cover time on randomly generated graphs, and (b) the cover time of a random walk vs. the cost of random policy.}
    
    \label{fig:c8_covertime} 
\end{figure}

In the previous section, I illustrated how the algebraic connectivity of an MDP's transition graph relates to the expected cover time. I now describe a simple experiment that further examines this relationship.

The experiment consists of two steps. First, generate 100 random connected graphs with 10 nodes with edge density fixed to $0.3$.\footnote{The graph generation process proceeds as follows. First, start with a single node. Pick one node from the existing graph and add an edge to connect to a new node. Follow this procedure for the number of nodes $n-1$, generating a random tree of size $n$. Then, pick an edge uniformly randomly from $E^c$ until the edge density reaches the threshold.} Second, approximate the expected cover time of a random walk on a random graph by sampling 10,000 trajectories induced by the random walk and computing their average cover time.

\autoref{fig:c8_connectivity-covertime} shows the relationship of the algebraic connectivity and the expected cover time of the random walk induced by the uniform random policy. The takeaway from these results is that the random walk tends to have smaller expected cover time when the underlying state-transition graph has larger algebraic connectivity. Conversely, \autoref{fig:c8_covertime-hits} shows the expected cost of a random policy from the initial state to reach the goal state. Here, observe that the cost of a random policy is correlated to the cover time, as expected.

\section{Covering Options}
\label{sec:c6_covering_options}

As discussed, computing the precise set of edges that minimize expected cover time in a graph is thought to be NP-Hard \cite{aldous1995reversible}. Even a good solution is hard to find due to Braess's paradox \cite{braess1968paradoxon,braess2005paradox} which states that the expected cover time does not monotonically decrease as edges are added to the graph.

Therefore, the expected cover time is often minimized indirectly by maximizing algebraic connectivity \cite{fiedler1973algebraic,chung1996spectral}.
In particular, the expected cover time is upper bounded by quantity related to the algebraic connectivity (\autoref{eq:boundl}), and by maximizing this quantity, the bound can be minimized \cite{broder1989bounds}.
Choosing the set of edges that maximize the algebraic connectivity of a given graph is known to be NP-hard \cite{mosk2008maximum}. However, \citet{ghosh2006growing} developed an approximation procedure for this problem that proceeds as follows.

\begin{enumerate}
    \item Compute the second smallest eigenvalue and its corresponding eigenvector (the Fiedler vector) of the Laplacian of the state transition graph $G$.
    
    \item Let $v_i$ and $v_j$ be the states with the largest and smallest values in the Fiedler vector respectively. Generate two point options, the first with $\mc{I}_o=\{v_i\}$, $\beta_{o_1}=\indic\{v_j\}$, and the second with $\mc{I}_{o_1}=\{v_j\}$ and $\beta_{o_2}=\indic\{v_i\}$. Each option policy is the optimal path from the initial state to the termination state.
    
    \item Set $G \leftarrow G \cup \{(v_i, v_j)\}$ and repeat the process until the number of options reaches $k$.
\end{enumerate}

\begin{figure}
    \centering
    \subfloat[Fiedler vector]{\includegraphics[width=0.45\columnwidth]{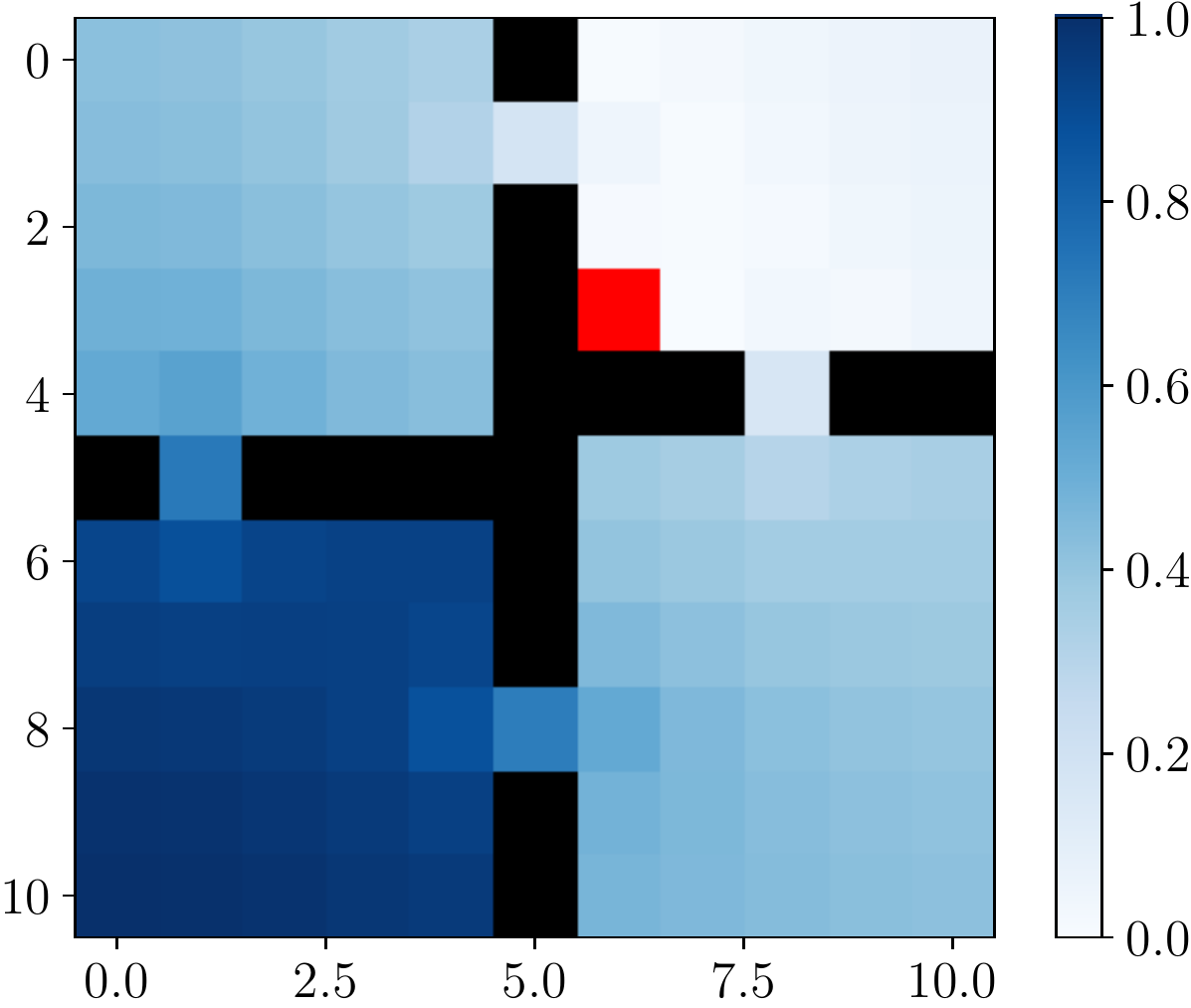}} \subfhspace
    \subfloat[Euclidean distance]{\includegraphics[width=0.45\columnwidth]{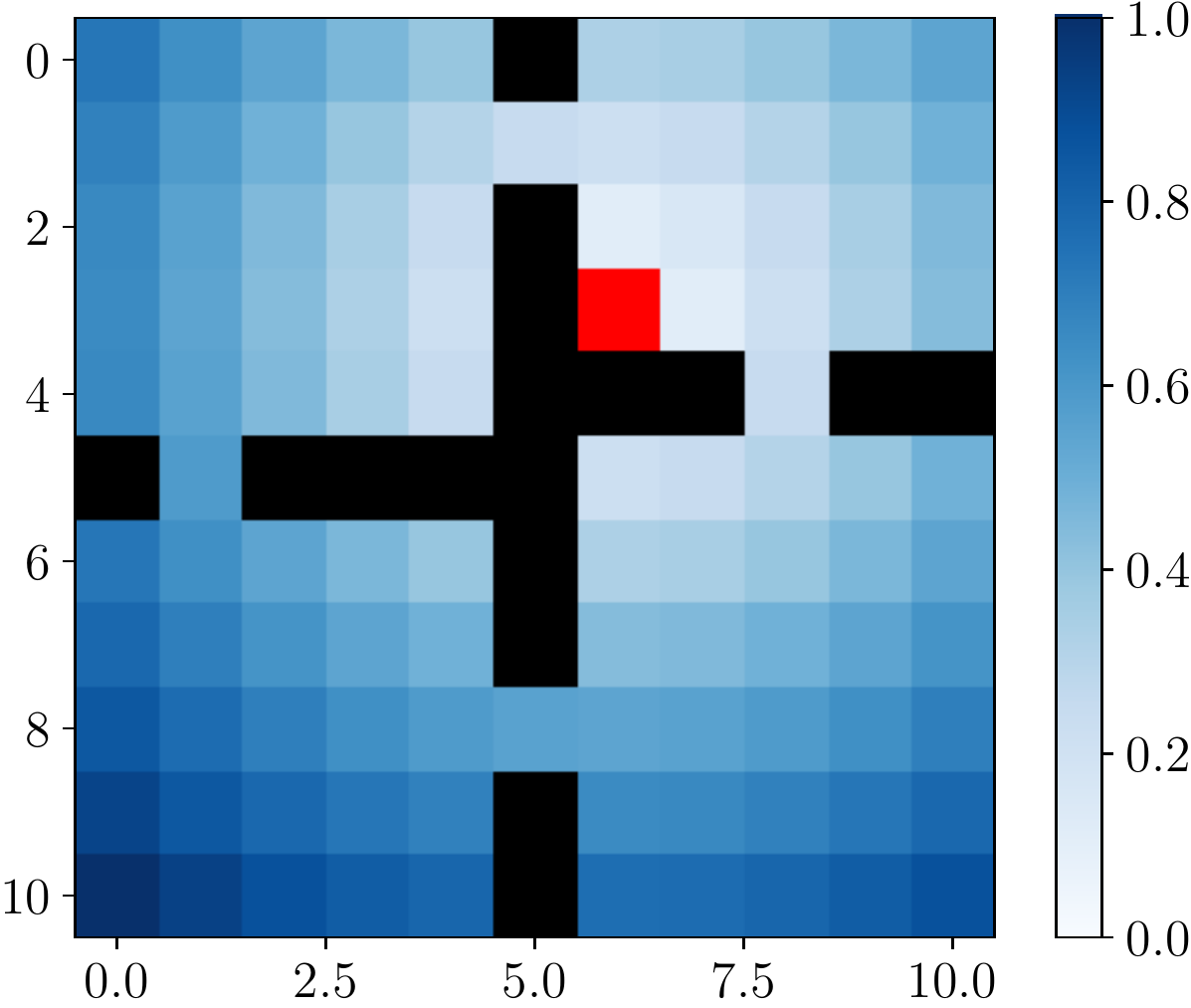}}
    \caption{The distance between the red state and all other states, measured via Fiedler vector (left) and Euclidean distance (right).}
    \label{fig:eigenvector}
\end{figure}

Intuitively, the algebraic connectivity represents how well the graph is connected. The Fiedler vector is an embedding of a graph to a line (that is, a real number) where nodes connected by an edge tend to be placed close to one another (for example, see \autoref{fig:eigenvector}). A pair of nodes with the maximum and minimum value in the Fiedler vector are the most distant nodes in this embedding space. Thus, our proposed method greedily connects the two most distant nodes in the embedding, thereby greedily maximizing the algebraic connectivity to a first order approximation \cite{ghosh2006growing}.

The result is a set of options that is guaranteed to minimize the lower bound on the expected cover time, as summarized by the following result.

\begin{theorem}
\label{thm:c8_option_cover_time}
    Assume that a random walk induced by a policy $\pi$ is a uniform random walk:
    \begin{equation}
    \label{eq:random-walk}
        W(u, v) := \begin{cases}
            1 / d_u  & \text{if $u$ and $v$ are adjacent,}\\
            0 & \text{otherwise},
            \end{cases}
    \end{equation}
    where $d_u$ is the degree of the node $u$.
    Adding two options by the algorithm improves the upper bound of the cover time if the multiplicity of the second smallest eigenvalue is one: 
    \begin{equation}
    \bE[C(G')] \leq \frac{n^2 \ln n}{\lambda_{2}(\mathcal{L}) + F} (1 + o(1)),
    \end{equation}
    where $\bE[C(G')]$ is the expected cover time of the augmented graph, $F = \frac{(v_i - v_j)^2}{6 / (\lambda_3 - \lambda_2) + 3 / 2}$, and $v_i, v_j$ are the maximum and minimum values of the Fiedler vector.
\end{theorem}

The proof was first introduced by \citet{jinnai2019opt_explore}---see Theorem 2.


The state transition graph $G$ may either be given to the agent as input or learned during interaction with the MDP. Throughout, we assume that the graph is undirected and strongly connected, so every state is reachable from every other state. As in the work by \citet{machado2017laplacian}, the proposed algorithm can be generalized to MDP's with rich state spaces using an incidence matrix instead of an adjacency matrix; such a generalization was recently carried out by \citet{jinnai2020exploration} to great effect.

To summarize the approach, the expected cover time of a random walk is a useful measure for approximating the exploration difficulty of goal-based MDPs. Under this approximation, we design an option discovery algorithm that decreases the cover time by choosing options that connect states that are most distant according to the Fiedler vector. I now discuss findings from an empirical study first carried out by \citet{jinnai2019opt_explore} that examines the utility of covering options.

\section{Experiments}

We conduct experiments with six finite MDPs, many of which should be familiar from previous chapters. These domains include the 9$\times$9 grid from \autoref{chap:options_for_planning}, the standard Four Rooms grid world, Parr's maze \cite{parr1998reinforcement}, Taxi from \autoref{chap:approx_state_abstr}, the classic disc puzzle Towers of Hanoi, and a discrete driving domain called Race Track. Towers of Hanoi is a classic game consisting of three pegs that can hold different size discs (sorted in decreasing order of disc diameter) on any of the pegs. The goal is to move all discs from a single initial peg to a goal peg while keeping the constraint that each smaller disc is placed above a larger one (or is the only disc on the peg). In Race Track, the agent must reach the finish line by driving a car without hitting the track's boundary. The agent may change the car's vertical and lateral velocity by +1, -1, or 0 in each time step. If the car hits the track boundary, it is moved back to the starting position.

In each MDP, we compare the performance of covering options, eigenoptions \cite{machado2017laplacian}, and betweenness options \cite{csimcsek2009skill}. These methods were chosen for similar reasons discussed in \autoref{chap:options_for_planning}: they are closely related option discovery methods that do not require reward information. More detail about eigenoptions and betweenness options is provided in \autoref{sec:action_abstr_survey}.

To make the comparison more direct, we experiment with a point option variant of the eigenoption method, though notably this was not the intend structure for eigenoptions. For the $k$-eigenvectors that correspond to the smallest $k$ eigenvalues, we generate a point option from a state with the highest (or lowest) value to a state the lowest (or highest) value in the eigenvector.
The point option constructed in this way minimizes the eigenvalue of each corresponding eigenvector.

\begin{table}[t!]
    \centering
    \begin{tabular}{lccccc}
        \toprule
        &\multicolumn{2}{c}{Four Rooms}&&\multicolumn{2}{c}{9$\times$9 grid} \\
        \cmidrule(lr){2-3} \cmidrule(lr){5-6}
        & $\lambda_2$ & $\bE[C(G')]$ && $\lambda_2$ & $\bE[C(G')]$ \\
        \midrule
        \algname{} & \textbf{0.065} & \textbf{672.0}&& \textbf{0.24} & \textbf{258.6} \\
        Eigenoptions    & 0.054 & 695.9&& 0.19 & 261.5 \\
        No options      & 0.023 & 1094.8&& 0.12 & 460.5 \\
        \bottomrule
    \end{tabular}
    \caption{Comparison of the algebraic connectivity and the expected cover time of covering options and eigenoptions.
    }
    \label{tab:c8_connectivity}
\end{table}

First, we present a simple quantitative evaluation measuring the impact covering options and eigenoptions have on the algebraic connectivity ($\lambda_2$) and expected cover time ($\bE[C(G')]$) in each of the Four Rooms and 9$\times$9 grid worlds. Results are presented in \autoref{tab:c8_connectivity}. In both domains the covering options achieve larger algebraic connectivity and smaller expected cover time than eigenoptions as desired, providing initial confirmation that covering options perform as expected.

\subsection{Visuals}

\begin{figure}[b!]
    \centering
    \subfloat[\algname{}]{\includegraphics[width=0.22\textwidth]{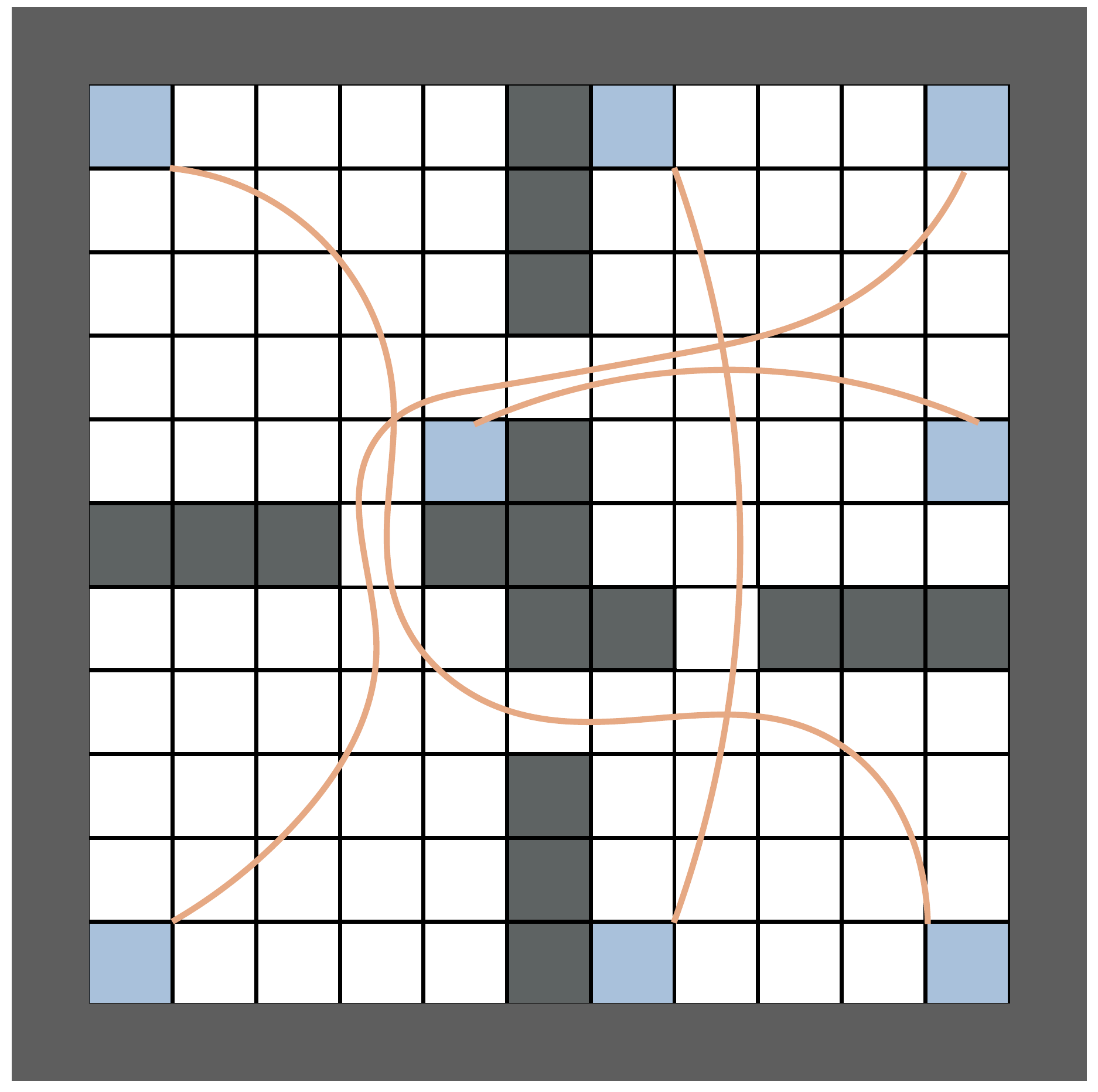}} \hspace{2mm}
    \subfloat[\algname{}]{\includegraphics[width=0.22\textwidth]{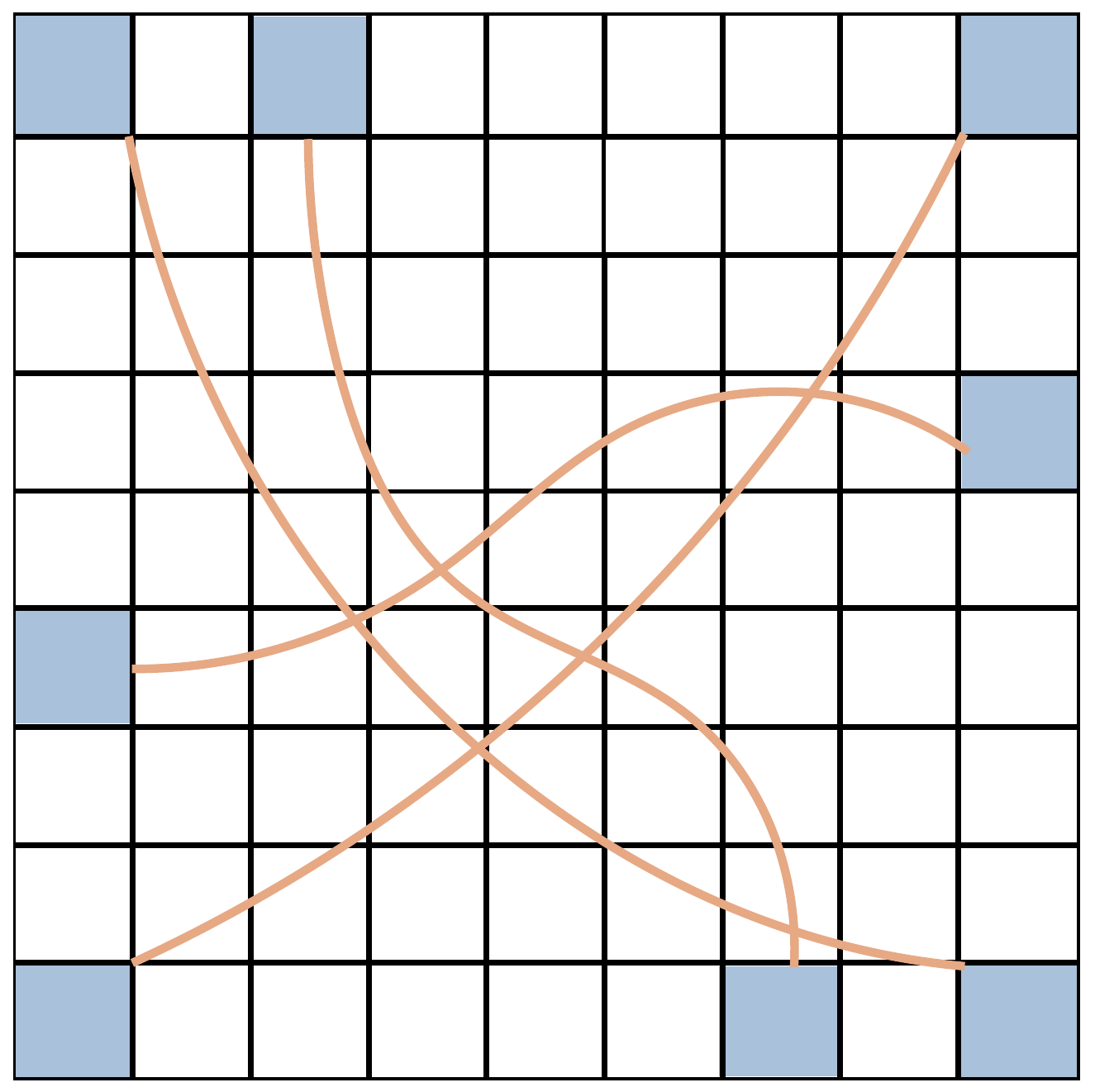}} \hspace{2mm}
    \subfloat[Eigenoptions]{\includegraphics[width=0.22\textwidth]{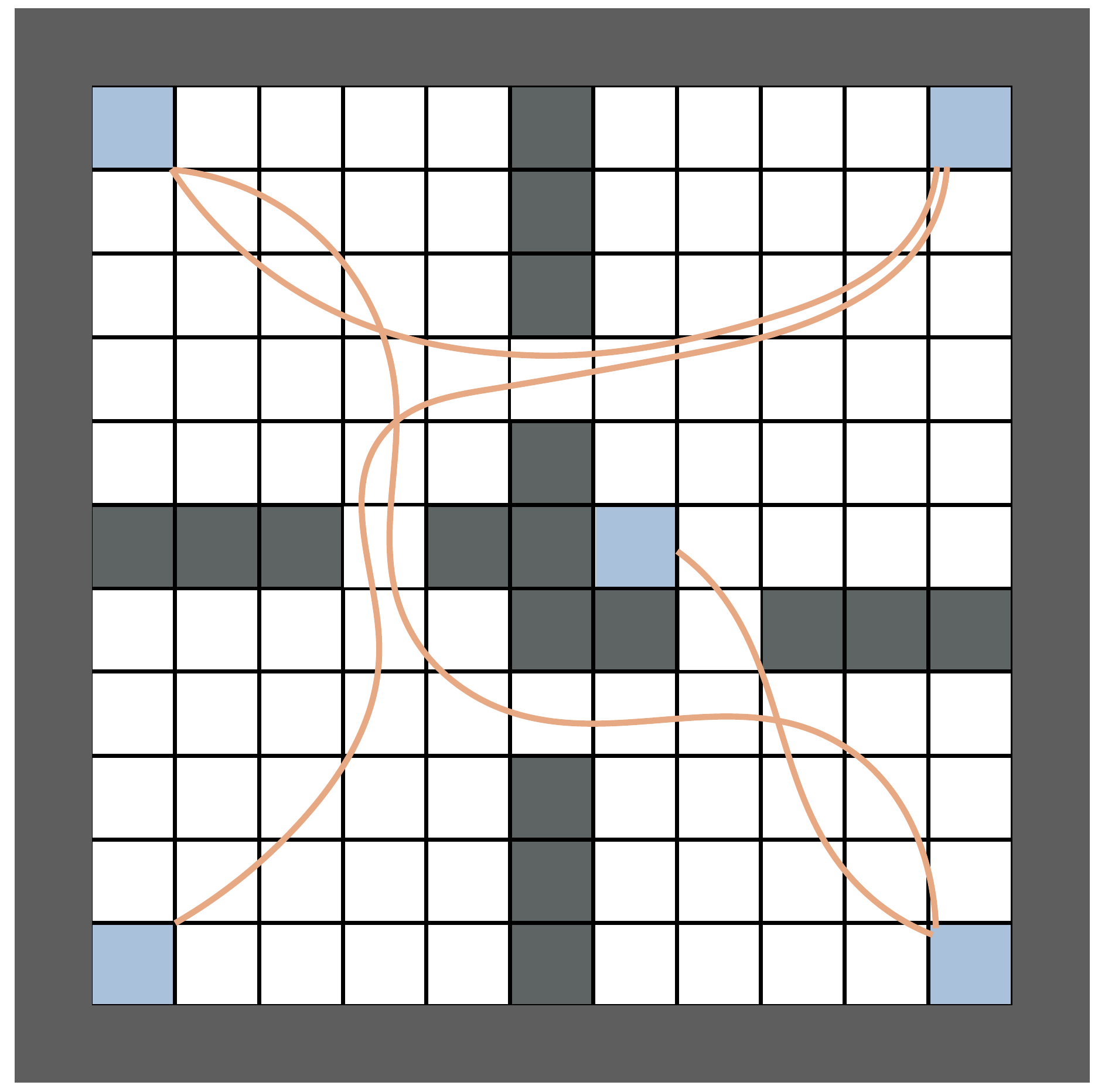}} \hspace{2mm}
    \subfloat[Eigenoptions]{\includegraphics[width=0.22\textwidth]{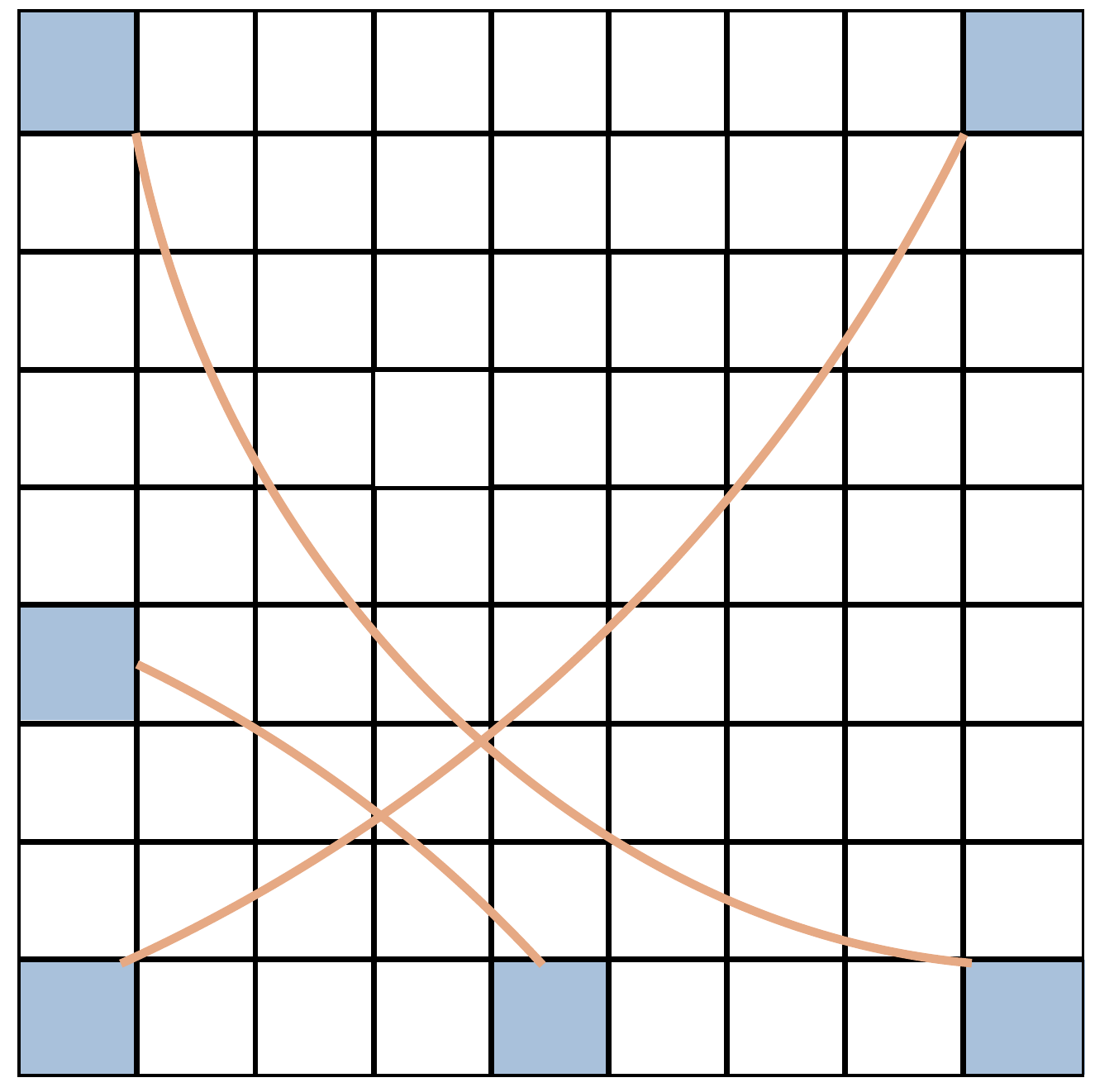}}
    
    \caption{Visualization of covering options and eigenoptions in Four Rooms and the 9$\times$9 grid world.}
    \label{fig:c8_visuals}
\end{figure}

We next present a series of visualizations that highlight important qualitative properties of covering options. \autoref{fig:c8_visuals} visualizes the eight covering options and eigenoptions found in Four Rooms and the 9$\times$9 grid world. Note that each algorithm may output many different sets of options, and we here choose to visualize just one set. Observe that in each MDP, both option types tend to connect options that are far apart in the underlying MDP. In Four Rooms, for instance, eigenoptions and covering options tend to connect the states in the opposite corners together. In the 9$\times$9 grid world, this quality is also present. The options found by both approaches tend to connect states that are far apart, suggesting that they each increase the algebraic connectivity of the MDP's transition graph.

\begin{figure}[b!]
    \centering
    \subfloat[\algname{} \newline (Four Rooms)]{\includegraphics[width=0.3\textwidth]{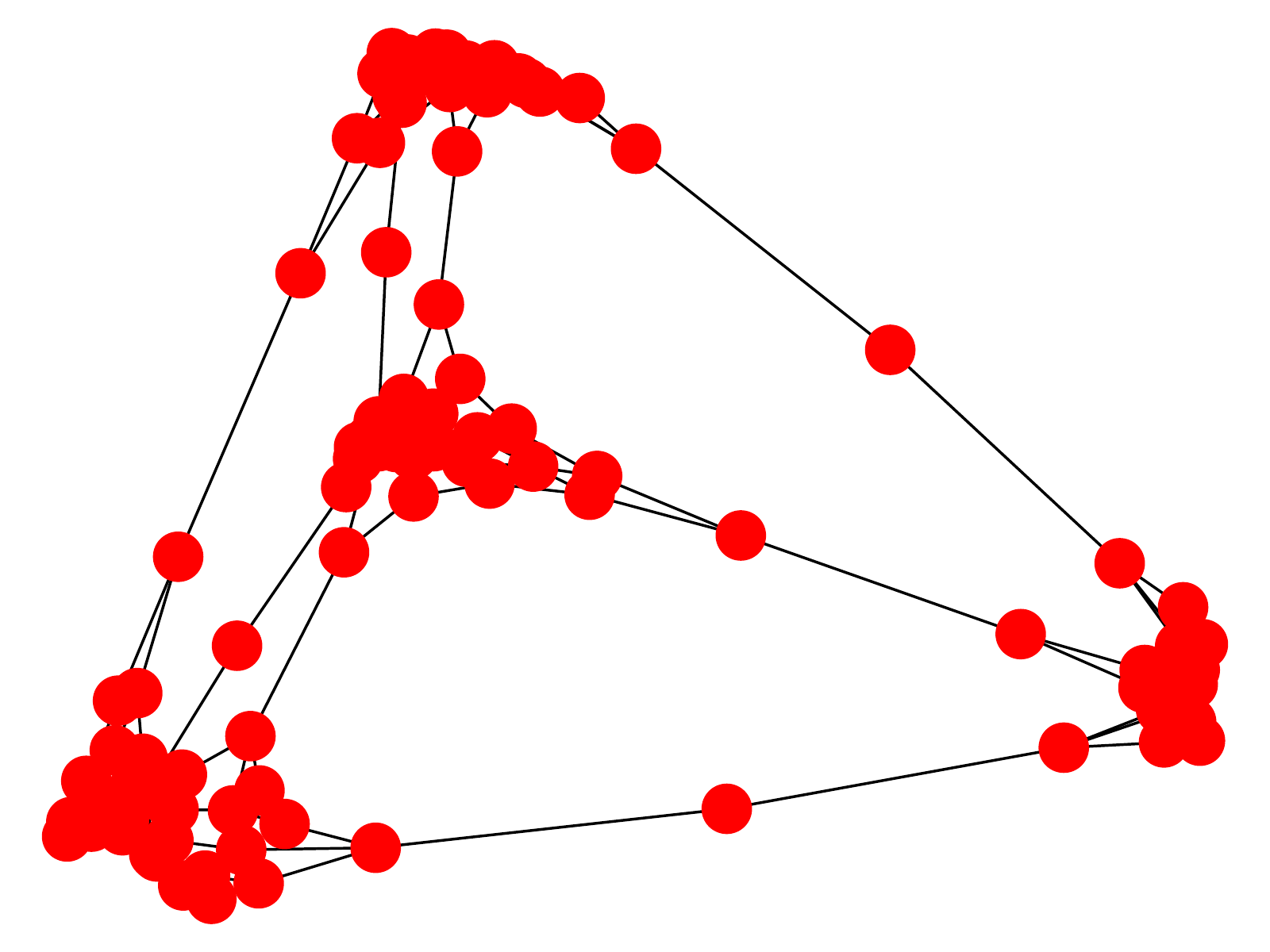}}\hspace{4mm}
    \subfloat[Eigenoptions (Four Rooms)]{\includegraphics[width=0.3\textwidth]{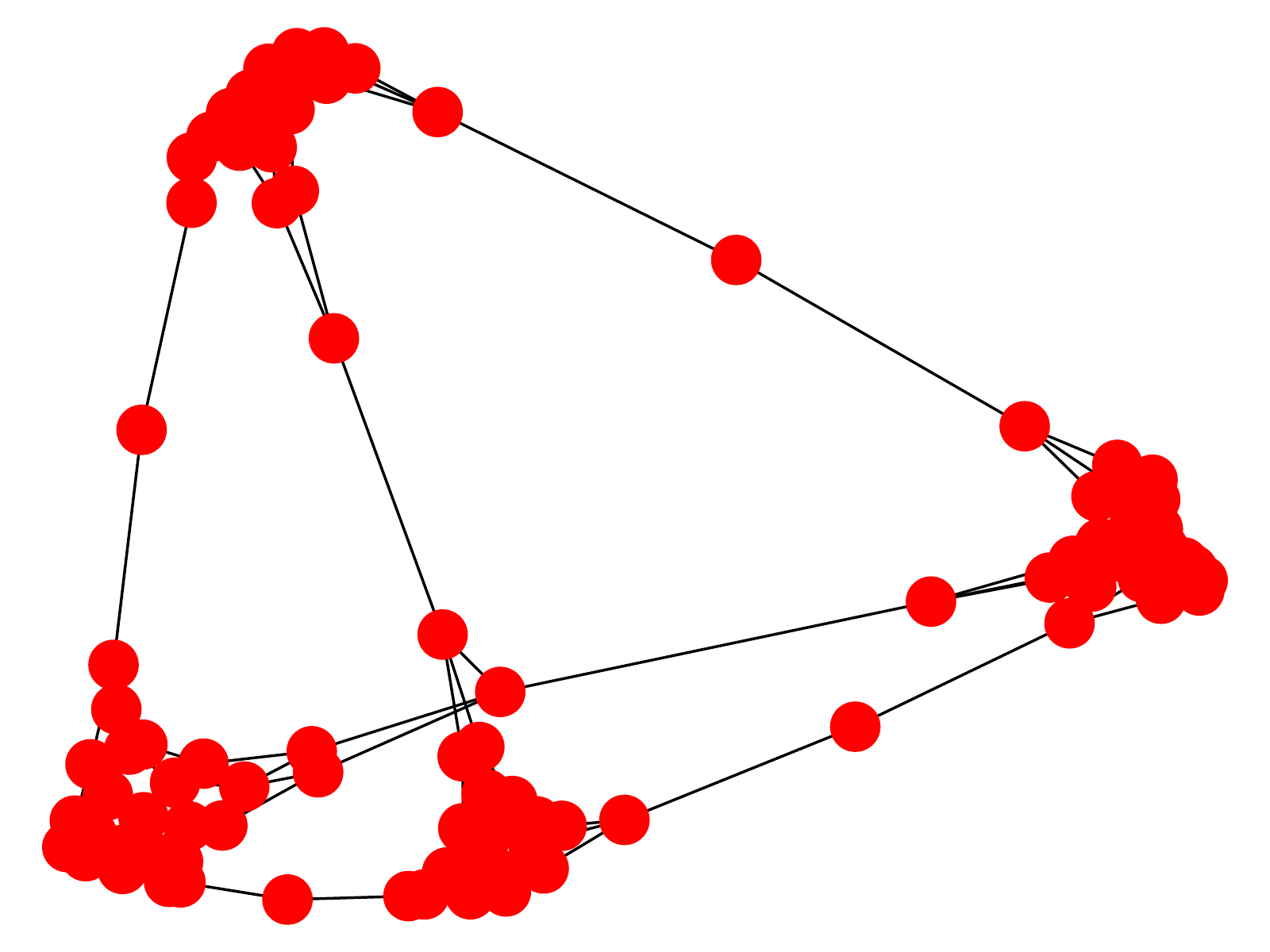}}\hspace{4mm}
    \subfloat[No options (Four Rooms)]{\includegraphics[width=0.3\textwidth]{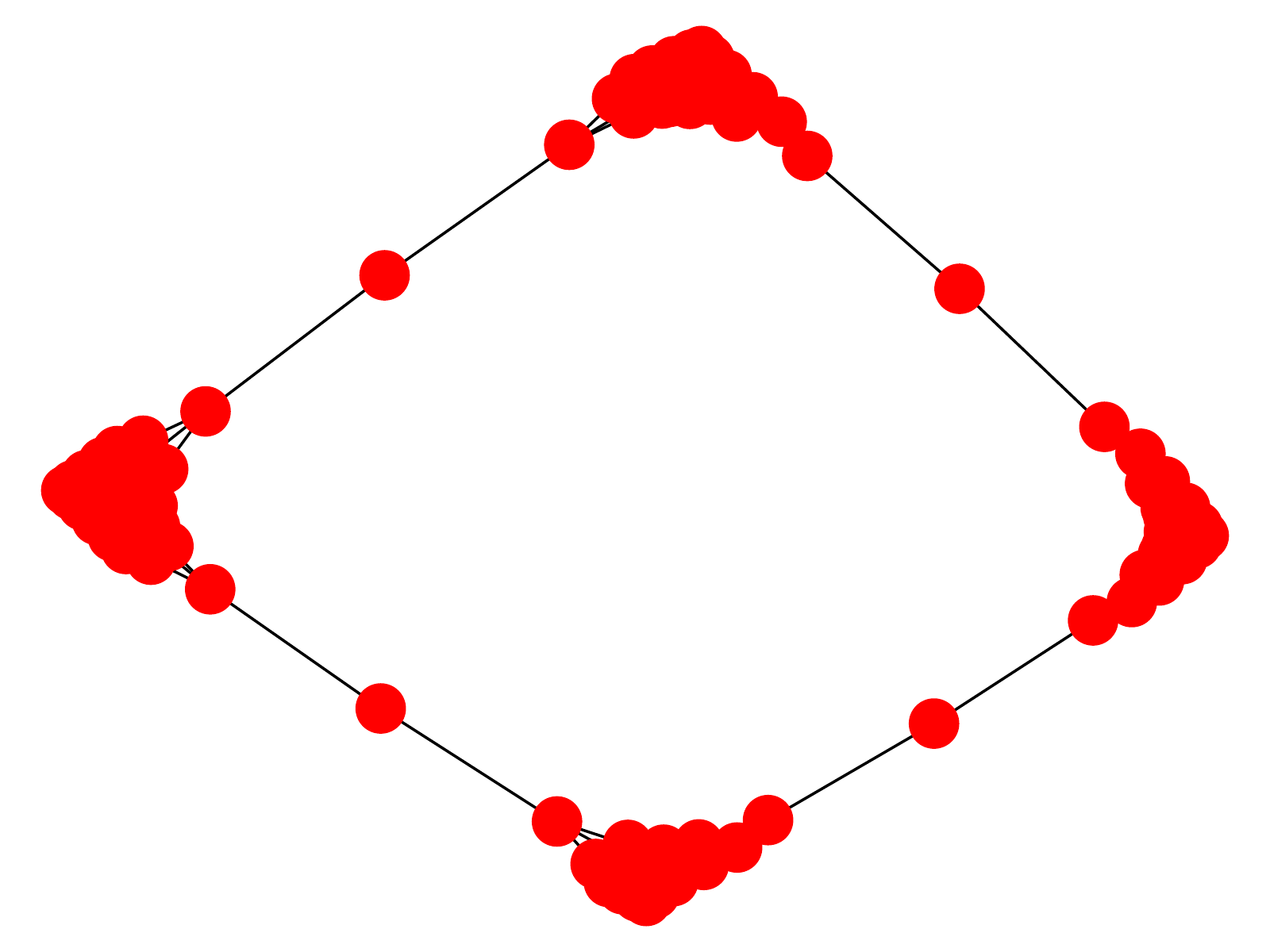}}
    
    \subfloat[\algname{} (9$\times$9 grid)]{\includegraphics[width=0.3\textwidth]{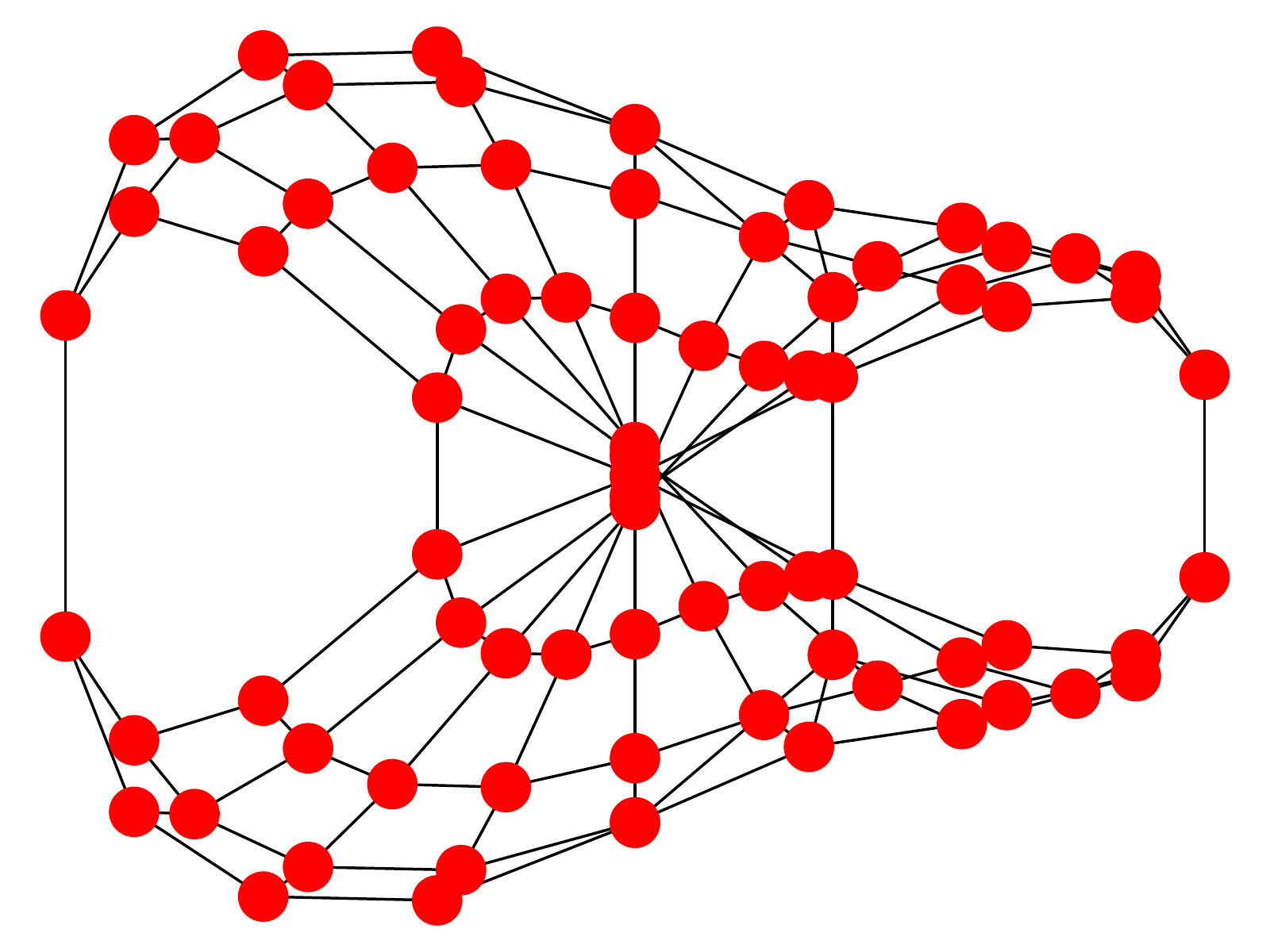}} \hspace{4mm}
    \subfloat[Eigenoptions (9$\times$9 grid)]{\includegraphics[width=0.3\textwidth]{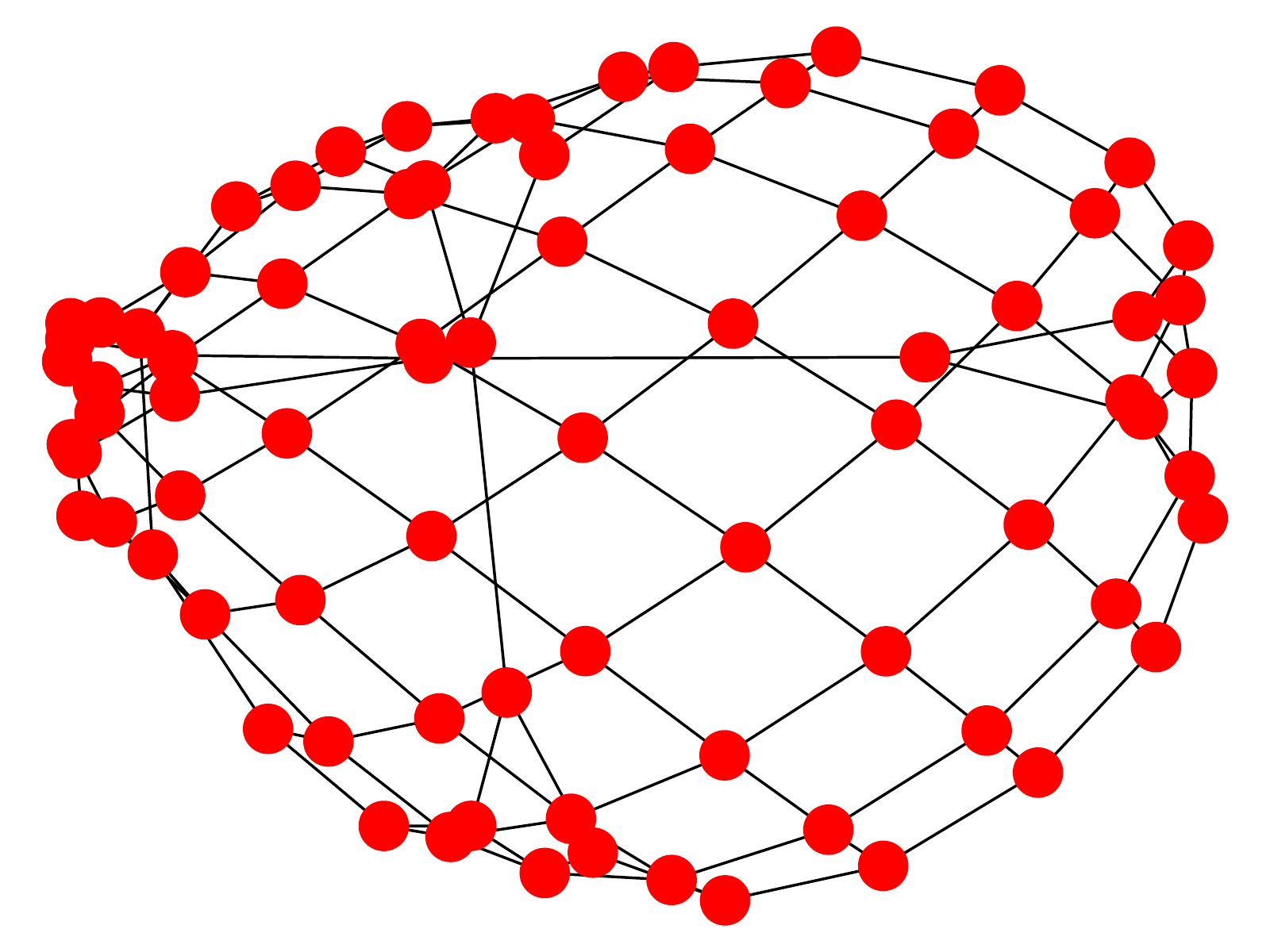}} \hspace{4mm}
    \subfloat[No options (9$\times$9 grid)]{\includegraphics[width=0.3\textwidth]{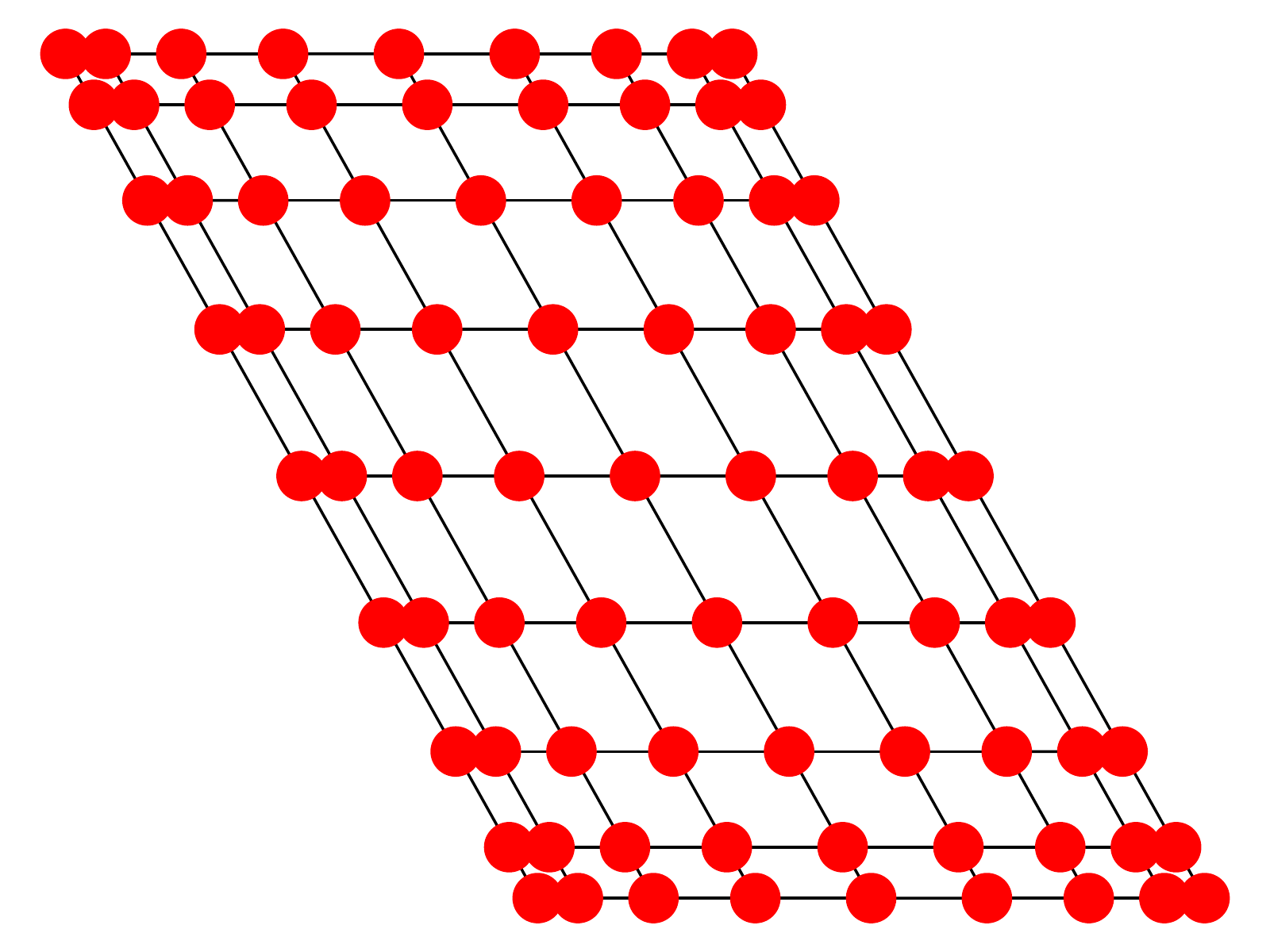}}
    \caption{Spectral graph drawing of the state-transition graph.}
    \label{fig:c8_spectral-drawing}
\end{figure}

Next, we further highlight the qualitative impact different options have on these two grid worlds. \autoref{fig:c8_spectral-drawing} presents the spectral graph drawing \cite{koren2003spectral} of the state-transition graph augmented with each option type. The spectral graph drawing is a technique that is used to visualize the graph topology using eigenvectors of the graph Laplacian. Each node $n$ in the state-space graph is placed at $(v_2(n), v_3(n))$ in the $(x, y)$-coordinate, where $v_i$ is the $i$-th smallest eigenvector of the graph Laplacian. These visuals provide further qualitative support for the hypothesis that the option generation methods are successfully connecting distant states.

\subsection{RL with Options}

We next present findings from two sets of experiments examining how the options found by our algorithm can impact RL.

\paragraph{Options Computed Offline.} In the first variant, we suppose that the transition graph is given to the option discovery method offline, and the computed options are given to an RL algorithm before learning begins. In each experiment we test with $Q$-learning as the underlying RL algorithm paired with different option types, where $\alpha=0.1, \gamma=0.95$. Each approach is run for $100$ episodes with $100$ steps per episode for the 9$\times$9 grid, and $500$ steps per episode in Four Rooms, Hanoi, and Taxi.

As one caveat, following the methods of \citet{machado2017laplacian}, we evaluate our method using a sample-based approach for option discovery in both Race Track and Parr's maze. That is, instead of giving the agent access to the whole adjacency matrix, the agent instead samples 100 trajectories of a uniform random policy in the MDP, and uses this data to form an incidence matrix. We sampled each trajectory for 1000 steps for Parr's maze and 100 steps for the Race Track domain, and use these data to generate an incidence matrix to inform option generation. As the agent has no prior knowledge on states not present in the incidence matrix, the agent terminates the option if it reaches a state outside of the incidence matrix.

\begin{figure}[t!]
    \centering
    \subfloat[9$\times$9 grid]{\includegraphics[width=0.3\textwidth]{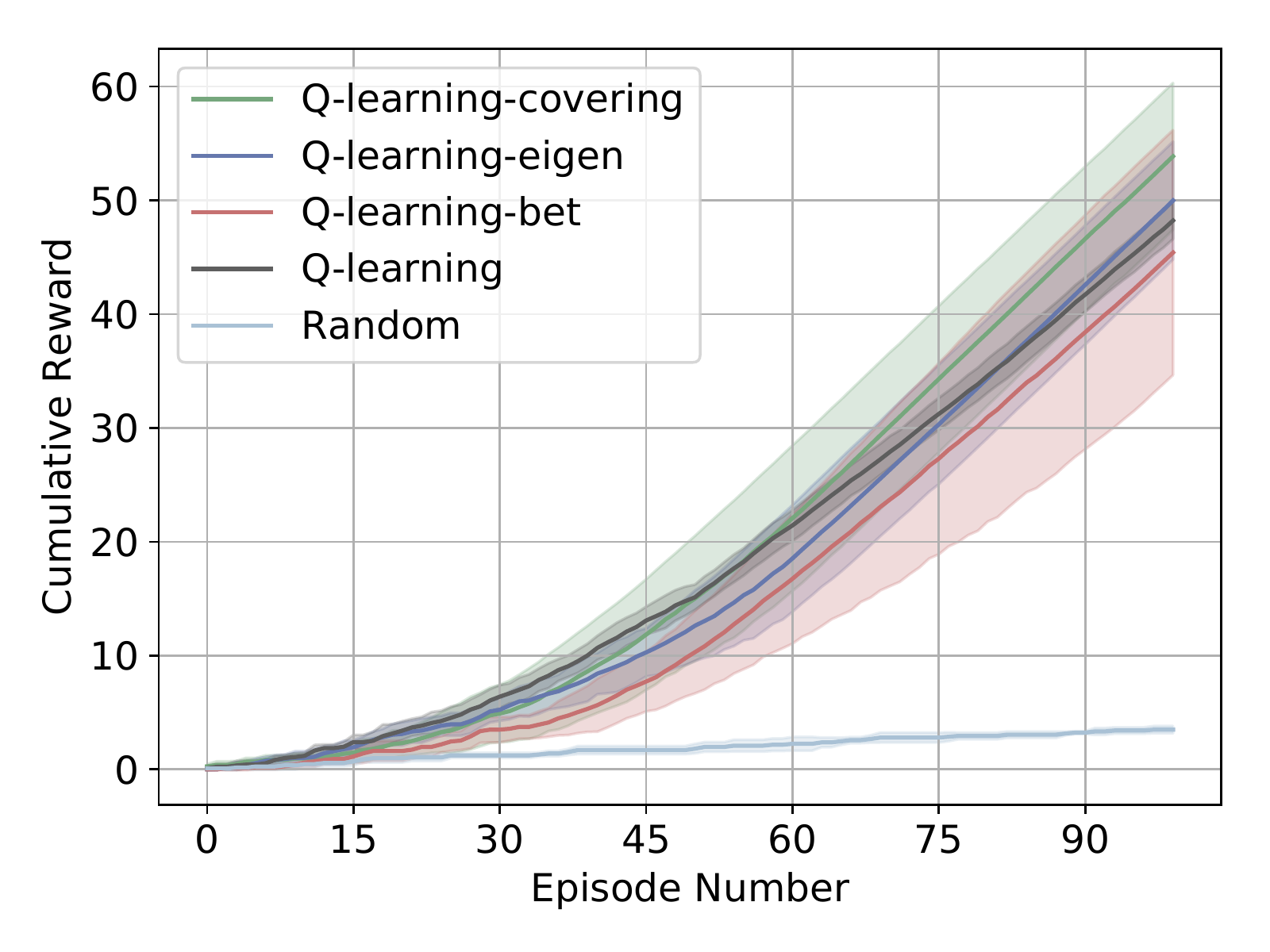}} \hspace{3mm}
    \subfloat[Four Rooms]{\includegraphics[width=0.33\textwidth]{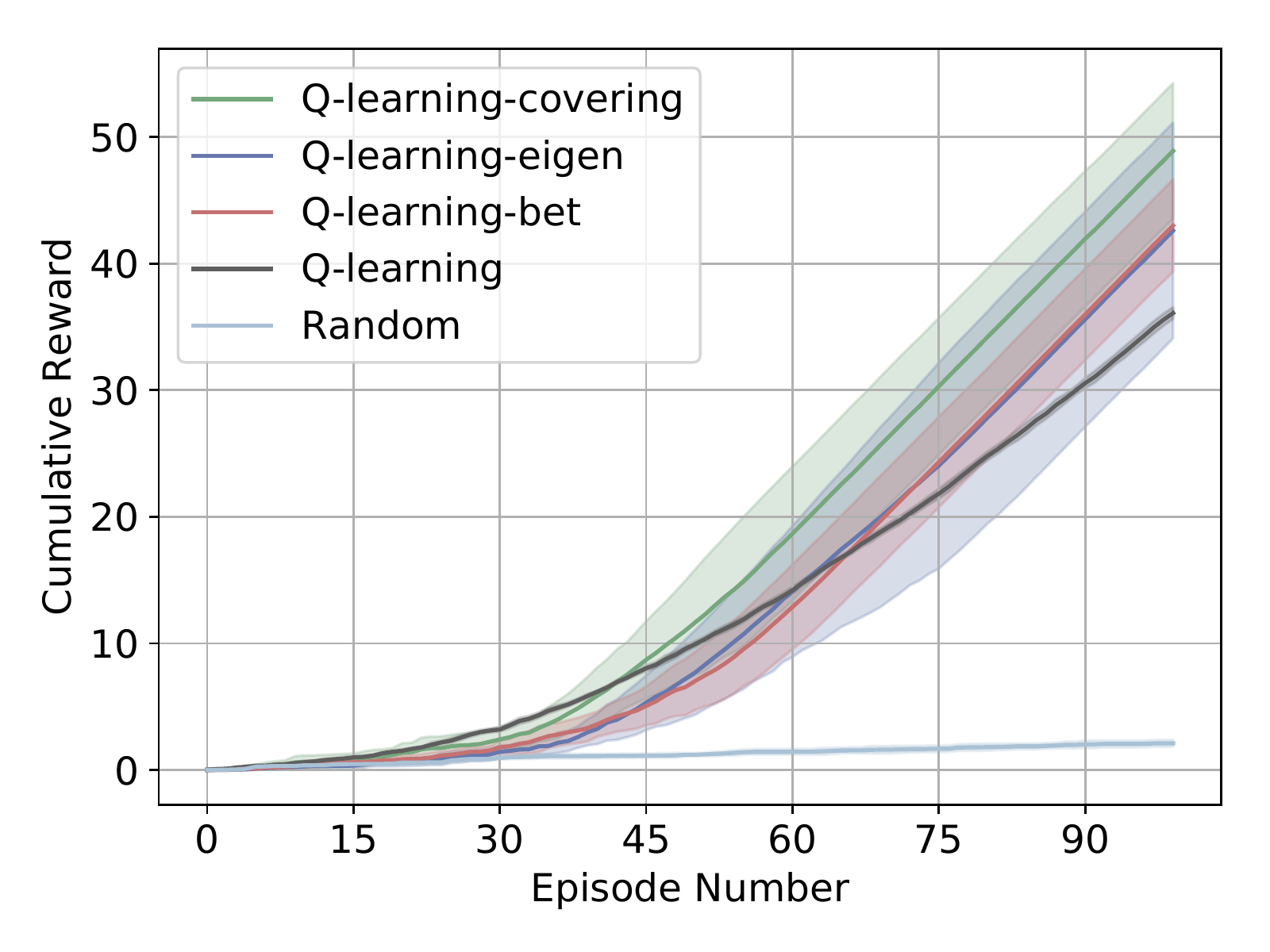}} \hspace{3mm}
    \subfloat[Towers of Hanoi]{\includegraphics[width=0.3\textwidth]{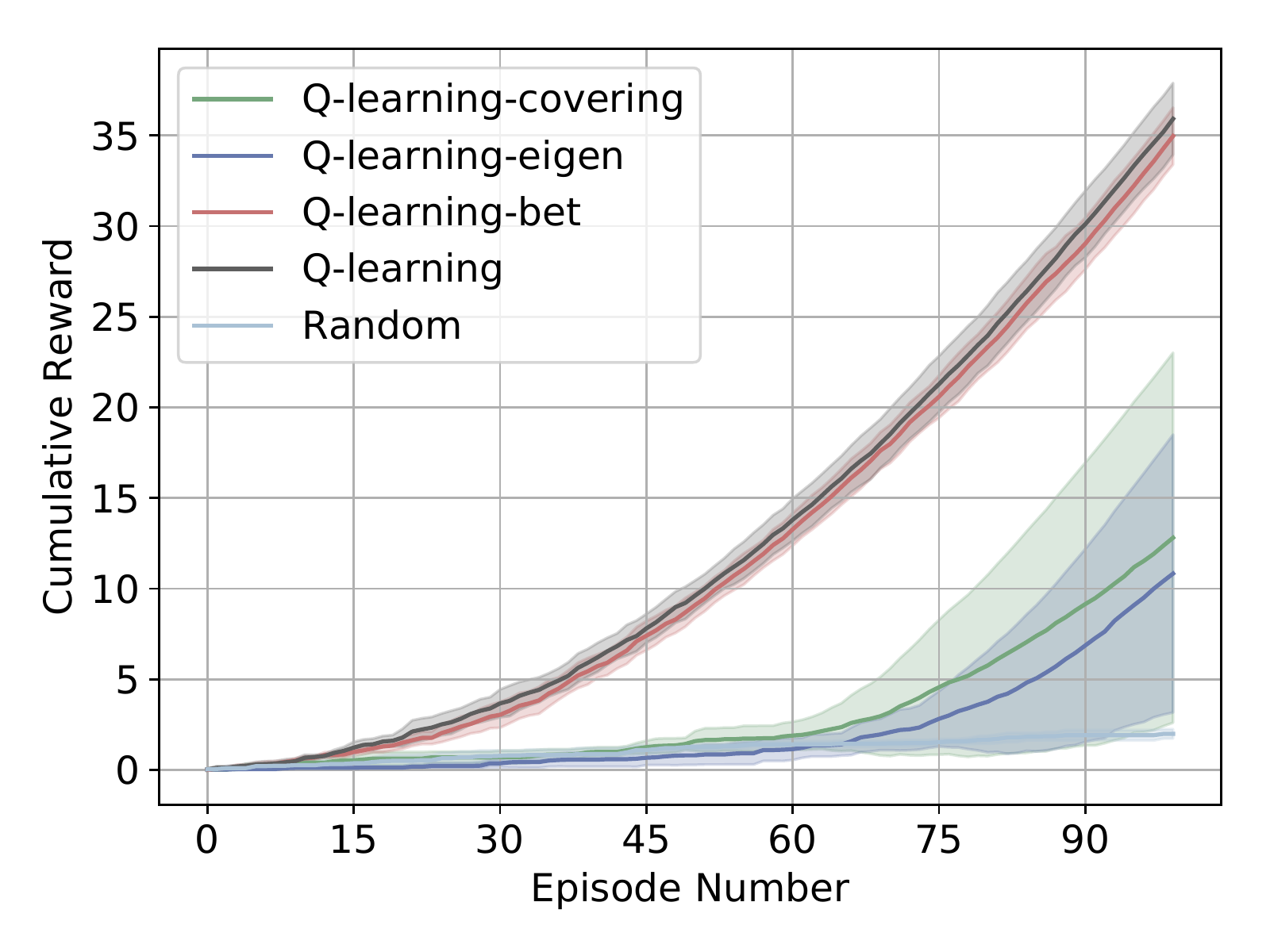}}
    
    \subfloat[Taxi]{\includegraphics[width=0.33\textwidth]{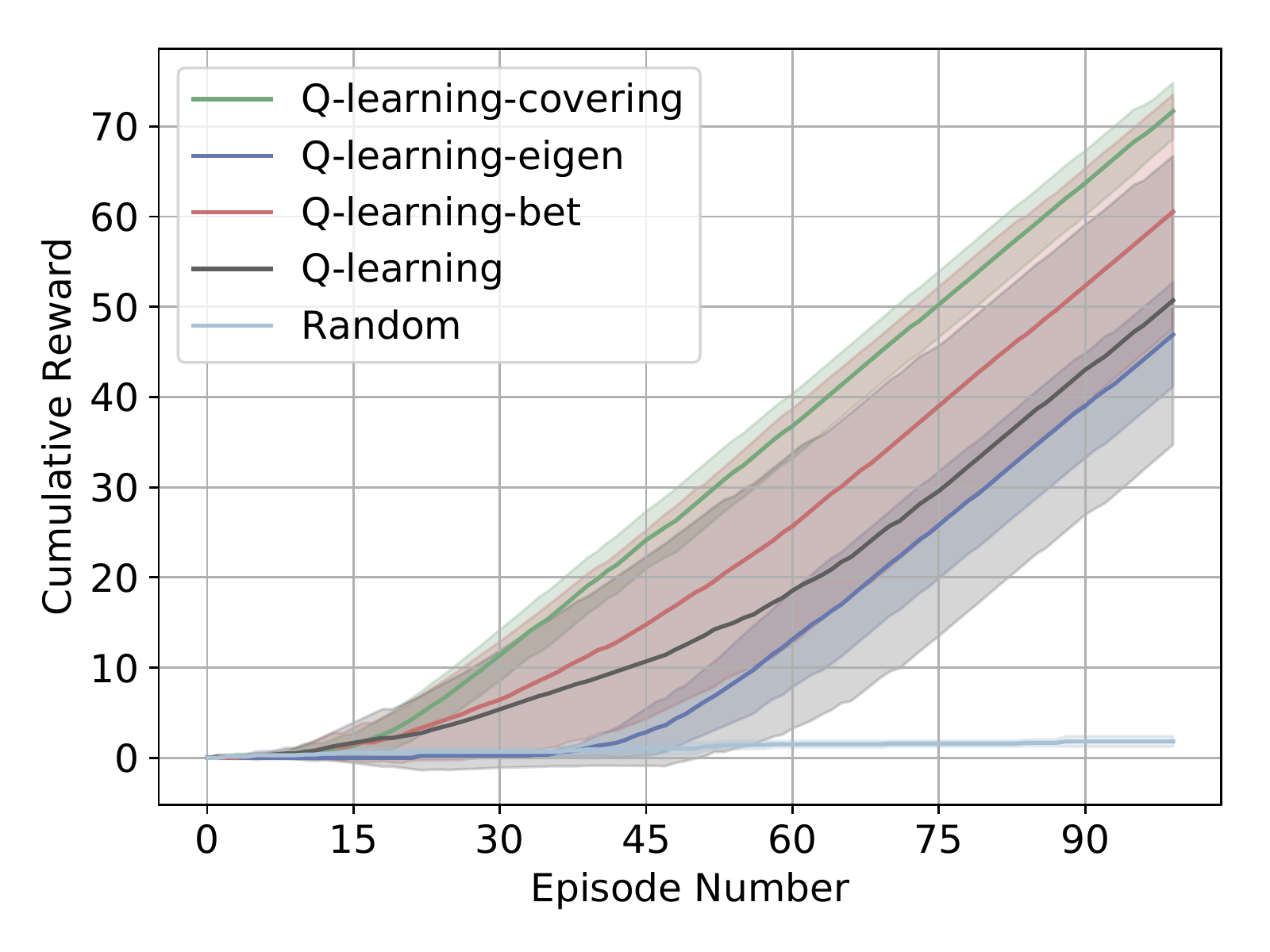}} \hspace{3mm}
    \subfloat[Parr's Maze]{\includegraphics[width=0.3\textwidth]{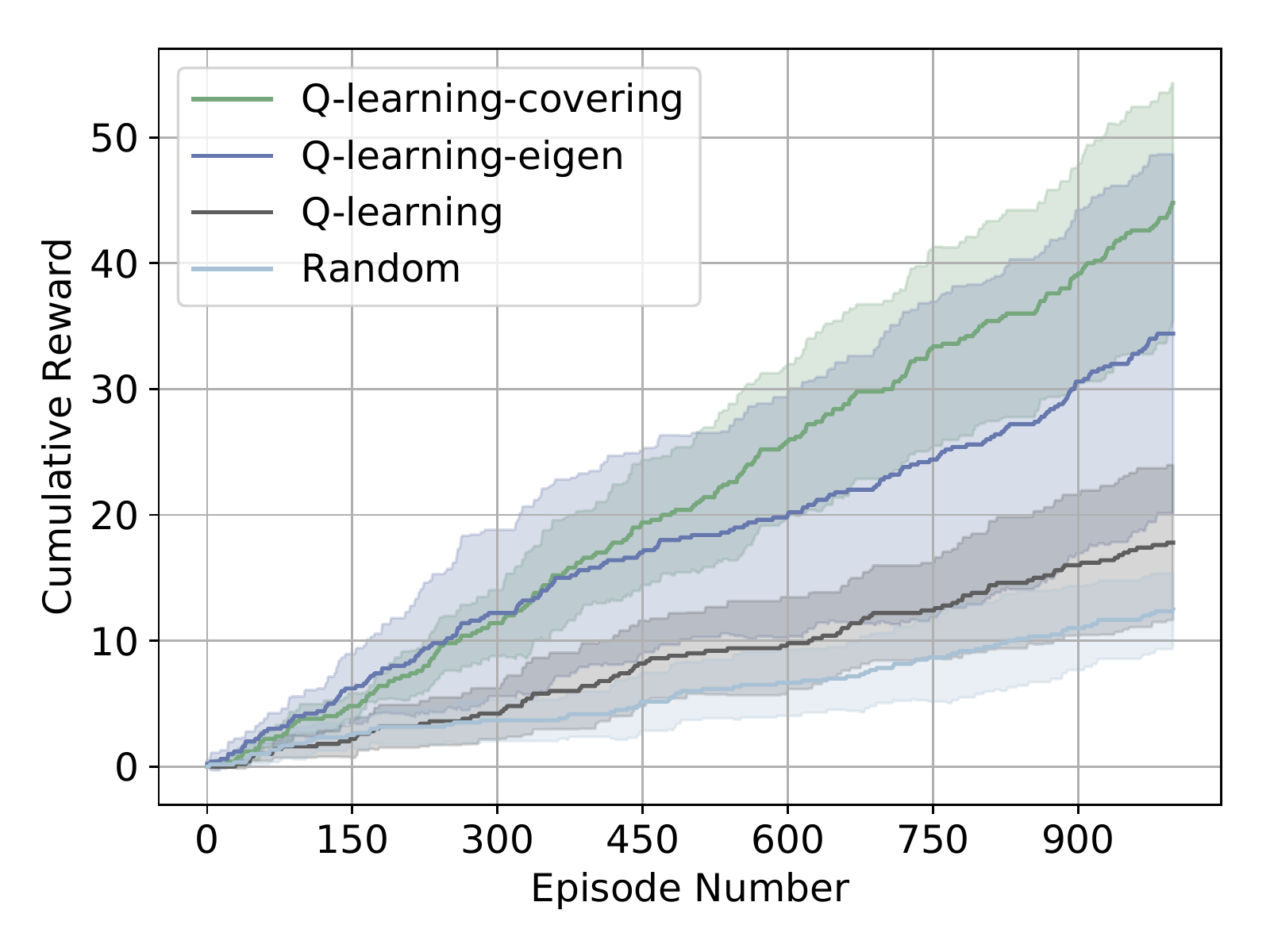}} \hspace{3mm}
    \subfloat[Race Track ]{\includegraphics[width=0.3\textwidth]{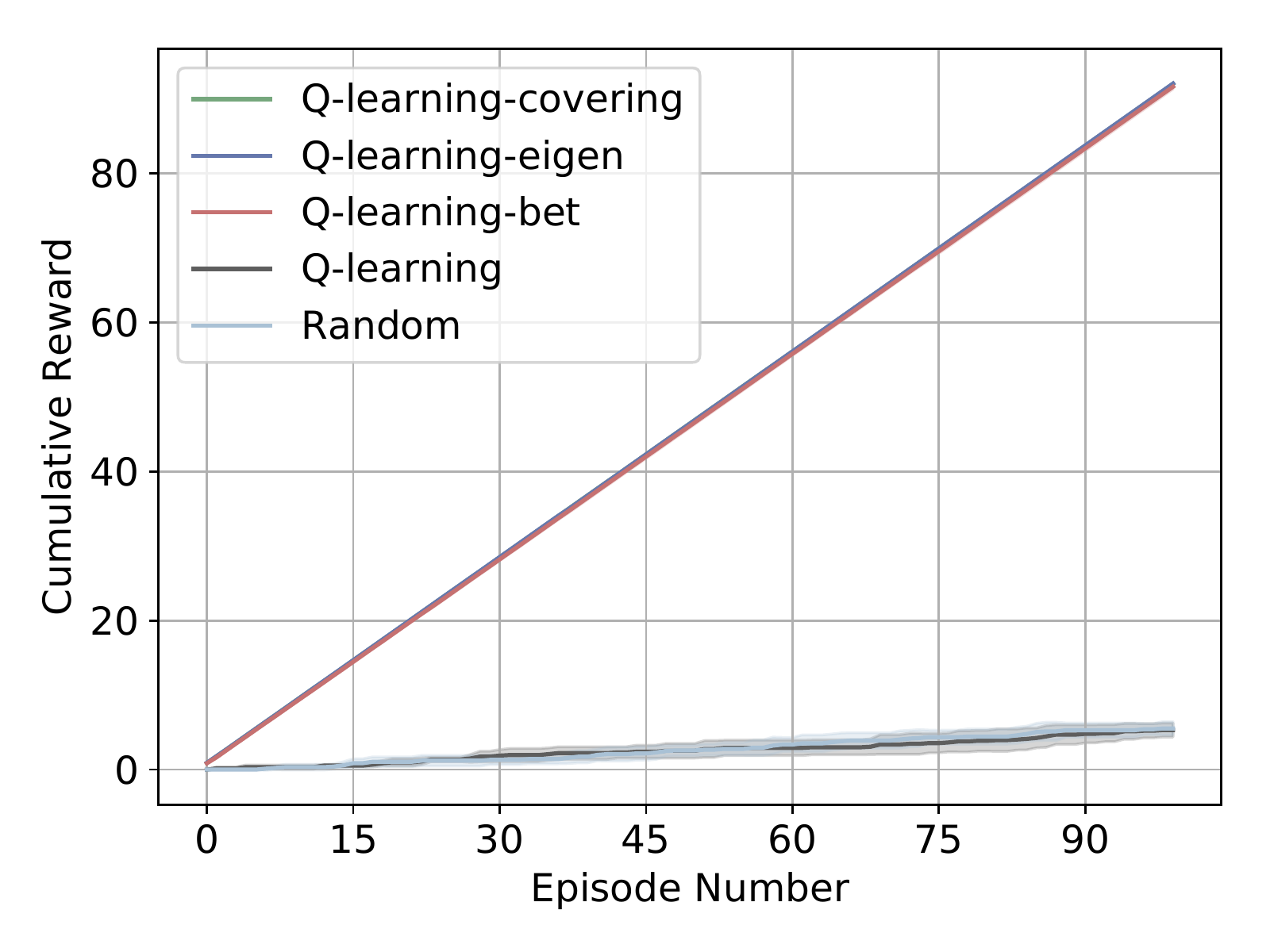}}
    
    \caption{Comparison of RL performance with different option generation methods.}
    \label{fig:c8_adjacency}
\end{figure}

\autoref{fig:c8_adjacency} presents the mean cumulative reward per averaged over five runs in each MDP. In some cases, the options neither accelerate nor deteriorate the learning---for instance, in both the $9\times9$ grid and Four Rooms, each of covering, eigen, and betweenness options all perform comparably to regular $Q$-learning. In Four Rooms, the quality of the policy learned by each of the variants including options does seem to be slightly higher on average than that of $Q$-learning, though it is not a statistically significant improvement. Conversely, in Race Track, we see covering options and betweenness options dramatically outperform all other methods, while in Towers of Hanoi, covering options and eigenoptions negatively impact learning. In summary, the data suggest that each option type can both help or hurt learning depending on the context, but options tend to help more often than hurt on the studied domains.

\paragraph{Options Computed Online.} Lastly, we study the case where options are discovered online during learning. Each agent generates four options to add to their option set every 10,000 step for Parr's maze and 500 steps for the Towers of Hanoi and Taxi, until $|\mc{O}| = 32$. Each agent is given 100 episodes consisting of 10,000 steps each in Parr's maze and 100 steps each for Hanoi and Taxi. The policy of each option is computed by forming the greedy policy relative to a $Q$ function learned by running $Q$-learning ($\alpha=0.1, \gamma=0.95$) on the sampled data to convergence. Here, I give an intrinsic reward of 1 to the agent when it reaches the subgoal state and ignore the rewards from the environment.

\begin{figure}[t!]
    \centering
	\subfloat[Parr's maze]{\includegraphics[width=0.3\textwidth]{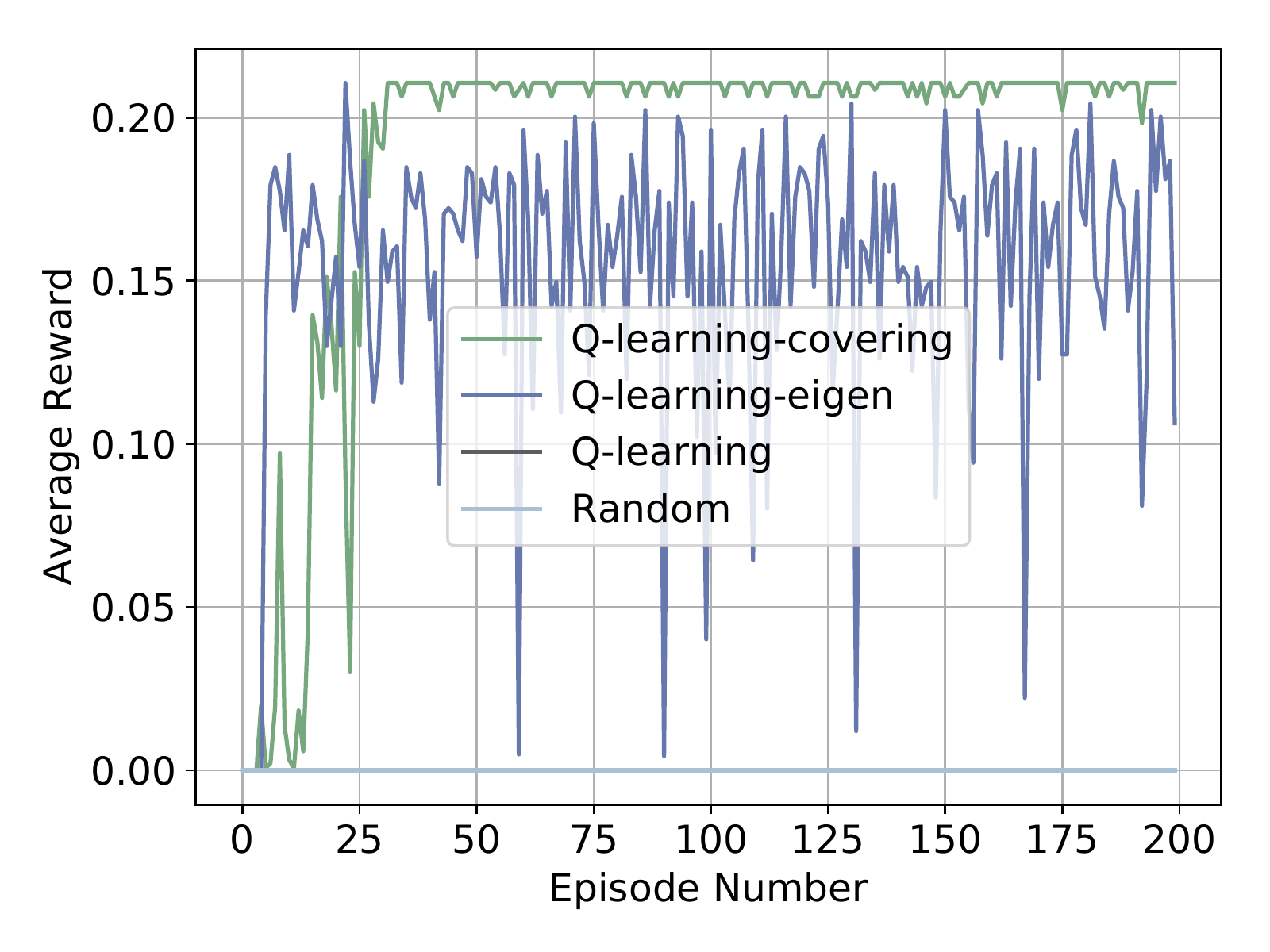} \label{fig:c8_parronline}} \hspace{3mm}
	\subfloat[Towers of Hanoi]{\includegraphics[width=0.3\textwidth]{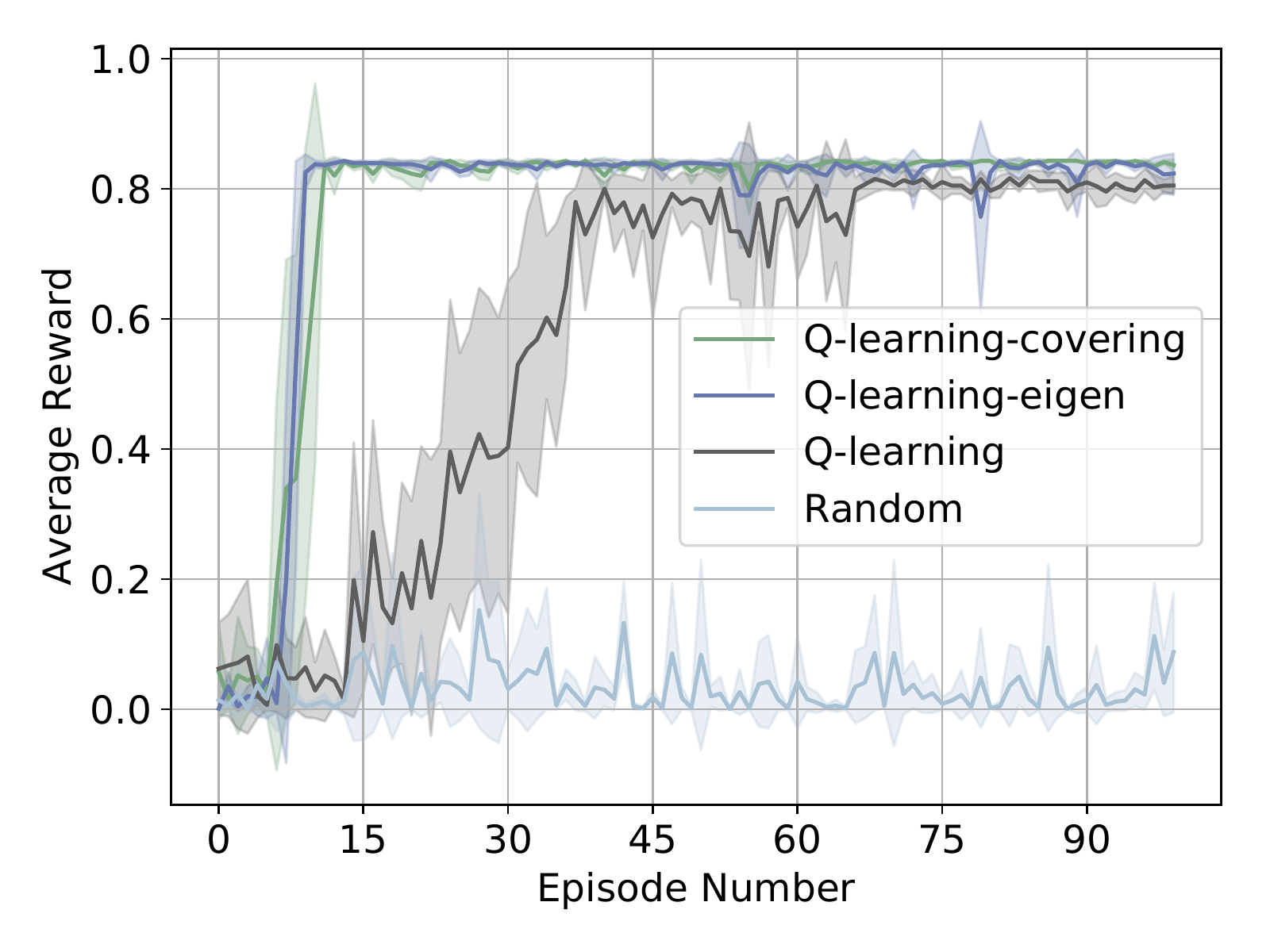} \label{fig:c8_hanoionline}} \hspace{3mm}
	\subfloat[Taxi]{\includegraphics[width=0.3\textwidth]{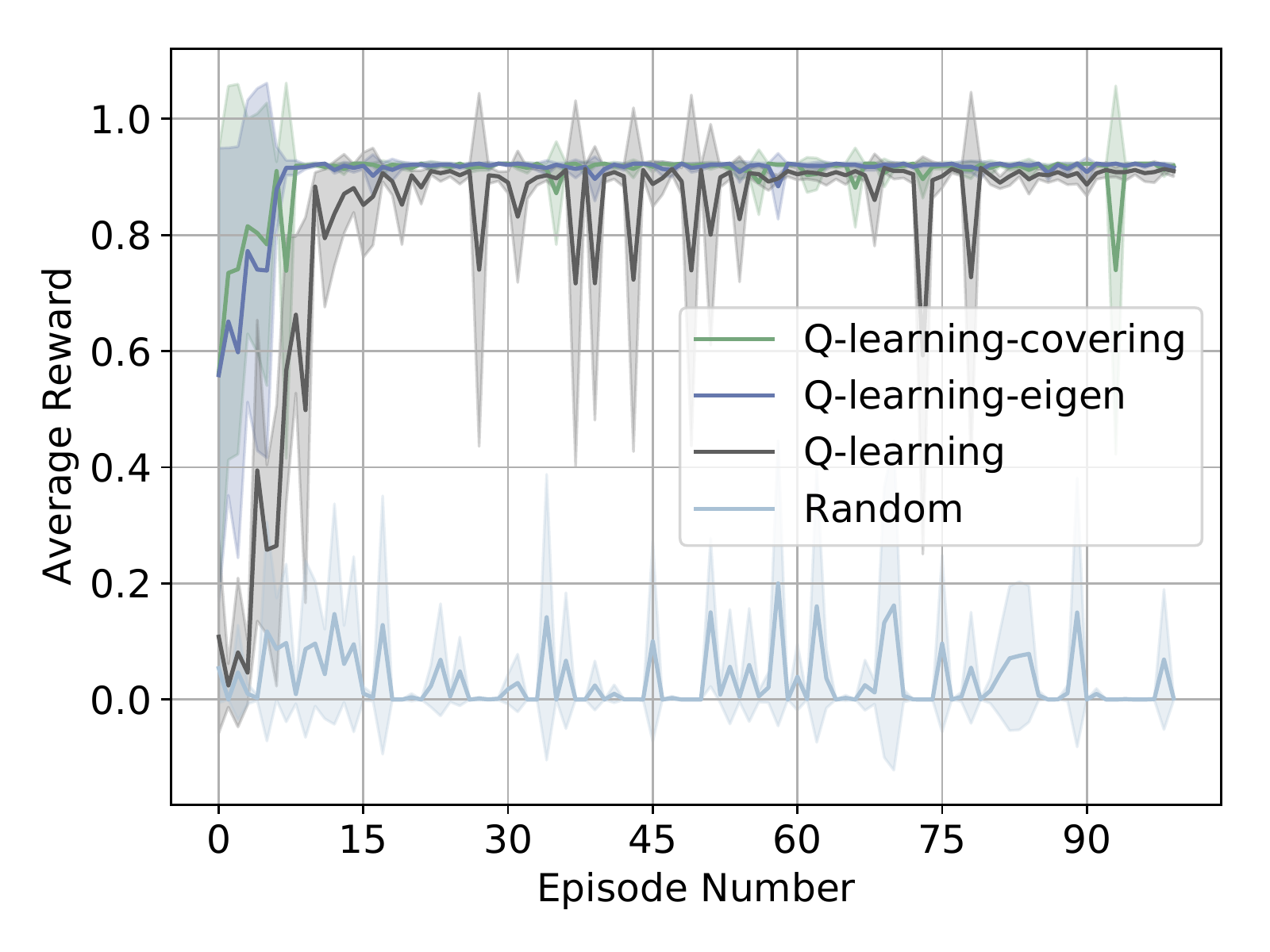}  \label{fig:c8_taxionline}}
    \caption{Comparison of online option generation methods.}
    \label{fig:c8_online}
\end{figure}

Results are presented in \autoref{fig:c8_online}, indicating the average reward per episode. Observe that across all three domains, $Q$-learning paired with covering options is able to reliably find a good policy, giving support to the claim that covering options can in fact accelerate exploration. In Parr's maze, for instance, a goal-base problem with a long horizon before any goal is obtained, the approach with covering options is able to find the goal a non-negligible fraction of episodes after around 25 episodes, whereas an agent with primitive actions is unable to find the goal throughout all of learning. Further observe that covering options and eigenoptions tend to perform similarly, suggesting that they can each be useful for accelerating exploration in RL.

\subsection{Concluding Remarks}
In this chapter, I illustrated the sense in which appropriate action abstractions can accelerate exploration in RL. In the previous two chapters, I concentrated on finding options that make planning efficient (\autoref{chap:options_for_planning}), and motivated an new alternative to the standard option models, (\autoref{chap:elm_options}). Collectively, the results established in this part of the dissertation offer support for the great potential of action abstraction to accelerate and improve RL, and provide concrete paths to action abstraction that can satisfy the desiderata. 


I now turn to the next and final part of the dissertation in which I study good combinations of state and action abstraction.

\vspace{4mm}
\begin{center}
\begin{minipage}{0.65\textwidth}
\resizebox{0.9\textwidth}{6pt}{%
  \begin{tikzpicture}
        \pgfornament[symmetry=h]{85} 
    \end{tikzpicture}
}
\end{minipage}
\end{center}

%% file: chapters/c9_hierarchical_abstraction.tex
\begin{center}
\begin{minipage}{0.8\textwidth}
\textit{This chapter is based on ``Value Preserving State-Action Abstractions" \cite{abel2020vpsa} with Nathan Umbanhowar, Khimya Khetarpal, Dilip Arumugam, Doina Precup, Michael L. Littman.}
\end{minipage}
\end{center}
\vspace{2mm}

In light of the separate benefits of state and action abstraction (see \autoref{part:state_abstraction} and \autoref{part:action_abstraction}), it has long been of interest as to how to appropriately combine the two methods. To this end, the focus of this chapter is on the following question.

\begin{center}
\begin{minipage}{0.75\textwidth}
\begin{center}
    \textit{Which combinations of state abstractions and options preserve representation of near-optimal policies?}
\end{center}
\end{minipage}
\end{center}

The main result of this chapter summarize new analysis addressing this question, providing a concrete step toward state-action abstractions that can satisfy all three desiderata. Specifically, I here introduce combinations of state abstractions ($\phi$) and options ($\mc{O}$) that are guaranteed to preserve representation of near-optimal policies in finite MDPs. These combinations, and the analysis thereof, resemble the classes of approximate state abstraction studied in \autoref{chap:approx_state_abstr}---the main theorem of the chapter (\autoref{thm:c9_vpsa_main_result}) highlights the general relationship between approximate knowledge used in forming these abstractions and the quality of the best policy representable in the abstract. I will then extend this result to the case of \textit{hierarchical} abstractions, providing a general scheme for characterizing value-preserving hierarchies under mild assumptions.

To perform this analysis, I first define $\phi$-relative options, a general formalism for analyzing the value loss of a state abstraction paired with a set of options. I then prove four sufficient conditions, along with one separate necessary condition, for $\phi$-relative options to preserve near-optimal behavior in any finite MDP. I further prove that $\phi$-relative options can be composed to induce a hierarchy that preserves near-optimal behavior under appropriate assumptions about the hierarchy's construction. I suggest these results can support the development of principled methods that learn and make use of value-preserving abstractions.

\begin{figure}[b!]
    \centering
    \includegraphics[width=0.6\textwidth]{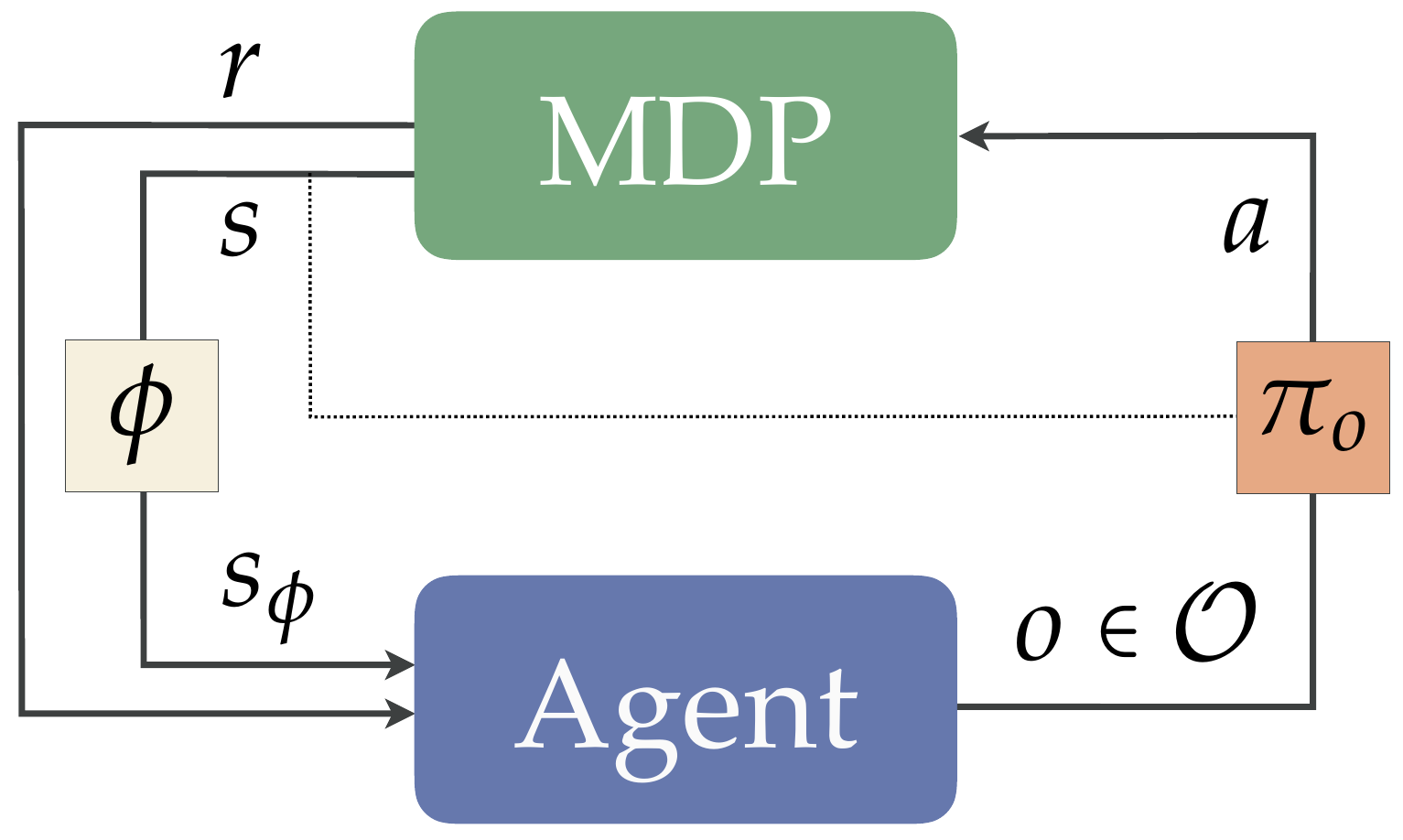}
    \caption{State and action abstraction in RL.}
    \label{fig:c9_sa_aa_rl}
\end{figure}

\section{Analysis: State-Action Abstractions}
\label{sec:analysis_phi_rel_opts}

I incorporate state and action abstraction into RL as pictured in \autoref{fig:c9_sa_aa_rl}. When the environment transitions to a new state $s$, the agent processes $s$ via $\phi$ yielding the abstract state, $s_\phi$. Then, the agent chooses an option from among those that initiate in $s_\phi$ and follows the chosen option's policy until termination, where this process repeats. In this way, an RL agent can reason in terms of abstract state and action alone, without knowing the true state or action space.

To analyze the value loss of these joint abstractions, I first introduce $\phi$-relative options, a simple means of combining state abstractions with options.

\ddef{$\phi$-Relative Option}{For a given $\phi$, an option $o$ is said to be $\bs{\phi}$\textbf{-relative} if and only if there is some $s_\phi \in \mc{S}_\phi$ such that, for all $s \in \mc{S}$:
\begin{equation}
    \mc{I}_o = \{s' \in s_\phi\}, \hspace{5mm} \beta_o(s) = \indic\{s \not \in s_\phi\}, \hspace{5mm} \pi_o \in \Pi_{s_\phi},
\end{equation}
where $\Pi_{s_\phi} = \left\{\pi : \{s' \in s_\phi\} \rightarrow \Delta(\mc{A})\right\}$ is the set of all ground policies defined over ground states in $s_\phi$, and $s \in s_\phi$ is shorthand for $s \in \{s' \in \mc{S} : \phi(s') = s_\phi\}$.}

Intuitively, these options initiate in exactly one abstract state and terminate when the option policy leaves the abstract state. I henceforth denote $(\phiop)$ as a state abstraction paired with a set of $\phi$-relative options, and denote $\mc{O}_\phi$ as any non-empty set that 1) contains only $\phi$-relative options, and 2) contains at least one option that initiates in each $s_\phi \in \mc{S}_\phi$.

\begin{figure}[t!]
    \centering
    \subfloat[Assignment of options to each $s_\phi$ via $\pi_{\mc{O}_\phi}$.\label{fig:c9_four_rooms}]{\includegraphics[width=0.42\textwidth]{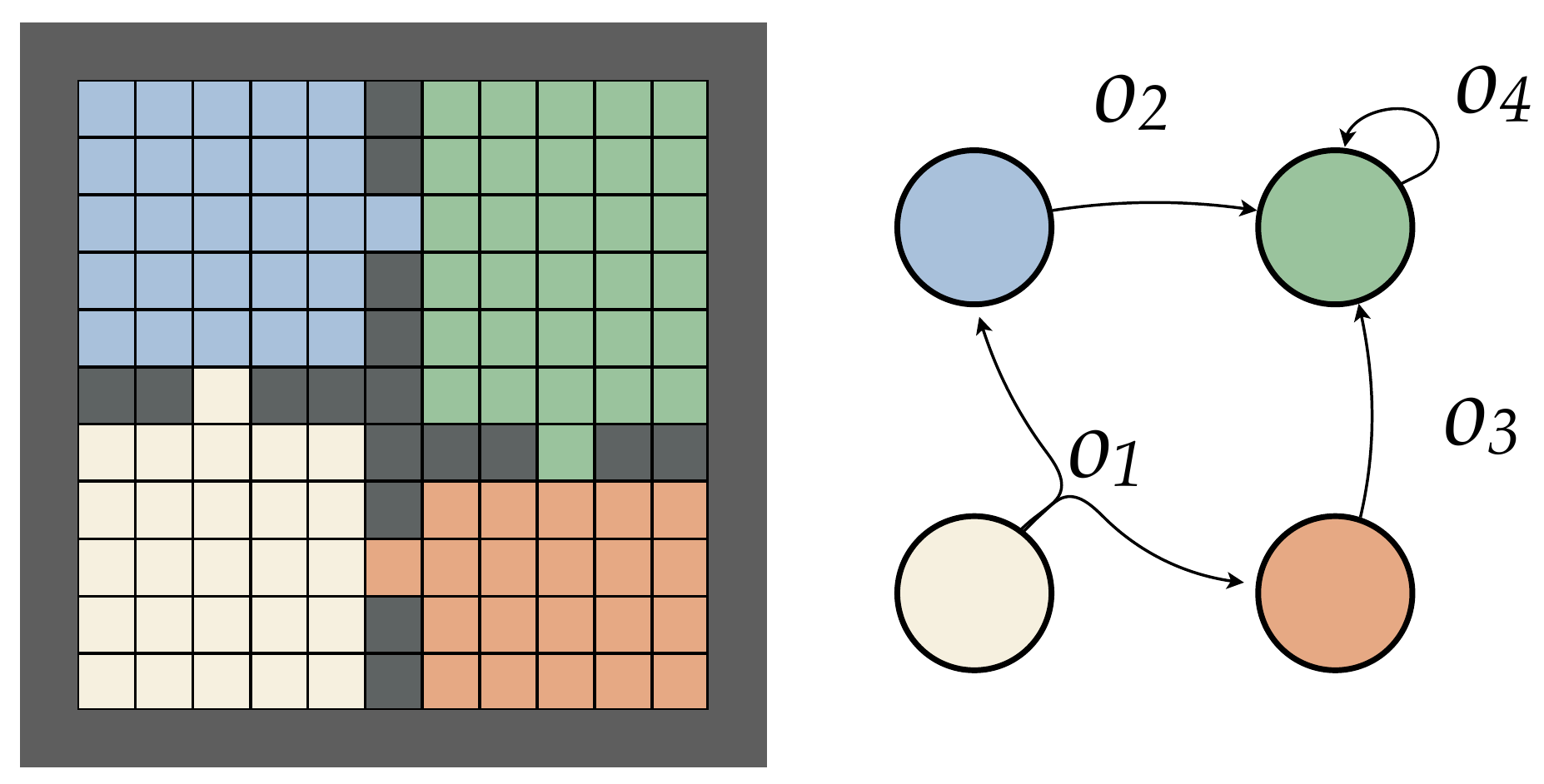}} \subfhspace
    \subfloat[Construction of $\pi_{\mc{O}_\phi}^\Downarrow$.]{\includegraphics[width=0.32\textwidth]{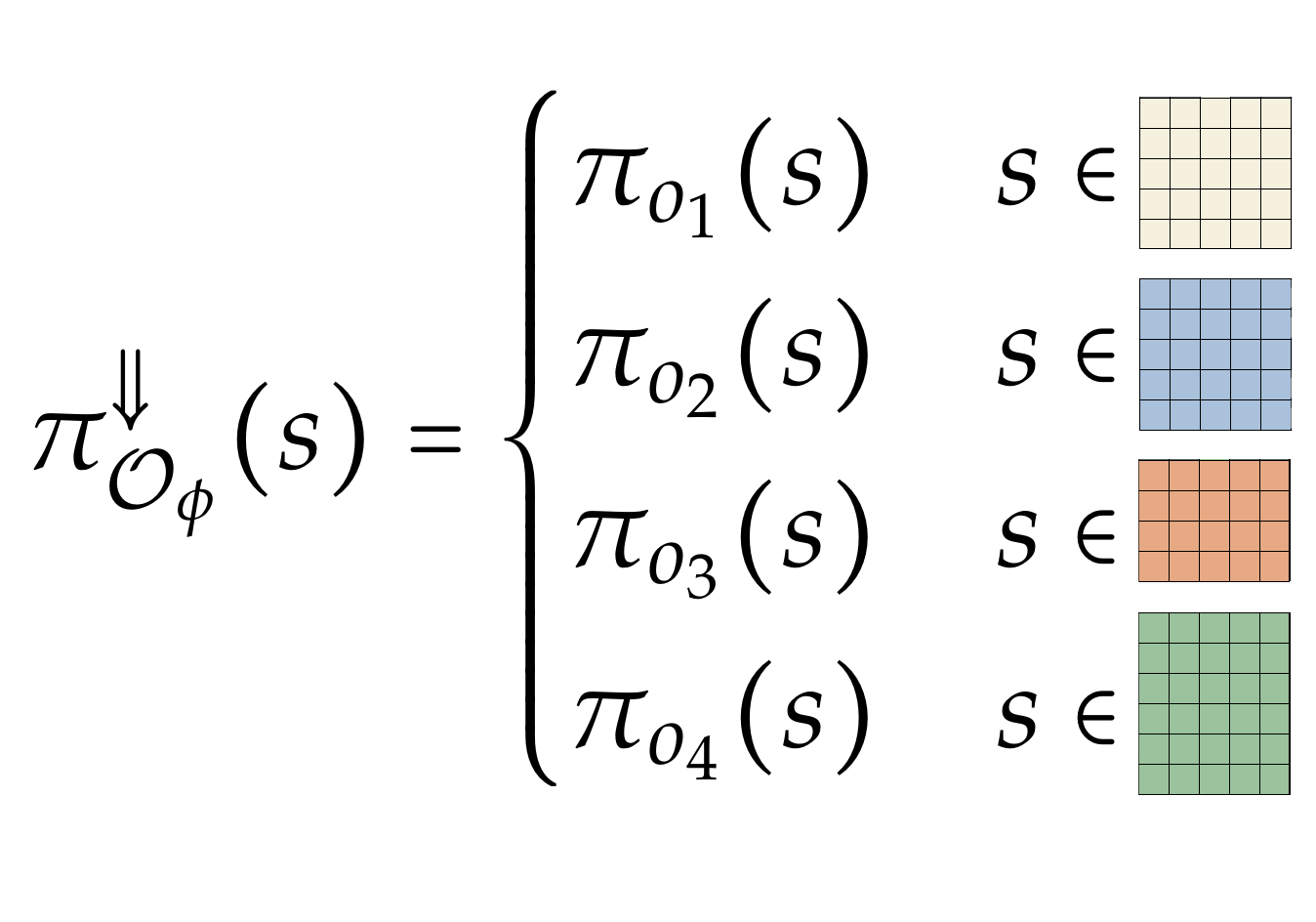}} 
    \caption{Grounding policy $\pi_{\mc{O}_\phi}$ to $\pi_{\mc{O}_\phi}^\Downarrow$.}
    \label{fig:c9_grounded_policy}
\end{figure}

\paragraph{Example.} Let us again consider the classical Four Rooms domain. Suppose that the state abstraction $\phi$ turns each room into an abstract state. Then any $\phi$-relative option in this domain is one that initiates anywhere in one of the rooms and terminates as soon as the agent leaves that room, as pictured in \autoref{fig:c9_four_rooms}. The only degree of flexibility in grounding a set of $\phi$-relative options for the given $\phi$, then, is which \textit{policies} are associated with each option, and how many options are available in each abstract state. If, for instance, the optimal policy $\pi^*$ were chosen for an option in the top right room, but the uniform random policy were available everywhere else, how might that impact the overall suboptimality of the policies induced by the abstraction? I now build toward the main result of the chapter (\autoref{thm:c9_vpsa_main_result}) that clarifies the precise conditions under which $\eps$-optimal policies are representable under different $(\phiop)$ pairs.

As discussed in \autoref{chap:background}, the value loss of an abstraction captures the gap in ground value between the best ground policy and best abstract policy. While value loss has a straightforward definition for state abstraction, it is not so clear for action abstraction. To analyze the value loss of state-action abstraction pairs $(\phiop)$, I first show that any such pair gives rise to an abstract policy over $\mc{S}_\phi$ and $\mc{O}_\phi$ that induces a unique policy in the original MDP (over the entire state space). Critically, this property does not hold for arbitrary options due to their semi-Markovian nature.

\begin{remark}
Every deterministic policy defined over abstract states and $\phi$-relative options, $\pi_{\mc{O}_\phi} : \mc{S}_\phi \rightarrow \mc{O}_\phi$, induces a unique Markov policy in the ground MDP, $\pi^{\Downarrow}_{\mc{O}_\phi} : \mc{S} \rightarrow \Delta(\mc{A})$. We let $\Pi_{\mc{O}_\phi}$ denote the set of abstract policies representable by the pair $(\phiop)$, and $\Pi^{\Downarrow}_{\mc{O}_\phi}$ denote the corresponding set of policies in the original MDP.
\label{rmk:phio_policy}
\end{remark}

\input{proofs/c9/c9_grounding_phio_policy}

This remark gives us a means of translating a policy over $\phi$-relative options into a policy over the original state and action space, $\mc{S}$ and $\mc{A}$.
Consequently, it is possible to extend the notion of value loss studied in previous chapters to apply to a set of options paired with a state abstraction: every $(\phiop)$ pair yields a set of policies in the original MDP, $\Pi^{\Downarrow}_{\mc{O}_\phi}$. The value loss of $(\phiop)$ as then the value loss of the best policy in this set.

\ddef{$(\phiop)$-Value Loss}{\label{def:phio_val_loss}The \textbf{value loss} of $(\phiop)$ is the smallest degree of suboptimality achievable:
\begin{equation}
    L(\phiop) := \min_{\pi_{\mc{O}_\phi} \in \Pi_{\mc{O}_\phi}} \left|\left| V^* - V^{\pi^{\Downarrow}_{\mc{O}_\phi}}\right|\right|_\infty.
\end{equation}}
Note that this notion of value loss is \textit{not} well defined for options in general, since they induce a semi-MDP: there is no well-formed \textit{ground} value function of a policy over options, but rather, a semi-Markov value function. As a simple illustration, consider a ground state $s_g$, two options $o_1$ and $o_2$ (either of which could be executing in $s_g$), and a policy $\pi_{\phi,o}$ over abstract states and options. It could be that $o_1$ or $o_2$ is currently executing when $s_g$ is entered or that either option has just terminated, requiring $\pi_{\phi,o}$ to select a new option. Each of these three cases induces a distinct value $V^{\pi_{\phi,o}}(s_g)$ which is then difficult to distill into a single ground value function. This is a key reason to restrict attention to $\phi$-relative options, each of which retains structure that couples with the corresponding state abstraction $\phi$ to yield value functions in the ground MDP.

\subsection{Four Classes of Value Preserving State-Action Abstractions}

I now show how different classes of $\phi$-relative options can represent near-optimal policies. We define an option class by a predicate $\lambda : \mc{O}_\phi \mapsto \{0,1\}$, and say that a set of $\phi$-relative options $\mc{O}_\phi$ belongs to the class $\mc{O}_{\phi, \lambda}$ if and only if $\lambda(\mc{O}_\phi) = 1$.

I begin by summarizing the four new $\phi$-relative option classes, drawing inspiration from other forms of abstraction~\citep{dean1997model,ravindran2004approximate,li2006towards,jiang2015abstraction,abel2016near,nachum2018near} discussed in more detail in \autoref{sec:c2_state_abstr_prior_work} and \autoref{sec:action_abstr_survey}. For each class, I will refer to the \textit{optimal} option in $s_\phi$, $o^*_{s_\phi}$, as the $\phi$-relative option that initiates in $s_\phi$ and executes $\pi^*$ until termination. These classes were chosen as they closely parallel existing properties studied in the literature. The four classes are as follows:
\begin{enumerate}
    \item \textit{Similar $Q^*$ Functions:} In each $s_\phi$, there is at least one option $o$ that has similar $Q^*$ to $o^*_{s_\phi}$.
    \item \textit{Similar Models:} In each $s_\phi$, there is at least one option $o$ that has a similar multi-time model~\citep{precup1998multi} to $o^*_{s_\phi}$.
    \item \textit{Similar $k$-Step Distributions:} In each $s_\phi$, there is at least one option $o$ that has a similar $k$-step termination state distribution to $o^*_{s_\phi}$, based off the hierarchical construction introduced by~\citet{nachum2018near}. Loss bounds will only apply to goal-based MDPs.
    \item \textit{Approximate MDP Homomorphisms:} Any deterministic $\pi_{\mc{O}_\phi}$ can encode an MDP homomorphism. The MDP homomorphism option class is defined by a guarantee on the quality of the resulting homomorphism.
\end{enumerate}

I now present each class in full technical detail. As stated, the first two classes guarantee $\eps$ closeness of values and models respectively. More concretely:

\paragraph{Similar $Q^*$-Functions ($\mc{O}_{\phi,Q_\eps^*}$).} The $\eps$-similar $Q^*$ predicate defines an option class where:
\begin{equation}
    \lambda(\mc{O}_\phi) \equiv \numberthis  \forall_{s_\phi \in \mc{S}_\phi} \exists_{o \in \mc{O}_\phi} : \max_{s\in s_\phi} |Q_{s_\phi}^*(s, o^*_{s_\phi}) - Q_{s_\phi}^*(s, o)| \leq \eps_Q,
\end{equation}
where
\begin{equation}
\label{eq:c9_option_q_func}
    Q_{s_\phi}^*(s, o) := R(s,\pi_o(s)) + \gamma \sum_{s' \in \mc{S}} T(s' \mid s,\pi_o(s)) \left(\indic(s' \in s_\phi)Q_{s_\phi}^*(s', o) + \indic(s' \not \in s_\phi)V^*(s')\right).
\end{equation}
This $Q$-function describes the expected return of starting in state $s$, executing a $\phi$-relative option $o$ until leaving $\phi(s)$, then following the optimal policy thereafter. More generally, this class of $\phiop$ pairs captures all cases where each abstract state has at least one option that is useful. 
Note that the identity state abstraction paired with the degenerate set of options that exactly encodes the execution of each primitive action will necessarily be an instance of this class.

\paragraph{Similar Models ($\mc{O}_{\phi, M_\eps}$).} The $\eps$-similar $T$ and $R$ predicate defines an option class where:
\begin{equation}
    \lambda(\mc{O}_\phi) \equiv  \forall_{s_\phi \in \mc{S}_\phi} \exists_{o \in \mc{O}_\phi} : \left|\left|T_{s,o^*_{s_\phi}}^{s'} - T_{s,o}^{s'}\right|\right|_\infty \leq \eps_T\ \text{and}\
    \left|\left|R_{s,o^*_{s_\phi}} - R_{s,o}\right|\right|_\infty \leq \eps_R,
\end{equation}
where $R_{s,o}$ and $T_{s,o}^{s'}$ are shorthand for the reward model and multi-time model of~\citet{sutton1999between}. Roughly, this class states that there is at least one option in each abstract state that behaves similarly to the optimal option in that abstract state, $o_{s_\phi}^*$, throughout its execution in the abstract state. 


I next derive two classes of $\phi$-relative options based on abstraction formalisms from existing literature. The first is based on the hierarchical construction introduced by~\citet{nachum2018near}, while the second shows that $\phi$-relative options can describe an MDP homomorphism~\citep{ravindran2004approximate}.

\paragraph{Similar k-Step Distributions ($\mc{O}_{\phi, \tau}$).} Let $p(s', k \mid s, o)$ denote the probability of option $o$ terminating in $s'$ after $k$ steps, given that it initiated in $s$. We define this class by the following predicate:
\begin{equation}
    \lambda(\mc{O}_\phi) \equiv  \forall_{s_\phi \in \mc{S}_\phi} \exists_{o \in \mc{O}_\phi} \forall_{k \in \mathbb{N}} : \max_{s\in s_\phi, s' \in \mc{S}}   |p(s', k \mid s, o^*_{s_\phi}) - p(s', k \mid s, o)| \leq \tau.
\end{equation}

Intuitively, this class of $\mc{O}_\phi$ states that in each abstract state, there exists an option that can induce sufficiently similar $k$-step state distributions to executing the optimal option in that abstract state, $o_{s_\phi}^*$.

\paragraph{Approximate MDP Homomorphisms ($\mc{O}_{\phi, H}$).} As discussed in \autoref{chap:background}, MDP homomorphisms define mappings from one MDP to another in a way that preserves desirable properties \cite{ravindran2002model}. The main idea behind these mappings, as with state and action abstraction, is to identify symmetries in the underlying environmental MDP that can be expressed through a simpler model than the original MDP. An approximate MDP homomorphism extends this notion of equivalence to similarity, thereby allowing greater opportunity to compress \cite{ravindran2004approximate}. To define this class of $\phi$-relative options, we first define the one-step abstract transition and reward functions for a $\phi$-relative option $o$. That is, for $w : \mc{S} \ra [0,1]$ any valid weighting function such that $\sum_{s \in s_\phi} w(s) = 1$:
\begin{align}
    T_\phi (s_\phi' \mid s_\phi, o) &= \sum_{s \in s_\phi} w(s) \sum_{s' \in s_\phi'} T(s' \mid s, \pi_o(s)), \\ R_\phi(s_\phi, o) &= \sum_{s \in s_\phi} w(s) R(s, \pi_o(s)).
\end{align}
Next, we introduce the quantities $K_p$ and $K_r$ of~\citet{ravindran2004approximate}:
\begin{align}
    K_p &= \max_{s \in \mc{S}, a \in \mc{A}} \sum_{s_\phi \in \mc{S}_\phi} \Big| \sum_{s' \in s_\phi} T(s' \mid s, a) - T_\phi(s_\phi \mid \phi(s), \pi_{\mc{O}_\phi}(\phi(s))) \Big|, \\
    K_r &= \max_{s \in \mc{S}, a \in \mc{A}} |R(s, a) - R_\phi(\phi(s), \pi_{\mc{O}_\phi}(\phi(s))) |. \numberthis
\end{align}

These capture the maximum discrepancy between the model of the ground MDP and the model of the induced abstract MDP defined according to $(\phiop)$. Using these quantities, the class of $\phi$-relative options is defined as follows.
\begin{equation}
    \lambda(\mc{O}_\phi) \equiv \forall\ \pi_{\mc{O}_\phi} \in \Pi_{\mc{O}_\phi} : K_p \leq \eps_p \; \text{and}\; K_r \leq \eps_r.
\end{equation}

These four classes will constitute four sufficient conditions for $(\phiop)$ pairs to yield bounded value loss.

\subsection{Main Result}

The main result of this chapter establishes the bounded value loss of pairs $(\phiop)$ where $\mc{O}_\phi$ belongs to any of these four classes, and the size of the bound depends on the degree of approximation ($\eps_Q$; $\eps_R$, $\eps_T$; $\tau$; and $\eps_r$, $\eps_p$).
\begin{theorem}
\label{thm:c9_vpsa_main_result}
(Main Result) For any $\phi$, the four introduced classes of $\phi$-relative options satisfy:
\begin{align}
    &L(\phi, \mc{O}_{\phi, Q_\eps^*}) \leq \frac{\eps_Q}{1-\gamma}, \\
    &L(\phi, \mc{O}_{\phi, M_\eps}) \leq \frac{\eps_{R} + |\mc{S}| \eps_{T} \textsc{RMax}}{(1-\gamma)^2}, \\
    &L(\phi, \mc{O}_{\phi, \tau}) \leq \frac{\tau \gamma |\mc{S}|}{(1-\gamma)^2}, \\
    &L(\phi, \mc{O}_{\phi, H}) \leq \frac{2}{1-\gamma} \left( \eps_r + \frac{\gamma\textsc{RMax}}{1 - \gamma} \frac{\eps_p}{2}\right), 
\end{align}
where the $L(\phi, \mc{O}_{\phi, \tau})$ bound holds in goal-based MDPs and the other three hold in any finite MDP.
\end{theorem}

\input{proofs/c9/c9_main_result}

Observe that when the approximation parameters are zero, many of the bounds collapse to $0$ as well. This illustrates the trade off made between the amount of knowledge used to construct the abstractions and the degree of optimality ensured, as was the case with approximate state abstractions in \autoref{chap:approx_state_abstr}. In a sense, this result is the spiritual successor to \autoref{thm:c3_approx_sa_main_result}, extended to options. Further note that the value loss of the state abstraction does not appear in any of the above bounds---indeed, $\phi$ will \textit{implicitly} affect the value loss as a function of the diameter of each abstract state. Finally, observe that, as with \autoref{thm:c3_approx_sa_main_result}, each of the above classes expresses a \textit{sufficient} condition needed for a pair $(\phi, \mc{O}_\phi)$ to preserve value.

It is useful, however, to identify not just sufficient conditions, but also necessary. To this end, we next establish one necessary condition of all (globally) value preserving $(\phiop)$ classes.
%
\begin{theorem}
\label{thm:c9_vpsa_nec_cond}
For any $(\phiop)$ pair with $L(\phiop) \leq \eta$, there exists at least one option per abstract state that is $\eta$-optimal in $Q$-value. Precisely, if $L(\phiop) \leq \eta$, then:
\begin{equation}
    \forall_{s_\phi \in \mc{S}_\phi} \forall_{s \in s_\phi} \exists_{o \in \mc{O}_\phi} : Q^*_{s_\phi}(s, o^*_{s_\phi}) - Q^*_{s_\phi}(s, o) \leq \eta.
\end{equation}
\end{theorem}

\input{proofs/c9/c9_necessary_condition}

This theorem tells us that for any agent acting using these join state-action abstractions, if there exists an abstract state for which there is not an $\eta$-optimal option, then the agent cannot represent a globally near-optimal policy.

\subsection{Experiment}
\label{sec:branch_experiment}
I next conduct a simple experiment to test whether value preserving options enable simple RL algorithms to find near-optimal policies. The experiment illustrates an important property of one of the introduced $\phiop$ classes, and is organized as follows. First, I construct a $\phiop$ pair belonging to the $\mc{O}_{\phi,Q_\eps^*}$ class using dynamic programming. I give this pair to one of four different RL algorithms: $Q$-learning \cite{watkins1992q}, SARSA \cite{rummery1994line}, Double $Q$-learning \cite{hasselt2010double}, and R-Max \cite{brafman2002rmax}. For each algorithm, I vary the number of interactions it is allowed to have with the environment, $N$, ranging from $N=100$ to $N=1,000,000$. As expected, the environment is the Four Rooms MDP with a single goal location in the top right and start location in the bottom left. The state abstraction $\phi$ maps each state into one of four abstract states, denoting each of the four rooms. I vary both the number of options added per abstract state ($|\mc{O}_\phi|$) and the sample budget given to each algorithm ($N$), and present the value of the policy discovered by the final episode for each setting of $|\mc{O}_\phi|$ and $N$. The code is publicly available for extension and reproduction. \footnote{\url{https://github.com/david-abel/vpsa_aistats2020}}

\begin{figure}[t!]
    \centering
    \subfloat[$Q$-learning]{\includegraphics[width=0.45\textwidth]{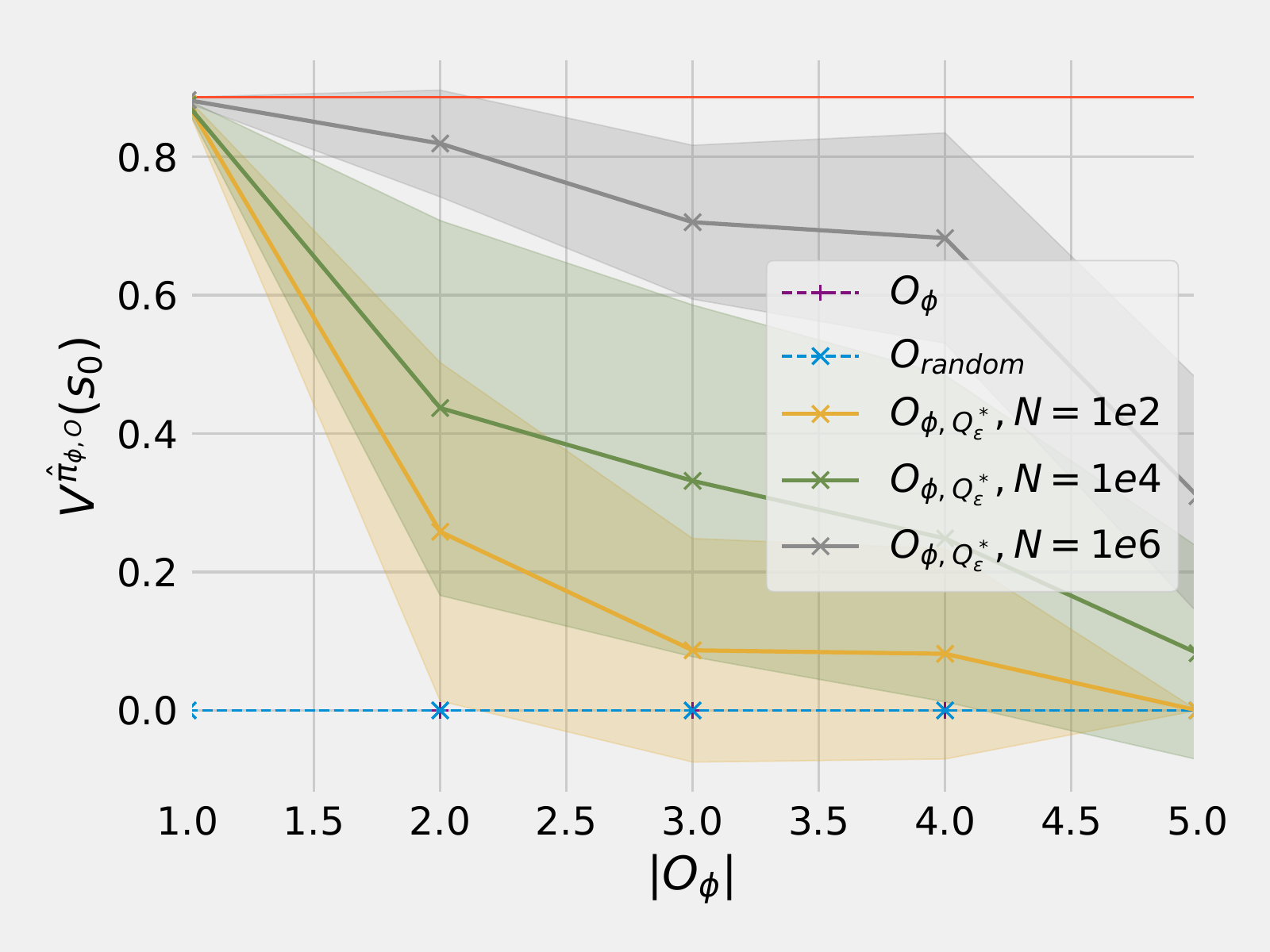}} \subfhspace
    \subfloat[SARSA]{\includegraphics[width=0.45\textwidth]{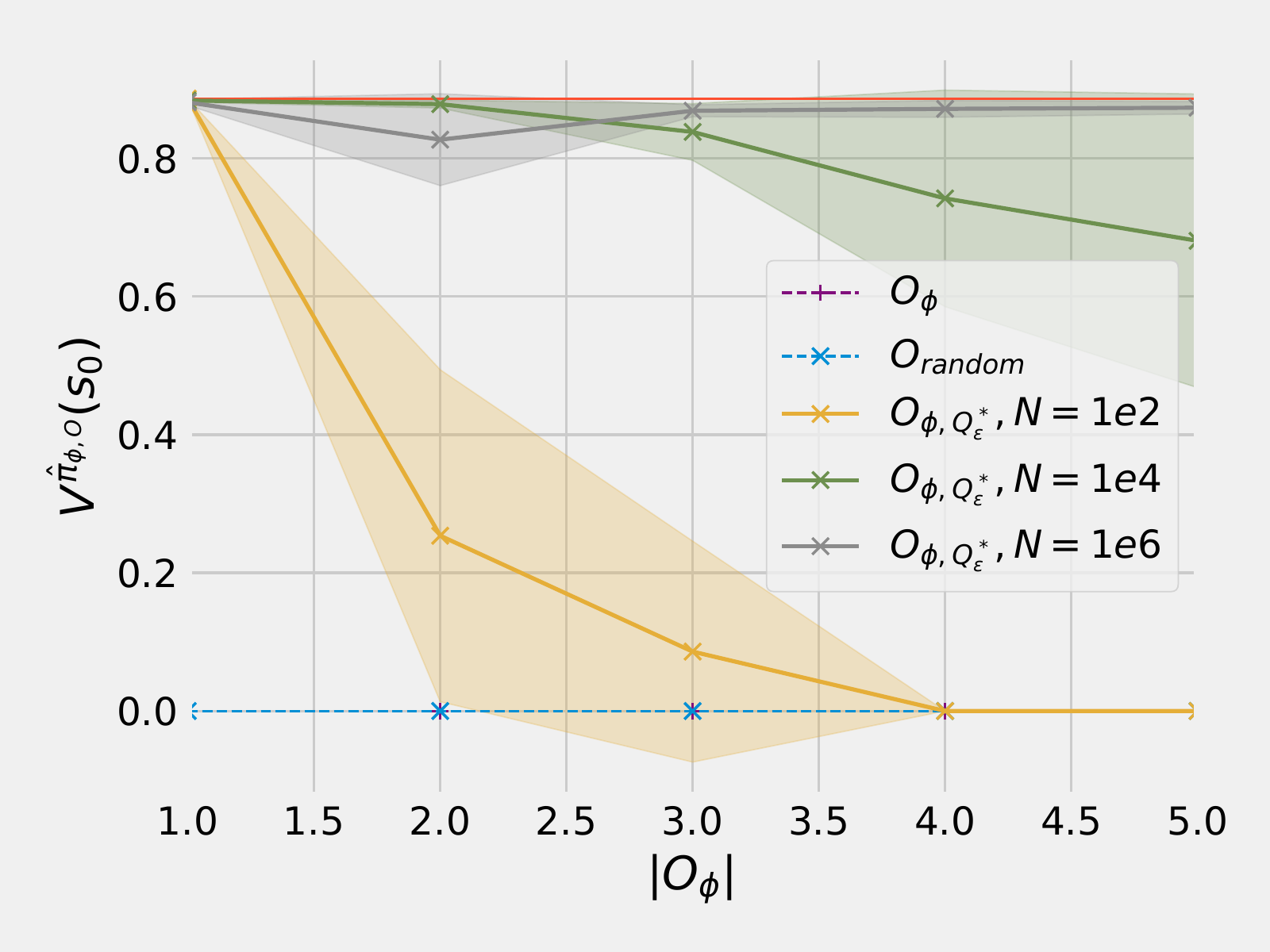}} \\
    \subfloat[Double $Q$]{\includegraphics[width=0.45\textwidth]{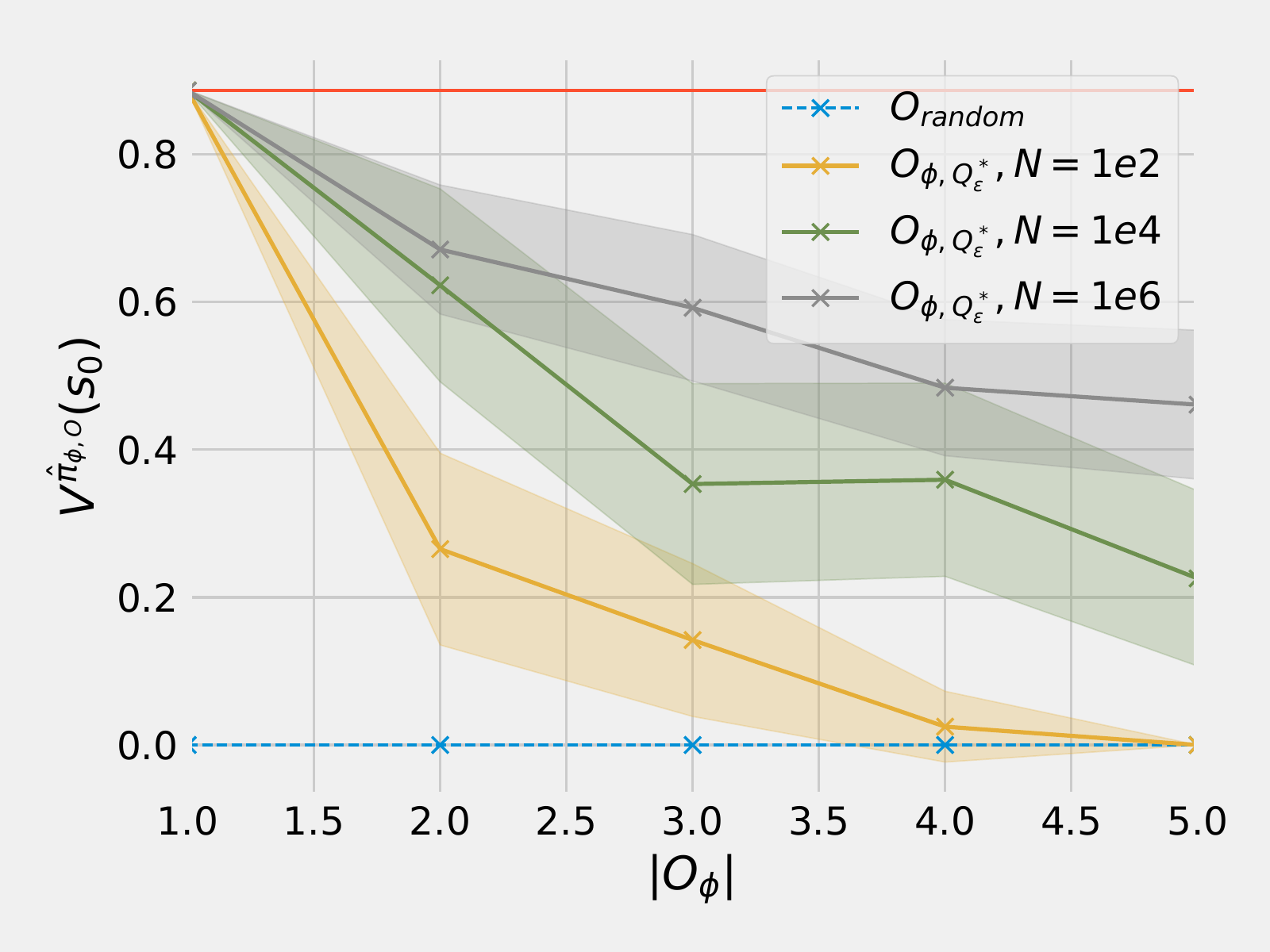}} \subfhspace
    \subfloat[R-Max]{\includegraphics[width=0.45\textwidth]{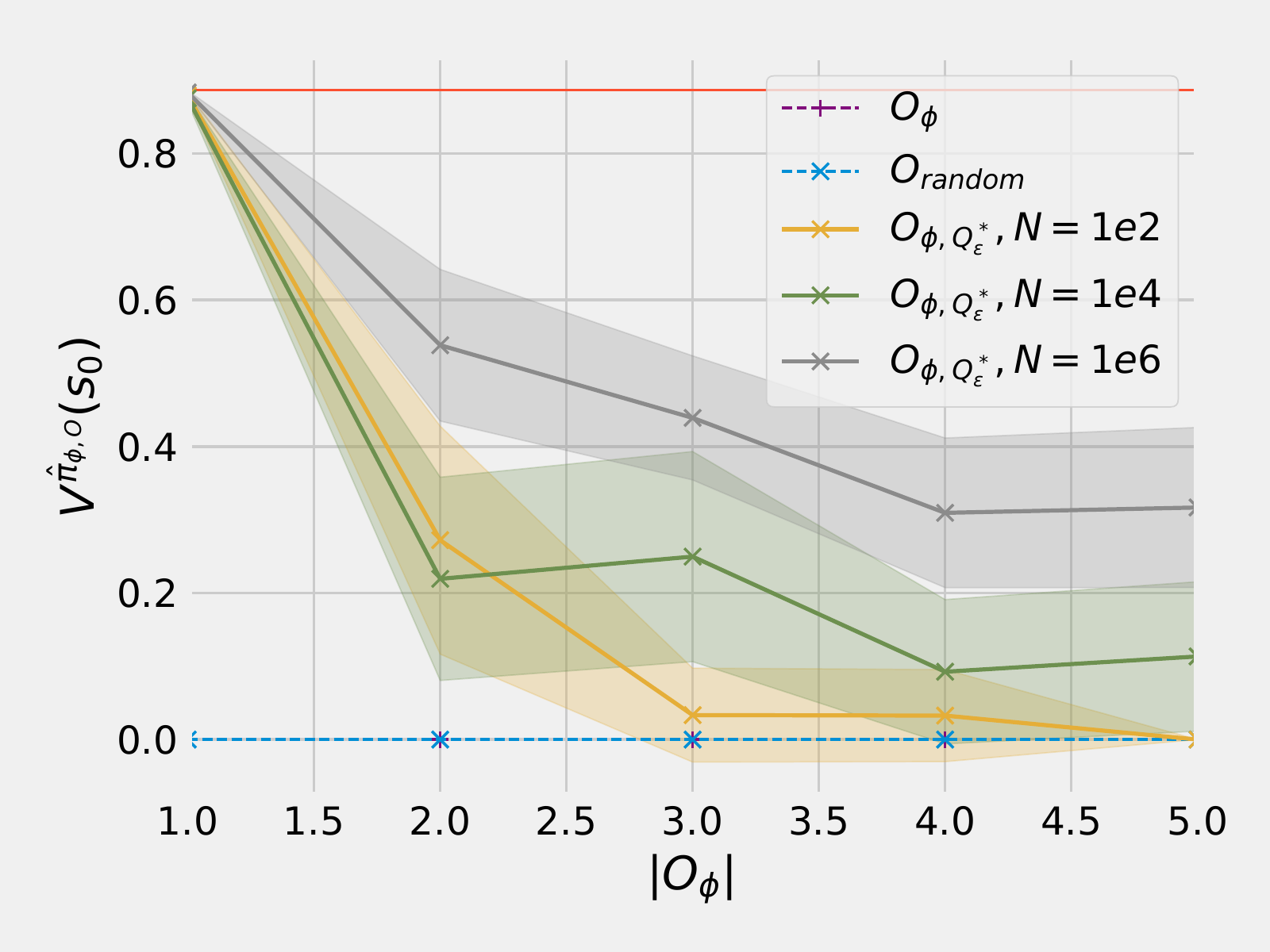}}
    \caption{Empirical evidence that the $\phiop$ pairs from Theorem \ref{thm:c9_vpsa_main_result} preserve value.}
    \label{fig:c9_branching_factor}
\end{figure}

\begin{figure}[b!]
    \centering
    \includegraphics[width=0.7\textwidth]{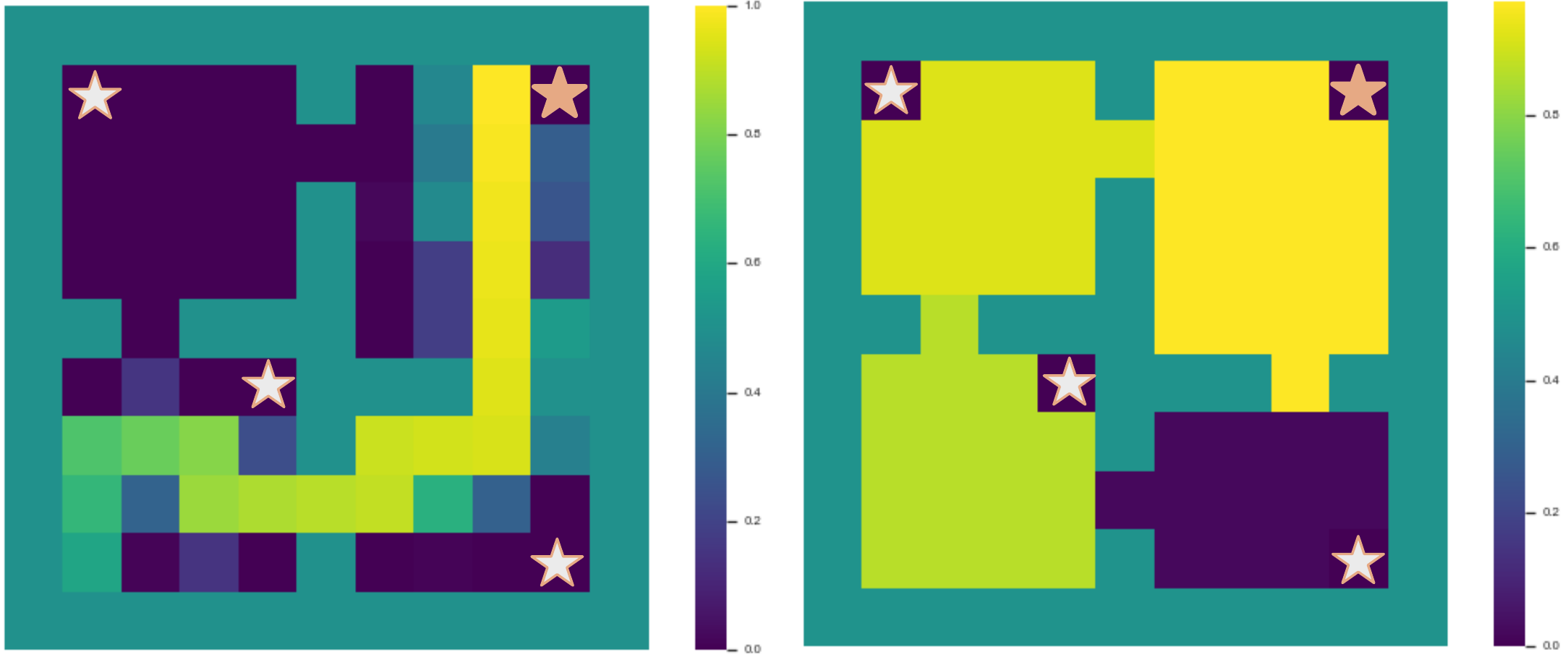}
    \caption{Comparison of the learned value function with regular $Q$-learning (left) and $Q$-learning with $\phiop$.}
    \label{fig:c9_multi_task_viz}
\end{figure}

Results are presented in \autoref{fig:c9_branching_factor}. First, note that with only one option per abstract state, all four algorithms can trivially find a near-optimal policy, even with a small sample budget. This is predicted by \autoref{thm:c9_vpsa_main_result}: the included options preserve value, and so any assignment of options to abstract states will yield a near-optimal policy. In contrast, if randomly chosen options are used instead (shown in blue, labeled as $O_{\text{random}}$), the learning algorithm fails to find a good policy even with a high sample budget ($N=1e6$ was used). Second, we find that as the number of options increases, the added branching factor causes each algorithm to find a lower value policy with the same number of samples. However, by \autoref{thm:c9_vpsa_main_result} we know each set of options preserves value; as the sample budget increases we see that the value of the discovered policy tends toward optimal in each algorithm. For SARSA, for instance, there is a dramatic difference between the lowest setting of $N$ and the higher two settings. In short, the $\phiop$ pairs defined by \autoref{thm:c9_vpsa_main_result} do in fact preserve value, but will also affect the sample budget required to find a good policy, with the exact extent changing depending on the RL algorithm. I foresee the combination of value preserving abstractions with those that lower learning complexity (see recent work by \citet{brunskill2014pac,fruit2017regret}) as a key direction for future work.

I further visualize the learned value function of $Q$-learning with and without $\phi$-relative options after the same sample budget, depicted in \autoref{fig:c9_multi_task_viz}. Notably, since $\phi$-relative options update entire blocks of states, we see large regions of the state space with the same learned value function. Conversely, $Q$-learning only tends to explore (and estimate the values of) a narrow region of the state space. The visual highlights this important qualitative difference between learning with and without action abstractions.

\section{Hierarchical Abstraction}
\label{sec:hierarch}

\begin{figure}[b!]
    \centering
    \includegraphics[width=0.65\textwidth]{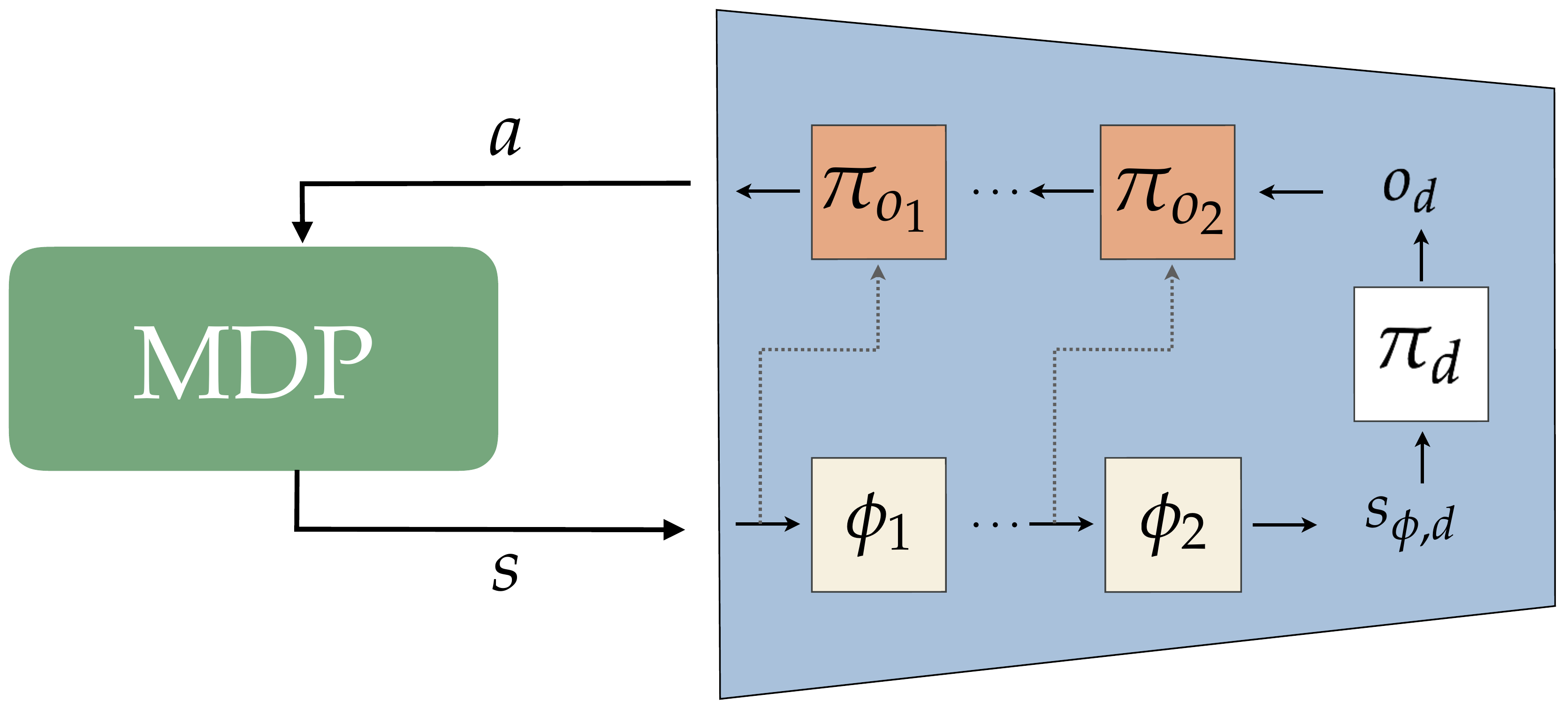}
    \caption{The construction of a hierarchy from $(\phiop)$ pairs.}
    \label{fig:c9_hierarch_depth}
\end{figure}

I now present an extension of \autoref{thm:c9_vpsa_main_result} that applies to hierarchies consisting of $(\phiop)$ pairs. I prove that the value loss compounds linearly if we are to construct a hierarchy using algorithms that generate a well-behaved $\phi$ and $\mc{O}_\phi$. To do so, we require two definitions and additional notation. We first define a \textit{hierarchy} as $n$ sets of $(\phiop)$ pairs, as pictured in \autoref{fig:c9_hierarch_depth}.

\ddef{$(\phiop)$-Hierarchy}{A \textbf{depth $\bs{n}$ hierarchy}, denoted $H_n$, is a list of $n$ state abstractions, $\phi^{(n)}$, and a list of $n$ sets of $\phi$-relative options, $\mc{O}^{(n)}_\phi$. The components $(\mc{I}, \beta, \pi)$ of each of the $i$-th set of options, $\mc{O}_{\phi,i}$ are defined over the $(i-1)$-th abstract state space,
\begin{equation}
    \mc{S}_{\phi,i-1}= \{\phi_{i-1}(\phi_{i-2}( \ldots \phi_1(s) \ldots )) : s \in \mc{S}\}.
\end{equation}}
I next introduce additional notation to refer to values, states, options, and policies at each level of the hierarchy. Let $\pi_n : \mc{S}_{\phi,n} \rightarrow \mc{O}_{\phi,n}$ denote the level $n$ policy encoded by the hierarchy, with $\Pi_n$ the space of all policies encoded in this way. I let $s_i$ denote shorthand for $s_{\phi,i} := \phi^i(s) = \phi_i( \ldots \phi_1(s) \ldots )$, with $s$ a state in the ground MDP. I further denote $V_i$ as the $i$-th level's value function, defined as follows for some ground state $s$.
\begin{equation}
        V_i^\pi(s) = \max_{o \in \mc{O}_i} \left(R_i\left(s_i,o\right) + \sum_{s' \in \mc{S}_i} T_i\left(s' \mid s_i, o\right) V_i^\pi(s')\right),
\end{equation}
where,
\begin{align}
    R_i(s_i, o) &:= \sum_{s_{i-1} \in s_i} w_i(s_{i-1}) R_{s_{i-1}, o},\\
    T_i(s_i' \mid s_i, o) &:= \sum_{s_{i-1} \in s_i} \sum_{s_{i-1}' \in \mc{S}_{i-1}} w_i(s_{i-1})  T_{s_{i-1}, o}^{s_{i-1}'}.
\end{align}
Again, $R_{s,o}$ and $T_{s,o}^{s'}$ are defined according to the multi-time model~\citep{sutton1999between}, $s_i \in \mc{S}_{\phi, i}$ is a level $i$ state resulting from $\phi^i(s)$, and $w_i$ is an aggregation weighting function for level $i$. Note that $V_0$ is the ground value function, which we refer to as $V$ for simplicity. The full list of notation for this section is presented in \autoref{tab:hierarch_abstr_notation}.

\subsection{Hierarchy Analysis}
I now extend \autoref{thm:c9_vpsa_main_result} to hierarchies of fixed but arbitrary depth, building on two key observations. First, any policy $\pi_n$ represented at the top level of a hierarchy $H_n$ also has a unique Markov policy in the ground MDP, which we denote $\pi_n^{\Downarrow}$ (in contrast to $\pi_n^{\downarrow}$, which moves the level $n$ policy to level $n-1$). 
I summarize this fact in the following remark:

\begin{remark}
Every deterministic policy $\pi_i$ defined by the $i$-th level of a hierarchy, $H_n$, induces a unique policy in the ground MDP, which we denote $\pi_i^{\Downarrow}$.
\end{remark}

\begin{figure}[h!]
    \centering
    \includegraphics[width=0.7\textwidth]{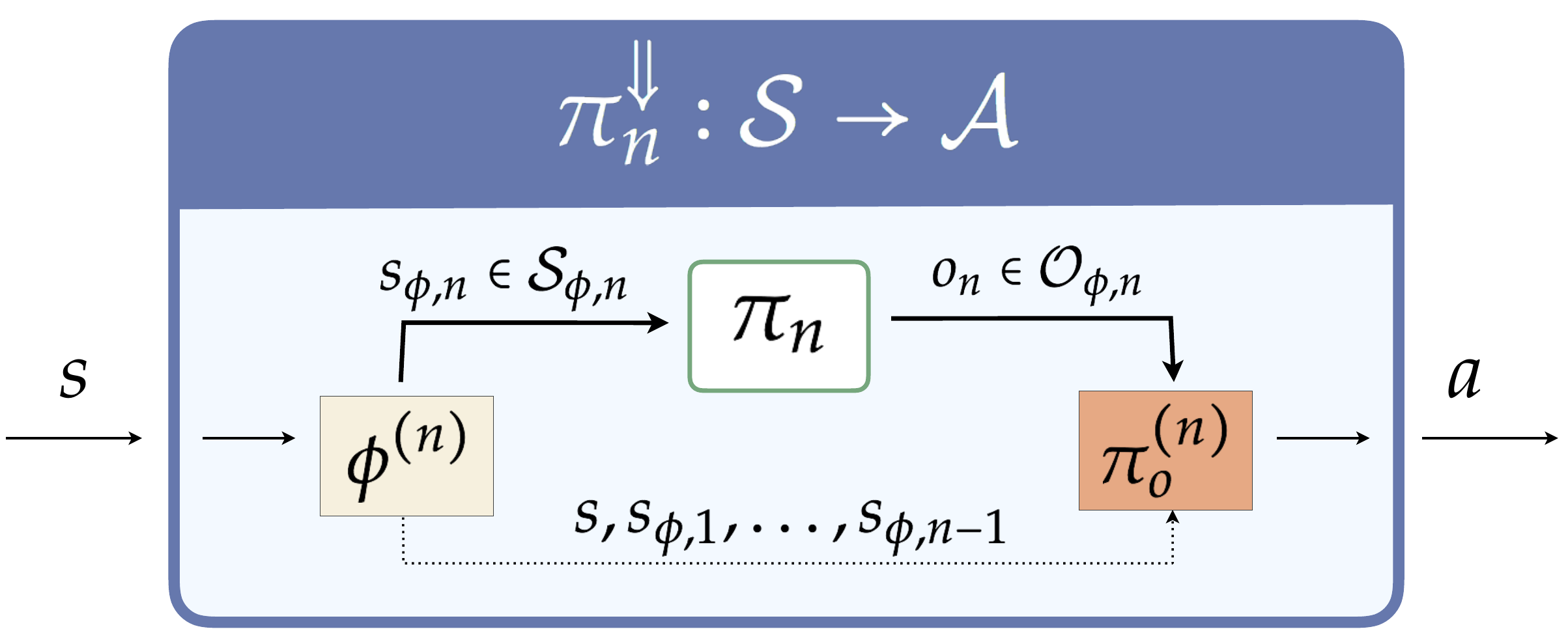}
    \caption{The process of grounding a hierarchical policy.}
    \label{fig:c9_hierarch_policy_grounding}
\end{figure}

To be precise, note that $\pi_i^\downarrow$ specifies the level $i$ policy $\pi_i$ mapped into level $\pi_{i-1}$, whereas $\pi^\Downarrow_i$ refers to the policy at $\pi_i$ mapped into $\pi_0$. The process of forming this ground policy $\pi^\Downarrow_n$ from a policy at the top level of the hierarchy $\pi_i$, is pictured in \autoref{fig:c9_hierarch_policy_grounding}.

\newpage
\vspace*{\fill}

\input{figures/tables/sa_aa_notation_table}
\vspace*{\fill}
\newpage

The second key insight is that the value loss of $(\phiop)$ pairs applies in a straightforward way to hierarchies, $H_n$.

\ddef{$H_n$-Value Loss}{The \textbf{value loss} of a depth $n$ hierarchy $H_n$ is the smallest degree of suboptimality across all policies representable at the top level of the hierarchy:
\begin{equation}
    L(H_n) := \min_{\pi_n \in \Pi_n} \left|\left|V^* - V^{\pi^{\Downarrow}_n}\right|\right|_\infty.
\end{equation}
}
This quantity denotes how suboptimal the best hierarchical policy is in the ground MDP. Therefore, the guarantee we present expresses a condition on \textit{global} optimality rather than \textit{recursive} or \textit{hierarchical} optimality \citep{dietterich2000hierarchical}.

I next show that there exist value-preserving hierarchies by bounding the above quantity for well constructed hierarchies. To prove this result, we require two assumptions.
\begin{assumption}
\label{asmpt:hierarchy_value}
The value function is consistent throughout the hierarchy. That is, for every level of the hierarchy $i \in \mathbb{N}_{\leq n-1}$, for any policy $\pi_i$ over states $\mc{S}_{\phi, i}$ and options $\mc{O}_{\phi, i}$, there is a small $\kappa \in \mathbb{R}_{\geq 0}$ such that: 
\begin{equation}
\max_{s \in \mc{S}} \left|V_{i-1}^{\pi^\downarrow_i}\left(\phi^{i-1}(s)\right) - V_i^{\pi_i}\left(\phi^{i}(s)\right)\right| \leq \kappa
\end{equation}
\end{assumption}
\begin{assumption}
\label{asmpt:vi_eps_options}
Subsequent levels of the hierarchy can represent policies similar in value to the best policy at the previous level. That is, for every $i \in \mathbb{N}_{\leq n-1}$, letting $\pi_i^\diamond = \argmin_{\pi_i \in \Pi_i} ||V_0^* - V_0^{\pi^\Downarrow_i}||_\infty$, there is a small $\ell \in \mathbb{R}_{\geq 0}$ such that:
\begin{equation}
    \min_{\pi^\downarrow_{i+1} \in \Pi^\downarrow_{i+1}}\left|\left|V_i^{\pi_i^\diamond} - V_i^{\pi^\downarrow_{i+1}}\right|\right|_\infty \leq \ell.
\end{equation}
\end{assumption}

It is likely that both assumptions are true given the right choice of state abstractions, options, and methods of constructing abstract MDPs. As some motivating evidence, a claim closely related to \autoref{asmpt:hierarchy_value} was proven in \autoref{chap:approx_state_abstr} as \autoref{eq:Q*Claim1}, and \autoref{asmpt:vi_eps_options} is of similar structure to ~\autoref{thm:c9_vpsa_main_result}. 
These two assumptions (along with \autoref{thm:c9_vpsa_main_result}) give rise to hierarchies that can represent near-optimal behavior.

\begin{theorem}
\label{thm:c9_hierarch_loss}
Consider two algorithms: 1) $\mathscr{A}_\phi$: given an MDP $M$, outputs a $\phi$, and 2) $\mathscr{A}_{\mc{O}_\phi}$: given $M$ and a $\phi$, outputs a set of options $\mc{O}_\phi$ such that there are constants $\kappa$ and $\ell$ for which \autoref{asmpt:hierarchy_value} and~\autoref{asmpt:vi_eps_options} are satisfied. Then, by repeated application of $\mathscr{A}_\phi$ and $\mathscr{A}_{\mc{O}_\phi}$, we can construct a hierarchy of depth $n$ such that
\begin{equation}
    L(H_n) \leq n (\kappa + \ell).
\end{equation}
\end{theorem}

\input{proofs/c9/c9_hierarch_loss}

This theorem offers a clear path for extending the guarantees of $\phi$-relative options beyond the typical two-timescale setup observed in recent work~\citep{bacon2017option,nachum2018near} to fully realize the benefits of (multi-level) hierarchical abstraction. Moreover, both \autoref{asmpt:hierarchy_value} and \autoref{asmpt:vi_eps_options} are sufficient---together with $\phi$-relative options that satisfy \autoref{thm:c9_vpsa_main_result}---to construct a hierarchy with low value loss. One conclusion to draw is that algorithms for leveraging hierarchies may want to explicitly search for structures that satisfy our assumptions: 1) value function smoothness up and down the hierarchy, and 2) policy richness at each level of the hierarchy.

I have here proven which state-action abstractions are guaranteed to preserve representation of near-optimal policies. To do so, I introduced $\phi$-relative options, a simple but expressive formalism for combining state abstractions with options. Under this formalism, I proposed four classes of $\phi$-relative options with bounded value loss. Lastly, I proved that under mild conditions, pairs of state-action abstractions can be recursively combined to induce hierarchies that also possess near-optimality guarantees.

I take these results, along with those established in \autoref{part:state_abstraction} and \autoref{part:action_abstraction}, to serve as a concrete path toward principled abstraction discovery and use in RL.

%% file: proofs/c9/c9_grounding_phio_policy.tex
\begin{dproof}[Remark \ref{rmk:phio_policy}]
Consider an arbitrary deterministic policy $\pi_{\mc{O}_\phi}$. By definition, this policy assigns one option to each abstract state. Let $\mc{O}_\pi$ denote the set of options this policy assigns. \\

By construction of $\phi$-relative options, for every ground state $s \in \mc{S}$ there is one unique option $o_{\phi(s)} \in \mc{O}_\pi$ that can be executed in $s$. \\

Therefore, we construct a policy $\pi_{\mc{O}_\phi}^\Downarrow$ as the combination of option policies in $\mc{O}_\pi$. Specifically, letting $\pi_{o_{\phi(s)}}$ denote the option policy of the option in $\mc{O_\pi}$ that is assigned to $\phi(s)$:
\begin{equation}
    \pi_{\mc{O}_\phi}^\Downarrow(s) = \pi_{o_{\phi(s)}}(s)
\end{equation}
\end{dproof}

%% file: proofs/c9/c9_main_result.tex
\begin{dproof}[Theorem \ref{thm:c9_vpsa_main_result}]
We prove this claim using four separate proofs, each targeting one class.

\input{proofs/c9/c9_phio_q_eps_loss}
\spacerule

\input{proofs/c9/c9_phio_model_loss}
\spacerule

\input{proofs/c9/c9_phio_k_step_loss}
\spacerule

\input{proofs/c9/c9_phio_homomorphism_loss}
\spacerule

Having proven the value loss for each $\mc{O}_\phi$ class, the result follows. \qedhere

\end{dproof}

%% file: proofs/c9/c9_phio_q_eps_loss.tex
\begin{proof}
    ($L(\phi, \mc{O}_{\phi, Q_\eps^*}) \leq \frac{\eps_Q}{1-\gamma}$) \\

    Consider $L(\phi, \mc{O}_{\phi, Q_\eps^*}) = \min_{\pi^\Downarrow_{\mc{O}_\phi} \in \Pi^\Downarrow_{\mc{O}_\phi}} \max_{s \in \mc{S}} |V^*(s) - V^{\pi^\Downarrow_{\mc{O}_\phi}}(s)|$. Since $V^*(s) \geq V^{\pi}(s)$ for all $\pi$, we henceforth drop the absolute value for convenience. \\
    
    To proceed, we recall that $o_{s_\phi}^*$ is the $\phi$-relative option that executes $\pi^*$ in every state and terminates when it leaves the abstract state $s_\phi$:
    \begin{align}
        o_{s_\phi}^* := \forall_{s \in \mc{S}} :\ &(\mc{I}(s) \equiv \phi(s) = s_\phi, \nonumber \\
                &\beta(s) \equiv \phi(s) \neq s_\phi, \nonumber \\
                &\pi(s) = \pi^*(s)).
    \end{align}
    Note that since $o_{s_\phi}^*$ always chooses actions according to $\pi^*$, that $Q_{s_\phi}^*(s, o_{s_\phi}^*) = V^*(s)$ (where $Q_{s_\phi}^*$ is defined according to \autoref{eq:c9_option_q_func}).\\
    
    Then, by the $Q_\eps^*$ predicate, we can construct a policy over abstract states and options $\mu_{\mc{O}_\phi} \in \Pi_{\mc{O}_\phi}$ with the following property:
    \begin{equation}
        \forall_{s_\phi \in \mc{S}_\phi, s \in s_\phi} : Q_{s_\phi}^*(s, o_{s_\phi}^*) - Q_{s_\phi}^*(s, \mu_{\mc{O}_\phi}(s_\phi)) \leq \eps_Q.
        \label{eq:mu_qstar_diff}
    \end{equation}
    Note that $\mu_{\mc{O}_\phi}(s_\phi)$ outputs an option. As in \autoref{eq:mu_qstar_diff}, we henceforth denote $s_\phi = \phi(s)$ and correspondingly $s_\phi' = \phi(s')$. \\
    
    Then it must be the case that
    \begin{equation}
        L(\phi, \mc{O}_{\phi, Q_\eps^*}) \leq \max_{s \in \mc{S}} V^*(s) - V^{\mu^\Downarrow_{\mc{O}_\phi}}(s).
    \end{equation}
    Let $Q_t^*(s, o)$ denote the expected discounted reward of executing option $o$, then executing $t$ options under $\mu_{\mc{O}_\phi}$, then following the optimal policy thereafter. Note that
    \begin{equation}
        \lim_{t \to \infty} Q_t^*(s, \mu_{\mc{O}_\phi}(s_\phi)) = V^{\mu^\Downarrow_{\mc{O}_\phi}}(s),
    \end{equation}
    because $Q_t^*(s, \mu_{\mc{O}_\phi}(s_\phi))$ is the expected discounted reward of executing $t+1$ options under $\mu_{\mc{O}_\phi}$, then following the optimal policy thereafter. \\
    
    We next show by induction on $t$ that 
    \begin{equation}
        \max_{s \in \mc{S}} V^*(s) - V^{\mu^\Downarrow_{\mc{O}_\phi}}(s) = \max_{s \in \mc{S}} \lim_{t \to \infty} V^*(s) - Q_t^*(s, \mu_{\mc{O}_\phi}(s_\phi)) \leq \frac{\eps_Q}{1 - \gamma}.
    \end{equation}
    In particular, we wish to show that
    \begin{equation}
        \forall_{t \in \mathbb{N}} : \max_{s \in \mc{S}} V^*(s) - Q_t^*(s, \mu_{\mc{O}_\phi}(s_\phi)) \leq \sum_{i=0}^{t} \eps_Q \gamma^i.
    \end{equation}
    
    \textbf{\textit{(Base Case)}} \\
    When $t = 0$, for all $s \in \mc{S}$,
    \begin{equation}
        Q_0^*(s, \mu_{\mc{O}_\phi}(s_\phi)) = Q_{s_\phi}^*(s, \mu_{\mc{O}_\phi}(s_\phi)),
    \end{equation}
    because both quantities represent the expected discounted reward of executing the option $\mu_{\mc{O}_\phi}(s_\phi)$ then following the optimal policy thereafter. It follows that
    \begin{align}
        \max_{s \in \mc{S}} V^*(s) - Q_0^*(s, \mu_{\mc{O}_\phi}(s_\phi)) &= \max_{s \in \mc{S}} V^*(s) - Q_{s_\phi}^*(s, \mu_{\mc{O}_\phi}(s_\phi)), \\
        &= \max_{s \in \mc{S}} Q_{s_\phi}^*(s, o_{s_\phi}^*) - Q_{s_\phi}^*(s, \mu_{\mc{O}_\phi}(s_\phi)), \\
        &\leq \eps_Q, \\
        &= \sum_{i=0}^{0} \eps_Q \gamma^0,
    \end{align}
    where the inequality holds by definition of $\mu_{\mc{O}_\phi}$.\\
    
    \textbf{\textit{(Inductive Case)}}\\
    We assume as the inductive hypothesis that 
    \begin{equation}
        \max_{s \in \mc{S}} V^*(s) - Q_k^*(s, \mu_{\mc{O}_\phi}(s_\phi)) \leq \sum_{i=0}^{k} \eps_Q \gamma^i,
    \end{equation}
    and want to show that
    \begin{equation}
        \max_{s \in \mc{S}} V^*(s) - Q_{k+1}^*(s, \mu_{\mc{O}_\phi}(s_\phi)) \leq \sum_{i=0}^{k+1} \eps_Q \gamma^i.
    \end{equation}
    
    To begin, fix $s \in \mc{S}$ and consider
    \begin{align}
        &V^*(s) - Q_{k+1}^*(s, \mu_{\mc{O}_\phi}(s_\phi)) \\
        &= V^*(s) - \left( R_o(s, \mu_{\mc{O}_\phi}(s_\phi)) + \sum_{s' \in \mc{S}} T_o(s' \mid s, \mu_{\mc{O}_\phi}(s_\phi)) Q_k^*(s', \mu_{\mc{O}_\phi}(s_\phi')) \right) \\
        &= V^*(s) -  R_o(s, \mu_{\mc{O}_\phi}(s_\phi)) - \sum_{s' \in \mc{S}} T_o(s' \mid s, \mu_{\mc{O}_\phi}(s_\phi)) Q_k^*(s', \mu_{\mc{O}_\phi}(s_\phi'))
    \end{align}
    where $R_o$ and $T_o$ indicate the reward and multi-time models.
    
    Now, subtract and add $\sum_{s' \in \mc{S}} T_o(s' | s, \mu_{\mc{O}_\phi}(s_\phi)) V^*(s')$:
    \begin{eqnarray}
        =  & & V^*(s) -  R_o(s, \mu_{\mc{O}_\phi}(s_\phi)) - \sum_{s' \in \mc{S}} T_o(s' \mid s, \mu_{\mc{O}_\phi}(s_\phi)) V^*(s') \\
        & & + \sum_{s' \in \mc{S}} T_o(s' \mid s, \mu_{\mc{O}_\phi}(s_\phi)) V^*(s') - \sum_{s' \in \mc{S}} T_o(s' \mid s, \mu_{\mc{O}_\phi}(s_\phi)) Q_k^*(s', \mu_{\mc{O}_\phi}(s_\phi')) \nonumber \\
        = & & V^*(s) - Q_{s_\phi}^*(s, \mu_{\mc{O}_\phi}(s_\phi)) \\
        & & + \sum_{s' \in \mc{S}} T_o(s' \mid s, \mu_{\mc{O}_\phi}(s_\phi)) \left[ V^*(s') - Q_k^*(s', \mu_{\mc{O}_\phi}(s_\phi') \right] \\
        = & & Q_{s_\phi}^*(s, o_{s_\phi}^*) - Q_{s_\phi}^*(s, \mu_{\mc{O}_\phi}(s_\phi)) \\
        & & + \sum_{s' \in \mc{S}} T_o(s' \mid s, \mu_{\mc{O}_\phi}(s_\phi)) \left[ V^*(s') - Q_k^*(s', \mu_{\mc{O}_\phi}(s_\phi') \right] \\
        \leq & & \eps_Q + \sum_{s' \in \mc{S}} T_o(s' \mid s, \mu_{\mc{O}_\phi}(s_\phi)) \left[ V^*(s') - Q_k^*(s', \mu_{\mc{O}_\phi}(s_\phi') \right],
    \end{eqnarray}
    by definition of $\mu_{\mc{O}_\phi}$. Continuing, we have that:
    \begin{eqnarray}
        = & & \eps_Q + \sum_{s' \in \mc{S}} \sum_{n=1}^{\infty} p(s', n \mid s, \mu_{\mc{O}_\phi}(s_\phi)) \gamma^n \left[ V^*(s') - Q_k^*(s', \mu_{\mc{O}_\phi}(s_\phi') \right] \\
        \leq & & \eps_Q + \sum_{s' \in \mc{S}} \sum_{n=1}^{\infty} p(s', n \mid s, \mu_{\mc{O}_\phi}(s_\phi)) \gamma^n \sum_{i=0}^{k} \eps_Q \gamma^i,
    \end{eqnarray}
    
    by the inductive hypothesis. Then:
    \begin{eqnarray}
        = & & \eps_Q + \gamma \sum_{s' \in \mc{S}} \sum_{n=0}^{\infty} p(s', n + 1 \mid s, \mu_{\mc{O}_\phi}(s_\phi)) \gamma^n \sum_{i=0}^{k} \eps_Q \gamma^i \\
        = & & \eps_Q + \gamma \sum_{i=0}^{k} \eps_Q \gamma^i \sum_{s' \in \mc{S}} \sum_{n=0}^{\infty} p(s', n + 1 \mid s, \mu_{\mc{O}_\phi}(s_\phi)) \gamma^n \\
        \leq & & \eps_Q + \gamma \sum_{i=0}^{k} \eps_Q \gamma^i \cdot 1 \\
        = & & \sum_{i=0}^{k+1} \eps_Q \gamma^i,
    \end{eqnarray}
    since $p(s', n + 1 \mid s, \mu_{\mc{O}_\phi}(s_\phi))$ is a probability distribution and $\gamma$ is less than $1$.\\
    
    All together, we've shown that $V^*(s) - Q_{k+1}^*(s, \mu_{\mc{O}_\phi}(s_\phi)) \leq \sum_{i=0}^{k+1} \eps_Q \gamma^i$ for all $s \in \mc{S}$, which implies that
    \begin{equation}
        \max_{s \in \mc{S}} V^*(s) - Q_{k+1}^*(s, \mu_{\mc{O}_\phi}(s_\phi)) \leq \sum_{i=0}^{k+1} \eps_Q \gamma^i,
    \end{equation}
    as desired.\\
    
    It follows by induction that
    \begin{equation}
        \forall_{t \in \mathbb{N}} : \max_{s \in \mc{S}} V^*(s) - Q_t^*(s, \mu_{\mc{O}_\phi}(s_\phi)) \leq \sum_{i=0}^{t} \eps_Q \gamma^i.
    \end{equation}
    
    Therefore,
    \begin{align}
        L(\phi, \mc{O}_{\phi, Q_\eps^*}) &\leq \max_{s \in \mc{S}} V^*(s) - V^{\mu^\Downarrow_{\mc{O}_\phi}}(s) \\
        &= \max_{s \in \mc{S}} \lim_{t \to \infty} V^*(s) - Q_t^*(s, \mu_{\mc{O}_\phi}(s_\phi)) \\
        &\leq \lim_{t \to \infty} \sum_{i=0}^{t} \eps_Q \gamma^i \\
        &= \frac{\eps_Q}{1 - \gamma},
    \end{align}
    which completes the proof.
\end{proof}

%% file: proofs/c9/c9_phio_model_loss.tex
\begin{proof}
    ($L(\phi, \mc{O}_{\phi, M_\eps}) \leq \frac{\eps_{R} + |\mc{S}| \eps_{T} \textsc{VMax}}{1-\gamma}$) \\
    
    We show that this class is a subclass of the $\mc{O}_{\phi, Q^*_\eps}$ class. Therefore, it stands to show that, given our class definition, there exists an option in every abstract state that is near-optimal in $Q$-value. \\
    
    Fix $s \in \mc{S}$. Let $s_\phi = \phi(s)$. By the $M_\eps$ predicate, there exists an option $o \in \mc{O}_\phi$ such that 
    
    \begin{equation}
        ||T_{s,o^*_{s_\phi}}^{s'} - T_{s,o}^{s'}||_\infty \leq \eps_T \;\text{and}\;  ||R_{s,o^*_{s_\phi}} - R_{s,o}||_\infty \leq \eps_R.
    \end{equation}
    
    Now, we consider the difference in optimal $Q$-values between $o^*_{s_\phi}$ and $o$. We first have that:
    \begin{equation}
    \begin{aligned}
        Q_{s_\phi}^*(s, o) &= R(s,\pi_{o}(s)) + \gamma \sum_{s' \in \mc{S}} T(s' \mid s,\pi_{o}(s)) \left(\indic(s' \in s_\phi)Q_{s_\phi}^*(s', o) + \indic(s' \not \in s_\phi)V^*(s')\right) \\
        &= R_o(s, o) + \sum_{s' \in \mc{S}} T_o(s' \mid s, o) V^*(s'),
    \end{aligned}
    \end{equation}
    with $R_o$ and $T_o$ denoting the standard multi-time model of~\citet{sutton1999between}.
    
    By symmetry,
    \begin{equation}
         Q_{s_\phi}^*(s, o^*_{s_\phi}) = R_o(s, o^*_{s_\phi}) + \sum_{s' \in \mc{S}} T_o(s' \mid s, o^*_{s_\phi}) V^*(s').
    \end{equation}
    
    Therefore,
    \begin{eqnarray}
        &&| Q_{s_\phi}^*(s, o^*_{s_\phi}) - Q_{s_\phi}^*(s, o) | \\
        &\leq& | R_o(s, o^*_{s_\phi}) - R_o(s, o) | + | \sum_{s' \in \mc{S}} \left( T_o(s'\mid s,o^*_{s_\phi}) - T_o(s'\mid s,o) \right) V^*(s') | \\
        &\leq& | R_o(s, o^*_{s_\phi}) - R_o(s, o) | +  \sum_{s' \in \mc{S}} | T_o(s'\mid s,o^*_{s_\phi}) - T_o(s'\mid s,o) | |V^*(s')|  \\
        &\leq& \eps_R + |\mc{S}| \eps_T \textsc{VMax},
    \end{eqnarray}
    by the model similarity assumption. We have now shown that any option with near-optimal models has a near-optimal $Q$-value with $\eps_Q = \eps_R + |\mc{S}| \eps_T \textsc{VMax}$. Therefore, by the previous result, 
    \begin{equation}
        L(\phi, O_{\phi, M_\eps}) \leq \frac{\eps_R + |\mc{S}| \eps_T \textsc{VMax}}{1-\gamma}.
    \end{equation}
    
\end{proof}

%% file: proofs/c9/c9_phio_k_step_loss.tex
\begin{proof}
($L(\phi, \mc{O}_{\phi, \tau}) \leq  \frac{\tau \gamma |\mc{S}|}{(1-\gamma)^2}$) \\

We first state rigorously our definition of a goal-based MDP. \\

\begin{definition}
  A goal-based MDP is an MDP with some number of goal states, denoted $\mc{S}_G \subseteq \mc{S}$. The reward function is such that $R(s,a) = 1$ if $s \in \mc{S}_G$, $R(s,a) = 0$ otherwise, and the episode terminates after receiving a reward in a goal state. Furthermore, we assume that each goal state exists in its own abstract state: $s \neq s_g \Rightarrow \phi(s_g) \neq \phi(s)$, where $s_g \in \mc{S}_G, s \in \mc{S}$.
\end{definition}

We show that this class is a subclass of the $\mc{O}_{\phi, Q^*_\eps}$ class in goal-based MDPs. In particular, it stands to show that given our class definition, there exists an option in every abstract state that is near-optimal in Q-value. \\

First, note that in the abstract states containing a goal state, any option is optimal since $R(s,a) = 1$ regardless of action. Therefore, we restrict our attention to an arbitrary $s \in \mc{S} \setminus \mc{S}_G$, fixing $s_\phi = \phi(s)$. Let $o$ be an option available in $s_\phi$ such that $\max_{s\in s_\phi, s' \in \mc{S}}   |p(s', k \mid s, o^*_{s_\phi}) - p(s', k \mid s, o)| \leq \tau$, by the option class definition. Then
\begin{align}
    &Q^*_{s_\phi}(s, o^*_{s_\phi}) - Q^*_{s_\phi}(s, o) \\
    &= R_o(s, o^*_{s_\phi}) + \sum_{s' \in \mc{S}} T_o(s' | s, o^*_{s_\phi}) V^*(s') - R_o(s, o) - \sum_{s' \in \mc{S}} T_o(s' | s, o) V^*(s') \\
    &= \sum_{s' \in \mc{S}} \left[ T_o(s' | s, o^*_{s_\phi}) - T_o(s' | s, o)\right] V^*(s'),
\end{align}
where we drop the $R_o$ terms since $s \not \in \mc{S}_G$, each goal state has its own abstract state, and $R(s,a)=0$ for $s \not \in \mc{S}_G$. Continuing, we have that
\begin{align}
    Q^*_{s_\phi}(s, o^*_{s_\phi}) - Q^*_{s_\phi}(s, o) &= \sum_{s' \in \mc{S}} \left[ \sum_{k=1}^{\infty} |p(s', k \mid s, o^*_{s_\phi}) - p(s', k \mid s, o)| \gamma^k \right] V^*(s),
\end{align}
writing out the multi-time model. This implies that
\begin{align}
    Q^*_{s_\phi}(s, o^*_{s_\phi}) - Q^*_{s_\phi}(s, o) \leq \sum_{s' \in \mc{S}} \frac{\tau \gamma}{1 - \gamma} V^*(s).
\end{align}

Now, note that $V^*(s') = \sum_{s_g \in \mc{S}_G} \sum_{t=0}^{\infty} p(s_g, t \mid s', \pi^*) \gamma^t$ in a goal-based MDP, where $p(s_g, t \mid s', \pi^*)$ is the probability of being in state $s_g$ after $t$ timesteps, starting from $s'$ and following $\pi^*$. Indeed, this gives that $V^*(s') \leq 1$ since $p(s_g, t \mid s', \pi^*)$ is a probability distribution and $\gamma$ is less than one. Therefore, 
\begin{equation}
    Q^*_{s_\phi}(s, o^*_{s_\phi}) - Q^*_{s_\phi}(s, o) \leq \frac{\tau \gamma |\mc{S}|}{1 - \gamma}.
\end{equation}

We have shown that there exists an option, $o$, in any abstract state that is near-optimal in Q-value, with $\eps_Q = \frac{\tau \gamma |\mc{S}|}{1 - \gamma}$. Therefore, by the $\mc{O}_{\phi, Q^*_\eps}$ bound,
\begin{equation}
    L(\phi, \mc{O}_{\phi, \tau}) \leq  \frac{\tau \gamma |\mc{S}|}{(1-\gamma)^2},
\end{equation}
as desired.
\end{proof}

%% file: proofs/c9/c9_phio_homomorphism_loss.tex
\begin{proof}
$\left(L(\phi, \mc{O}_{\phi, H}) \leq \frac{2}{1-\gamma} \left( \eps_r + \frac{\gamma\textsc{RMax}}{1 - \gamma}  \frac{\eps_p}{2}\right)\right)$ \\

We prove this result by illustrating the connection between our formalisms and the work of~\citet{ravindran2004approximate}. To do so, we first restate their definition of an approximate homomorphism.


\begin{definition}
    An \textbf{approximate MDP homomorphism} \cite{ravindran2004approximate} $h : M \mapsto M^{\prime}$ from an MDP $M=(\mc{S}, \mc{A}, \Psi, T, R, \gamma )$ to an MDP
    \(M^{\prime}=\left( \mc{S}^{\prime}, \mc{A}^{\prime}, \Psi^{\prime}, T^{\prime}, R^{\prime}, \gamma \right)\) is a surjection from \(\Psi\) to \(\Psi^{\prime},\) defined by a tuple of surjections \(\left\langle f,\left\{g_{s} : s \in S\right\}\right\rangle,\)
    with \( h((s, a))=\left(f(s), g_{s}(a)\right),\) where \(f : S \rightarrow S^{\prime}\) and \(g_{s} : \mc{A}_{s} \rightarrow \mc{A}_{f(s)}^{\prime}\) for \(s \in S,\) such that for all \(s, s^{\prime}\) in
    \(\mc{S}\) and \(a \in \mc{A}_{s}:\)
    
    \begin{equation}
    T^{\prime}\left(f\left(s^{\prime}\right) \mid f(s), g_{s}(a) \right)=\sum_{(q, b) \in[(s, a)]_{h}} w_{q b} \sum_{s^{\prime \prime} \in\left[s^{\prime}\right]_{f}} T\left(s^{\prime \prime} \mid q, b \right)
    \end{equation}
    
    \begin{equation}
    R^{\prime}\left(f(s), g_{s}(a)\right)=\sum_{(q, b) \in[(s, a)]_{h}} w_{q b}\ R(q, b),
    \end{equation}

where $[(s,a)]_h$ denotes the preimage of $h((s,a))$, $[s']_f$ denotes the preimage of $f(s')$, and $\sum_{(q, b) \in[(s, a)]_{h}} w_{q b}=1$. Furthermore, $\Psi$ and $\Psi'$ denote the sets of admissible state-action pairs in the ground and abstract MDP respectively. Based on $\Psi$ and $\Psi'$, $A_s$ denotes the set of actions available in state $s$ of the ground MDP, and $A'_{f(s)}$ denotes the set of abstract actions available in state $f(s)$ of the abstract MDP.
\end{definition}

We now illustrate how our definitions of $\phi, R_\phi, T_\phi$ with respect to a given $\pi_{\mc{O}_\phi}$ induce an approximate homomorphism. First, note that our $\phi$ precisely corresponds to their definition of $f$, a state abstraction. Then, fix $s_\phi \in \mc{S}_\phi$, and let $A'_{s_\phi} = \{\pi_{\mc{O}_\phi}(s_\phi)\}$ with $g_s(a) = \pi_{\mc{O}_\phi}(s_\phi) \forall_{s \in s_\phi} \ \forall_{a \in A}$. \\

We now consider our definitions of $T_\phi$ and $R_\phi$:
\begin{equation}
    T_\phi (s_\phi' \mid s_\phi, o) = \sum_{s \in s_\phi} w(s) \sum_{s' \in s_\phi'} T(s' \mid s, \pi_o(s)) \hspace{1cm} R_\phi(s_\phi, o) = \sum_{s \in s_\phi} w(s) R(s, \pi_o(s)),
\end{equation}

We note that these are precisely an instance of $P'$ and $R'$ as defined above, with $w_{q b} = 0$ whenever $b \neq \pi_o(q)$
. We write $w(s)$ to denote this choice of weighting function, which depends only on the action prescribed by $\pi_o$. We select this choice of weighting function (as opposed to a weighting dependent on all available actions) in order to faithfully represent the 1-step behavior of executing an option in the abstract MDP. \\

By these connections, a deterministic policy $\pi_{\mc{O}_\phi}$ over $\phi$-relative options coupled with our choice of weighting function defines an approximate homomorphism. We further adapt their definitions of $K_p$ and $K_r$ to our notational setting, which describe the maximum discrepancy in models between the ground and abstract MDPs.

\begin{equation}
    K_p = \max_{s \in \mc{S}, a \in \mc{A}} \sum_{s_\phi \in \mc{S}_\phi} \Big| \sum_{s' \in s_\phi} T(s' \mid s, a) - T_\phi(s_\phi \mid \phi(s), \pi_{\mc{O}_\phi}(\phi(s))) \Big|,
\end{equation}
\begin{equation}
    K_r = \max_{s \in \mc{S}, a \in \mc{A}} |R(s, a) - R_\phi(\phi(s), \pi_{\mc{O}_\phi}(\phi(s))) |.
\end{equation}

The main theorem of~\citet{ravindran2004approximate} guarantees that the value loss of the optimal policy in the abstract MDP $\mc{M}'$ is upper-bounded by
\[
\frac{2}{1-\gamma}\left(K_{r}+\frac{\gamma}{1-\gamma} \delta_{r^{\prime}} \frac{K_{p}}{2}\right),
\]
where $\delta_{r^{\prime}}$ is upper-bounded by $\textsc{RMax}$. Let $\mu_{\mc{O}_\phi} \in \Pi_{\mc{O}_\phi}$ denote the optimal policy in the abstract MDP. By our option class definition, all abstract policies $\pi_{\mc{O}_\phi} \in \Pi_{\mc{O}_\phi}$ induce homomorphisms with bounded $K_p, K_r$. Thus, $\mu_{\mc{O}_\phi}$ has bounded $K_p, K_r$. Then:
\begin{align}
    L(\phi, \mc{O}_{\phi, H}) &= \min_{\pi_{\mc{O}_\phi} \in \Pi_{\mc{O}_\phi}} \left|\left| V^* - V^{\pi^{\Downarrow}_{\mc{O}_\phi}}\right|\right|_\infty, \\
    &\leq \left|\left| V^* - V^{\mu^{\Downarrow}_{\mc{O}_\phi}}  \right|\right|_\infty, \\
    &\leq \frac{2}{1-\gamma}\left(K_{r}+\frac{\gamma}{1-\gamma} \delta_{r^{\prime}} \frac{K_{p}}{2}\right) \\
    &\leq \frac{2}{1-\gamma} \left( \eps_r + \frac{\gamma\textsc{RMax}}{1 - \gamma}  \frac{\eps_p}{2}\right),
\end{align}
as desired.
\end{proof}

%% file: proofs/c9/c9_necessary_condition.tex
\begin{dproof}[Theorem \ref{thm:c9_vpsa_nec_cond}]
Let $\mu_{\mc{O}_\phi} = \argmin_{\pi_{\mc{O}_\phi} \in \Pi_{\mc{O}_\phi}} \left|\left| V^* - V^{\pi^{\Downarrow}_{\mc{O}_\phi}}\right|\right|_\infty.$

Suppose, for a contradiction, that there exists an abstract state $s_\phi$ for which there is no $\eta$-optimal option in $\mc{O}_\phi$. Then it must be the case that
\begin{equation}
    Q^*_{s_\phi}(s, o^*) - Q^*_{s_\phi}(s, \mu_{\mc{O}_\phi}(s)) > \eta
\end{equation}
for some $s \in s_\phi$. \\

By, $Q^*_{s_\phi}(s, o^*) = V^*(s)$, this implies that
\begin{equation}
    V^*(s) - Q^*_{s_\phi}(s, \mu_{\mc{O}_\phi}(s)) > \eta.
\end{equation}

Then, note that $Q^*_{s_\phi}(s, \mu_{\mc{O}_\phi}(s)) \geq V^{\mu_{\mc{O}_\phi}^\Downarrow}(s)$ because $Q^*_{s_\phi}$ describes the expected return of executing option $\mu_{\mc{O}_\phi}(s)$, then switching to optimal behavior, whereas $V^{\mu_{\mc{O}_\phi}^\Downarrow}$ describes the expected return of executing $\mu_{\mc{O}_\phi}(s)$ then continuing to execute options according to $\mu_{\mc{O}_\phi}$.

Noticing that $V^*(s) \geq Q^*_{s_\phi}(s, \mu_{\mc{O}_\phi}(s)) \geq V^{\mu_{\mc{O}_\phi}^\Downarrow}(s)$, we have that
\begin{equation}
    V^*(s) - V^{\mu_{\mc{O}_\phi}^\Downarrow}(s) > \eta.
\end{equation}

This implies that 
\begin{align}
    L(\phiop) &= \min_{\pi_{\mc{O}_\phi} \in \Pi_{\mc{O}_\phi}} \left|\left| V^* - V^{\pi^{\Downarrow}_{\mc{O}_\phi}}\right|\right|_\infty \\
    &= V^*(s) - V^{\mu_{\mc{O}_\phi}^\Downarrow}(s) \\
    &> \eta,
\end{align}
which contradicts the premise. Therefore, it must be true that
\begin{equation}
    \forall_{s_\phi \in \mc{S}_\phi} \forall_{s \in s_\phi} \exists_{o \in \mc{O}_\phi} : Q^*_{s_\phi}(s, o^*) - Q^*_{s_\phi}(s, o) \leq \eta. \qedhere
\end{equation}

\end{dproof}

%% file: figures/tables/sa_aa_notation_table.tex
\begin{table*}[!h]
\vspace{8mm}
\def\arraystretch{1.23}
    \centering
    \begin{tabular}{lll}
        \toprule
        $\phi$&A state abstraction function. \\
        $\mc{O}_\phi$&A set of $\phi$-relative options.\\
        $L(\phiop)$& The value loss of the $\phiop$ pair. \\
        \midrule
        $\pi_{\mc{O}_\phi}$&A policy that maps each abstract state to an option. \\
        $\pi_{\mc{O}_\phi}^\Downarrow$&A policy over $\mc{S}$ and $\mc{A}$, induced by $\pi_{\mc{O}_\phi}$. \\
        \midrule
        $H_n$& A hierarchy of depth $n$, denoting the pair of lists $(\phi^{(n)}, \mc{O}_\phi^{(n)})$. \\
        $\phi^{(n)}$&A list of $n$ state abstractions, where $\phi_i : \mc{S}_{\phi,i-1} \ra \mc{S}_{\phi,i}$. \\
        $\phi_i$&The $i$-th state abstraction in a list $\phi^{(n)}$. \\
        $\phi^i$&The result of applying the first $i$ state abstractions to $s$, $\phi_i(\ldots \phi_1(s) \ldots)$.  \\
        $\mc{S}_{\phi,i}$&The $i$-th abstract state space, with $\mc{S}_{\phi, 0}$ the ground state space. \\
        $s_i$& A state belonging to $\mc{S}_{\phi,i}$ \\
        $V_i^\pi$& Value of level $i$ under policy $\pi$, defined according to $R_i$ and $T_i$. \\
        $\mc{O}_{\phi,i}$& Options at level $i$, with each component defined over states in $\mc{S}_{\phi, i-1}$. \\
        $R_i$& A reward function over level $i$ states and options. \\
        $T_i$& A transition function over level $i$ states and options. \\
         $\pi_i$& The policy over level $i$ of the hierarchy such that $\pi_i : \mc{S}_i \ra \mc{O}_{\phi,i}$.  \\
        $\pi_{i}^\downarrow$&A policy over $\mc{S}_{\phi, i-1}$ and $\mc{O}_{\phi, i-1}$, induced by $\pi_{i}$. \\
        $\pi_{i}^\Downarrow$&A policy over $\mc{S}$ and $\mc{A}$, induced by $\pi_{i}$. \\
        \bottomrule
    \end{tabular}
    \vspace{1mm}
    \caption{Hierarchical abstraction notation.}
    \label{tab:hierarch_abstr_notation}
\end{table*}

%% file: proofs/c9/c9_hierarch_loss.tex
\begin{dproof}[Theorem \ref{thm:c9_hierarch_loss}]
We present the proof of the bound for a two level hierarchy, but the same strategy generalizes to $n$ levels via induction. \\

\noindent Let $\ell$ be the known upper bound for $L(\phi, \mc{O})$, obtained by any of the $(\phiop)$ pairs from \autoref{thm:c9_vpsa_main_result}. 
\begin{align}
    &\text{By definition of $L(\phi, \mc{O})$:} \nonumber\\
       &\hspace{16mm} \min_{\pi_1 \in \Pi_1} ||V_0^* - V_0^{\pi^\downarrow_1}||_\infty \leq \ell. \\
&\text{By \autoref{asmpt:hierarchy_value}:} \nonumber\\
    &\hspace{16mm} \forall_{\pi_1 \in \Pi_1} : ||V_0^{\pi^\downarrow_1} - V_1^{\pi_1}||_\infty \leq \kappa. \\
&\text{Letting $\pi_1^\diamond = \argmin_{\pi_1 \in \Pi_1} ||V_0^* - V_0^{\pi^\downarrow_1}||_\infty$}, \text{ by \autoref{asmpt:vi_eps_options}:} \nonumber\\
    &\hspace{16mm} \min_{\pi_2^{\downarrow} \in \Pi^{\downarrow}_2} ||V_1^{\pi_1^\diamond} - V_1^{\pi^\downarrow_2}||_\infty \leq \ell. \\
&\text{By \autoref{asmpt:hierarchy_value}:} \nonumber \\
    &\hspace{16mm} \forall_{\pi^\downarrow_2 \in \Pi^\downarrow_2} : ||V_1^{\pi^\downarrow_2} - V_0^{\pi^{\Downarrow}_2}||_\infty \leq \kappa. \\
&\text{Therefore, by the triangle inequality:} \nonumber \\
    &\hspace{16mm} \min_{\pi_2 \in \Pi_2}||V_0^* - V_0^{\pi^{\Downarrow}_2}||_\infty \leq 2(\kappa + \ell). \qedhere
\end{align}
\end{dproof}

%% file: chapters/c10_conclusion.tex
The thesis of this dissertation is that insights from computational complexity, decision-theoretic planning, and information theory can shape principled abstraction discovery algorithms that empower RL agents. I defended this thesis on two fronts: 1) state abstraction (\autoref{part:state_abstraction}), and 2) action abstraction (\autoref{part:action_abstraction}), with a final note on the tightly woven connections between these two (\autoref{part:state_action_abstraction}). In this final chapter, I offer broader outlooks on abstraction and its role in both AI and RL.

\section{Why Abstraction?}

In \autoref{chap:introduction}, I suggested that the process of abstraction is critical to the success of any adaptive sequential decision making agent. Naturally, such a claim is speculative, as the space of all possible agents is vast. It is not yet clear which properties precisely separate effective agents from the ineffective ones. I now revisit this point with the results established in the dissertation in hand. I set out to present the strongest argument about the potential for abstraction to contribute to solutions to the RL problem, though naturally much of the present discussion remains speculative.

I take there to be three senses in which understanding the role of abstraction in RL is useful:
\begin{enumerate}
    \item Abstraction is sufficient (and perhaps, necessary) for grounding simulated states and actions to observation and behavior.
    
    \item If the space of relevant worlds can be characterized by simple underlying laws, then abstraction may be viewed (charitably, perhaps) as the process of recovering this simplicity from an agent-centric experience of the world.
    
    \item Even if the best RL agents do not explicitly abstract, there is likely to be \textit{implicit} abstraction taking place in their computation. Furthermore, understanding the implicit mechanisms that support effective agency is still of deep scientific importance.
\end{enumerate}

I now expand on each point in more detail.

First, abstraction is \textit{sufficient} for grounding simulated states and actions to the observation and behavior space defined by the world. As highlighted throughout the dissertation and in prior work, these internal representations can be immensely useful for unleashing the power \textit{and} reliability of computation onto sequential decision making problems. \citet{konidaris2019necessity} presents a compelling case for the necessity of abstraction from this perspective, too, arguing that ``a necessary but understudied requirement for general intelligence is the ability to form task-specific abstract representations" (\citeyear{konidaris2019necessity}, p. 1). Like the views presented in this dissertation, Konidaris goes on to suggest that RL is an appropriate paradigm to formalize and investigate abstraction in the context of agency. Abstraction is at least one vehicle for carrying out simulated decision making in fictional, but grounded, state-action space. Using this capacity, agents are empowered to consider past counterfactuals or inform present decision from alternative courses of future behavior. Such practices appear critical to effective agency. 

Second, abstraction can also be viewed as the recovery of simple underlying world laws from agent-centric experience. To expand on this point, let us make two assumptions. First, the space of worlds of interest are those with exceedingly simple descriptions that give rise to complex phenomena. Second, that agents of interest are resource-bounded, as has been articulated by many \cite{simon1957models,simon1972theories,gigerenzer1996reasoning,Russell1995,lewis2014computational,gershman2015computational}. Which kinds of resource-bounded agents will be most successful in such a space of worlds? An agent using a simple model can support more valuable computation per time step; it is likely that those models that recover more of the underlying laws will be more capable of making informed predictions about their surroundings. For more on this point, I expand on this argument in detail elsewhere \cite{abel2019philrl}.

Third, let us suppose that the most reliably successful RL agents are those that do not make explicit use of abstraction. I suggest that some form of abstraction is likely to be taking place \textit{implicitly} within such agents.\footnote{This conjecture is supported in part by point 2: simple underlying worlds yield simple explanatory models.} That is, the more an agent can specialize to a particular distribution of worlds, the more effective they can be at learning and solving tasks in those worlds. I speculate that it is likely that the process of abstraction is necessarily tied to this process of specialization. This reasoning is supported in part by the bias-complexity trade off discussed throughout the dissertation; as an abstraction becomes more aggressive, the space of representable entities becomes less rich, thereby making many problems critical to learning and decision making easier, but potentially compromising the quality of the best entities representable. In this sense, I conjecture that the most effective learning agents will carry out abstraction in some form, whether implicitly or explicitly.

One might worry that without a clear protocol for testing for the implicit use of abstraction, this claim is unfalsifiable. However, one straightforward mechanism for testing whether an agent abstracts implicitly is to again turn to information theory. As has been argued elsewhere (see work by \citet{legg2007universal} and \citet{dowe2011compression} and references therein) \textit{compression} may be tied to fundamental aspects of intelligent systems. Thus, let us suppose that some form of compression-based test will suffice for determining whether abstraction is implicitly used by an agent. Then, I claim, if it is feasible to determine which \textit{kinds} of compression are essential to effective agency, and which are not. In this sense, abstraction remains critical.

There are many other reasons to take abstraction-based approaches to RL as promising beyond those discussed thus far. With abstractions, there is a more direct path toward shared knowledge among a community of agents, simple communication between the agents, or mechanisms that test for the reliability, failure modes, safety, and robustness of an agent. As hinted at in \autoref{chap:introduction}, a hiker that first learns walking into a tree is painful is sure to share this finding with their community. As such discoveries become bigger and more significant, abstraction is essential to effectively convey knowledge across broad populations of agents. In addition to communication, the role of abstraction in relation to these other properties is also deserving of attention. Indeed, there are many fundamental questions left to address.


\section{The Road Ahead}

There are many remaining steps to realize the potential of abstraction in RL. The primary contributions of this thesis target finite MDPs, the class of state aggregation functions $\phi : \mc{S} \ra \mc{S}_\phi$, and action abstractions $\omega : \mc{A} \mapsto \mc{O}$ where the set $\mc{O}$ is assumed to operate on these finite state spaces. The analysis and algorithms describe methods for finding and using abstractions that are guaranteed to retain desirable properties, with empirical support illustrating the potential for these abstractions to accelerate learning and planning. However, these results are restricted in several ways.

First, the kinds of functions studied are themselves relatively weak. The act of state aggregation or discretization can only go so far to simplify state spaces. This formalism fundamentally lacks the capacity to express the kinds of powerful relations or descriptions that appear in common language. The most natural expansion of the results presented in \autoref{part:state_abstraction} will go beyond aggregation functions to richer function families that can define objects, relations, predicates, and their kin. Establishing the same degree of understanding about these more powerful function families is essential to a comprehensive understanding of how agents come to act in complex environments.
The same can be said of options.

Second, the primary focus of this dissertation is on finite MDPs. As discussed briefly in \autoref{chap:background}, there are many schemes for defining the space of relevant worlds, with finite MDPs being one suitable choice. Much of the analysis and the vast majority of the experimental study in this dissertation takes place in the context of simple grid worlds and their kin, with only a few exceptions. Thus, a second critical direction is to expand the primary analysis beyond finite state-action spaces. Some preliminary directions toward this goal were summarized in \autoref{sec:c5_rlit_extensions}, but there is more to be done. 

Third, this dissertation focuses on the learning problem facing a single agent. In reality, of course, many agents of relevance occupy a community. These agents learn, act, explore, and plan based on the beliefs and behaviors of other agents. How might the abstraction desiderata change if two agents or more are learning cooperatively in the same world? It is likely, for instance, that agents ought to specialize their abstractions while retaining enough overlap to allow for communication. Understanding abstraction when multiple agents are present is a key direction for further work.



\section{Concluding Remarks}
In conclusion, I take understanding abstraction in RL to be of fundamental importance to a holistic science of AI. The formalisms and analysis of this dissertation build on a long line of research to provide new clarity on how to discover and use good abstractions in RL. There is still much to be done, but the road ahead is an exciting one.

\vspace{4mm}
\begin{center}
\begin{minipage}{0.65\textwidth}
\resizebox{0.9\textwidth}{6pt}{%
  \begin{tikzpicture}
        \pgfornament[symmetry=h]{85} 
    \end{tikzpicture}
}
\end{minipage}
\end{center}